\documentclass[10pt,twocolumn,letterpaper]{article}

\usepackage{cvpr}              

\usepackage[table]{xcolor}
\usepackage{algorithmic}            
\usepackage{algorithm}
\usepackage{tikz}
\usepackage{amsmath}
\usepackage{amssymb}
\usepackage{amsthm}
\usetikzlibrary{matrix}
\usetikzlibrary{arrows.meta}
\usepackage{graphicx}
\usepackage{caption}
\usetikzlibrary{shapes.arrows}
\usepackage{svg}
\usetikzlibrary{shadows, backgrounds}
\usepackage{pgfplots}
\pgfplotsset{compat=newest}
\usepgfplotslibrary{groupplots}
\usepgfplotslibrary{colormaps}
\usetikzlibrary{pgfplots.groupplots}
\usepackage{booktabs}
\usepackage[percent]{overpic}
\usepackage{subcaption}
\usepackage[most]{tcolorbox}
\usetikzlibrary{shadows.blur}
\usepackage{multirow}
\usepackage{makecell}
\usepackage{rotating}
\usepackage{siunitx}
\newcolumntype{T}[1]{S[table-format=#1]}

\definecolor{cvprblue}{rgb}{0.21,0.49,0.74}
\usepackage[pagebackref,breaklinks,colorlinks,allcolors=cvprblue]{hyperref}

\def\paperID{31639} 
\def\confName{CVPR}
\def\confYear{2026}

\title{SegQuant: A Semantics-Aware and Generalizable Quantization Framework for Diffusion Models}

\author{
Jiaji Zhang\textsuperscript{1} \quad
Ruichao Sun\textsuperscript{1} \quad
Hailiang Zhao\textsuperscript{1}\thanks{Corresponding authors} \quad
Jiaju Wu\textsuperscript{2} \quad
Peng Chen\textsuperscript{1} \\
Hao Li\textsuperscript{3} \quad
Yuying Liu\textsuperscript{3} \quad
Kingsum Chow\textsuperscript{1} \quad
Gang Xiong\textsuperscript{3} \quad
Shuiguang Deng\textsuperscript{1}\footnotemark[1] \\[2mm]
\textsuperscript{1}Zhejiang University \quad
\textsuperscript{2}Nanyang Technological University \quad
\textsuperscript{3}Kuaishou Technology \\[1mm]
{\tt\small \{zhangjiaji, csunray, hliangzhao, pgchen, kingsum.chow, dengsg\}@zju.edu.cn} \\
{\tt\small wujiajucn@gmail.com \quad \{lihao36, liuyuying, xionggang\}@kuaishou.com}
}

\begin{document}
\maketitle

\begin{abstract}
Diffusion models have demonstrated exceptional generative capabilities but are computationally intensive, posing significant challenges for deployment in resource-constrained or latency-sensitive environments.
Quantization offers an effective means to reduce model size and computational cost, with post-training quantization (PTQ) being particularly appealing due to its compatibility with pre-trained models without requiring retraining or training data.
However, existing PTQ methods for diffusion models often rely on manual, architecture-specific heuristics that limit their generalizability and hinder integration with industrial deployment pipelines.
To address these limitations, we propose SegQuant, a deployment-aware quantization framework that adaptively combines complementary techniques to enhance cross-model versatility.
SegQuant consists of a segment-aware, graph-based quantization strategy (SegLinear) that captures structural semantics and spatial heterogeneity, along with a dual-scale quantization scheme (DualScale) that preserves polarity-asymmetric activations using a hardware-native dual-path computation, minimizing performance penalties associated with custom implementations, which is crucial for maintaining visual fidelity in generated outputs.
SegQuant is broadly applicable beyond Transformer-based diffusion models, achieving strong performance while ensuring seamless compatibility with mainstream deployment tools.
\end{abstract}    

\section{Introduction}
\label{sec:intro}

Diffusion models~\cite{ddpm2020} have emerged as a dominant class of generative models, demonstrating impressive performance across various applications including image synthesis, inpainting, video generation, etc.
Despite their impressive performance, deploying diffusion models at scale remains challenging, particularly in high-concurrency settings where service providers must balance computational efficiency with output quality.

To address these challenges, quantization~\cite{quant2021} has emerged as a practical solution for reducing the computational burden of diffusion inference.
One promising approach to improving deployment efficiency is post-training quantization~\cite{ptq2018CVPR} (PTQ), which reduces model precision without requiring retraining or extensive fine-tuning.
Quantization significantly improves inference speed and memory usage.
However, reducing numerical precision often leads to performance degradation, especially in complex models like diffusion models that rely on iterative refinement over multiple denoising steps.
Therefore, developing effective quantization strategies that preserve generation fidelity while maximizing efficiency has become a critical research direction.

Recent efforts in PTQ for diffusion models~\cite{li2023qdiffusionquantizingdiffusionmodels, wu2024ptq4dit} have achieved strong results, but often rely on strategies incompatible with modern, graph-based AI compilers~\cite{torch2019, modelopt}. 
These deployment tools depend on \textit{static graph analysis} for optimization. 
However, methods like PTQ4DiT~\cite{wu2024ptq4dit} use heuristics based on \textit{runtime-dynamic} data (e.g., timestep-varying activations), while others like Q-Diffusion~\cite{li2023qdiffusionquantizingdiffusionmodels} require \textit{manual}, architecture-specific rules (e.g., for UNet skip-connections). 
This creates a \textit{Compiler Gap}: techniques that are effective in isolation but resist automated, large-scale deployment.

\begin{figure*}[htb!]
\centering
\includegraphics[width=0.7\textwidth]{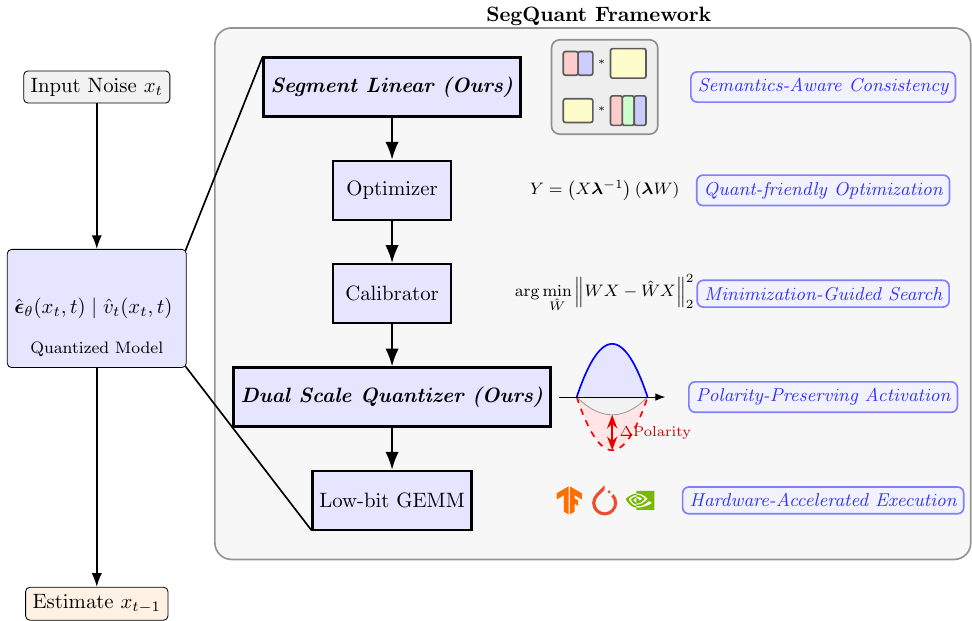}
\caption{SegQuant framework follows a top-down workflow that effectively integrates existing quantization techniques with our novel contributions.}
\label{fig:quant_pipeline}
\end{figure*}

In this work, we bridge this gap with \textbf{SegQuant}, a framework designed to be both high-performing and compiler-native. 
Instead of relying on dynamic values, our core method, SegLinear, derives its quantization strategy \emph{purely from the static computation graph} (e.g., \texttt{torch.fx} representation). 
By analyzing inherent structural semantics, it aligns advanced quantization with the operational paradigm of modern deployment toolchains, ensuring both accuracy and automated integration.
To better understand the unique challenges of quantizing diffusion models, we begin our investigation with DiT-based architectures~\cite{dit2023}, which are widely adopted and representative of modern diffusion backbones. By examining the spatial semantic heterogeneity in special layers such as AdaNorm, we identify a phenomenon of semantic segmentation emerging in certain computational patterns. Additionally, we observe that the negative activation behavior introduced by activation functions like SiLU significantly impacts the final visual quality. To address these issues, we propose:

\begin{enumerate}
    \item \textbf{SegQuant}, a deployment-oriented and modular framework (Figure~\ref{fig:quant_pipeline}). It functions as a top-down platform, integrating diverse techniques via adaptive search and enabling practical, versatile deployment.
    \item \textbf{SegLinear}, a fully automatic graph-based semantic segmentation method. Unlike manual-rule methods~\cite{li2023qdiffusionquantizingdiffusionmodels}, it algorithmically partitions weights based on static graph semantics, generalizing effectively across diverse architectures.
    \item \textbf{DualScale}, a hardware-native polarity preservation technique. It maintains the fidelity of asymmetric activations (e.g., from SiLU) without requiring performance-penalizing \textit{custom hardware implements}, ensuring compatibility with standard inference engines.
\end{enumerate}

Together, SegQuant achieves an improved trade-off between quantization accuracy and deployment flexibility. Extensive experiments demonstrate that our framework not only enhances the performance of quantized diffusion models but also generalizes well to other architectures.

\section{Related Work}
\label{sec:related_work}

\paragraph{Quantization}
Quantization is a key technique for compressing deep models and enabling efficient inference. Existing works mainly target linear layers and can be grouped into \textit{weight-only} and \textit{joint weight–activation} quantization. 
Weight-only approaches minimize quantization error via analytical or blockwise reconstruction~\cite{adaround2020, brecq2021, gptq2022}, while joint methods introduce smoothing or decomposition strategies to suppress activation outliers~\cite{smoothquant2023, svdquant2025}. 
Industrial toolkits such as TensorRT~\cite{modelopt} provide deployment-ready implementations with kernel fusion and calibration but struggle to handle the complex graph structure of diffusion or transformer-based models.

\paragraph{Quantization for Diffusion Models}
Diffusion models present unique challenges due to their multi-step denoising and heterogeneous architectures. Prior works are largely split by their information source. 
1) Architecture-specific methods, like Q-Diffusion~\cite{li2023qdiffusionquantizingdiffusionmodels}, use \textit{manual, hardcoded rules}. Its shortcut-splitting correctly identifies the bimodal distributions from UNet's skip-connections but is a non-generalizable fix, tightly coupled to that specific architecture. 
2) Data-dependent methods, like PTQ4DiT~\cite{wu2024ptq4dit}, rely on \textit{runtime-dynamic values}, such as timestep-varying activations or salient channels. These numerical heuristics are effective but fundamentally incompatible with modern static graph compilers, creating a ``Compiler Gap" that prevents automated deployment.

Our framework offers a distinct alternative that bridges this gap. SegQuant, through its SegLinear component, also addresses structural heterogeneity, but does so via \emph{automatic graph-based segmentation}. Unlike Q-Diffusion, this segmentation is algorithmically derived from the static graph, not manually defined, enabling generalization beyond UNet. Unlike PTQ4DiT, our analysis is based purely on the \emph{static computation graph structure}, not dynamic numerical distributions. This makes our approach both compiler-native and architecturally general. 
Therefore, temporal reconstruction methods~\cite{tfmq2024, edadm2024} are orthogonal and can be integrated as a SegQuant \textit{Calibrator}.

\paragraph{Video-oriented Quantization}
Approaches like ViDiT-Q and Q-VDiT~\cite{viditq2024, qvdit2024} leverage video-specific heuristics and token-level temporal adaptation. 
SegQuant contrasts by focusing on image synthesis via automatic, general-purpose computation-graph analysis, requiring no such domain-specific heuristics.

\paragraph{Vision Transformer Quantization}
Another line of work~\cite{ptq4vit2023, tsptq2024, adalog2024, adfq2025, ahcptq2025} often introduces nonstandard data formats (e.g., logarithmic) or complex Hessian-based search, targeting edge/FPGA deployment. 
These methods limit standard GPU efficiency. In contrast, our \emph{DualScale} preserves polarity via simple calibration while maintaining native GPU GEMM operations (including CUDA epilogue fusion), prioritizing simplicity and scalable inference.

\paragraph{Optimizer and Calibrator Perspective}
SegQuant unifies diverse quantization techniques through two abstractions: \textit{Optimizer} and \textit{Calibrator}. 
Optimizers such as SmoothQuant, SVDQuant, and DMQ~\cite{smoothquant2023, svdquant2025, dmq2025} reshape activation distributions via analytical scaling, while rotation-based methods like SpinQuant~\cite{spinquant2024} enhance numerical alignment. 
Calibrators such as GPTQ~\cite{gptq2022} and native amax schemes formulate quantization as optimization or matching problems. 
We adopt SmoothQuant and SVDQuant as representative Optimizers, and GPTQ or native amax as Calibrators, to highlight SegQuant’s extensibility and compatibility rather than to compete with any single algorithm.

\paragraph{Our Work}
In contrast to prior work, SegQuant avoids dependence on rigid architectural assumptions (e.g., UNet) or runtime-dynamic heuristics (e.g., timestep variance). 
Instead, it leverages intrinsic semantic structures identified directly from the static computation graph. 
This principled approach, combining automatic semantic segmentation (SegLinear) and hardware-native activation handling (DualScale), enables high-fidelity quantization in a manner that is both generalizable across diverse architectures and inherently compatible with modern deployment pipelines.

\section{Preliminaries}

\paragraph{Quantization}

Quantization reduces model size and computational cost by replacing full-precision floats with low-precision integers or floats.
Two primary paradigms exist: quantization-aware training (QAT), which incorporates quantization effects during training, and post-training quantization (PTQ), which applies quantization after training without requiring fine-tuning.
PTQ is particularly favored in real-world deployments for its simplicity and practicality.

For any given full-precision number $x$, PTQ can apply either \textit{symmetric} or \textit{asymmetric} quantization, both described by:

\[
    \hat{x} = \mathrm{clip}\left(\mathrm{round}\left( \frac{x - z}{s} \right), q_{\min}, q_{\max} \right),
\]
where $\mathrm{round}(\cdot)$ denotes rounding the input to the nearest integer; $s$, $z$, and $[q_{\min}, q_{\max}]$ represent the scale, zero-point, and quantization range, respectively.
The original value $x$ can be approximately recovered by the inverse operation: $x \approx s \cdot (\hat{x} + z)$.

In symmetric quantization, the zero-point $z$ is zero, simplifying the formula and hardware implementation.
Asymmetric quantization allows a nonzero $z$ for greater flexibility but at the cost of increased complexity.

\paragraph{Diffusion Models}
Diffusion models~\cite{ddpm2020} have emerged as a dominant class of generative models for high-fidelity image synthesis.
They learn to reverse a gradual noising process by predicting and removing injected perturbations from data samples.
In standard formulations such as Denoising Diffusion Probabilistic Models (DDPMs), the network is trained to estimate the noise (or equivalently, the reconstruction error) added to an input at each timestep.
More recent variants extend this view: for example, Flow Matching~\cite{flowmatching2023} and Rectified Flow~\cite{rectifiedflow2022} reformulate the objective as vector-field prediction over continuous time, leading to more stable and efficient training dynamics.

Modern diffusion backbones have shifted from UNet-based~\cite{unet2015} designs toward Transformer architectures, notably the DiT family~\cite{dit2023}.
DiT operates directly in the latent space of a pretrained variational autoencoder (VAE)~\cite{vae2022}, treating latent patches as input tokens.
It integrates temporal embeddings to encode diffusion timesteps and employs Adaptive LayerNorm~\cite{adanorm2018} for conditional normalization.
Cross-attention modules allow conditioning on text or other modalities, while a final linear projection predicts the per-token noise residual (i.e., the error term) used to refine the sample iteratively.
This architecture achieves strong scalability and flexibility, making it a representative backbone for large-scale diffusion-based generation systems.

\begin{figure}[h!]
    \centering
    \includegraphics[width=0.3\textwidth]{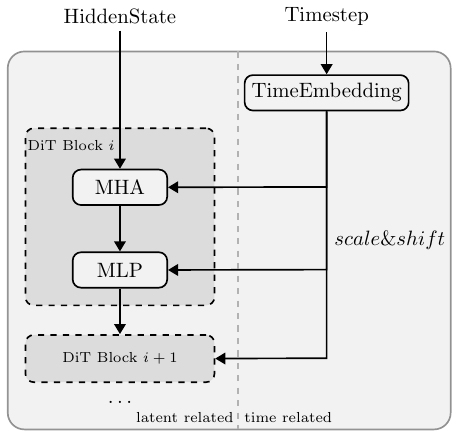}
    \caption{Structural overview of the DiT diffusion model, highlighting latent-related modules (left) and time-related modules (right).}
    \label{fig:dit_structure}
\end{figure}

\paragraph{Graph}

Modern deep learning frameworks such as PyTorch~\cite{torch2019} represent neural networks as directed acyclic graphs (DAGs), where nodes correspond to operations and edges denote tensor dependencies.
This graph abstraction enables various graph-based optimizations such as operator fusion, memory reuse, and scheduling used by AI compilers~\cite{tvm2018,magpy2024}.
In this work, we exploit the structured nature of computation graphs to guide our quantization strategy.
By identifying semantically meaningful operator regions based on patterns in the graph, we enable more effective and context-aware quantization decisions.
\section{SegQuant Framework}

\subsection{SegQuant Overview}

In this section, we present \textbf{SegQuant}, a unified top-down quantization framework designed through an in-depth analysis of both diffusion-specific and general-purpose quantization techniques (Figure~\ref{fig:quant_pipeline}).

SegQuant is a modular framework built on two plug-in components adapted from existing work: the \textit{Optimizer}~\cite{smoothquant2023, dmq2025, svdquant2025} and the \textit{Calibrator}~\cite{gptq2022}. We enhance this base with our proposed \textit{SegLinear} and \textit{DualScale} modules. These interfaces allow users to customize the default (SmoothQuant + GPTQ) by substituting any PTQ method. Finally, the framework includes efficient CUDA kernels to accelerate deployment without compromising quantization integrity.

By combining graph-based analysis with automated configuration, SegQuant provides a flexible and extensible foundation for quantizing diverse model architectures.
Although our empirical evaluation focuses on DiT-based diffusion models, the framework is model-agnostic and can be readily applied to other model architectures.

\subsection{SegLinear}

\subsubsection{Semantic Heterogeneity in Linear Layers}

We posit a general principle: linear layers within complex neural architectures often operate on heterogeneous inputs, where different segments of the input vector encode semantically distinct information. Applying a uniform quantization strategy across such concatenated segments is inherently suboptimal, as these segments may possess unique data distributions and varying sensitivities to quantization noise.

\begin{figure}[h!]
    \centering
    \includegraphics[width=0.9\linewidth]{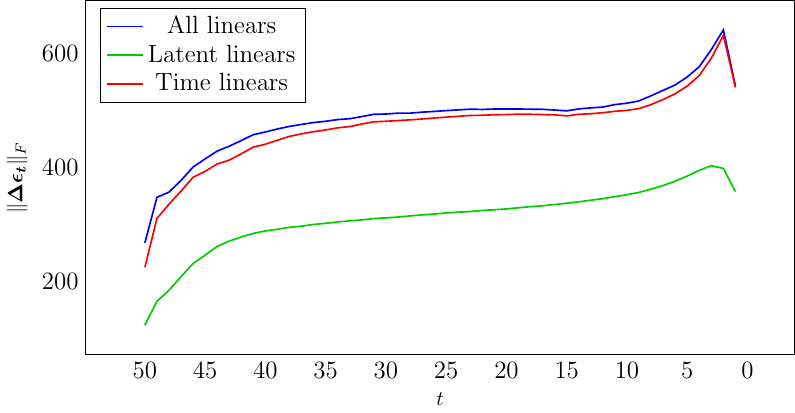}
    \caption{Frobenius norm of error $\|\boldsymbol{\Delta \epsilon_t}\|_F$ over timesteps for INTW8A8 vs. FP16 across linear layers.}
    \label{fig:quant_metrics}
\end{figure}

This challenge is clearly manifested in DiT-based diffusion models, which serve as a strong motivating example. As shown in Figure~\ref{fig:dit_structure}, the diffusion model consists of time-related and latent-related submodules with distinct roles. Our analysis reveals that these components exhibit different degrees of sensitivity to quantization. Specifically, when we apply uniform INTW8A8 quantization across all layers, we find that time-related layers suffer significantly higher error (Figure~\ref{fig:quant_metrics}).

\begin{figure}[h!]
    \centering
    \includegraphics[width=\linewidth]{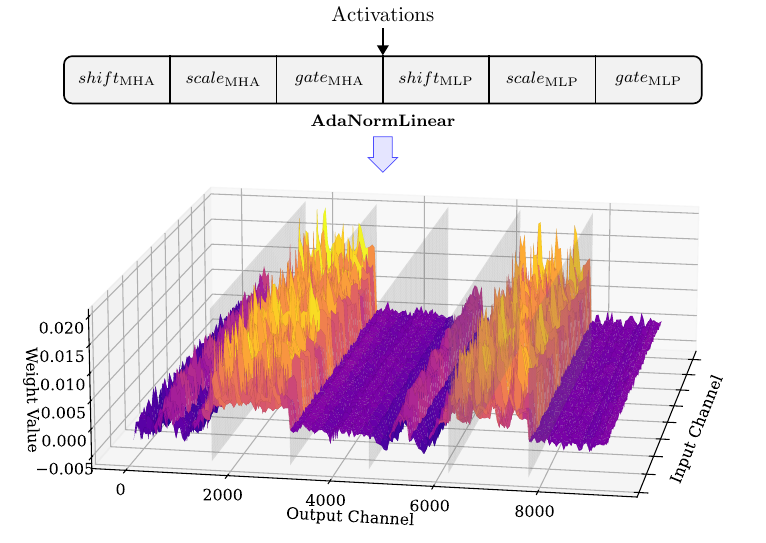}
    \caption{Visualization of weights in \texttt{AdaNorm} within the \texttt{TimeEmbedding} module. The distribution reveals distinct semantic patterns.}
    \label{fig:timeembedding}
\end{figure}

We traced this high sensitivity to the internal structure of key components, such as the \texttt{TimeEmbedding} module. As shown in Figure~\ref{fig:timeembedding}, the weights of linear transformations within this module exhibit distinct semantic patterns. This heterogeneity implies that preceding operations, such as \texttt{Chunk} and \texttt{Split}, implicitly partition the input into multiple semantic branches, each requiring tailored quantization treatment. Uniform quantization fails to account for this internal structure, leading to a significant increase in error.

To address this general problem, such as that found in the DiT time embedding, we propose \textit{SegLinear} (Segmented quantization for linear layers). It is a structure-aware quantization strategy designed to handle this semantic heterogeneity and is applicable across diverse architectures.

\subsubsection{Semantics-Aware Quantization}

SegLinear is a core component of the SegQuant framework, designed to reduce quantization error in computation graphs where linear operations interact with semantic partitioning patterns.
These include operations such as \texttt{chunk}/\texttt{split} (output fragmentation) and \texttt{stack}/\texttt{concat} (input aggregation).
Such operations indicate that a single linear layer operates over semantically distinct segments, which often possess disparate data distributions.
Applying uniform quantization across these segments creates \emph{quantization interference}, where one segment's numerical properties degrade the precision of another.

To prevent this interference, SegLinear performs fine-grained, segment-wise quantization based on these structural semantics.
Specifically, it partitions the weight matrix and corresponding activations according to the graph structure, and crucially, applies quantization independently within each segment.
This isolation preserves the fidelity of all distinct data pathways.
SegLinear supports two primary modes:

\begin{figure*}[htb]
\centering
\includegraphics[width=0.85\linewidth]{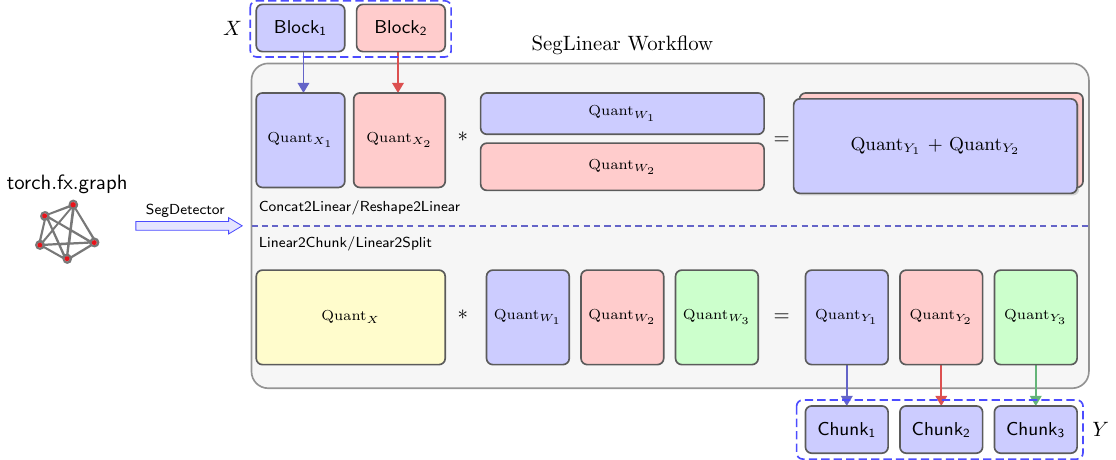}
\caption{\textit{SegLinear} reveals two semantic patterns in the weight matrix that guide quantization.}
\label{fig:segweight}
\end{figure*}

\begin{enumerate}
    \item \textbf{Output-Segmented Quantization.} In the following, we denote by $\hat{\mathbf{M}}$ the  quantized version of any matrix $\mathbf{M}$.
    When the output of a linear layer is followed by operations like \texttt{chunk} or \texttt{split}, we partition the output space and apply quantization independently to each segment.
    Formally, given a linear transformation \( \mathbf{Y} = \mathbf{XW} \), where \( \mathbf{X} \in \mathbb{R}^{m \times k} \) and \( \mathbf{W} \in \mathbb{R}^{k \times n} \), we decompose the weight matrix as:
    \[
    \mathbf{W} = [\mathbf{W}_1, \mathbf{W}_2, \cdots, \mathbf{W}_N], \quad \mathbf{W}_i \in \mathbb{R}^{k \times d_i}, 
    \]
    where \( d_i \) denotes the output dimension of the \( i \)-th partition, such that \( \sum_{i=1}^N d_i = n \).
    Each sub-matrix \( \mathbf{W}_i \) is quantized separately as $\hat{\mathbf{W}}_i$, and the final output is constructed by concatenating segment outputs:
    \[
    \hat{\mathbf{Y}} = [\hat{\mathbf{X}}\hat{\mathbf{W}}_1, \hat{\mathbf{X}}\hat{\mathbf{W}}_2, \cdots, \hat{\mathbf{X}}\hat{\mathbf{W}}_N].
    \]

    \item \textbf{Input-Segmented Quantization.} When the input to a linear layer comes from operations like \texttt{reshape}\footnote{We regard reshape operations involving input dimensions as originating from distinct sources, such as the linear layer merging multiple heads in MHA (multi-head attention).} or \texttt{concat}, we partition the input accordingly and adjust the weight matrix to match.
    Suppose \( \mathbf{X} = [\mathbf{X}_1, \mathbf{X}_2, \cdots, \mathbf{X}_N] \), where each \( \mathbf{X}_i \in \mathbb{R}^{m \times d_i} \) and \( \sum_{i=1}^N d_i = k \).
    Then, the weight matrix is decomposed as:
    \[
    \mathbf{W} = \left[ \mathbf{W}_1^\text{T},\mathbf{W}_2^\text{T},\cdots,
    \mathbf{W}_N^\text{T} \right]^\text{T},
    \quad \mathbf{W}_i \in \mathbb{R}^{d_i \times n}.
    \]
    Each segment is quantized and multiplied separately:
    \[
    \hat{\mathbf{Y}} = \sum_{i=1}^N \hat{\mathbf{X}}_i \hat{\mathbf{W}}_i.
    \]
\end{enumerate}

The segment sizes $d_i$ are automatically inferred from the computation graph via pattern matching over operators such as \texttt{chunk}, \texttt{reshape}, and \texttt{concat} (Figure~\ref{fig:segweight}). 
Unlike Q-Diffusion , which uses a manual rule to address the specific bimodal distributions arising from UNet's skip-connection concatenations, SegLinear performs fully \emph{automatic graph analysis} applicable to arbitrary architectures, including AdaNorm and MHA.
This automation not only generalizes the idea of structural segmentation but also ensures consistency across models without heuristic layer selection.

Moreover, SegLinear complements per-channel quantization by optimizing shared hyperparameters (e.g., migration strength $\alpha$ in SmoothQuant\cite{smoothquant2023}) across semantically coherent channel groups. 
While per-channel quantization treats each output channel independently, SegLinear captures higher-level inter-channel semantics derived from the computation graph, enabling more stable optimization and better scalability to low-bit settings.

\subsection{DualScale}

\subsubsection{Polarity Asymmetric}
\begin{figure}[h!]
\centering
\includegraphics[width=0.8\linewidth]{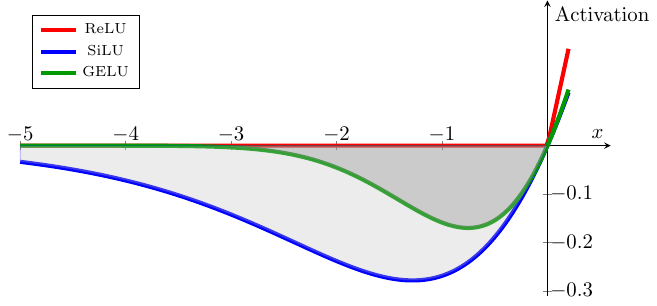}
\caption{Activation curves of SiLU, GELU, and ReLU. The shaded regions show how SiLU and GELU retain negative values, while ReLU suppresses them.}
\label{fig:activation_curves}
\end{figure}

Modern Transformer-based diffusion models, such as DiT~\cite{dit2023}, Stable Diffusion 3~\cite{esser2024scalingrectifiedflowtransformers}, and FLUX.1~\cite{flux2024}, commonly employ polarity-asymmetric activations like \texttt{SiLU} and \texttt{GELU}.
Unlike \texttt{ReLU}, which suppresses negative values, these functions retain dense low-magnitude negatives critical for fine-grained semantics.
Their output is highly skewed, featuring broad positive values and narrow negative ranges (Figure~\ref{fig:activation_curves}), which poses challenges for low-bit quantization where limited bins over-compress negatives.

Existing methods~\cite{ptq4vit2023, tsptq2024, adalog2024, ahcptq2025} that address polarity asymmetry are primarily tailored for ViT or SAM models dominated by \texttt{Softmax} and \texttt{GELU}.
They often redefine bit layouts, scaling bases, or logarithmic quantizers, which require hardware-specific kernels and extensive grid search.
Such customized representations disrupt fused kernel scheduling and are incompatible with high-throughput GPU infrastructures, particularly Tensor Cores that rely on fixed-width PTX (Parallel Thread Execution) and CUDA's epilogue fusion mechanism.

In contrast, DualScale is the first polarity-aware post-training quantization scheme designed for diffusion models.
It preserves the native GPU computation path by maintaining standard bit width and fully leveraging Tensor Core parallelism and CUDA epilogue fusion.
This allows asymmetric quantization without altering data representation or breaking GEMM execution, ensuring high efficiency and scalability across modern GPU architectures.

\begin{table}[h!]
\centering
\caption{Polarity statistics of \texttt{SiLU} and \texttt{GELU} activations from SD3.5-ControlNet on COCO, averaged over 30 timesteps. ``Neg/Pos Ratio'' shows the asymmetry in activation distributions.}
\resizebox{\columnwidth}{!}{%
\begin{tabular}{lccc}
\toprule
\textbf{Layer (Module)} & \textbf{Activation} & \textbf{Channels} & \textbf{Neg/Pos Ratio} \\
\midrule
AdaNorm (DiT) & \texttt{SiLU} & 1536 & 0.955 / 0.021 \\
AdaNorm (Ctrl.) & \texttt{SiLU} & 1536 & 0.645 / 0.338 \\
FFN (DiT) & \texttt{GELU} & 6144 & 0.744 / 0.256 \\
FFN (Ctrl.) & \texttt{GELU} & 6144 & 0.589 / 0.400 \\
\bottomrule
\end{tabular}
}
\label{tab:activation_stats}
\end{table}

\begin{figure}[h!]
  \centering
  \includegraphics[width=\linewidth]{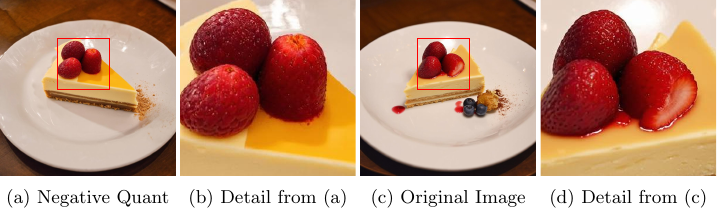}
  \caption{
    Visual impact of \textit{negative-range quantization} in SD3.5 (timestep 60, COCO). 
    Only the negative part of activations is quantized to isolate its contribution to image details. 
    (a) and (c) show full images; (b) and (d) zoom in to highlight detail and range loss.
  }
  \label{fig:quant_effect}
\end{figure}

We also analyzed activation outputs channel-wise across several representative layers from SD3.5-ControlNet using the COCO dataset (averaged over 30 timesteps).
As shown in Table~\ref{tab:activation_stats}, a large proportion of channels consistently exhibit negative values.
For instance, over 60\%-70\% of channels in certain AdaNorm and FFN layers remain predominantly negative during inference.
However, conventional PTQ methods typically apply a single global scale (either symmetric or asymmetric) across the entire activation range.
Due to the large spread of positive values, compression of the narrower negative range (e.g., $[-0.3, 0]$) can become relatively stronger. For instance, in some activation layers, positive values may reach up to $3.5$, whereas negatives rarely exceed $-0.26$.

Even with asymmetric quantization, which shifts the zero-point, the uniform distribution of bins leads to poor resolution for negative values.
Figure~\ref{fig:quant_effect} shows that negative activations reserve high-frequency details and textural consistency, which is crucial for high-fidelity generation.

\subsubsection{Polarity Preserving Quantization via DualScale}

To address the issue of polarity asymmetry, we propose the so-called \textit{DualScale} strategy.
Unlike standard methods that use a single scaling factor across the full activation range, DualScale applies distinct scales to the negative and non-negative regions, thereby preserving resolution in both regions, particularly in the narrow but semantically important negative range.

Let $x \in \mathbb{R}$ denote an activation value along the computation path from an activation function to its subsequent linear operator (referred to as the \textit{act-to-linear} segment).
The dual-scale quantization function $Q_{\text{dual}}(x)$ is defined as:
\[
Q_{\text{dual}}(x) = 
\begin{cases}
\mathrm{round}\left( \frac{x}{s_{-}} \right), & x < 0 \\
\mathrm{round}\left( \frac{x}{s_{+}} \right), & x \geq 0,
\end{cases}
\]
where $s_{-}$ and $s_{+}$ denote the step sizes for the negative and positive regions, respectively. These are computed as:
\[
s_{-} = \frac{|\min(x)|}{q_{\min}}, \quad
s_{+} = \frac{\max(x)}{q_{\max}},
\]
where $q_{\min}$ and $q_{\max}$ denote the quantization range.

The dual-scale quantization is applied to the activation matrix \( \mathbf{X} \in \mathbb{R}^{m \times k} \) in a linear layer, where \( \mathbf{W} \in \mathbb{R}^{k \times n} \) is the weight matrix.
Since polarity asymmetry mainly comes from activations, we apply dual-scale quantization to \( \mathbf{X} \) only, keeping \( \mathbf{W} \) in standard low precision.
To preserve resolution and avoid destructive rounding across polarities, we decompose \( \mathbf{X} \) into its non-negative and negative parts using element-wise masks:
\[
\mathbf{X}_{+} = \max(\mathbf{X}, 0), \quad \mathbf{X}_{-} = \min(\mathbf{X}, 0).
\]

The key idea is to separately quantize and process the positive and negative channels, then linearly combine the results after matrix multiplication:
\begin{align*}
\mathbf{Y} = \mathbf{X} \mathbf{W} \notag &\approx \operatorname{DeQuant}(Q_{\text{dual}}(\mathbf{X_{+} + \mathbf{X_{-}}}) Q(\mathbf{W})) \notag \\
           &= \left( s_{+}\cdot \mathbf{\hat{X}_{+}} + s_{-}\cdot \mathbf{\hat{X}_{-}} \right)\cdot \left(s_{w}\cdot\mathbf{\hat{W}}\right) \notag \\
           &= s_{+} s_{w}\cdot\left(\mathbf{\hat{X}_{+}}\mathbf{\hat{W}}\right) + s_{-} s_{w}\cdot \left(\mathbf{\hat{X}_{-}}\mathbf{\hat{W}}\right),
\end{align*}
where $\mathbf{\hat{A}} = \mathrm{round}\left( \mathbf{A}/{s_A} \right)$ denotes the quantized version of matrix $\mathbf{A}$ scaled by $s_A$.

This dual-scale quantization avoids inverse zero-point correction, enabling simple output reconstruction with fixed positive and negative scales (see Appendix).
Crucially, the apparent overhead of two matrix multiplications is eliminated in practice. The computation of $\hat{X}_{+}\hat{W}$ and $\hat{X}_{-}\hat{W}$ is implemented as a single, highly efficient BatchedGEMM operation using libraries like CUTLASS~\cite{cutlass2025}. This allows both computations to execute in parallel within one kernel launch, after which the scaled results are combined in a fused epilogue. This ensures our method fully leverages hardware acceleration for standard integer matrix multiplication without introducing latency from custom operators or extra kernel launches. It preserves small negative values often lost in standard quantization and integrates into Transformer MLPs, AdaNorm, and diffusion embeddings without retraining or custom operations.

\section{Experiments}
\subsection{Main Results}
\paragraph{Setup.} We evaluate three representative text-to-image models: Stable Diffusion 3.5 Medium (2B)~\cite{esser2024scalingrectifiedflowtransformers}, FLUX.1-dev (12B, DiT-based)~\cite{flux2024}, and SDXL (UNet-based)~\cite{sdxl2023}. 
We compare our SegQuant-A (AMax) and SegQuant-G (GPTQ) variants against five PTQ baselines: Q-Diffusion~\cite{li2023qdiffusionquantizingdiffusionmodels}, PTQD~\cite{zju2023ptqd}, PTQ4DiT~\cite{wu2024ptq4dit}, a reinforced baseline Smooth+ (combining SmoothQuant~\cite{smoothquant2023} with GPTQ~\cite{gptq2022}), TAC-Diffusion~\cite{yao2024timestepaware}, and SVDQuant~\cite{svdquant2025}. 
Following recent protocols~\cite{svdquant2025}, evaluation is performed on 5,000 randomly sampled images from COCO~\cite{lin2015microsoft}, MJHQ-30K~\cite{mjhq30k2024}, and DCI~\cite{dci2024}. 
Metrics include FID~\cite{fid2017}, LPIPS~\cite{lpips2018}, PSNR, SSIM~\cite{ssim2004}, and Image Reward~\cite{imagereward2023} to cover distributional, perceptual, and pixel-level fidelity. 
All experiments are conducted on GPUs with Ada Lovelace architecture, equipped with 24GB and 48GB of VRAM.
Calibration uses 256 images for SD3/SDXL, 64 for 8-bit FLUX, and 32 for 4-bit, with all models using 50 sampling steps under the default scheduler. 
We report both SegQuant-A (AMax) and SegQuant-G (GPTQ) variants; implementation details appear in Appendix~\ref{appendix:exp-detail}.

\newcommand{\firstmainimgwidth}{0.139\textwidth}
\newcommand{\firstmaincapwidth}{0.98\textwidth}
\setlength{\fboxsep}{0pt}
\setlength{\fboxrule}{0.2pt}
\begin{figure*}[htbp]
    \centering
    \begin{minipage}[t]{0.495\textwidth}
        \centering
        \begin{minipage}[t]{\firstmainimgwidth}
            \centering \tiny{FP16}
        \end{minipage}
        \hfill
        \begin{minipage}[t]{\firstmainimgwidth}
            \centering \tiny{PTQD}
        \end{minipage}%
        \begin{minipage}[t]{\firstmainimgwidth}
            \centering \tiny{PTQ4DiT}
        \end{minipage}%
        \begin{minipage}[t]{\firstmainimgwidth}
            \centering \tiny{TAC-Diffusion}
        \end{minipage}%
        \begin{minipage}[t]{\firstmainimgwidth}
            \centering \tiny{Smooth+}
        \end{minipage}%
        \begin{minipage}[t]{\firstmainimgwidth}
            \centering \tiny{\textbf{SegQuant-A}}
        \end{minipage}%
        \begin{minipage}[t]{\firstmainimgwidth}
            \centering \tiny{\textbf{SegQuant-G}}
        \end{minipage}%
    \end{minipage}
    \hfill
    \begin{minipage}[t]{0.495\textwidth}
        \centering
        \begin{minipage}[t]{\firstmainimgwidth}
            \centering \tiny{FP16}
        \end{minipage}
        \hfill
        \begin{minipage}[t]{\firstmainimgwidth}
            \centering \tiny{PTQD}
        \end{minipage}%
        \begin{minipage}[t]{\firstmainimgwidth}
            \centering \tiny{PTQ4DiT}
        \end{minipage}%
        \begin{minipage}[t]{\firstmainimgwidth}
            \centering \tiny{TAC-Diffusion}
        \end{minipage}%
        \begin{minipage}[t]{\firstmainimgwidth}
            \centering \tiny{Smooth+}
        \end{minipage}%
        \begin{minipage}[t]{\firstmainimgwidth}
            \centering \tiny{\textbf{SegQuant-A}}
        \end{minipage}%
        \begin{minipage}[t]{\firstmainimgwidth}
            \centering \tiny{\textbf{SegQuant-G}}
        \end{minipage}%
    \end{minipage}

    \vspace{0.6ex}

    \begin{minipage}[t]{0.495\textwidth}
        \centering
        \fbox{\includegraphics[width=\firstmainimgwidth]{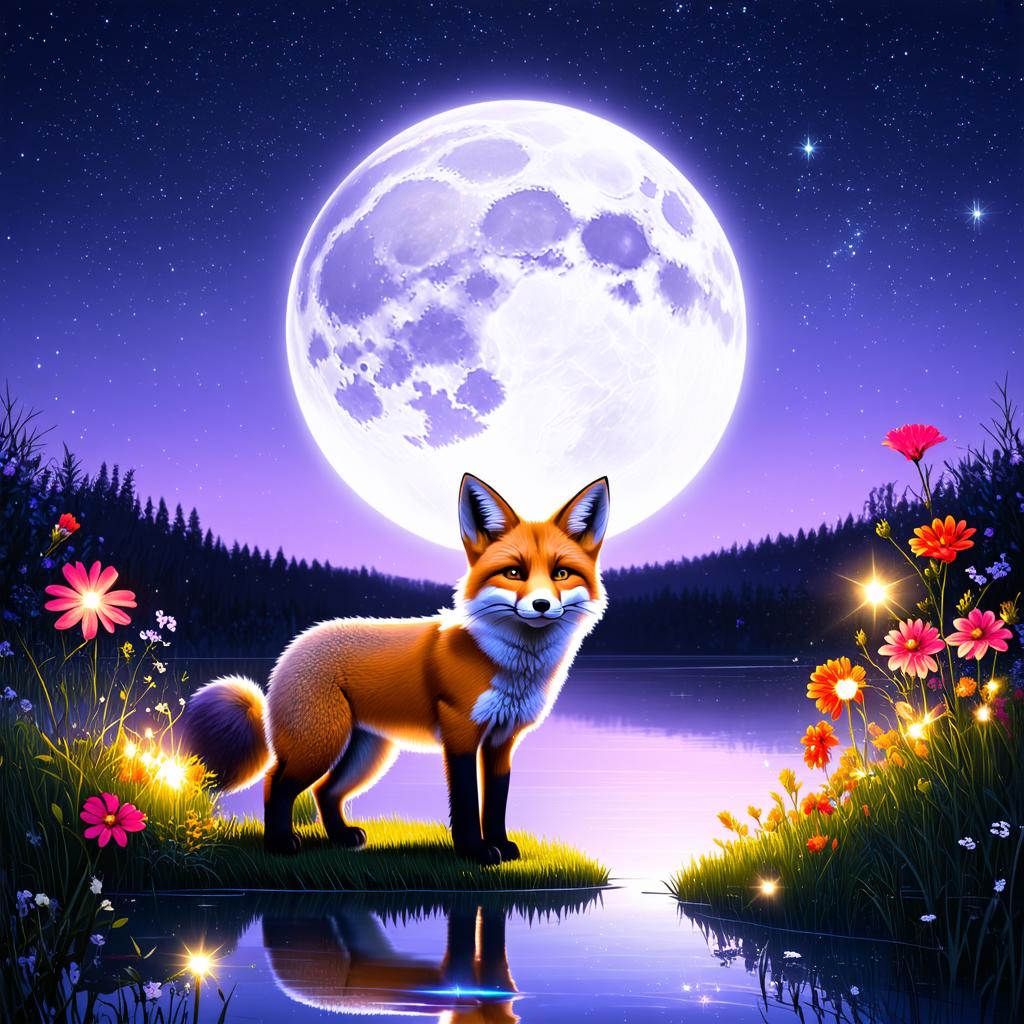}}
        \hfill
        \fbox{\includegraphics[width=\firstmainimgwidth]{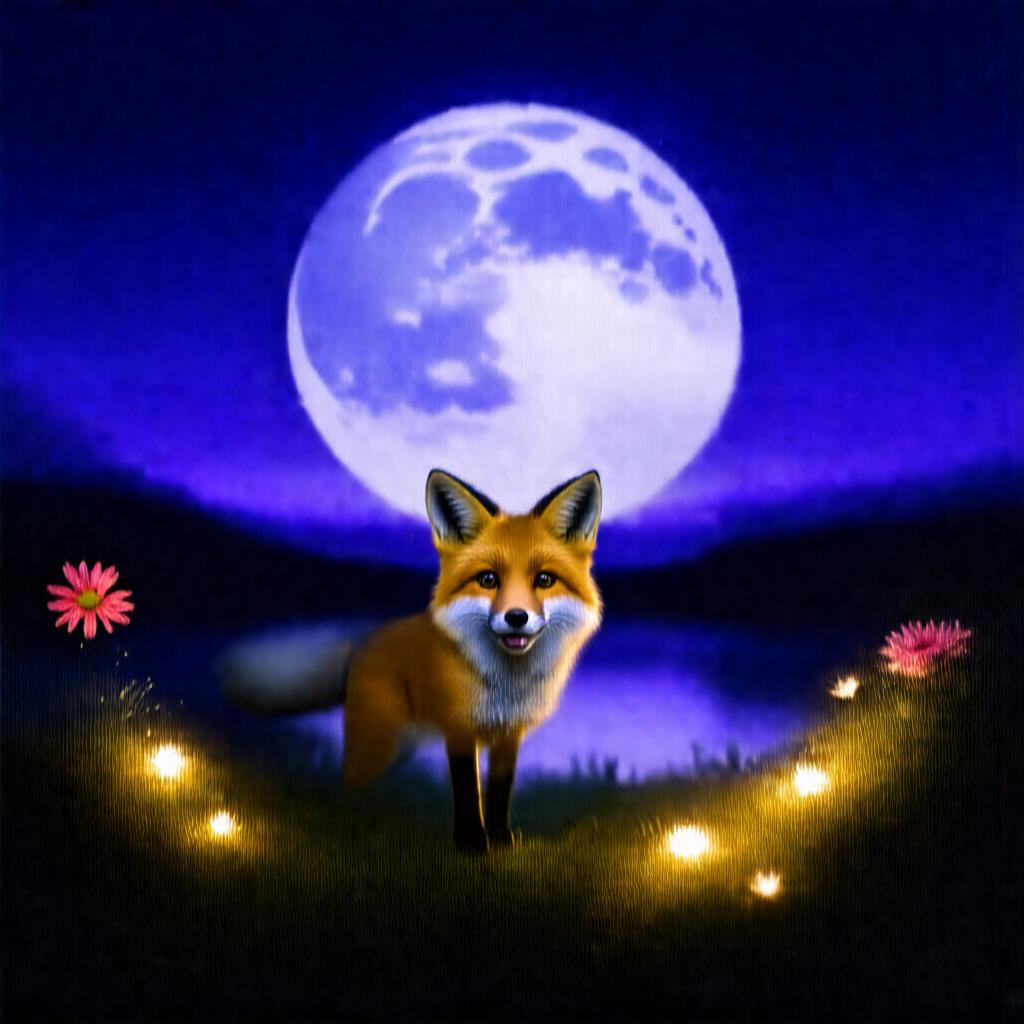}}%
        \fbox{\includegraphics[width=\firstmainimgwidth]{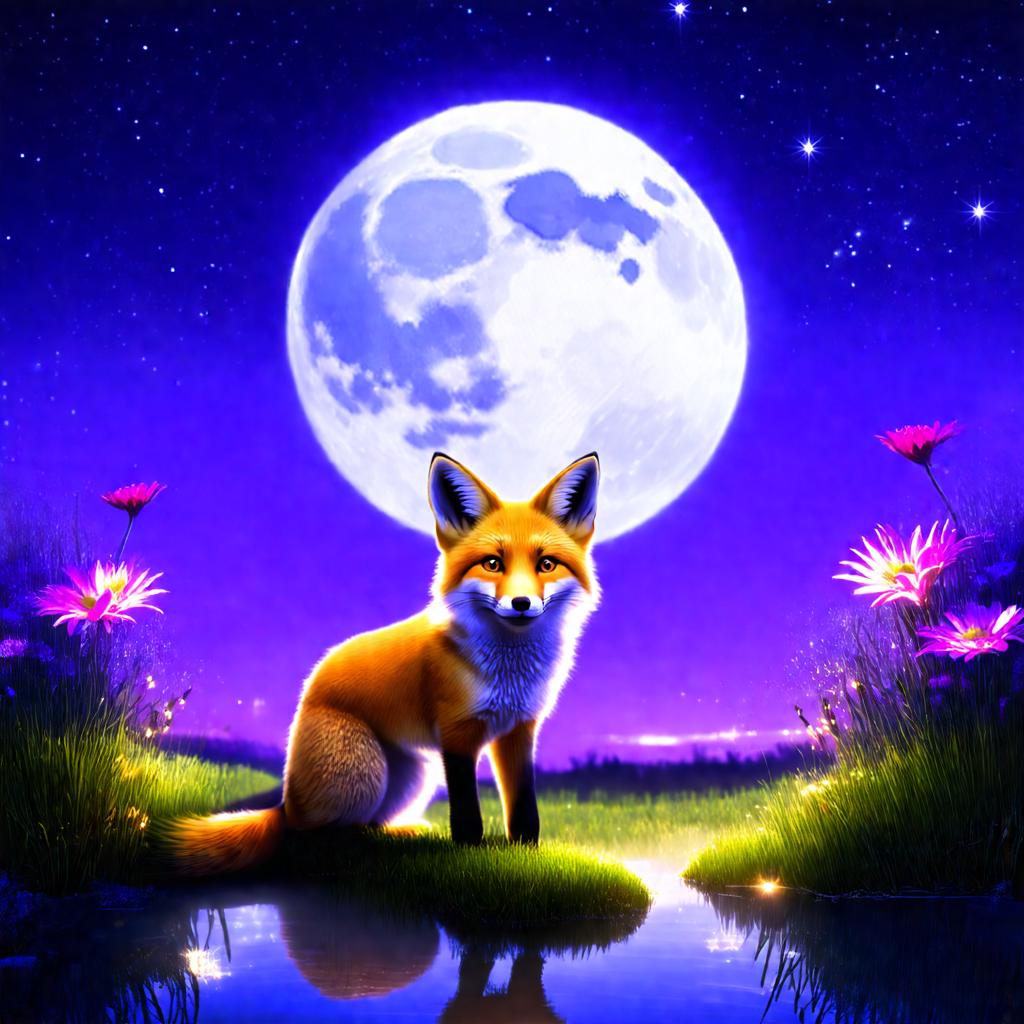}}%
        \fbox{\includegraphics[width=\firstmainimgwidth]{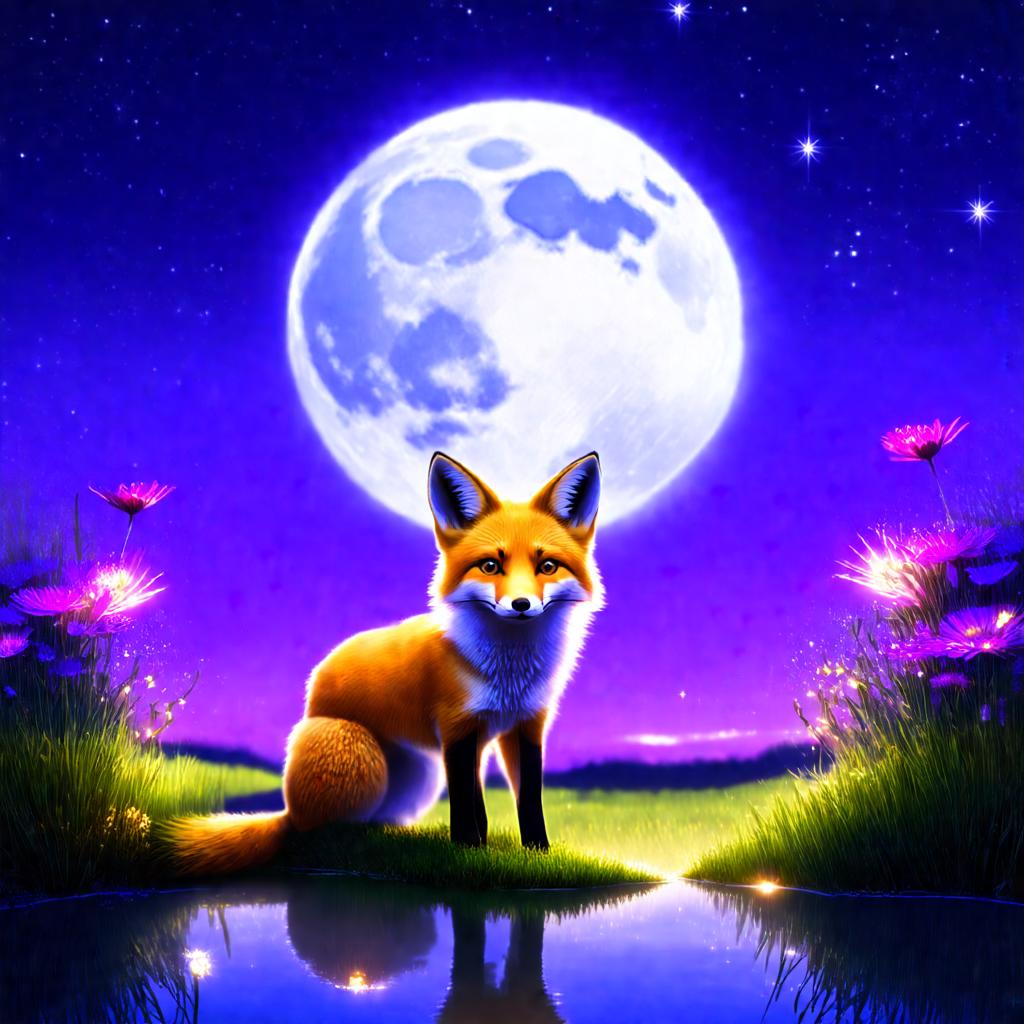}}%
        \fbox{\includegraphics[width=\firstmainimgwidth]{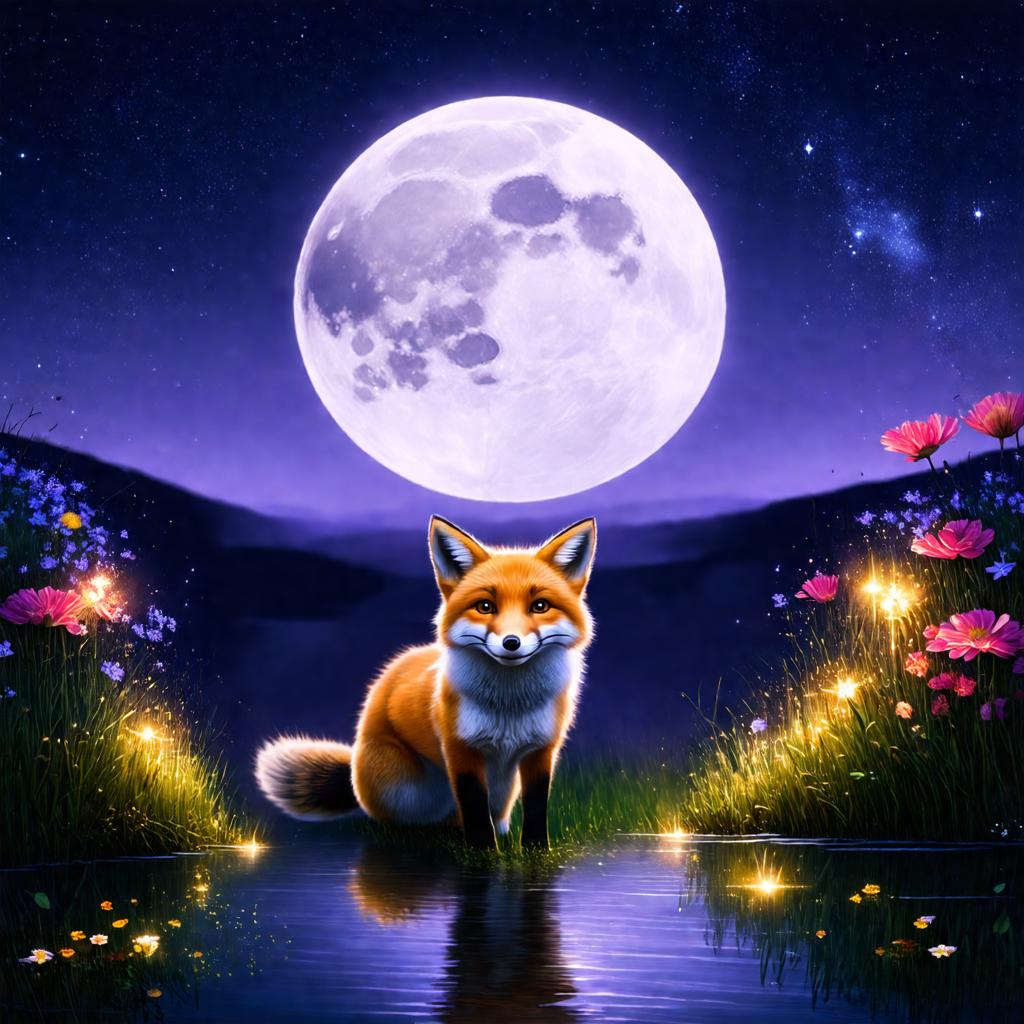}}%
        \fbox{\includegraphics[width=\firstmainimgwidth]{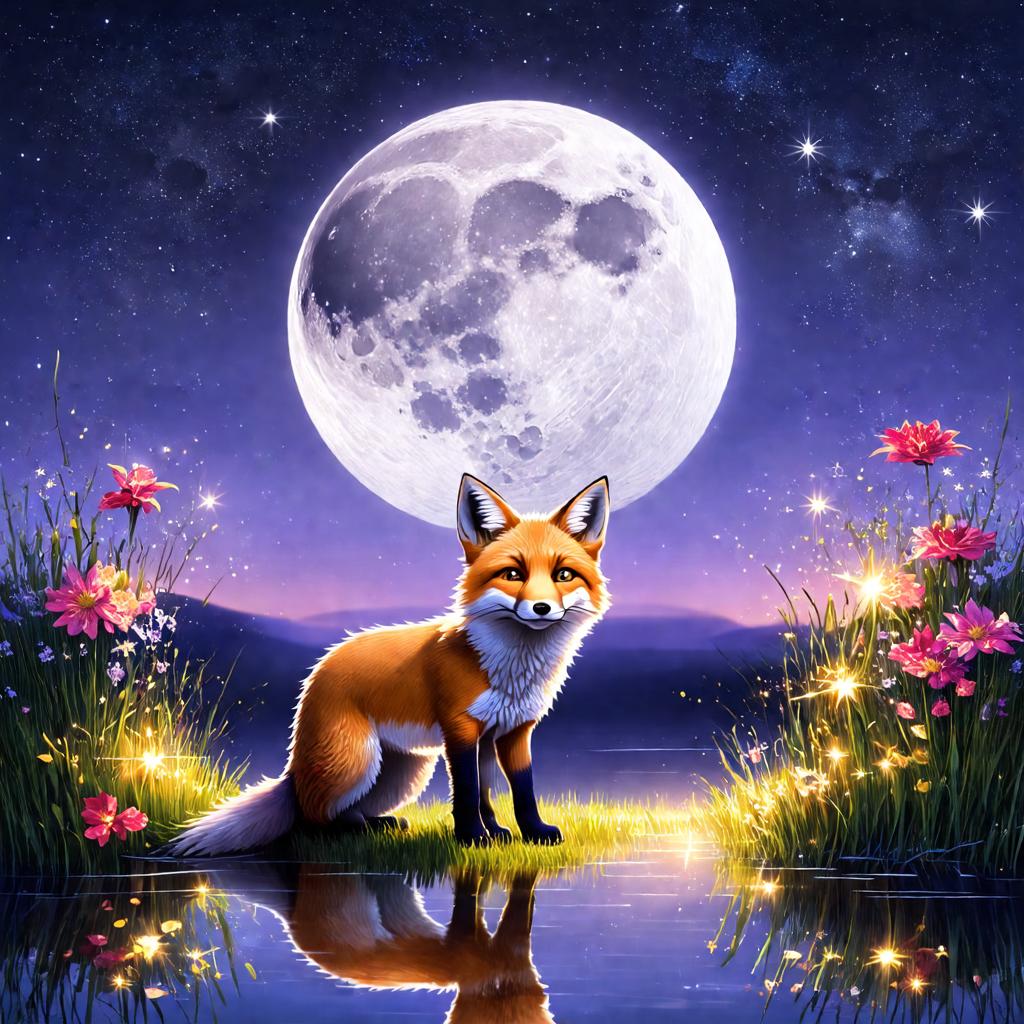}}%
        \fbox{\includegraphics[width=\firstmainimgwidth]{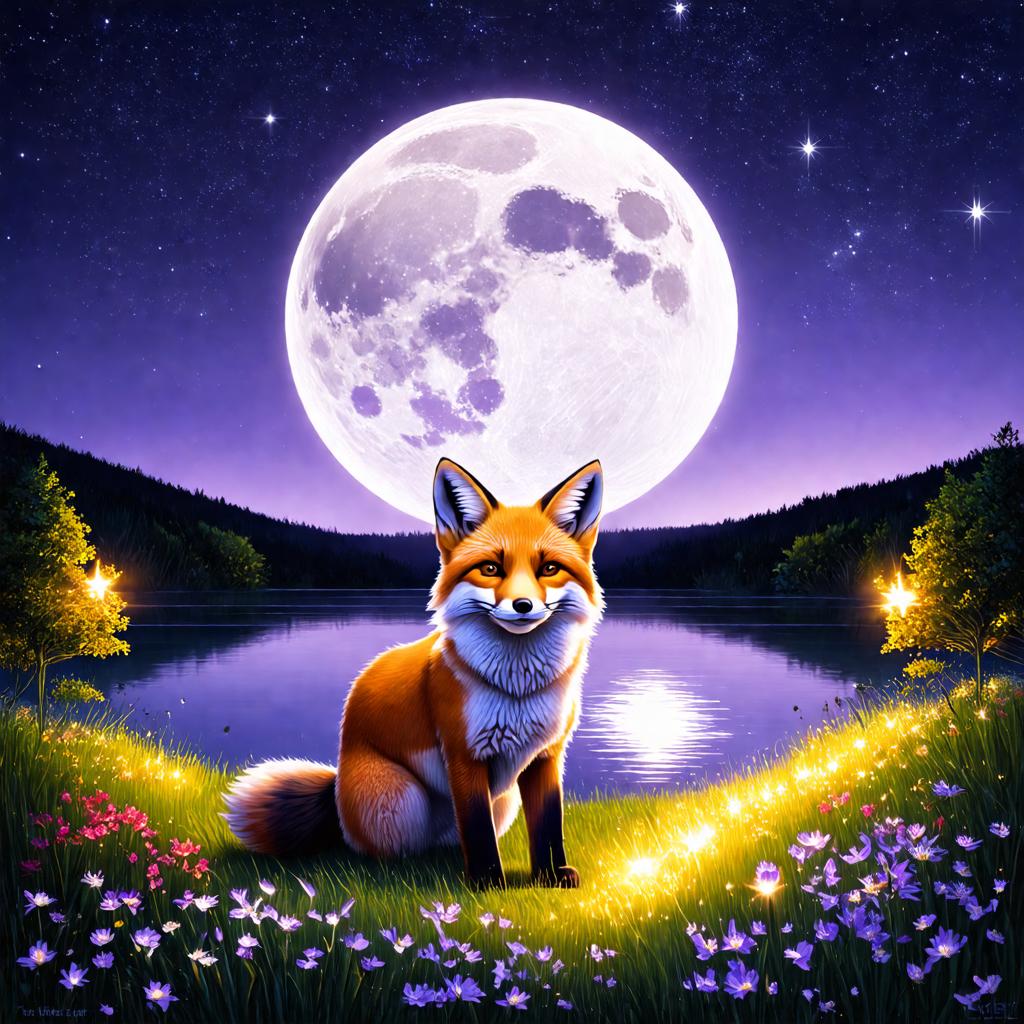}}\\[0.5ex]
        \hfill
        \vspace{-20pt}
        \caption*{
            \begin{minipage}{\firstmaincapwidth}
            \centering
                \tiny{Prompt: \textit{thor with drone, with neon lights in the background, Intricate, Highly detailed, Sharp focus, Digital painting, Artstation, Concept art, inspired by blade runner, ghost in the shell and cyberpunk 2077, art by rafal wechterowicz and khyzyl saleem}}
            \end{minipage}
        }
    \end{minipage}
    \hfill
    \begin{minipage}[t]{0.495\textwidth}
        \centering
        \fbox{\includegraphics[width=\firstmainimgwidth]{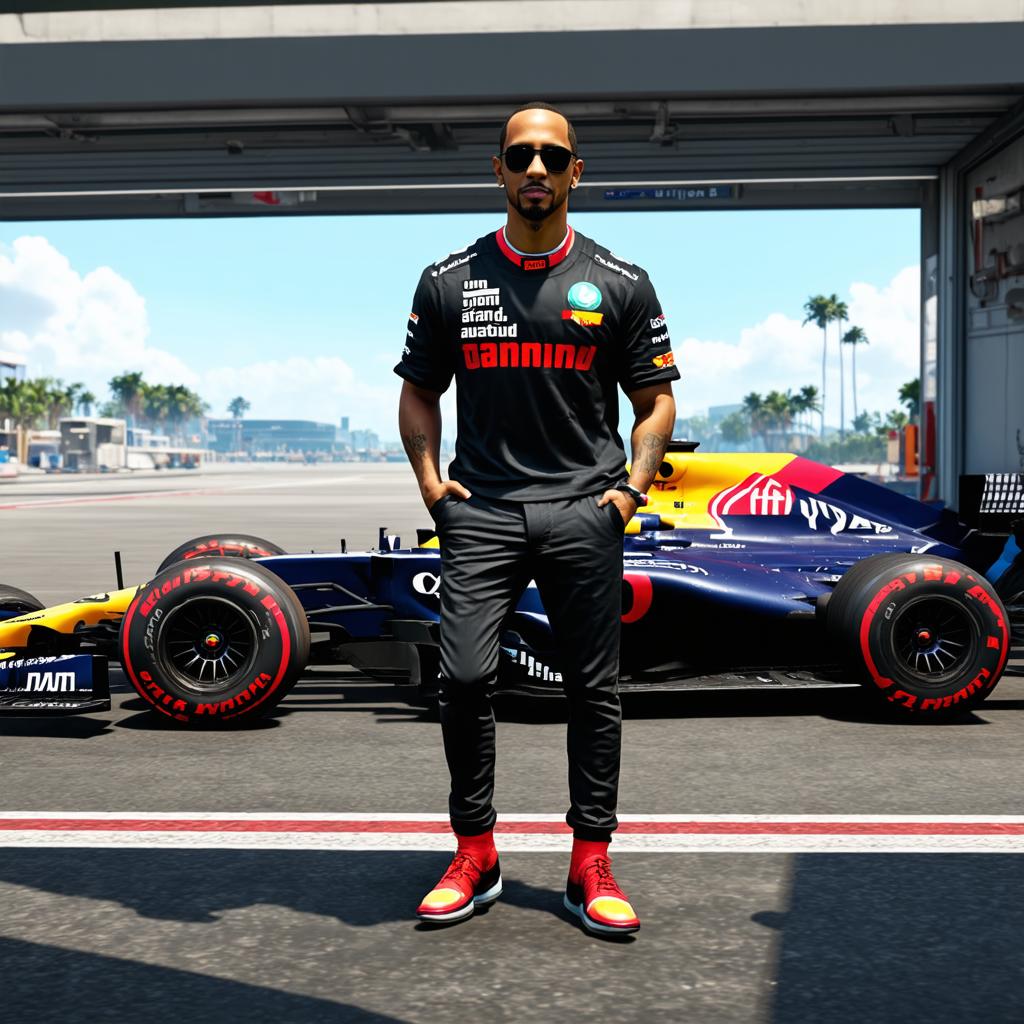}}
        \hfill
        \fbox{\includegraphics[width=\firstmainimgwidth]{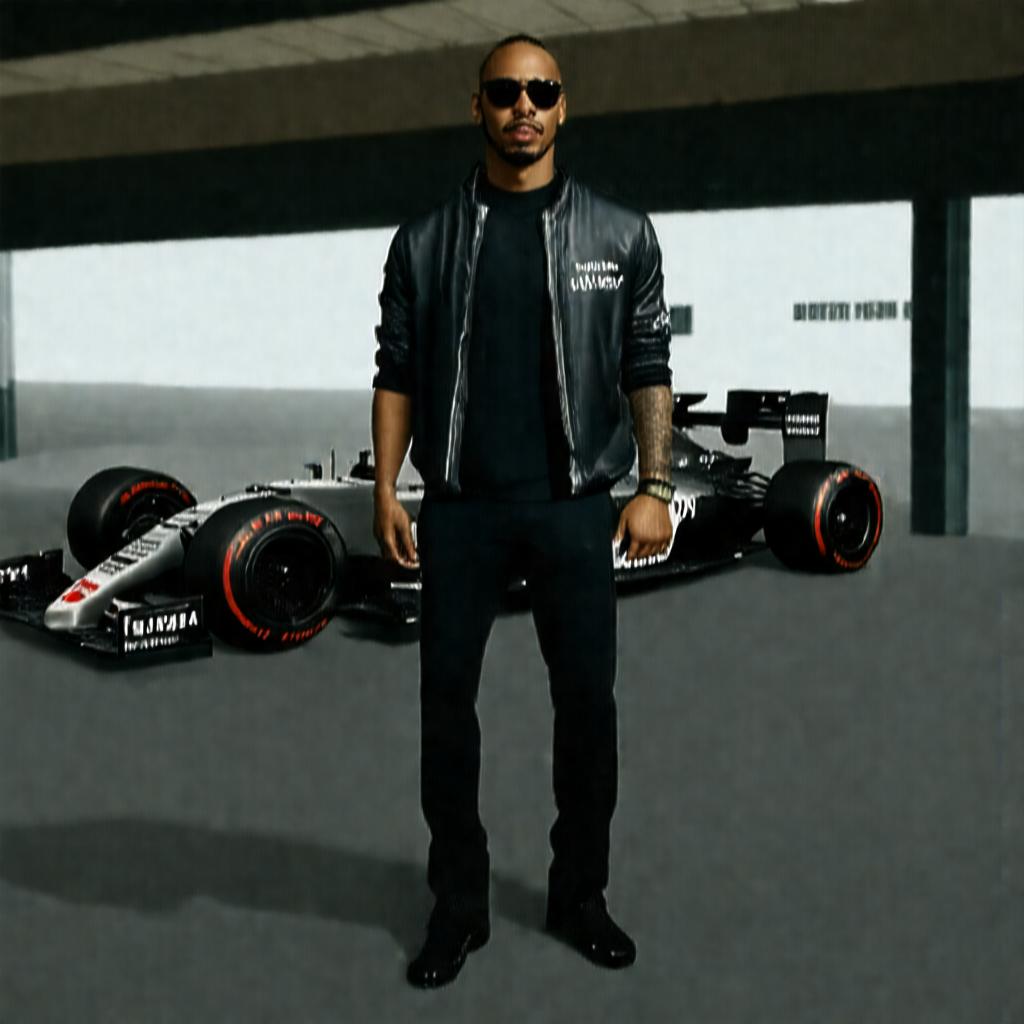}}%
        \fbox{\includegraphics[width=\firstmainimgwidth]{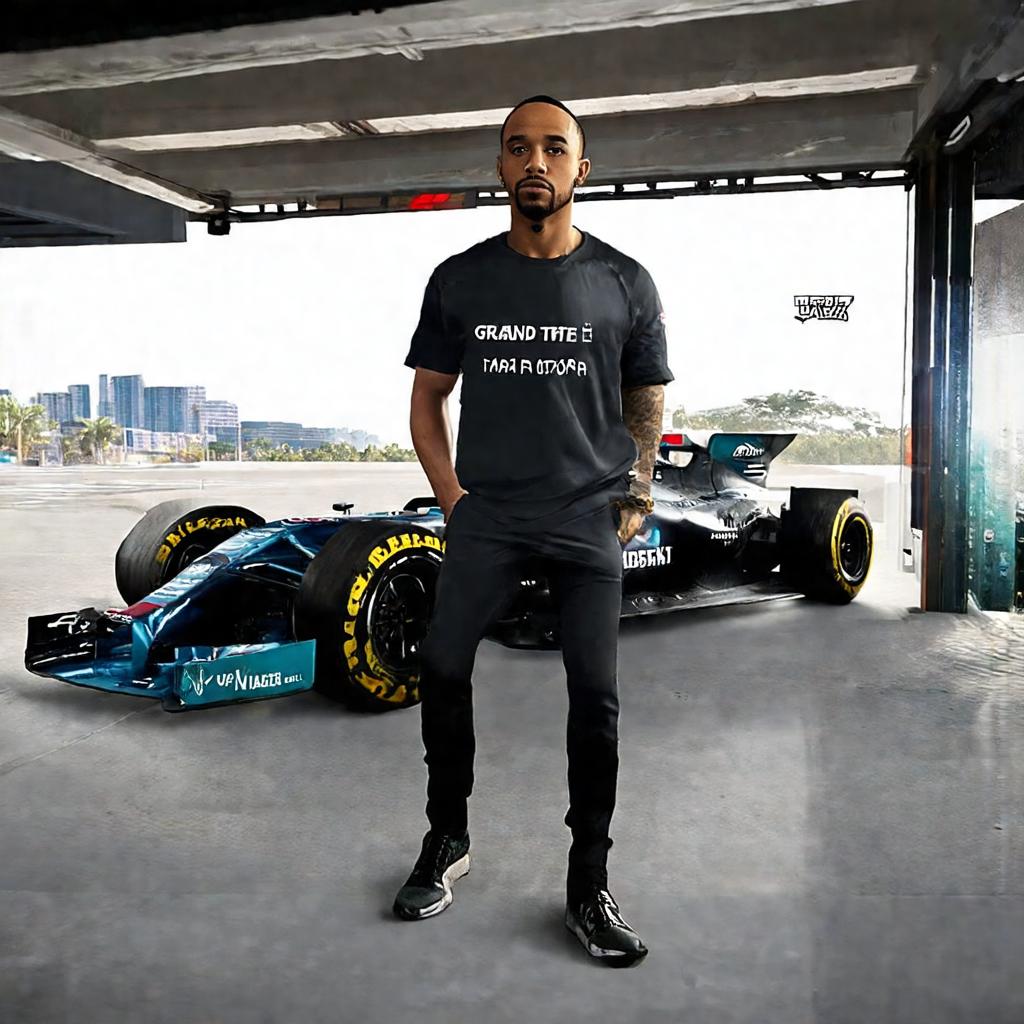}}%
        \fbox{\includegraphics[width=\firstmainimgwidth]{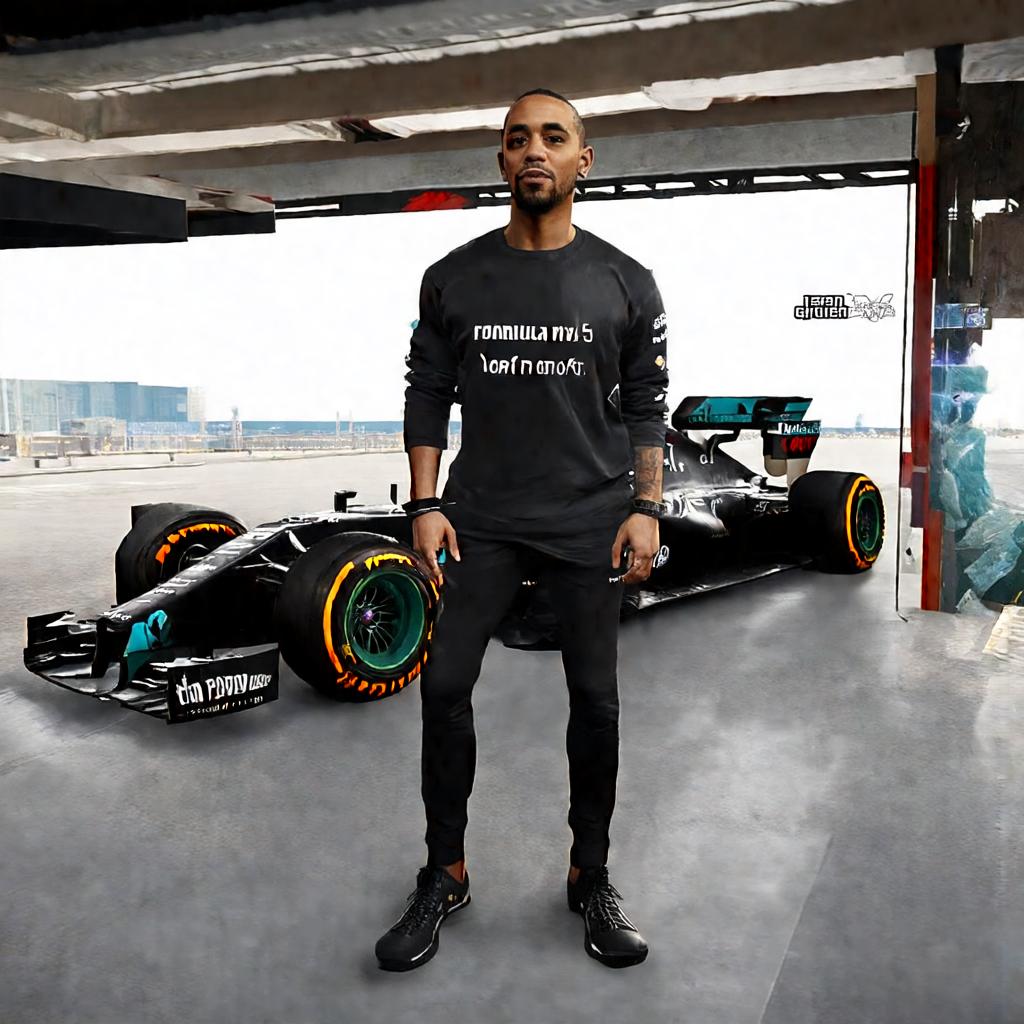}}%
        \fbox{\includegraphics[width=\firstmainimgwidth]{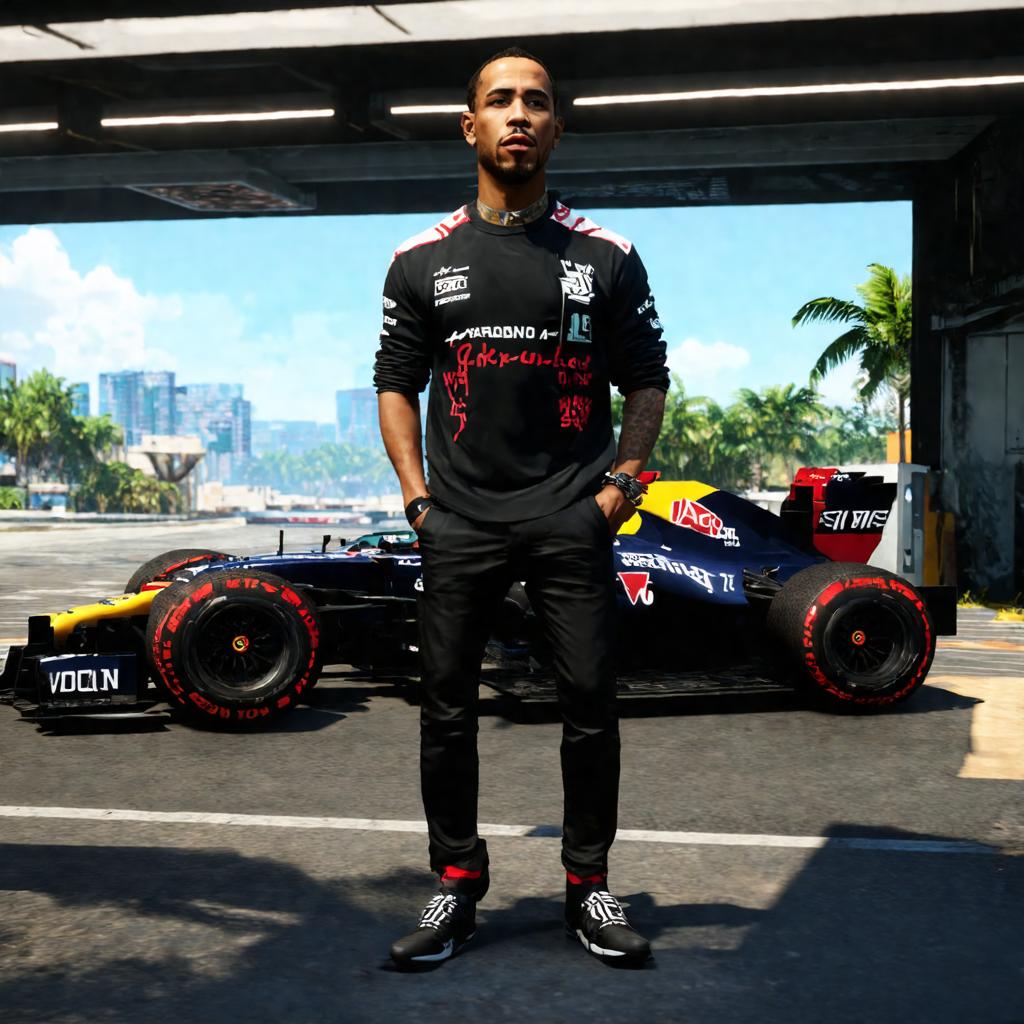}}%
        \fbox{\includegraphics[width=\firstmainimgwidth]{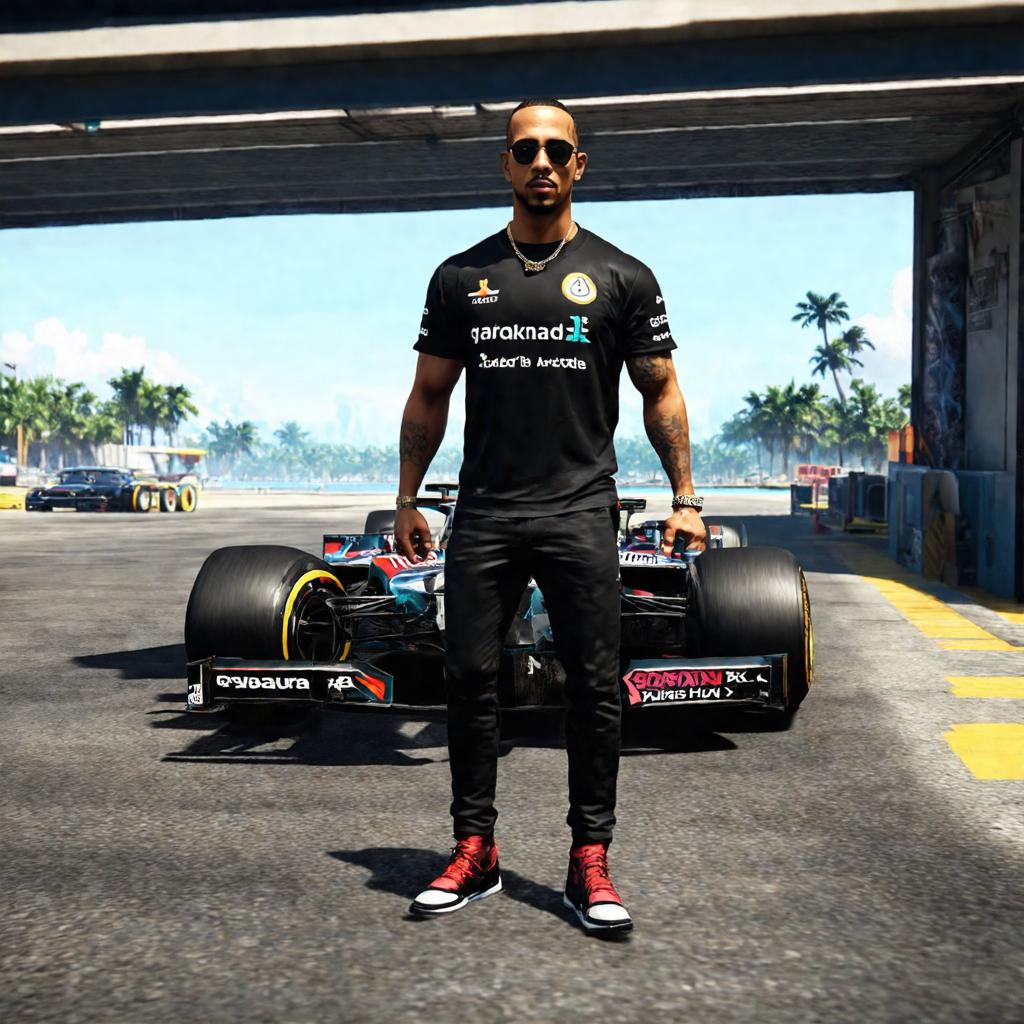}}%
        \fbox{\includegraphics[width=\firstmainimgwidth]{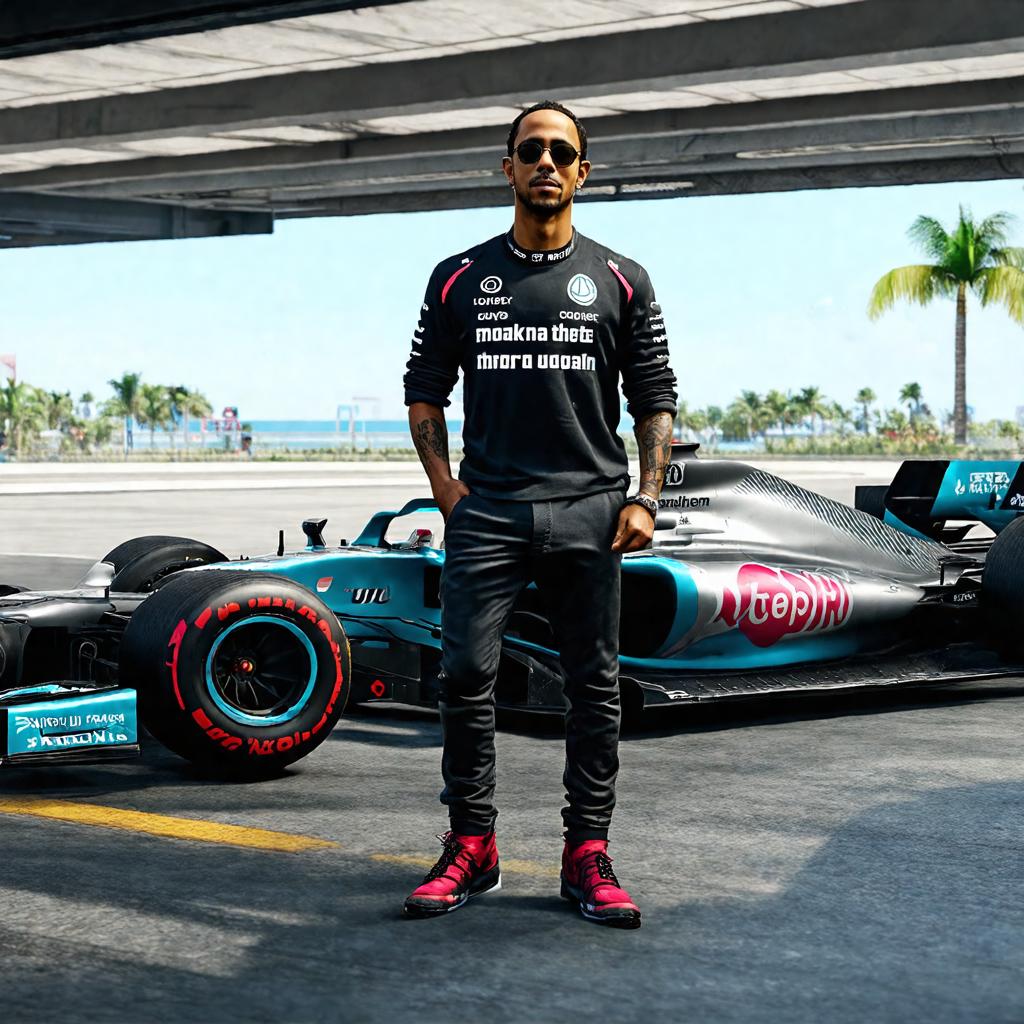}}\\[0.5ex]
        \hfill
        \vspace{-20pt}
        \caption*{
            \begin{minipage}{\firstmaincapwidth}
            \centering
                \tiny{Prompt: \textit{Lewis Hamilton standing infront of his Formula 1 pit garage and Formula 1 car in the style of a GTA 5 grand theft auto 5 loading screen miami}}
            \end{minipage}
        }
    \end{minipage}
    \vspace{-8pt}
    \caption{
        Partial visualization of main results on the MJHQ dataset with SD3.5 and W8A8 DiT quantization.
    }
    \label{fig:main_pics}
\end{figure*}

\newcommand{\imgwidth}{0.19\textwidth}
\newcommand{\capwidth}{0.95\textwidth}
\setlength{\fboxsep}{0pt}
\setlength{\fboxrule}{0.2pt}
\begin{figure*}[htbp]
    \centering
    \begin{minipage}[t]{0.49\textwidth}
        \centering
        \fbox{\includegraphics[width=\imgwidth]{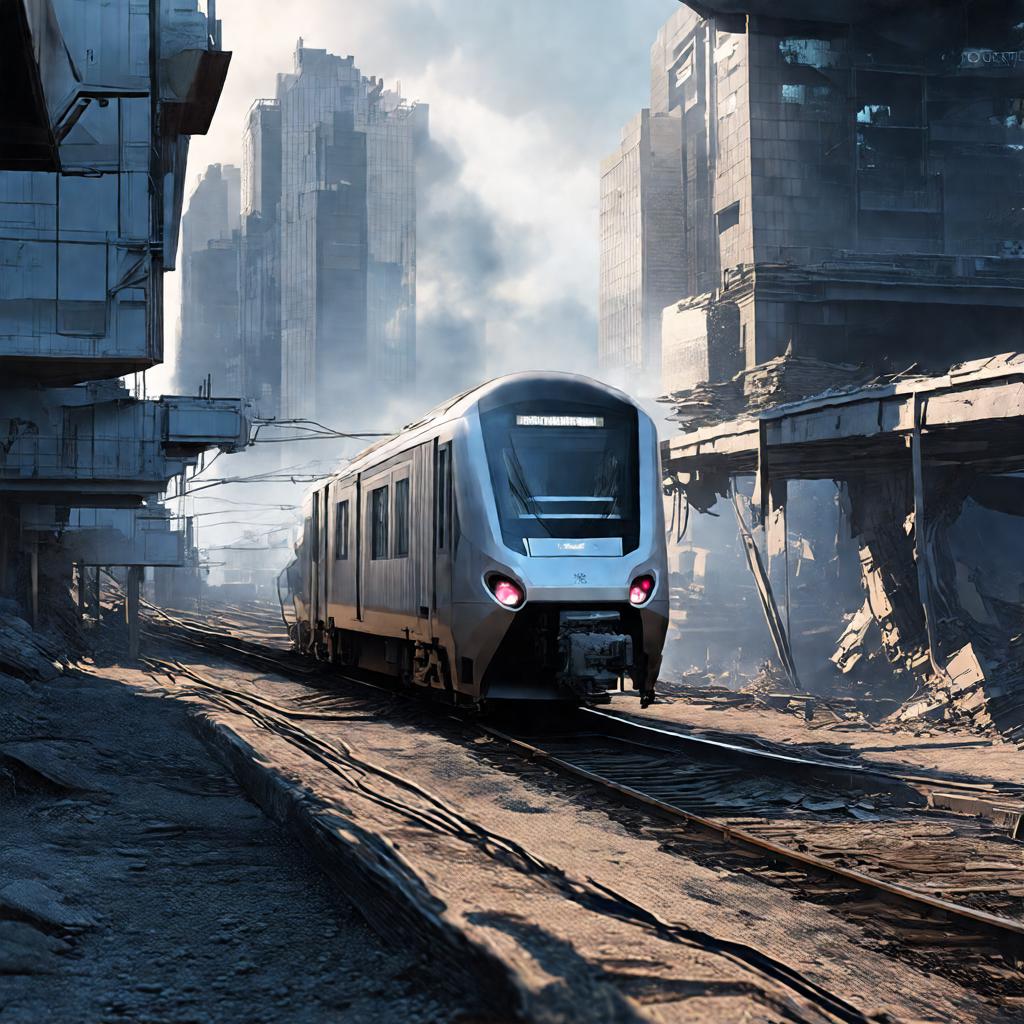}}
        \fbox{\includegraphics[width=\imgwidth]{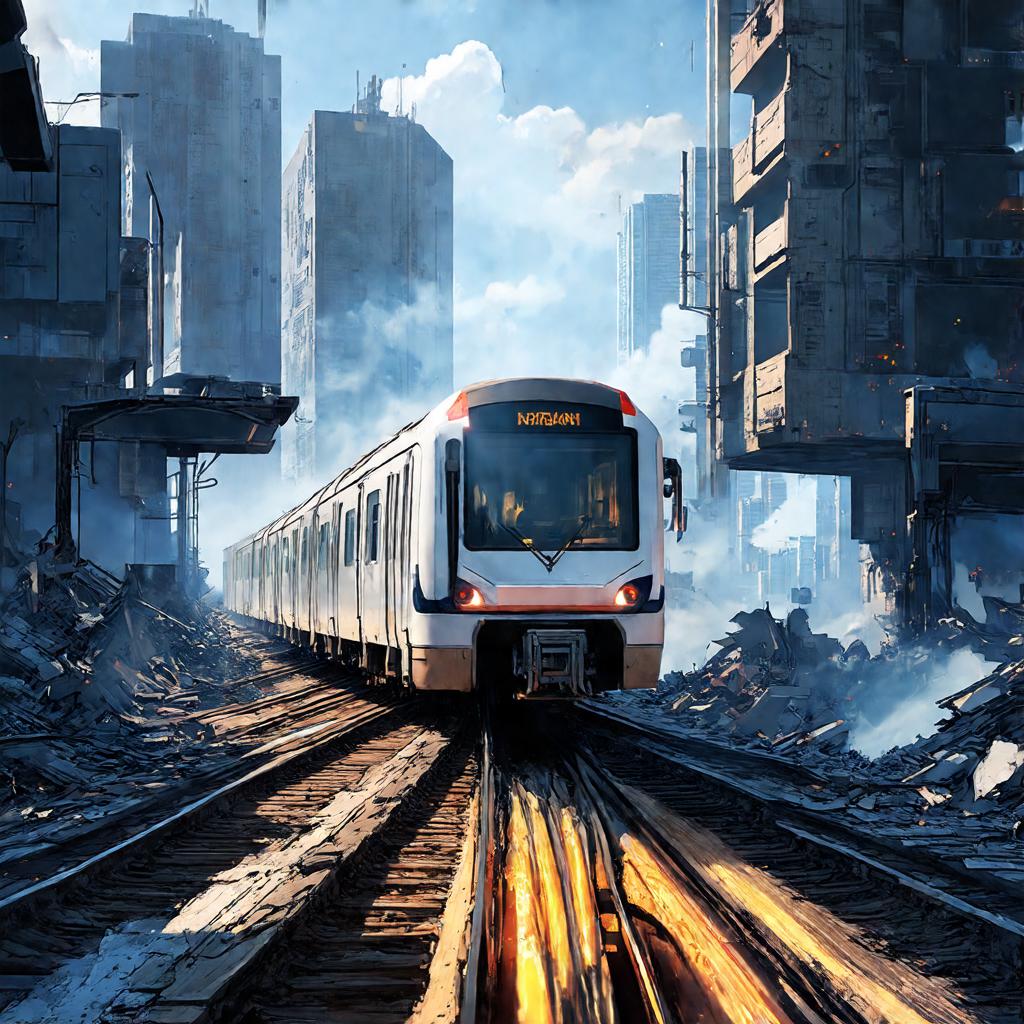}}
        \fbox{\includegraphics[width=\imgwidth]{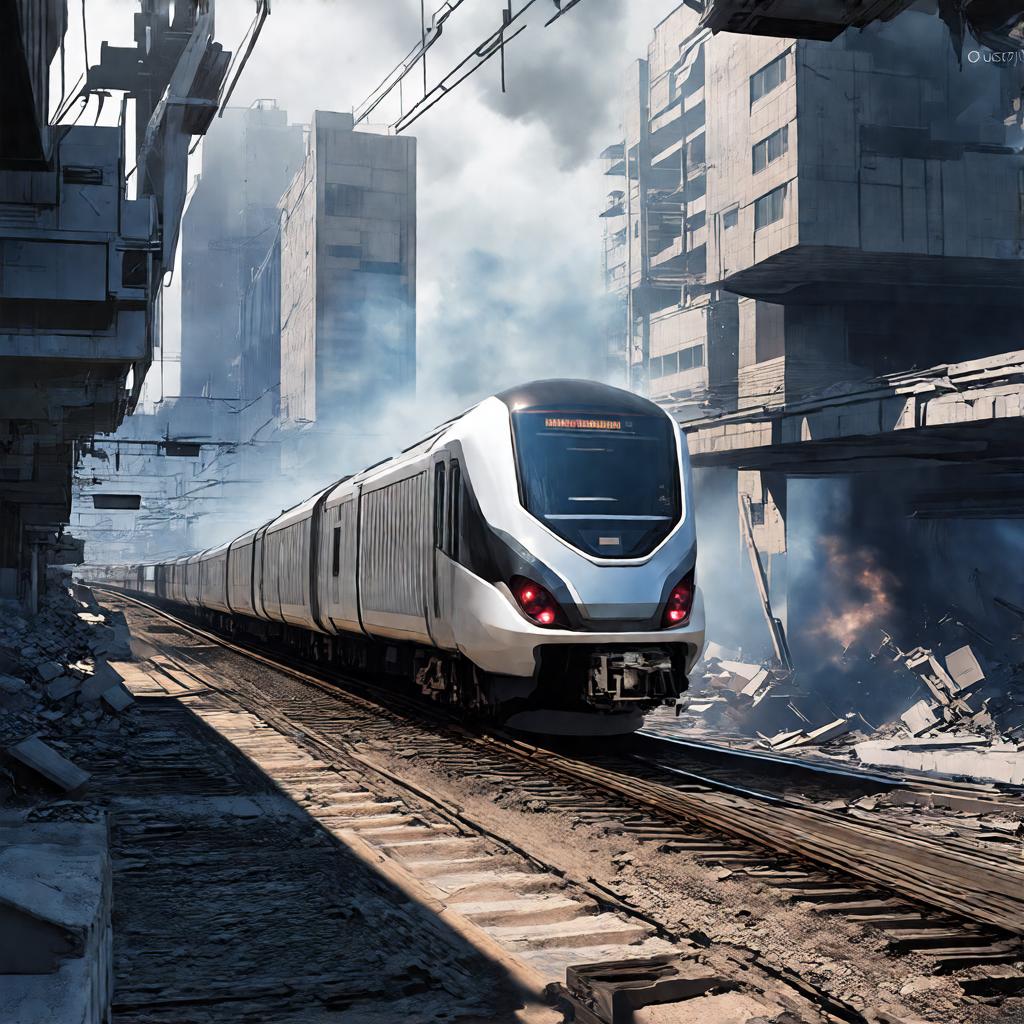}}
        \fbox{\includegraphics[width=\imgwidth]{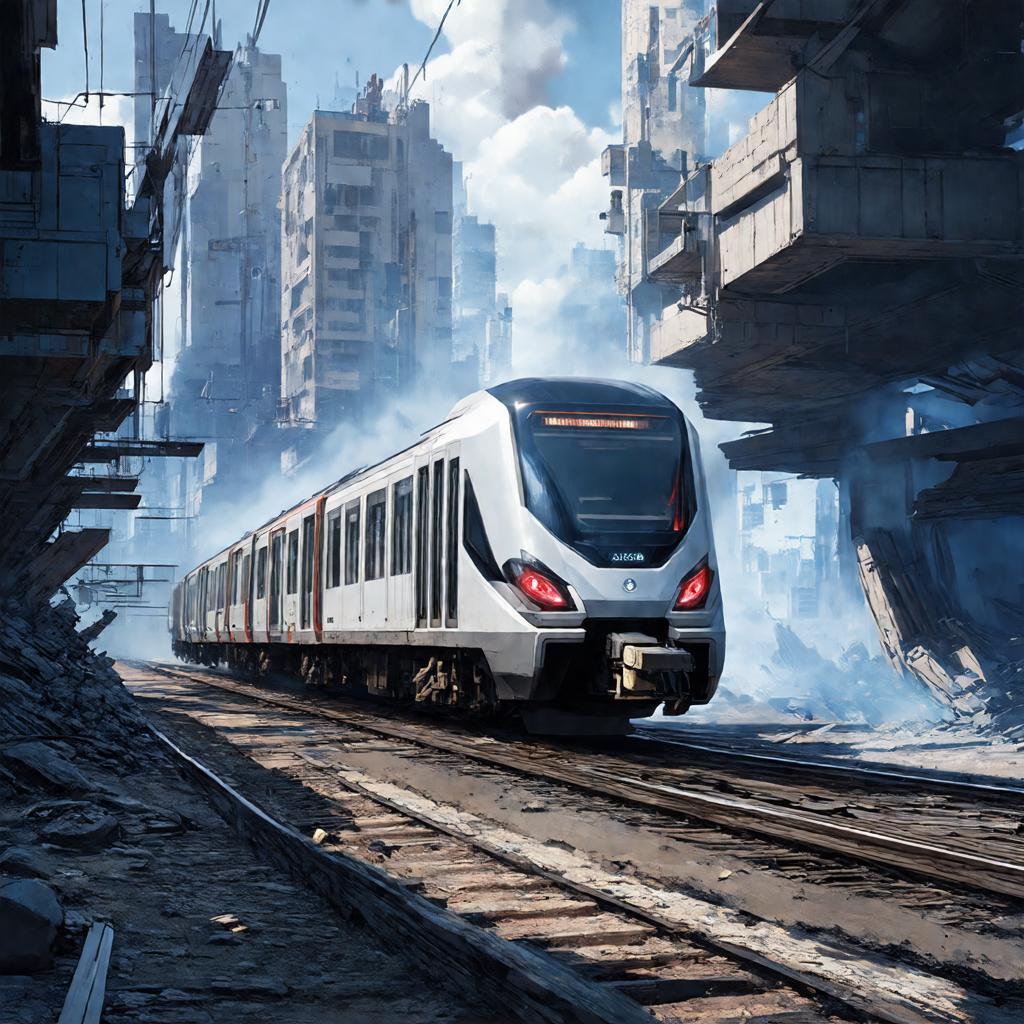}}
        \fbox{\includegraphics[width=\imgwidth]{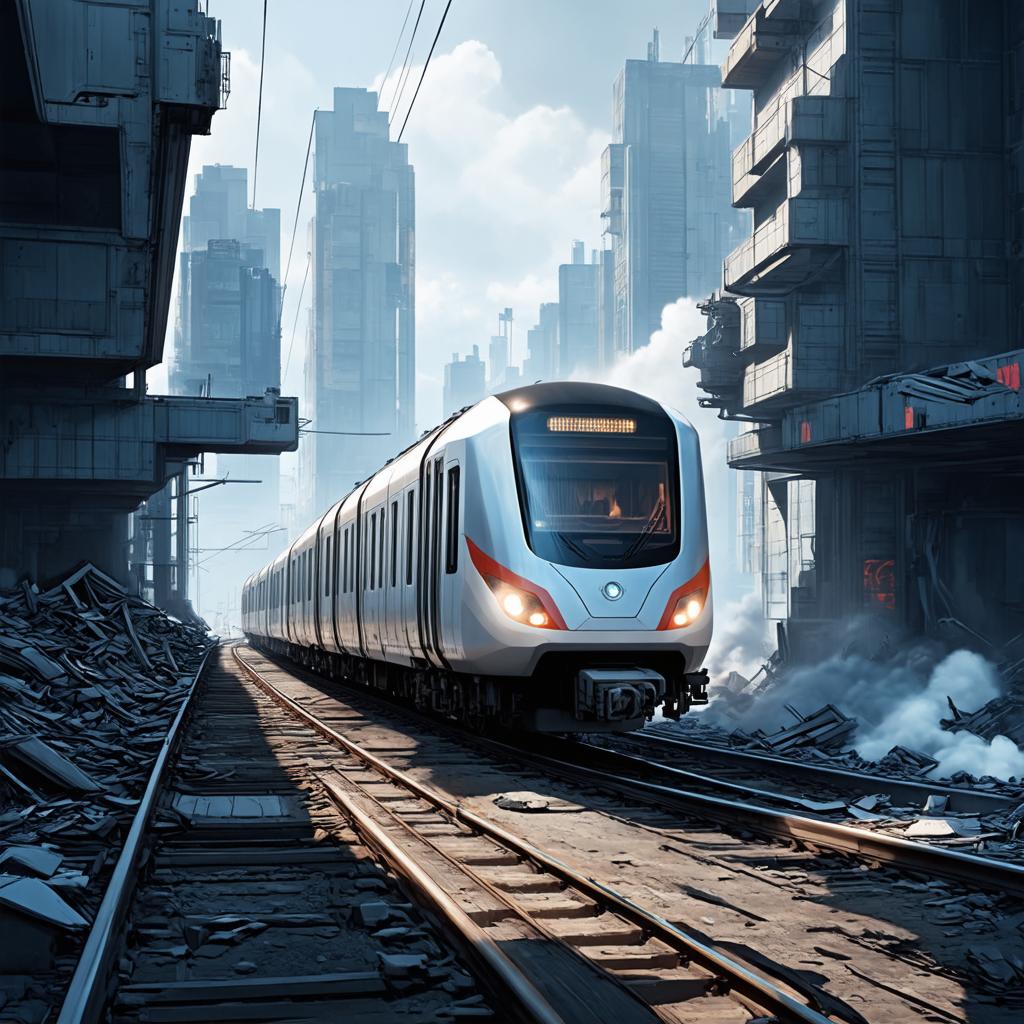}}\\[0.5ex]
        \vspace{-8pt}
        \caption*{
            \begin{minipage}{\capwidth}
                \centering
                \scriptsize{Prompt: \textit{a futuristic train travelling through a cyberpunk city that has considerable earthquake damage}}
            \end{minipage}
        }
    \end{minipage}%
    \hfill
    \begin{minipage}[t]{0.49\textwidth}
        \centering
        \fbox{\includegraphics[width=\imgwidth]{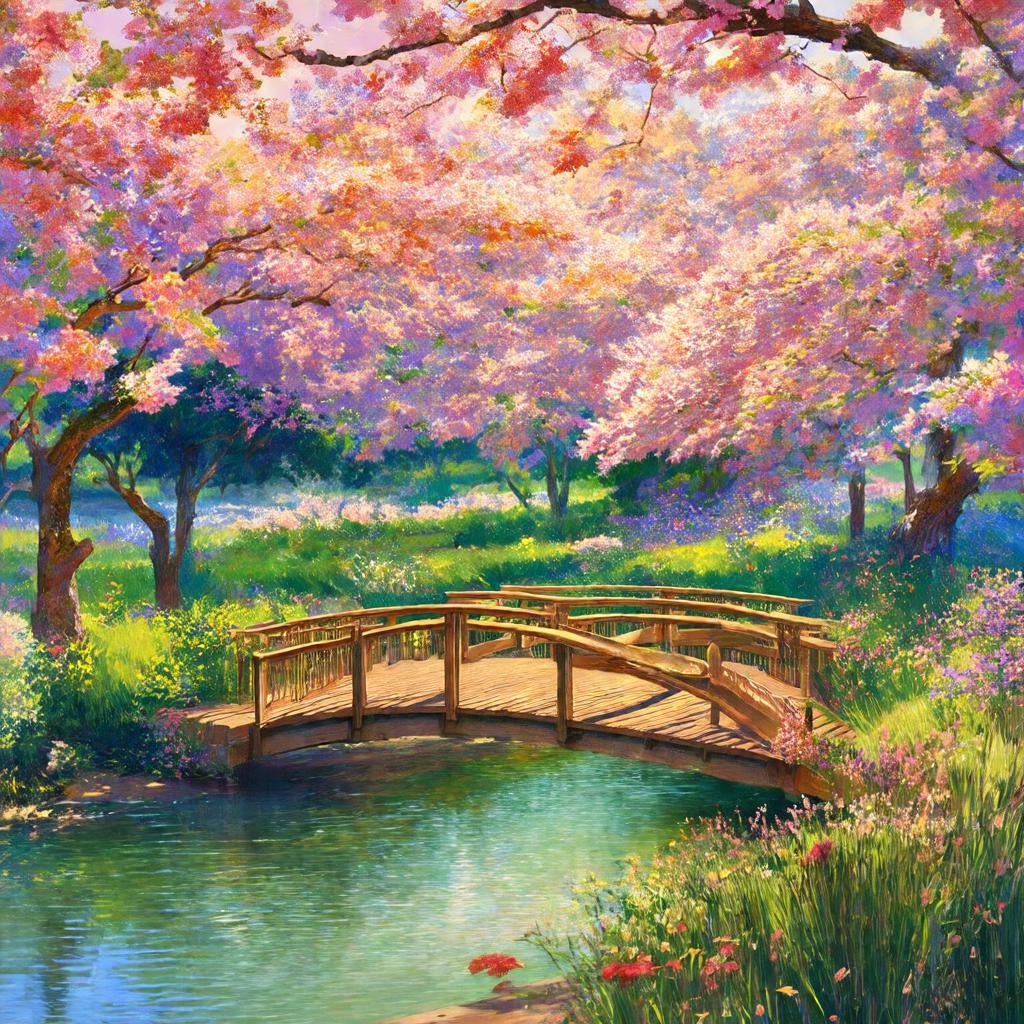}}
        \fbox{\includegraphics[width=\imgwidth]{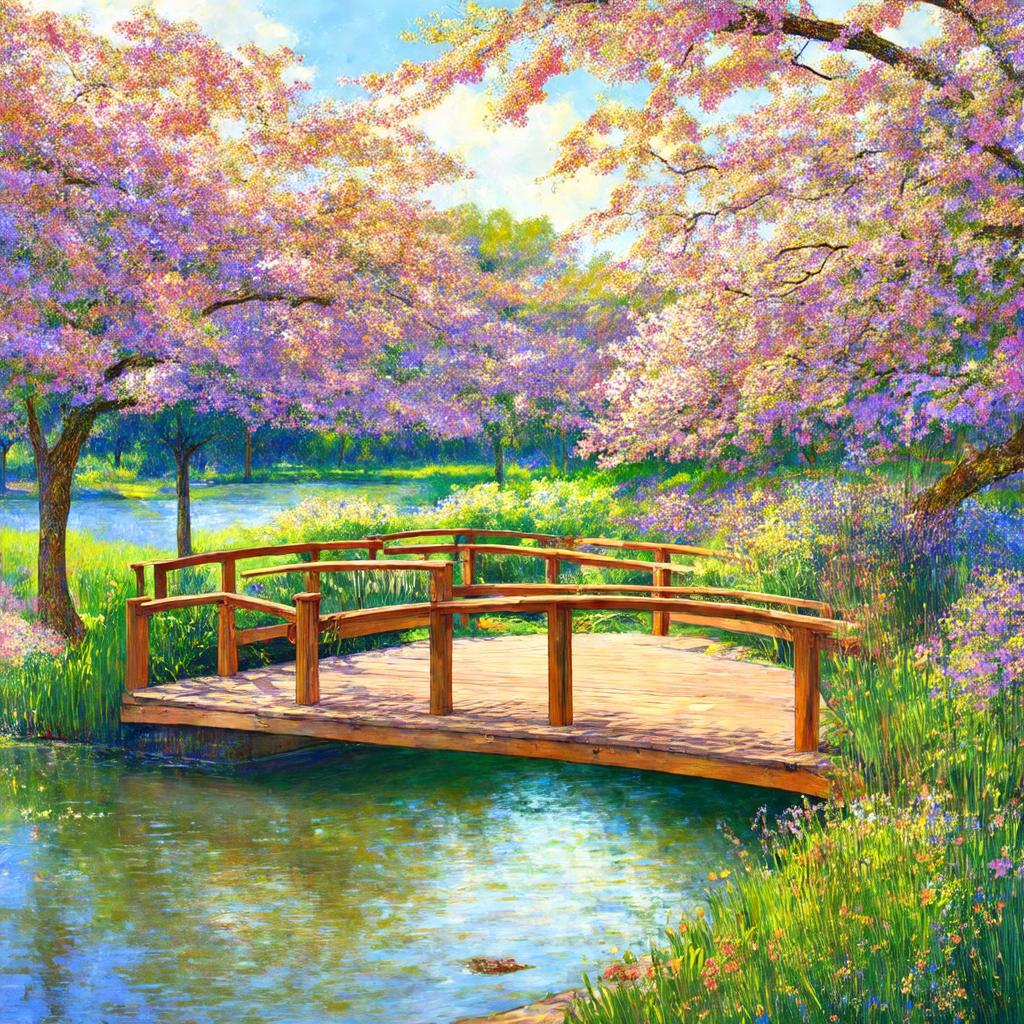}}
        \fbox{\includegraphics[width=\imgwidth]{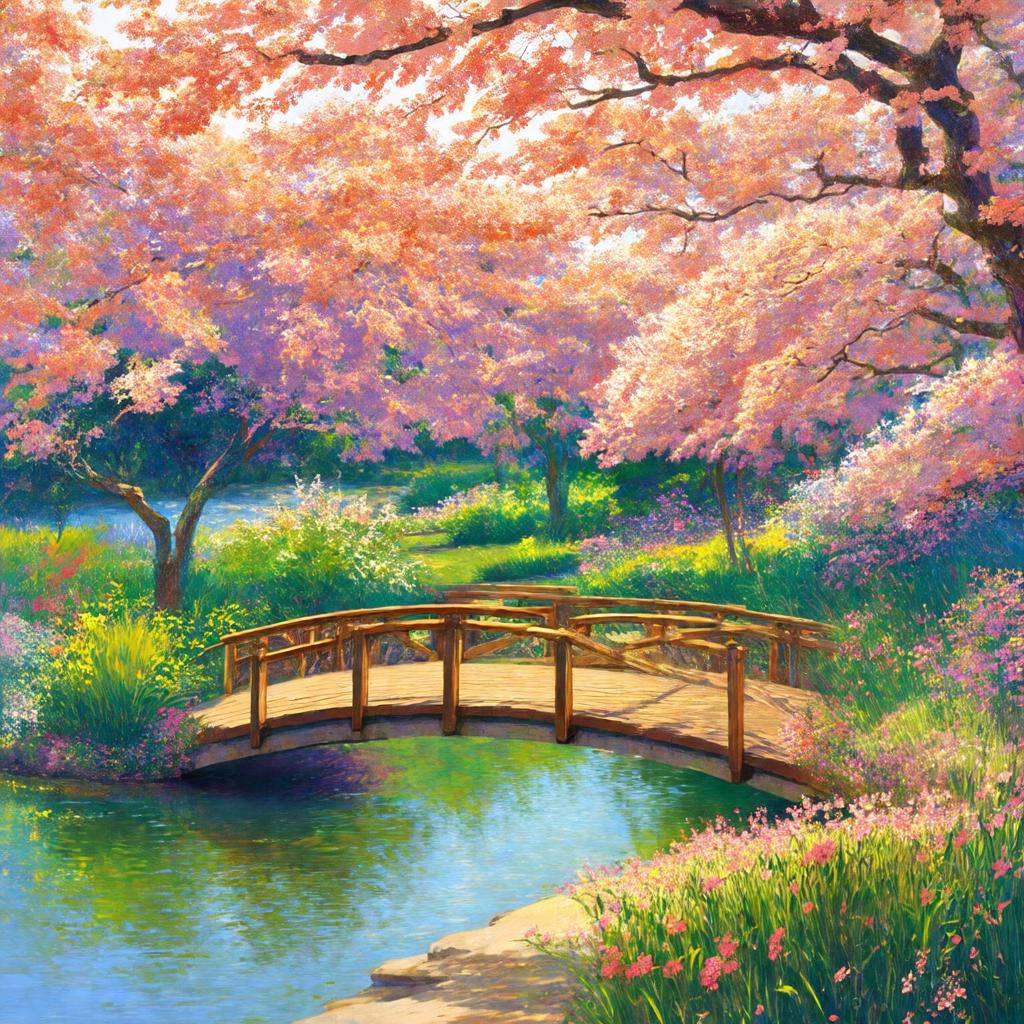}}
        \fbox{\includegraphics[width=\imgwidth]{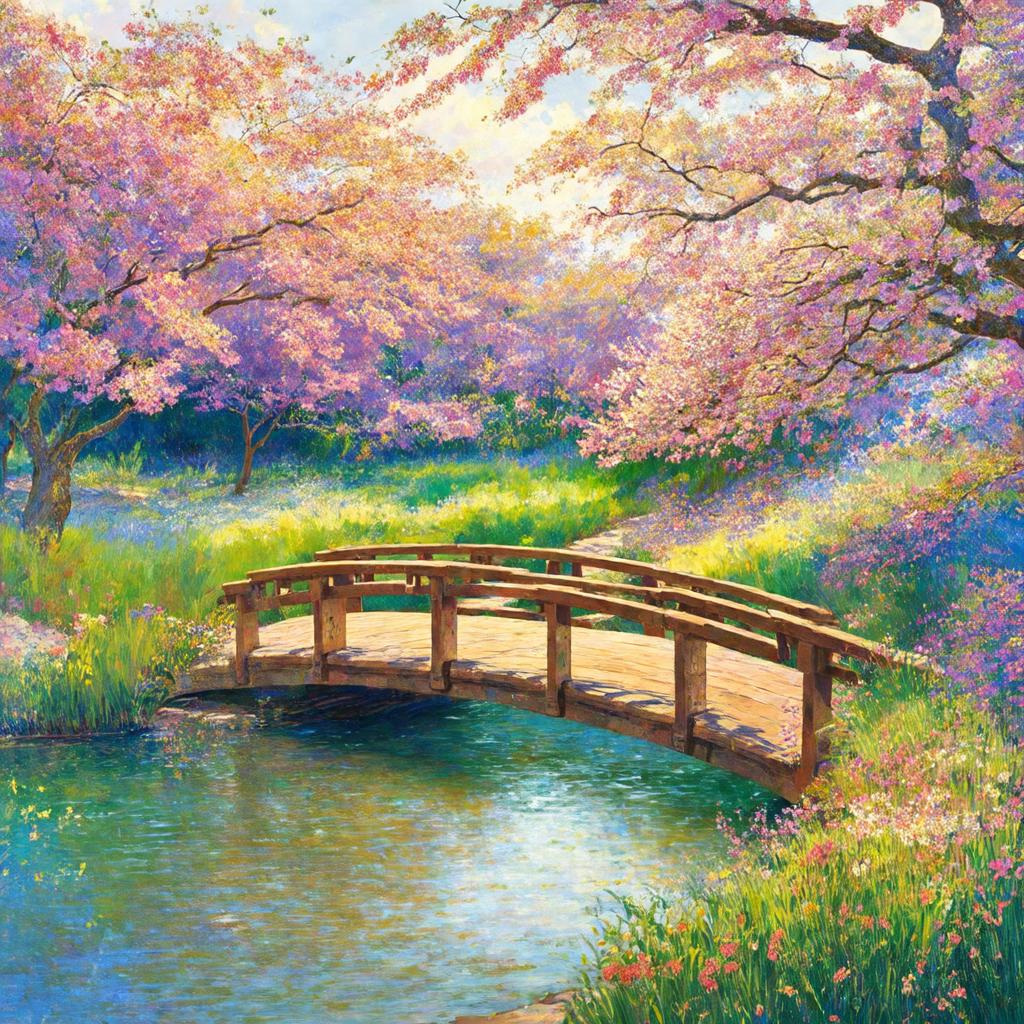}}
        \fbox{\includegraphics[width=\imgwidth]{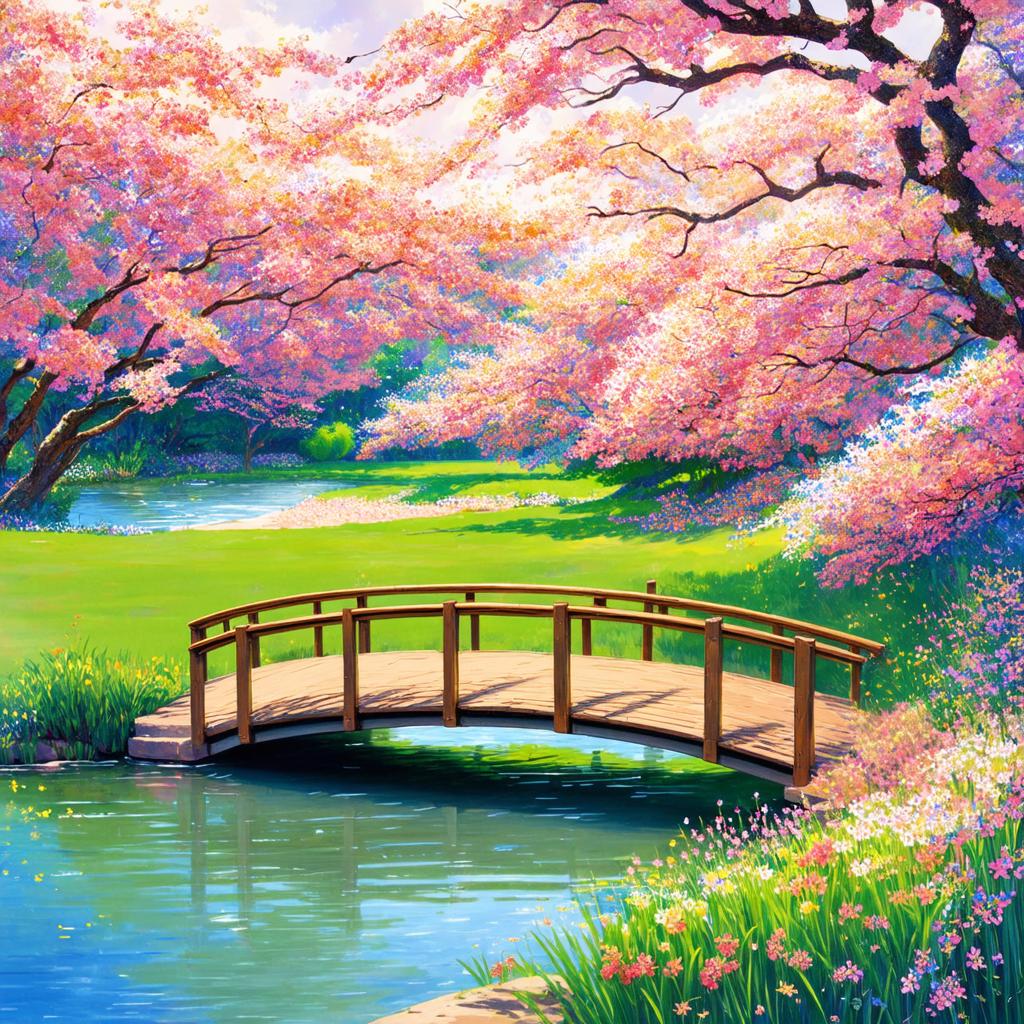}}\\[0.5ex]
        \vspace{-8pt}
        \caption*{
            \begin{minipage}{\capwidth}
                \centering
                \scriptsize{Prompt: \textit{a park with a beautiful wooden bridge over a pond, flowering trees around the banks, beautiful color correction, 16k, Alphonse Mucha style}}
            \end{minipage}
        }
    \end{minipage}

    \caption{
        Partial visualization of ablation results on the MJHQ dataset with SD3.5 and W8A8 DiT quantization. From left to right: baseline, \textsc{Seg.}, \textsc{Dual.}, \textsc{Seg.}+\textsc{Dual.} and FP16.
    }
    \vspace{-8pt}
    \label{fig:ablation_pics}
\end{figure*}

\begin{figure*}[htb!]
\centering
\includegraphics[width=\textwidth]{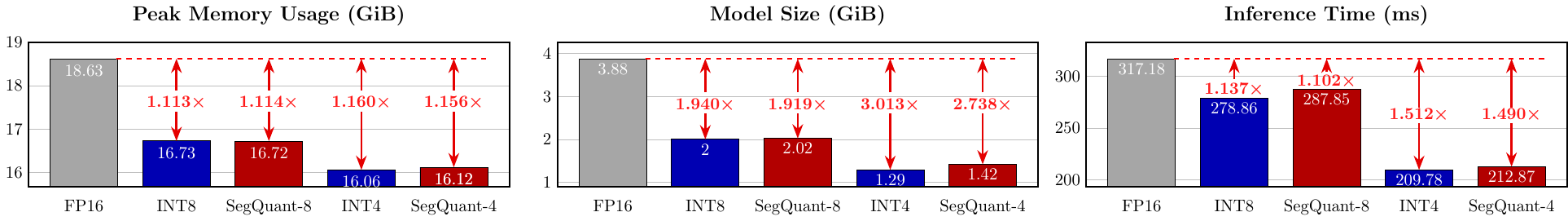}
\caption{Performance of quantization strategies on SD3.5 (RTX 4090). INT8 (W8A8) uses SmoothQuant; INT4 (W4A4) uses SVDQuant. Model size includes only the backbone; inference time is per-step (end-to-end).}
\vspace{-8pt}
\label{fig:time_memory}
\end{figure*}

\definecolor{lightgray}{gray}{0.93}
\begin{table}[htbp]
\centering
\caption{Main results across different backbones and models on the MJHQ-30K dataset.}
\resizebox{\columnwidth}{!}{%
\begin{tabular}{ccc
  T{3.2}  
  T{1.3}  
  T{1.3}  
  T{2.2}  
  T{1.3}  
}
\toprule

\multirow{2}{*}{\textbf{Backbone}} & \multirow{2}{*}{\textbf{W/A}} & \multirow{2}{*}{\textbf{Method}} & \multicolumn{2}{c}{\textbf{Quality}} & \multicolumn{3}{c}{\textbf{Similarity}} \\

\cmidrule(lr){4-5}
\cmidrule(lr){6-8}

& & &
\multicolumn{1}{c}{FID$\downarrow$} &
\multicolumn{1}{c}{IR$\uparrow$} &
\multicolumn{1}{c}{LPIPS$\downarrow$} &
\multicolumn{1}{c}{PSNR$\uparrow$} &
\multicolumn{1}{c}{SSIM$\uparrow$} \\
\midrule
\multirow{11}{*}{SD3.5-DiT} & FP16 & Baseline & 23.70 & 0.952 & {N/A} & {N/A} & {N/A}  \\
\cmidrule(lr){2-8}

 & \multirow{6}{*}{8/8(int)} &  PTQD     & 36.84 & 0.309 & 0.520 & 10.20 & 0.417  \\
& & PTQ4DiT   & 25.66 & 0.752 & 0.426 & 12.18 & 0.532\\ 
& & TAC       & 26.17 & 0.702 & 0.440 & 11.99 & 0.520\\ 
& & Smooth+   & 24.10 & 0.851 & 0.404 & 12.16 & 0.552\\ 
& & \cellcolor{lightgray}SegQuant-A & \cellcolor{lightgray}24.33 & \cellcolor{lightgray}\bfseries 0.924 & \cellcolor{lightgray}0.384 & \cellcolor{lightgray}12.78 & \cellcolor{lightgray}0.563 \\ 
& & \cellcolor{lightgray}SegQuant-G & \cellcolor{lightgray}\bfseries23.94 & \cellcolor{lightgray}0.859 & \cellcolor{lightgray}\bfseries 0.383 & \cellcolor{lightgray}\bfseries 12.83 & \cellcolor{lightgray}\bfseries 0.564 \\ 

\cmidrule(lr){2-8}

& \multirow{3}{*}{4/8(int)} & PTQ4DiT & 60.47 & -0.190 & 0.577 & 10.06 & 0.429 \\
& & SVDQuant & 27.95 & 0.725 & 0.456 & \bfseries11.76 & \bfseries0.523 \\
& & \cellcolor{lightgray}SegQuant-G & \cellcolor{lightgray}\bfseries27.30 & \cellcolor{lightgray}\bfseries0.762 & \cellcolor{lightgray}\bfseries0.453 & \cellcolor{lightgray}11.69 & \cellcolor{lightgray}0.521 \\

\midrule
\multirow{8}{*}{SDXL-UNet} & FP16 & Baseline & 17.10 & 0.910  & {N/A} & {N/A} & {N/A}\\
\cmidrule(lr){2-8}


 & \multirow{4}{*}{8/8(int)} & PTQ4DiT & 19.15 & 0.736 & 0.191 & 19.66 & 0.691 \\
& & Smooth+   & 18.97 & 0.746 & 0.161 & 20.63 & 0.719\\ 
& & \cellcolor{lightgray}SegQuant-A & \cellcolor{lightgray}18.90 & \cellcolor{lightgray}\bfseries0.775 & \cellcolor{lightgray}0.141 & \cellcolor{lightgray}21.33 & \cellcolor{lightgray}0.742 \\
& & \cellcolor{lightgray}SegQuant-G & \cellcolor{lightgray}\bfseries18.83 & \cellcolor{lightgray}0.764 & \cellcolor{lightgray}\bfseries0.134 & \cellcolor{lightgray}\bfseries21.60 & \cellcolor{lightgray}\bfseries0.750 \\

\cmidrule(lr){2-8}

& \multirow{3}{*}{8/8(fp)} & Q-Diffusion & 17.04 & 0.897 & 0.093 & 24.31 & 0.827  \\
& & \cellcolor{lightgray}SegQuant-A & \cellcolor{lightgray}17.06 & \cellcolor{lightgray}0.901 & \cellcolor{lightgray} 0.093 & \cellcolor{lightgray}24.28 & \cellcolor{lightgray}0.827  \\
& & \cellcolor{lightgray}SegQuant-G & \cellcolor{lightgray}\bfseries 17.03 & \cellcolor{lightgray}\bfseries 0.903 & \cellcolor{lightgray}\bfseries0.082 & \cellcolor{lightgray}\bfseries 24.84 & \cellcolor{lightgray}\bfseries 0.838  \\

\midrule

\multirow{8}{*}{FLUX-DiT} & BF16 & Baseline & 23.21 & 0.837  & {N/A} & {N/A} & {N/A} \\
\cmidrule(lr){2-8}
 & \multirow{5}{*}{8/8(int)} & Q-Diffusion & 23.99 & 0.732 & 0.299 & 15.87 & 0.633 \\
& & PTQ4DiT & 27.34 & 0.630 & 0.325 & 15.36 & 0.611 \\
& & Smooth+   & 26.21 & 0.684 & 0.539 & 15.56 & 0.620\\ 
& & \cellcolor{lightgray}SegQuant-A & \cellcolor{lightgray}\bfseries22.85 & \cellcolor{lightgray}\bfseries0.835 & \cellcolor{lightgray}0.150 & \cellcolor{lightgray}19.84 & \cellcolor{lightgray}0.770 \\
& & \cellcolor{lightgray}SegQuant-G & \cellcolor{lightgray}23.07 & \cellcolor{lightgray}0.822 & \cellcolor{lightgray}\bfseries0.138 & \cellcolor{lightgray}\bfseries20.32 & \cellcolor{lightgray}\bfseries0.782 \\

\cmidrule(lr){2-8}

& \multirow{3}{*}{4/8(int)} & PTQ4DiT & 41.02 & -0.039 & 0.540 & 12.09 & 0.540 \\
& & SVDQuant & 23.61 & 0.783 & 0.232 & 17.29 & 0.697 \\
& & \cellcolor{lightgray}SegQuant-G & \cellcolor{lightgray}\bfseries23.45 & \cellcolor{lightgray}\bfseries0.789 & \cellcolor{lightgray}\bfseries0.225 & \cellcolor{lightgray}\bfseries17.48 & \cellcolor{lightgray}\bfseries0.702 \\
\bottomrule
\end{tabular}
}
\label{tab:main_expr}
\vspace{-4mm}
\end{table}


We evaluate SegQuant on the MJHQ dataset across various models and precision levels.
As shown in Table~\ref{tab:main_expr}, it consistently achieves better image quality and higher fidelity to the original model.
Additional visualizations are presented in Fig.~\ref{fig:main_pics}, with more results in Appendix \ref{appendix:exp-detail}.
We also analyze efficiency in Fig.~\ref{fig:time_memory}, showing comparable memory usage to naive quantization and a modest runtime increase from segmentation and dual-scale steps, which bring notable quality gains.




\subsection{Ablation Study}

We ablate SegQuant's two components: semantic-aware segmentation (\textit{SegLinear}) and adaptive scaling (\textit{DualScale}). 
Table~\ref{tab:opt_cal_stories} shows \textit{SegLinear} consistently reduces layer-wise Frobenius error across optimizers (e.g., GPTQ error on \texttt{DiT.11.norm1\_context} drops from 3.02 to 1.76 in SD3.5). 
Furthermore, Table~\ref{tab:ablation_seg} indicates both components independently improve MJHQ-30K generation quality. Their combination performs best, improving FID (23.35$\to$22.54) and Image Reward (0.877$\to$0.952), proving they complementarily enhance quantization robustness.

\begin{table}[h!]
\centering
\caption{Frobenius norm of quantization error for single linear layers in SD3.5, comparing SmoothQuant (W8A8, tuned $\alpha$), SVDQuant (W4A8, per-channel), and GPTQ (fixed $\alpha$=0.5).}
\resizebox{\columnwidth}{!}{%
\begin{tabular}{lccc}
\toprule
\multirow{2}{*}{\textbf{Layer Name}} & \multirow{2}{*}{\textbf{Method}} & \multicolumn{2}{c}{\textbf{F-norm}} \\
\cmidrule(lr){3-4}
& & \text{w/o}~\textsc{Seg.} & \text{w/}~\textsc{Seg.} \\
\midrule
\texttt{DiT.0.norm1} & \multirow{2}{*}{\textsc{SmoothQuant}} & 0.7041 & \textbf{0.5381} \\
\texttt{DiT.11.attn.out} & & 2273.3 & \textbf{1879.3} \\
\texttt{DiT.0.norm1} & \multirow{2}{*}{\textsc{SVDQuant}} & 0.0861 & \textbf{0.0714} \\
\texttt{DiT.11.attn.out} & & 2031.6 & \textbf{1810.7} \\
\midrule
\texttt{DiT.0.norm1} & \multirow{4}{*}{\textsc{GPTQ}} & 0.8350 & \textbf{0.4441} \\
\texttt{DiT.0.norm1\_context} & & 1.5166 & \textbf{0.7441} \\
\texttt{DiT.11.norm1} & & 1.1719 & \textbf{0.9419} \\
\texttt{DiT.11.norm1\_context} & & 3.0176 & \textbf{1.7637} \\
\bottomrule
\end{tabular}
}
\label{tab:opt_cal_stories}
\end{table}

\begin{table}[h!]
\centering
\caption{
Ablation study on MJHQ-30K with SD3.5 and W8A8 DiT quantization (32 images calibrating). 
Using SmoothQuant as our optimizer (fixed $\alpha$=0.5) and AMax.}
\resizebox{\columnwidth}{!}{%
\begin{tabular}{lccccc}
\toprule

\multirow{2}{*}{\textbf{Method}} & \multicolumn{2}{c}{\textbf{Quality}} & \multicolumn{3}{c}{\textbf{Similarity}} \\

\cmidrule(lr){2-3}
\cmidrule(lr){4-6}

& FID$\downarrow$ & IR$\uparrow$ & LPIPS$\downarrow$ & PSNR$\uparrow$ & SSIM$\uparrow$ \\

\midrule
Baseline & 23.35 & 0.877 & 0.419 & 11.93 & 0.536 \\
+\textsc{Seg.} & 23.36 & 0.899 & \underline{0.395} & 12.03 & \underline{0.554} \\
+\textsc{Dual.} & \underline{22.61} & \underline{0.909} & 0.401 & \underline{12.14} & 0.551 \\
+\textsc{Seg.}+\textsc{Dual.} & \textbf{22.54} & \textbf{0.952} & \textbf{0.377} & \textbf{12.50} & \textbf{0.567} \\
\bottomrule
\end{tabular}
}
\label{tab:ablation_seg}
\end{table}
\section{Conclusion}

We propose SegQuant, a semantics-aware quantization framework leveraging feature segmentation and polarity-sensitive scaling. Evaluations show it significantly enhances image quality in DiT-based diffusion models and generalizes well to other architectures.

\section*{Acknowledgments}

This work was supported in part by the National Science Foundation of China (62502441, 62125206) and the Major Program of the National Natural Science Foundation of Zhejiang (LD25F020002). Hailiang Zhao’s work was supported in part by the Zhejiang University Education Foundation Qizhen Scholar Foundation.
{
    \small
    \bibliographystyle{ieeenat_fullname}
    \bibliography{main}

@String(CVPR= {IEEE Conf. Comput. Vis. Pattern Recog.})

@String(ICCV= {Int. Conf. Comput. Vis.})

@String(ECCV= {Eur. Conf. Comput. Vis.})

@String(NIPS= {Adv. Neural Inform. Process. Syst.})

@String(AAAI = {AAAI})

@String(CVPR  = {CVPR})

@String(ICCV  = {ICCV})

@String(ECCV  = {ECCV})

@String(NIPS  = {NeurIPS})

@inbook{quant2021,
  author    = {Amir Gholami and Sehoon Kim and Zhen Dong and Zhewei Yao and Michael W. Mahoney and Kurt Keutzer},
  title     = {A Survey of Quantization Methods for Efficient Neural Network Inference},
  booktitle = {Low-Power Computer Vision},
  publisher = {Chapman and Hall/CRC},
  year      = {2022},
  chapter   = {13},
  doi       = {10.1201/9781003162810-13}
}

@InProceedings{ptq2018CVPR,
author = {Jacob, Benoit and Kligys, Skirmantas and Chen, Bo and Zhu, Menglong and Tang, Matthew and Howard, Andrew and Adam, Hartwig and Kalenichenko, Dmitry},
title = {Quantization and Training of Neural Networks for Efficient Integer-Arithmetic-Only Inference},
booktitle = {Proceedings of the IEEE Conference on Computer Vision and Pattern Recognition (CVPR)},
month = {June},
year = {2018}
}

@inproceedings{ddpm2020,
author = {Ho, Jonathan and Jain, Ajay and Abbeel, Pieter},
title = {Denoising diffusion probabilistic models},
year = {2020},
isbn = {9781713829546},
publisher = {Curran Associates Inc.},
address = {Red Hook, NY, USA},
booktitle = {Proceedings of the 34th International Conference on Neural Information Processing Systems},
articleno = {574},
numpages = {12},
location = {Vancouver, BC, Canada},
series = {NIPS '20}
}

@INPROCEEDINGS{dit2023,
  author={Peebles, William and Xie, Saining},
  booktitle={2023 IEEE/CVF International Conference on Computer Vision (ICCV)}, 
  title={Scalable Diffusion Models with Transformers}, 
  year={2023},
  volume={},
  number={},
  pages={4172-4182},
  doi={10.1109/ICCV51070.2023.00387}}

@InProceedings{unet2015,
author="Ronneberger, Olaf
and Fischer, Philipp
and Brox, Thomas",
editor="Navab, Nassir
and Hornegger, Joachim
and Wells, William M.
and Frangi, Alejandro F.",
title="U-Net: Convolutional Networks for Biomedical Image Segmentation",
booktitle="Medical Image Computing and Computer-Assisted Intervention -- MICCAI 2015",
year="2015",
publisher="Springer International Publishing",
address="Cham",
pages="234--241",
isbn="978-3-319-24574-4"
}

@inproceedings{adanorm2018,
author = {Perez, Ethan and Strub, Florian and de Vries, Harm and Dumoulin, Vincent and Courville, Aaron},
title = {FiLM: visual reasoning with a general conditioning layer},
year = {2018},
isbn = {978-1-57735-800-8},
publisher = {AAAI Press},
booktitle = {Proceedings of the Thirty-Second AAAI Conference on Artificial Intelligence and Thirtieth Innovative Applications of Artificial Intelligence Conference and Eighth AAAI Symposium on Educational Advances in Artificial Intelligence},
articleno = {483},
numpages = {10},
location = {New Orleans, Louisiana, USA},
series = {AAAI'18/IAAI'18/EAAI'18}
}

@misc{vae2022,
      title={Auto-Encoding Variational Bayes}, 
      author={Diederik P Kingma and Max Welling},
      year={2022},
      eprint={1312.6114},
      archivePrefix={arXiv},
      primaryClass={stat.ML},
      url={https://arxiv.org/abs/1312.6114}, 
}

@InProceedings{smoothquant2023,
    title = {{S}mooth{Q}uant: Accurate and Efficient Post-Training Quantization for Large Language Models},
    author = {Xiao, Guangxuan and Lin, Ji and Seznec, Mickael and Wu, Hao and Demouth, Julien and Han, Song},
    booktitle = {Proceedings of the 40th International Conference on Machine Learning},
    year = {2023}
}

@inproceedings{
  svdquant2025,
  title={SVDQuant: Absorbing Outliers by Low-Rank Components for 4-Bit Diffusion Models},
  author={Li*, Muyang and Lin*, Yujun and Zhang*, Zhekai and Cai, Tianle and Li, Xiuyu and Guo, Junxian and Xie, Enze and Meng, Chenlin and Zhu, Jun-Yan and Han, Song},
  booktitle={The Thirteenth International Conference on Learning Representations},
  year={2025}
}

@inproceedings{tvm2018,
author = {Chen, Tianqi and Moreau, Thierry and Jiang, Ziheng and Zheng, Lianmin and Yan, Eddie and Cowan, Meghan and Shen, Haichen and Wang, Leyuan and Hu, Yuwei and Ceze, Luis and Guestrin, Carlos and Krishnamurthy, Arvind},
title = {TVM: an automated end-to-end optimizing compiler for deep learning},
year = {2018},
isbn = {9781931971478},
publisher = {USENIX Association},
address = {USA},
booktitle = {Proceedings of the 13th USENIX Conference on Operating Systems Design and Implementation},
pages = {579–594},
numpages = {16},
location = {Carlsbad, CA, USA},
series = {OSDI'18}
}

@inproceedings {magpy2024,
author = {Chen Zhang and Rongchao Dong and Haojie Wang and Runxin Zhong and Jike Chen and Jidong Zhai},
title = {{MAGPY}: Compiling Eager Mode {DNN} Programs by Monitoring Execution States},
booktitle = {2024 USENIX Annual Technical Conference (USENIX ATC 24)},
year = {2024},
isbn = {978-1-939133-41-0},
address = {Santa Clara, CA},
pages = {683--698},
url = {https://www.usenix.org/conference/atc24/presentation/zhang-chen},
publisher = {USENIX Association},
month = jul
}

@inproceedings{torch2019,
author = {Paszke, Adam and Gross, Sam and Massa, Francisco and Lerer, Adam and Bradbury, James and Chanan, Gregory and Killeen, Trevor and Lin, Zeming and Gimelshein, Natalia and Antiga, Luca and Desmaison, Alban and K\"{o}pf, Andreas and Yang, Edward and DeVito, Zach and Raison, Martin and Tejani, Alykhan and Chilamkurthy, Sasank and Steiner, Benoit and Fang, Lu and Bai, Junjie and Chintala, Soumith},
title = {PyTorch: an imperative style, high-performance deep learning library},
year = {2019},
booktitle = {Proceedings of the 33rd International Conference on Neural Information Processing Systems},
articleno = {721},
pages = {8026--8037},
publisher = {Curran Associates Inc.},
address = {Red Hook, NY, USA}
}

@INPROCEEDINGS{li2023qdiffusionquantizingdiffusionmodels,
  author={Li, Xiuyu and Liu, Yijiang and Lian, Long and Yang, Huanrui and Dong, Zhen and Kang, Daniel and Zhang, Shanghang and Keutzer, Kurt},
  booktitle={2023 IEEE/CVF International Conference on Computer Vision (ICCV)}, 
  title={Q-Diffusion: Quantizing Diffusion Models}, 
  year={2023},
  volume={},
  number={},
  pages={17489-17499},
  doi={10.1109/ICCV51070.2023.01608}}

@misc{modelopt,
  title = {Model Optimizer},
  author = "{NVIDIA Corporation}",
  howpublished = {\url{https://nvidia.github.io/TensorRT-Model-Optimizer}},
  note = {Accessed: 2025-05-03},
  year = {2023}
}

@inproceedings{esser2024scalingrectifiedflowtransformers,
author = {Esser, Patrick and Kulal, Sumith and Blattmann, Andreas and Entezari, Rahim and M\"{u}ller, Jonas and Saini, Harry and Levi, Yam and Lorenz, Dominik and Sauer, Axel and Boesel, Frederic and Podell, Dustin and Dockhorn, Tim and English, Zion and Rombach, Robin},
title = {Scaling rectified flow transformers for high-resolution image synthesis},
year = {2024},
publisher = {JMLR.org},
booktitle = {Proceedings of the 41st International Conference on Machine Learning},
articleno = {503},
numpages = {28},
location = {Vienna, Austria},
series = {ICML'24}
}

@misc{flux2024,
  author       = {{Black Forest Labs}},
  title        = {{Flux.1}},
  year         = {2024},
  url          = {https://blackforestlabs.ai/},
  note         = {Accessed: 2025-05-05}
}

@inproceedings{zju2023ptqd,
author = {He, Yefei and Liu, Luping and Liu, Jing and Wu, Weijia and Zhou, Hong and Zhuang, Bohan},
title = {PTQD: accurate post-training quantization for diffusion models},
year = {2023},
publisher = {Curran Associates Inc.},
address = {Red Hook, NY, USA},
booktitle = {Proceedings of the 37th International Conference on Neural Information Processing Systems},
articleno = {580},
numpages = {13},
location = {New Orleans, LA, USA},
series = {NIPS '23}
}

@InProceedings{tfmq2024,
    author    = {Huang, Yushi and Gong, Ruihao and Liu, Jing and Chen, Tianlong and Liu, Xianglong},
    title     = {TFMQ-DM: Temporal Feature Maintenance Quantization for Diffusion Models},
    booktitle = {Proceedings of the IEEE/CVF Conference on Computer Vision and Pattern Recognition (CVPR)},
    month     = {June},
    year      = {2024},
    pages     = {7362-7371}
}

@inproceedings{wu2024ptq4dit,
  title={PTQ4DiT: Post-training Quantization for Diffusion Transformers},
  author={Wu, Junyi and Wang, Haoxuan and Shang, Yuzhang and Shah, Mubarak and Yan, Yan},
  booktitle={NeurIPS},
  year={2024}
}

@inproceedings{
viditq2024,
title={ViDiT-Q: Efficient and Accurate Quantization of Diffusion Transformers for Image and Video Generation},
author={Tianchen Zhao and Tongcheng Fang and Haofeng Huang and Rui Wan and Widyadewi Soedarmadji and Enshu Liu and Shiyao Li and Zinan Lin and Guohao Dai and Shengen Yan and Huazhong Yang and Xuefei Ning and Yu Wang},
booktitle={The Thirteenth International Conference on Learning Representations},
year={2025},
url={https://openreview.net/forum?id=E1N1oxd63b}
}

@inproceedings{yao2024timestepaware,
author = {Yao, Yuzhe and Tian, Feng and Chen, Jun and Lin, Haonan and Dai, Guang and Liu, Yong and Wang, Jingdong},
title = {Timestep-Aware Correction for Quantized Diffusion Models},
year = {2024},
isbn = {978-3-031-72847-1},
publisher = {Springer-Verlag},
address = {Berlin, Heidelberg},
url = {https://doi.org/10.1007/978-3-031-72848-8_13},
doi = {10.1007/978-3-031-72848-8_13},
booktitle = {Computer Vision – ECCV 2024: 18th European Conference, Milan, Italy, September 29–October 4, 2024, Proceedings, Part LXVI},
pages = {215–232},
numpages = {18},
location = {Milan, Italy}
}

@InProceedings{lin2015microsoft,
author="Lin, Tsung-Yi
and Maire, Michael
and Belongie, Serge
and Hays, James
and Perona, Pietro
and Ramanan, Deva
and Doll{\'a}r, Piotr
and Zitnick, C. Lawrence",
editor="Fleet, David
and Pajdla, Tomas
and Schiele, Bernt
and Tuytelaars, Tinne",
title="Microsoft COCO: Common Objects in Context",
booktitle="Computer Vision -- ECCV 2014",
year="2014",
publisher="Springer International Publishing",
address="Cham",
pages="740--755",
isbn="978-3-319-10602-1"
}

@misc{mjhq30k2024,
      title={Playground v2.5: Three Insights towards Enhancing Aesthetic Quality in Text-to-Image Generation}, 
      author={Daiqing Li and Aleks Kamko and Ehsan Akhgari and Ali Sabet and Linmiao Xu and Suhail Doshi},
      year={2024},
      eprint={2402.17245},
      archivePrefix={arXiv},
      primaryClass={cs.CV}
}

@InProceedings{dci2024,
    author    = {Urbanek, Jack and Bordes, Florian and Astolfi, Pietro and Williamson, Mary and Sharma, Vasu and Romero-Soriano, Adriana},
    title     = {A Picture is Worth More Than 77 Text Tokens: Evaluating CLIP-Style Models on Dense Captions},
    booktitle = {Proceedings of the IEEE/CVF Conference on Computer Vision and Pattern Recognition (CVPR)},
    month     = {June},
    year      = {2024},
    pages     = {26700-26709}
}

@inproceedings{
flowmatching2023,
title={Flow Matching for Generative Modeling},
author={Yaron Lipman and Ricky T. Q. Chen and Heli Ben-Hamu and Maximilian Nickel and Matthew Le},
booktitle={The Eleventh International Conference on Learning Representations },
year={2023},
url={https://openreview.net/forum?id=PqvMRDCJT9t}
}

@inproceedings{
gptq2022,
title={{OPTQ}: Accurate Quantization for Generative Pre-trained Transformers},
author={Elias Frantar and Saleh Ashkboos and Torsten Hoefler and Dan Alistarh},
booktitle={The Eleventh International Conference on Learning Representations },
year={2023},
url={https://openreview.net/forum?id=tcbBPnfwxS}
}

@inproceedings{lpips2018,
  title={The Unreasonable Effectiveness of Deep Features as a Perceptual Metric},
  author={Zhang, Richard and Isola, Phillip and Efros, Alexei A and Shechtman, Eli and Wang, Oliver},
  booktitle={CVPR},
  year={2018}
}

@inproceedings{imagereward2023,
  title={ImageReward: learning and evaluating human preferences for text-to-image generation},
  author={Xu, Jiazheng and Liu, Xiao and Wu, Yuchen and Tong, Yuxuan and Li, Qinkai and Ding, Ming and Tang, Jie and Dong, Yuxiao},
  booktitle={Proceedings of the 37th International Conference on Neural Information Processing Systems},
  pages={15903--15935},
  year={2023}
}

@inproceedings{fid2017,
author = {Heusel, Martin and Ramsauer, Hubert and Unterthiner, Thomas and Nessler, Bernhard and Hochreiter, Sepp},
title = {GANs trained by a two time-scale update rule converge to a local nash equilibrium},
year = {2017},
isbn = {9781510860964},
publisher = {Curran Associates Inc.},
address = {Red Hook, NY, USA},
booktitle = {Proceedings of the 31st International Conference on Neural Information Processing Systems},
pages = {6629–6640},
numpages = {12},
location = {Long Beach, California, USA},
series = {NIPS'17}
}

@inproceedings{clipscore2021,
  title={{CLIPScore:} A Reference-free Evaluation Metric for Image Captioning},
  author={Hessel, Jack and Holtzman, Ari and Forbes, Maxwell and Bras, Ronan Le and Choi, Yejin},
  booktitle={EMNLP},
  year={2021}
}

@ARTICLE{ssim2004,
  author={Zhou Wang and Bovik, A.C. and Sheikh, H.R. and Simoncelli, E.P.},
  journal={IEEE Transactions on Image Processing}, 
  title={Image quality assessment: from error visibility to structural similarity}, 
  year={2004},
  volume={13},
  number={4},
  pages={600-612},
  doi={10.1109/TIP.2003.819861}}

@inproceedings{adaround2020,
author = {Nagel, Markus and Amjad, Rana Ali and Van Baalen, Mart and Louizos, Christos and Blankevoort, Tijmen},
title = {Up or down? adaptive rounding for post-training quantization},
year = {2020},
publisher = {JMLR.org},
booktitle = {Proceedings of the 37th International Conference on Machine Learning},
articleno = {667},
numpages = {10},
series = {ICML'20}
}

@inproceedings{
brecq2021,
title={{\{}BRECQ{\}}: Pushing the Limit of Post-Training Quantization by Block Reconstruction},
author={Yuhang Li and Ruihao Gong and Xu Tan and Yang Yang and Peng Hu and Qi Zhang and Fengwei Yu and Wei Wang and Shi Gu},
booktitle={International Conference on Learning Representations},
year={2021},
url={https://openreview.net/forum?id=POWv6hDd9XH}
}

@inproceedings{
sdxl2023,
title={{SDXL}: Improving Latent Diffusion Models for High-Resolution Image Synthesis},
author={Dustin Podell and Zion English and Kyle Lacey and Andreas Blattmann and Tim Dockhorn and Jonas M{\"u}ller and Joe Penna and Robin Rombach},
booktitle={The Twelfth International Conference on Learning Representations},
year={2024},
url={https://openreview.net/forum?id=di52zR8xgf}
}

@inproceedings{iqa2022,
    author = {Wang, Jianyi and Chan, Kelvin CK and Loy, Chen Change},
    title = {Exploring CLIP for Assessing the Look and Feel of Images},
    booktitle = {AAAI},
    year = {2023}
}

@article{edadm2024,
  title={Enhanced distribution alignment for post-training quantization of diffusion models},
  author={Liu, Xuewen and Li, Zhikai and Xiao, Junrui and Gu, Qingyi},
  journal={arXiv e-prints},
  pages={arXiv--2401},
  year={2024}
}

@misc{qvdit2024,
      title={Q-VDiT: Towards Accurate Quantization and Distillation of Video-Generation Diffusion Transformers}, 
      author={Weilun Feng and Chuanguang Yang and Haotong Qin and Xiangqi Li and Yu Wang and Zhulin An and Libo Huang and Boyu Diao and Zixiang Zhao and Yongjun Xu and Michele Magno},
      year={2025},
      eprint={2505.22167},
      archivePrefix={arXiv},
      primaryClass={cs.CV},
      url={https://arxiv.org/abs/2505.22167}, 
}

@misc{ptq4vit2023,
      title={PTQ4ViT: Post-training quantization for vision transformers with twin uniform quantization}, 
      author={Zhihang Yuan and Chenhao Xue and Yiqi Chen and Qiang Wu and Guangyu Sun},
      year={2024},
      eprint={2111.12293},
      archivePrefix={arXiv},
      primaryClass={cs.CV},
      url={https://arxiv.org/abs/2111.12293}, 
}

@misc{tsptq2024,
      title={TSPTQ-ViT: Two-scaled post-training quantization for vision transformer}, 
      author={Yu-Shan Tai and Ming-Guang Lin and An-Yeu and Wu},
      year={2023},
      eprint={2305.12901},
      archivePrefix={arXiv},
      primaryClass={eess.IV},
      url={https://arxiv.org/abs/2305.12901}, 
}

@misc{adalog2024,
      title={AdaLog: Post-Training Quantization for Vision Transformers with Adaptive Logarithm Quantizer}, 
      author={Zhuguanyu Wu and Jiaxin Chen and Hanwen Zhong and Di Huang and Yunhong Wang},
      year={2024},
      eprint={2407.12951},
      archivePrefix={arXiv},
      primaryClass={cs.CV},
      url={https://arxiv.org/abs/2407.12951}, 
}

@misc{adfq2025,
      title={ADFQ-ViT: Activation-Distribution-Friendly Post-Training Quantization for Vision Transformers}, 
      author={Yanfeng Jiang and Ning Sun and Xueshuo Xie and Fei Yang and Tao Li},
      year={2024},
      eprint={2407.02763},
      archivePrefix={arXiv},
      primaryClass={cs.CV},
      url={https://arxiv.org/abs/2407.02763}, 
}

@misc{ahcptq2025,
      title={AHCPTQ: Accurate and Hardware-Compatible Post-Training Quantization for Segment Anything Model}, 
      author={Wenlun Zhang and Yunshan Zhong and Shimpei Ando and Kentaro Yoshioka},
      year={2025},
      eprint={2503.03088},
      archivePrefix={arXiv},
      primaryClass={cs.CV},
      url={https://arxiv.org/abs/2503.03088}, 
}

@misc{dmq2025,
      title={DMQ: Dissecting Outliers of Diffusion Models for Post-Training Quantization}, 
      author={Dongyeun Lee and Jiwan Hur and Hyounguk Shon and Jae Young Lee and Junmo Kim},
      year={2025},
      eprint={2507.12933},
      archivePrefix={arXiv},
      primaryClass={cs.CV},
      url={https://arxiv.org/abs/2507.12933}, 
}

@misc{spinquant2024,
      title={SpinQuant: LLM quantization with learned rotations}, 
      author={Zechun Liu and Changsheng Zhao and Igor Fedorov and Bilge Soran and Dhruv Choudhary and Raghuraman Krishnamoorthi and Vikas Chandra and Yuandong Tian and Tijmen Blankevoort},
      year={2025},
      eprint={2405.16406},
      archivePrefix={arXiv},
      primaryClass={cs.LG},
      url={https://arxiv.org/abs/2405.16406}, 
}

@misc{rectifiedflow2022,
      title={Flow Straight and Fast: Learning to Generate and Transfer Data with Rectified Flow}, 
      author={Xingchao Liu and Chengyue Gong and Qiang Liu},
      year={2022},
      eprint={2209.03003},
      archivePrefix={arXiv},
      primaryClass={cs.LG},
      url={https://arxiv.org/abs/2209.03003}, 
}

@misc{cutlass2025,
  author       = {{NVIDIA Corporation}},
  title        = {CUTLASS: CUDA Templates and Python DSLs for High-Performance Linear Algebra},
  year         = {2025},
  publisher    = {GitHub},
  journal      = {GitHub repository},
  howpublished = {\url{https://github.com/NVIDIA/cutlass}}
}
}

\clearpage
\setcounter{page}{1}

\appendix
\onecolumn

\section{Quantization Recovery Comparison}
\label{appendix:quant-recovery}

In quantized linear layers, the forward pass is typically composed of three steps: quantizing the input and weight tensors, performing matrix multiplication in the quantized domain, and then recovering the output via dequantization. Formally, for input $\mathbf{X}$ and weight $\mathbf{W}$, where \( \mathbf{X} \in \mathbb{R}^{m \times k} \) and \( \mathbf{W} \in \mathbb{R}^{k \times n} \):
\[
\mathbf{Y} \approx \mathrm{DeQuant}(\mathrm{Quant}(\mathbf{X}) \cdot \mathrm{Quant}(W)),
\]
where $\mathrm{Quant}(\cdot)$ and $\mathrm{DeQuant}(\cdot)$ denote quantization and dequantization, respectively.

This section compares three representative quantization strategies from the perspective of how easily the original scale can be recovered after matrix multiplication:
\begin{itemize}
    \item \textbf{Symmetric quantization}: both inputs and weights are quantized using zero-centered uniform scales without offsets. Specifically, for input $\mathbf{X}$ and weight $\mathbf{W}$, the quantization is defined as:
    \[
    \hat{\mathbf{X}} = \mathrm{round}\left(\frac{\mathbf{X}}{s_X}\right), \quad
    \hat{\mathbf{W}} = \mathrm{round}\left(\frac{\mathbf{W}}{s_w}\right),
    \]
    where $s_X$ and $s_W$ are the quantization scales for input and weight, respectively.  
    The low-bits matrix multiplication yields:
    \[
    \hat{\mathbf{Y}} = \hat{\mathbf{X}} \hat{\mathbf{W}},
    \]
    and the recovered output is simply:
    \[
    \mathbf{Y} \approx s_X s_W \cdot \hat{\mathbf{Y}},
    \]
    which is efficient and scale-preserving due to the absence of zero-points.

    \item \textbf{Asymmetric quantization}: unlike the symmetric case, asymmetric quantization introduces non-zero offsets (zero-points), which shifts the quantized representation. For input $\mathbf{X}$ and weight $\mathbf{W}$, the quantization process is:
    \[
    \hat{\mathbf{X}} = \mathrm{round}\left(\frac{\mathbf{X}}{s_X}\right) + z_X, \quad
    \hat{\mathbf{W}} = \mathrm{round}\left(\frac{\mathbf{W}}{s_W}\right) + z_W,
    \]
    where $s_X$, $s_W$ are the quantization scales, and $z_X$, $z_W$ are the zero-points for input and weight, respectively.
    
    The low-bits matrix multiplication gives:
    \[
    \hat{\mathbf{Y}} = \hat{\mathbf{X}} \hat{\mathbf{W}}.
    \]
    To recover the output in full precision, we must subtract the effects of the zero-points:
    \begin{align*}
    \mathbf{Y} &= \mathbf{X} \mathbf{W} \\
    &\approx s_X\left(\hat{\mathbf{X}} - z_{X}\right)\cdot s_W\left(\hat{\mathbf{W}} - z_{W}\right)  \\
    &=s_X s_W \left( \hat{\mathbf{X}} \cdot \hat{\mathbf{W}} - z_{X}\mathbf{J}_{m\times k}\hat{\mathbf{W}} - z_{W}\hat{\mathbf{X}}\mathbf{J}_{k\times n} + k z_X z_W  \mathbf{J}_{m\times n} \right)\\
    &=s_X s_W \left( \hat{\mathbf{X}} \cdot \hat{\mathbf{W}} - z_X \cdot \mathbf{J}_{m \times 1} \cdot \operatorname{rowsum}(\hat{\mathbf{W}}) - z_{W}\operatorname{rowsum}(\hat{\mathbf{X}}) \cdot \mathbf{J}_{1 \times n} + k z_X z_W \right),\\
    \end{align*}
    where \(\mathbf{J}\) is a matrix of ones.
    
    As shown, asymmetric quantization introduces additional terms involving zero-points, which require extra additions and broadcasted summations during the recovery of the GEMM output. Although this scheme can improve accuracy when data distributions are significantly shifted, it increases computational complexity and reduces implementation efficiency.

    \item \textbf{Dual-scale quantization (ours)}: decomposes the input into positive and negative channels before quantization, preserving directional fidelity. This allows for scale-aligned matrix multiplication and avoids zero-point corrections. The final output is recovered by linearly combining the quantized results (see the dual-scale quantization formula).
\end{itemize}

Our dual-scale quantization method achieves a favorable trade-off between accuracy and computational efficiency. By decomposing the input into positive and negative channels with separate quantization scales, it preserves directional information that symmetric quantization often loses, thereby improving precision. Compared to asymmetric quantization, it avoids costly zero-point corrections and broadcasted summations during recovery. As a result, the output dequantization is simpler and more efficient, making our method well-suited for practical quantized neural network implementations.

\section{More Experimental Details and Results}
\label{appendix:exp-detail}

\paragraph{Image Quality Evaluation Metrics}
We employ several standard metrics to evaluate the quality and similarity of generated images against reference images.

\begin{itemize}
    \item \textbf{PSNR (Peak Signal-to-Noise Ratio)}: Measures fidelity by quantifying the ratio between the maximum possible power of a signal and the power of corrupting noise. It is based on the \textbf{Mean Squared Error (MSE)}. A higher value is better.
    
    First, given a noise-free $m \times n$ image $I$ and its noisy approximation $K$, MSE is defined as:
    \[ MSE = \frac{1}{mn} \sum_{i=0}^{m-1} \sum_{j=0}^{n-1} [I(i, j) - K(i, j)]^2 \]
    
    The PSNR (in dB) is then defined using $MSE$ and the maximum pixel value $MAX_I$ (e.g., 255 for 8-bit images):
    \[ PSNR = 10 \cdot \log_{10} \left( \frac{MAX_I^2}{MSE} \right) \]
    
    \item \textbf{SSIM (Structural Similarity Index Measure)}: Measures perceptual similarity based on luminance ($\mu$), contrast ($\sigma$), and structure ($\sigma_{xy}$). A higher value (closer to 1) is better.
    \[ SSIM(x, y) = \frac{(2\mu_x\mu_y + C_1)(2\sigma_{xy} + C_2)}{(\mu_x^2 + \mu_y^2 + C_1)(\sigma_x^2 + \sigma_y^2 + C_2)} \]
    where $\mu_x, \mu_y$ are the pixel means, $\sigma_x, \sigma_y$ are the standard deviations, $\sigma_{xy}$ is the covariance of $x$ and $y$, and $C_1, C_2$ are stabilization constants.
    
    \item \textbf{FID (Fréchet Inception Distance)}: Measures the distributional similarity between real ($r$) and generated ($g$) images in the Inception-V3 feature space. A lower value is better.
    \[ FID = ||\mu_r - \mu_g||_2^2 + \text{Tr}(C_r + C_g - 2(C_r C_g)^{\frac{1}{2}}) \]
    where $\mu_r, \mu_g$ are the feature means, $C_r, C_g$ are the covariance matrices, and $\text{Tr}$ is the trace of the matrix.
    
    \item \textbf{LPIPS (Learned Perceptual Image Patch Similarity)}: A learned metric computing the distance between deep features ($l$) of two images ($x, x_0$). A lower value is better.
    \[ d(x, x_0) = \sum_l \frac{1}{H_l W_l} \sum_{h,w} || w_l \odot (\hat{y}_{hw}^l - \hat{y}_{0hw}^l) ||_2^2 \]
    where $l$ is the layer, $\hat{y}^l, \hat{y}_{0hw}^l$ are the normalized feature activations from that layer, $w_l$ are the learned channel weights, and $H_l, W_l$ are the feature map dimensions.
    
    \item \textbf{C.SCR (CLIP Score)}: Measures semantic alignment between an image $I$ and text $T$. A higher value is better.
    \[ \text{CLIPScore}(I, T) = 100 \cdot \cos(E_I, E_T) \]
    where $E_I$ and $E_T$ are the image and text feature embeddings from the CLIP model, and $\cos$ is the cosine similarity.
    
    \item \textbf{C.IQA \& IR (CLIP-IQA \& Image Reward)}: No-Reference (NR) metrics that predict human preference scores. They are learned models (Reward Models) and do not have a static formula. A higher value is better.
\end{itemize}

\paragraph{Extra Hyperparameter Settings and Implementation Details}

In all experiments involving diffusion models, we enable classifier-free guidance and set the guidance scale to 7. Regarding quantization granularity: all 8-bit quantization uses a per-tensor scheme. In 4-bit settings, weights are quantized per-channel, and activations are quantized dynamically per-token.

For SegQuant, the SmoothQuant algorithm is applied with the $\alpha$ parameter individually selected for each linear layer.
We sweep $\alpha$ in the range from 0.0 to 1.0 with a step size of 0.1, choosing the value that minimizes the mean squared error (MSE) between the quantized and full-precision layer outputs.
For 4-bit quantization, SegQuant uses SVDQuant as the optimizer instead of SmoothQuant.
In the \textit{DualScale} scheme, we focus on polarity-sensitive activation functions, specifically SiLU, GELU, and GEGLU, to ensure asymmetric activations are well preserved.

For SVDQuant, the low-rank setting is fixed at 64. For the FLUX model, singular value decomposition (SVD) is performed using float32 precision due to implementation constraints, while all other models use float64 precision for better numerical stability.
For PTQD and TAC-Diffusion, calibration is applied only to the unconditional branch of the model, with 32 images used for sampling.
For TAC-Diffusion, we use $\lambda_1{=}0.8$, $\lambda_2{=}0.1$, and a threshold of 4, following the original implementation.
Additional implementation details and full configuration files can be found in our released codebase.

To facilitate future research and ensure full reproducibility of the results presented in this paper, we have released our complete implementation at: {\url{https://github.com/OptiSys-ZJU/segquant}}

\paragraph{Dual Scale Sensitivity Analysis}

To evaluate the sensitivity of our dual-scale quantization to outlier activations, we conducted an analysis comparing the strict Min/Max calibration strategy ($p=1.0$) against percentile clipping ($p \le 0.9999$). As detailed in Table~\ref{tab:dualscale_sensitivity}, using the absolute maximum ($p=1.0$) consistently yields the lowest relative F-norm error, avoiding the destructive effects of clipping. This finding implies that high-magnitude outlier activations in diffusion models carry critical semantic signals. Furthermore, our dual-scale approach significantly outperforms standard quantization at $p=1.0$, demonstrating that splitting dimensions according to polarity ($X_+, X_-$) is structurally much more effective for heavy-tailed distributions than conventional clipping methods. These results remain stable across calibration set sizes of $N=100$ and $N=200$, confirming our method does not rely on massive calibration data.

\begin{table}[h]
    \centering
    \caption{Dual Scale Sensitivity Analysis on SD3. We report the relative F-norm error across different calibration percentile settings and set sizes ($N$).}
    \label{tab:dualscale_sensitivity}
    \begin{tabular}{lccc}
        \toprule
        \multirow{2}{*}{\textbf{Method}} & \multirow{2}{*}{\textbf{Percentile}} & \multicolumn{2}{c}{\textbf{Rel F-Norm} $\downarrow$} \\
        \cmidrule(lr){3-4}
         & & \textbf{N=100} & \textbf{N=200} \\
        \midrule
        Standard & 1.0 (Max) & 0.0150 & 0.0132 \\
        \textbf{Ours (DualScale)} & \textbf{1.0 (Max)} & \textbf{0.0108} & \textbf{0.0102} \\
        Ours (DualScale) & 0.9999 & 0.0115 & 0.0146 \\
        Ours (DualScale) & 0.9990 & 0.0412 & 0.0583 \\
        \bottomrule
    \end{tabular}
\end{table}

\paragraph{Pattern Detection Algorithm}

To ensure robustness across diverse model architectures and compiler transformations, SegQuant employs a topological pattern matching algorithm based on \texttt{torch.fx} symbolic tracing.
Algorithm~\ref{alg:segquant_detection} details the procedure.
The detector first performs symbolic tracing and shape propagation to obtain a computational graph with metadata.
It then iterates through the graph to identify two categories of patterns:
\begin{itemize}
    \item \textbf{Weight Segmentation (Forward-Tracing):} We identify \texttt{Linear} nodes followed immediately by \texttt{chunk} or \texttt{split} operations. The segment sizes are explicitly inferred from the operator arguments (e.g., \texttt{chunks} count or \texttt{split\_size}).
    \item \textbf{Input Segmentation (Backward-Tracing):} We identify \texttt{Linear} nodes whose inputs originate from \texttt{concat}, \texttt{stack}, or \texttt{reshape} operations. For \texttt{Concat}/\texttt{Stack}, segment sizes are inferred by retrieving the shapes of constituent input tensors; for \texttt{Reshape} (common in multi-head merges), we verify if the flattened input dimension $D_{in}$ decomposes into $N \times S$ to infer the segment size $S$.
\end{itemize}

\begin{algorithm}[h]
    \caption{SegQuant Graph Pattern Detection}
    \label{alg:segquant_detection}
    \begin{algorithmic}[1]
        \REQUIRE Pre-trained model $\mathcal{M}$, Example inputs $\mathcal{X}$.
        \ENSURE Set of detected patterns $\mathcal{P} = \mathcal{P}_{\text{weight}} \cup \mathcal{P}_{\text{input}}$.
        \STATE \textbf{Initialization:}
        \STATE $\mathcal{G} \leftarrow \text{torch.fx.symbolic\_trace}(\mathcal{M})$ \hfill $\triangleright$ Flatten module hierarchy
        \STATE $\text{ShapeProp}(\mathcal{G}, \mathcal{X})$ \hfill $\triangleright$ Propagate tensor metadata
        \STATE $\mathcal{P} \leftarrow \emptyset$
        \STATE \textbf{Main Loop:}
        \FOR{each node $n \in \mathcal{G}.\text{nodes}$}
            \STATE \textbf{Case 1: Weight Segmentation (Forward-Tracing)}
            \IF{$n.\text{op}$ is \texttt{chunk} or \texttt{split}}
                \STATE $n_{\text{prev}} \leftarrow \text{TraceForward}(n.\text{args}[0])$ \hfill $\triangleright$ Skip transparent ops like Dropout/To
                \IF{$n_{\text{prev}}$ is \texttt{nn.Linear}}
                    \STATE $D_{\text{out}} \leftarrow n_{\text{prev}}.\text{out\_features}$
                    \IF{$n$ is \texttt{chunk}}
                        \STATE $K \leftarrow n.\text{args}[\text{chunks}]$
                        \STATE Infer segments: $S_i \approx D_{\text{out}} / K$ for $i \in \{1..K\}$
                    \ELSIF{$n$ is \texttt{split}}
                        \STATE Infer segments: $S_i \leftarrow n.\text{args}[\text{split\_size}]$
                    \ENDIF
                    \STATE Add $(n_{\text{prev}}, \text{Type:Weight}, \{S_i\})$ to $\mathcal{P}$
                \ENDIF
            \ENDIF
            \STATE \textbf{Case 2: Input Segmentation (Backward-Tracing)}
            \IF{$n$ is \texttt{nn.Linear}}
                \STATE $n_{\text{in}} \leftarrow \text{TraceBackward}(n.\text{args}[0])$ \hfill $\triangleright$ Skip transparent ops like Dropout/To
                \STATE $D_{\text{in}} \leftarrow n.\text{in\_features}$
                \IF{$n_{\text{in}}$ is \texttt{concat} or \texttt{stack}}
                    \STATE Retrieve input tensors $\{T_1, T_2, ...\}$ from $n_{\text{in}}.\text{args}$
                    \STATE Infer segments: $S_j \leftarrow T_j.\text{meta}[\text{`shape'}][-1]$
                    \STATE Add $(n, \text{Type:Input}, \{S_j\})$ to $\mathcal{P}$
                \ELSIF{$n_{\text{in}}$ is \texttt{reshape}}
                    \STATE Retrieve input shape $[..., N, S]$ from $n_{\text{in}}.\text{args}[0].\text{meta}$
                    \IF{$N \times S == D_{\text{in}}$}
                        \STATE Infer segments: $N$ segments of size $S$ \hfill $\triangleright$ Detects Reshape Merges
                        \STATE Add $(n, \text{Type:Input}, \{S\} \times N)$ to $\mathcal{P}$
                    \ENDIF
                \ENDIF
            \ENDIF
        \ENDFOR
        \RETURN $\mathcal{P}$
    \end{algorithmic}
\end{algorithm}

\paragraph{Validity of Semantic Graph Alignment}

By semantic alignment, we mean quantizing according to the model's intrinsic semantic boundaries (e.g., time versus latent features). To isolate the gains brought by this semantic alignment, we ablated the quantization of the \texttt{DiT.0.norm1} layer in SD3. As shown in Table~\ref{tab:semantics_ablation}, our topology-aware segmentation (\textit{SegLinear}) correctly infers $K=6$ chunks and achieves the lowest F-norm error. In contrast, random chunking with the same number of chunks yields significantly higher error, confirming that aligning the quantization process with the model's topological semantics is vital rather than just performing fine-grained segmentation arbitrarily.

\begin{table}[h]
    \centering
    \caption{Semantics Ablation on SD3 \texttt{DiT.0.norm1}. Comparison of our topology-aware segmentation against random and mismatched chunking strategies.}
    \label{tab:semantics_ablation}
    \begin{tabular}{lcc}
        \toprule
        \textbf{Strategy} & \textbf{Chunks} & \textbf{F-Norm Error}$\downarrow$ \\
        \midrule
        \textbf{SegLinear (Ours)} & \textbf{6 (Inferred)} & \textbf{0.5415} \\
        Random Chunk & 6 (Same $K$) & 0.7080 \\
        Mismatched & 8 (Misalign) & 0.5713 \\
        \bottomrule
    \end{tabular}
\end{table}

\paragraph{Quantization Runtime and Memory Consumption.}
To ensure reproducibility and transparency, we analyze the computational resources required by SegQuant.
For the SD3 model, the complete quantization process takes approximately 2.5 hours on a single NVIDIA L20 GPU, utilizing 256 calibration images and 50 diffusion steps.
Enabling hyperparameter search (e.g., with an $\alpha$ step size of 0.1) extends the runtime to $\sim$25 hours due to iterative calibration passes.
The total duration is primarily determined by the calibration set size, diffusion steps, search granularity, and the specific optimizer used (e.g., SVDQuant incurs an additional $\sim$0.5 hours for singular value decomposition).

Regarding memory usage, the calibration process typically requires peak memory equivalent to $\sim$2$\times$ the model size when fully loaded (approx. 40 GB for SD3).
To accommodate the larger FLUX model within a 48 GB budget (e.g., on NVIDIA L20), we employ a sequential swap-in/swap-out strategy that avoids quality degradation; without this, profiling would necessitate $\sim$80 GB (e.g., on NVIDIA H800).
Crucially, as detailed in Table~\ref{tab:layer_stats}, SegQuant achieves efficiency by selectively targeting structure-sensitive layers—applying DualScale to only 12.0\%--28.8\% of the network—rather than processing the entire model uniformly.
Consequently, the memory overhead for storing separate FP16 quantization scales is negligible; for instance, applying input segmentation to 48 attention layers in SD3 adds only $\sim$3.4 MB of parameter storage.
These results confirm that SegQuant significantly enhances granularity with minimal cost, ensuring compatibility with standard deployment infrastructures.

\begin{table}[h]
    \centering
    \caption{Statistics of Optimized Layers. We report the number of linear layers identified and processed by each component of SegQuant. The ``Total Linears" column indicates the number of quantizable linear layers. The ``Est.~Overhead" column estimates the additional memory required to store the fine-grained quantization scales (calculated in FP16), which is negligible (less than 0.3\% of model size).}
    \label{tab:layer_stats}
    \begin{tabular}{lccccc}
        \toprule
        \textbf{Model} & \textbf{Total Linears} & \textbf{DualScale} & \textbf{SegLinear (Weight)} & \textbf{SegLinear (Input)} & \textbf{Est. Overhead} \\
        \midrule
        SD3.5-Medium       & 340 & 98  \small{(28.8\%)} & 49 \small{(14.4\%)} & 48 \small{(14.1\%)} & $\sim$3.4 MB \\
        FLUX.1-dev         & 504 & 118 \small{(23.4\%)} & 77 \small{(15.3\%)} & 38 \small{(7.5\%)}  & $\sim$5.4 MB \\
        SDXL-base-1.0      & 743 & 89  \small{(12.0\%)} & 78 \small{(10.5\%)} & 156 \small{(21.0\%)}& $\sim$5.6 MB \\
        \bottomrule
    \end{tabular}
\end{table}

\paragraph{Performance Optimization and Implementation Efficiency}

To further enhance runtime efficiency, we implement several system-level optimizations in SegQuant. Both \textit{SegLinear} and \textit{DualScale} are designed with high parallelism in mind. Their segmentation-aware structure allows the quantization, calibration, and inference processes to be fully \textbf{vectorized} using native PyTorch tensor operations, significantly accelerating layer-wise tracing and calibration.

For quantized inference, we develop a series of customized \texttt{Quant}/\texttt{DeQuant} and batched GEMM kernels with precision-aware execution paths, which effectively reduce latency under low-bit computation. Although \textit{DualScale} requires two consecutive GEMM operations, these computations are fused and parallelized via \texttt{BatchedGEMM}/\texttt{ArrayGEMM} primitives. We implement these operations using the CUTLASS library to enable efficient dual-scale computation while maintaining compatibility with standard GPU architectures.

In our W8A8 online inference benchmark (see Table~\ref{tab:perf_dualscale}), we compare the proposed DualScale implementation against the PTQ4ViT\cite{ptq4vit2023} method applied to DiT’s AdaNorm layer, which exhibits both \textit{output-segmented} and \textit{polarity-asymmetric} properties. The results demonstrate that our kernel achieves higher inference speed while maintaining equivalent numerical accuracy.

Looking ahead, we plan to introduce optimizer- and calibrator-specific kernel customization, along with sparsity-aware and memory-efficient variants of DualScale, to further mitigate memory bottlenecks and enhance deployment scalability.

\begin{table}[htbp]
\centering
\caption{Runtime comparison of AdaNorm layer quantization and inference (W8A8) on an NVIDIA RTX 4090 GPU.}
\vspace{4pt}
\begin{tabular}{cc
    T{4.2}
    T{4.2}
    T{4.2}
}
\toprule
\textbf{Model}& \textbf{Hidden Dim} & \multicolumn{1}{c}{\textbf{PTQ4ViT} ($\mu$s)} & \multicolumn{1}{c}{\textbf{DualScale} ($\mu$s)}  & \multicolumn{1}{c}{\textbf{SegLinear + DualScale} ($\mu$s)} \\
\midrule
FLUX & 3072   & 4201.05 & \bfseries1074.93 & \textit{1476.85} \\
SD3.5 & 1536  & 1962.20 & \bfseries908.05  & \textit{1360.18} \\
\bottomrule
\end{tabular}
\label{tab:perf_dualscale}
\end{table}

\paragraph{Results on Other Datasets and Visual Evidence}
Beyond the main results, we further evaluate SegQuant on additional datasets and provide corresponding visual analyses.
Specifically, we report CLIP-based metrics—CLIP Score~\cite{clipscore2021} and CLIP-IQA~\cite{iqa2022}—on MJHQ (Table~\ref{tab:mjhq_expr}), and compare SegQuant with other baselines on COCO (Table~\ref{tab:coco_expr}) and DCI (Table~\ref{tab:dci_expr}), using the \texttt{openai/clip-vit-large-patch14} model.
Detailed visual comparisons, including both baseline and ablation studies, are shown in Tables~\ref{tab:more_1}, \ref{tab:more_3} and Figures~\ref{fig:app-mjhq-sd3}, \ref{fig:app-mjhq-sdxl}, \ref{fig:app-mjhq-flux}, \ref{fig:app-dci-flux}, and~\ref{fig:app-mjhq-abl}.
SegQuant consistently preserves semantic structure and visual fidelity under aggressive quantization settings.
In particular, SegQuant-G produces more coherent textures and stronger semantic alignment, demonstrating the advantage of incorporating semantic cues into the quantization process.

\begin{table}[htbp]
\centering
\caption{COCO}
\begin{tabular}{ccc
  T{3.2}  
  T{1.3}  
  T{1.3}  
  T{2.2}  
  T{1.3}  
  T{2.2}  
  T{1.3}  
}
\toprule

\multirow{2}{*}{\textbf{Backbone}} & \multirow{2}{*}{\textbf{W/A}} & \multirow{2}{*}{\textbf{Method}} & \multicolumn{4}{c}{\textbf{Quality}} & \multicolumn{3}{c}{\textbf{Similarity}} \\

\cmidrule(lr){4-7}
\cmidrule(lr){8-10}

& & &
\multicolumn{1}{c}{FID$\downarrow$} &
\multicolumn{1}{c}{IR$\uparrow$} &
\multicolumn{1}{c}{C.IQA$\uparrow$} &
\multicolumn{1}{c}{C.SCR$\uparrow$} &
\multicolumn{1}{c}{LPIPS$\downarrow$} &
\multicolumn{1}{c}{PSNR$\uparrow$} &
\multicolumn{1}{c}{SSIM$\uparrow$} \\
\midrule
\multirow{10}{*}{SD3.5-DiT} & FP16 & Baseline & 34.70 & 1.027 & 0.476 & 16.86  & {N/A} & {N/A} & {N/A}   \\
\cmidrule(lr){2-10}
 & \multirow{6}{*}{8/8(int)} &  Q-Diffusion & 184.66 & -2.197 & 0.395 & 16.02 & 0.704 & 8.21 & 0.388   \\
& & PTQD        & 33.78 & 0.822 & 0.434 & \bfseries16.47 & 0.454 & 10.79 & 0.523 \\ 
& & PTQ4DiT     & 31.79 & 0.912 & 0.458 & 16.37 & 0.403 & 12.69 & 0.593 \\ 
& & TAC         & 32.22 & 0.863 & 0.458 & 16.31 & 0.417 & 12.53 & 0.583  \\ 
& & \cellcolor{lightgray}SegQuant-A  & \cellcolor{lightgray}32.45 & \cellcolor{lightgray}\bfseries1.020 & \cellcolor{lightgray}\bfseries0.467 & \cellcolor{lightgray}16.35 & \cellcolor{lightgray}\bfseries0.362 & \cellcolor{lightgray}\bfseries13.27 & \cellcolor{lightgray}\bfseries0.618 \\ 
& & \cellcolor{lightgray}SegQuant-G  & \cellcolor{lightgray}\bfseries32.02 & \cellcolor{lightgray}0.991 & \cellcolor{lightgray}0.457 & \cellcolor{lightgray}16.36 & \cellcolor{lightgray}0.376 & \cellcolor{lightgray}13.14 & \cellcolor{lightgray}0.601 \\ 

\cmidrule(lr){2-10}

& \multirow{3}{*}{4/8(int)} & PTQ4DiT & 69.89 & 0.085 & 0.411 & 16.51 & 0.570 & 10.83 & 0.504 \\
& & SVDQuant & \bfseries39.22 & \bfseries0.855 & \bfseries0.454 & 16.40 & \bfseries0.432 & 12.32 & \bfseries0.582 \\
& & \cellcolor{lightgray}SegQuant-G & \cellcolor{lightgray}40.70 & \cellcolor{lightgray}0.843 & \cellcolor{lightgray}0.438 & \cellcolor{lightgray}\bfseries16.45 & \cellcolor{lightgray}0.434 & \cellcolor{lightgray}\bfseries12.40 & \cellcolor{lightgray}0.578 \\

\midrule

\multirow{8}{*}{SDXL-UNet} & FP16 & Baseline & 28.44 & 0.841 & 0.429 & 16.54 & {N/A} & {N/A} & {N/A} \\
\cmidrule(lr){2-10}
 & \multirow{4}{*}{8/8(int)} & Q-Diffusion & 68.06 & -1.574 & 0.396 & 16.48 & 0.537 & 13.68 & 0.517  \\
& & PTQ4DiT & 28.15 & 0.639 & 0.407 & 16.70 & 0.213 & 19.17 & 0.725 \\
& & \cellcolor{lightgray}SegQuant-A & \cellcolor{lightgray}28.37 & \cellcolor{lightgray}0.645 & \cellcolor{lightgray}0.408 & \cellcolor{lightgray}16.71 & \cellcolor{lightgray}0.145 & \cellcolor{lightgray}21.32 & \cellcolor{lightgray}0.785 \\
& & \cellcolor{lightgray}SegQuant-G & \cellcolor{lightgray}\bfseries28.13 & \cellcolor{lightgray}\bfseries0.652 & \cellcolor{lightgray}\bfseries0.408 & \cellcolor{lightgray}\bfseries16.71 & \cellcolor{lightgray}\bfseries0.138 & \cellcolor{lightgray}\bfseries21.63 & \cellcolor{lightgray}\bfseries0.793 \\

\cmidrule(lr){2-10}

& \multirow{3}{*}{8/8(fp)} & Q-Diffusion & 28.24 & 0.843 & 0.428 & 16.46 & 0.104 & 23.32 & 0.832   \\
& & \cellcolor{lightgray}SegQuant-A & \cellcolor{lightgray}28.36 & \cellcolor{lightgray}\bfseries0.844 & \cellcolor{lightgray}\bfseries0.428 & \cellcolor{lightgray}\bfseries16.46 & \cellcolor{lightgray}0.104 & \cellcolor{lightgray}23.31 & \cellcolor{lightgray}0.832   \\
& & \cellcolor{lightgray}SegQuant-G & \cellcolor{lightgray}\bfseries28.00 & \cellcolor{lightgray}0.839 & \cellcolor{lightgray}0.427 & \cellcolor{lightgray}16.46 & \cellcolor{lightgray}\bfseries0.093 & \cellcolor{lightgray}\bfseries23.90 & \cellcolor{lightgray}\bfseries0.842  \\

\midrule

\multirow{8}{*}{FLUX-DiT} & BF16 & Baseline & 36.25 & 0.914 & 0.445 & 16.89  & {N/A} & {N/A} & {N/A}  \\
\cmidrule(lr){2-10}
& \multirow{4}{*}{8/8(int)} & Q-Diffusion & 36.04 & 0.883 & 0.446 & 16.52 & 0.302 & 15.23 & 0.624 \\
& & PTQ4DiT & 38.82 & 0.716 & \bfseries0.460 & 16.45 & 0.328 & 14.73 & 0.620 \\
& & \cellcolor{lightgray}SegQuant-A & \cellcolor{lightgray}36.48 & \cellcolor{lightgray}\bfseries0.907 & \cellcolor{lightgray}0.445 & \cellcolor{lightgray}16.52 & \cellcolor{lightgray}0.155 & \cellcolor{lightgray}19.40 & \cellcolor{lightgray}0.770 \\
& & \cellcolor{lightgray}SegQuant-G & \cellcolor{lightgray}\bfseries36.02 & \cellcolor{lightgray}0.900 & \cellcolor{lightgray}0.444 & \cellcolor{lightgray}\bfseries16.54 & \cellcolor{lightgray}\bfseries0.143 & \cellcolor{lightgray}\bfseries19.94 & \cellcolor{lightgray}\bfseries0.784 \\

\cmidrule(lr){2-10}

& \multirow{3}{*}{4/8(int)} & PTQ4DiT & 41.16 & 0.096 & 0.417 & 16.65 & 0.561 & 11.63 & 0.537 \\
& & SVDQuant & 36.98 & 0.880 & 0.436 & \bfseries16.57 & 0.242 & 16.85 & 0.693 \\
& & \cellcolor{lightgray}SegQuant-G & \cellcolor{lightgray}\bfseries36.88 & \cellcolor{lightgray}\bfseries0.882 & \cellcolor{lightgray}\bfseries0.439 & \cellcolor{lightgray}16.56 & \cellcolor{lightgray}\bfseries0.232 & \cellcolor{lightgray}\bfseries17.03 & \cellcolor{lightgray}\bfseries0.700 \\
\bottomrule
\end{tabular}
\label{tab:coco_expr}
\end{table}

\begin{table}[h!]
\centering
\caption{MJHQ-30K}
\begin{tabular}{ccc
  T{1.3}  
  T{2.2}  
  cc
  T{1.3}  
  T{2.2}  
}
\toprule
\multirow{2}{*}{\textbf{Backbone}} & \multirow{2}{*}{\textbf{W/A}} & \multirow{2}{*}{\textbf{Method}} & \multicolumn{2}{c}{\textbf{Quality}} & \multirow{2}{*}{\textbf{W/A}} & \multirow{2}{*}{\textbf{Method}} & \multicolumn{2}{c}{\textbf{Quality}} \\
\cmidrule(lr){4-5}
\cmidrule(lr){8-9}
& & &
\multicolumn{1}{c}{C.IQA$\uparrow$} &
\multicolumn{1}{c}{C.SCR$\uparrow$} & & &
\multicolumn{1}{c}{C.IQA$\uparrow$} &
\multicolumn{1}{c}{C.SCR$\uparrow$} \\
\midrule


\multirow{7}{*}{SD3.5-DiT} & FP16 & Baseline & 0.481 & 15.91 \\
\cmidrule(lr){2-5}

 & \multirow{6}{*}{8/8(int)} &  PTQD     & 0.441  & 15.91 & \multirow{6}{*}{4/8(int)} & \multirow{2}{*}{PTQ4DiT} & {\multirow{2}{*}{0.428}} & {\multirow{2}{*}{15.86}} \\
& & PTQ4DiT   & 0.461 & \bfseries15.91 & \\ 
& & TAC       & 0.460 & 15.90 & & \multirow{2}{*}{SVDQuant} & {\multirow{2}{*}{\bfseries0.452}} & {\multirow{2}{*}{15.85}} \\ 
& & Smooth+     & 0.458 & 15.85 \\ 
& & \cellcolor{lightgray}SegQuant-A & \cellcolor{lightgray}\bfseries 0.468 & \cellcolor{lightgray}15.86 & & \cellcolor{lightgray} & \cellcolor{lightgray} & \cellcolor{lightgray} \\ 
& & \cellcolor{lightgray}SegQuant-G & \cellcolor{lightgray}0.466 & \cellcolor{lightgray}15.88 & & \multirow{-2}{*}{\cellcolor{lightgray}SegQuant-G} & {\multirow{-2}{*}{\cellcolor{lightgray}0.452}} & {\multirow{-2}{*}{\cellcolor{lightgray}\bfseries15.89}} \\

\midrule


\multirow{4}{*}{SDXL-UNet} & FP16 & Baseline & 0.434 & 15.77 \\
\cmidrule(lr){2-5}

 & \multirow{4}{*}{8/8(int)} & PTQ4DiT & 0.430 & 15.76 & \multirow{4}{*}{8/8(fp)} & \multirow{2}{*}{Q-Diffusion} & {\multirow{2}{*}{0.417}} & {\multirow{2}{*}{15.71}} \\

& & Smooth+ & 0.430 & 15.76 \\

& & \cellcolor{lightgray}SegQuant-A & \cellcolor{lightgray}\bfseries0.433 & \cellcolor{lightgray}15.77 & & \cellcolor{lightgray}SegQuant-A & \cellcolor{lightgray}0.417 & \cellcolor{lightgray}\bfseries 15.71 \\
& & \cellcolor{lightgray}SegQuant-G & \cellcolor{lightgray}0.433 & \cellcolor{lightgray}\bfseries15.78 & & \cellcolor{lightgray}SegQuant-G & \cellcolor{lightgray}\bfseries 0.418 & \cellcolor{lightgray}15.71 \\

\midrule

\multirow{5}{*}{FLUX-DiT} & BF16 & Baseline & 0.440 & 16.00 \\
\cmidrule(lr){2-5}

 & \multirow{5}{*}{8/8(int)} & Q-Diffusion & 0.444 & \bfseries15.98 & \multirow{5}{*}{4/8(int)} & \multirow{2}{*}{PTQ4DiT} & {\multirow{2}{*}{0.416}} & {\multirow{2}{*}{15.98}}  \\
& & PTQ4DiT & \bfseries0.461 & 15.88 & & & & \\

& & Smooth+ & 0.431 & 15.86 & & \multirow{2}{*}{SVDQuant} & {\multirow{2}{*}{0.435}} & {\multirow{2}{*}{15.99}} \\

& & \cellcolor{lightgray}SegQuant-A & \cellcolor{lightgray}0.440 & \cellcolor{lightgray}15.94 &  \\
& & \cellcolor{lightgray}SegQuant-G & \cellcolor{lightgray}0.440 & \cellcolor{lightgray}15.93 & & \cellcolor{lightgray}SegQuant-G & \cellcolor{lightgray}\bfseries0.437 & \cellcolor{lightgray}\bfseries16.01 \\

\bottomrule
\end{tabular}
\label{tab:mjhq_expr}
\end{table}

\begin{table}[h!]
\centering
\caption{DCI}
\begin{tabular}{ccc
  T{3.2}  
  T{1.3}  
  T{1.3}  
  T{2.2}  
  T{1.3}  
  T{2.2}  
  T{1.3}  
}
\toprule

\multirow{2}{*}{\textbf{Backbone}} & \multirow{2}{*}{\textbf{W/A}} & \multirow{2}{*}{\textbf{Method}} & \multicolumn{4}{c}{\textbf{Quality}} & \multicolumn{3}{c}{\textbf{Similarity}} \\

\cmidrule(lr){4-7}
\cmidrule(lr){8-10}

& & &
\multicolumn{1}{c}{FID$\downarrow$} &
\multicolumn{1}{c}{IR$\uparrow$} &
\multicolumn{1}{c}{C.IQA$\uparrow$} &
\multicolumn{1}{c}{C.SCR$\uparrow$} &
\multicolumn{1}{c}{LPIPS$\downarrow$} &
\multicolumn{1}{c}{PSNR$\uparrow$} &
\multicolumn{1}{c}{SSIM$\uparrow$} \\
\midrule
\multirow{10}{*}{SD3.5-DiT} & FP16 & Baseline & 22.17 & 0.685 & 0.451 & 18.05  & {N/A} & {N/A} & {N/A} \\
\cmidrule(lr){2-10}
 & \multirow{6}{*}{8/8(int)} &  Q-Diffusion & 143.36 & -1.973 & 0.460 & 17.92 & 0.691 & 7.80 & 0.254   \\
& & PTQD & 65.42 & -0.454 & 0.416 & 18.08 & 0.582 & 10.00 & 0.296 \\ 
& & PTQ4DiT   & 27.01 & 0.485 & 0.430 & 18.08 & 0.445 & 12.33 & 0.492 \\ 
& & TAC & 28.33 & 0.430 & 0.432 & 18.10 & 0.461 & 12.11 & 0.478 \\ 
& & \cellcolor{lightgray}SegQuant-A & \cellcolor{lightgray}\bfseries24.13 & \cellcolor{lightgray}0.639 & \cellcolor{lightgray}0.440 & \cellcolor{lightgray}\bfseries18.07 & \cellcolor{lightgray}\bfseries0.407 & \cellcolor{lightgray}\bfseries12.74 & \cellcolor{lightgray}\bfseries0.521 \\ 
& & \cellcolor{lightgray}SegQuant-G & \cellcolor{lightgray}24.58 & \cellcolor{lightgray}\bfseries0.639 & \cellcolor{lightgray}\bfseries0.447 & \cellcolor{lightgray}18.06 & \cellcolor{lightgray}0.412 & \cellcolor{lightgray}12.40 & \cellcolor{lightgray}0.516 \\ 

\cmidrule(lr){2-10}

& \multirow{3}{*}{4/8(int)} & PTQ4DiT & 61.68 & -0.314 & 0.403 & \bfseries18.21 & \bfseries0.566 & 10.94 & 0.386 \\
& & SVDQuant & 25.22 & 0.518 & 0.430 & 18.14 & 0.446 & \bfseries12.34 & 0.495 \\
& & \cellcolor{lightgray}SegQuant-G & \cellcolor{lightgray}\bfseries23.50 & \cellcolor{lightgray}\bfseries0.576 & \cellcolor{lightgray}\bfseries0.435 & \cellcolor{lightgray}18.14 & \cellcolor{lightgray}0.438 & \cellcolor{lightgray}12.25 & \cellcolor{lightgray}\bfseries0.497 \\

\midrule

\multirow{8}{*}{SDXL-UNet} & FP16 & Baseline & 24.27 & 0.495 & 0.417 & 17.94  & {N/A} & {N/A} & {N/A} \\
\cmidrule(lr){2-10}
 & \multirow{4}{*}{8/8(int)} & Q-Diffusion & 73.39 & -1.466 & 0.402 & 17.65 & 0.552 & 13.64 & 0.448 \\
& & PTQ4DiT & 24.34 & 0.472 & 0.415 & 17.94 & 0.197 & 19.63 & 0.686 \\
& & \cellcolor{lightgray}SegQuant-A & \cellcolor{lightgray}24.44 & \cellcolor{lightgray}0.487 & \cellcolor{lightgray}0.416 & \cellcolor{lightgray}17.94 & \cellcolor{lightgray}0.132 & \cellcolor{lightgray}21.69 & \cellcolor{lightgray}0.752 \\
& & \cellcolor{lightgray}SegQuant-G & \cellcolor{lightgray}\bfseries24.12 & \cellcolor{lightgray}\bfseries0.504 & \cellcolor{lightgray}\bfseries0.416 & \cellcolor{lightgray}\bfseries17.94 & \cellcolor{lightgray}\bfseries0.123 & \cellcolor{lightgray}\bfseries22.11 & \cellcolor{lightgray}\bfseries0.763 \\

\cmidrule(lr){2-10}

& \multirow{3}{*}{8/8(fp)} & Q-Diffusion & 26.98 & 0.474 & \bfseries0.406 & 17.89 & 0.102 & 23.28 & 0.780   \\
& & \cellcolor{lightgray}SegQuant-A & \cellcolor{lightgray}26.13 & \cellcolor{lightgray}\bfseries0.475 & \cellcolor{lightgray}0.406 & \cellcolor{lightgray}\bfseries17.89 & \cellcolor{lightgray}0.101 & \cellcolor{lightgray}23.34 & \cellcolor{lightgray}0.782   \\
& & \cellcolor{lightgray}SegQuant-G & \cellcolor{lightgray}\bfseries26.10 & \cellcolor{lightgray}0.474 & \cellcolor{lightgray}0.405 & \cellcolor{lightgray}17.89 & \cellcolor{lightgray}\bfseries0.092 & \cellcolor{lightgray}\bfseries23.82 & \cellcolor{lightgray}\bfseries0.793   \\

\midrule

\multirow{8}{*}{FLUX-DiT} & BF16 & Baseline & 22.10 & 0.618 & 0.451 & 18.03 & {N/A} & {N/A} & {N/A} \\
\cmidrule(lr){2-10}
 & \multirow{4}{*}{8/8(int)} & Q-Diffusion & 23.67 & 0.513 & 0.449 & 18.01 & 0.298 & 14.40 & 0.563 \\
& & PTQ4DiT & 26.39 & 0.438 & \bfseries0.472 & 17.97 & 0.552 & 9.22 & 0.321 \\
& & \cellcolor{lightgray}SegQuant-A & \cellcolor{lightgray}22.28 & \cellcolor{lightgray}\bfseries0.592 & \cellcolor{lightgray}0.452 & \cellcolor{lightgray}18.01 & \cellcolor{lightgray}0.149 & \cellcolor{lightgray}18.29 & \cellcolor{lightgray}0.715 \\
& & \cellcolor{lightgray}SegQuant-G & \cellcolor{lightgray}\bfseries22.19 & \cellcolor{lightgray}0.590 & \cellcolor{lightgray}0.451 & \cellcolor{lightgray}\bfseries18.02 & \cellcolor{lightgray}\bfseries0.131 & \cellcolor{lightgray}\bfseries18.97 & \cellcolor{lightgray}\bfseries0.737 \\

\cmidrule(lr){2-10}

& \multirow{3}{*}{4/8(int)} & PTQ4DiT & 72.86 & -0.393 & 0.418 & 17.88 & 0.592 & 10.98 & 0.465  \\
& & SVDQuant & \bfseries22.57 & \bfseries0.575 & 0.442 & 18.03 & 0.224 & 16.19 & 0.635 \\
& & \cellcolor{lightgray}SegQuant-G & \cellcolor{lightgray}23.31 & \cellcolor{lightgray}0.566 & \cellcolor{lightgray}\bfseries0.446 & \cellcolor{lightgray}\bfseries18.03 & \cellcolor{lightgray}\bfseries0.219 & \cellcolor{lightgray}\bfseries16.31 & \cellcolor{lightgray}\bfseries0.639 \\
\bottomrule
\end{tabular}
\label{tab:dci_expr}
\end{table}


\begin{table*}[htbp]
\centering
\begin{minipage}[c]{0.52\textwidth}
\centering
\caption{Additional results on quantization errors of single layers.}
\label{tab:more_1}
\resizebox{\columnwidth}{!}{%
\begin{tabular}{lccc}
\toprule
\multirow{2}{*}{\textbf{Layer Name}} & \multirow{2}{*}{\textbf{Method}} & \multicolumn{2}{c}{\textbf{F-norm}} \\
\cmidrule(lr){3-4}
& & w/o~\textsc{Seg.} & w/~\textsc{Seg.} \\
\midrule
\texttt{DiT.6.norm1\_context} & \multirow{2}{*}{\textsc{SmoothQuant}} & 1.7670 & \textbf{1.0541} \\
\texttt{DiT.22.attn.add\_out} & & 313.96 & \textbf{283.04} \\
\midrule
\texttt{DiT.6.norm1\_context} & \multirow{2}{*}{\textsc{SVDQuant}} & 0.2454 & \textbf{0.2265} \\
\texttt{DiT.22.attn.add\_out} & & 326.17 & \textbf{285.67} \\
\bottomrule
\end{tabular}
}
\end{minipage}
\hfill
\begin{minipage}[c]{0.44\textwidth}
\centering
\caption{Example result of the computation-graph-based automatic semantic search for layer segmentation, using the torch FX graph of SD3.5 model. The listed layer names correspond to those defined in the Diffusers implementation.}
\label{tab:more_3}
\resizebox{\columnwidth}{!}{%
\begin{tabular}{lc}
\toprule
\textbf{Layer Name} & \textbf{Detection Result} \\
\midrule
\texttt{DiT.*.norm1} & \multirow{3}{*}{\textsc{Output-Segmented}} \\
\texttt{DiT.*.norm1\_context} &  \\
\texttt{DiT.*.norm\_out} &  \\
\midrule
\texttt{DiT.*.attn.out} & \multirow{2}{*}{\textsc{Input-Segmented}} \\
\texttt{DiT.*.add\_out} &  \\
\midrule
\texttt{TimeEmbedder.linear\_2} & \multirow{7}{*}{\textsc{Polarity-Asymmetric}} \\
\texttt{TextProjection.linear\_2} &  \\
\texttt{DiT.*.norm1} &  \\
\texttt{DiT.*.norm1\_context} &  \\
\texttt{DiT.*.attn.ff.2} &  \\
\texttt{DiT.*.attn.ff\_context.2} &  \\
\texttt{DiT.*.norm\_out} &  \\
\bottomrule
\end{tabular}
}
\end{minipage}
\end{table*}

\newcommand{\mainimgwidth}{0.139\textwidth}
\newcommand{\maincapwidth}{0.98\textwidth}
\setlength{\fboxsep}{0pt}
\setlength{\fboxrule}{0.2pt}
\begin{figure}[h!]
    \centering
    \begin{minipage}[t]{0.495\textwidth}
        \centering
        \begin{minipage}[t]{\mainimgwidth}
            \centering \tiny{FP16}
        \end{minipage}
        \hfill
        \begin{minipage}[t]{\mainimgwidth}
            \centering \tiny{PTQD}
        \end{minipage}%
        \begin{minipage}[t]{\mainimgwidth}
            \centering \tiny{PTQ4DiT}
        \end{minipage}%
        \begin{minipage}[t]{\mainimgwidth}
            \centering \tiny{TAC-Diffusion}
        \end{minipage}%
        \begin{minipage}[t]{\mainimgwidth}
            \centering \tiny{Smooth+}
        \end{minipage}%
        \begin{minipage}[t]{\mainimgwidth}
            \centering \tiny{\textbf{SegQuant-A}}
        \end{minipage}%
        \begin{minipage}[t]{\mainimgwidth}
            \centering \tiny{\textbf{SegQuant-G}}
        \end{minipage}%
    \end{minipage}
    \hfill
    \begin{minipage}[t]{0.495\textwidth}
        \centering
        \begin{minipage}[t]{\mainimgwidth}
            \centering \tiny{FP16}
        \end{minipage}
        \hfill
        \begin{minipage}[t]{\mainimgwidth}
            \centering \tiny{PTQD}
        \end{minipage}%
        \begin{minipage}[t]{\mainimgwidth}
            \centering \tiny{PTQ4DiT}
        \end{minipage}%
        \begin{minipage}[t]{\mainimgwidth}
            \centering \tiny{TAC-Diffusion}
        \end{minipage}%
        \begin{minipage}[t]{\mainimgwidth}
            \centering \tiny{Smooth+}
        \end{minipage}%
        \begin{minipage}[t]{\mainimgwidth}
            \centering \tiny{\textbf{SegQuant-A}}
        \end{minipage}%
        \begin{minipage}[t]{\mainimgwidth}
            \centering \tiny{\textbf{SegQuant-G}}
        \end{minipage}%
    \end{minipage}

    \vspace{0.6ex}

    \begin{minipage}[t]{0.495\textwidth}
        \centering
        \fbox{\includegraphics[width=\mainimgwidth]{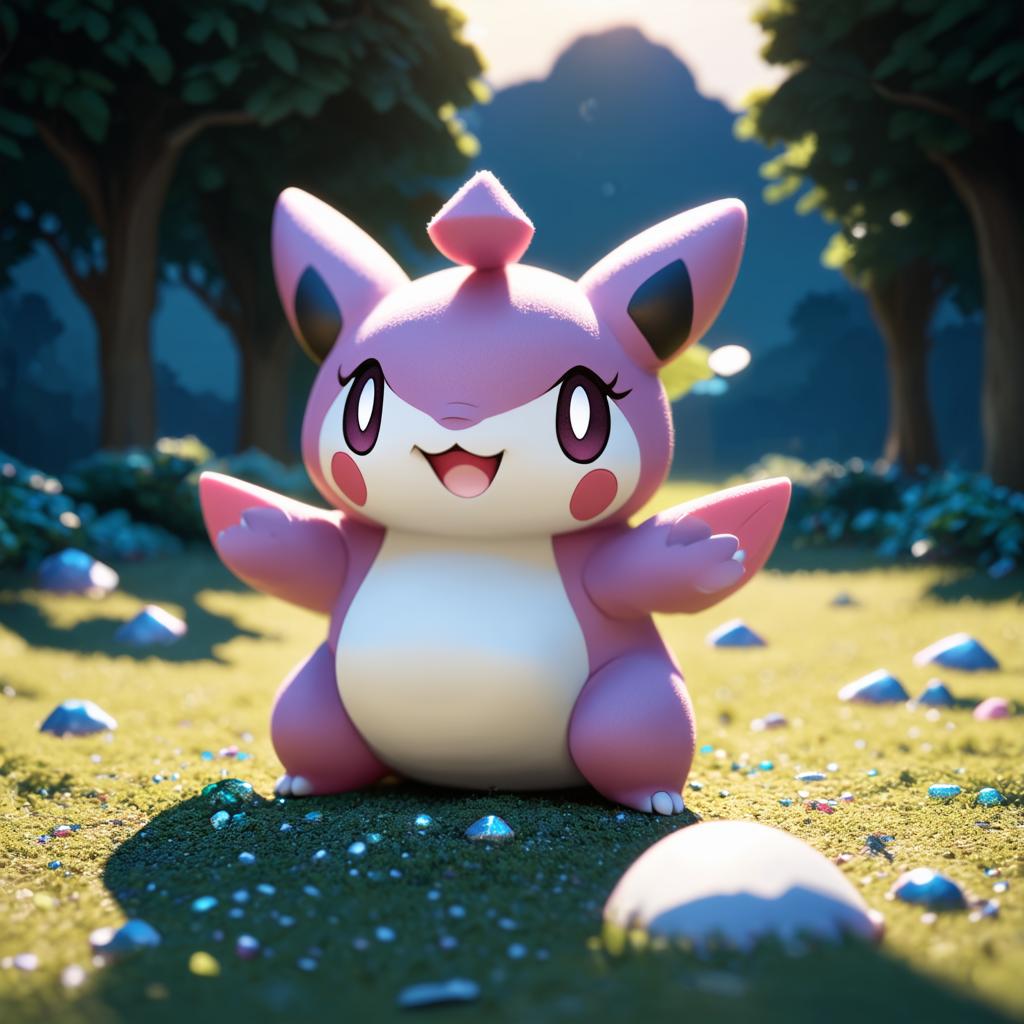}}
        \hfill
        \fbox{\includegraphics[width=\mainimgwidth]{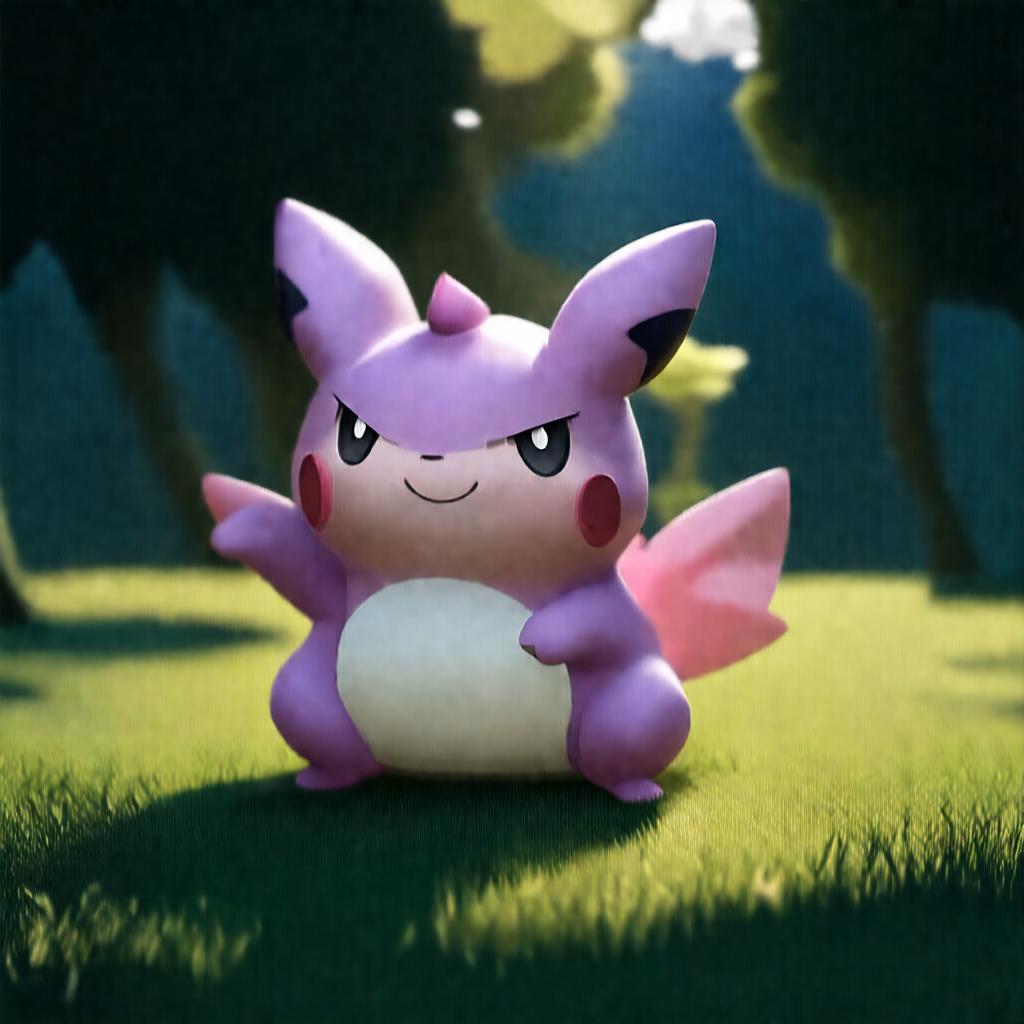}}%
        \fbox{\includegraphics[width=\mainimgwidth]{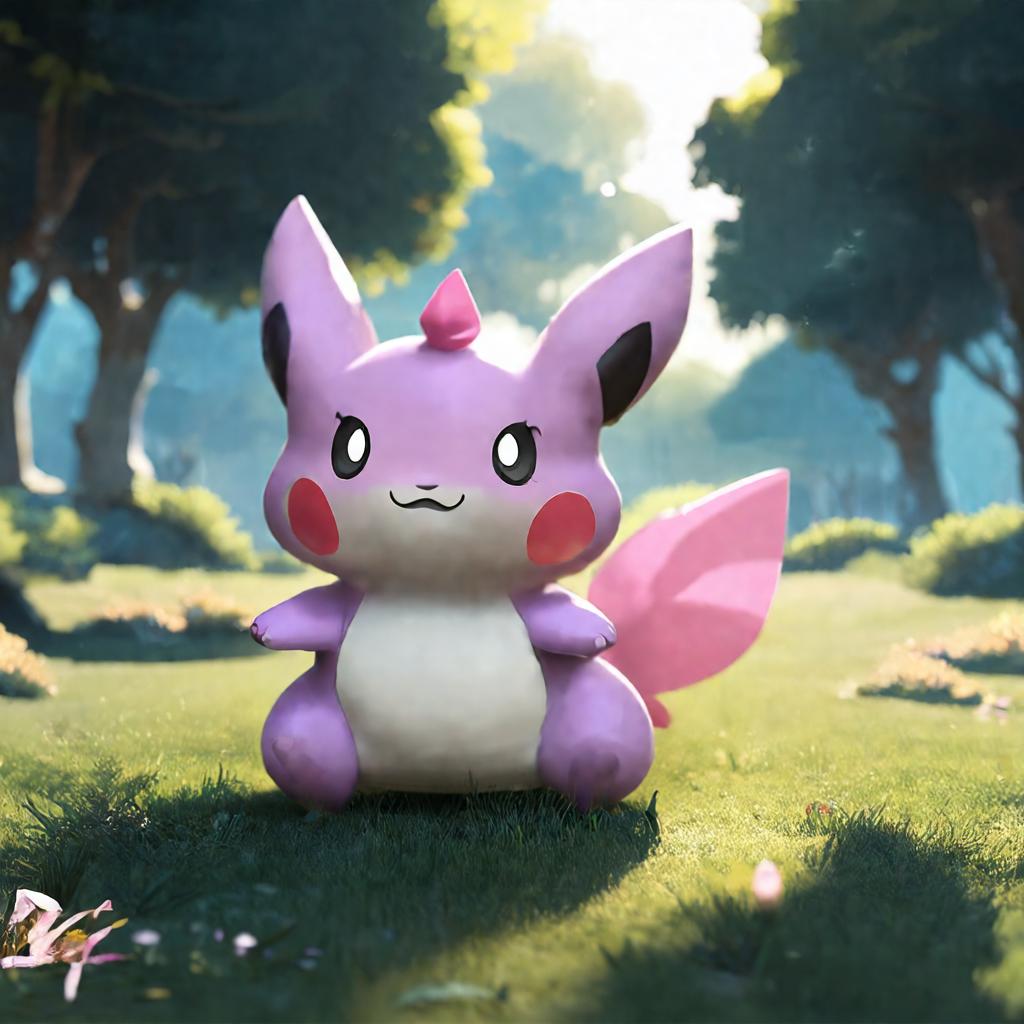}}%
        \fbox{\includegraphics[width=\mainimgwidth]{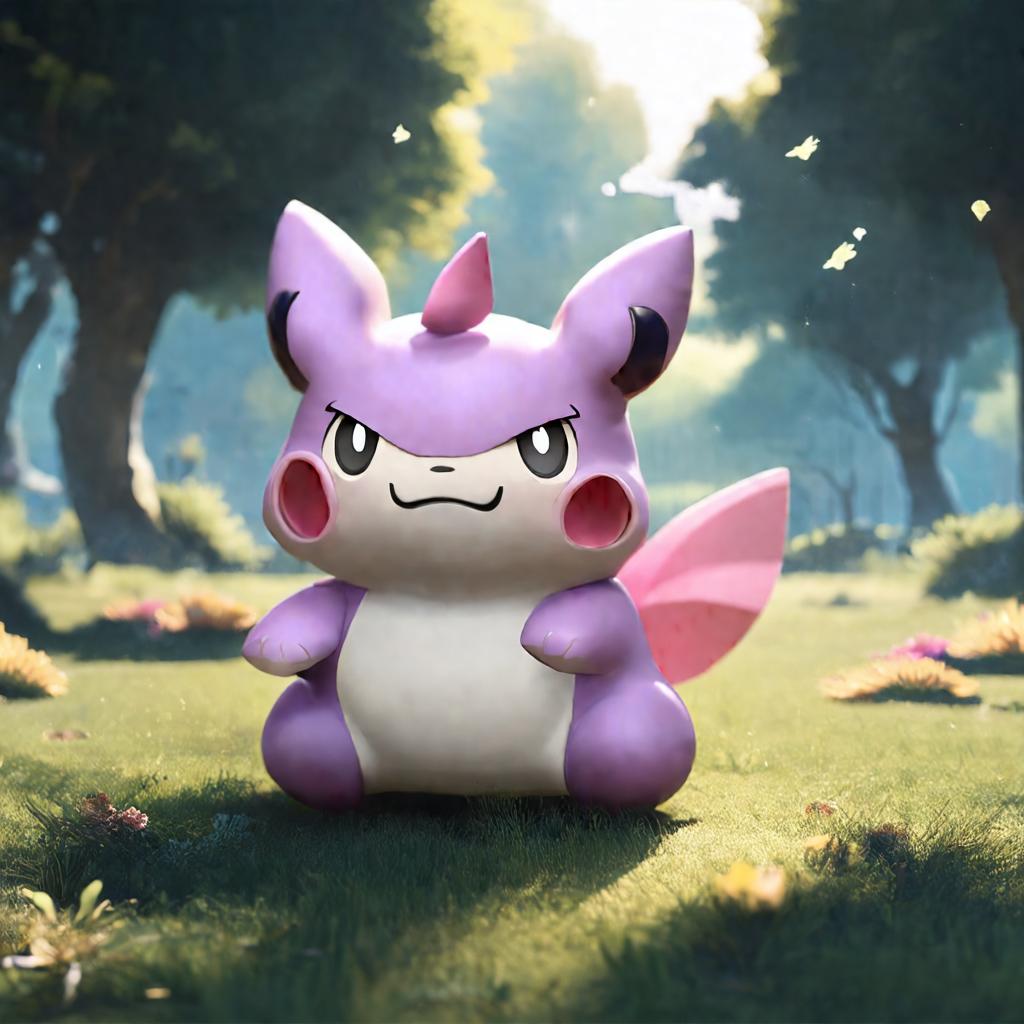}}%
        \fbox{\includegraphics[width=\mainimgwidth]{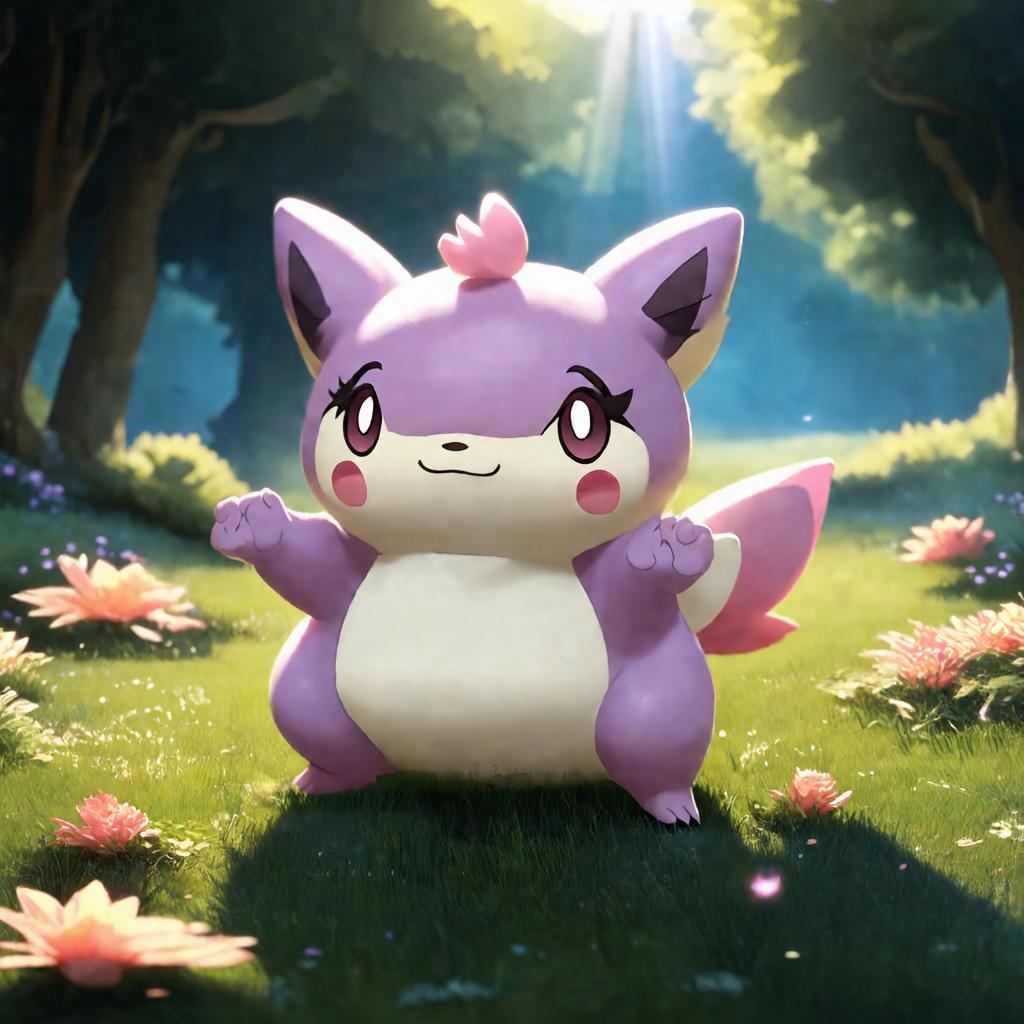}}%
        \fbox{\includegraphics[width=\mainimgwidth]{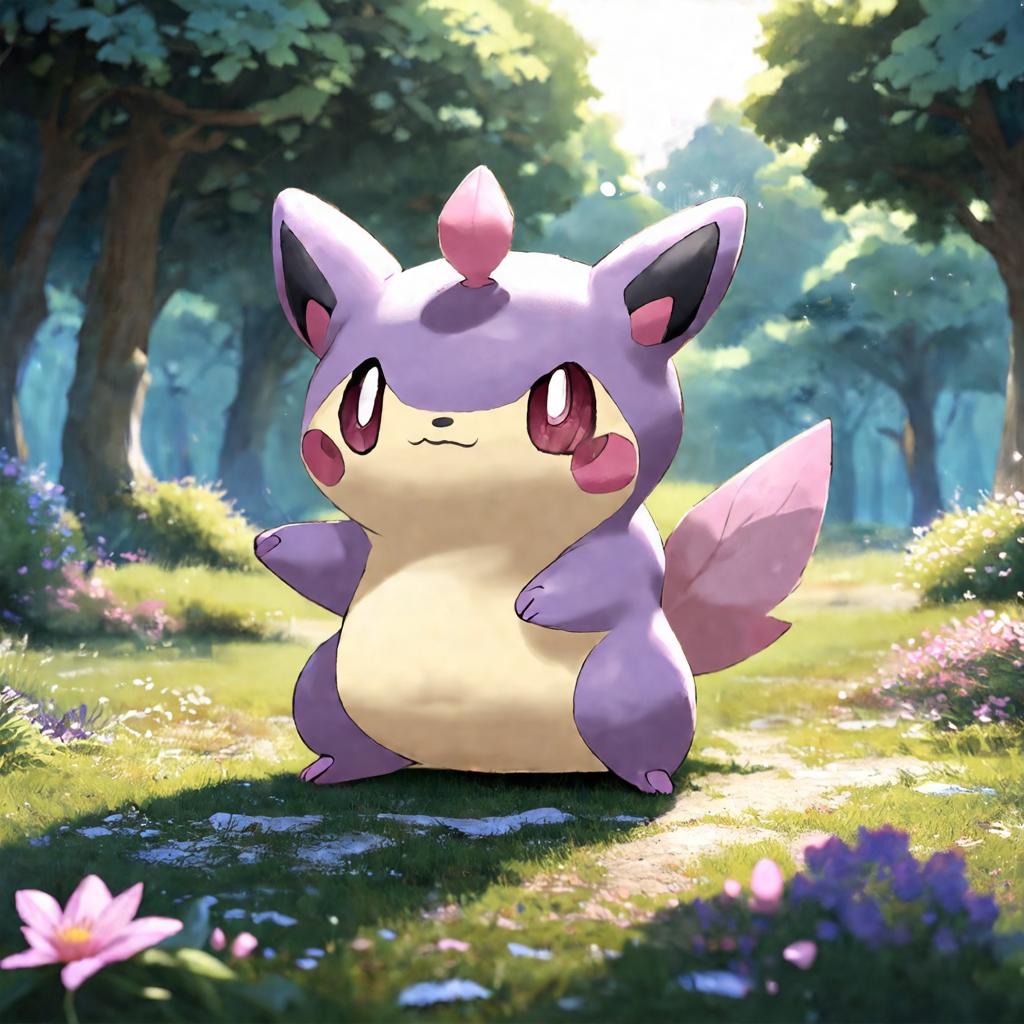}}%
        \fbox{\includegraphics[width=\mainimgwidth]{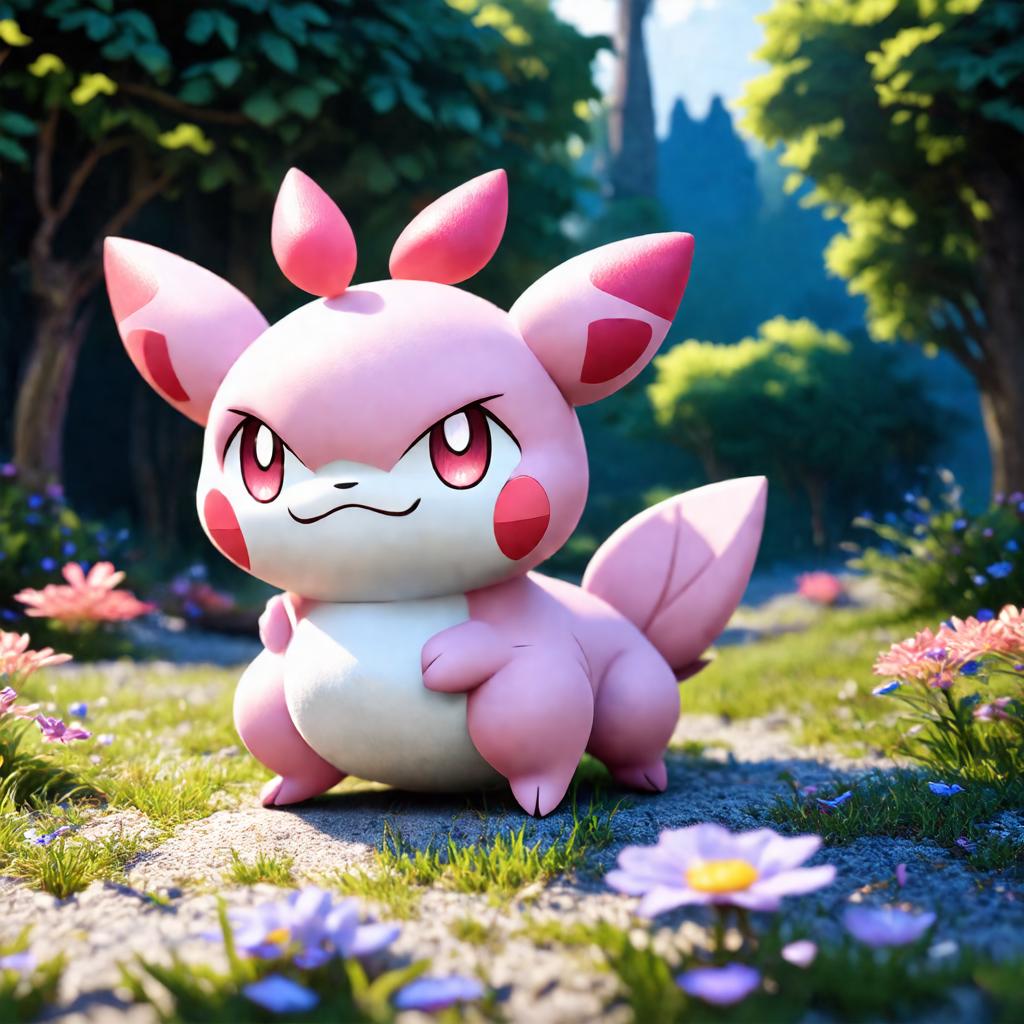}}\\[0.5ex]
        \hfill
        \vspace{-18pt}
        \caption*{
            \begin{minipage}{\maincapwidth}
            \centering
                \tiny{Prompt: \textit{Detailed photography of Pokemon Clefable pick up on the ground some moondust. The final image should be ultradetailed. colors graded with a muted palette of fresh and warm tones to convey the dramatic, exciting essence of the scene. Enhance the captivating charadesign with postproduction techniques such as tone mapping, chromatic aberration, and ambient occlusion for a polished, evocative masterpiece, happy hour, cinematic, award winning photo of the year, dramatic lighting, photorealistic, 50x65, 8k}}
            \end{minipage}
        }
    \end{minipage}
    \hfill
    \begin{minipage}[t]{0.495\textwidth}
        \centering
        \fbox{\includegraphics[width=\mainimgwidth]{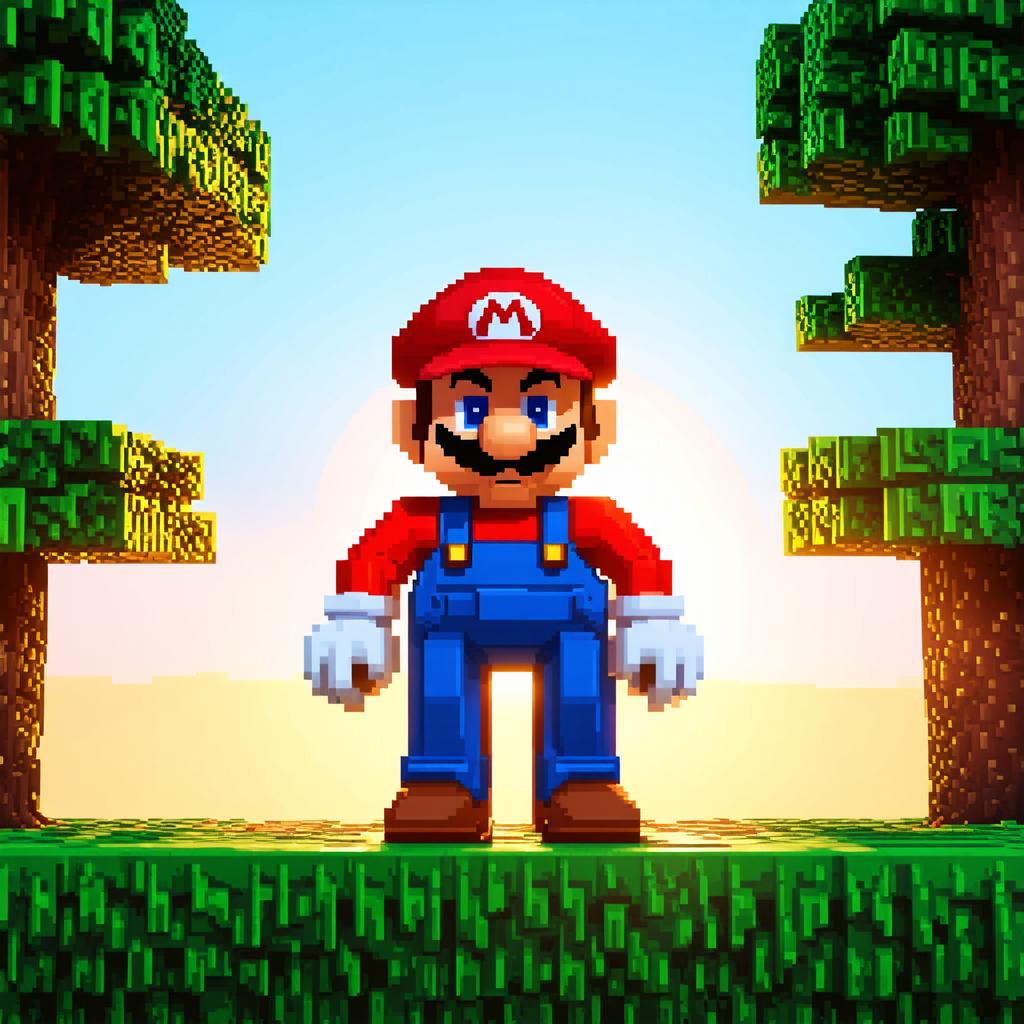}}
        \hfill
        \fbox{\includegraphics[width=\mainimgwidth]{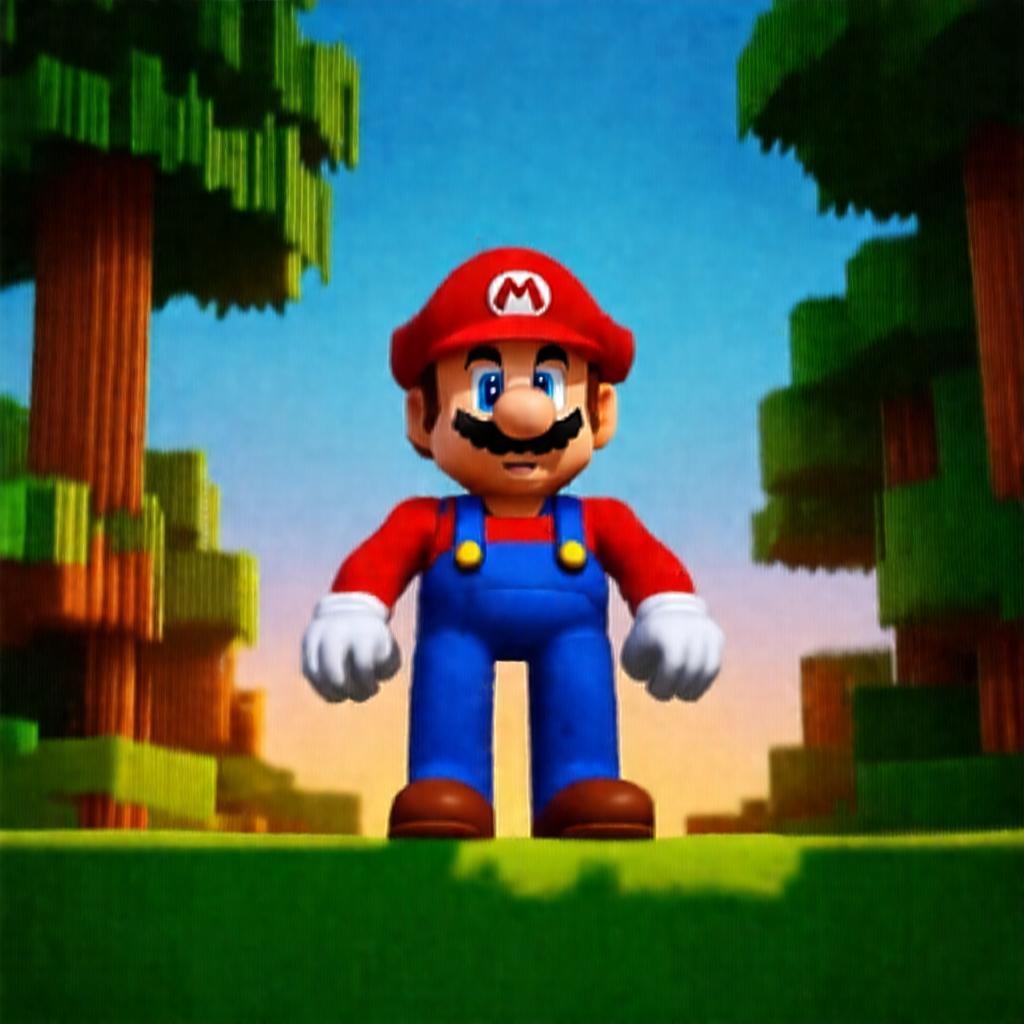}}%
        \fbox{\includegraphics[width=\mainimgwidth]{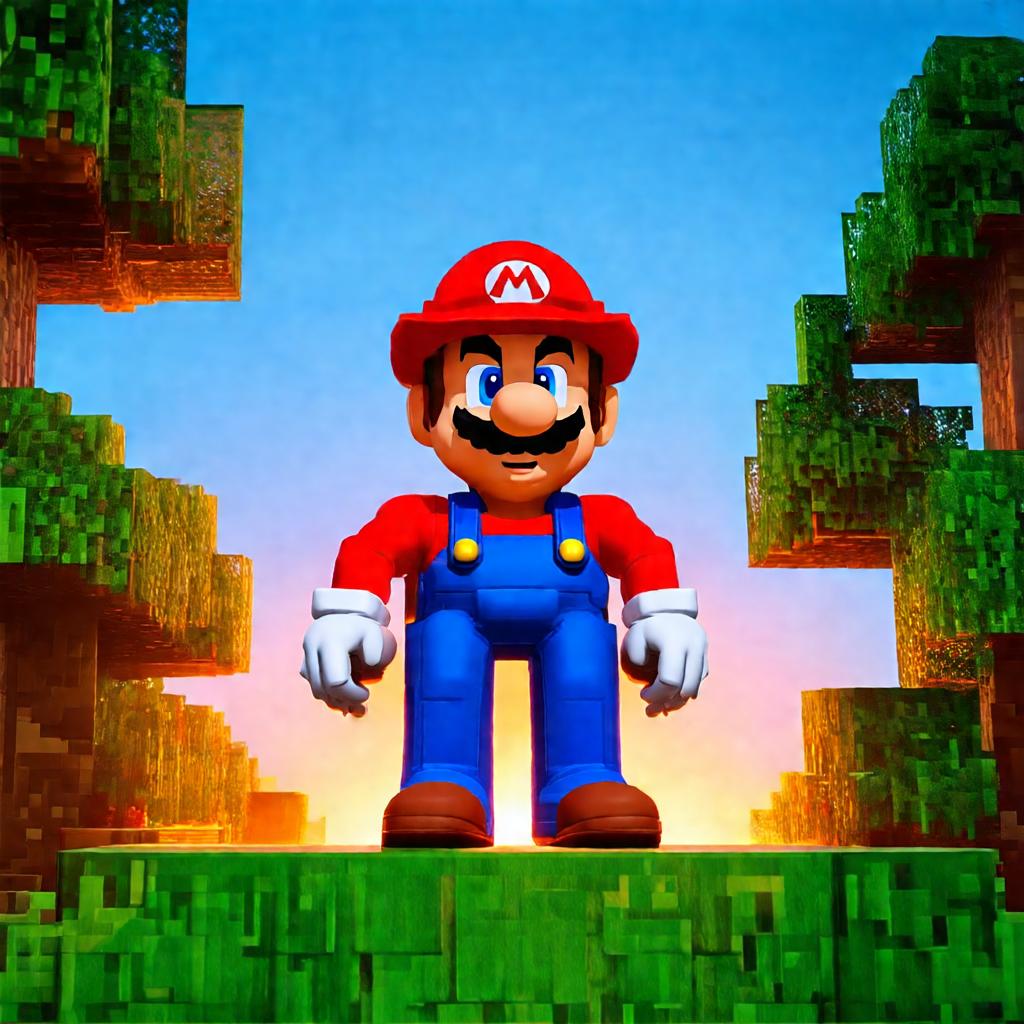}}%
        \fbox{\includegraphics[width=\mainimgwidth]{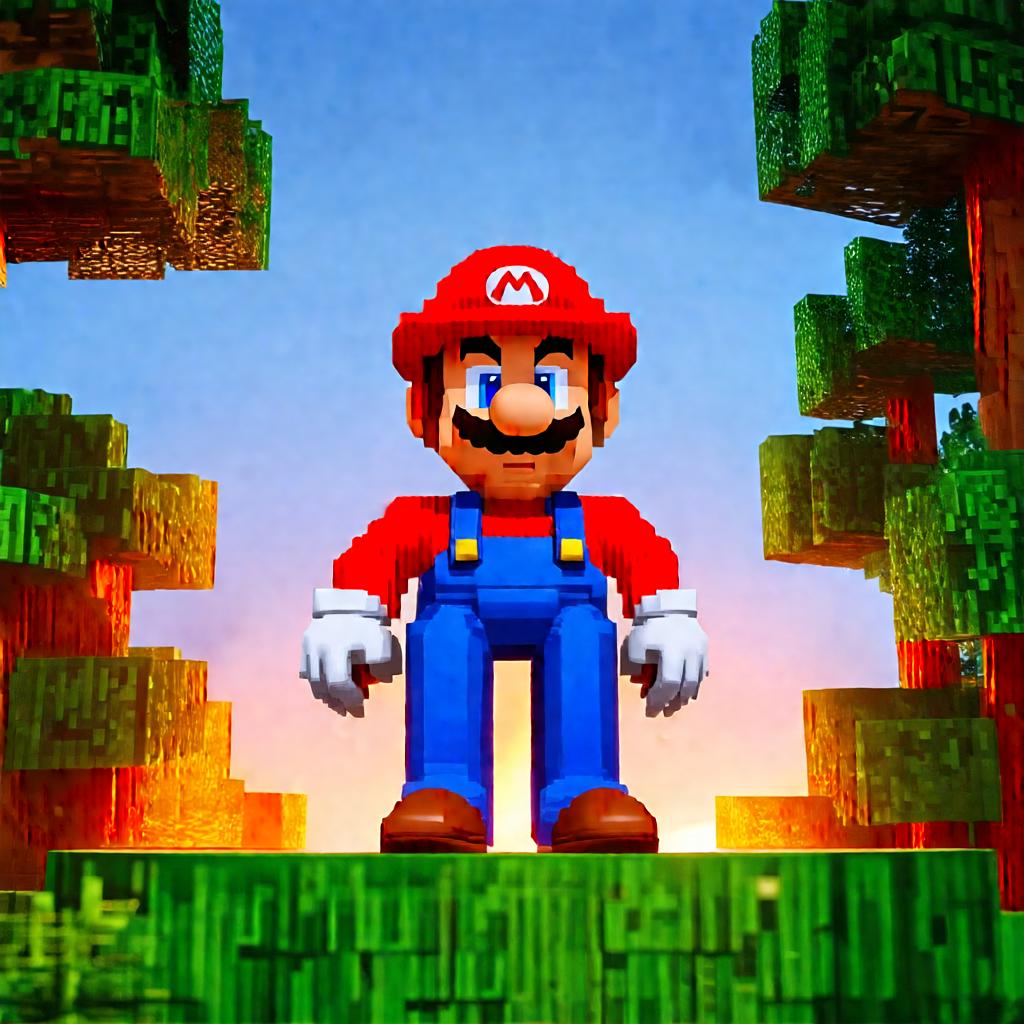}}%
        \fbox{\includegraphics[width=\mainimgwidth]{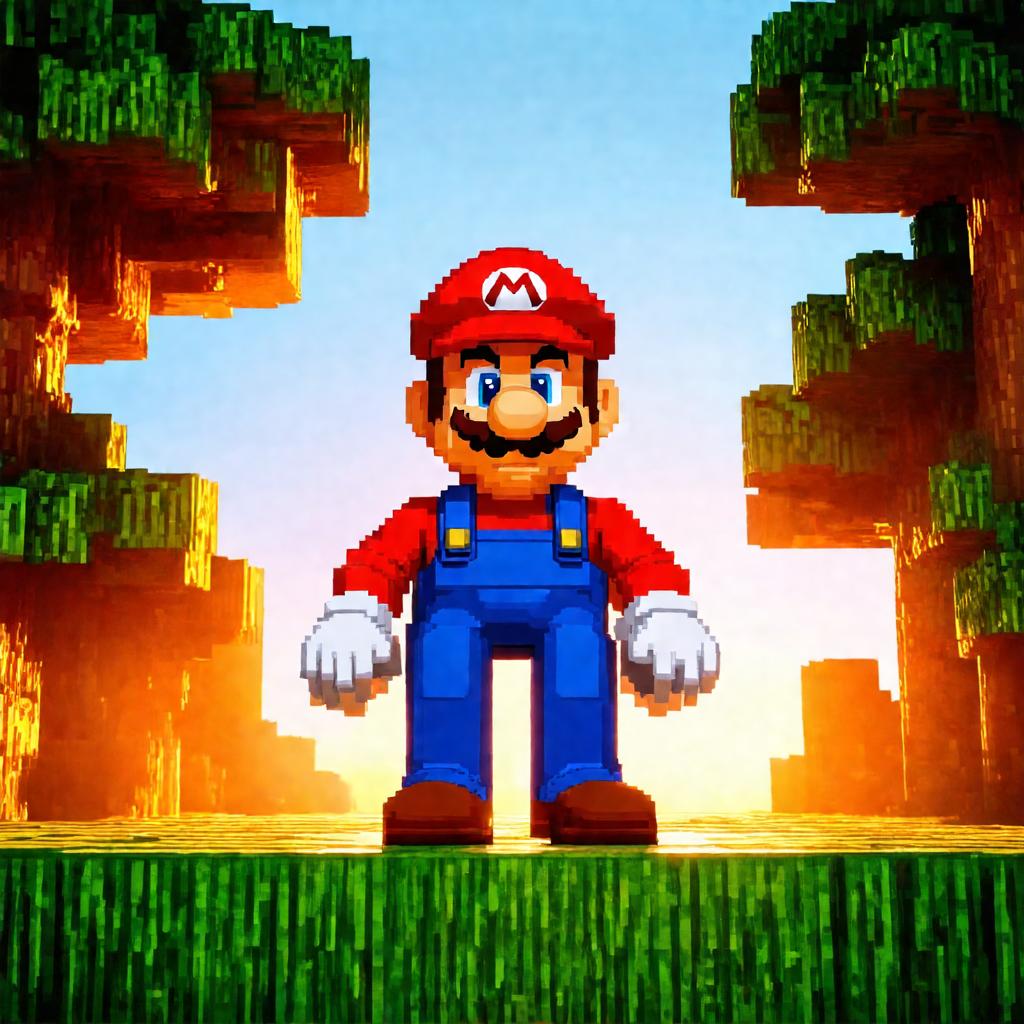}}%
        \fbox{\includegraphics[width=\mainimgwidth]{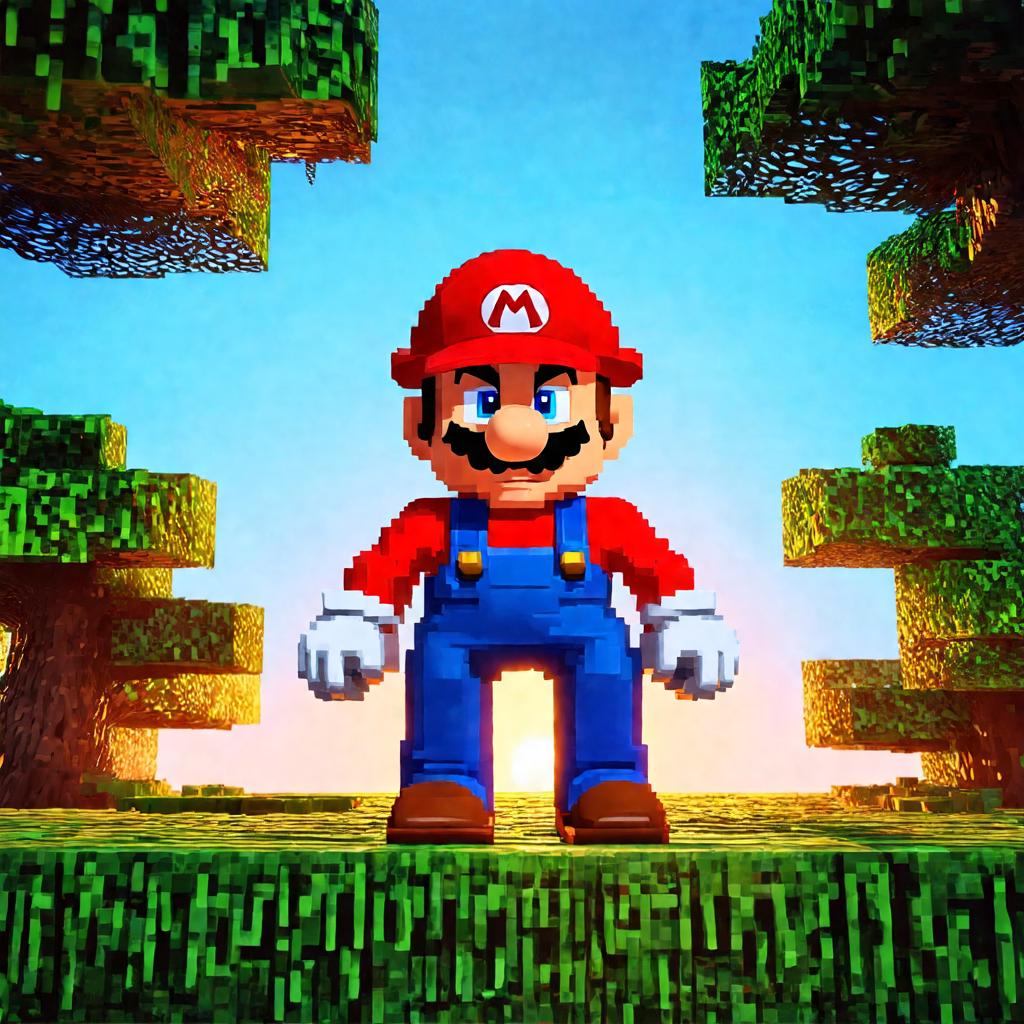}}%
        \fbox{\includegraphics[width=\mainimgwidth]{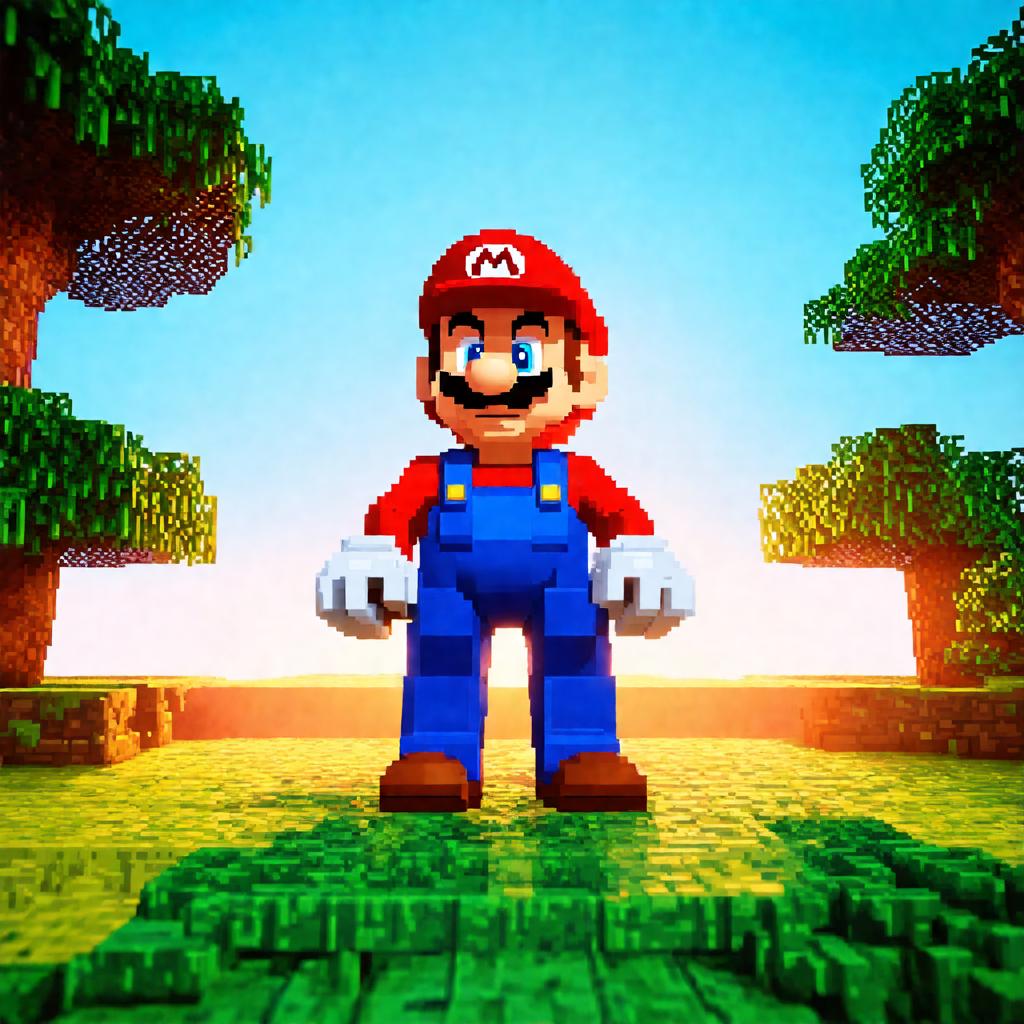}}\\[0.5ex]
        \hfill
        \vspace{-18pt}
        \caption*{
            \begin{minipage}{\maincapwidth}
            \centering
                \tiny{Prompt: \textit{Mario in a 3D Minecraft world, blocky world, Minecraft world, minecraft trees, sunset in the horizon, colourful lighting, centered, front view, red hat, white gloves, blue pants, moustache, ultrarealistic, unreal engine 5, HDR, Ray tracing, cinematic, depth of field, sharp focus, natural light, concept art, super resolution, cartoon style, pixar style, flat coloring.}}
            \end{minipage}
        }
    \end{minipage}
    
    \begin{minipage}[t]{0.495\textwidth}
        \centering
        \fbox{\includegraphics[width=\mainimgwidth]{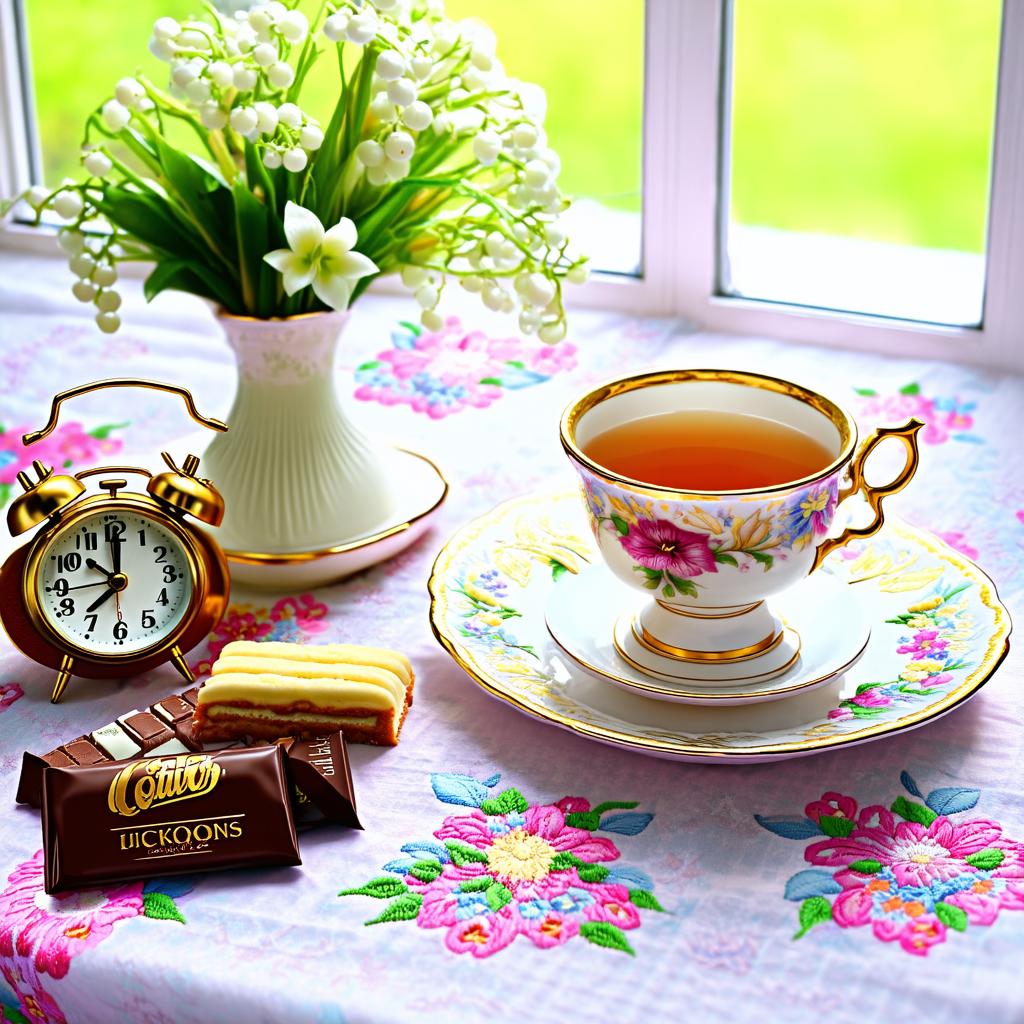}}
        \hfill
        \fbox{\includegraphics[width=\mainimgwidth]{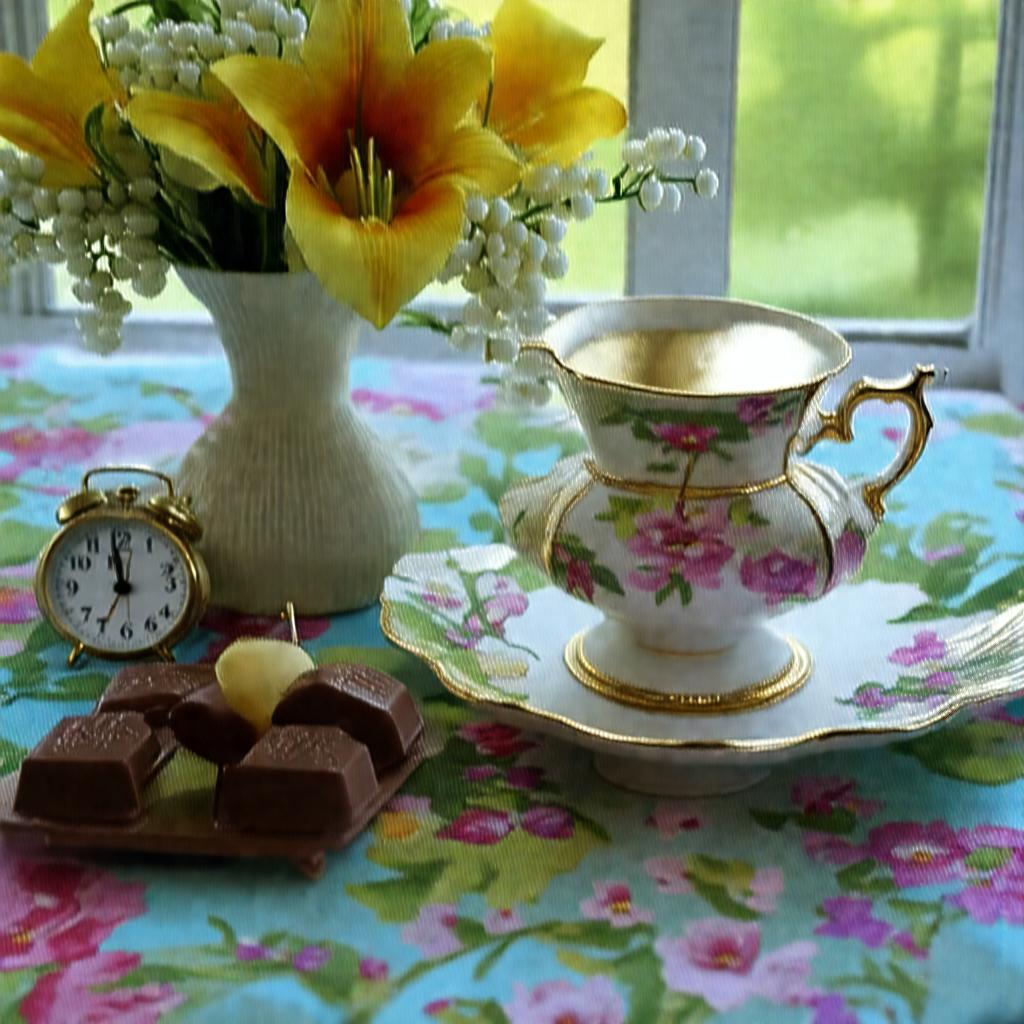}}%
        \fbox{\includegraphics[width=\mainimgwidth]{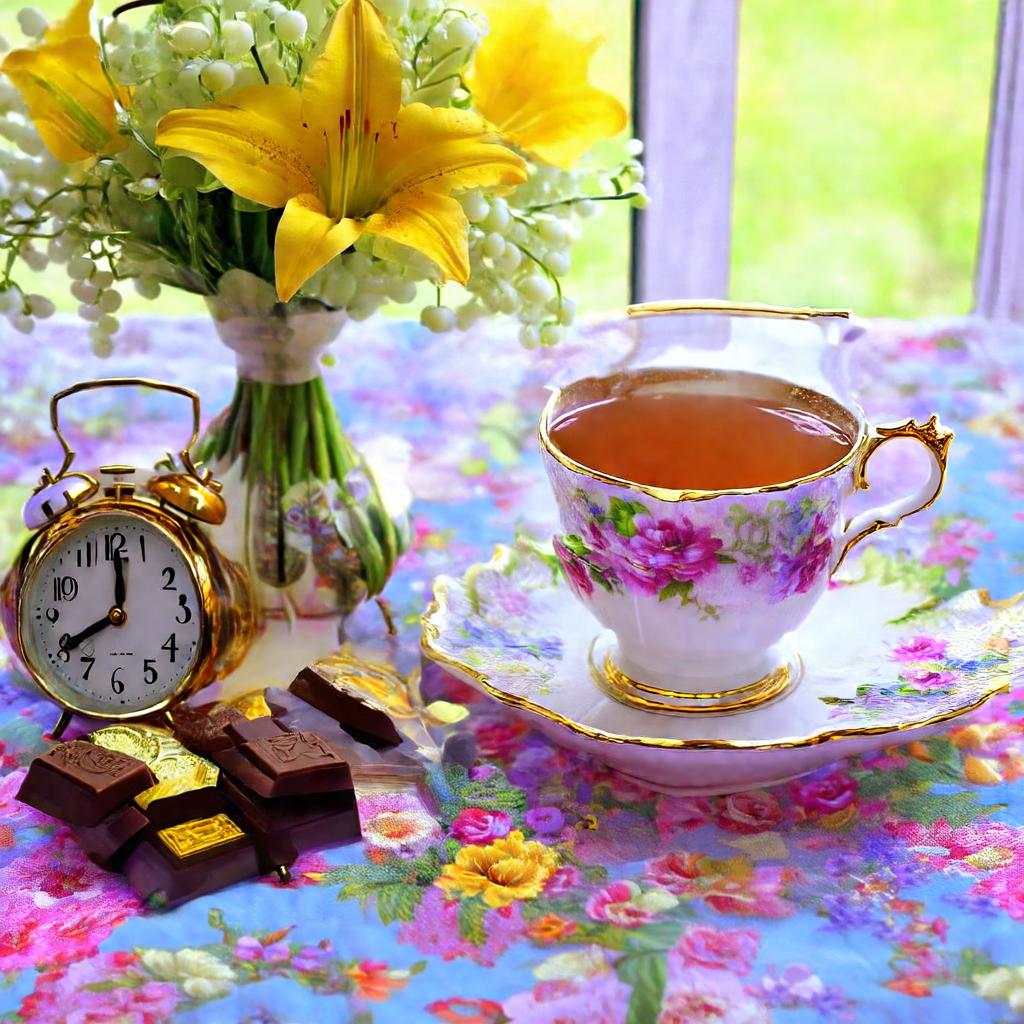}}%
        \fbox{\includegraphics[width=\mainimgwidth]{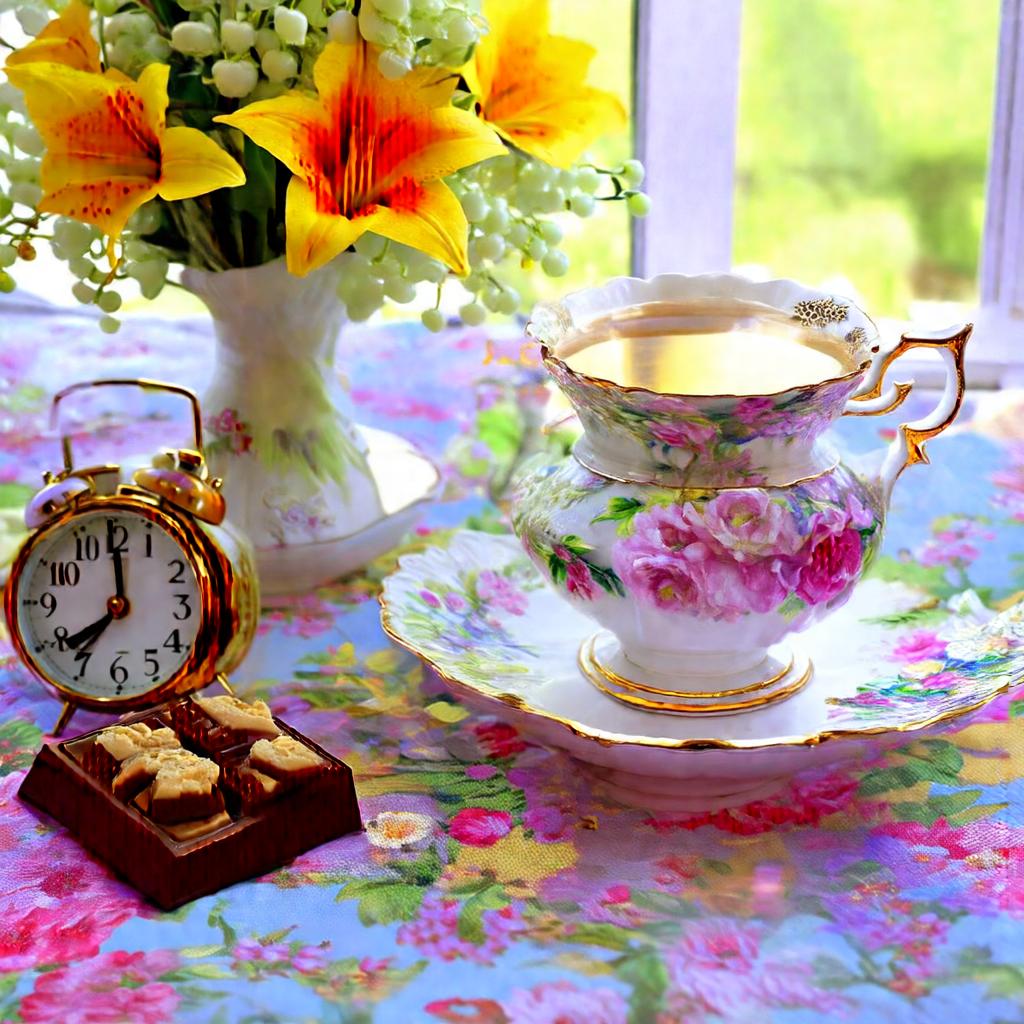}}%
        \fbox{\includegraphics[width=\mainimgwidth]{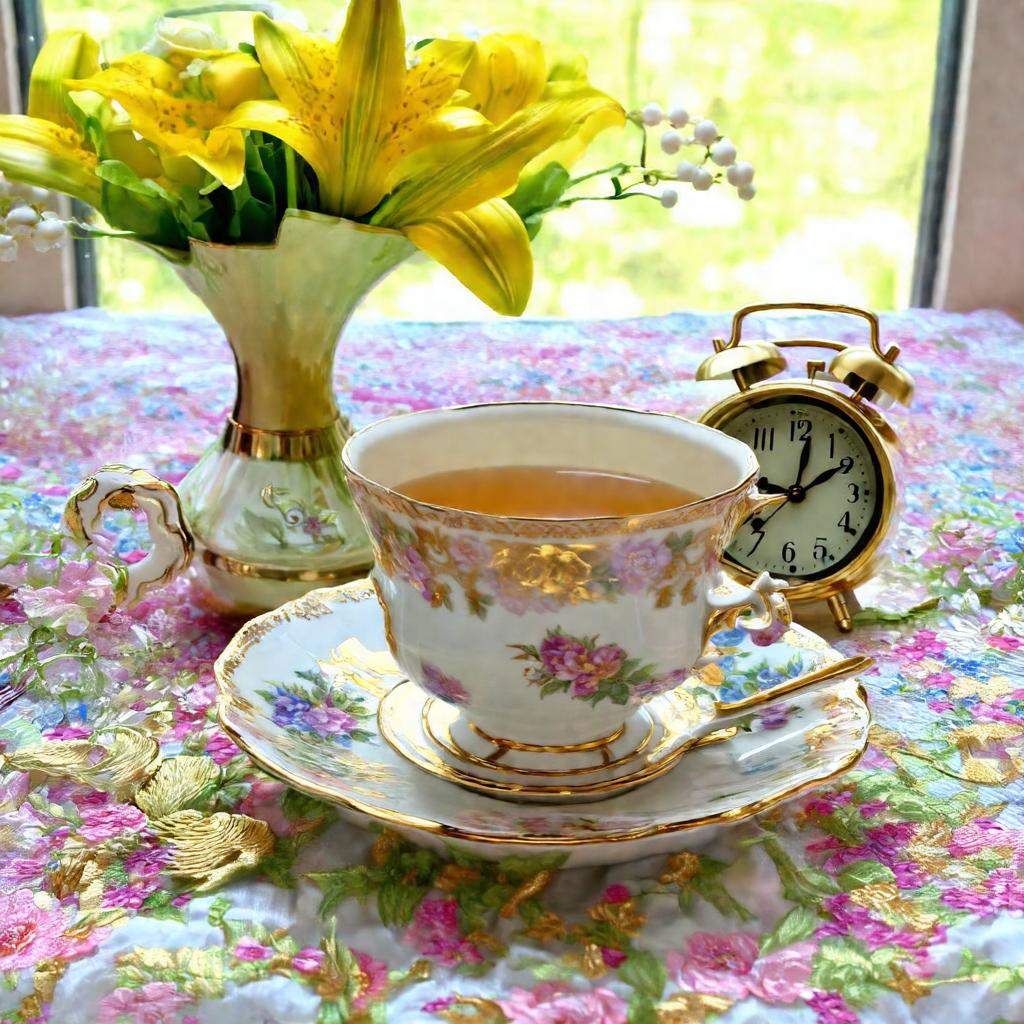}}%
        \fbox{\includegraphics[width=\mainimgwidth]{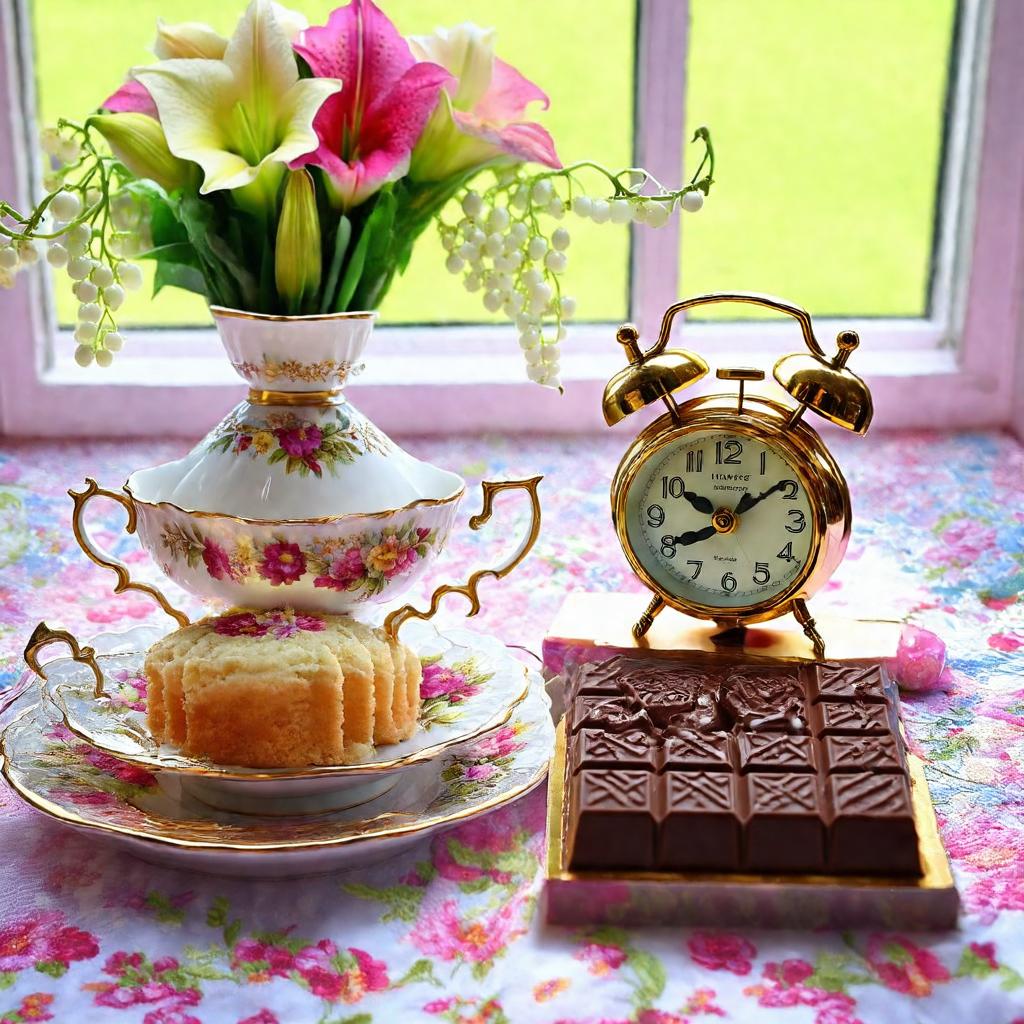}}%
        \fbox{\includegraphics[width=\mainimgwidth]{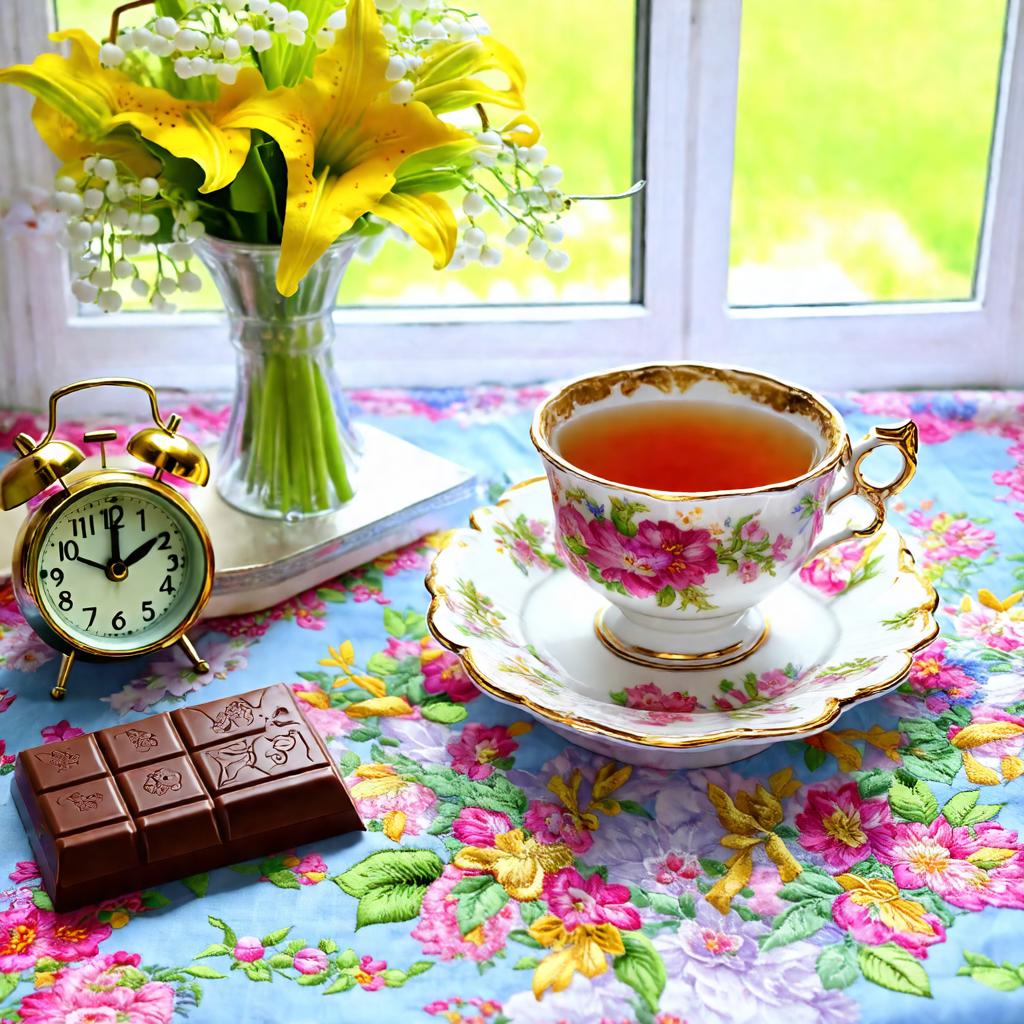}}\\[0.5ex]
        \hfill
        \vspace{-18pt}
        \caption*{
            \begin{minipage}{\maincapwidth}
            \centering
                \tiny{Prompt: \textit{Porcelain antique Cup of tea, multicolored cakes, samovar, a bouquet of lilies of the valley, a chocolate bar, an alarm clock, the sun, joy, a beautifully embroidered chintz tablecloth, the sun shines brightly outside the window}}
            \end{minipage}
        }
    \end{minipage}
    \hfill
    \begin{minipage}[t]{0.495\textwidth}
        \centering
        \fbox{\includegraphics[width=\mainimgwidth]{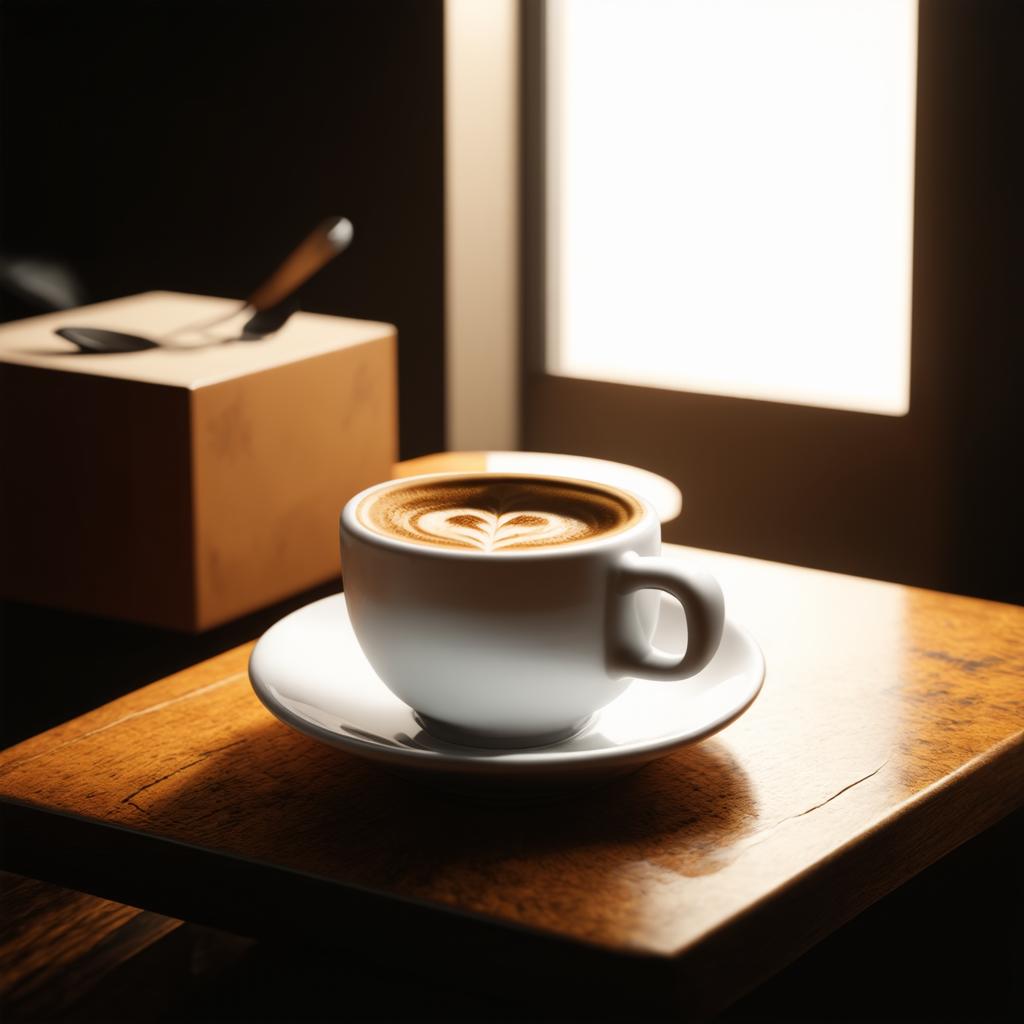}}
        \hfill
        \fbox{\includegraphics[width=\mainimgwidth]{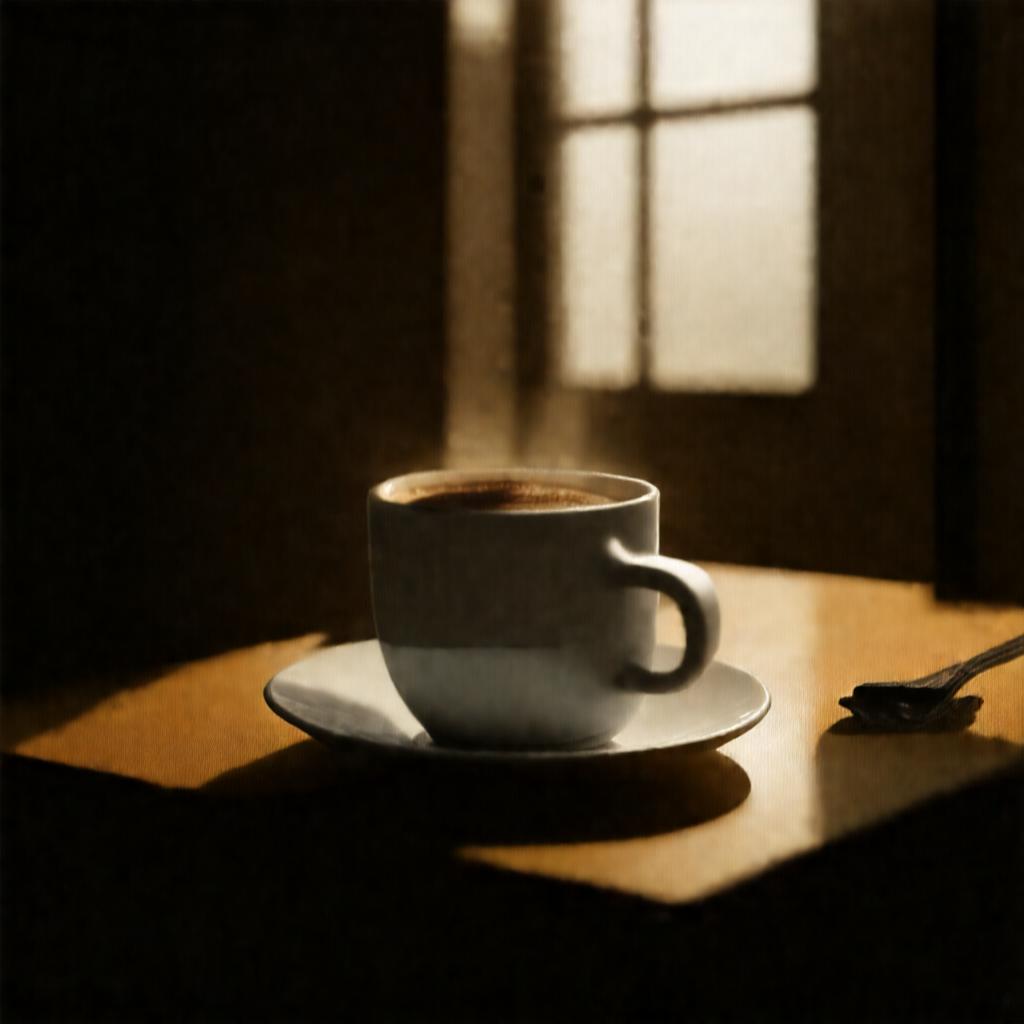}}%
        \fbox{\includegraphics[width=\mainimgwidth]{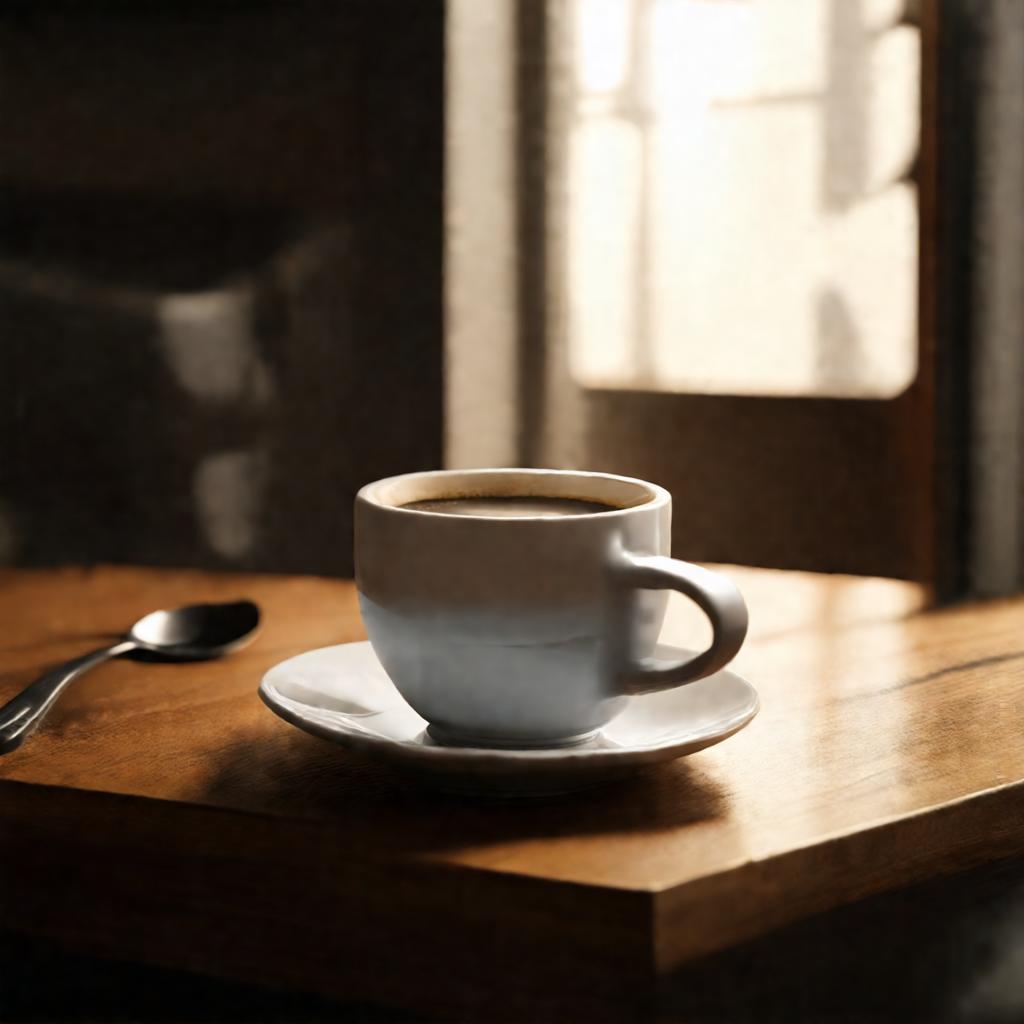}}%
        \fbox{\includegraphics[width=\mainimgwidth]{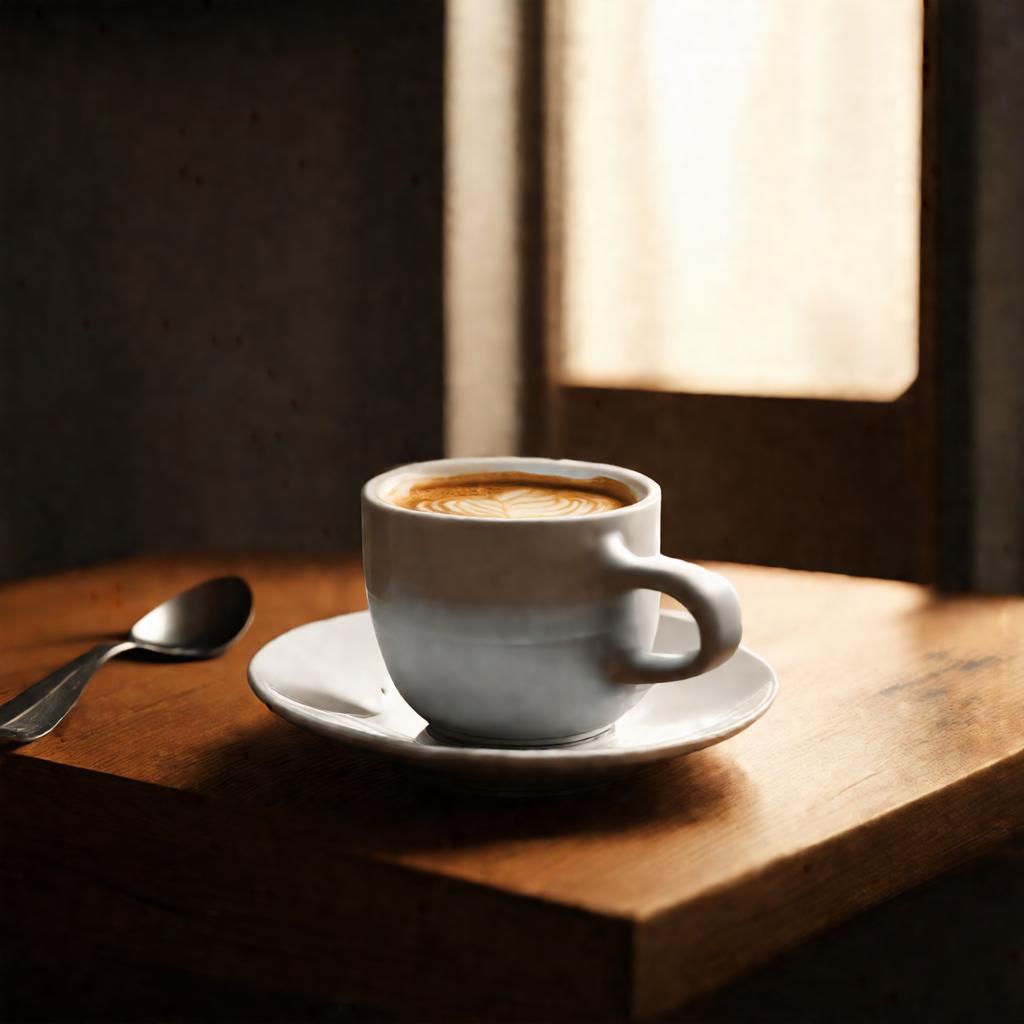}}%
        \fbox{\includegraphics[width=\mainimgwidth]{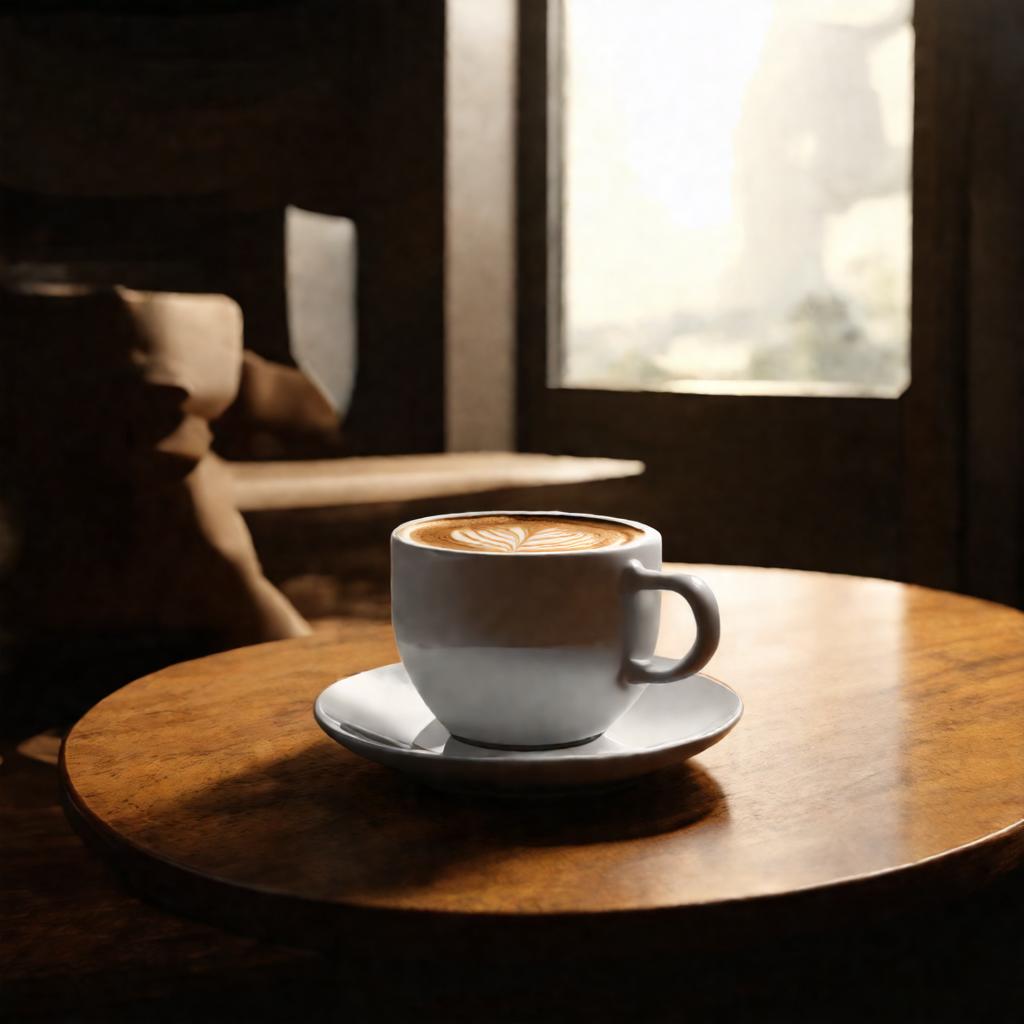}}%
        \fbox{\includegraphics[width=\mainimgwidth]{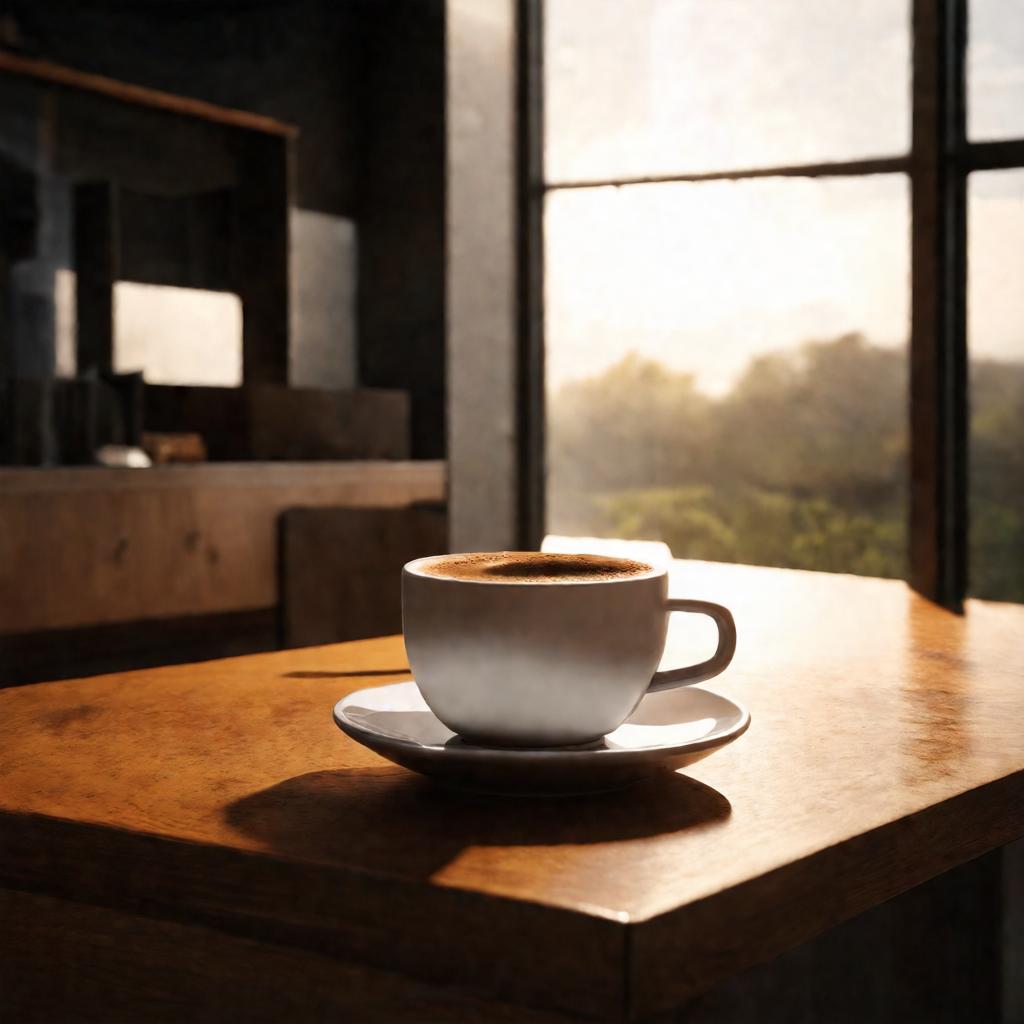}}%
        \fbox{\includegraphics[width=\mainimgwidth]{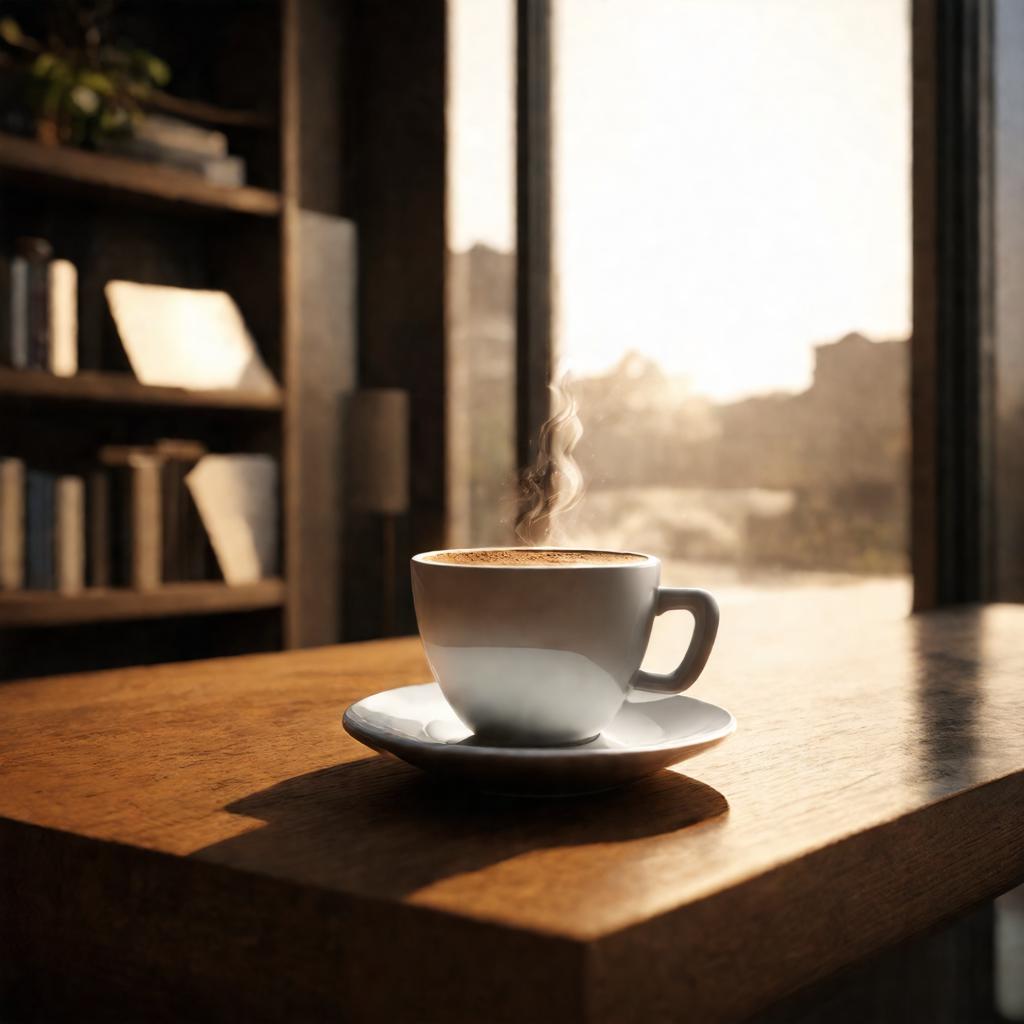}}\\[0.5ex]
        \hfill
        \vspace{-18pt}
        \caption*{
            \begin{minipage}{\maincapwidth}
            \centering
                \tiny{Prompt: \textit{morning coffee, morning light, unreal engine}}
            \end{minipage}
        }
    \end{minipage}

    \begin{minipage}[t]{0.495\textwidth}
        \centering
        \fbox{\includegraphics[width=\mainimgwidth]{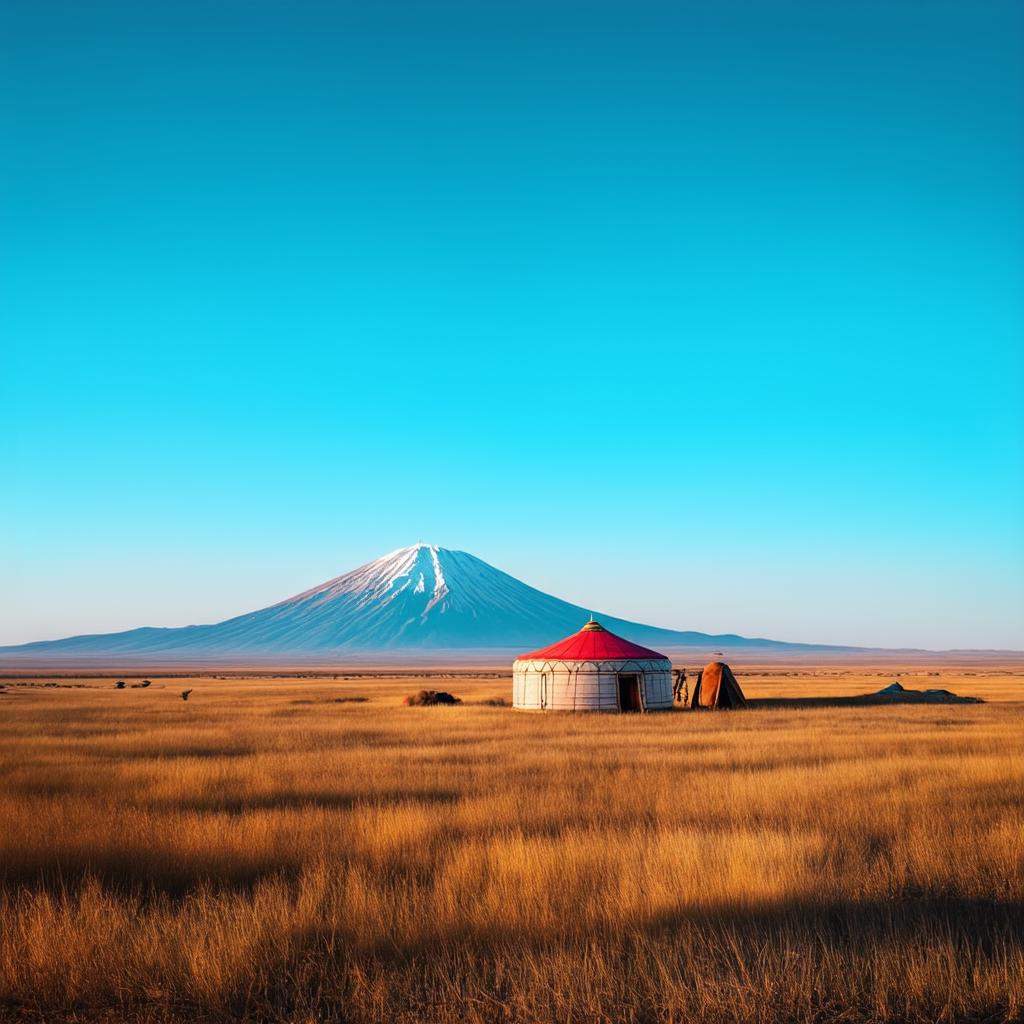}}
        \hfill
        \fbox{\includegraphics[width=\mainimgwidth]{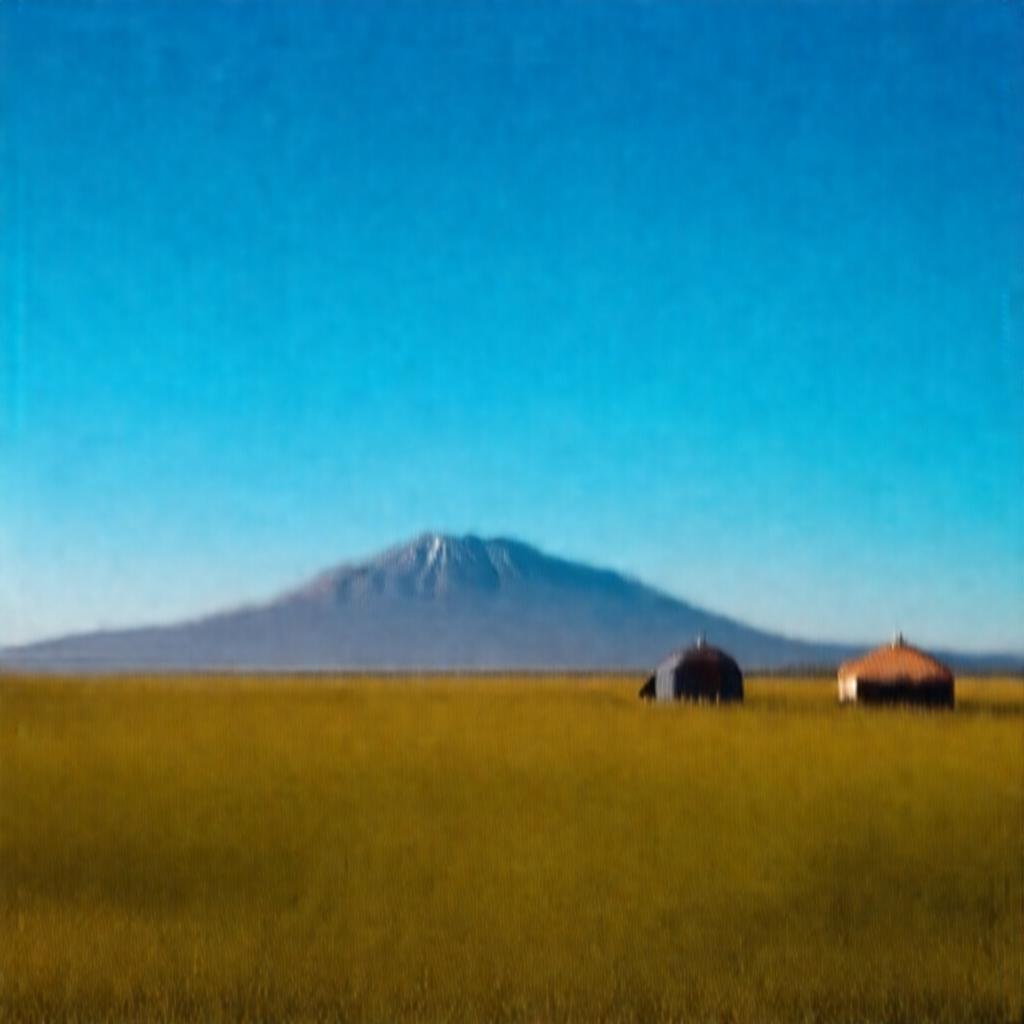}}%
        \fbox{\includegraphics[width=\mainimgwidth]{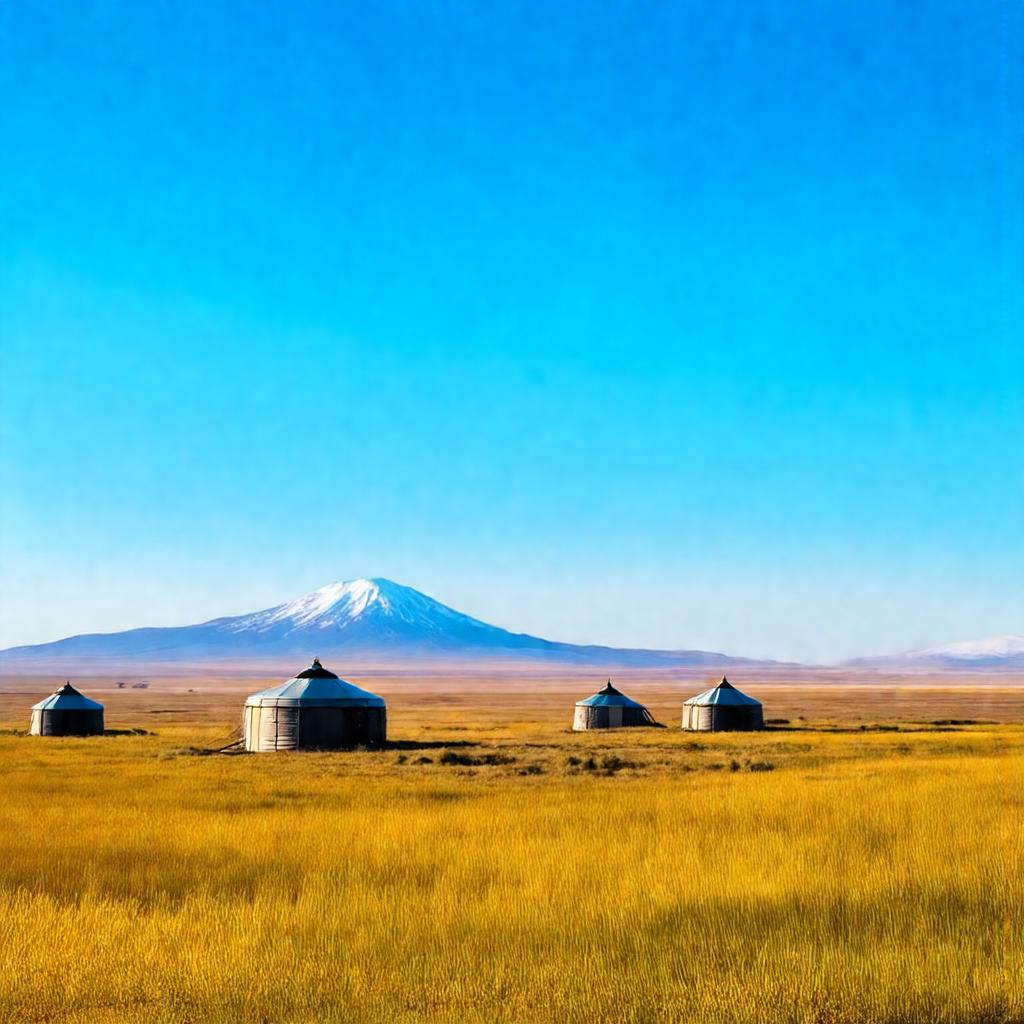}}%
        \fbox{\includegraphics[width=\mainimgwidth]{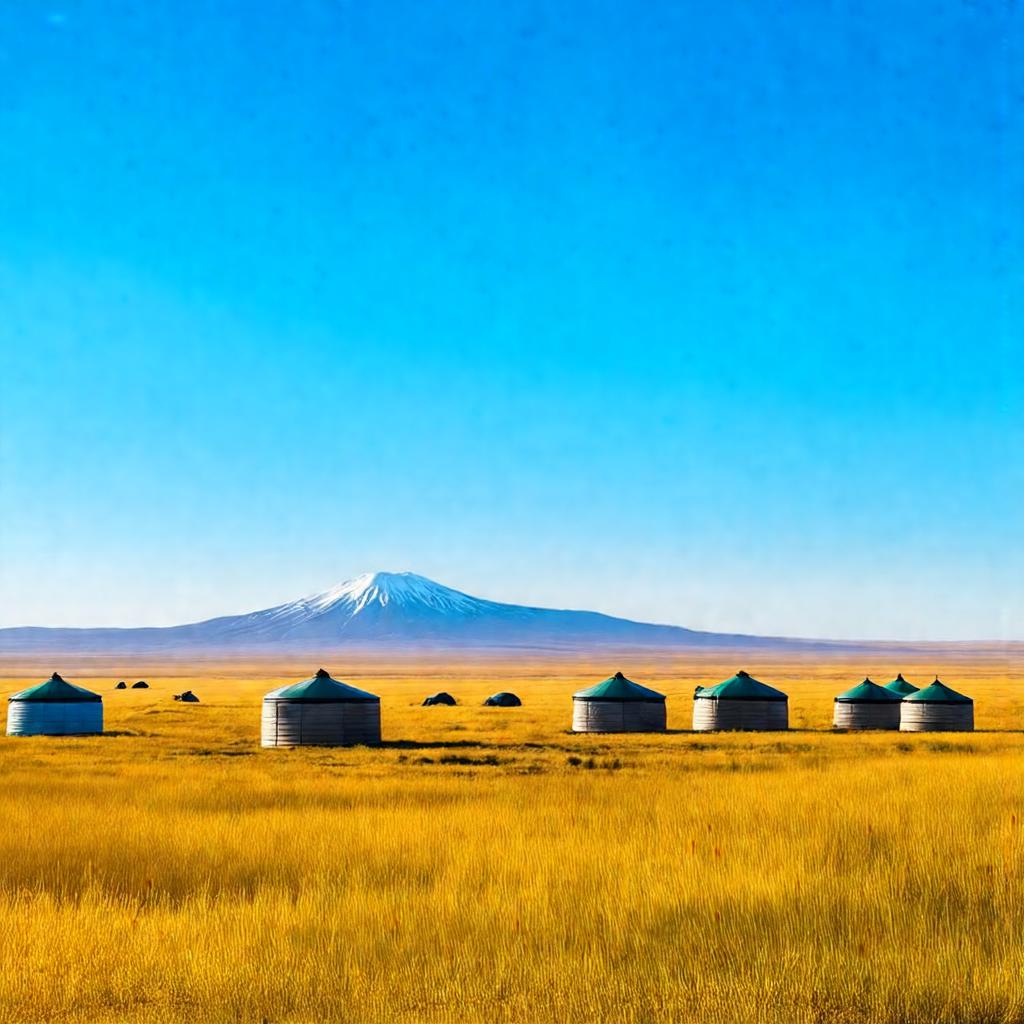}}%
        \fbox{\includegraphics[width=\mainimgwidth]{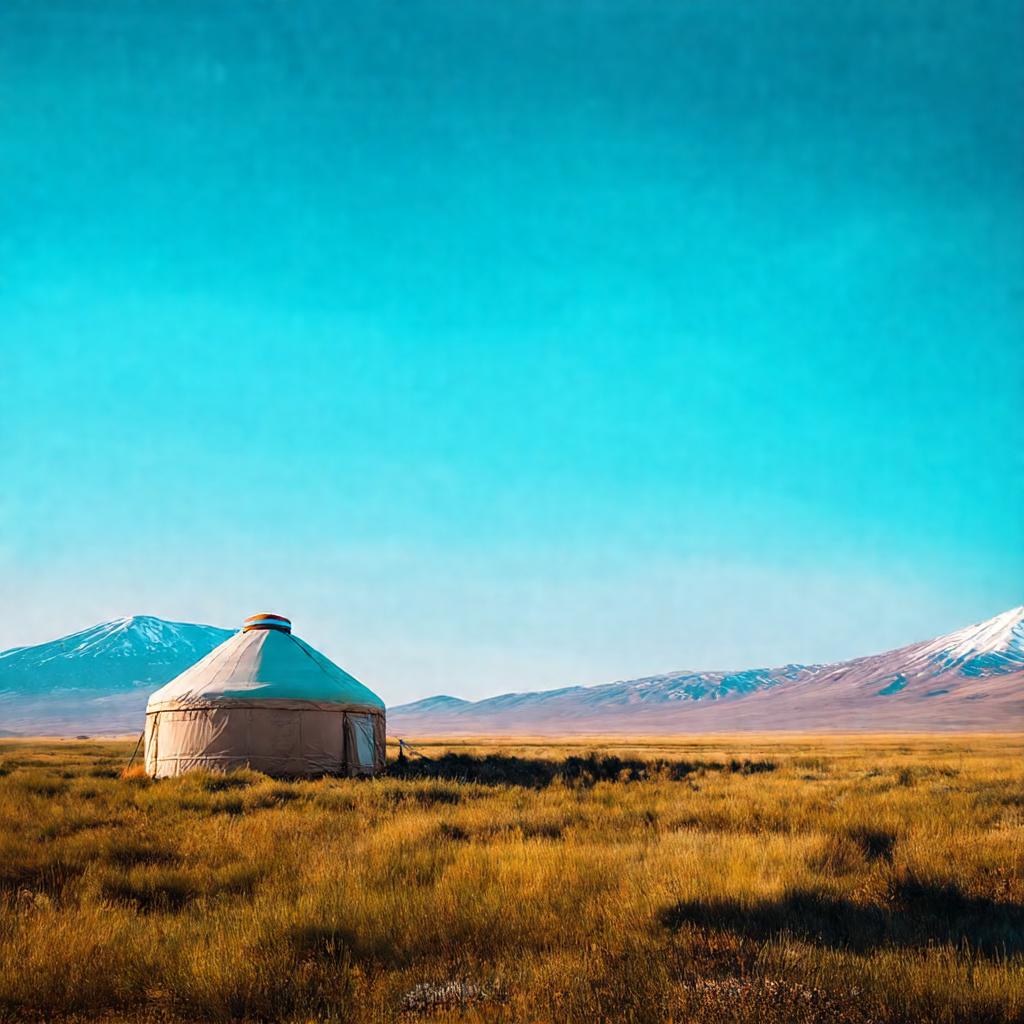}}%
        \fbox{\includegraphics[width=\mainimgwidth]{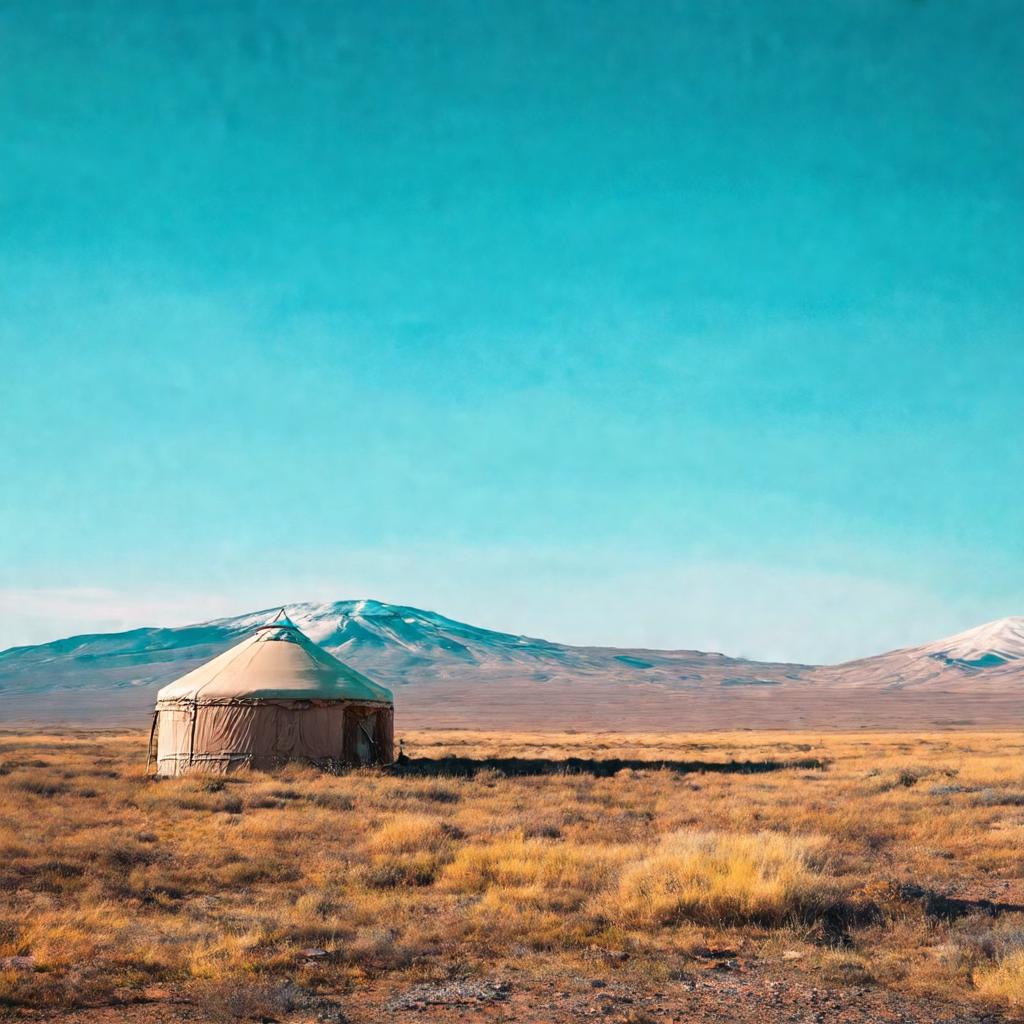}}%
        \fbox{\includegraphics[width=\mainimgwidth]{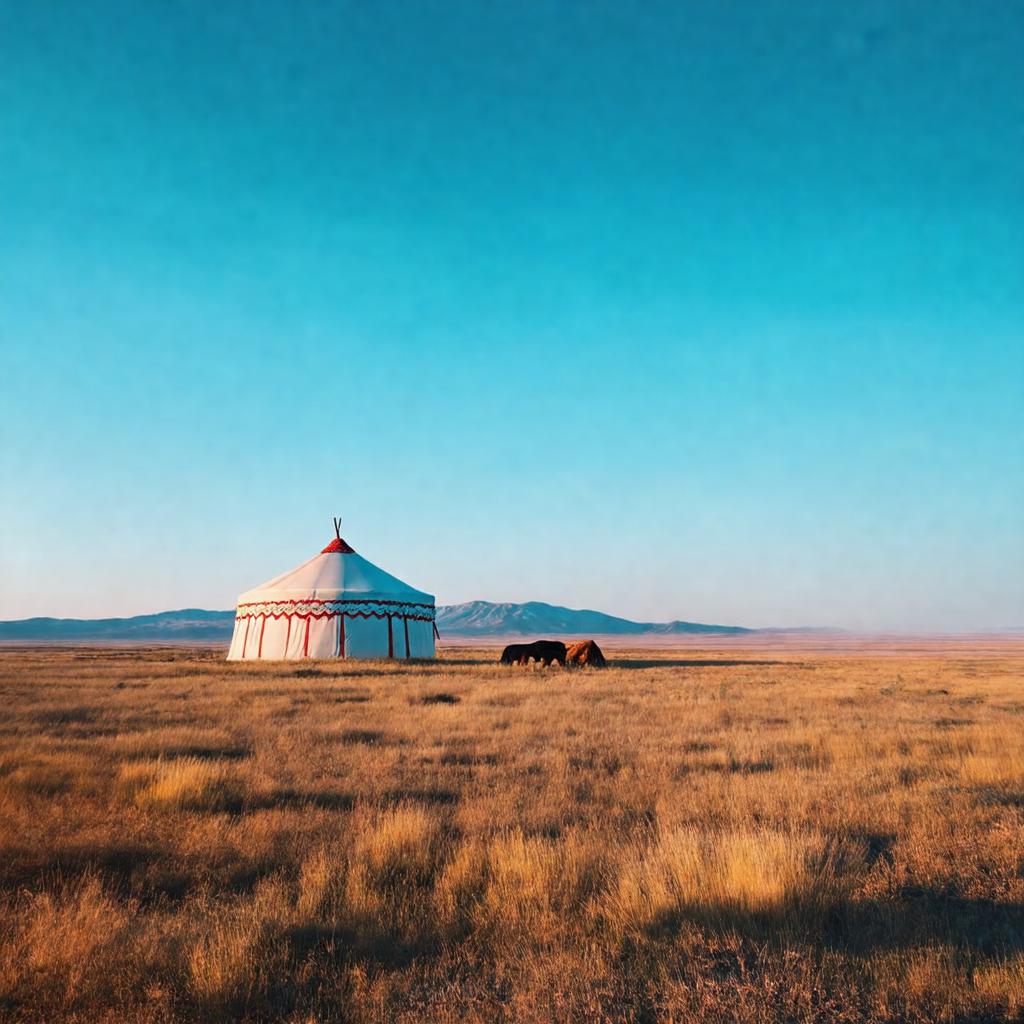}}\\[0.5ex]
        \hfill
        \vspace{-18pt}
        \caption*{
            \begin{minipage}{\maincapwidth}
            \centering
                \tiny{Prompt: \textit{Mongolian nature with blue sky, mountain, steppe, Mongolian traditional yurt, hyperrealistic, photorealistic, cinematic colour grading, cinematic frame composition, 8K, shot on konica centuria film Leica M6}}
            \end{minipage}
        }
    \end{minipage}
    \hfill
    \begin{minipage}[t]{0.495\textwidth}
        \centering
        \fbox{\includegraphics[width=\mainimgwidth]{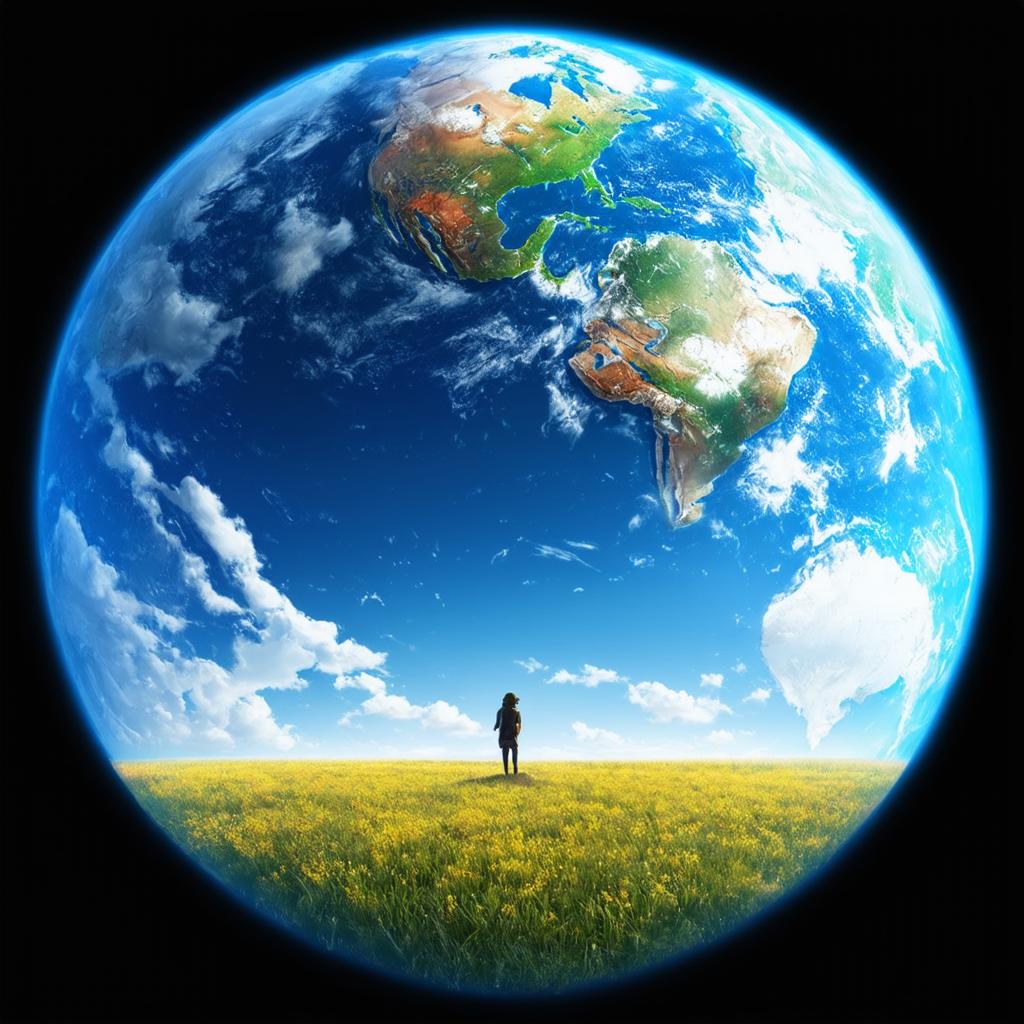}}
        \hfill
        \fbox{\includegraphics[width=\mainimgwidth]{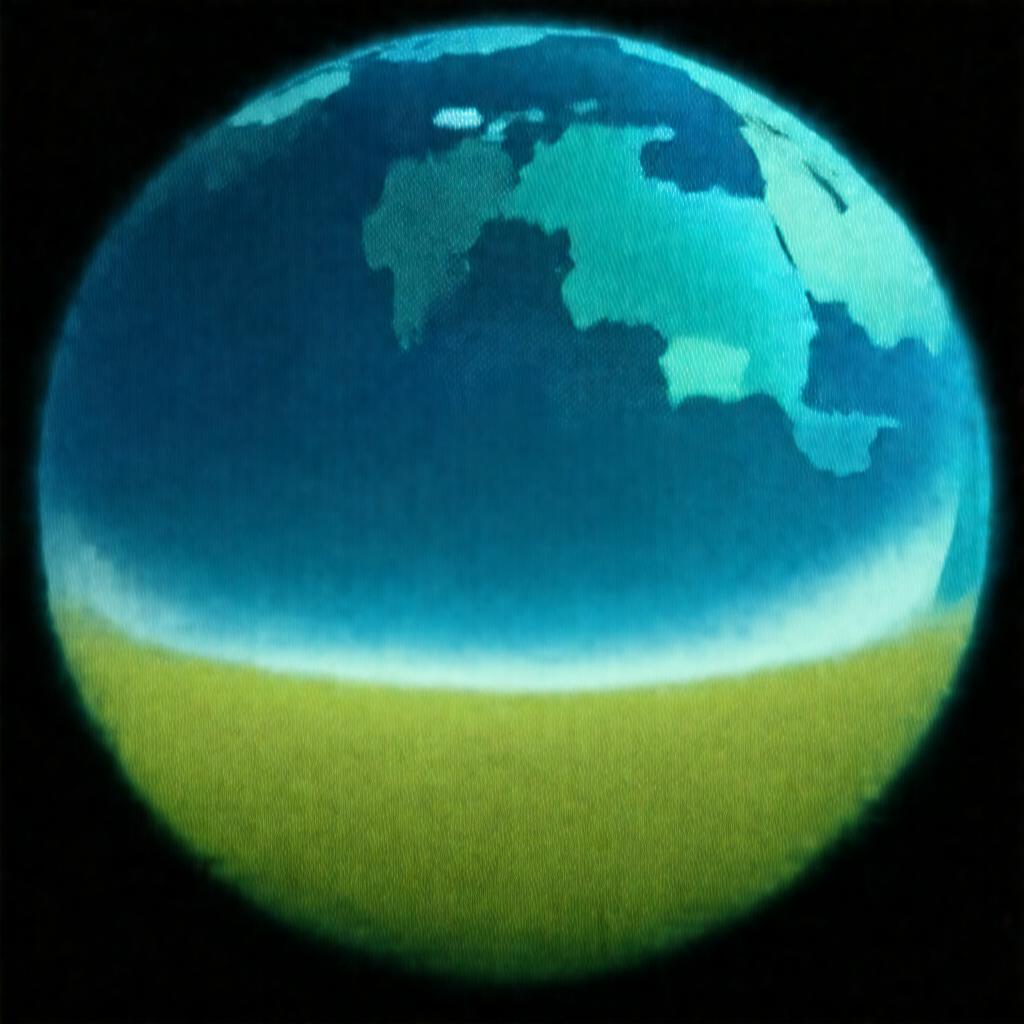}}%
        \fbox{\includegraphics[width=\mainimgwidth]{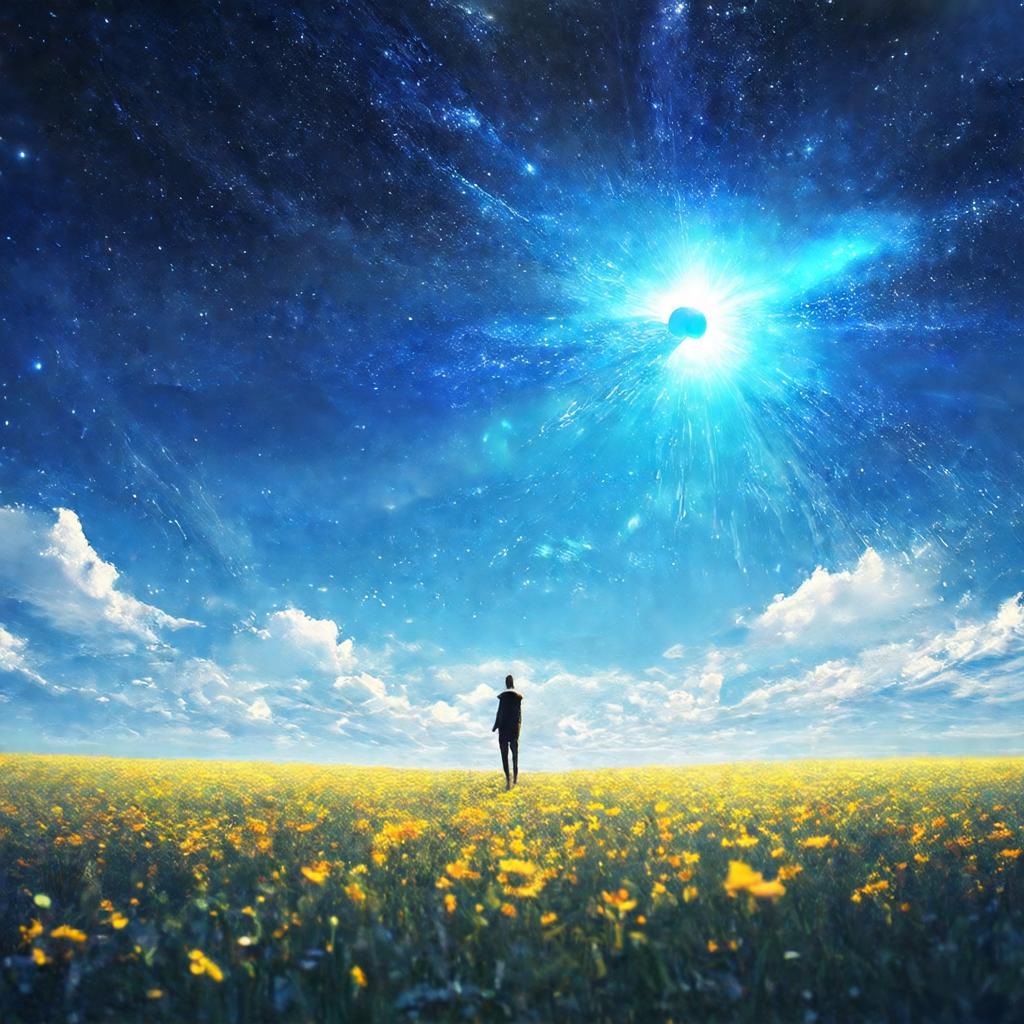}}%
        \fbox{\includegraphics[width=\mainimgwidth]{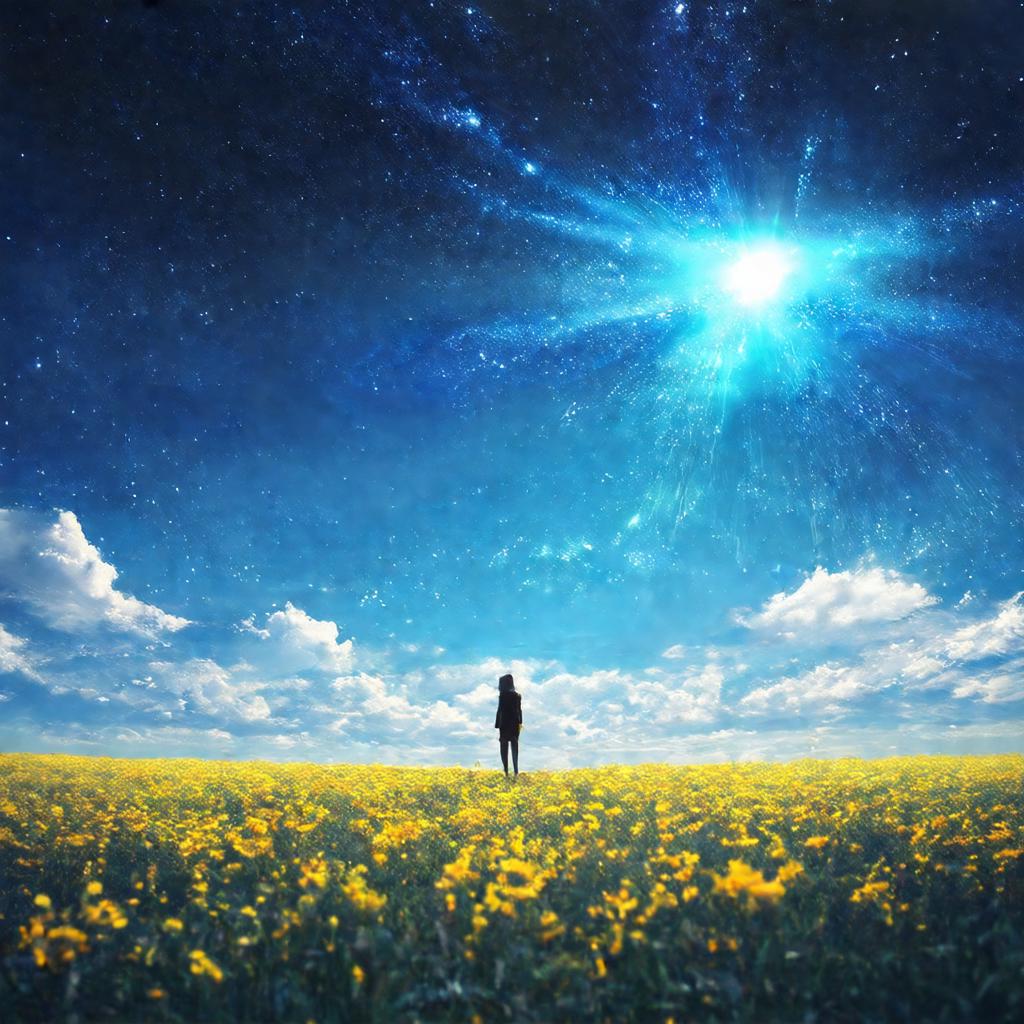}}%
        \fbox{\includegraphics[width=\mainimgwidth]{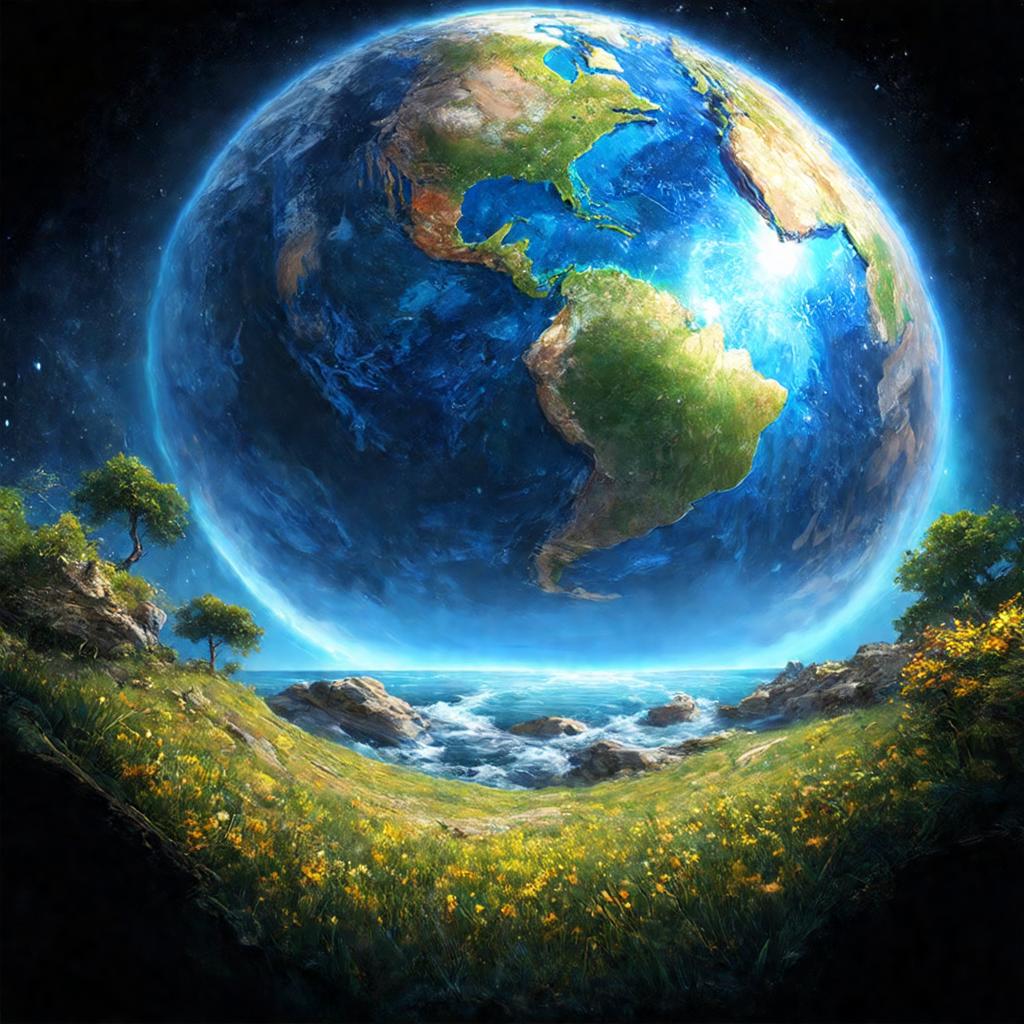}}%
        \fbox{\includegraphics[width=\mainimgwidth]{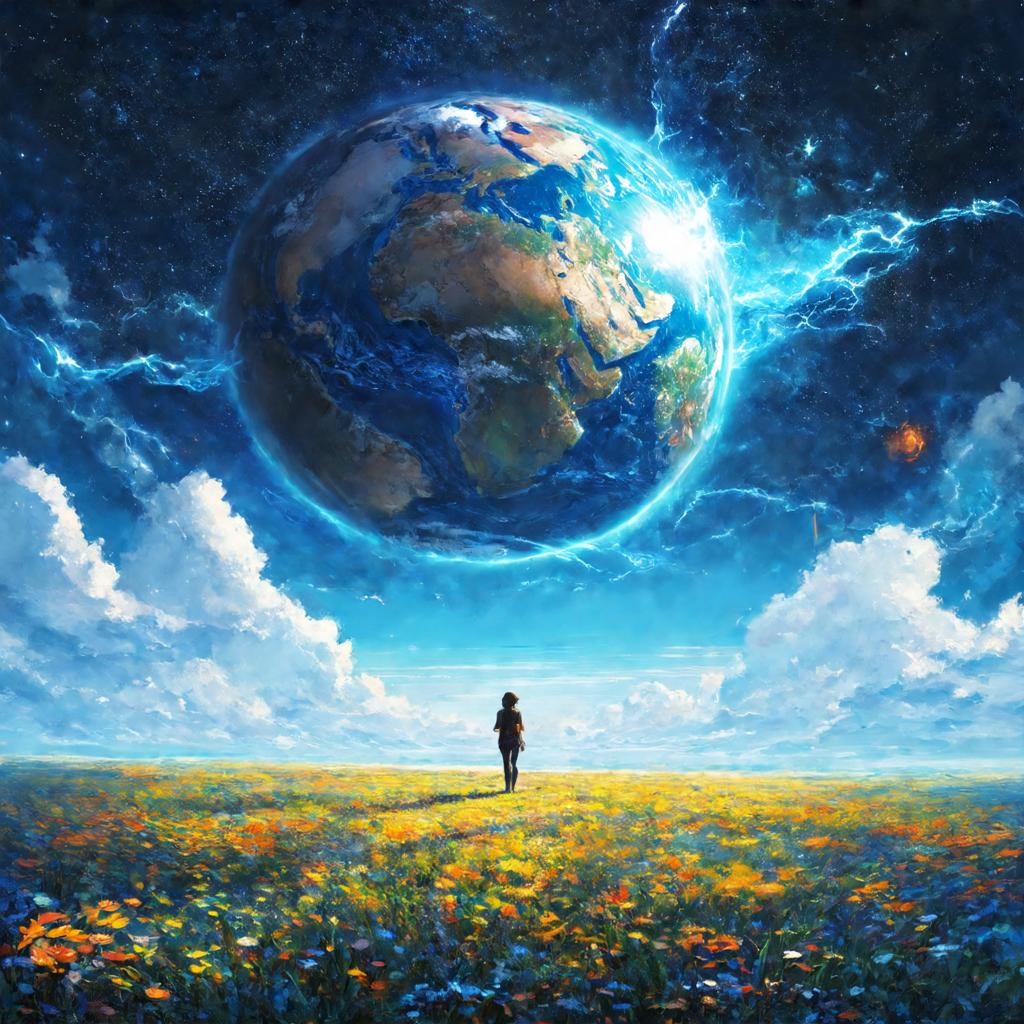}}%
        \fbox{\includegraphics[width=\mainimgwidth]{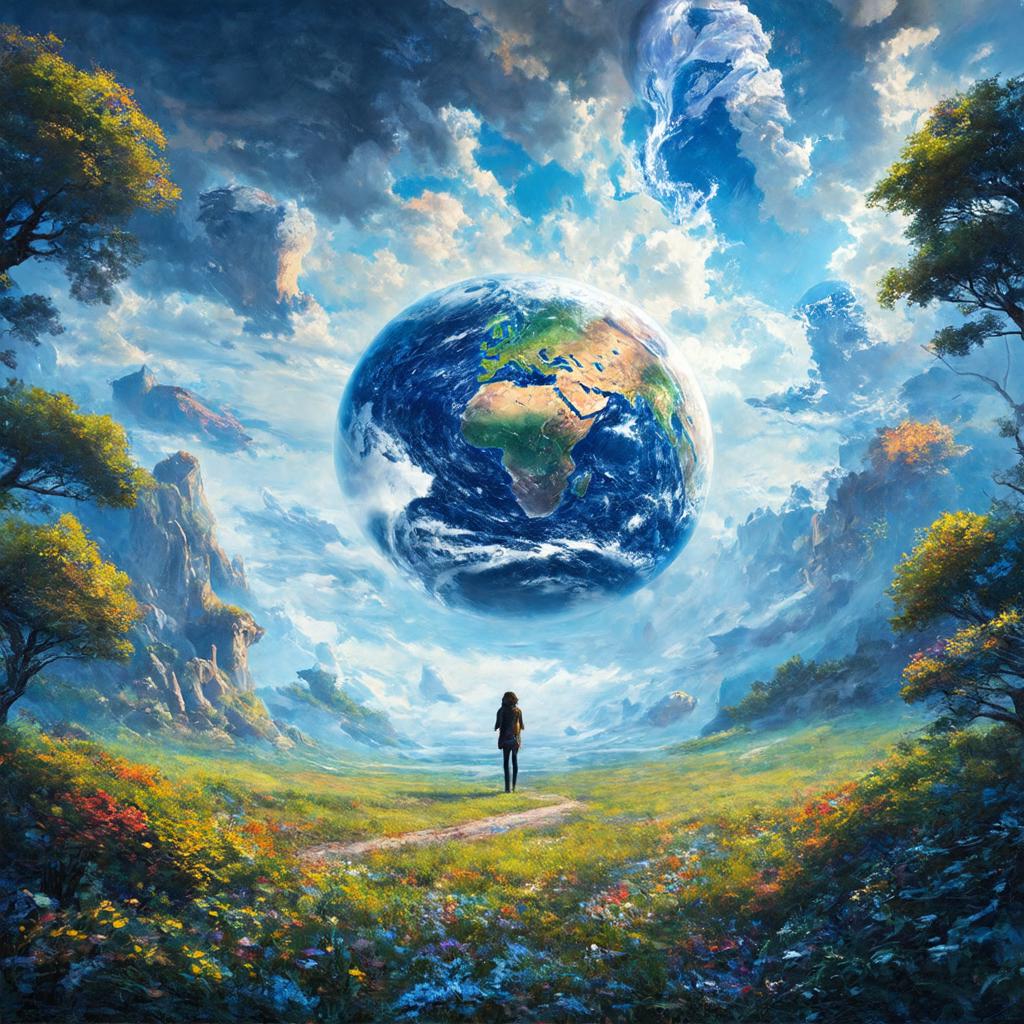}}\\[0.5ex]
        \hfill
        \vspace{-18pt}
        \caption*{
            \begin{minipage}{\maincapwidth}
            \centering
                \tiny{Prompt: \textit{This world, a part of infinity, Contains an endless variety. Yet it stands alone, not part of any, An enigma, a mystery.}}
            \end{minipage}
        }
    \end{minipage}

    \begin{minipage}[t]{0.495\textwidth}
        \centering
        \fbox{\includegraphics[width=\mainimgwidth]{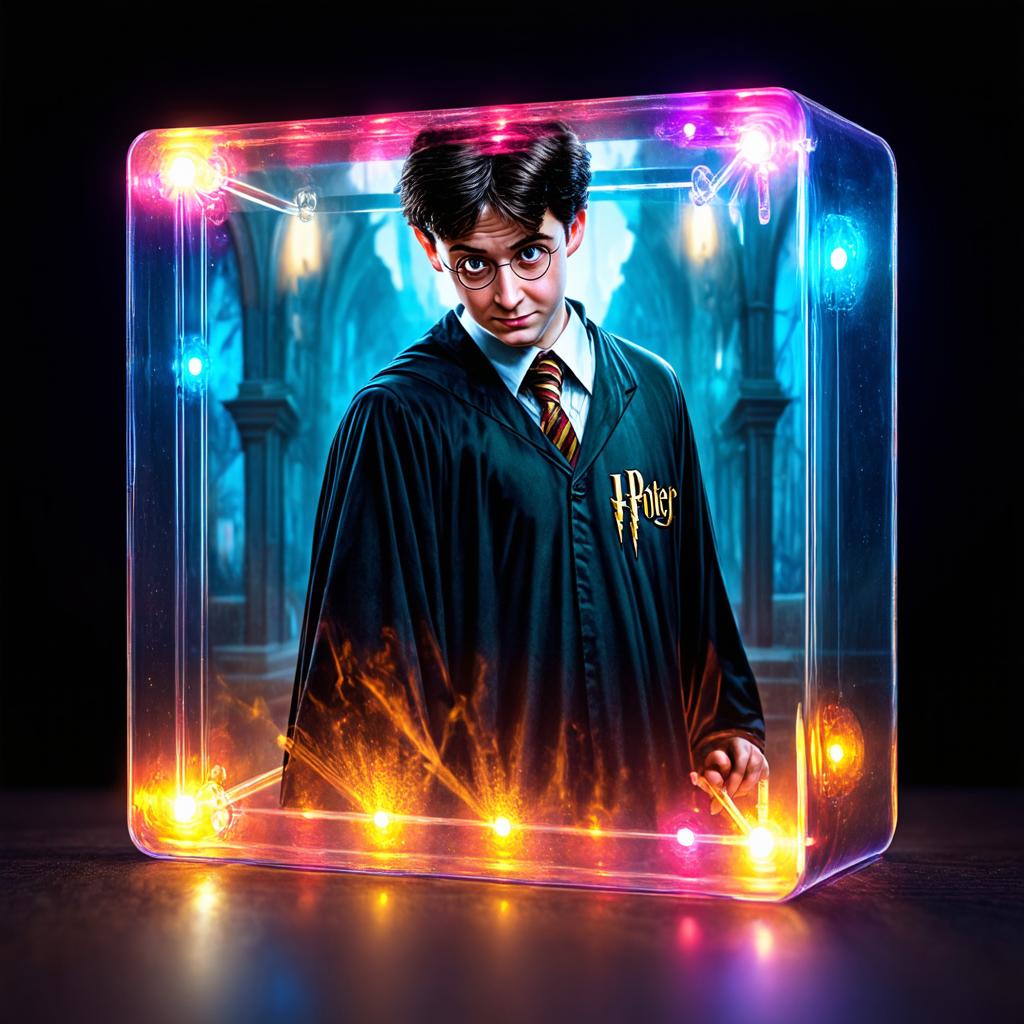}}
        \hfill
        \fbox{\includegraphics[width=\mainimgwidth]{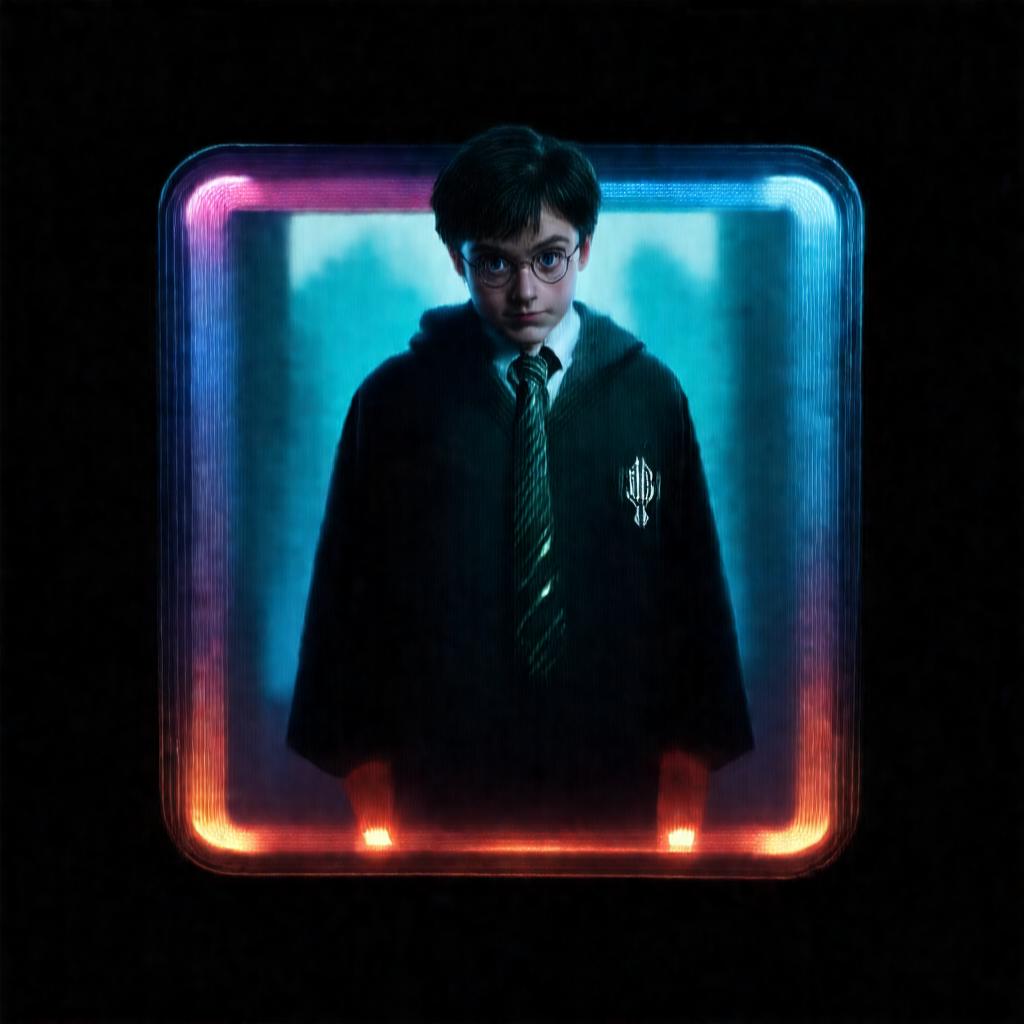}}%
        \fbox{\includegraphics[width=\mainimgwidth]{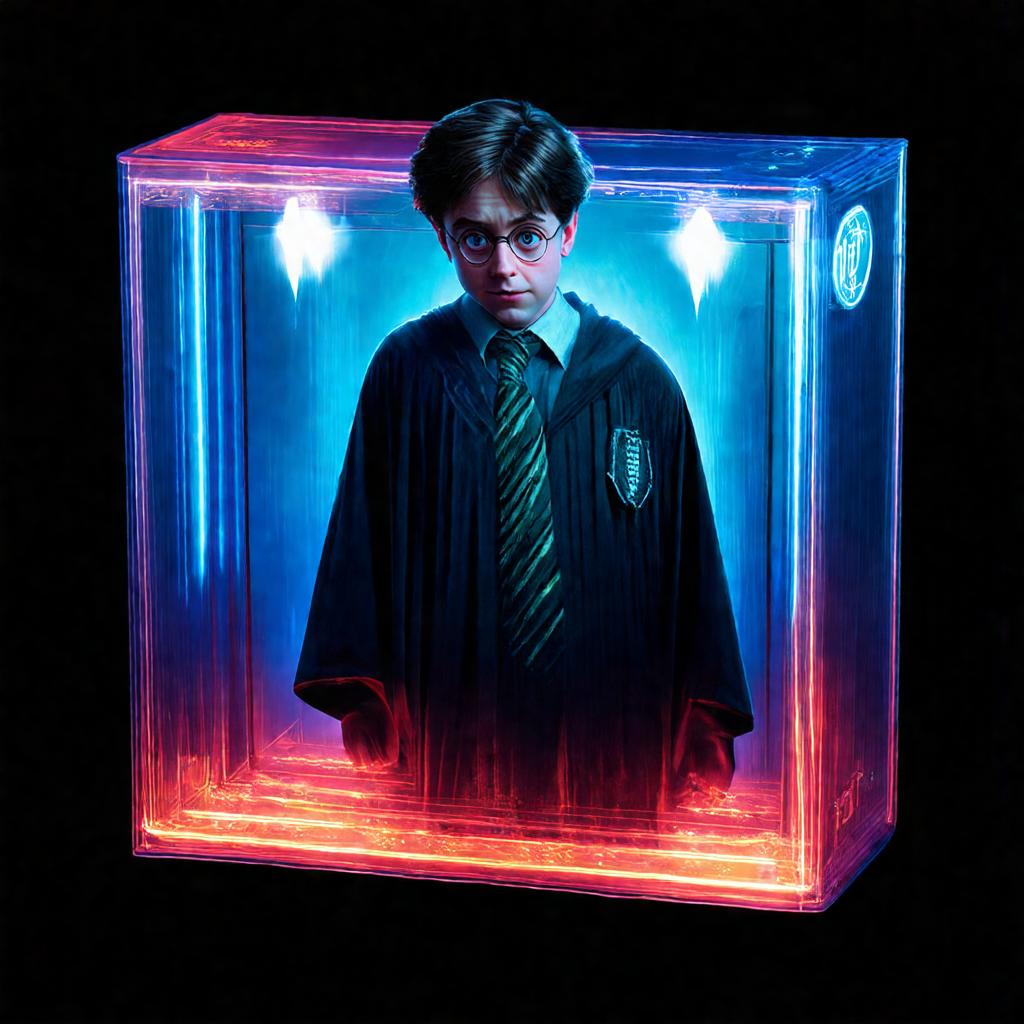}}%
        \fbox{\includegraphics[width=\mainimgwidth]{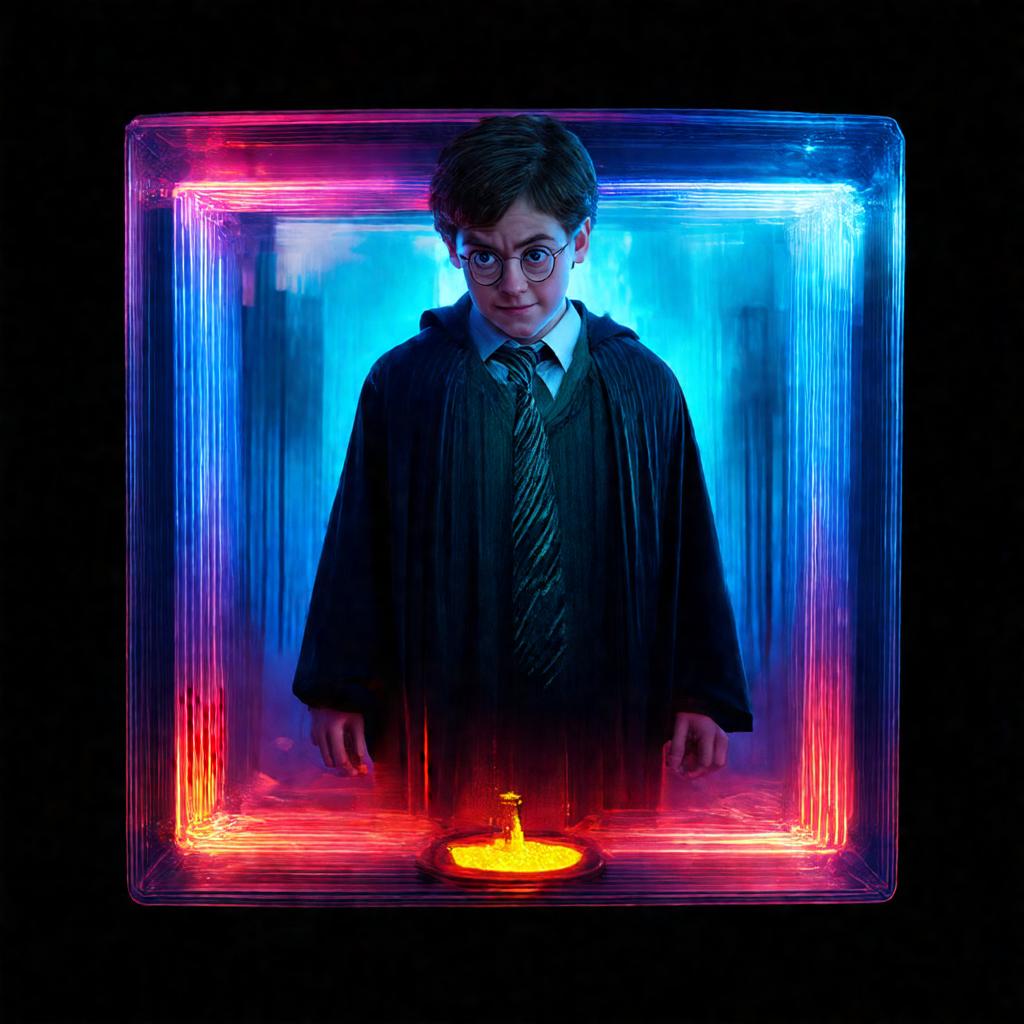}}%
        \fbox{\includegraphics[width=\mainimgwidth]{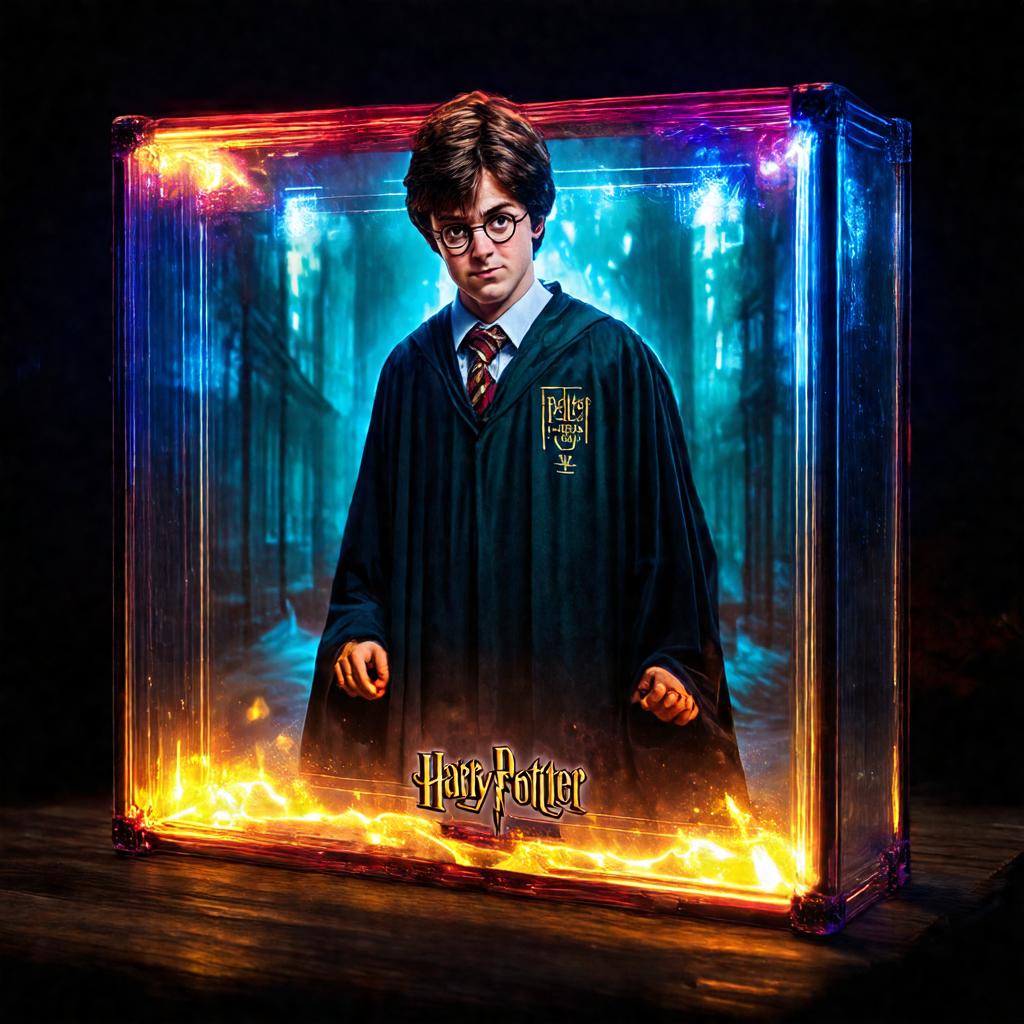}}%
        \fbox{\includegraphics[width=\mainimgwidth]{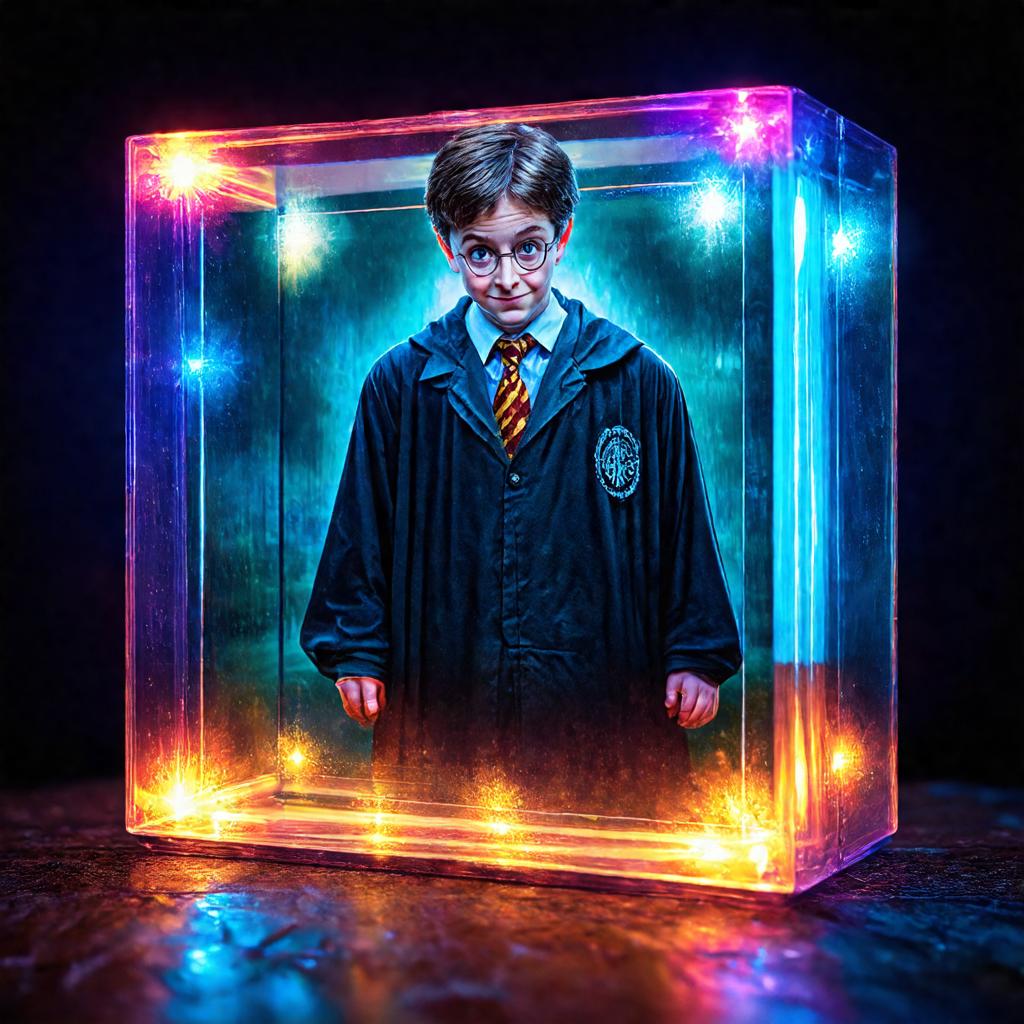}}%
        \fbox{\includegraphics[width=\mainimgwidth]{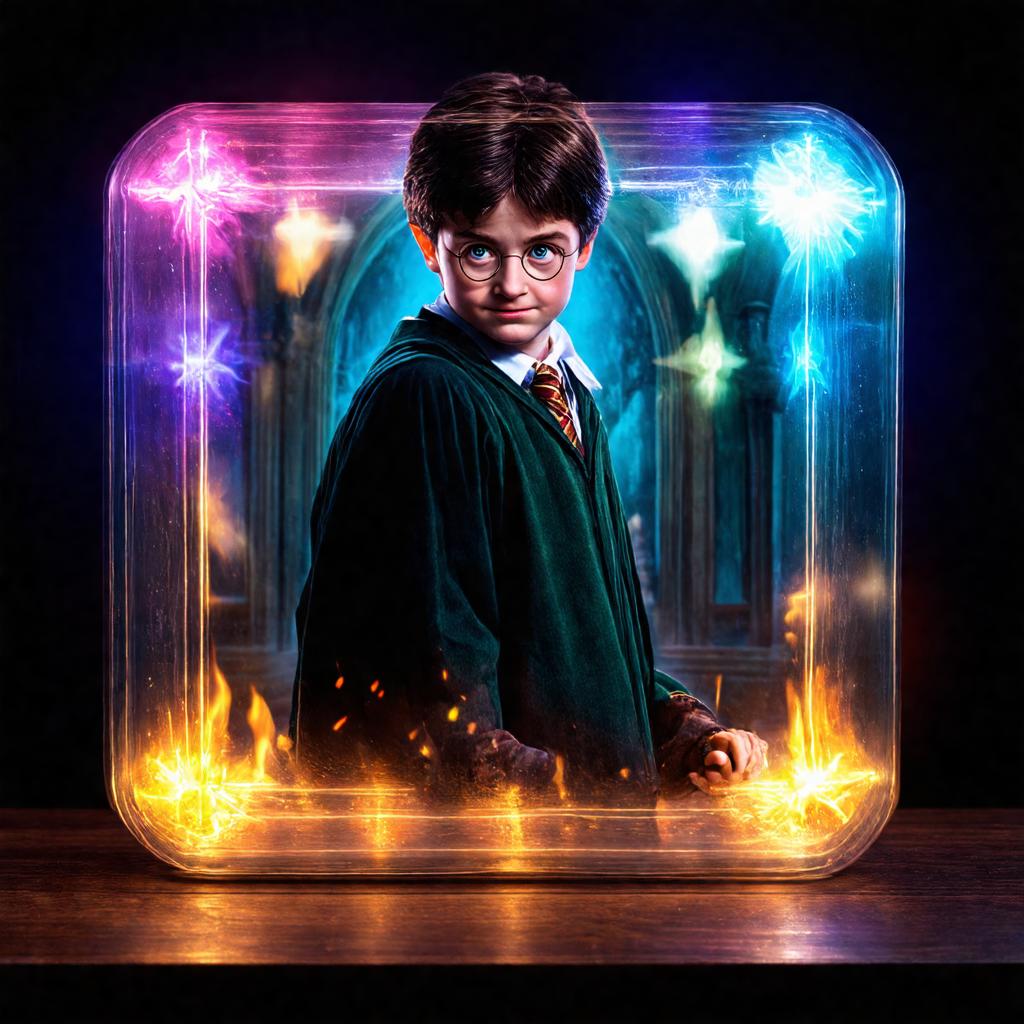}}\\[0.5ex]
        \hfill
        \vspace{-18pt}
        \caption*{
            \begin{minipage}{\maincapwidth}
            \centering
                \tiny{Prompt: \textit{hyperrealistic image of Harry Potter poster, lit by colored lights, in a case made in transparent plastic, full square picture, light background, extreme details, cinematic, vector, 8K resolution, HD, 34 view, high detail}}
            \end{minipage}
        }
    \end{minipage}
    \hfill
    \begin{minipage}[t]{0.495\textwidth}
        \centering
        \fbox{\includegraphics[width=\mainimgwidth]{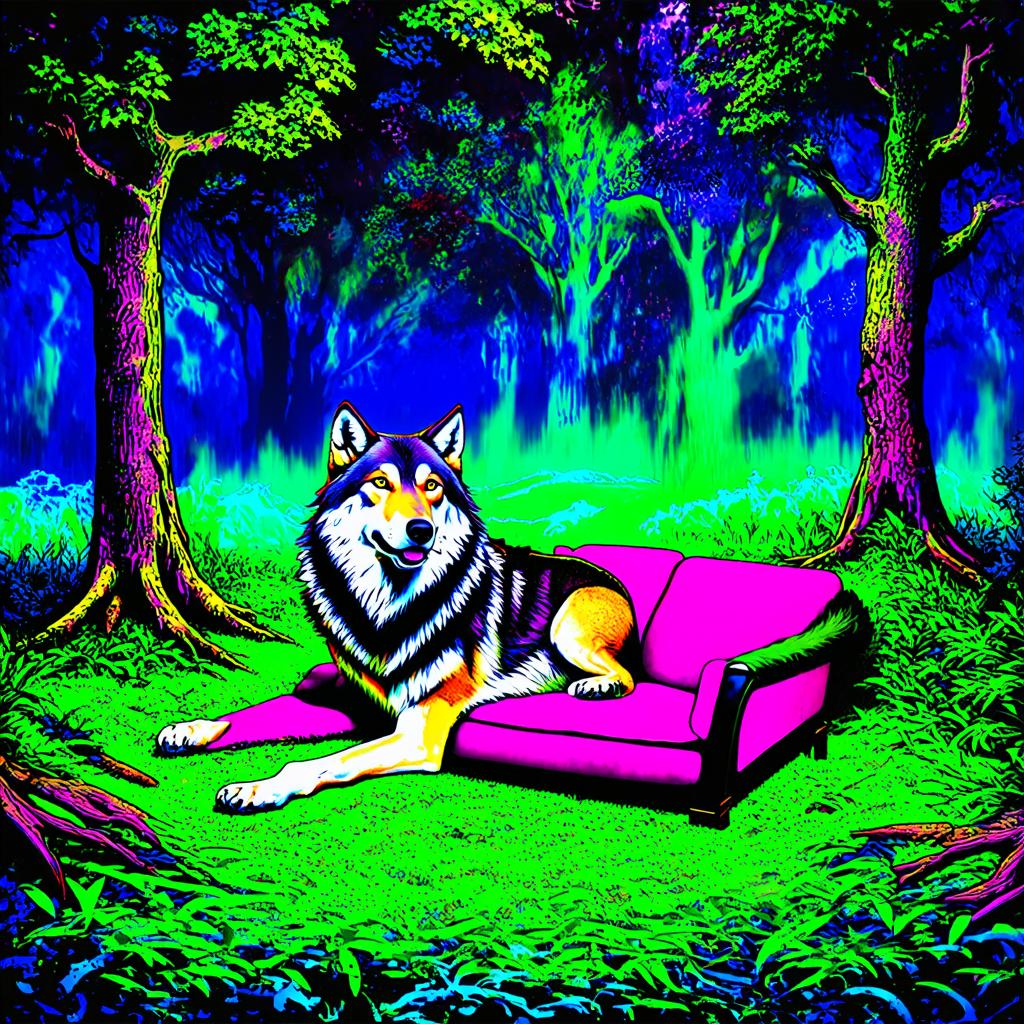}}
        \hfill
        \fbox{\includegraphics[width=\mainimgwidth]{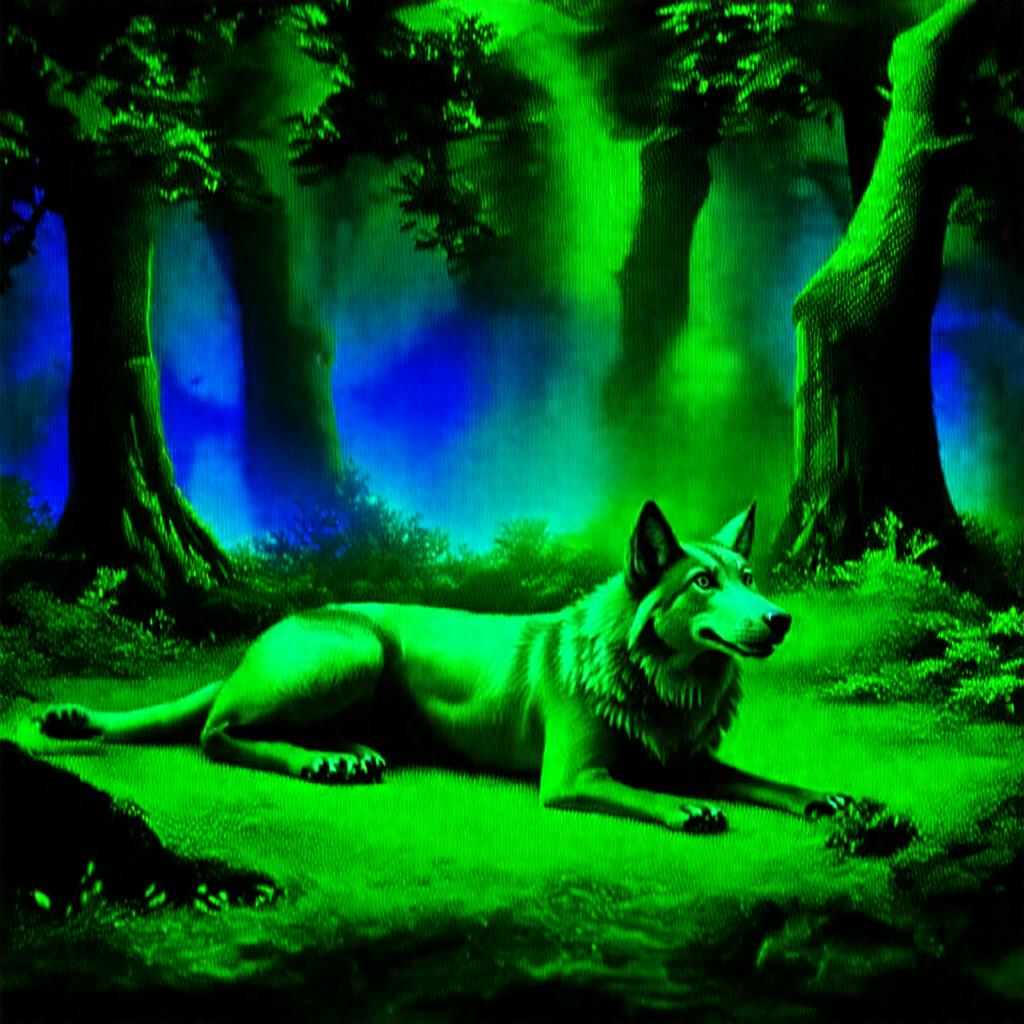}}%
        \fbox{\includegraphics[width=\mainimgwidth]{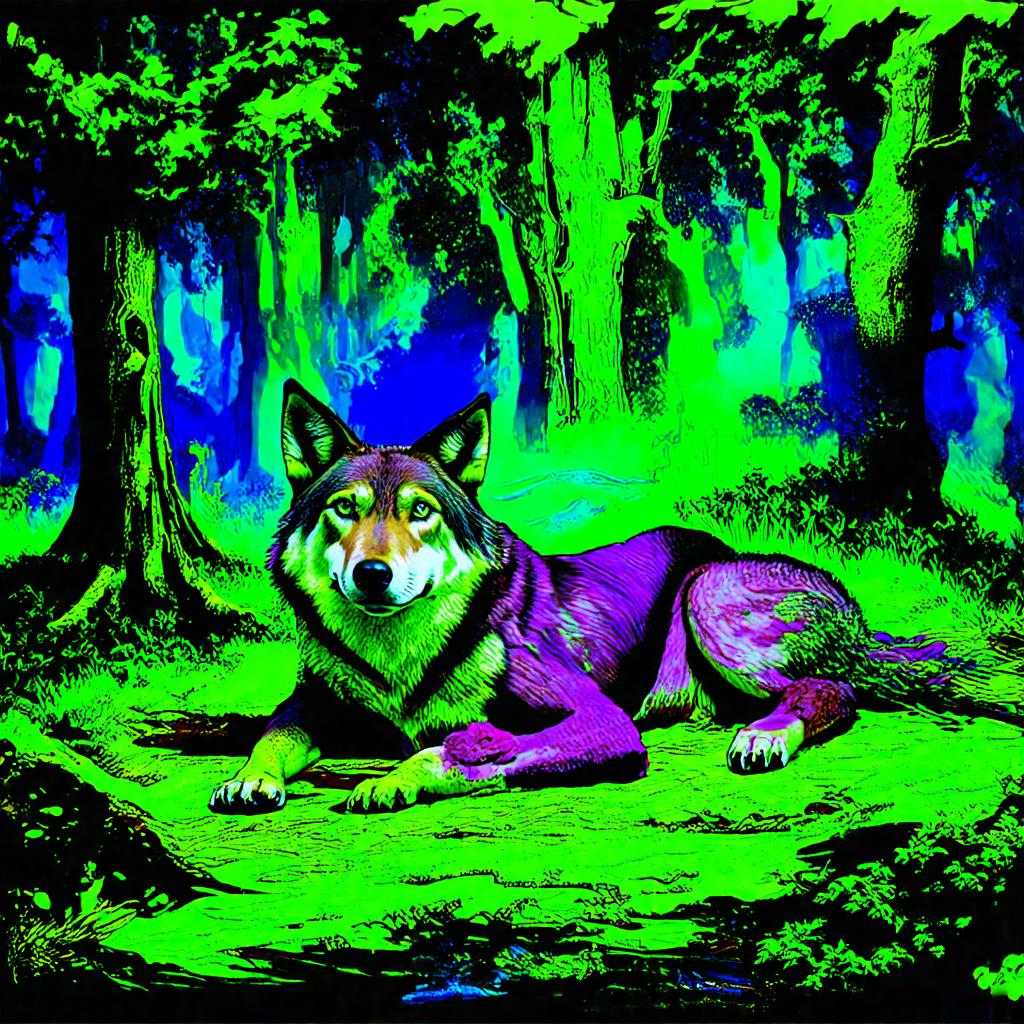}}%
        \fbox{\includegraphics[width=\mainimgwidth]{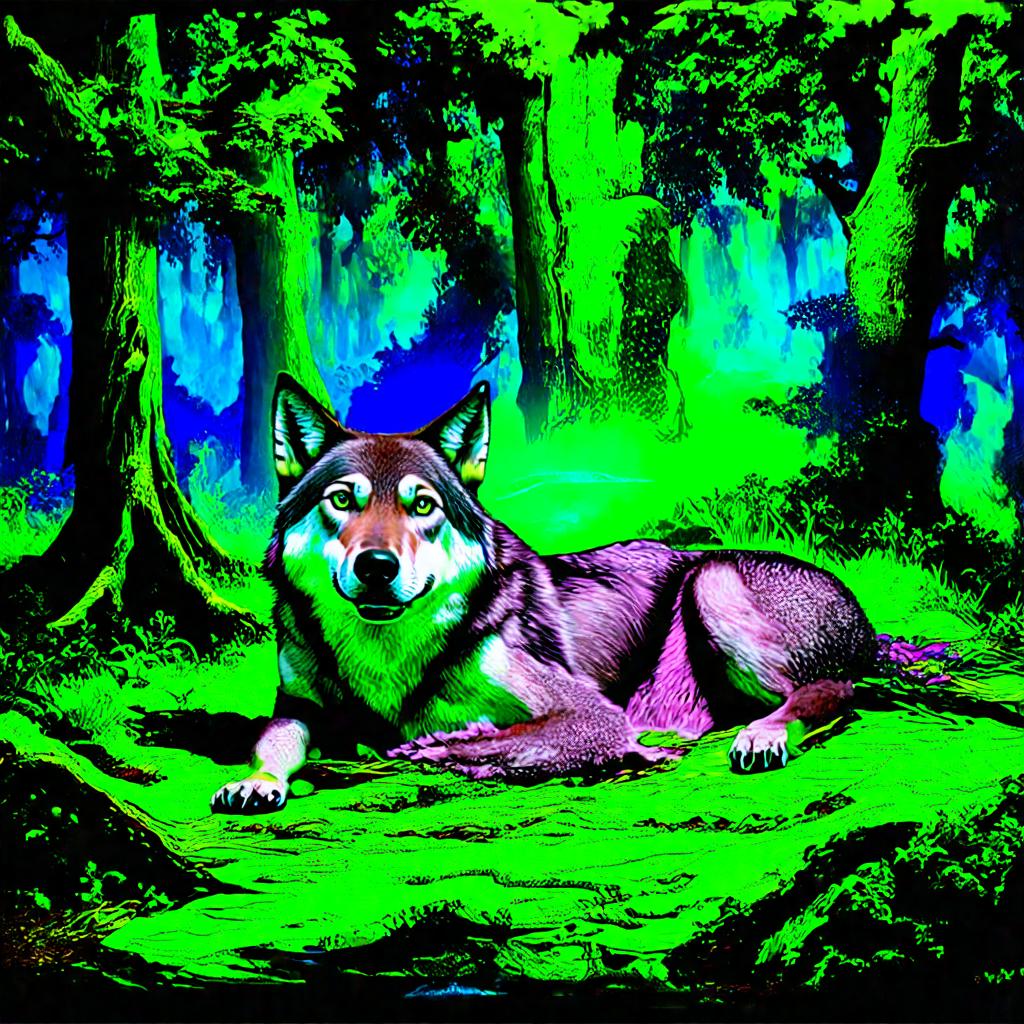}}%
        \fbox{\includegraphics[width=\mainimgwidth]{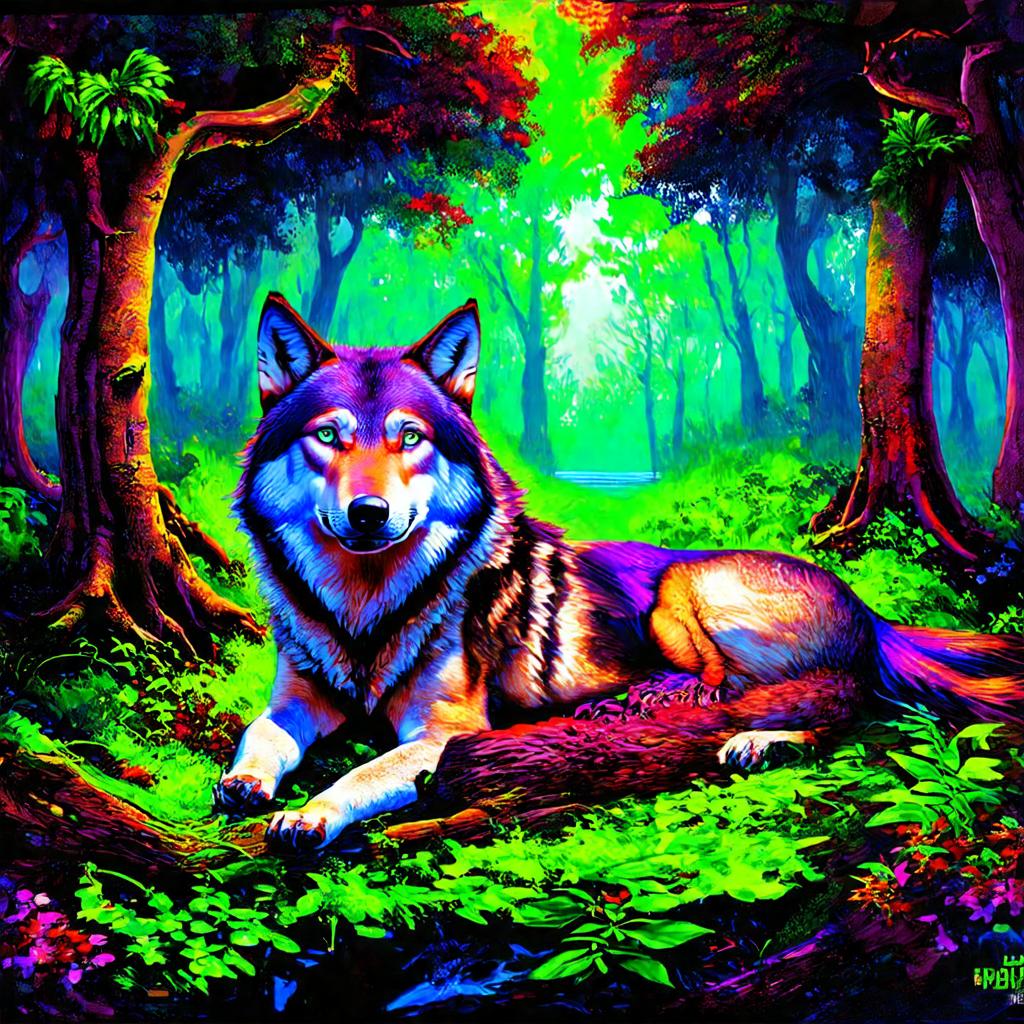}}%
        \fbox{\includegraphics[width=\mainimgwidth]{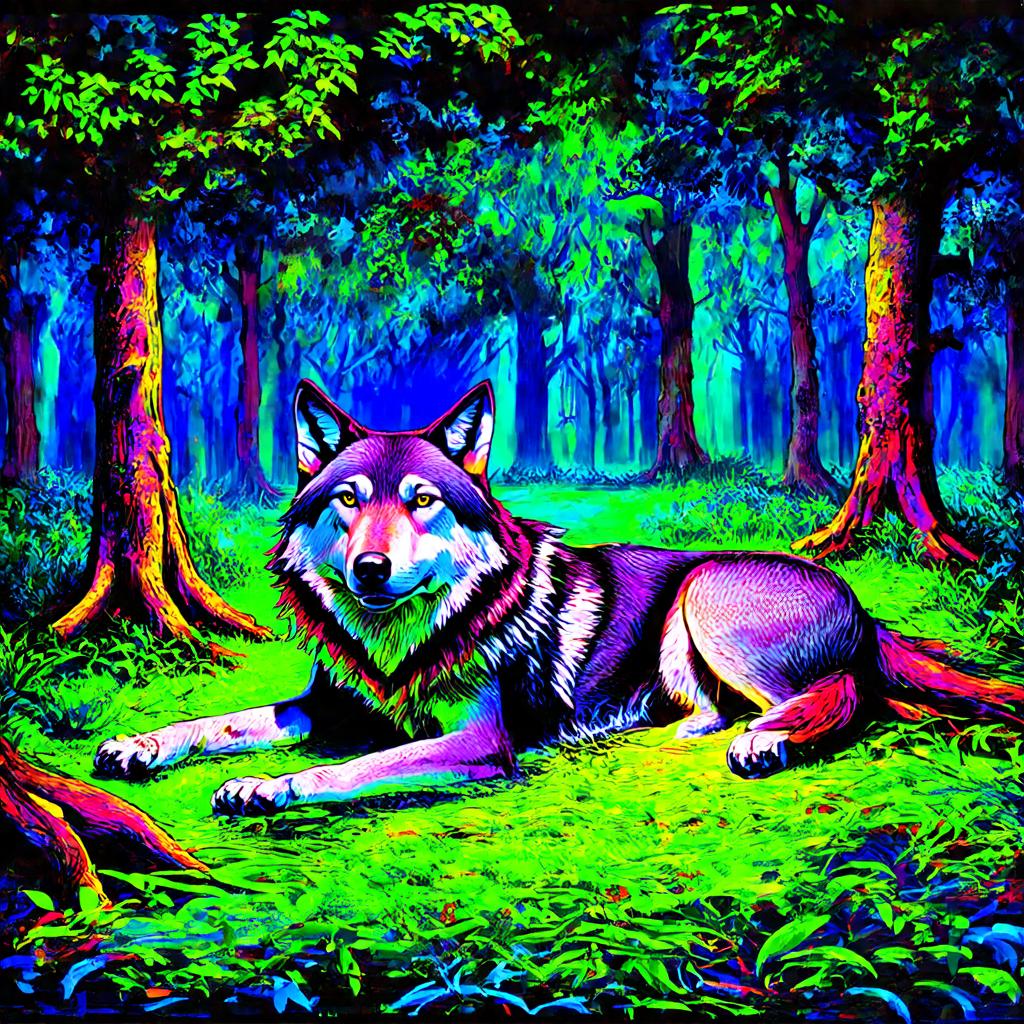}}%
        \fbox{\includegraphics[width=\mainimgwidth]{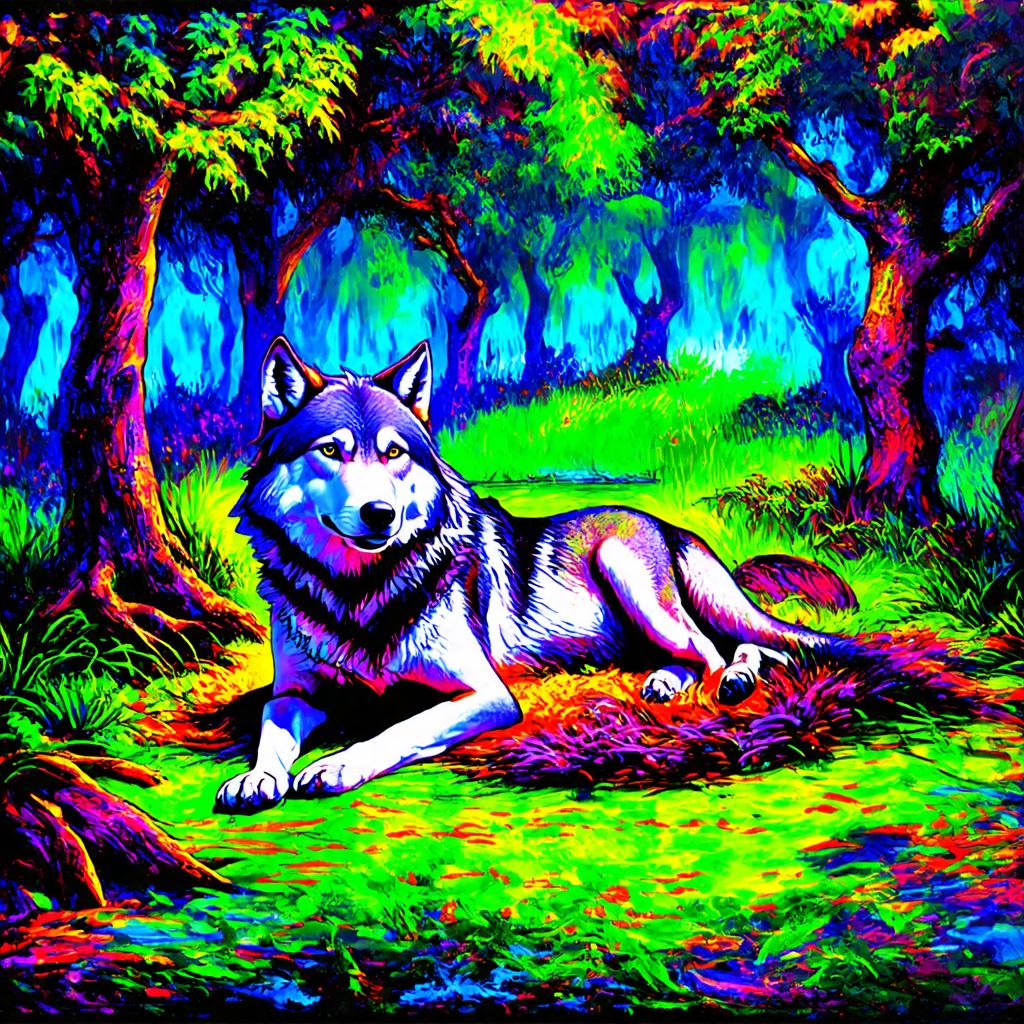}}\\[0.5ex]
        \hfill
        \vspace{-18pt}
        \caption*{
            \begin{minipage}{\maincapwidth}
            \centering
                \tiny{Prompt: \textit{wolf lounging in surreal scene of forest, drawn by johnen vasquez, pop art, high impact, wild and vivid colors, neon accents, blacklight reflective, magic and wonder, vivid colors, psychedelic, super art, high detail}}
            \end{minipage}
        }
    \end{minipage}

    \begin{minipage}[t]{0.495\textwidth}
        \centering
        \fbox{\includegraphics[width=\mainimgwidth]{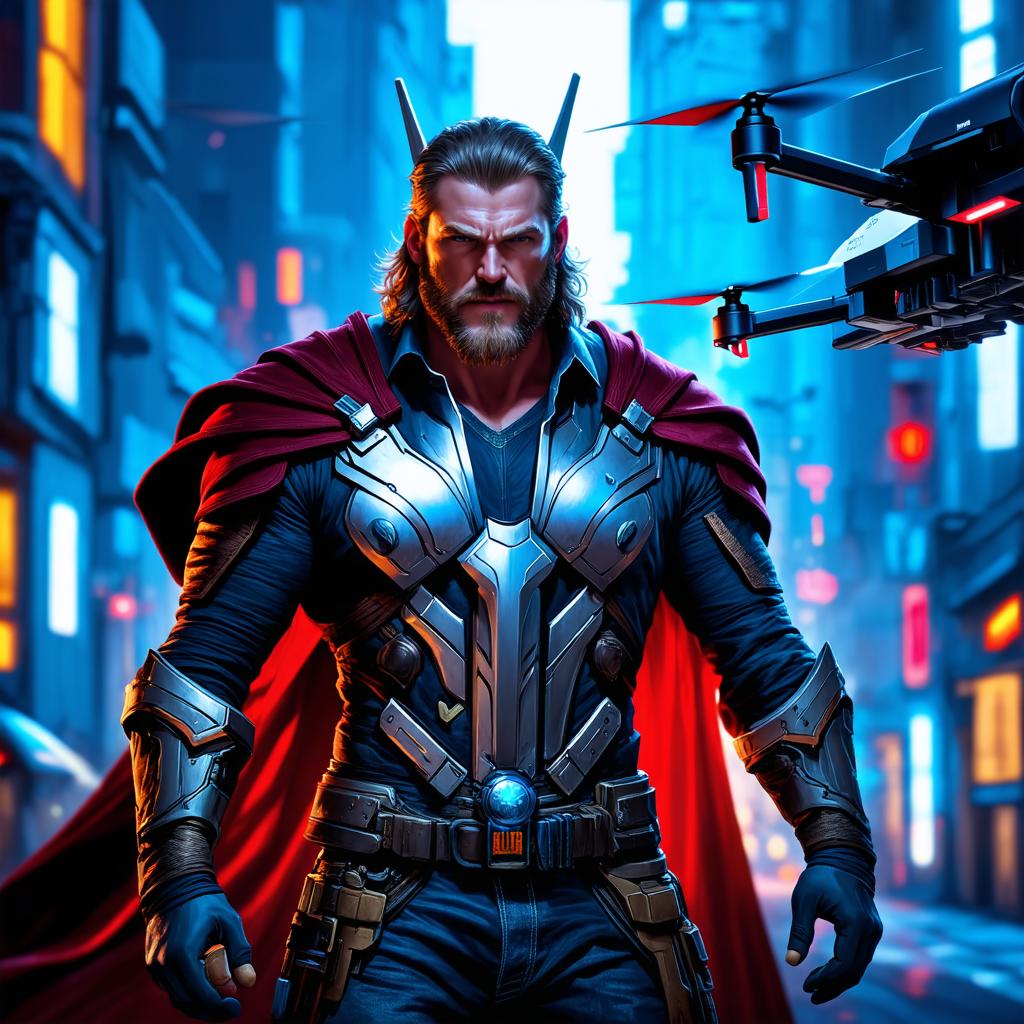}}
        \hfill
        \fbox{\includegraphics[width=\mainimgwidth]{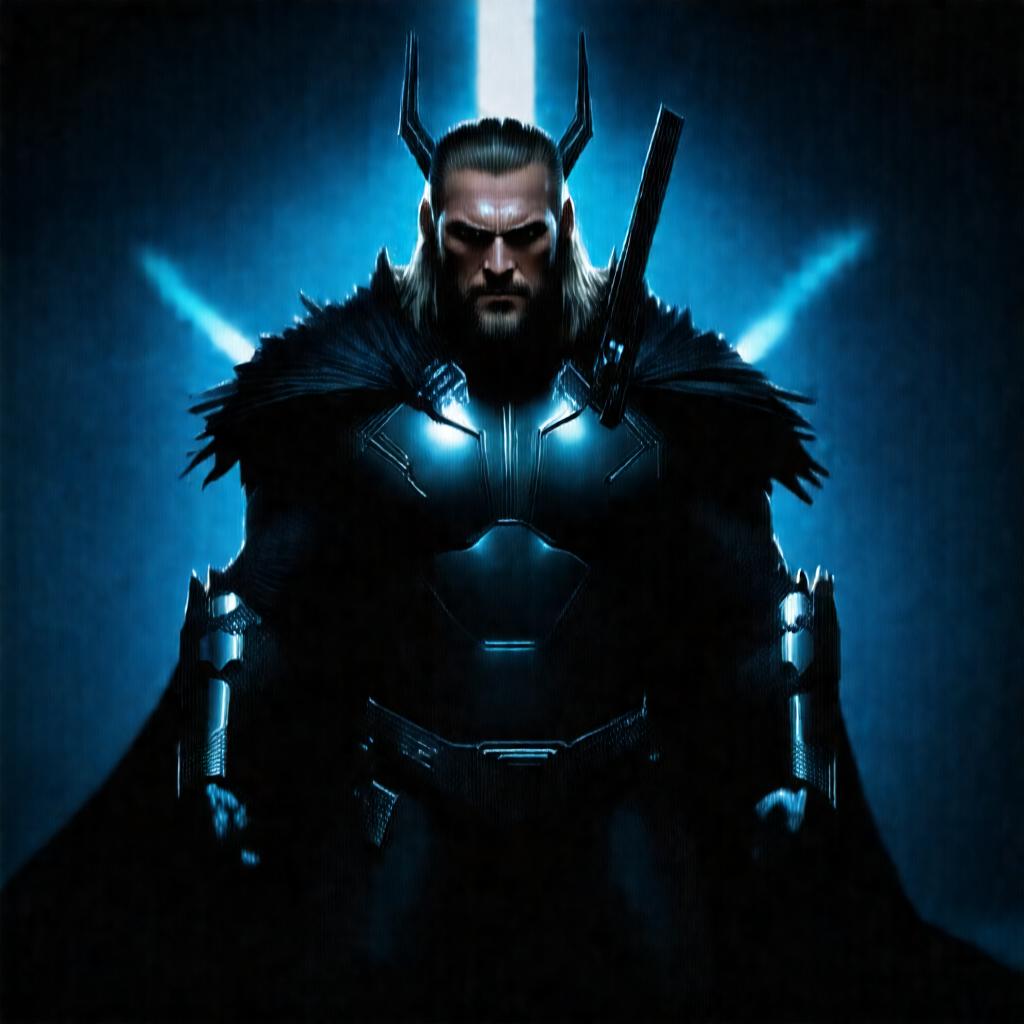}}%
        \fbox{\includegraphics[width=\mainimgwidth]{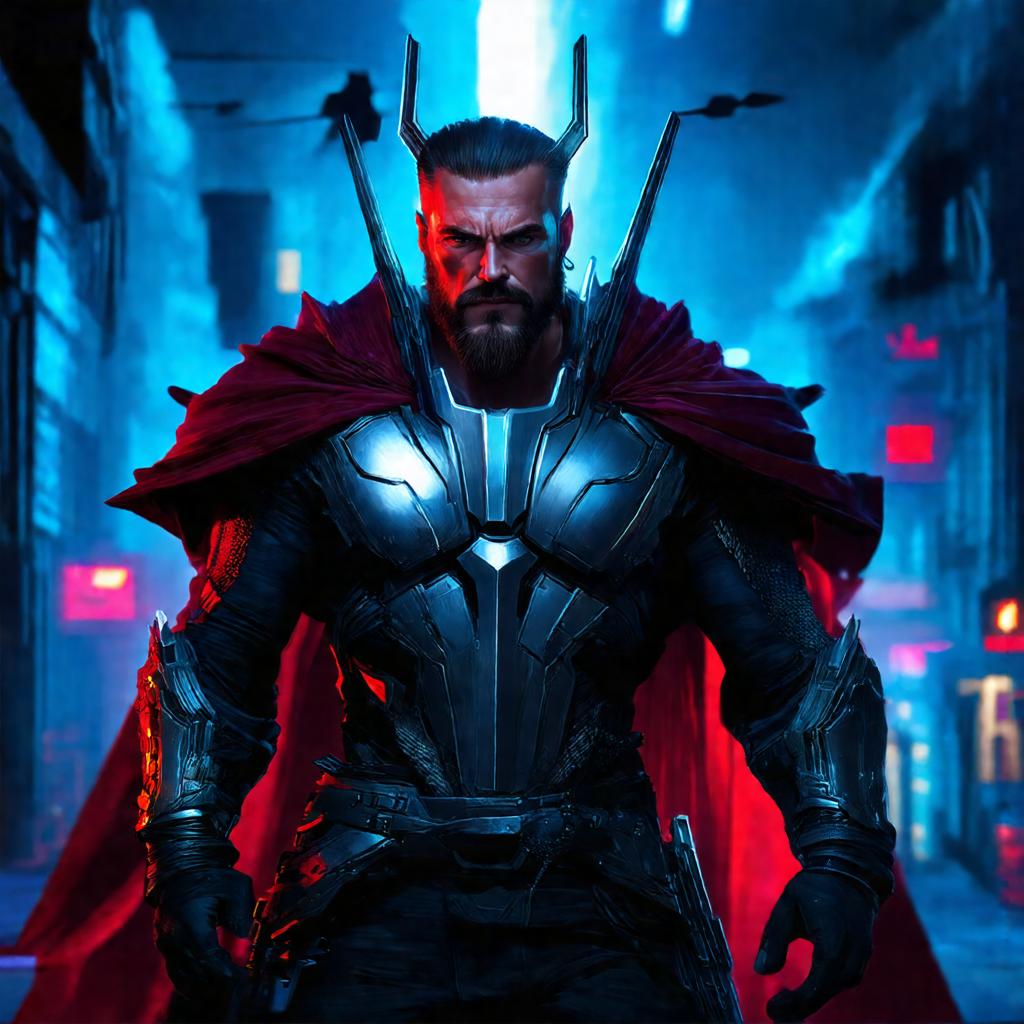}}%
        \fbox{\includegraphics[width=\mainimgwidth]{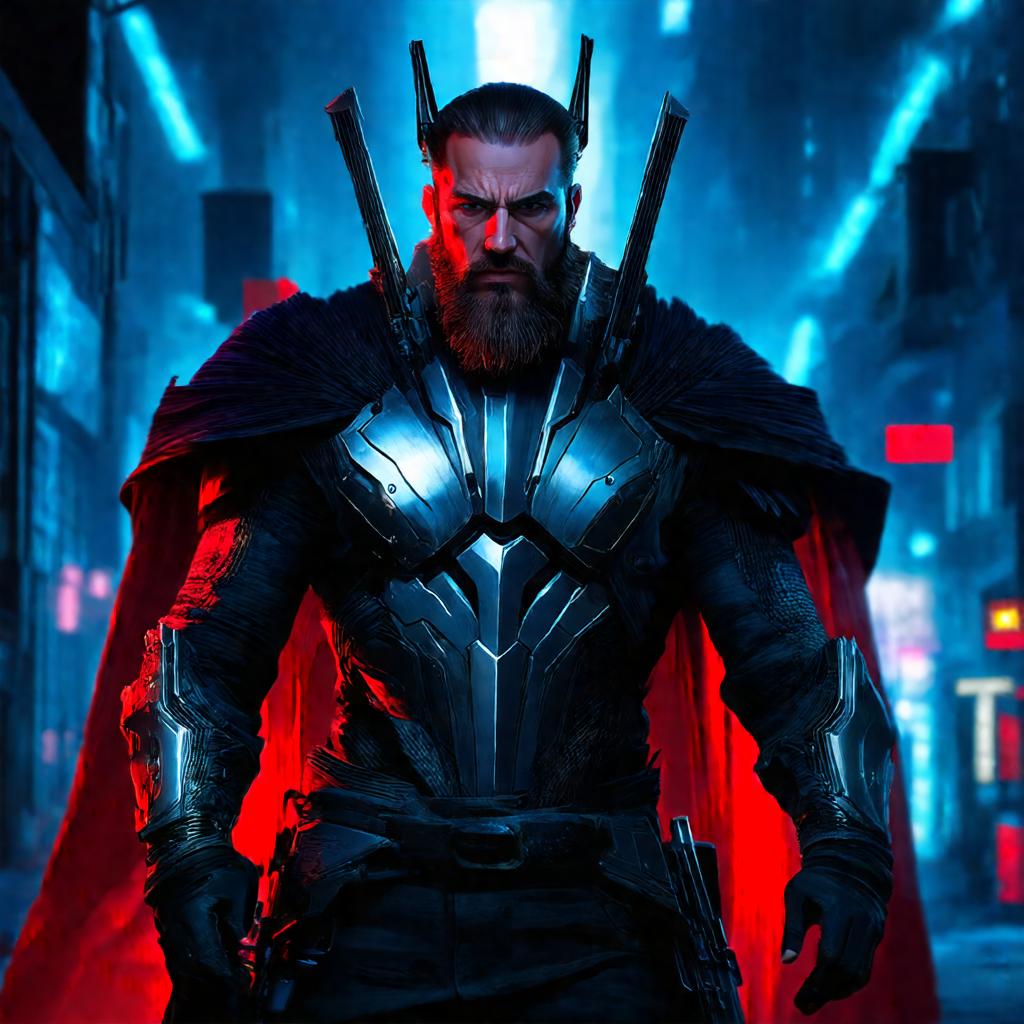}}%
        \fbox{\includegraphics[width=\mainimgwidth]{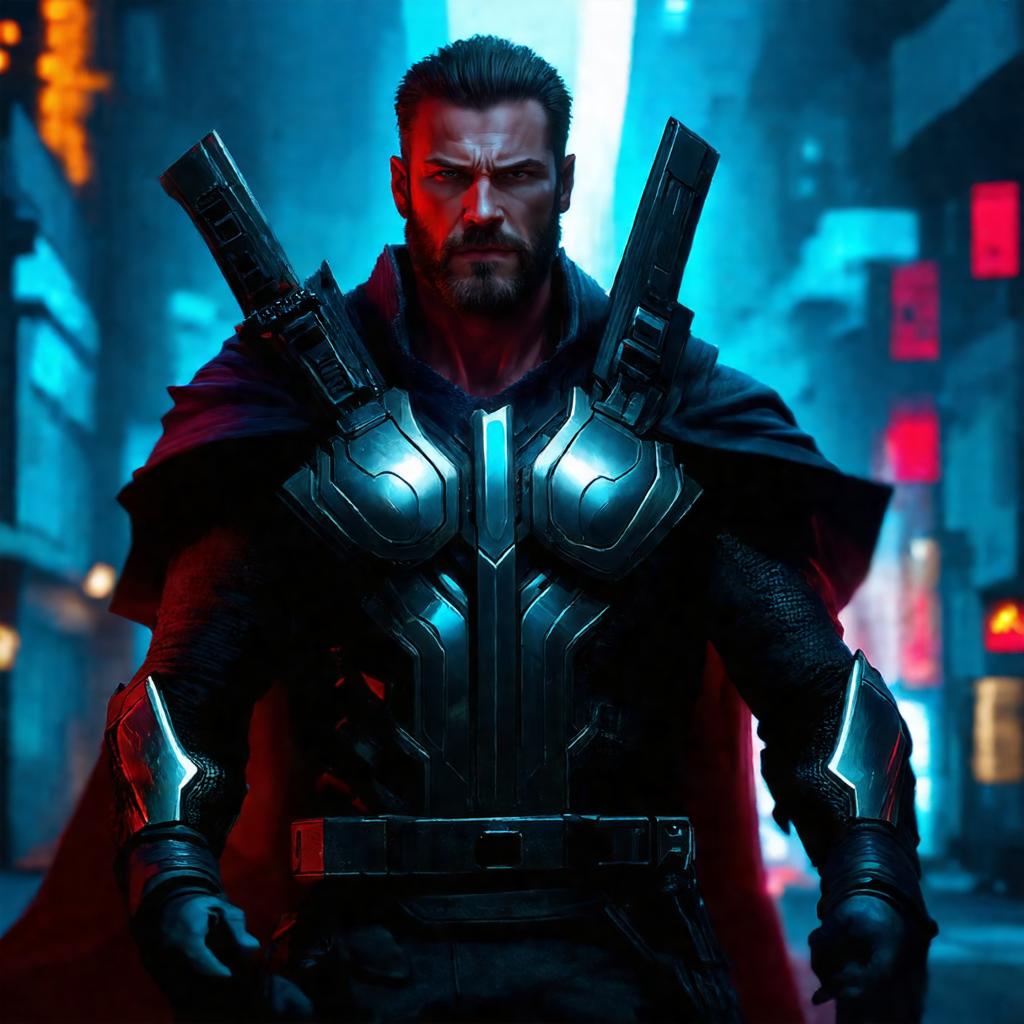}}%
        \fbox{\includegraphics[width=\mainimgwidth]{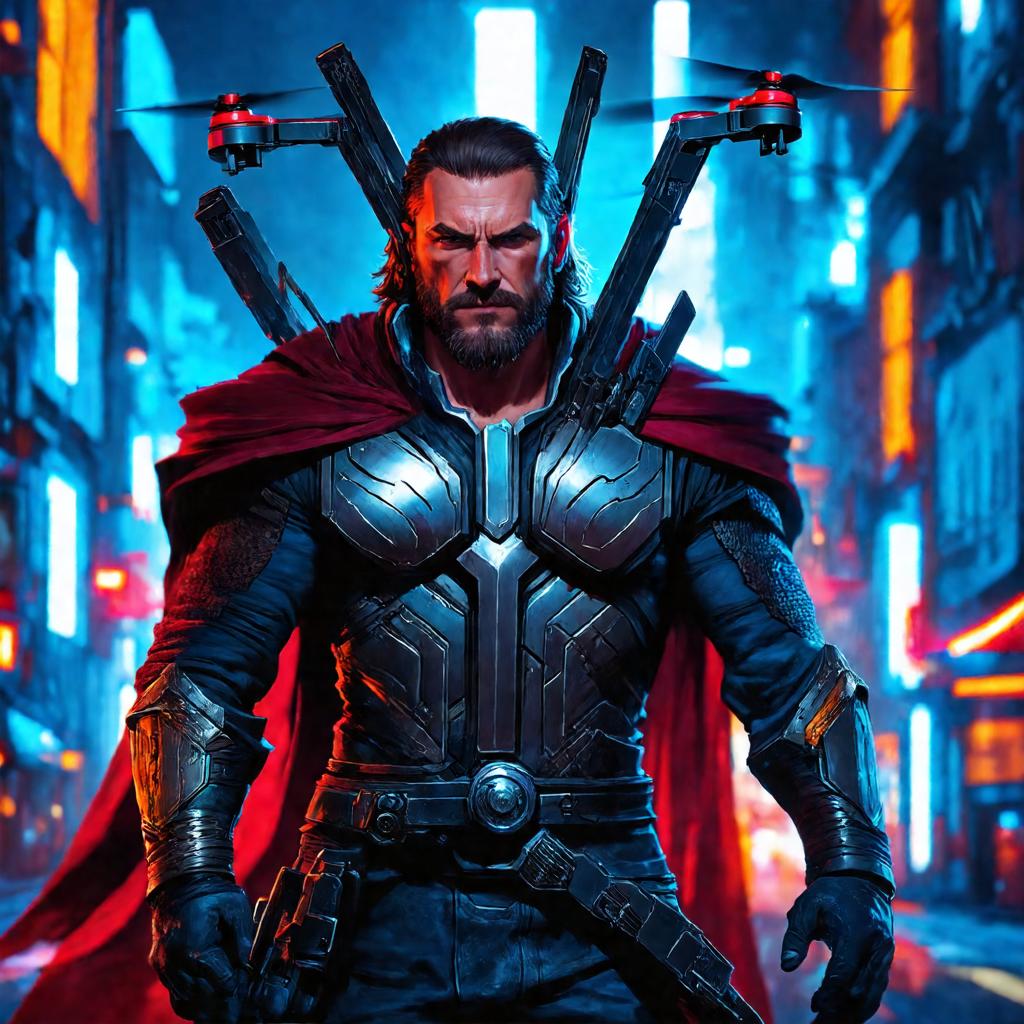}}%
        \fbox{\includegraphics[width=\mainimgwidth]{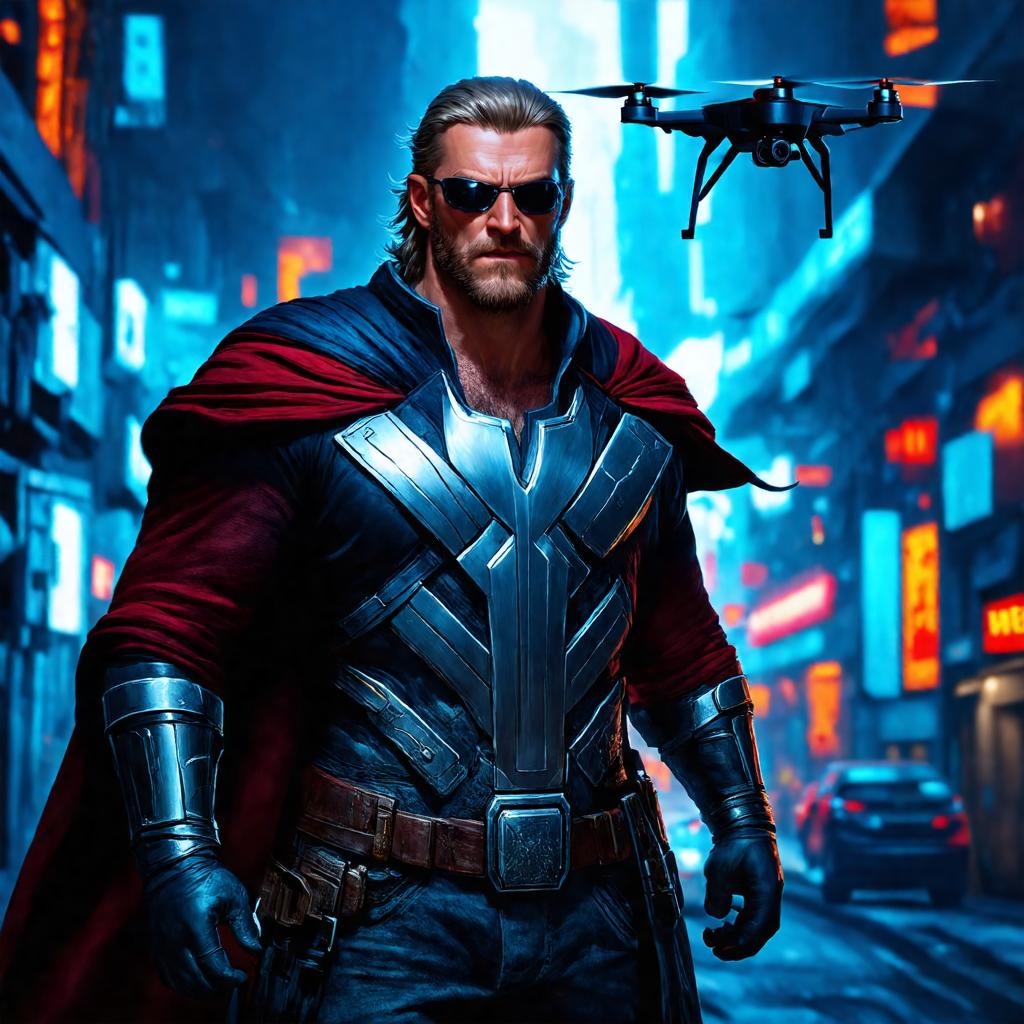}}\\[0.5ex]
        \hfill
        \vspace{-18pt}
        \caption*{
            \begin{minipage}{\maincapwidth}
            \centering
                \tiny{Prompt: \textit{thor with drone, with neon lights in the background, Intricate, Highly detailed, Sharp focus, Digital painting, Artstation, Concept art, inspired by blade runner, ghost in the shell and cyberpunk 2077, art by rafal wechterowicz and khyzyl saleem}}
            \end{minipage}
        }
    \end{minipage}
    \hfill
    \begin{minipage}[t]{0.495\textwidth}
        \centering
        \fbox{\includegraphics[width=\mainimgwidth]{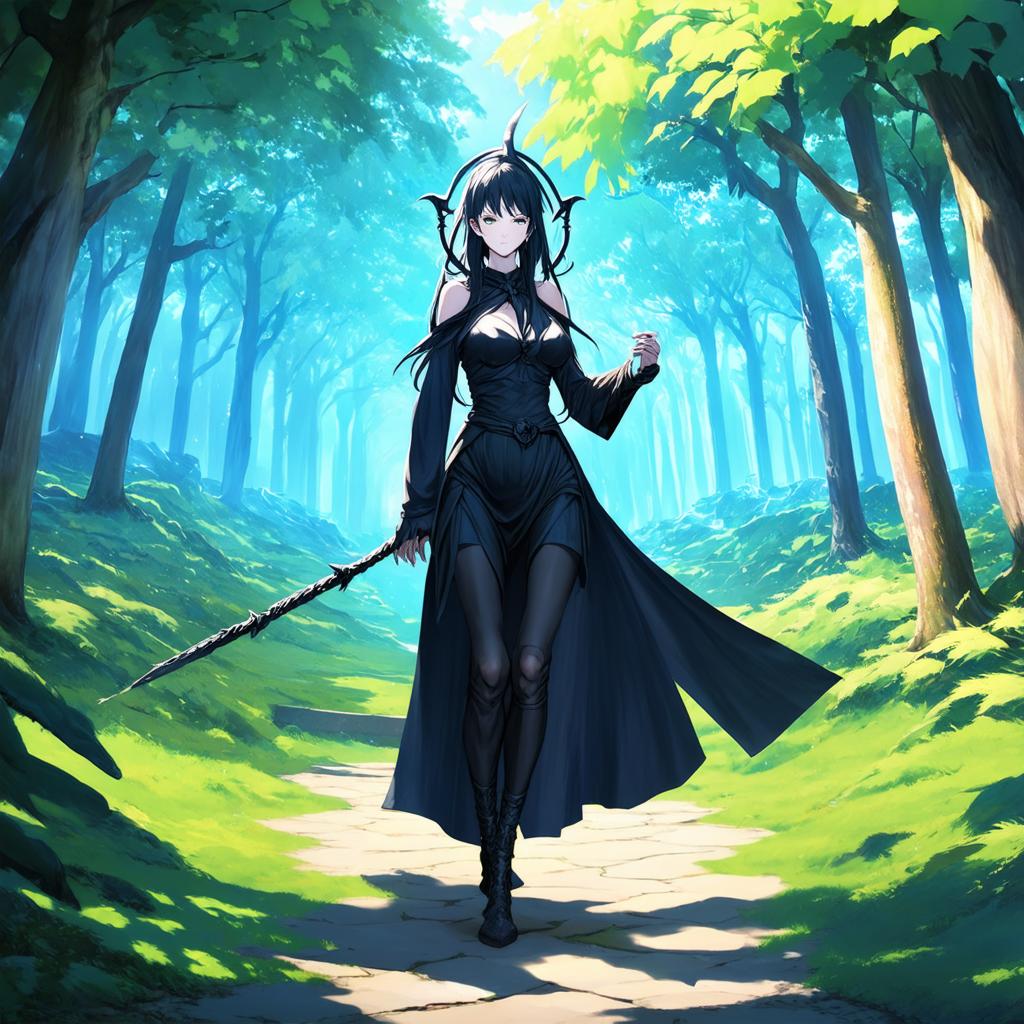}}
        \hfill
        \fbox{\includegraphics[width=\mainimgwidth]{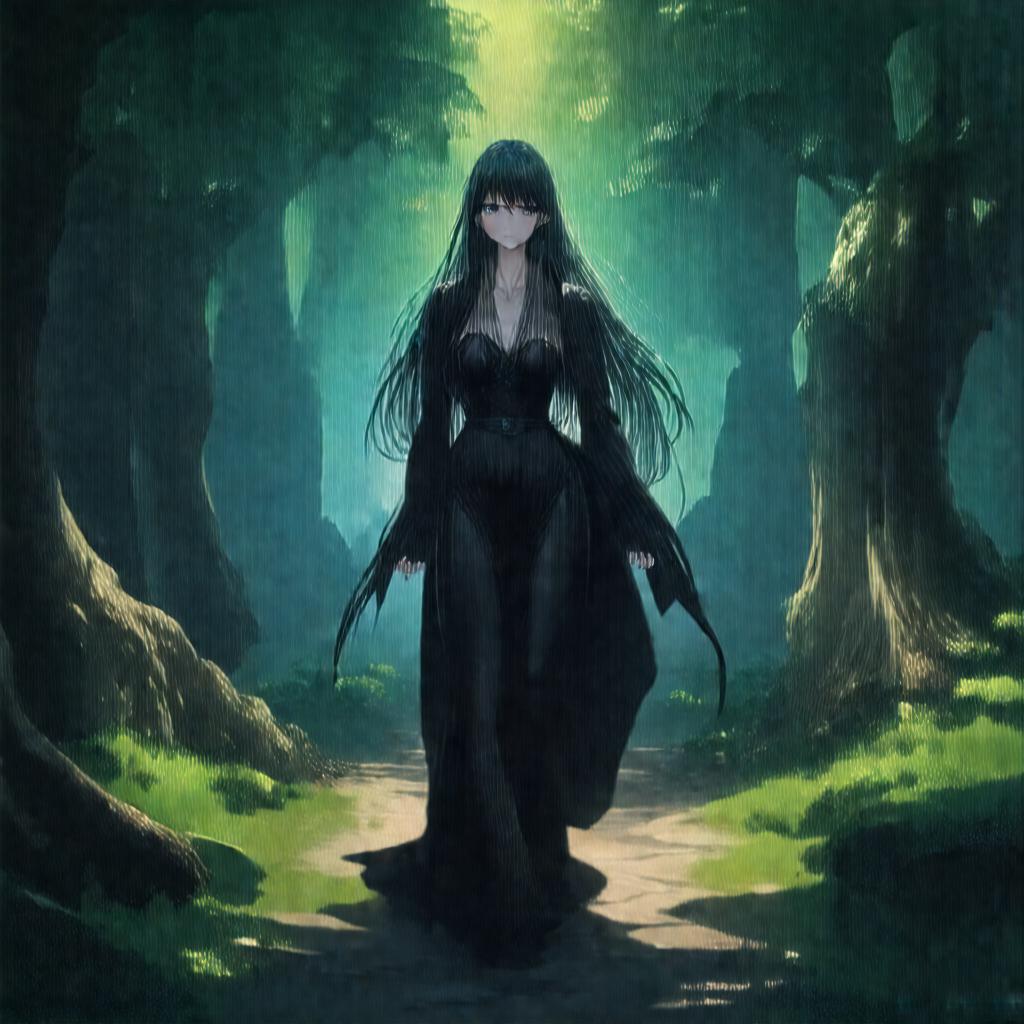}}%
        \fbox{\includegraphics[width=\mainimgwidth]{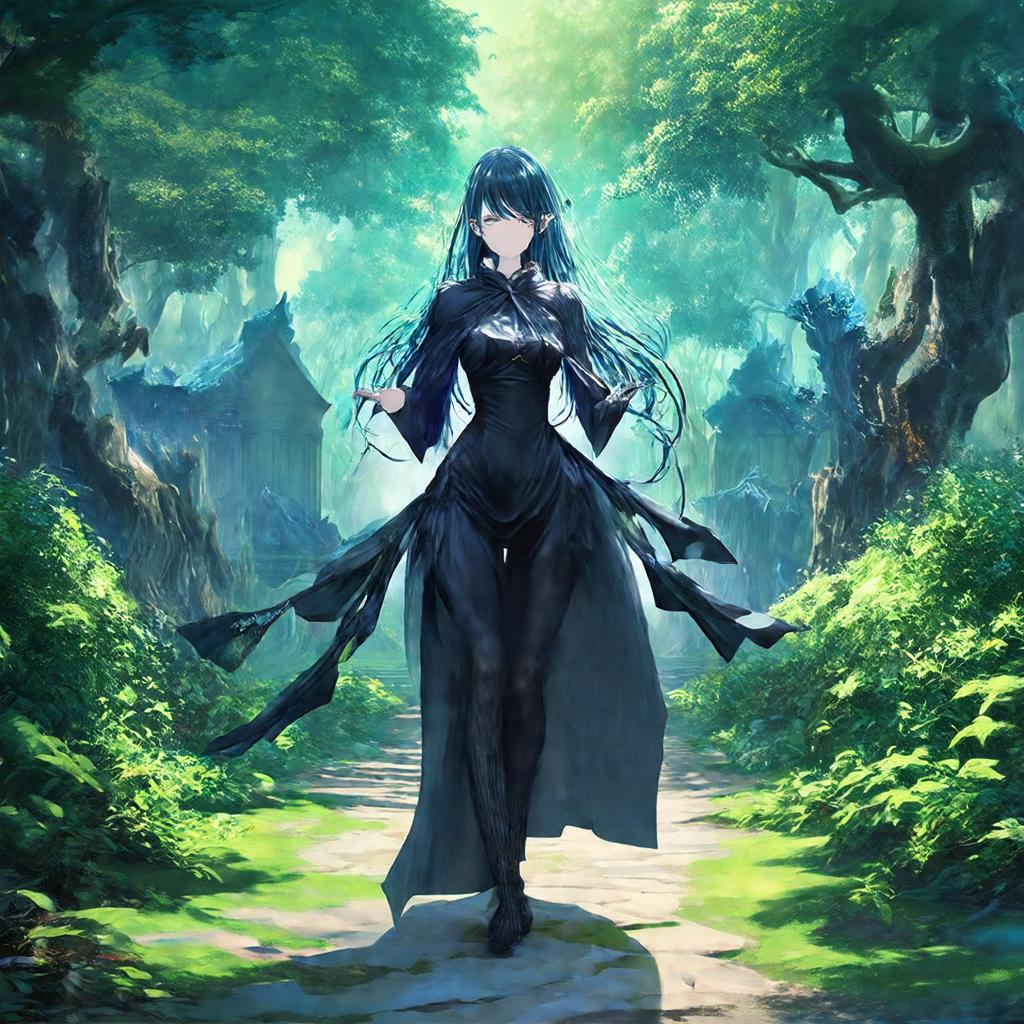}}%
        \fbox{\includegraphics[width=\mainimgwidth]{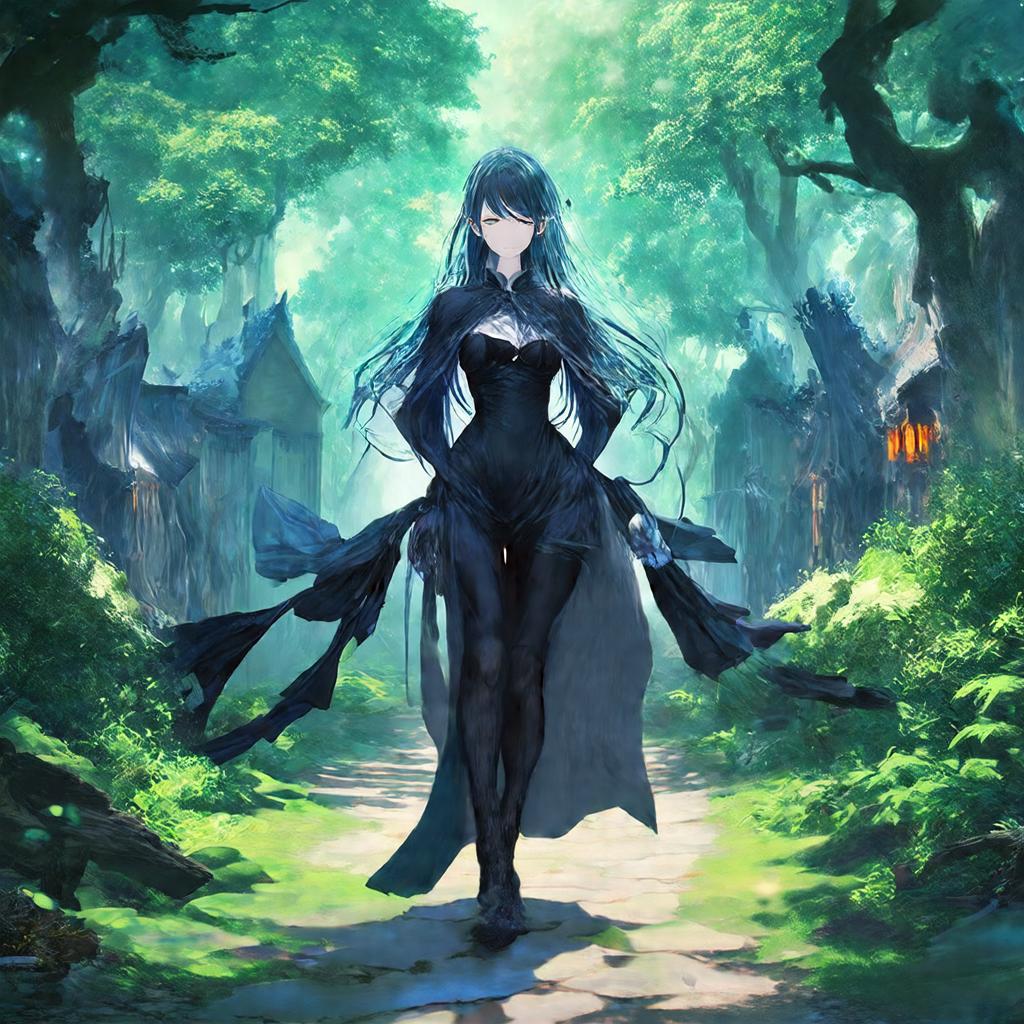}}%
        \fbox{\includegraphics[width=\mainimgwidth]{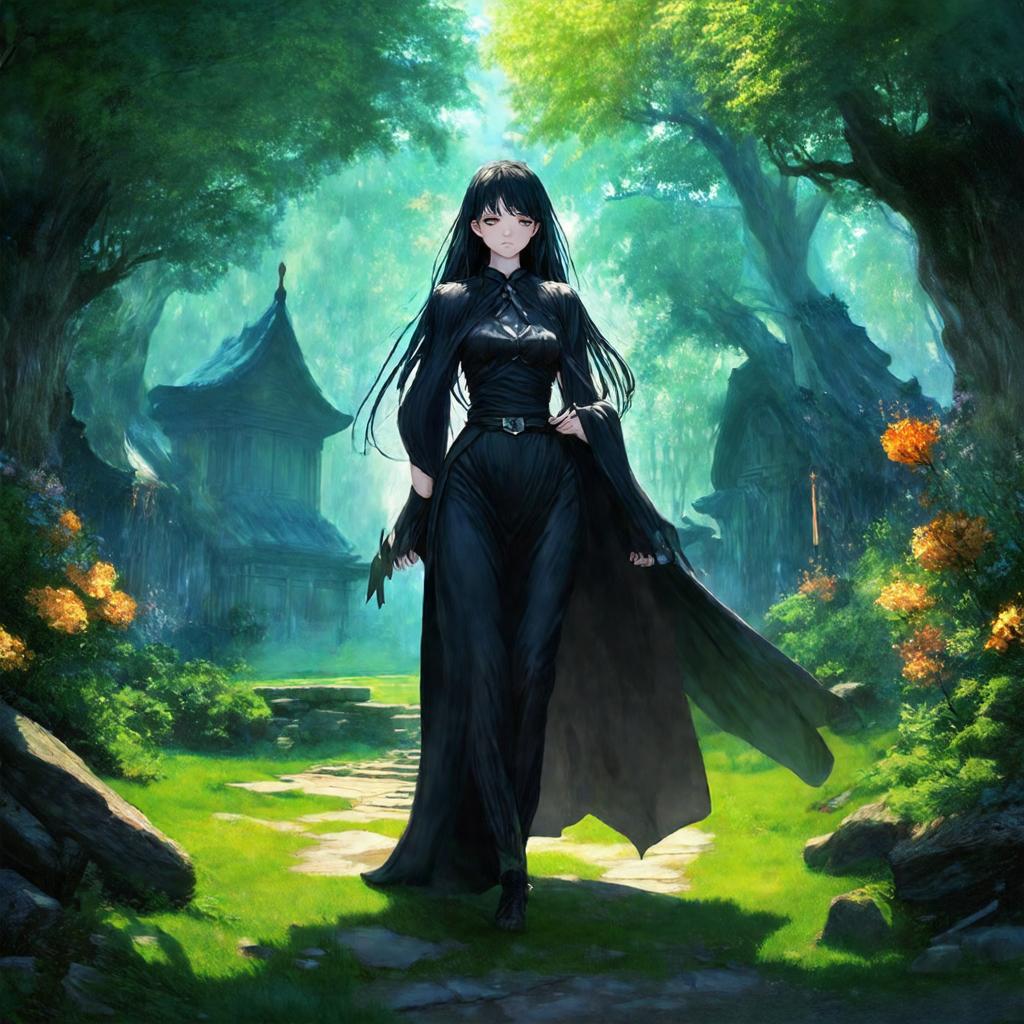}}%
        \fbox{\includegraphics[width=\mainimgwidth]{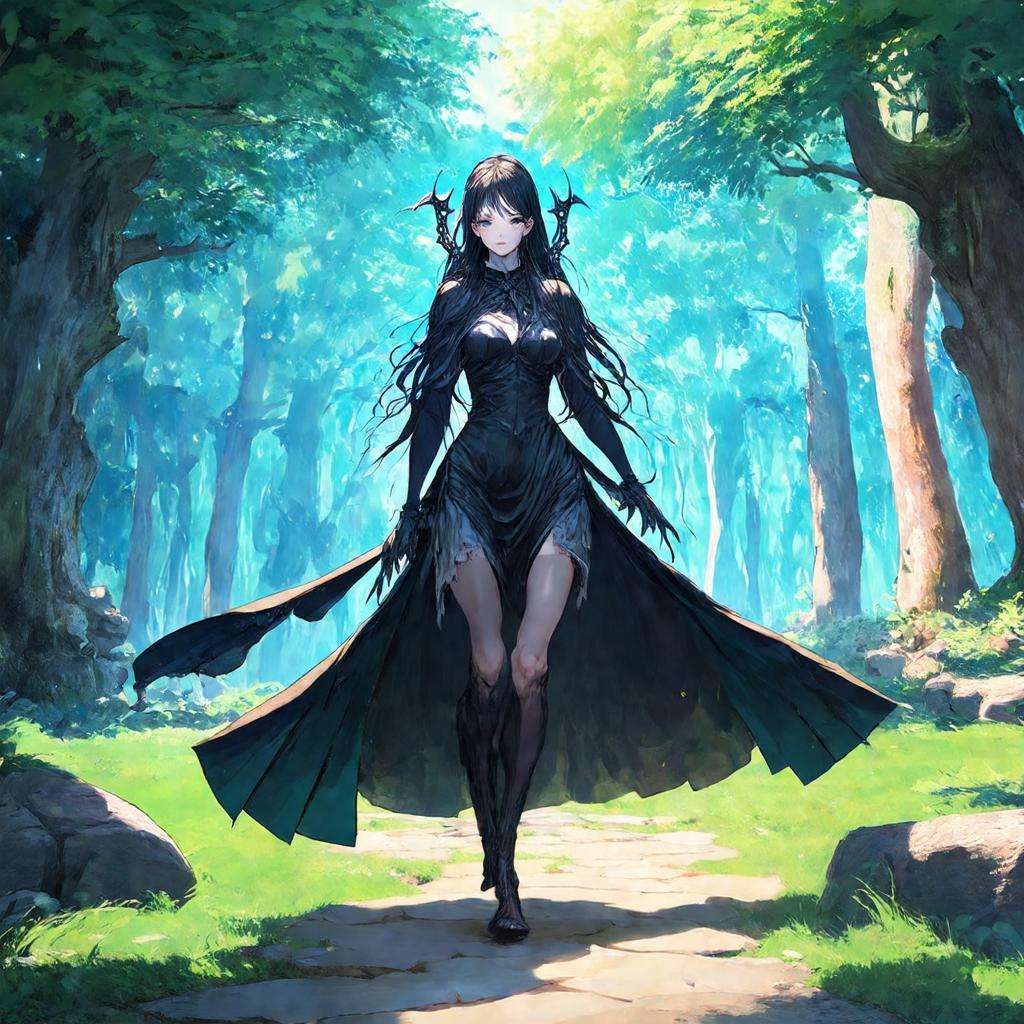}}%
        \fbox{\includegraphics[width=\mainimgwidth]{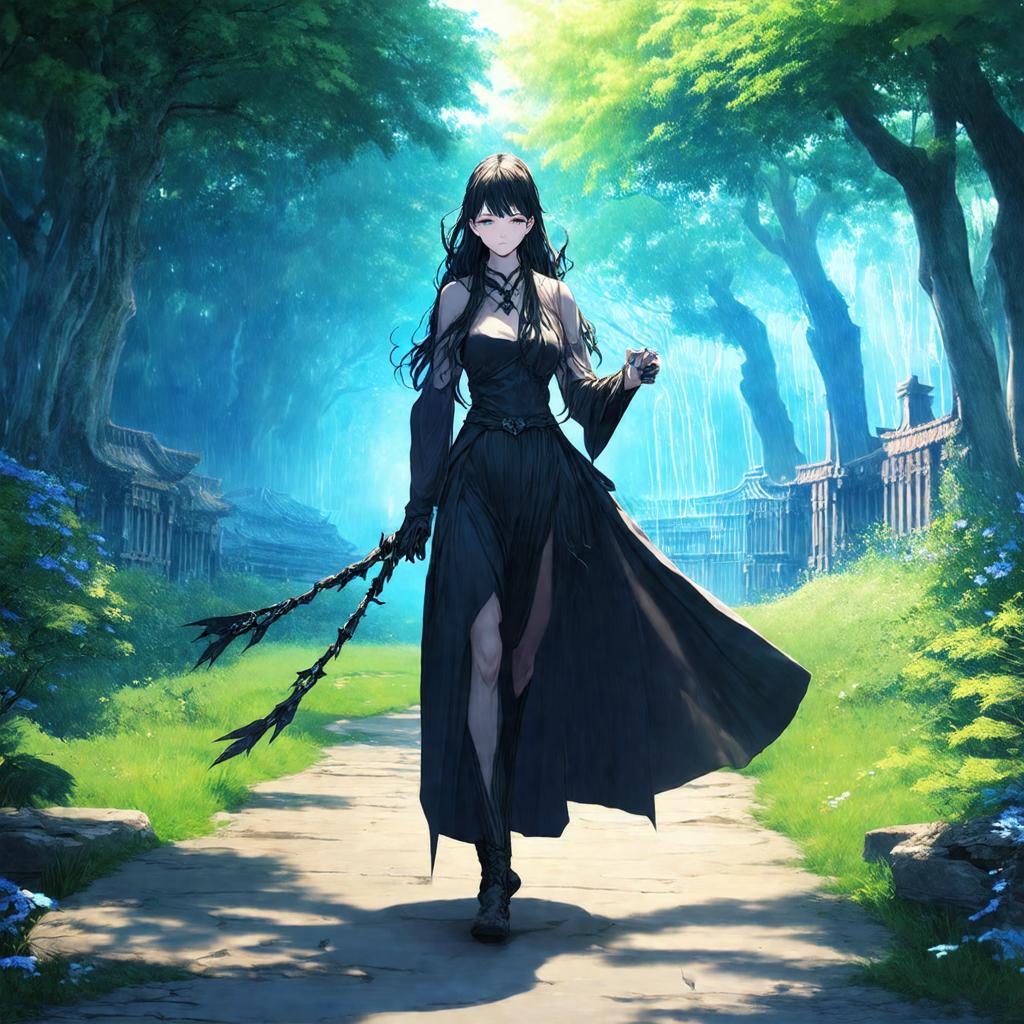}}\\[0.5ex]
        \hfill
        \vspace{-18pt}
        \caption*{
            \begin{minipage}{\maincapwidth}
            \centering
                \tiny{Prompt: \textit{hecate walking in a temple forest, anime, magical realism, hd, photorealistic, anime}}
            \end{minipage}
        }
    \end{minipage}

    \begin{minipage}[t]{0.495\textwidth}
        \centering
        \fbox{\includegraphics[width=\mainimgwidth]{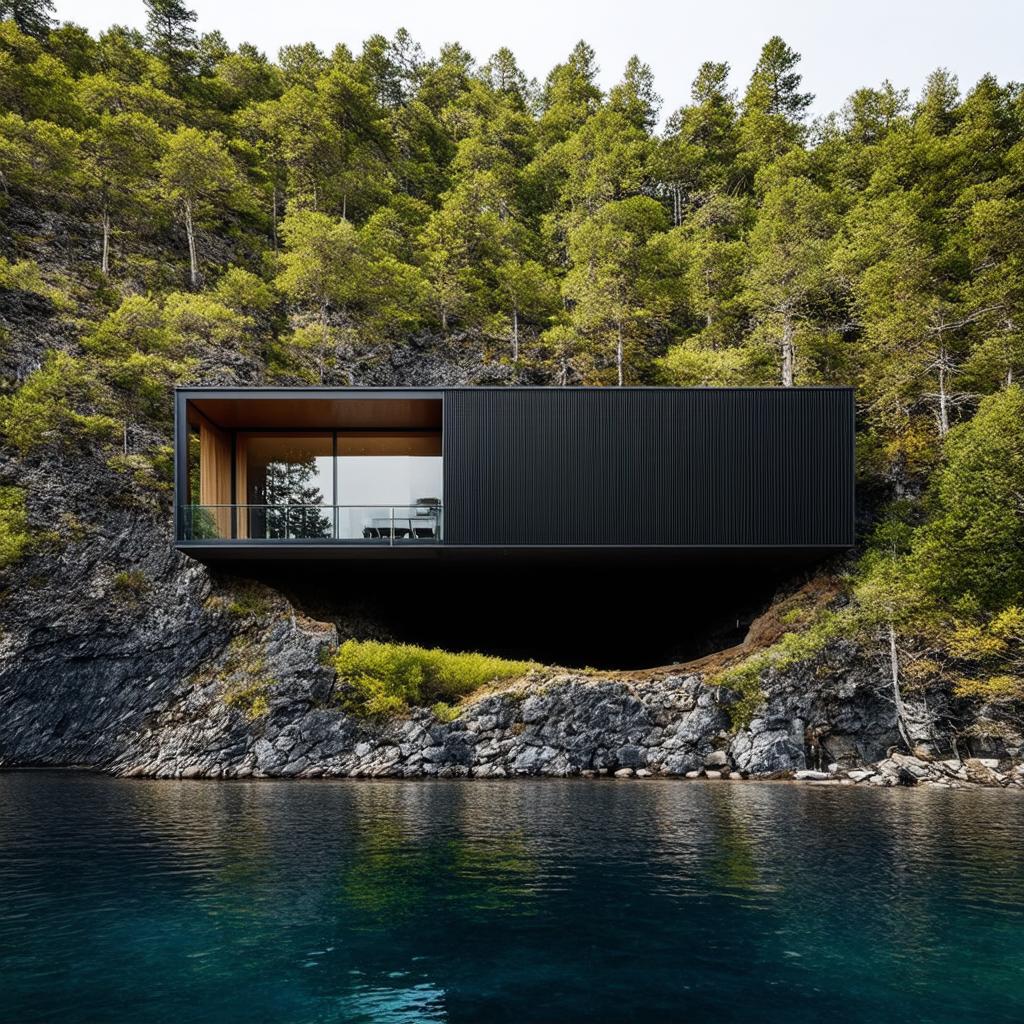}}
        \hfill
        \fbox{\includegraphics[width=\mainimgwidth]{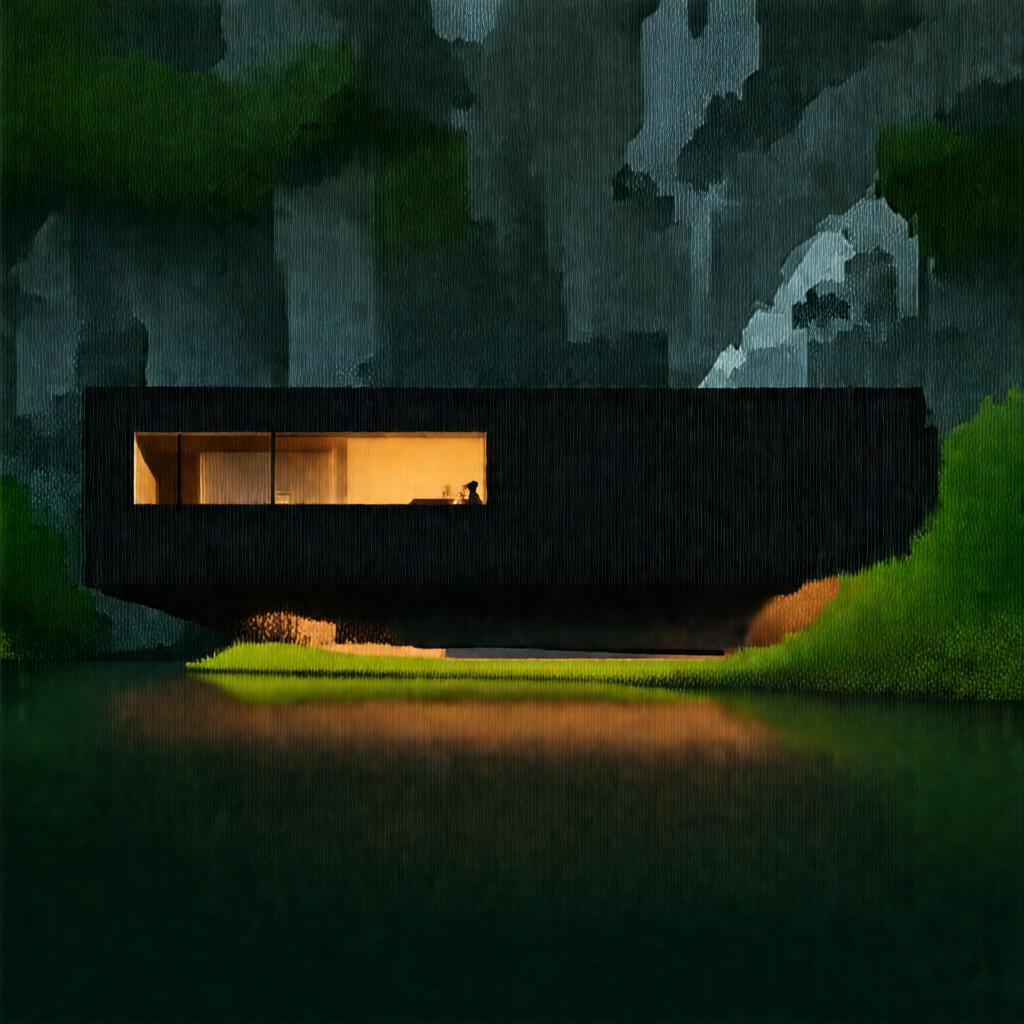}}%
        \fbox{\includegraphics[width=\mainimgwidth]{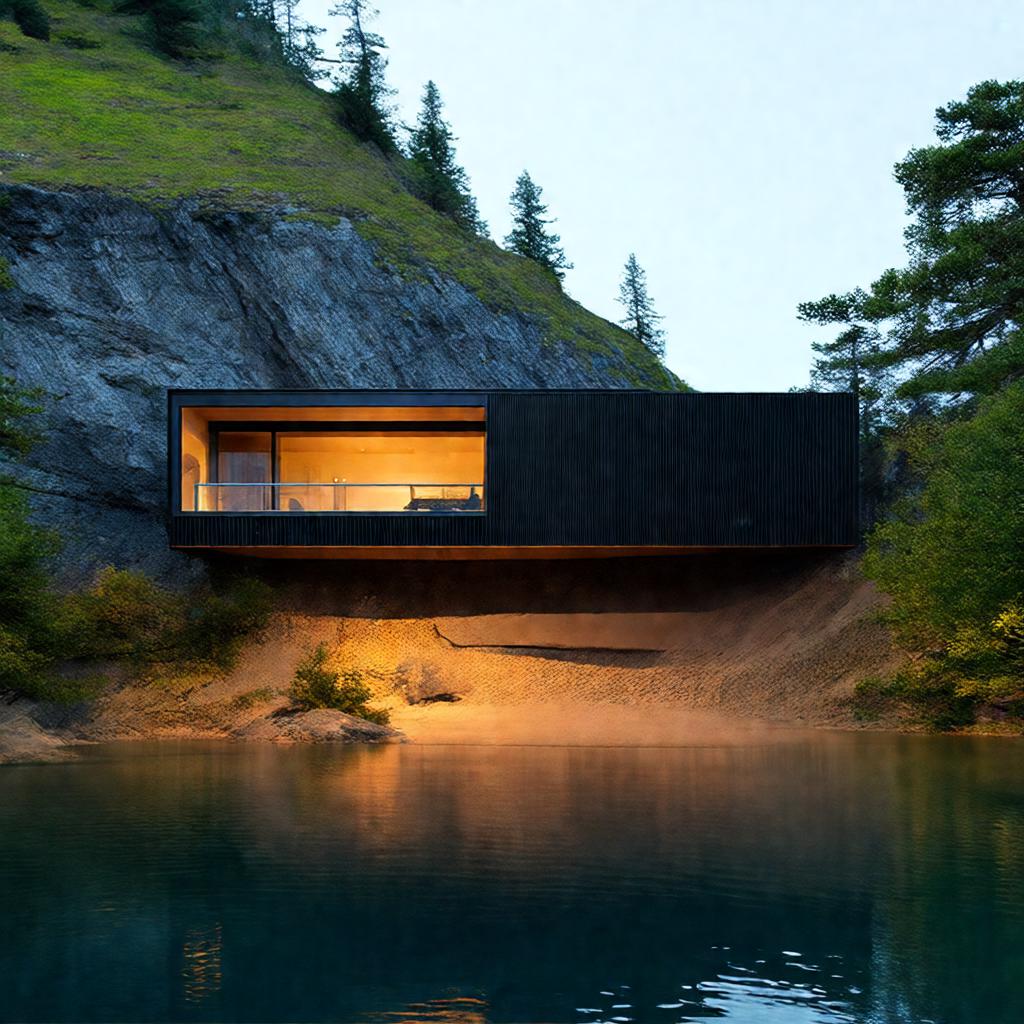}}%
        \fbox{\includegraphics[width=\mainimgwidth]{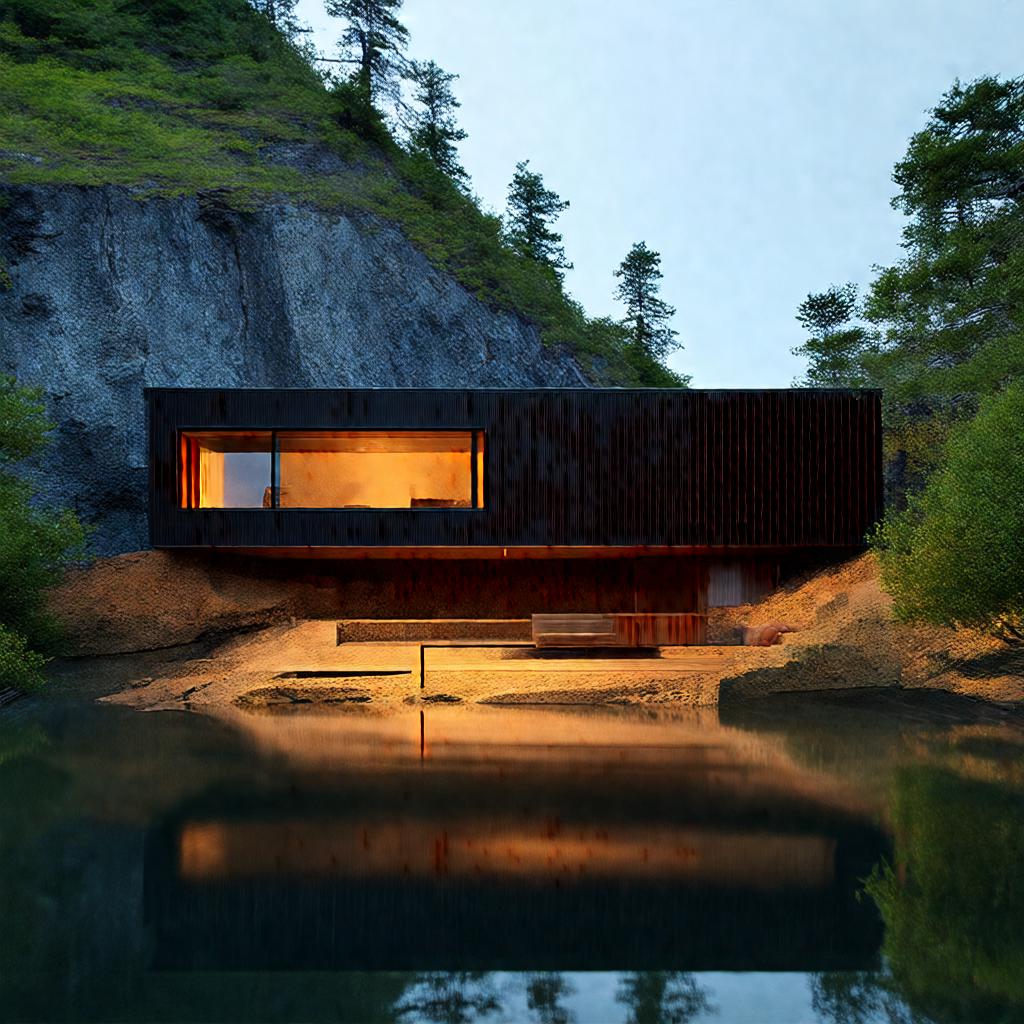}}%
        \fbox{\includegraphics[width=\mainimgwidth]{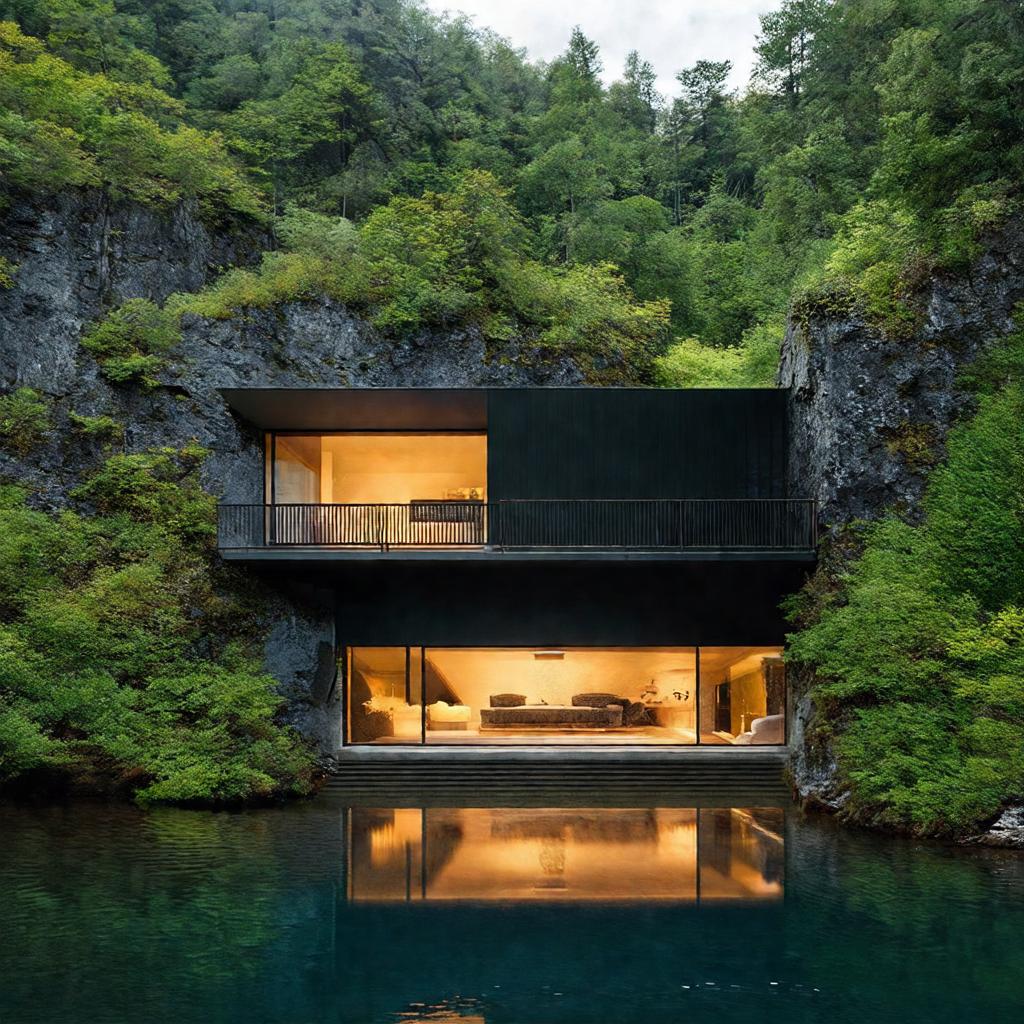}}%
        \fbox{\includegraphics[width=\mainimgwidth]{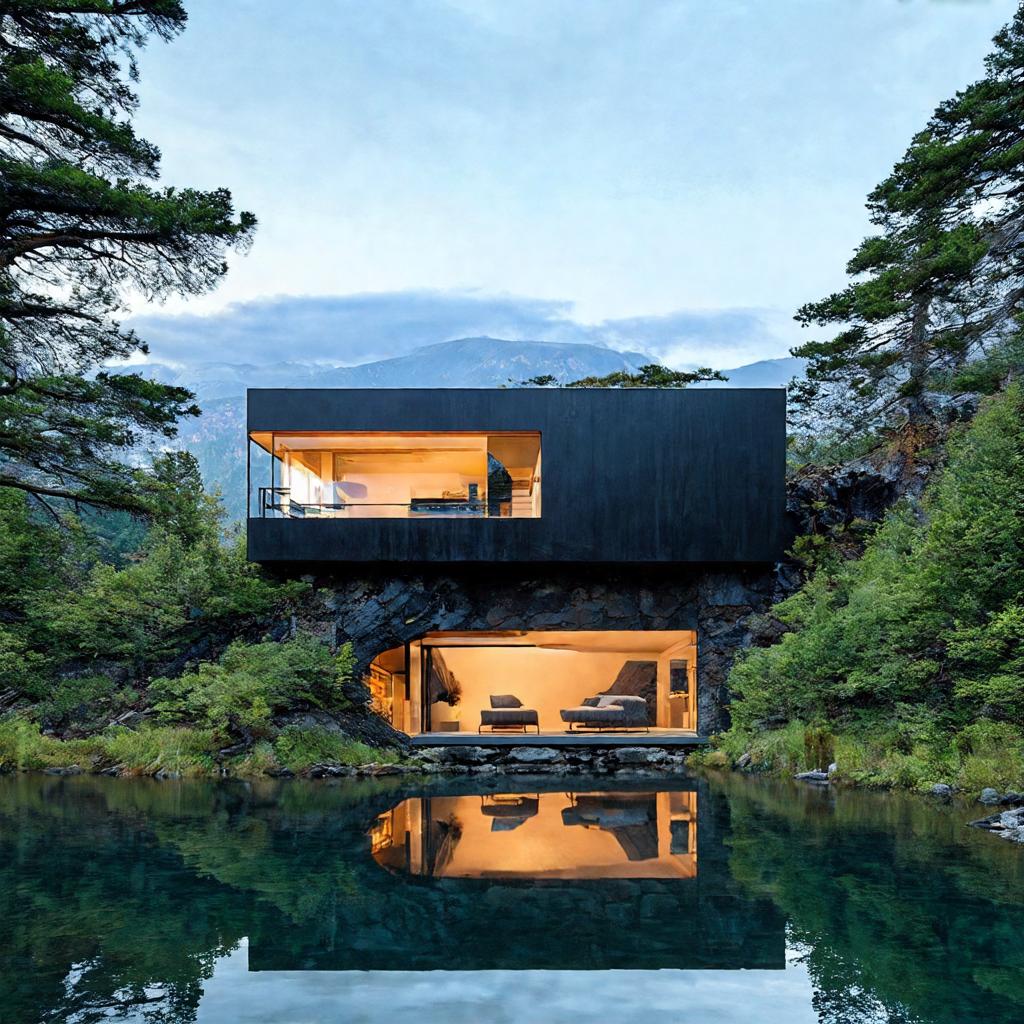}}%
        \fbox{\includegraphics[width=\mainimgwidth]{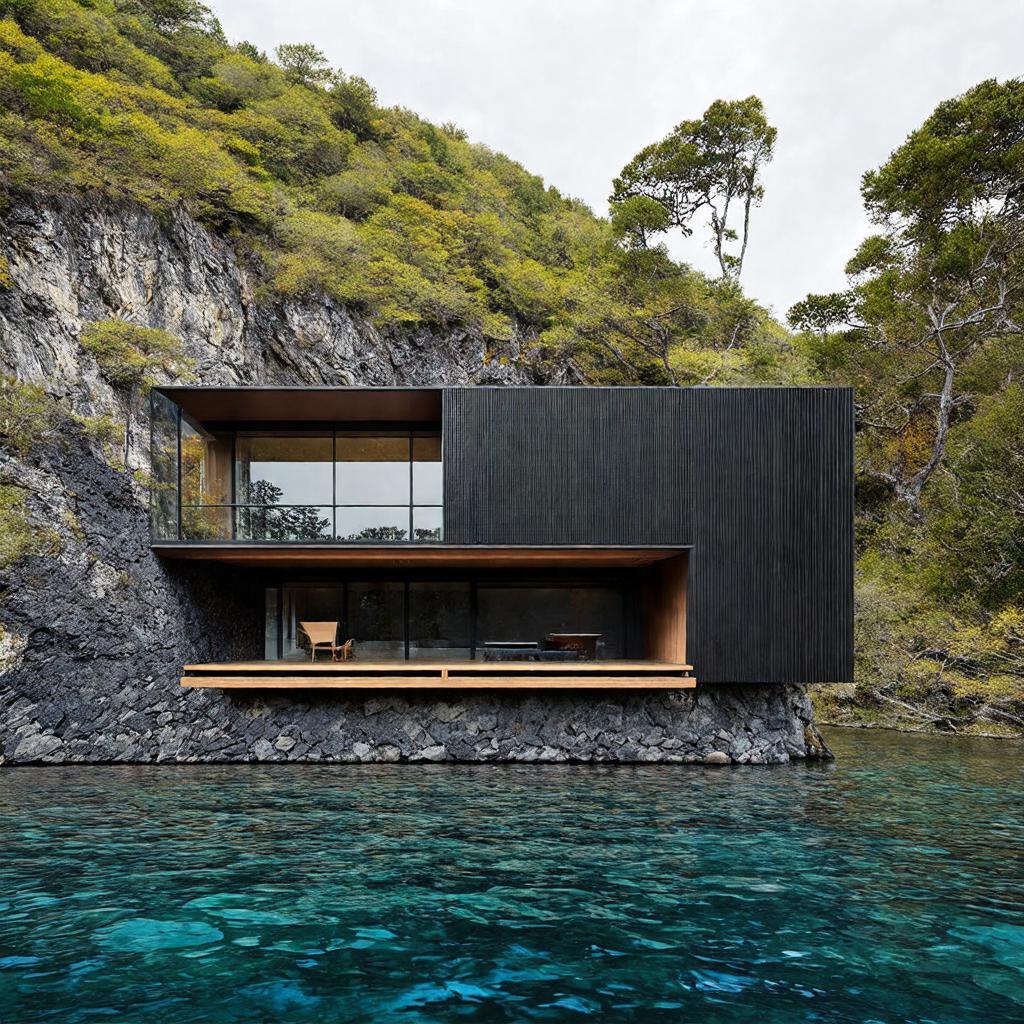}}\\[0.5ex]
        \hfill
        \vspace{-18pt}
        \caption*{
            \begin{minipage}{\maincapwidth}
            \centering
                \tiny{Prompt: \textit{modern one store house buried in the green mountain in forest cliff rectangle long facade with a deck, with black stone, inside of a mountain, warm light, perspective view with a reflexion in a lake}}
            \end{minipage}
        }
    \end{minipage}
    \hfill
    \begin{minipage}[t]{0.495\textwidth}
        \centering
        \fbox{\includegraphics[width=\mainimgwidth]{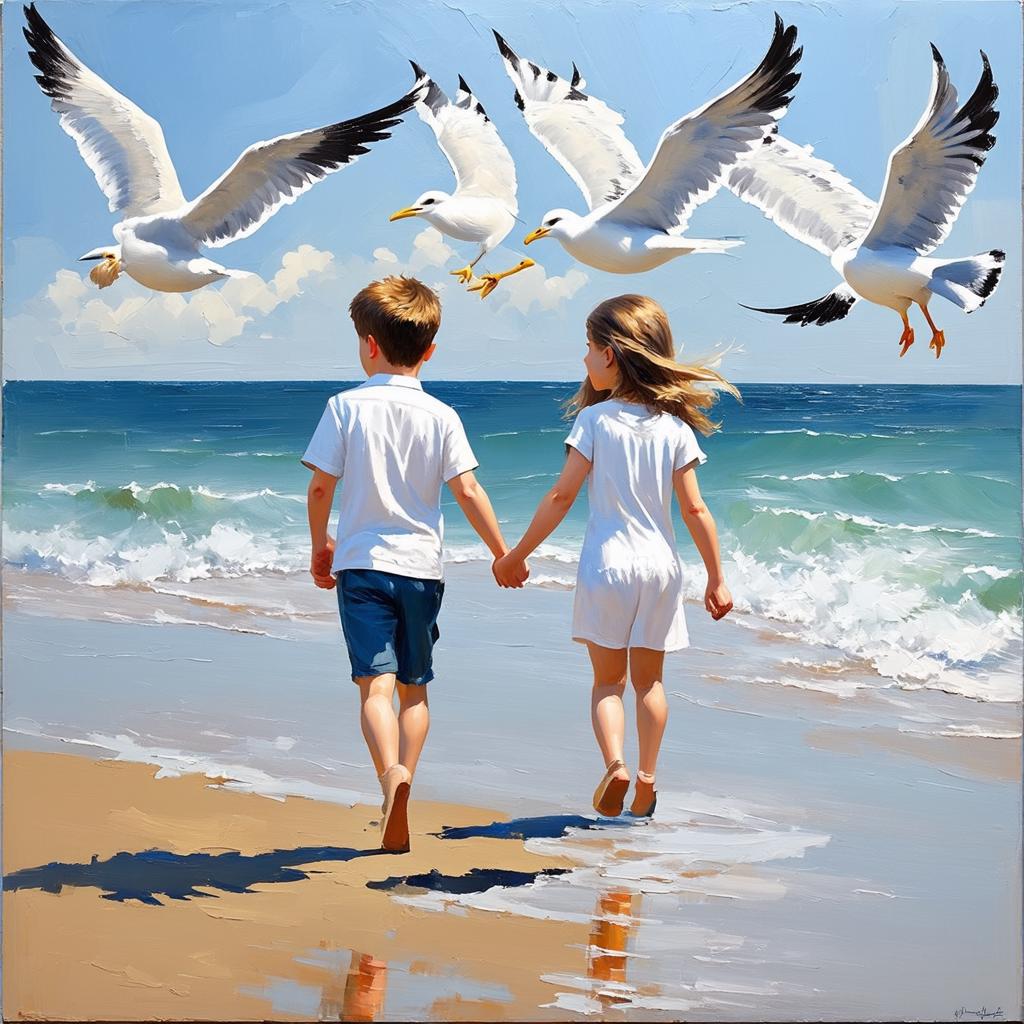}}
        \hfill
        \fbox{\includegraphics[width=\mainimgwidth]{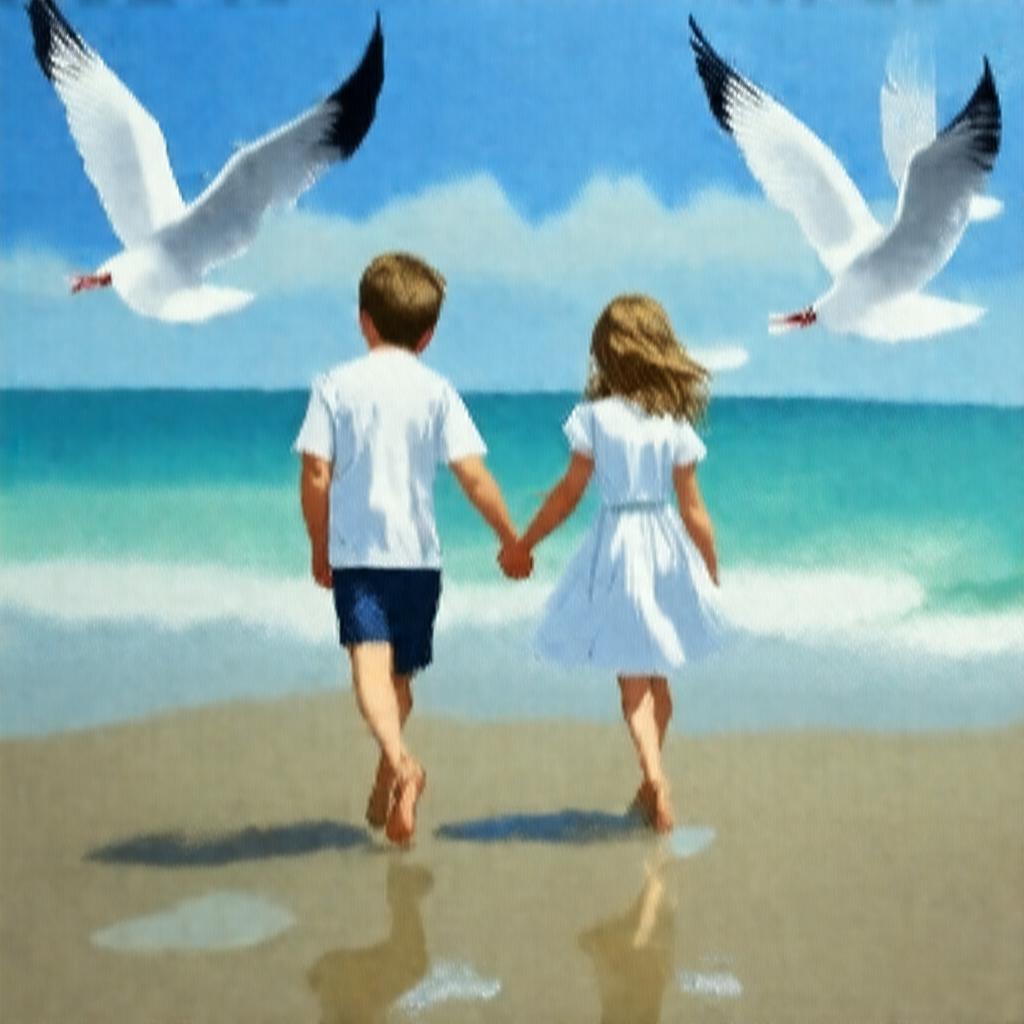}}%
        \fbox{\includegraphics[width=\mainimgwidth]{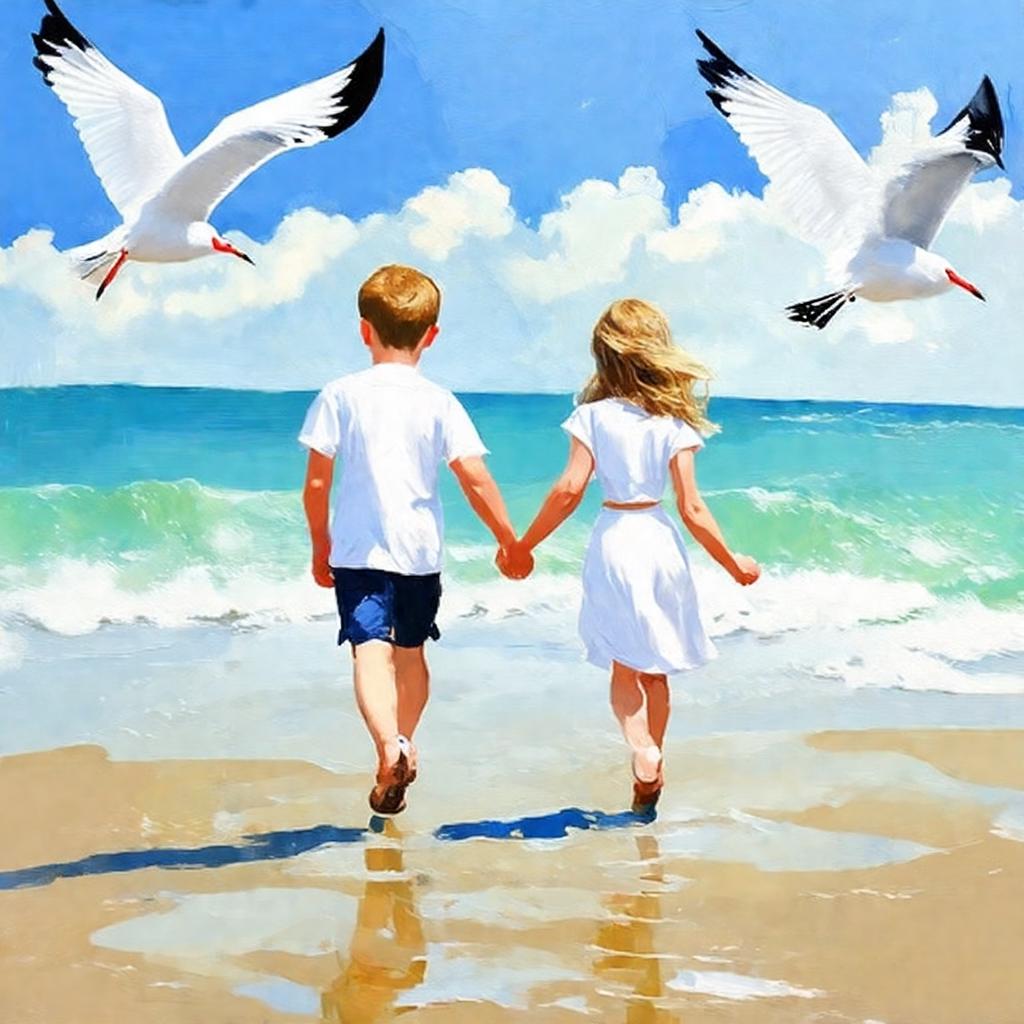}}%
        \fbox{\includegraphics[width=\mainimgwidth]{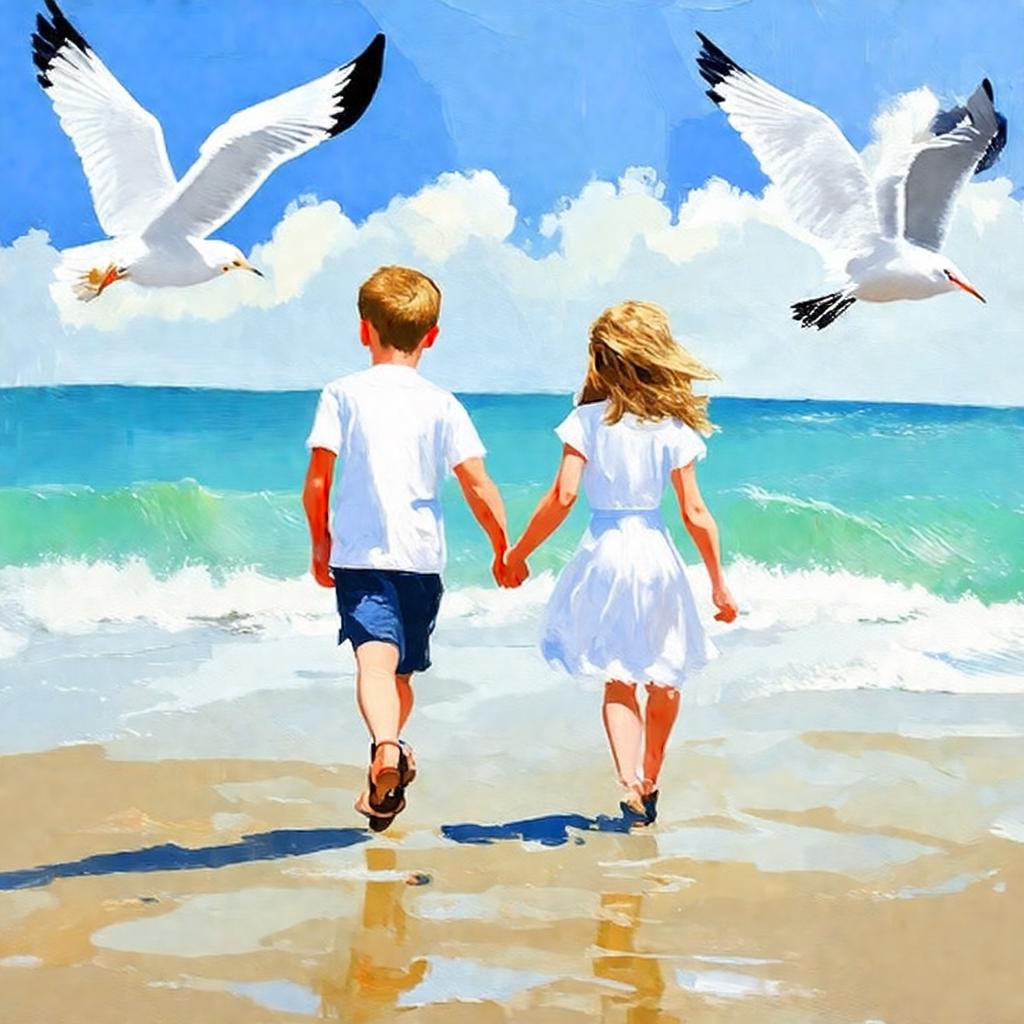}}%
        \fbox{\includegraphics[width=\mainimgwidth]{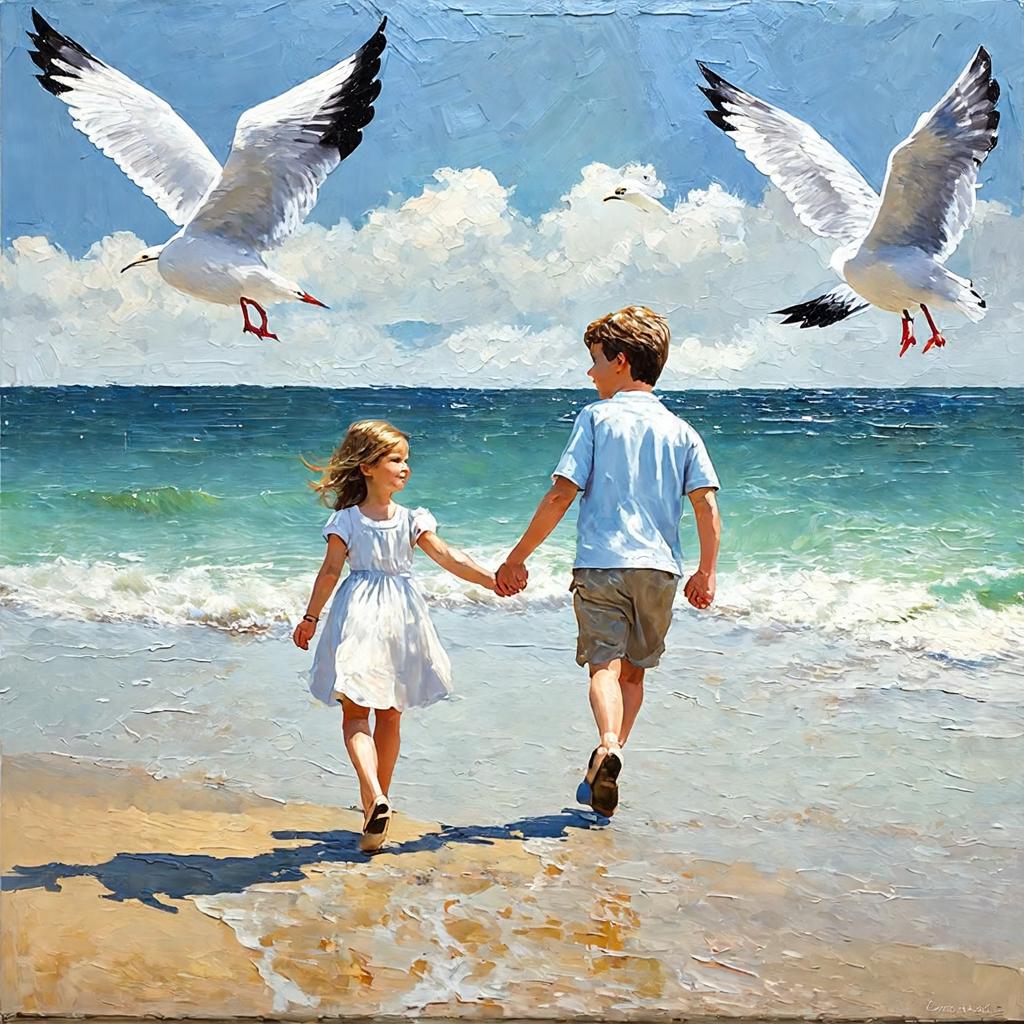}}%
        \fbox{\includegraphics[width=\mainimgwidth]{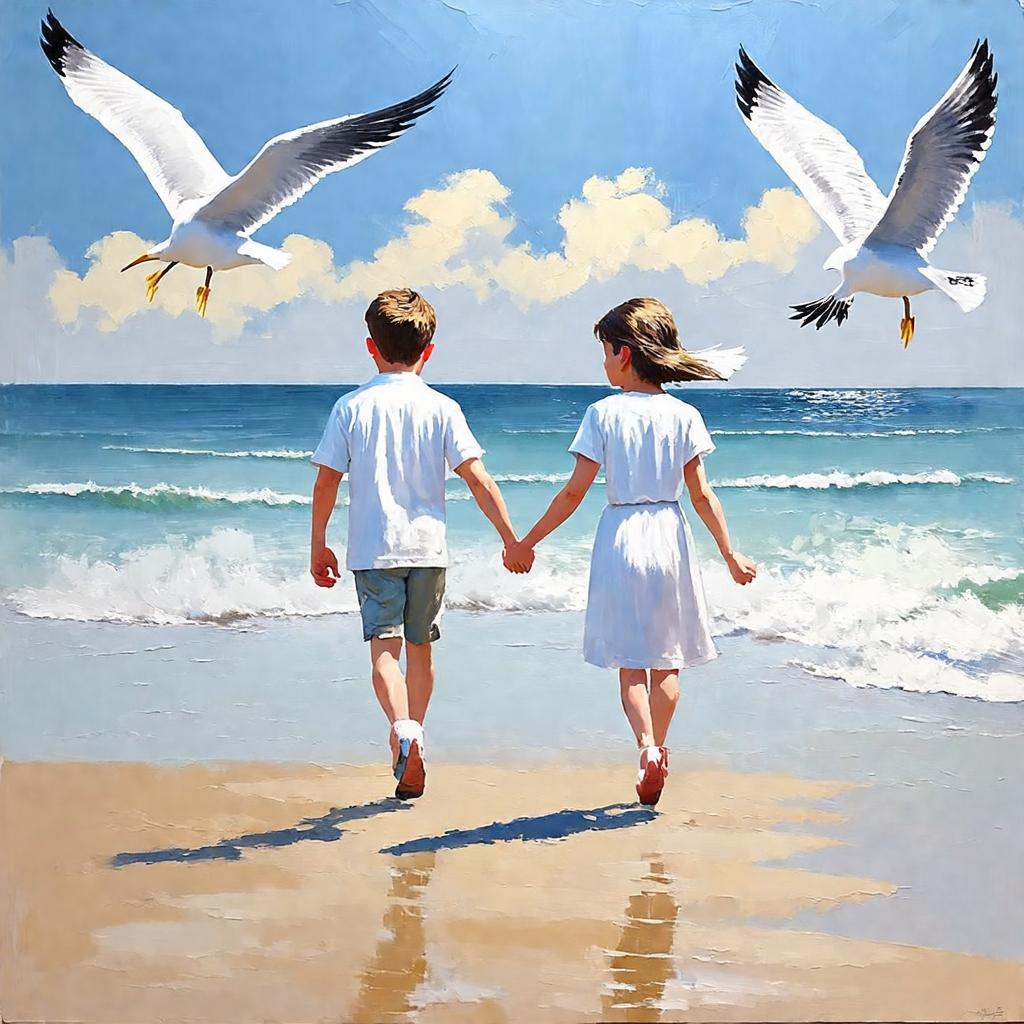}}%
        \fbox{\includegraphics[width=\mainimgwidth]{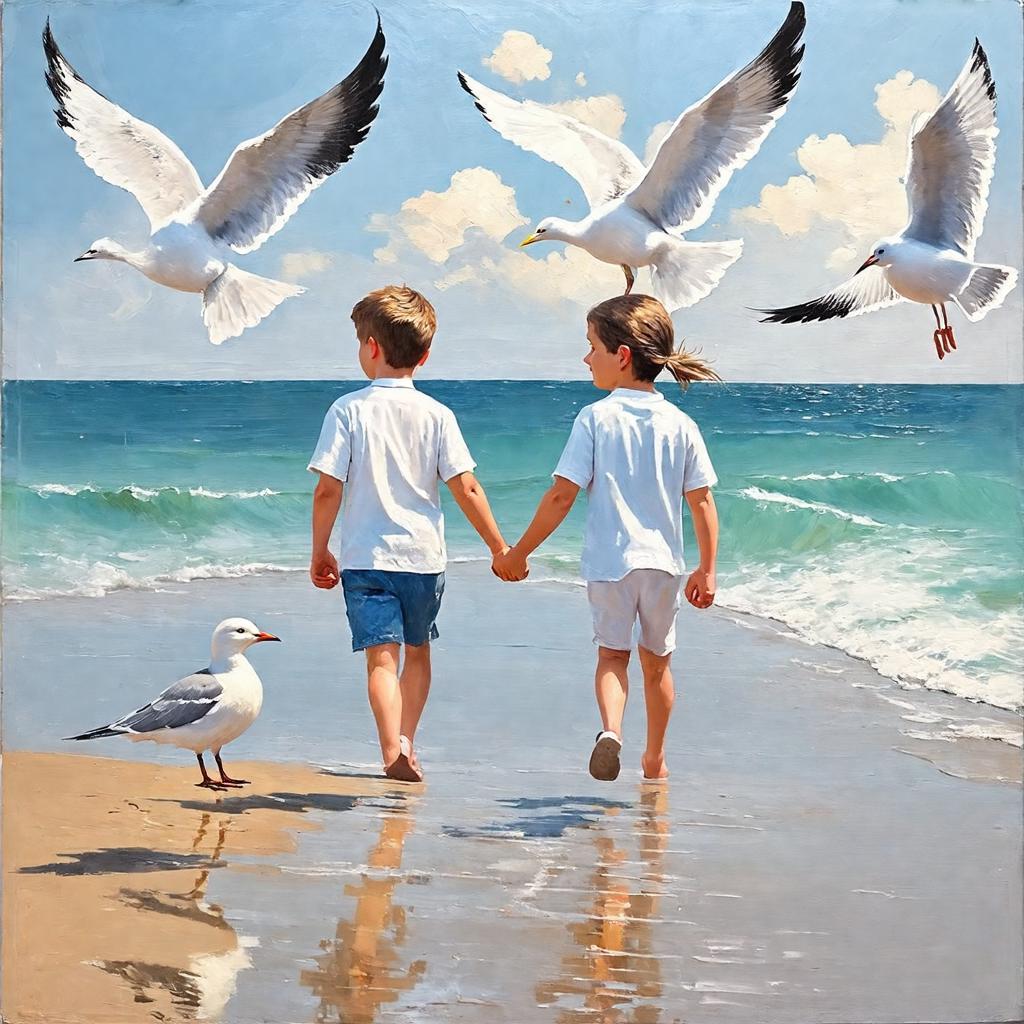}}\\[0.5ex]
        \hfill
        \vspace{-18pt}
        \caption*{
            \begin{minipage}{\maincapwidth}
            \centering
                \tiny{Prompt: \textit{oil painting in sally swatland style, boy and girl holding hands, walking along the sea shore, sea gulls flying}}
            \end{minipage}
        }
    \end{minipage}
    
    \caption{
        More visualization of main results on the MJHQ dataset with SD3.5 and W8A8 DiT quantization.
    }
    \label{fig:app-mjhq-sd3}
\end{figure}

\newcommand{\mainsdxlimgwidth}{0.19\textwidth}
\newcommand{\mainsdxlcapwidth}{0.98\textwidth}
\setlength{\fboxsep}{0pt}
\setlength{\fboxrule}{0.2pt}
\begin{figure}[h!]
    \centering
    \begin{minipage}[t]{0.325\textwidth}
        \centering
        \begin{minipage}[t]{\mainsdxlimgwidth}
            \centering \tiny{FP16}
        \end{minipage}
        \hfill
        \begin{minipage}[t]{\mainsdxlimgwidth}
            \centering \tiny{PTQ4DiT}
        \end{minipage}%
        \begin{minipage}[t]{\mainsdxlimgwidth}
            \centering \tiny{Smooth+}
        \end{minipage}%
        \begin{minipage}[t]{\mainsdxlimgwidth}
            \centering \tiny{\textbf{SegQuant-A}}
        \end{minipage}%
        \begin{minipage}[t]{\mainsdxlimgwidth}
            \centering \tiny{\textbf{SegQuant-G}}
        \end{minipage}
    \end{minipage}
    \hfill
    \begin{minipage}[t]{0.325\textwidth}
        \centering
        \begin{minipage}[t]{\mainsdxlimgwidth}
            \centering \tiny{FP16}
        \end{minipage}
        \hfill
        \begin{minipage}[t]{\mainsdxlimgwidth}
            \centering \tiny{PTQ4DiT}
        \end{minipage}%
        \begin{minipage}[t]{\mainsdxlimgwidth}
            \centering \tiny{Smooth+}
        \end{minipage}%
        \begin{minipage}[t]{\mainsdxlimgwidth}
            \centering \tiny{\textbf{SegQuant-A}}
        \end{minipage}%
        \begin{minipage}[t]{\mainsdxlimgwidth}
            \centering \tiny{\textbf{SegQuant-G}}
        \end{minipage}
    \end{minipage}
    \hfill
    \begin{minipage}[t]{0.325\textwidth}
        \centering
        \begin{minipage}[t]{\mainsdxlimgwidth}
            \centering \tiny{FP16}
        \end{minipage}
        \hfill
        \begin{minipage}[t]{\mainsdxlimgwidth}
            \centering \tiny{PTQ4DiT}
        \end{minipage}%
        \begin{minipage}[t]{\mainsdxlimgwidth}
            \centering \tiny{Smooth+}
        \end{minipage}%
        \begin{minipage}[t]{\mainsdxlimgwidth}
            \centering \tiny{\textbf{SegQuant-A}}
        \end{minipage}%
        \begin{minipage}[t]{\mainsdxlimgwidth}
            \centering \tiny{\textbf{SegQuant-G}}
        \end{minipage}
    \end{minipage}
    \vspace{0.6ex}

    \begin{minipage}[t]{0.325\textwidth}
        \centering
        \fbox{\includegraphics[width=\mainsdxlimgwidth]{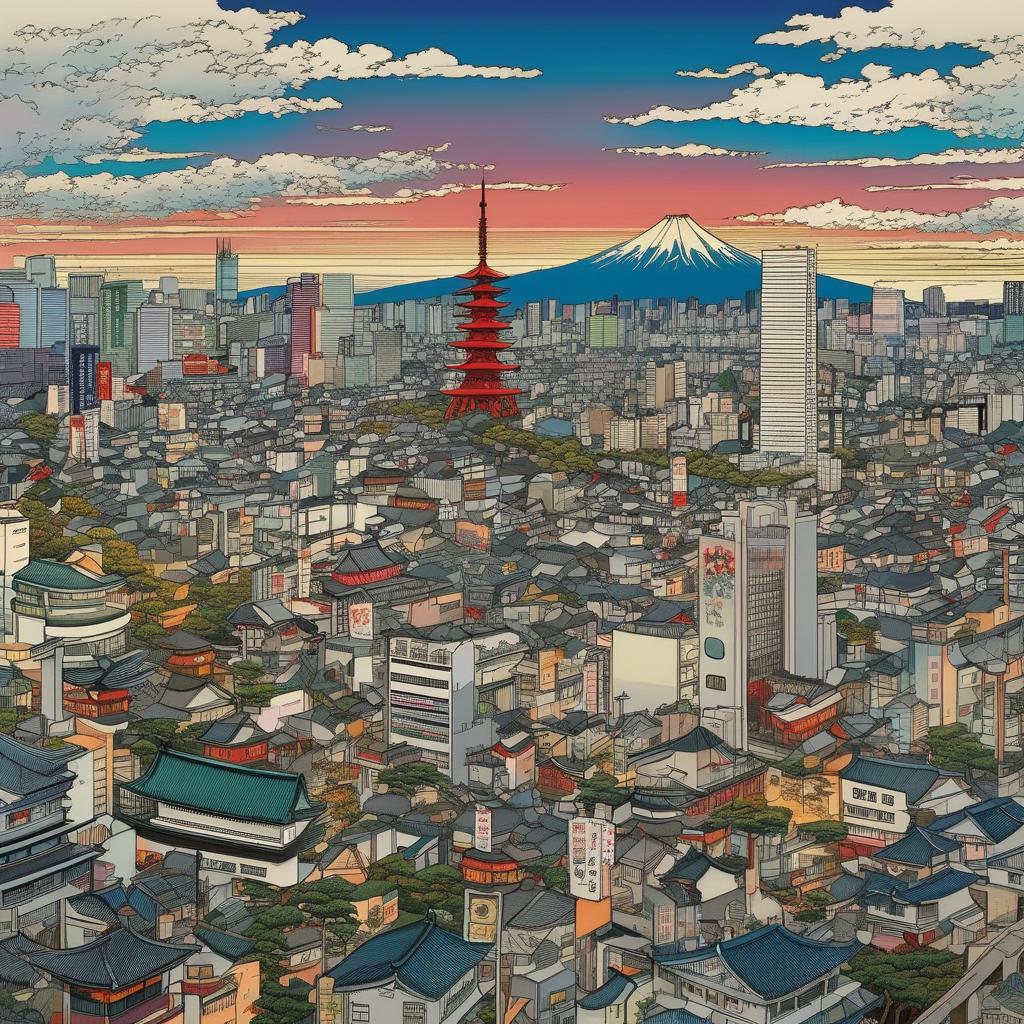}}
        \hfill
        \fbox{\includegraphics[width=\mainsdxlimgwidth]{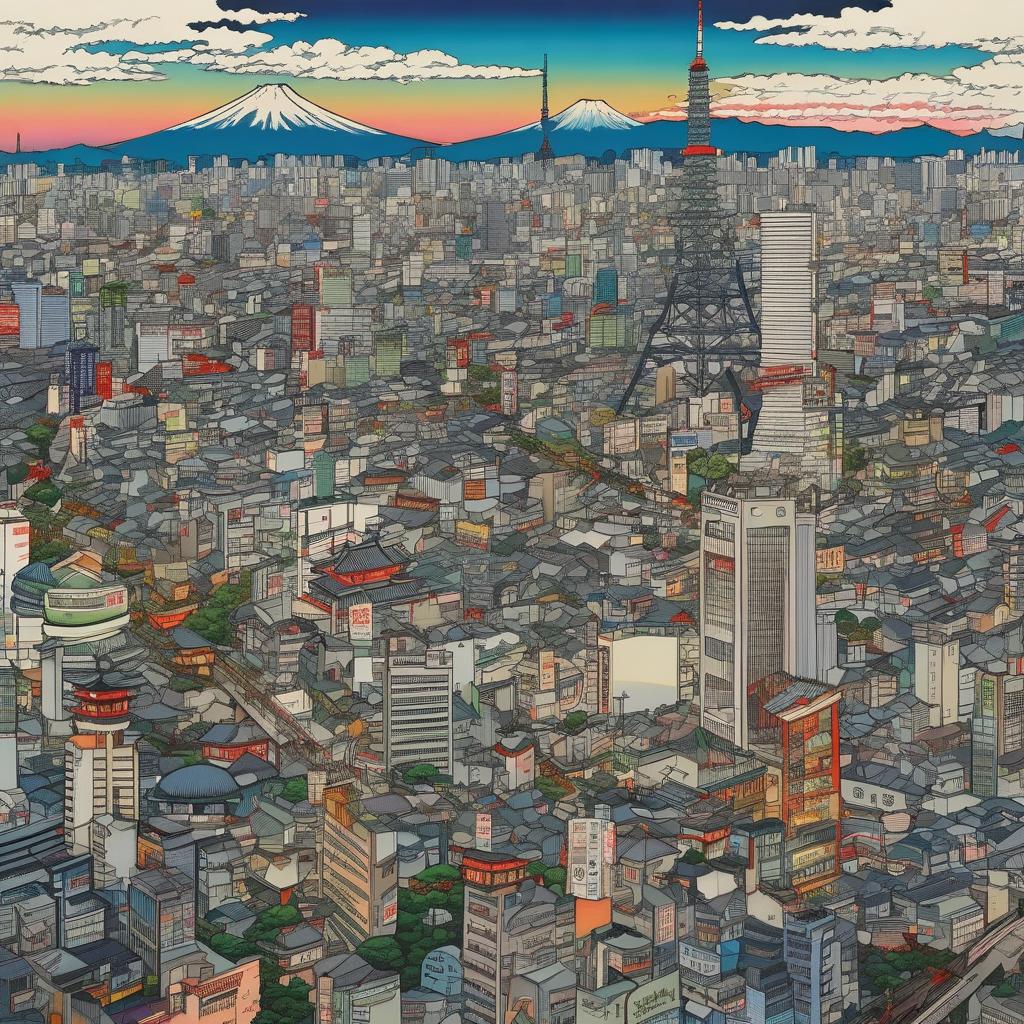}}%
        \fbox{\includegraphics[width=\mainsdxlimgwidth]{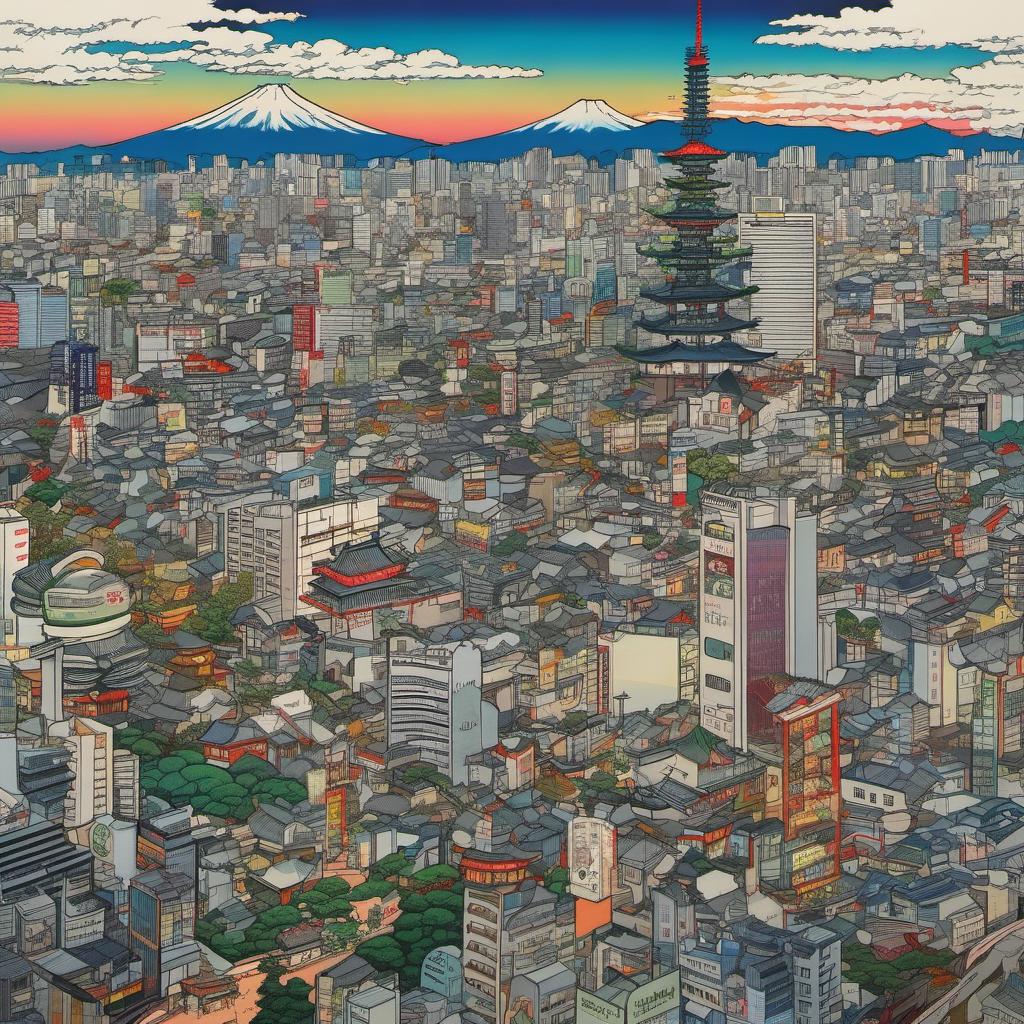}}%
        \fbox{\includegraphics[width=\mainsdxlimgwidth]{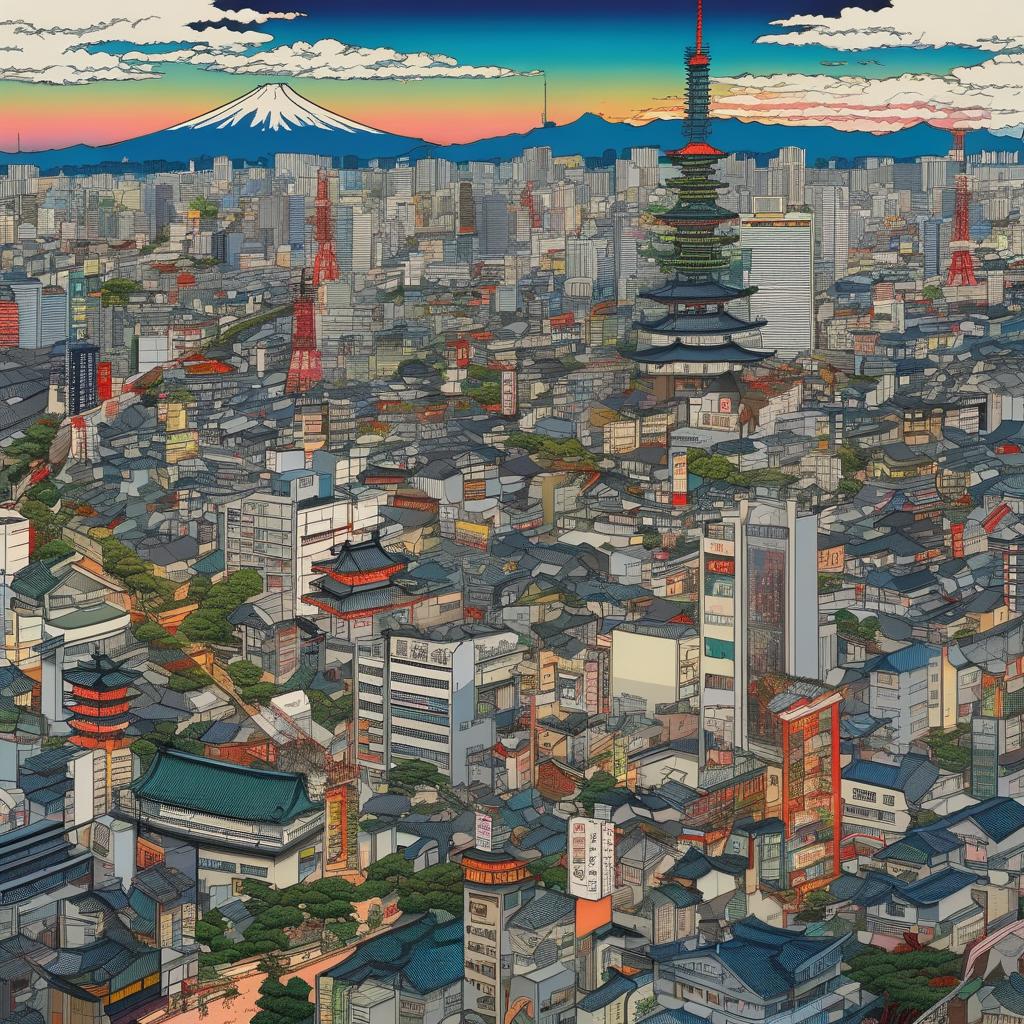}}%
        \fbox{\includegraphics[width=\mainsdxlimgwidth]{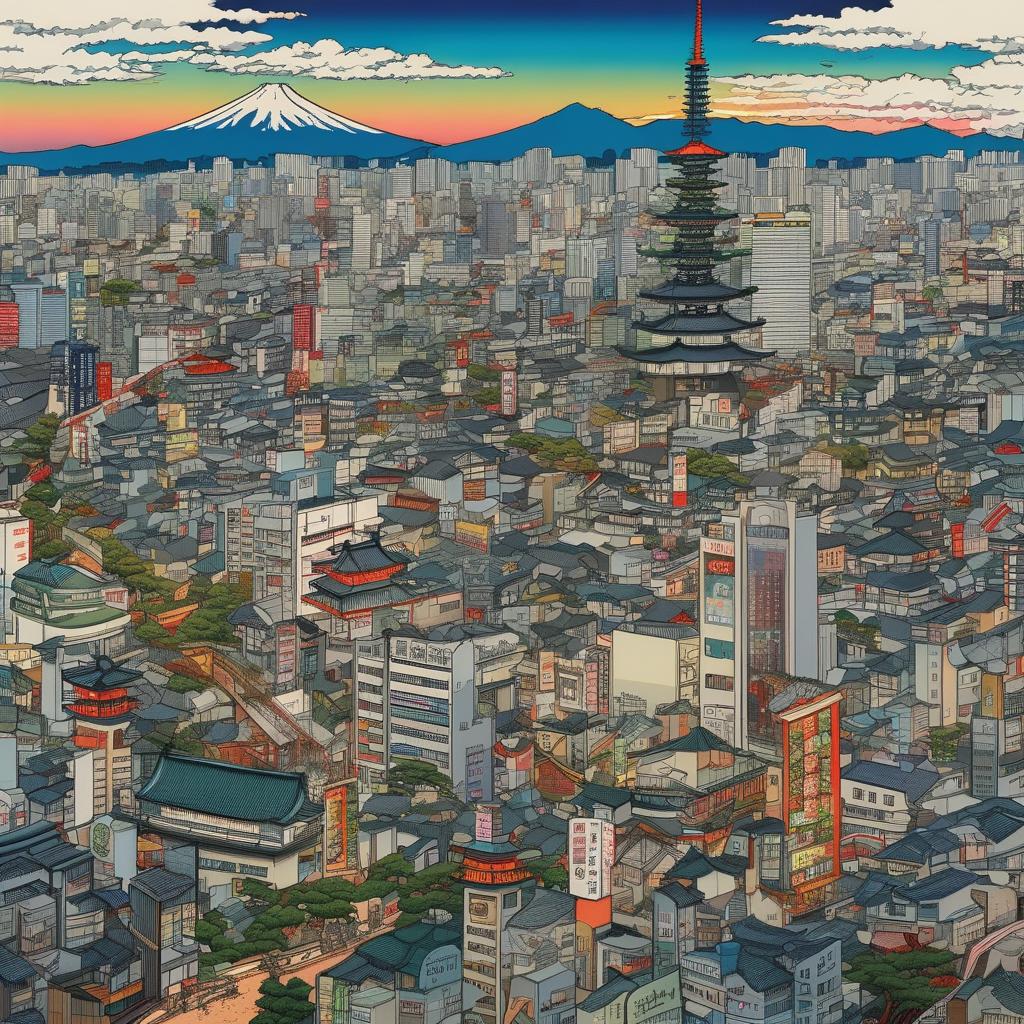}}\\[0.5ex]
        \hfill
        \vspace{-18pt}
        \caption*{
            \begin{minipage}{\mainsdxlcapwidth}
            \centering
                \tiny{Prompt: \textit{ukiyoe painting of tokyo city, colorful, high resolution}}
            \end{minipage}
        }
    \end{minipage}
    \hfill
    \begin{minipage}[t]{0.325\textwidth}
        \centering
        \fbox{\includegraphics[width=\mainsdxlimgwidth]{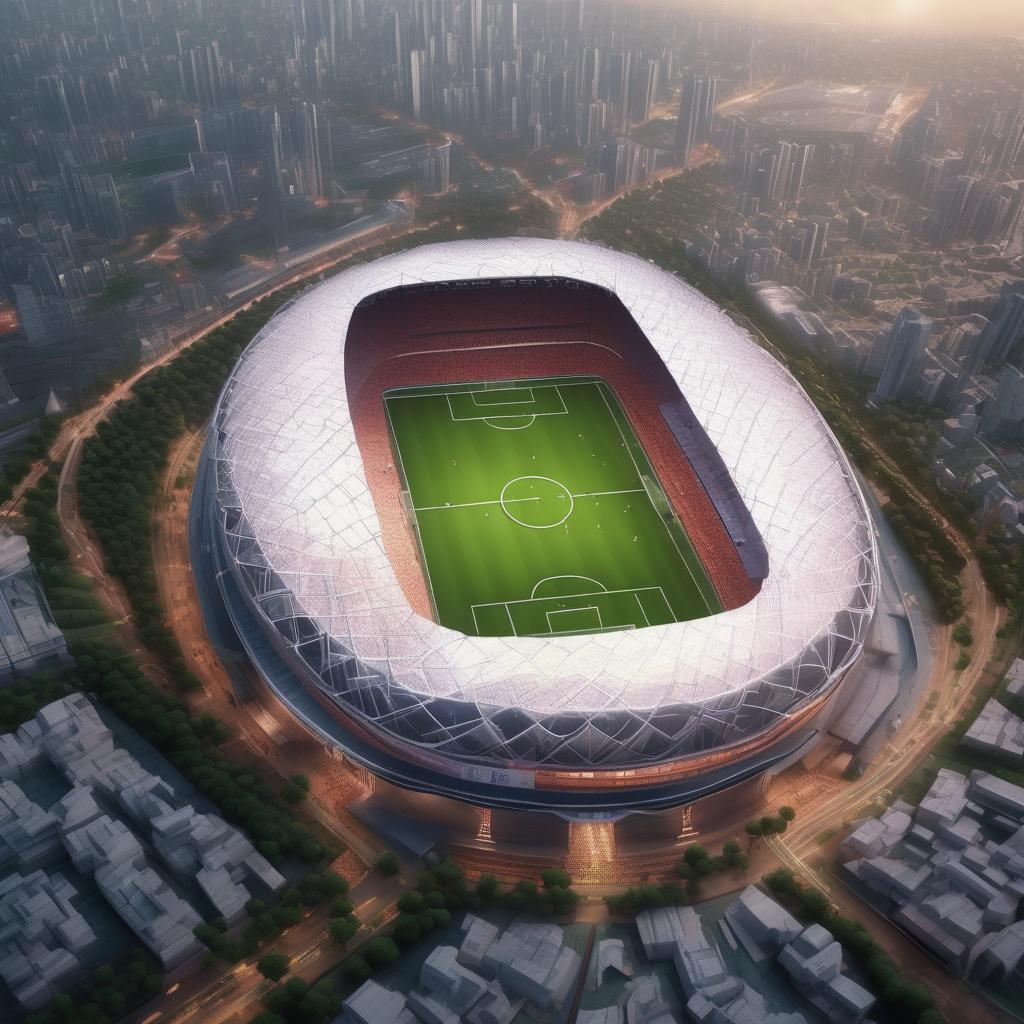}}
        \hfill
        \fbox{\includegraphics[width=\mainsdxlimgwidth]{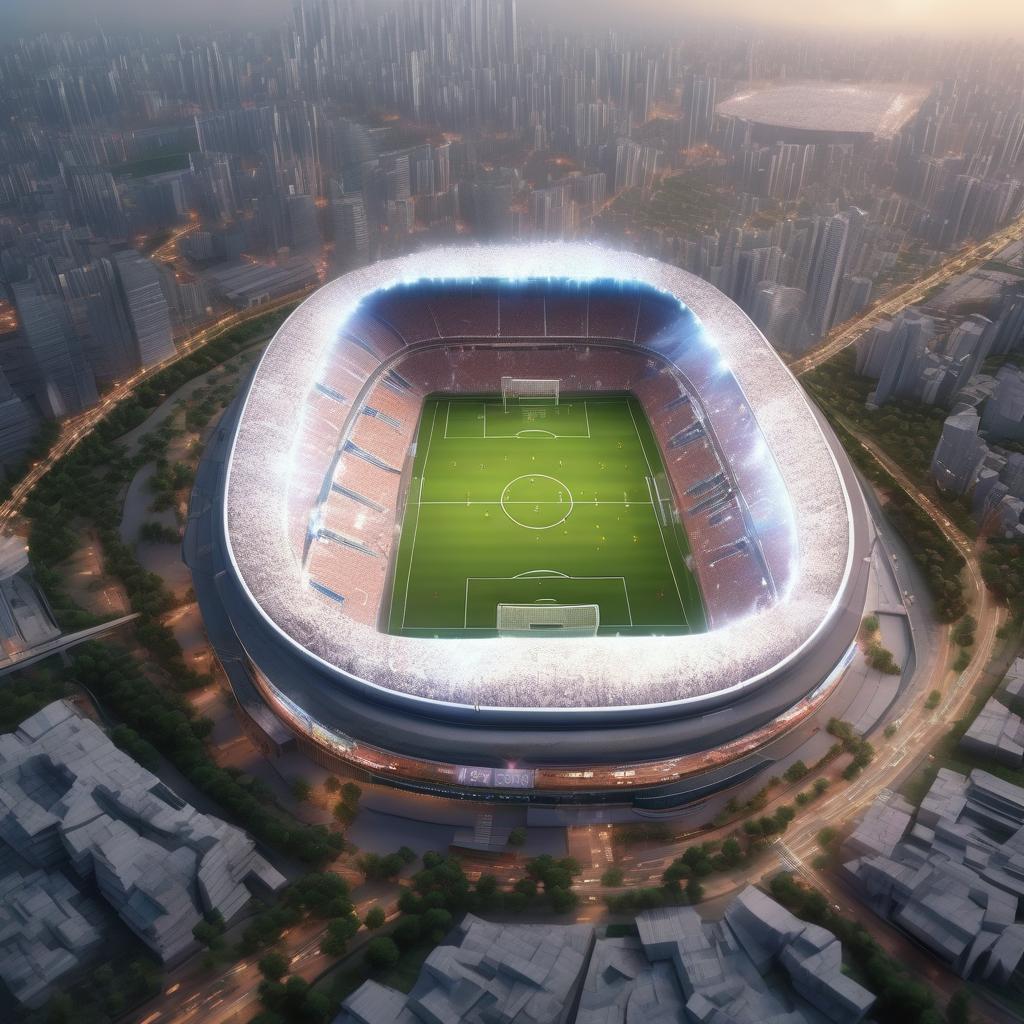}}%
        \fbox{\includegraphics[width=\mainsdxlimgwidth]{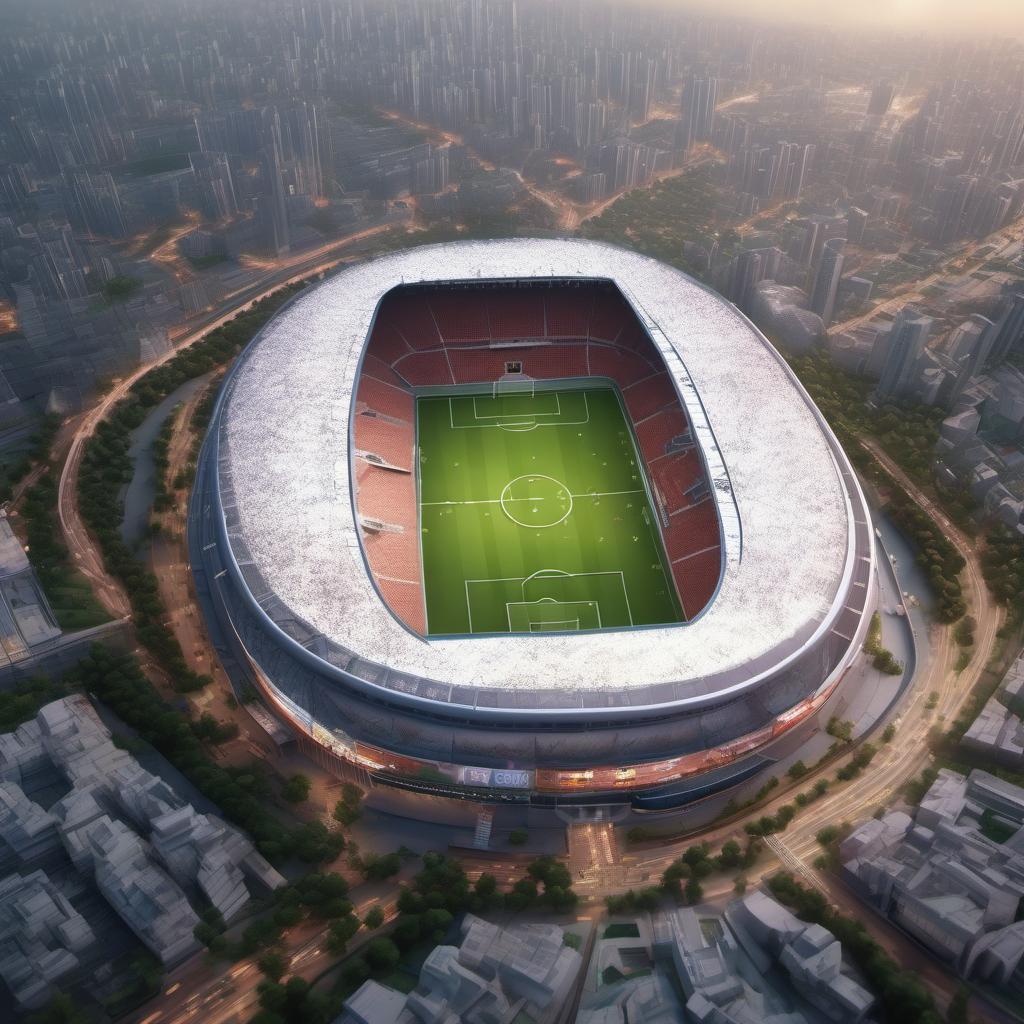}}%
        \fbox{\includegraphics[width=\mainsdxlimgwidth]{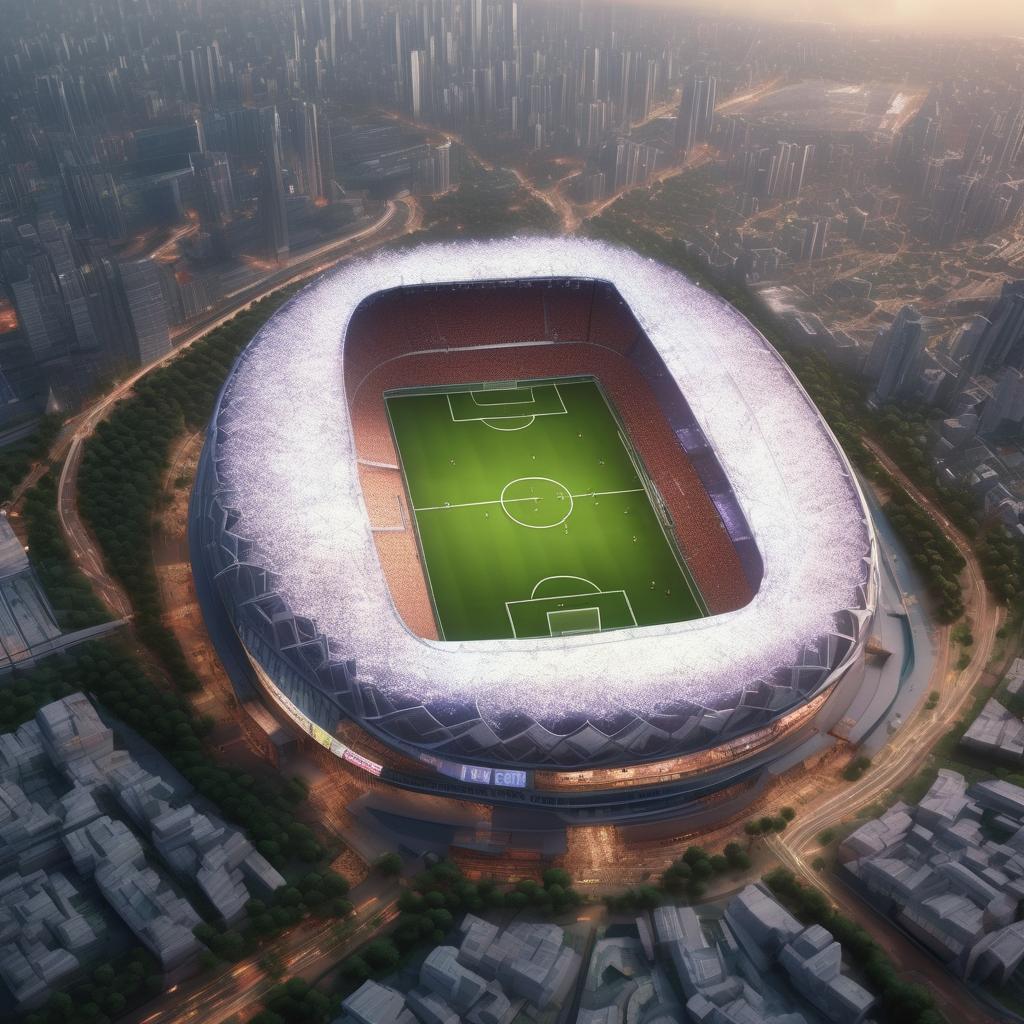}}%
        \fbox{\includegraphics[width=\mainsdxlimgwidth]{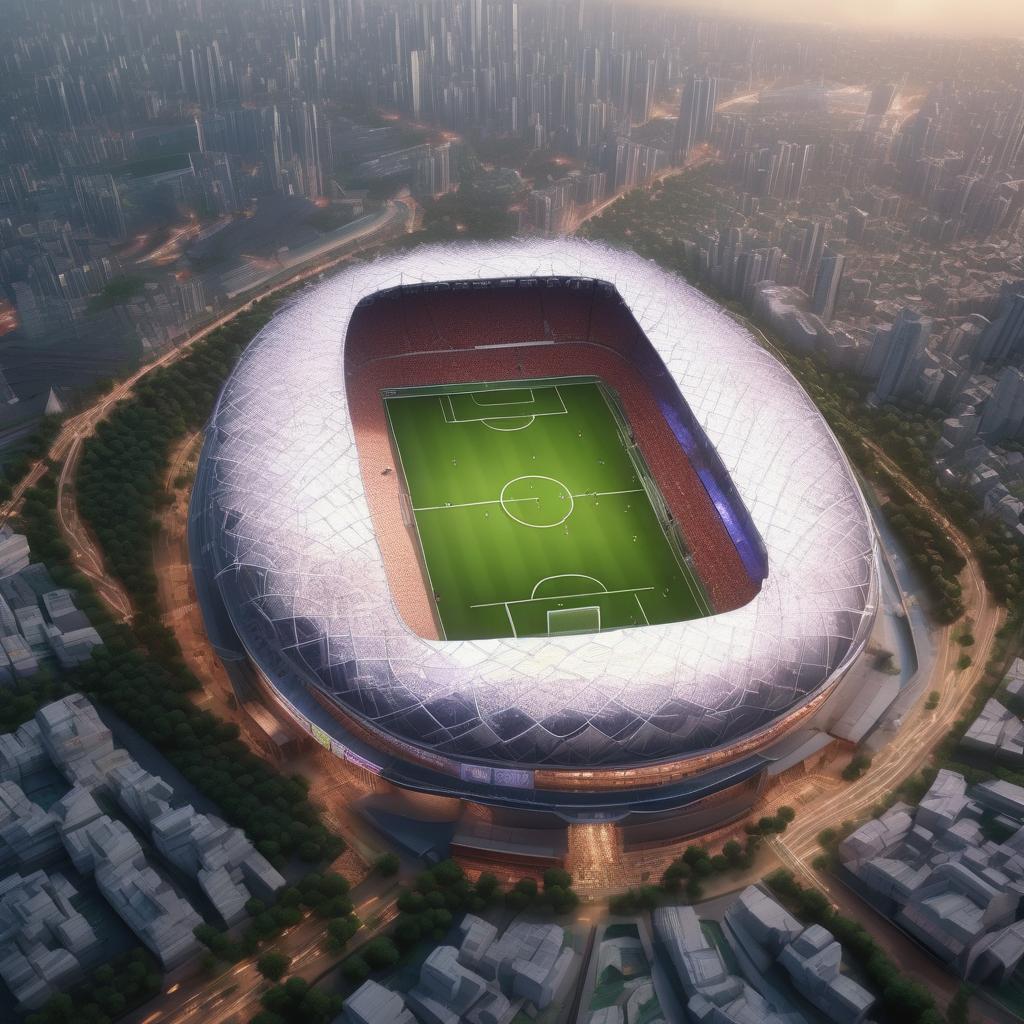}}\\[0.5ex]
        \hfill
        \vspace{-18pt}
        \caption*{
            \begin{minipage}{\mainsdxlcapwidth}
            \centering
                \tiny{Prompt: \textit{Photorealistic, Realistic aerial view 3D big render soccer stadium, asian style, summer time, fantastical lightning, in the middle of an big asian city, full detailed, resulting in an intricately detailed and wondrous image, 8k.}}
            \end{minipage}
        }
    \end{minipage}
    \hfill
    \begin{minipage}[t]{0.325\textwidth}
        \centering
        \fbox{\includegraphics[width=\mainsdxlimgwidth]{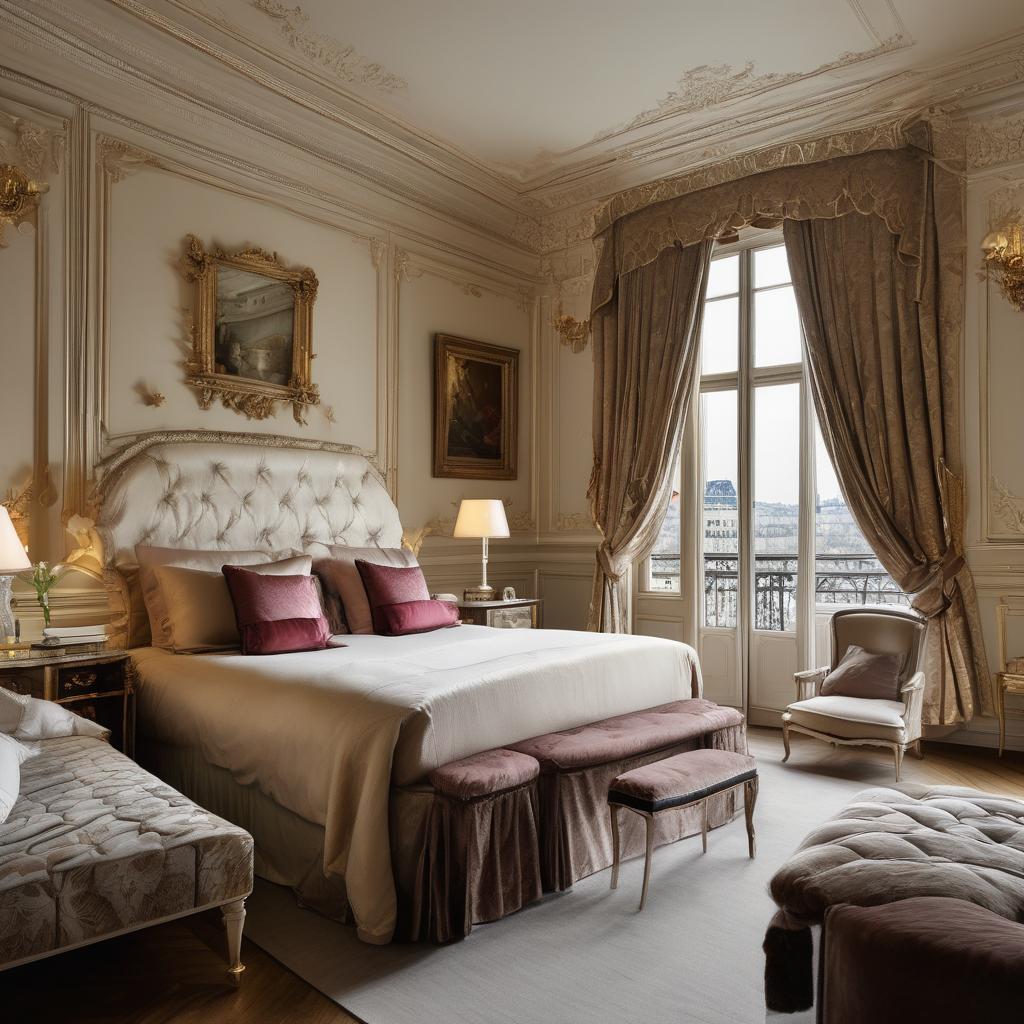}}
        \hfill
        \fbox{\includegraphics[width=\mainsdxlimgwidth]{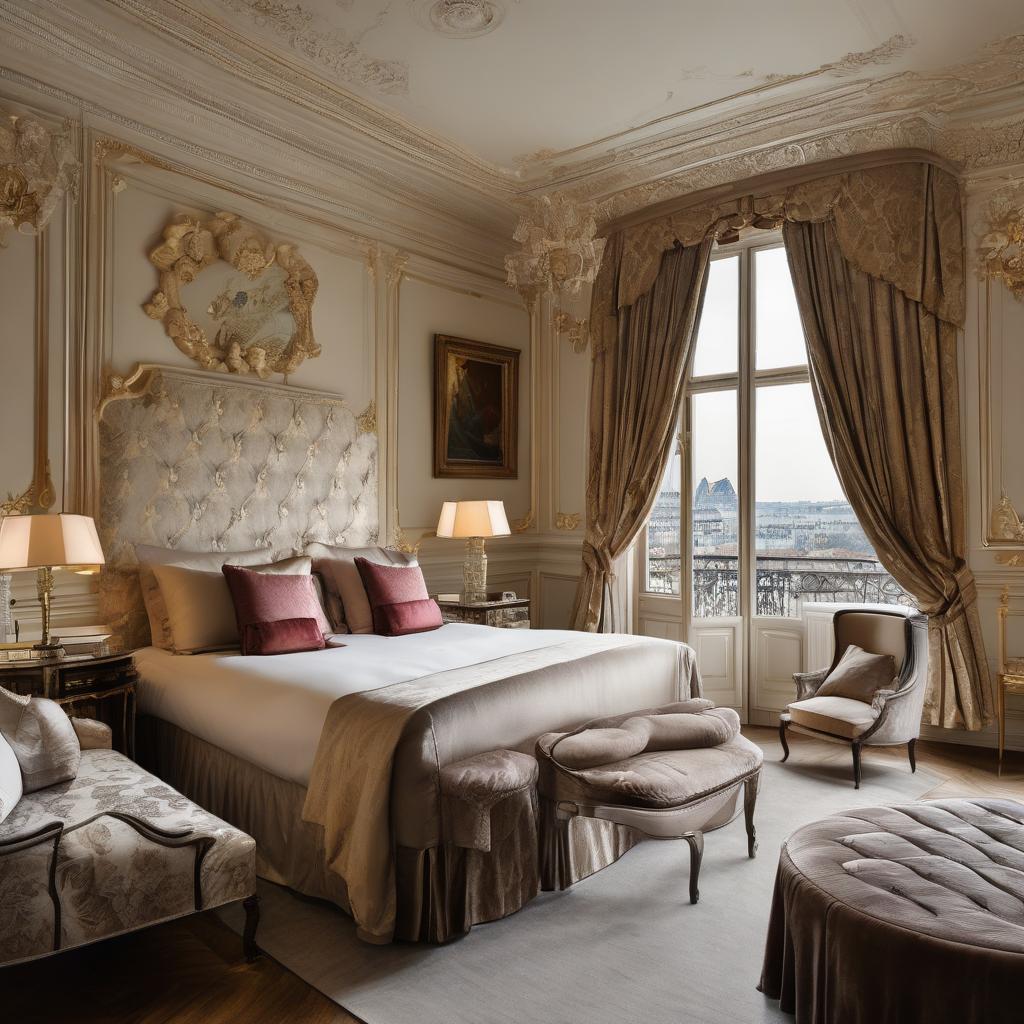}}%
        \fbox{\includegraphics[width=\mainsdxlimgwidth]{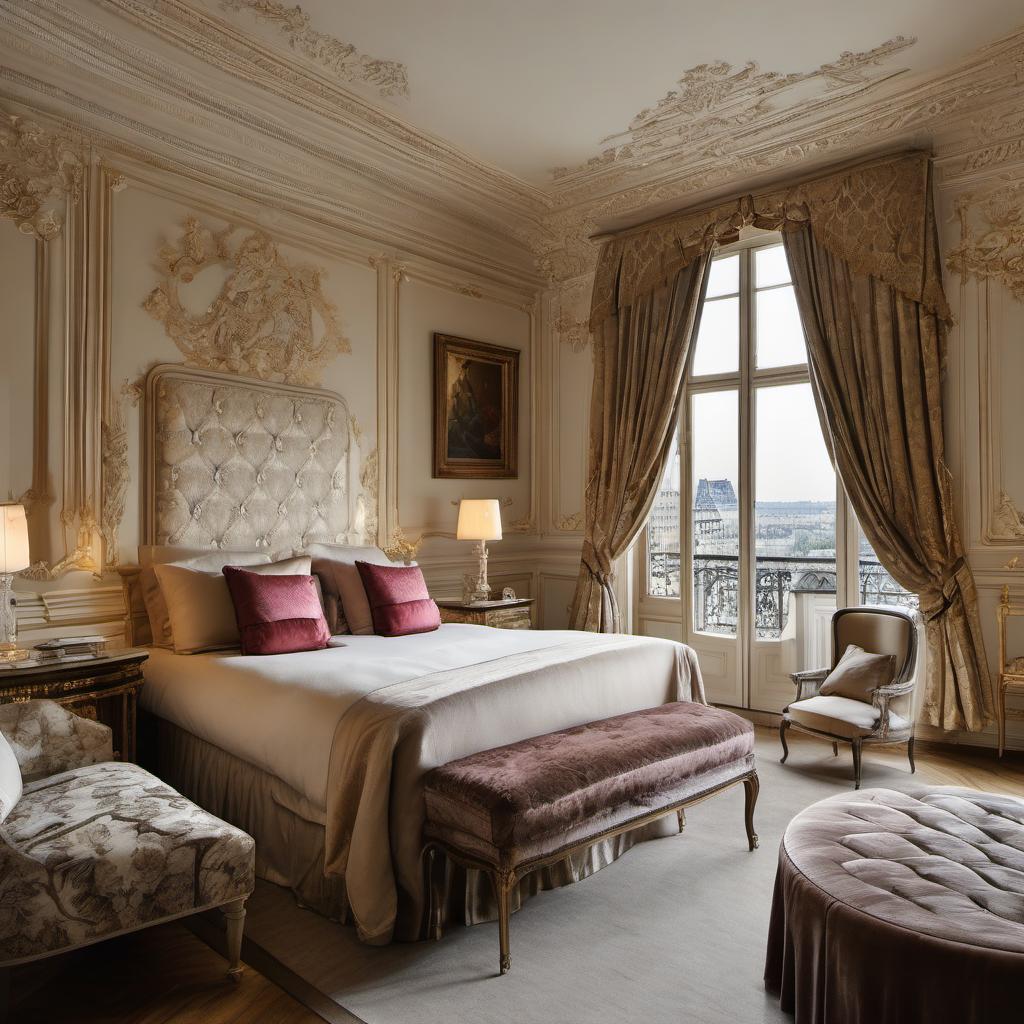}}%
        \fbox{\includegraphics[width=\mainsdxlimgwidth]{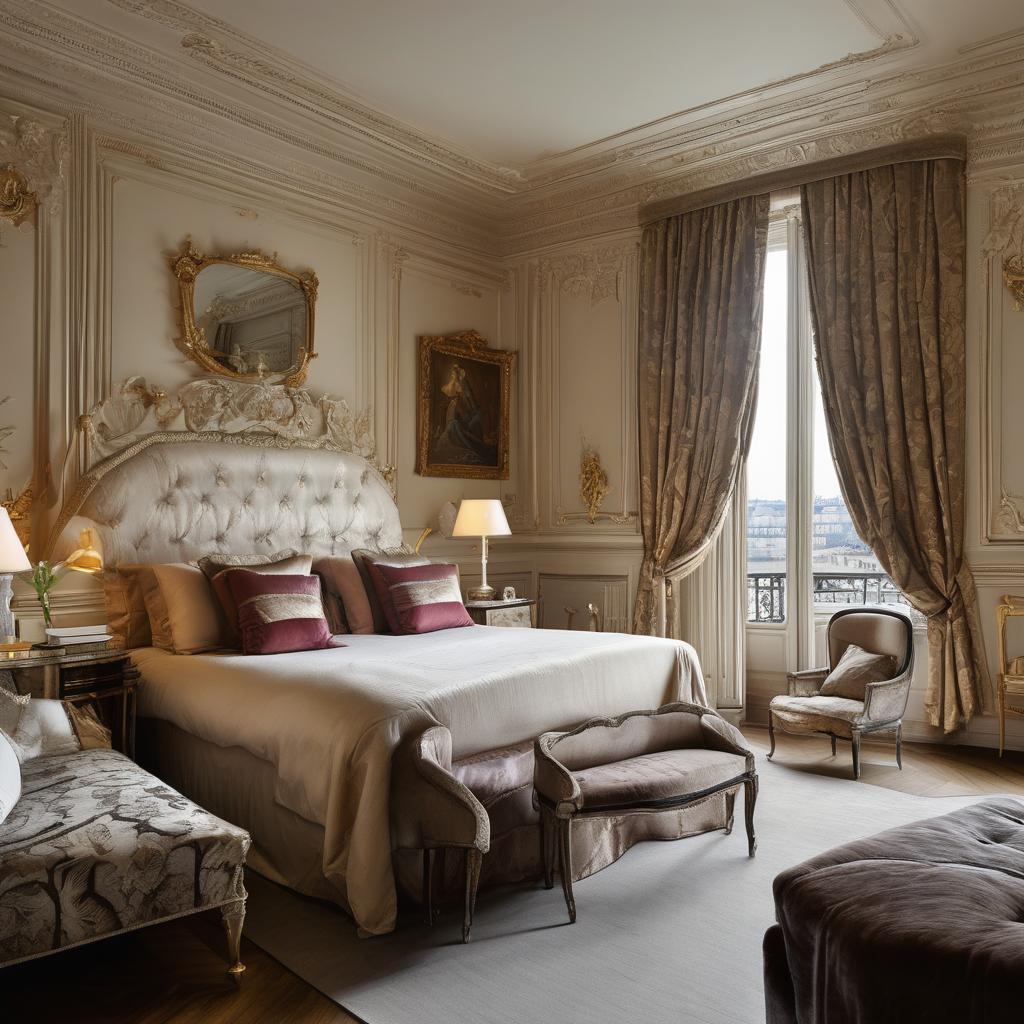}}%
        \fbox{\includegraphics[width=\mainsdxlimgwidth]{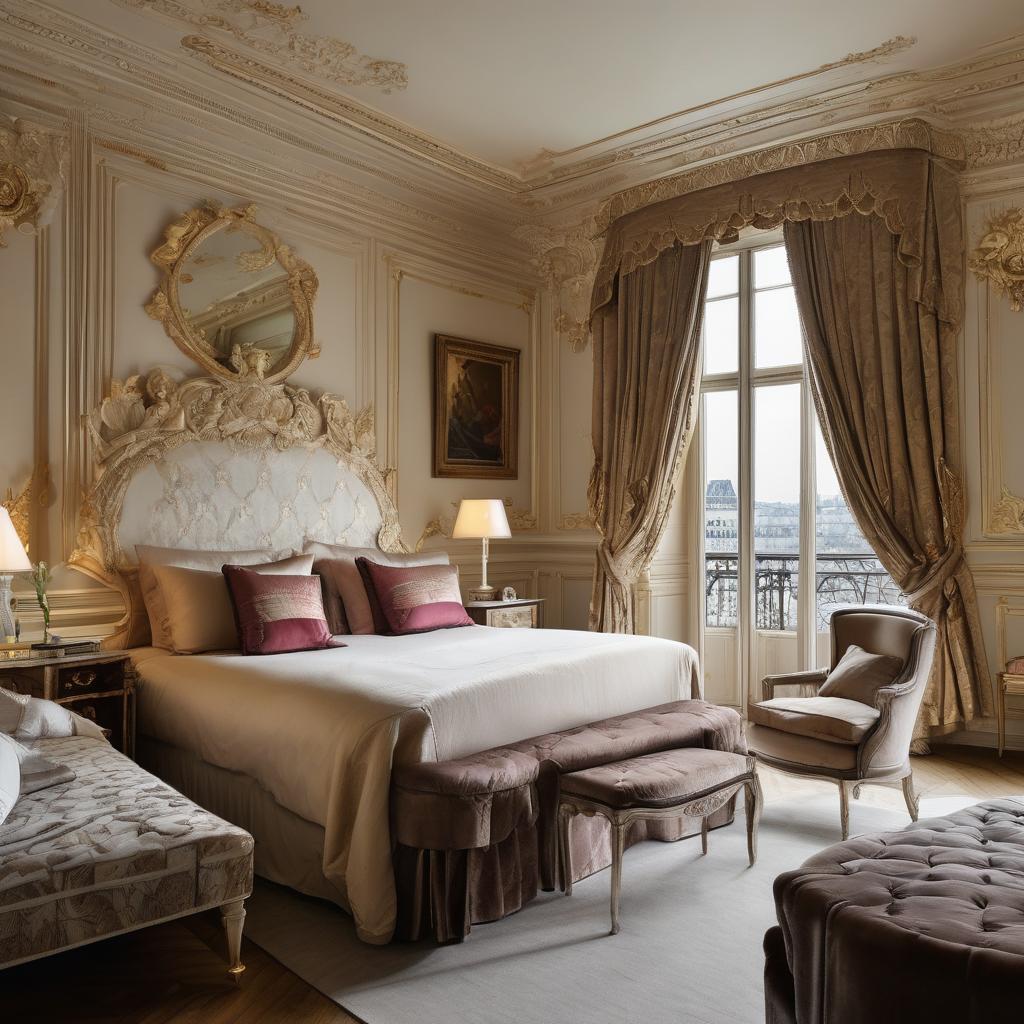}}\\[0.5ex]
        \hfill
        \vspace{-18pt}
        \caption*{
            \begin{minipage}{\mainsdxlcapwidth}
            \centering
                \tiny{Prompt: \textit{a luxury bedroom with a view of paris}}
            \end{minipage}
        }
    \end{minipage}

    \begin{minipage}[t]{0.325\textwidth}
        \centering
        \fbox{\includegraphics[width=\mainsdxlimgwidth]{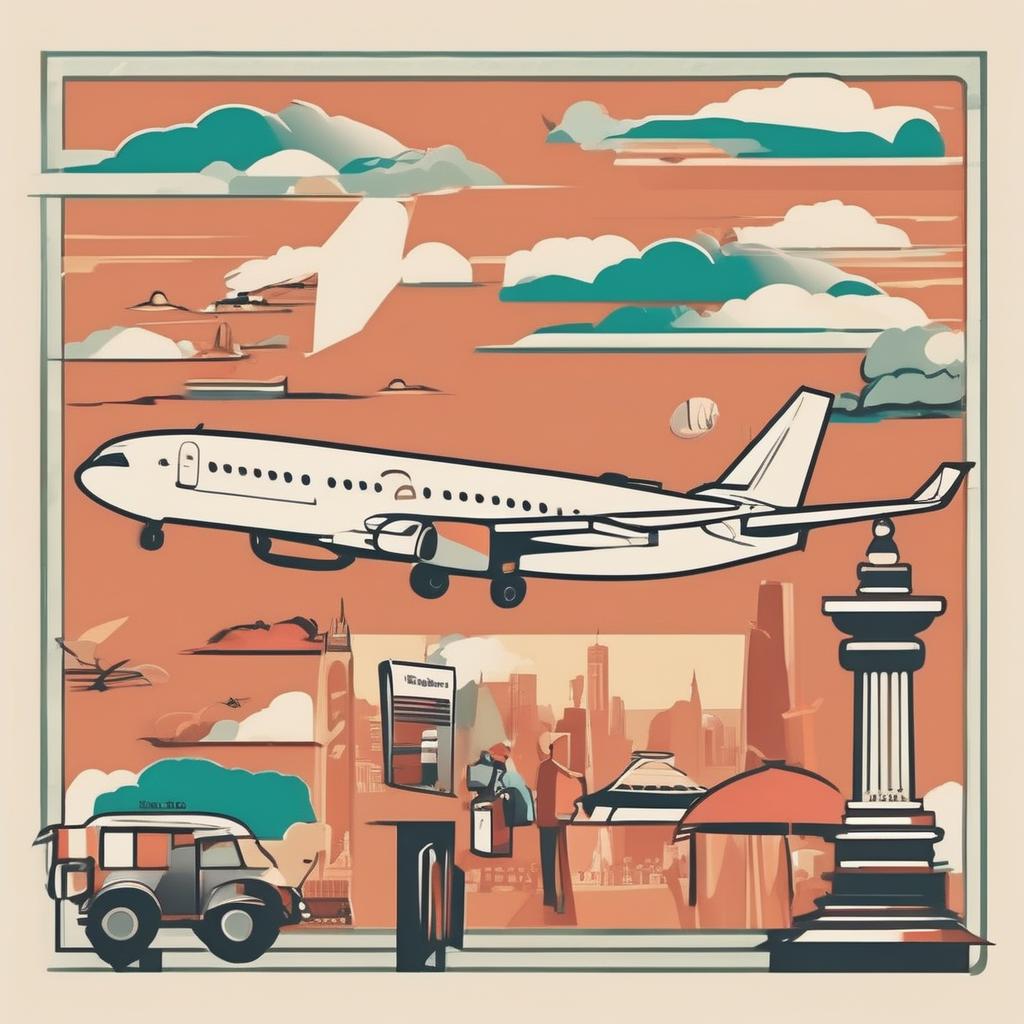}}
        \hfill
        \fbox{\includegraphics[width=\mainsdxlimgwidth]{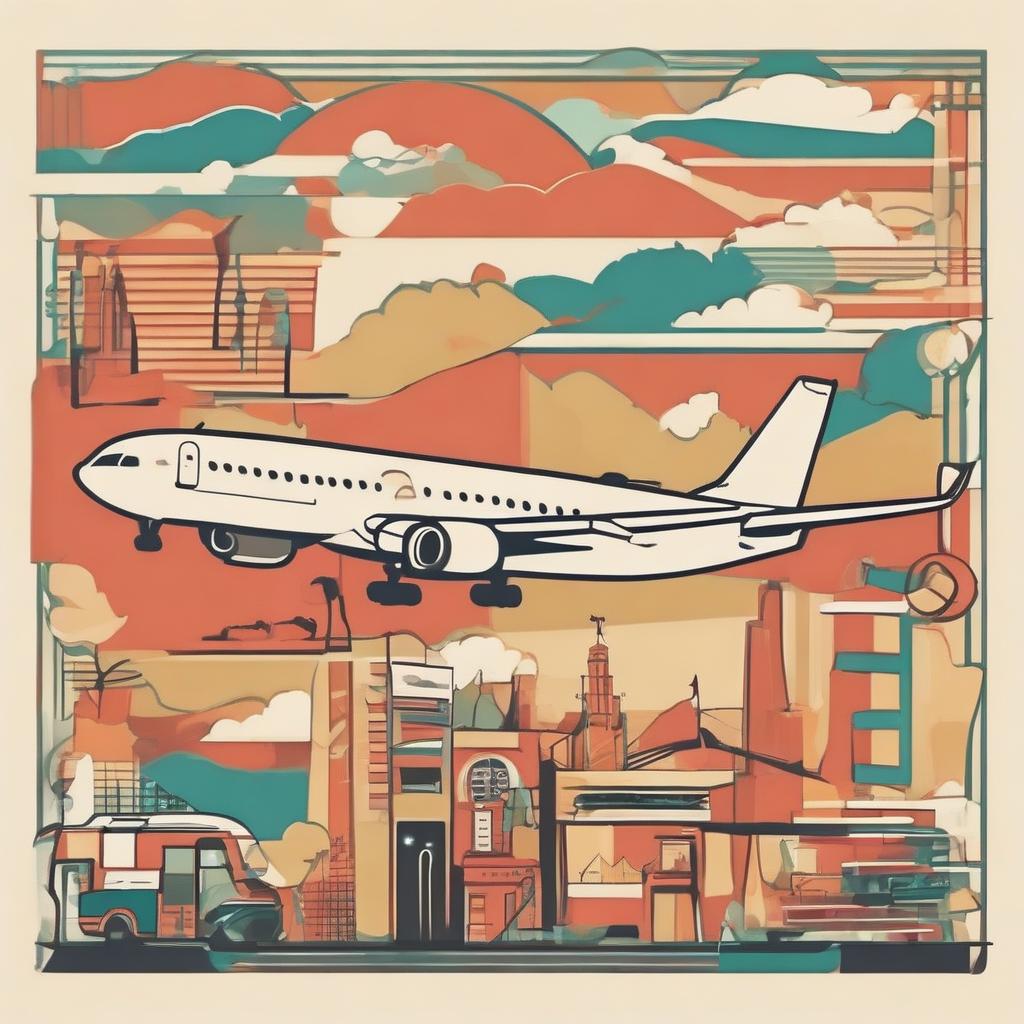}}%
        \fbox{\includegraphics[width=\mainsdxlimgwidth]{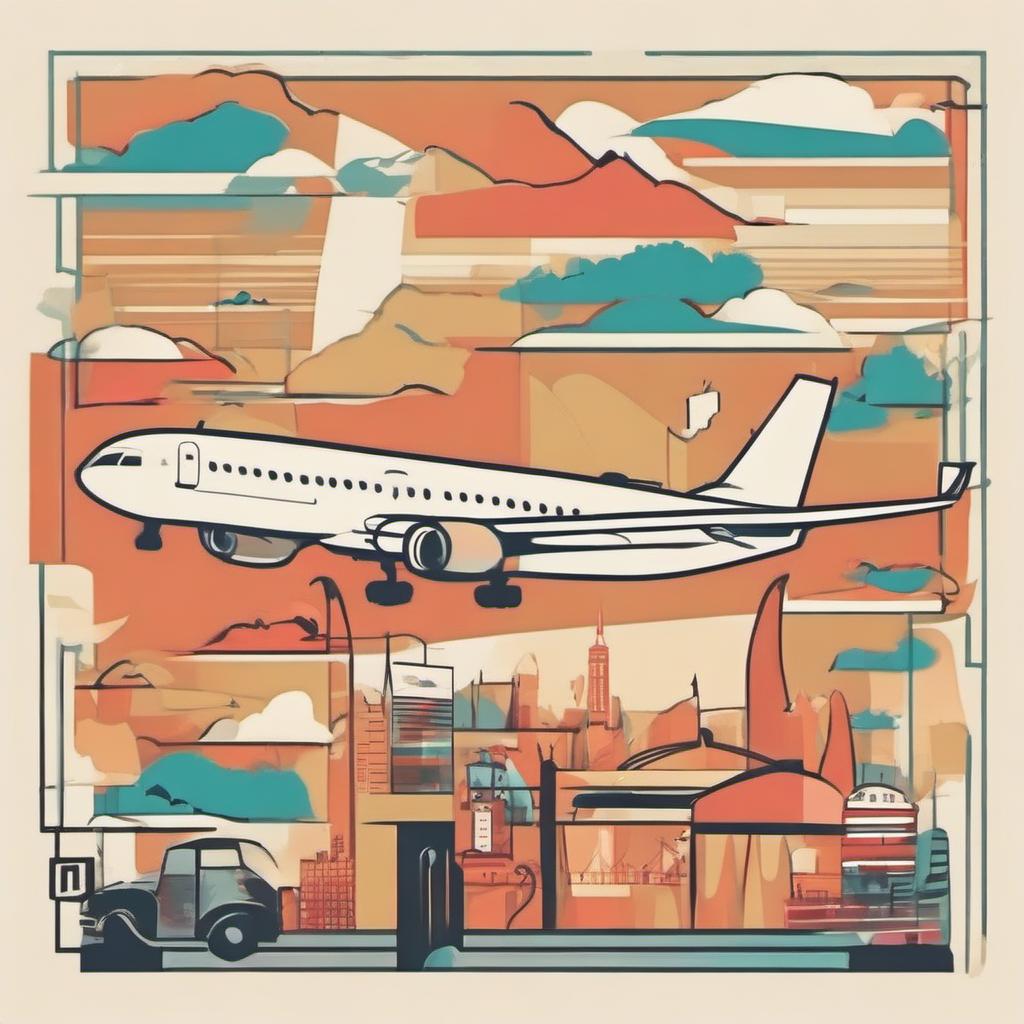}}%
        \fbox{\includegraphics[width=\mainsdxlimgwidth]{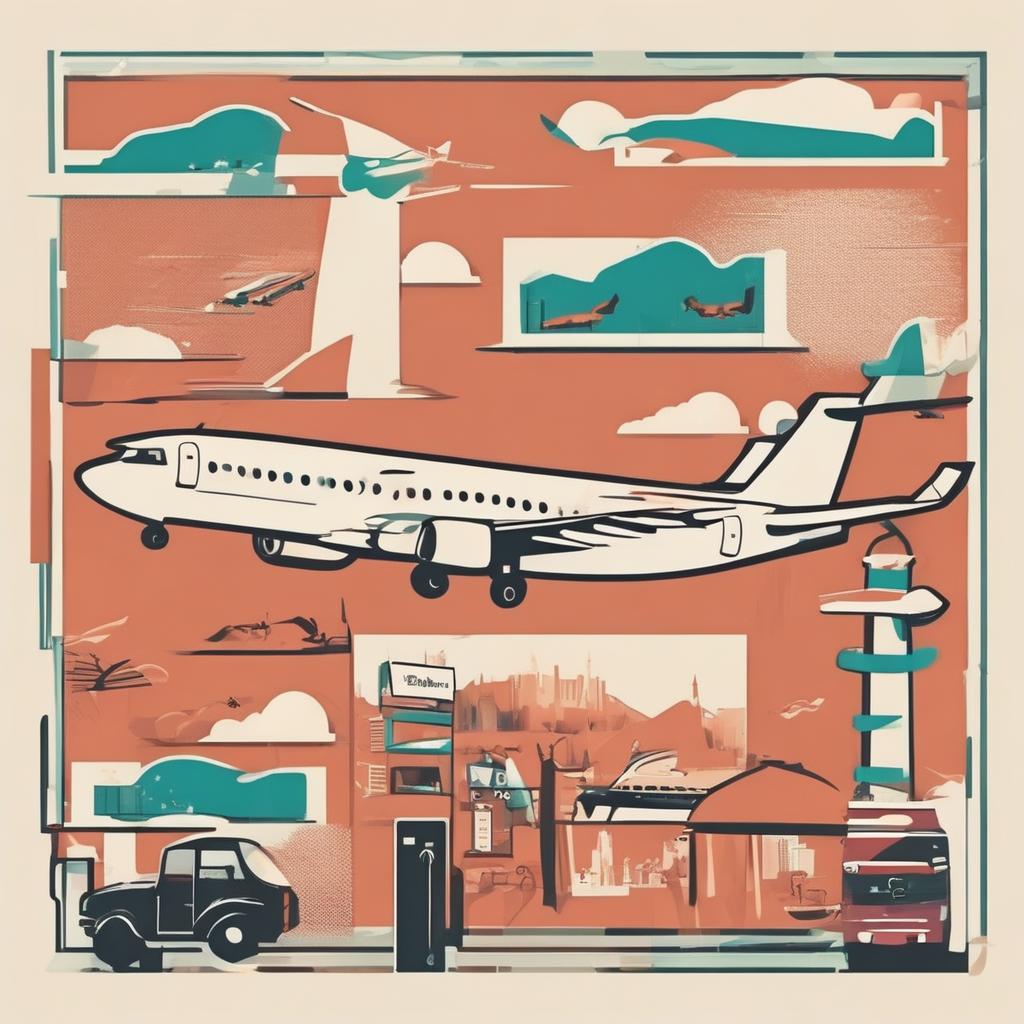}}%
        \fbox{\includegraphics[width=\mainsdxlimgwidth]{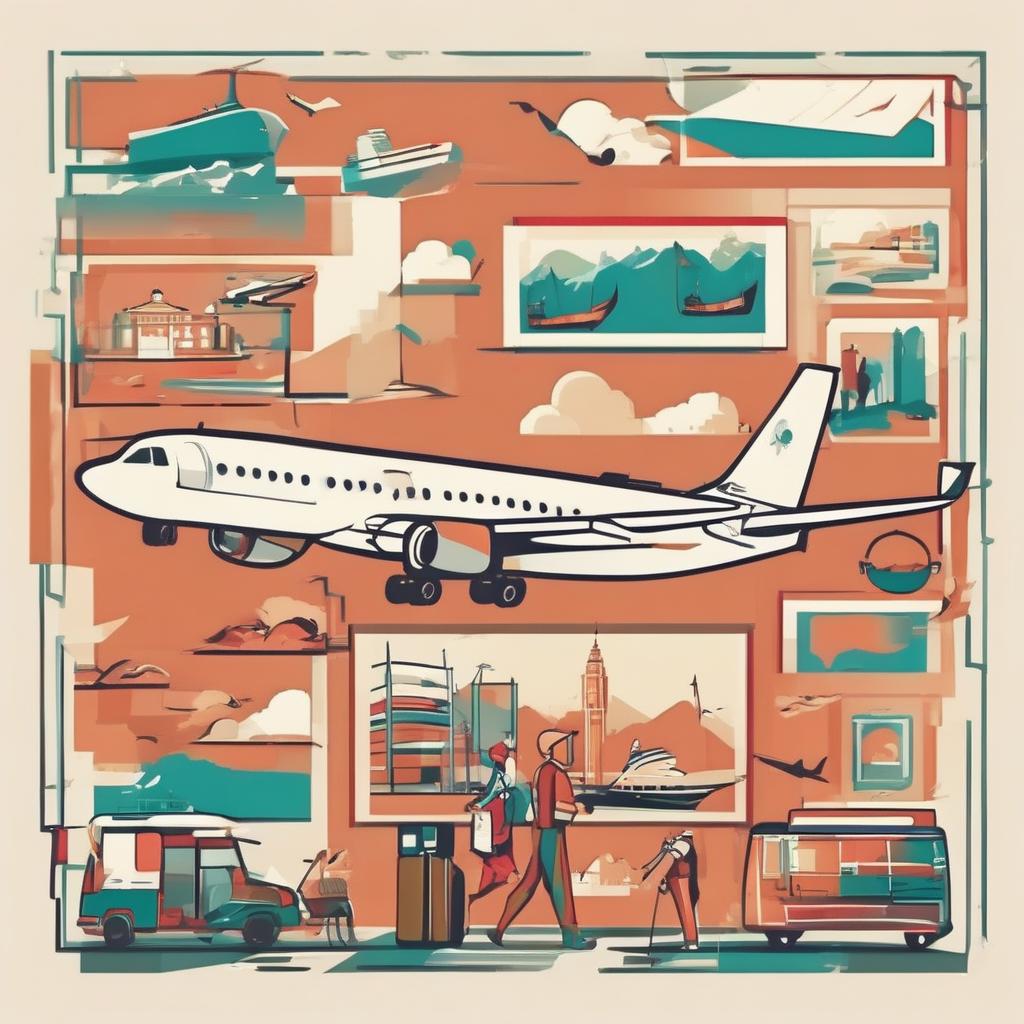}}\\[0.5ex]
        \hfill
        \vspace{-18pt}
        \caption*{
            \begin{minipage}{\mainsdxlcapwidth}
            \centering
                \tiny{Prompt: \textit{a minimalistic 3 colored and artistic illustration that describes a travel company without any letters or text included. make it in the format of 916}}
            \end{minipage}
        }
    \end{minipage}
    \hfill
    \begin{minipage}[t]{0.325\textwidth}
        \centering
        \fbox{\includegraphics[width=\mainsdxlimgwidth]{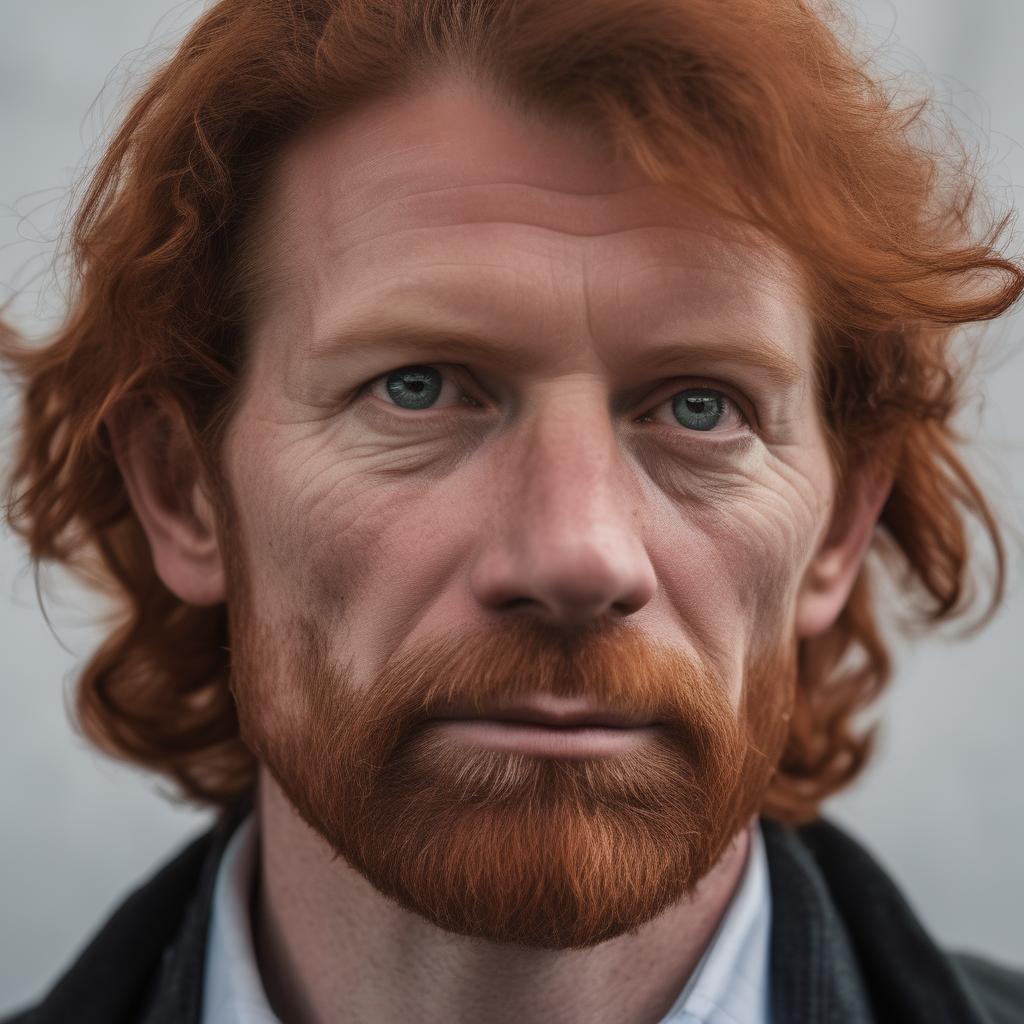}}
        \hfill
        \fbox{\includegraphics[width=\mainsdxlimgwidth]{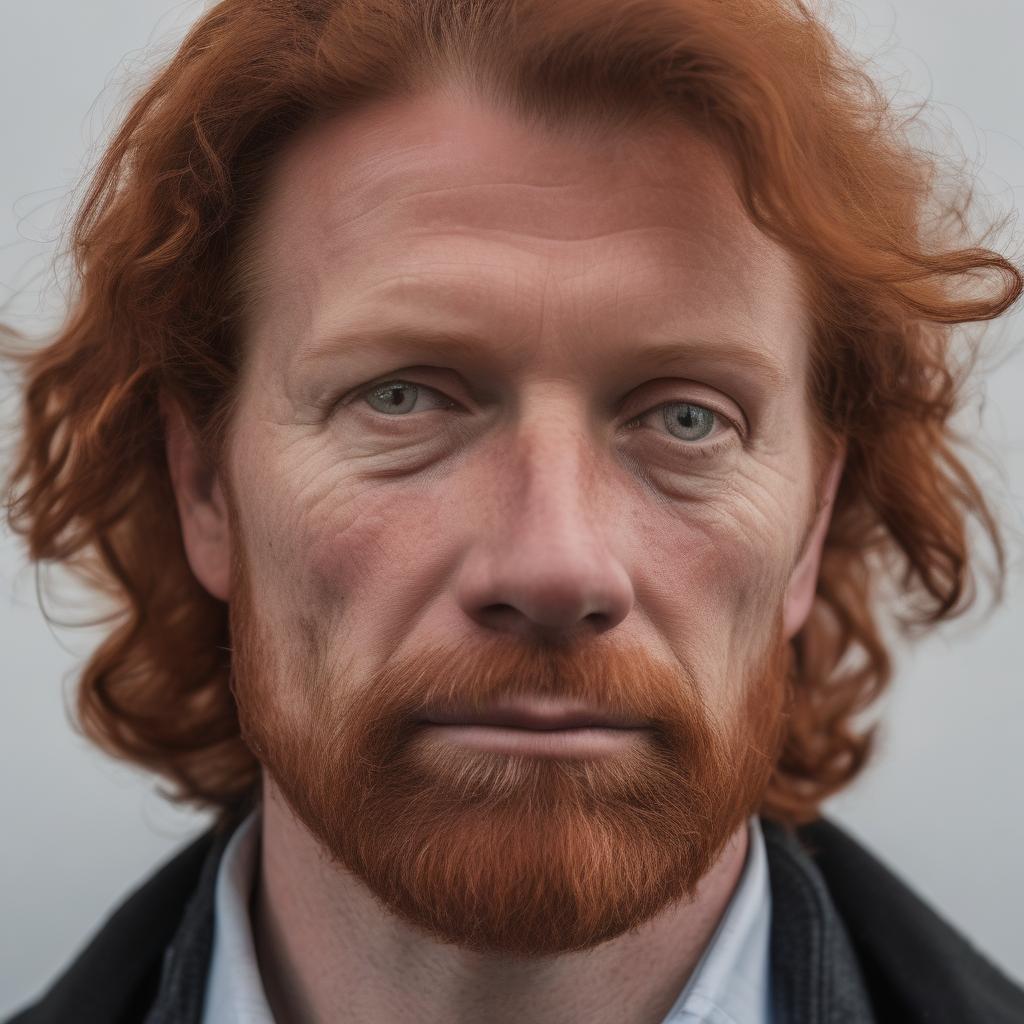}}%
        \fbox{\includegraphics[width=\mainsdxlimgwidth]{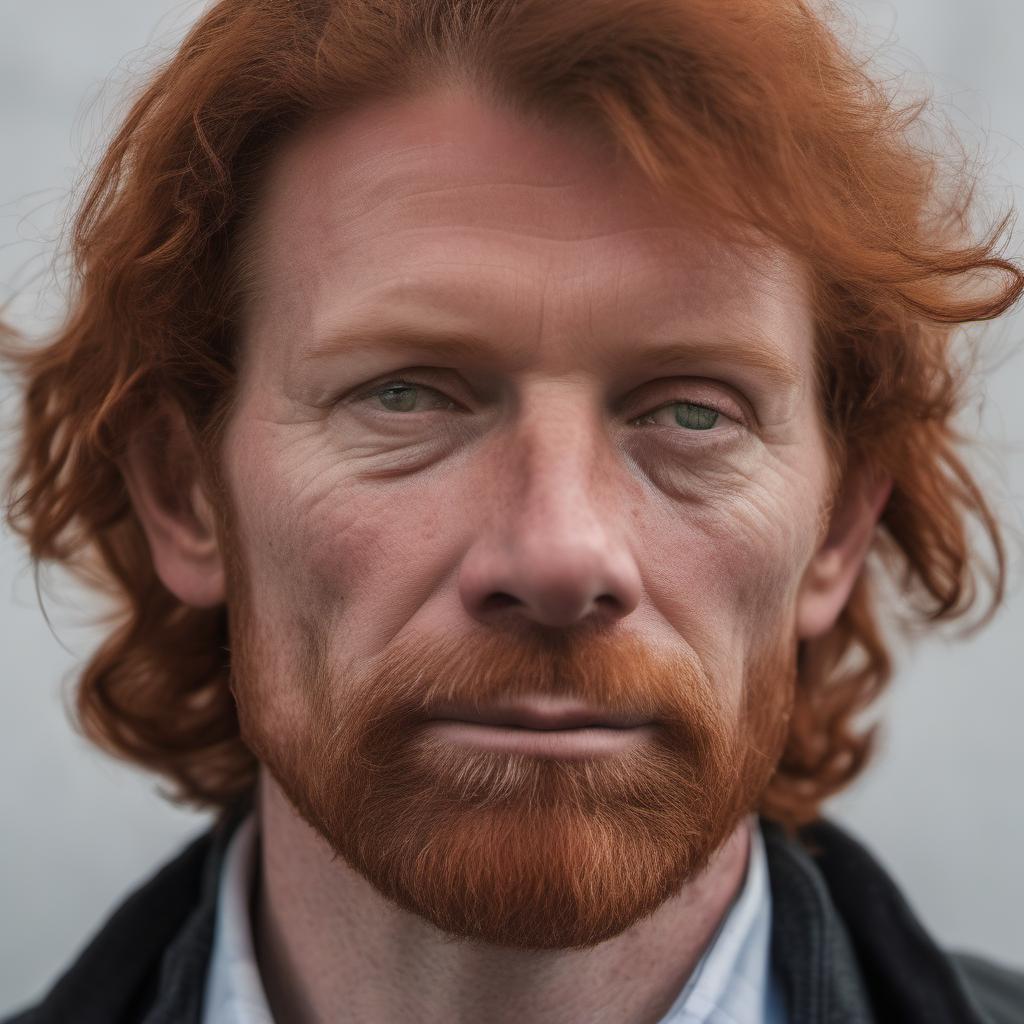}}%
        \fbox{\includegraphics[width=\mainsdxlimgwidth]{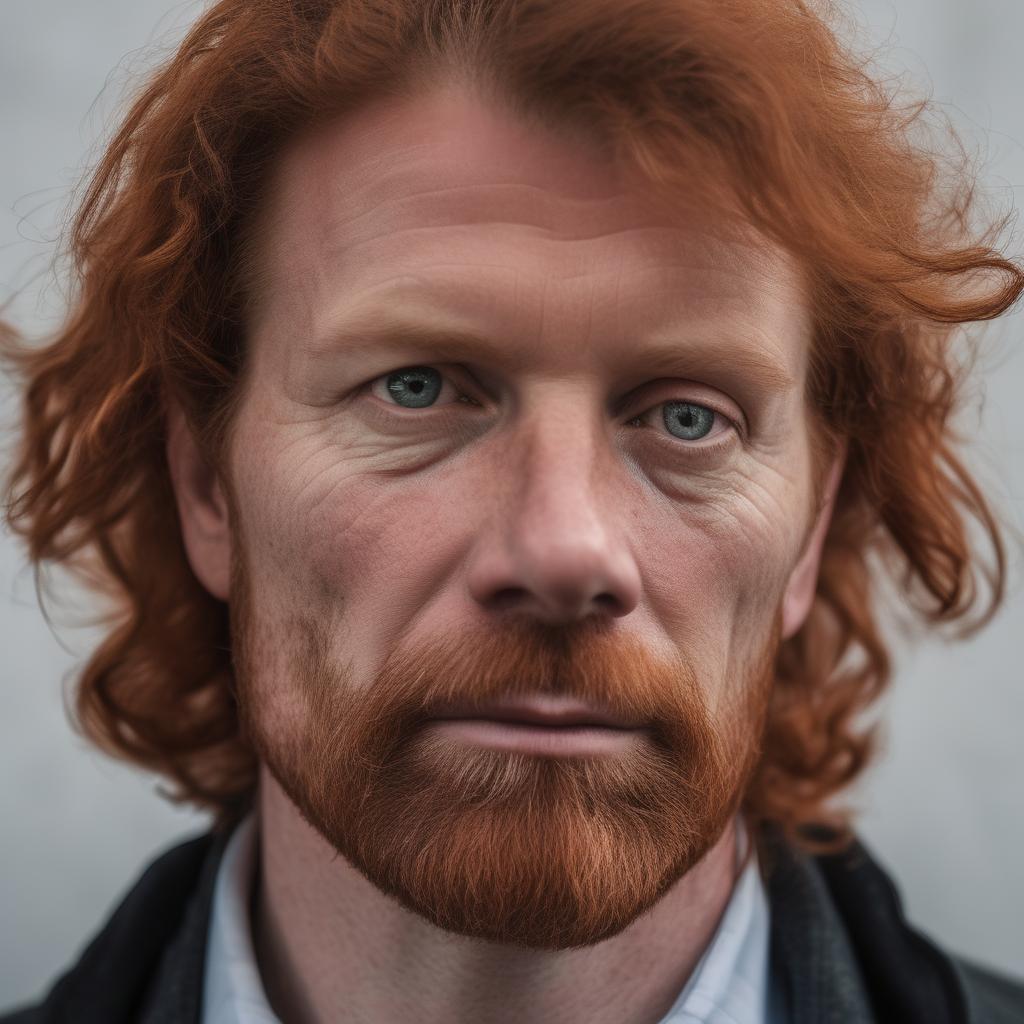}}%
        \fbox{\includegraphics[width=\mainsdxlimgwidth]{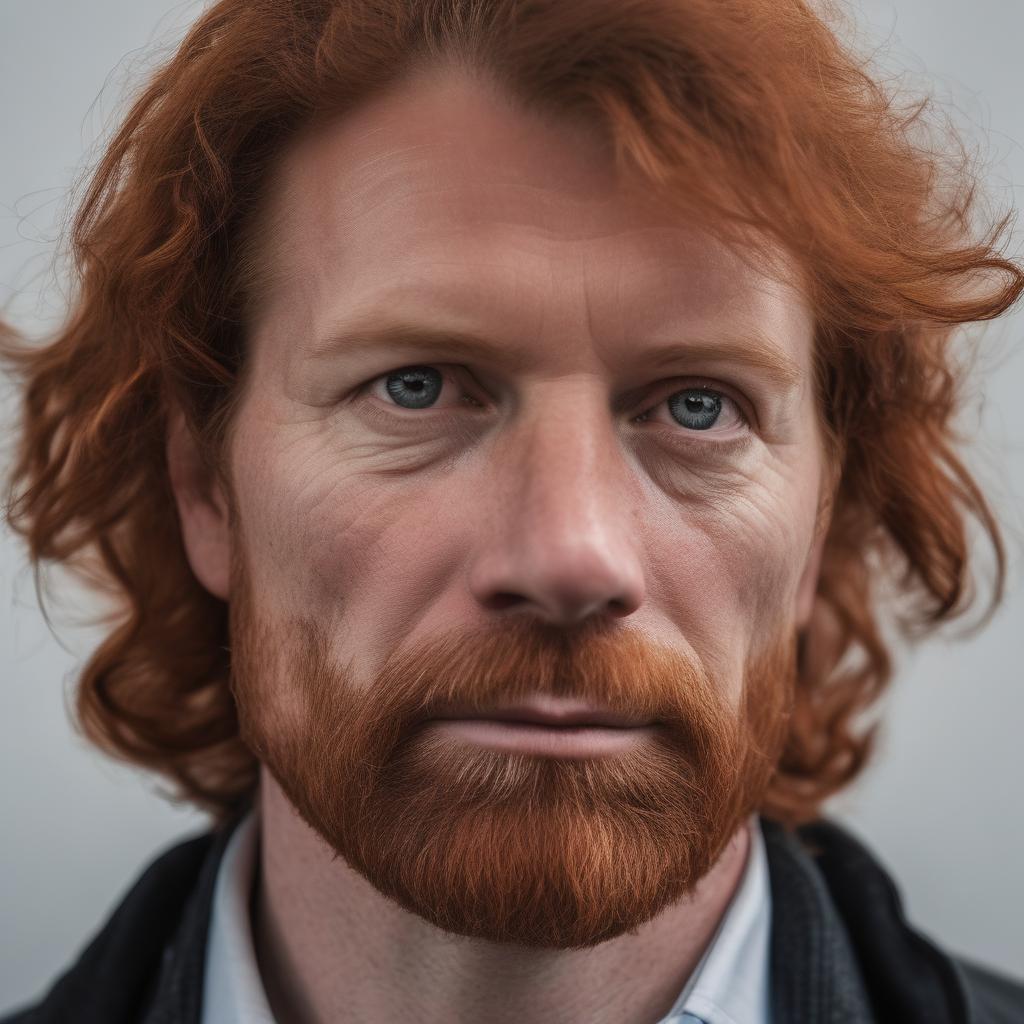}}\\[0.5ex]
        \hfill
        \vspace{-18pt}
        \caption*{
            \begin{minipage}{\mainsdxlcapwidth}
            \centering
                \tiny{Prompt: \textit{photograph of a private Detective, age 41, irish, big, reddish hair, ruddy skin complexion, taken with a fujifilm xt5, ultra realistic, highly detailed, 8k}}
            \end{minipage}
        }
    \end{minipage}
    \hfill
    \begin{minipage}[t]{0.325\textwidth}
        \centering
        \fbox{\includegraphics[width=\mainsdxlimgwidth]{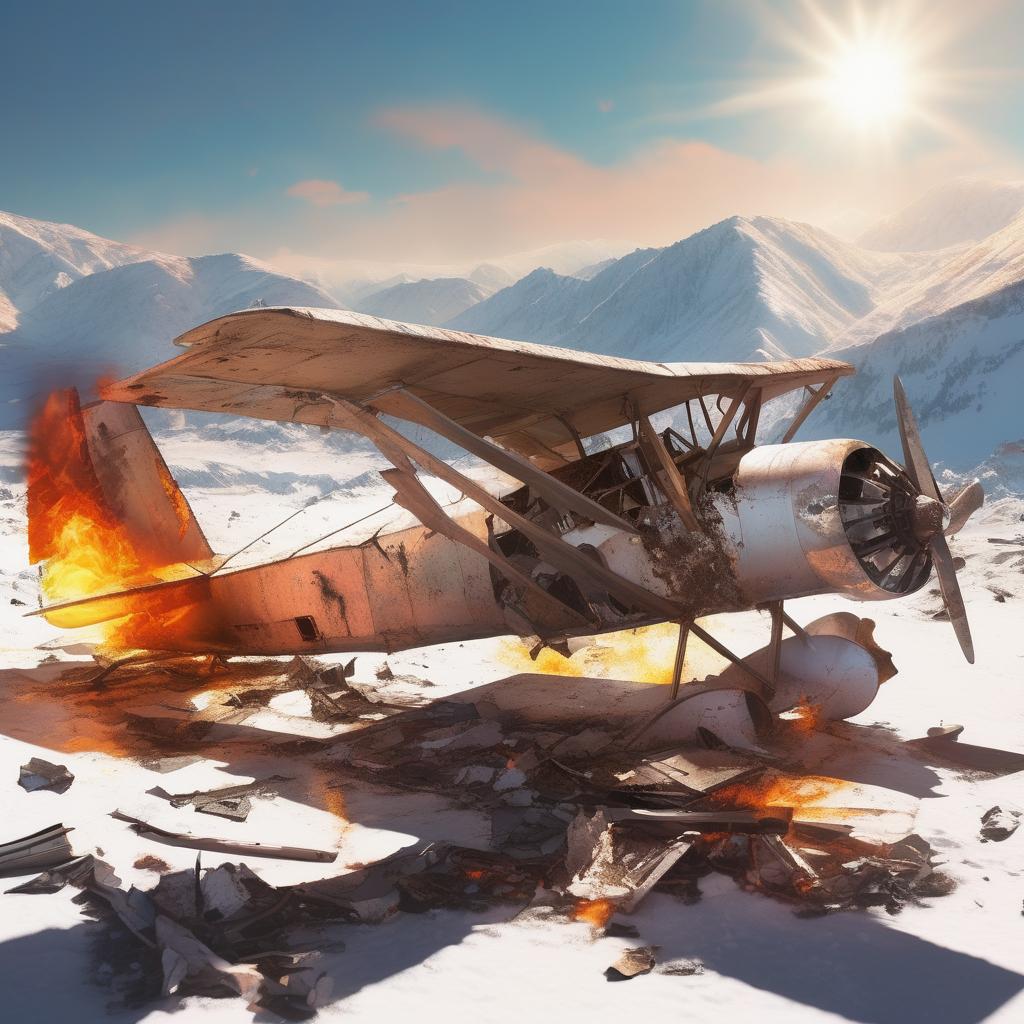}}
        \hfill
        \fbox{\includegraphics[width=\mainsdxlimgwidth]{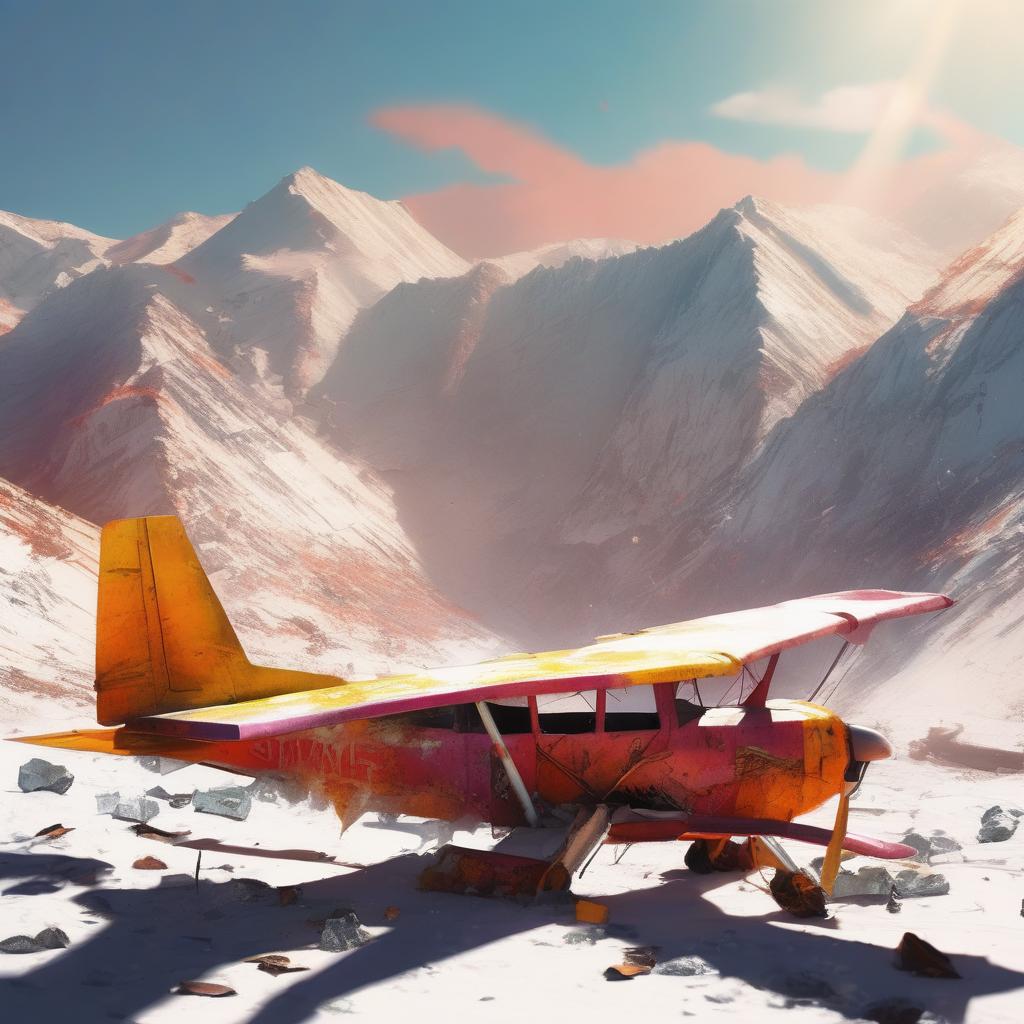}}%
        \fbox{\includegraphics[width=\mainsdxlimgwidth]{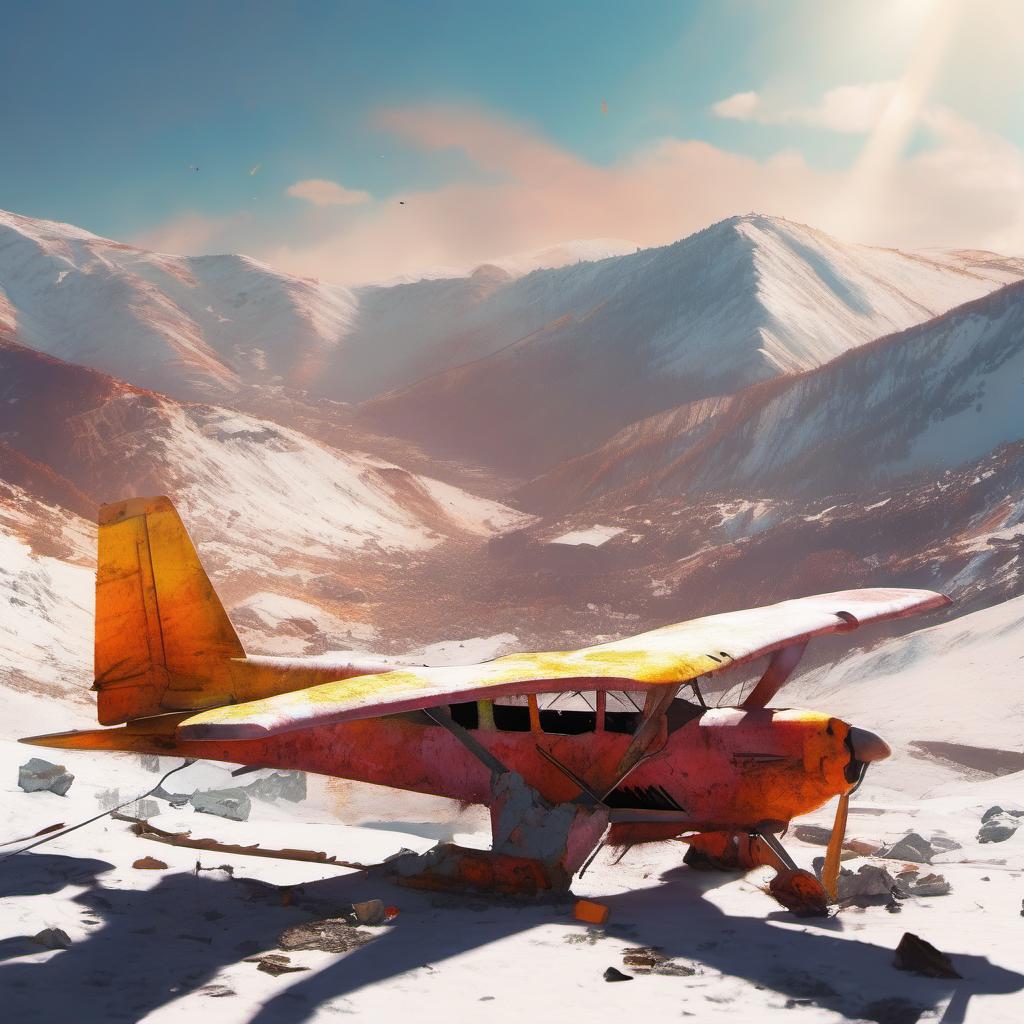}}%
        \fbox{\includegraphics[width=\mainsdxlimgwidth]{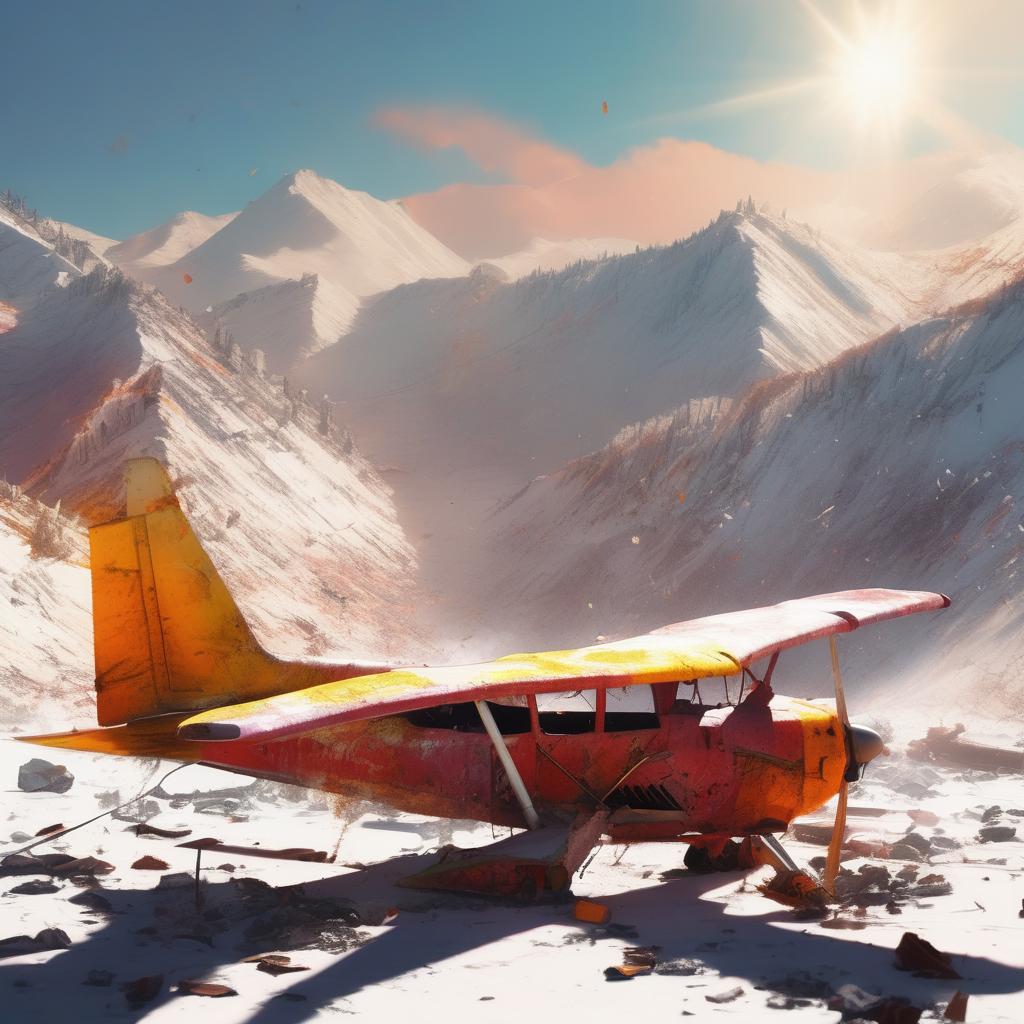}}%
        \fbox{\includegraphics[width=\mainsdxlimgwidth]{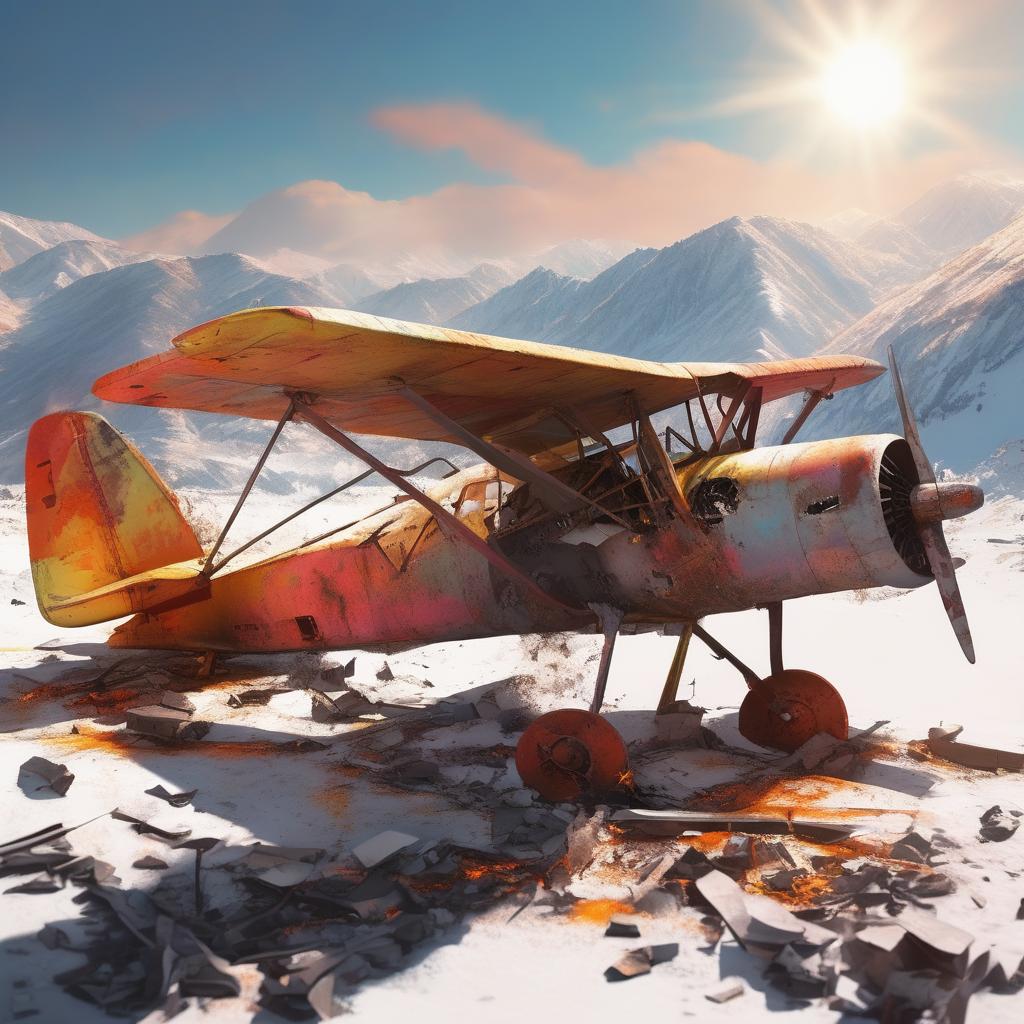}}\\[0.5ex]
        \hfill
        \vspace{-18pt}
        \caption*{
            \begin{minipage}{\mainsdxlcapwidth}
            \centering
                \tiny{Prompt: \textit{broken plane parked on snowy mountains with burning debris, Bright color palette, 150mm, realistic, gritty, sun shafts, v5}}
            \end{minipage}
        }
    \end{minipage}
    
    \caption{
        More visualization of main results on the MJHQ dataset with SDXL and W8A8 DiT quantization.
    }
    \label{fig:app-mjhq-sdxl}
\end{figure}

\newcommand{\mjhqfluximgwidth}{0.16\textwidth}
\newcommand{\mjhqfluxrightimgwidth}{0.24\textwidth}
\newcommand{\mjhqfluxcapwidth}{0.95\textwidth}
\setlength{\fboxsep}{0pt}
\setlength{\fboxrule}{0.2pt}
\begin{figure}[h!]
    \centering
    \begin{minipage}[t]{0.6\textwidth}
        \centering
        \begin{minipage}[t]{\mjhqfluximgwidth}
            \centering \scriptsize{FP16}
        \end{minipage}%
        \hspace{0.5mm}
        \begin{minipage}[t]{\mjhqfluximgwidth}
            \centering \scriptsize{Q-Diffusion}
        \end{minipage}%
        \begin{minipage}[t]{\mjhqfluximgwidth}
            \centering \scriptsize{PTQ4DiT}
        \end{minipage}%
        \begin{minipage}[t]{\mjhqfluximgwidth}
            \centering \scriptsize{Smooth+}
        \end{minipage}%
        \begin{minipage}[t]{\mjhqfluximgwidth}
            \centering \scriptsize{\textbf{SegQuant-A}}
        \end{minipage}%
        \begin{minipage}[t]{\mjhqfluximgwidth}
            \centering \scriptsize{\textbf{SegQuant-G}}
        \end{minipage}%
    \end{minipage}
    \hfill
    \begin{minipage}[t]{0.395\textwidth}
        \centering
        \hfill
        \begin{minipage}[t]{\mjhqfluxrightimgwidth}
            \centering \scriptsize{Q-Diffusion}
        \end{minipage}%
        \begin{minipage}[t]{\mjhqfluxrightimgwidth}
            \centering \scriptsize{SVDQuant}
        \end{minipage}%
        \begin{minipage}[t]{\mjhqfluxrightimgwidth}
            \centering \scriptsize{\textbf{SegQuant-G}}
        \end{minipage}%
        \hspace{1.5mm}
        \begin{minipage}[t]{\mjhqfluxrightimgwidth}
            \centering \scriptsize{FP16}
        \end{minipage}%
    \end{minipage}

    \vspace{0.6ex}

    \begin{minipage}[t]{0.595\textwidth}
        \centering
        \fbox{\includegraphics[width=\mjhqfluximgwidth]{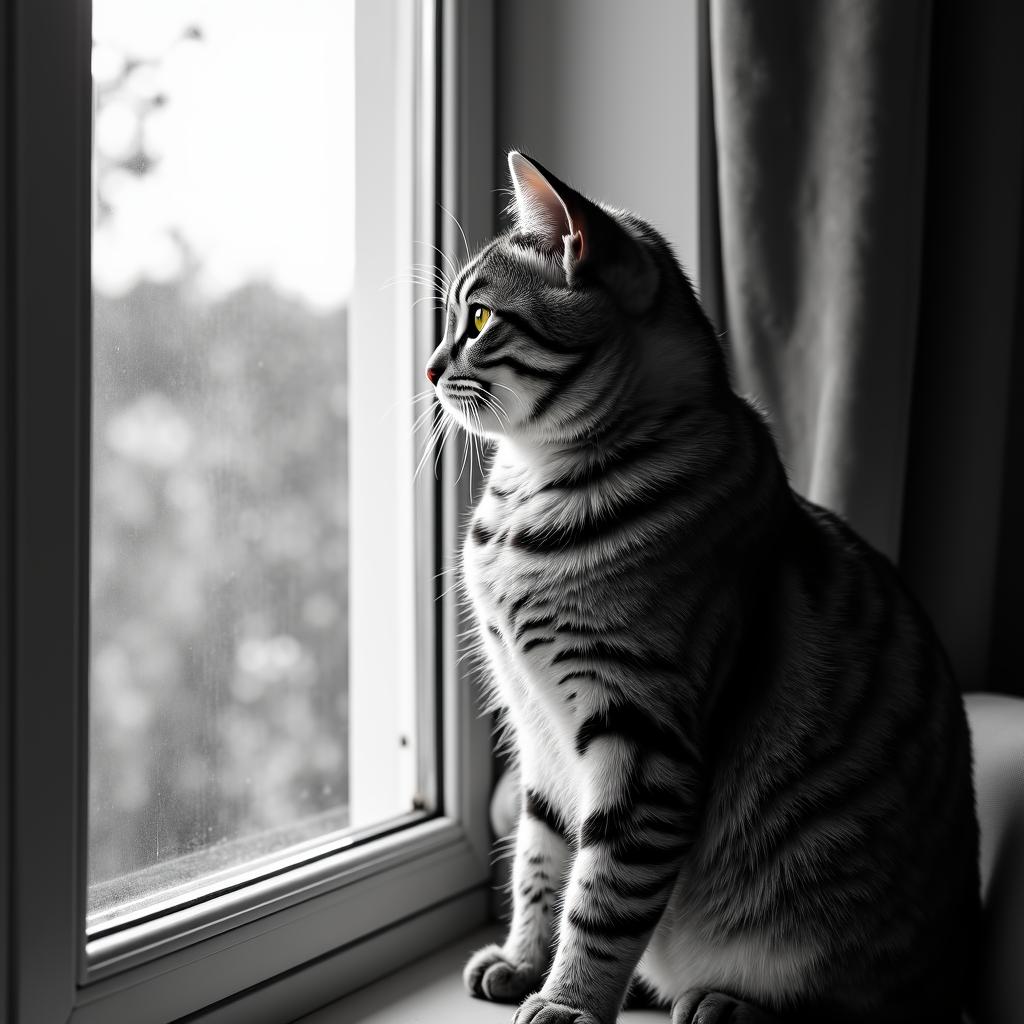}}%
        \hspace{0.5mm}
        \fbox{\includegraphics[width=\mjhqfluximgwidth]{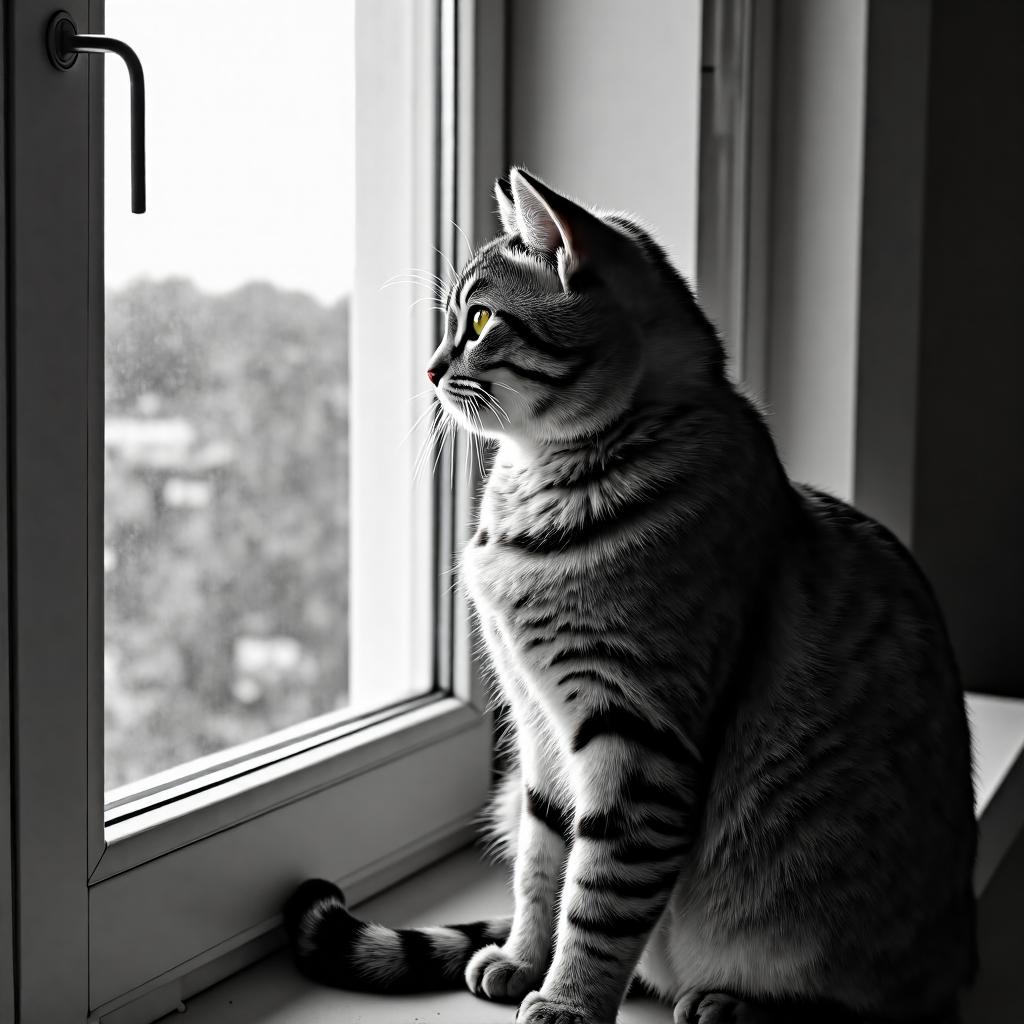}}%
        \fbox{\includegraphics[width=\mjhqfluximgwidth]{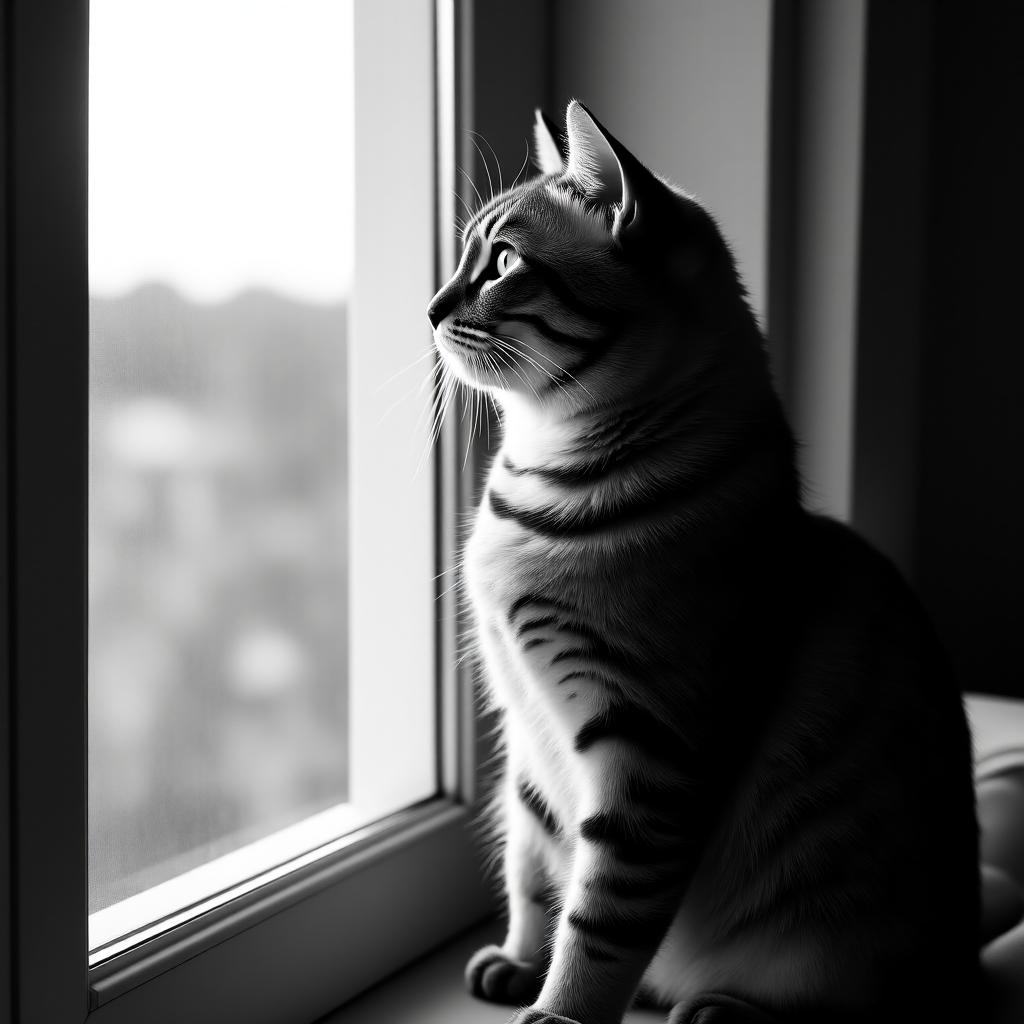}}%
        \fbox{\includegraphics[width=\mjhqfluximgwidth]{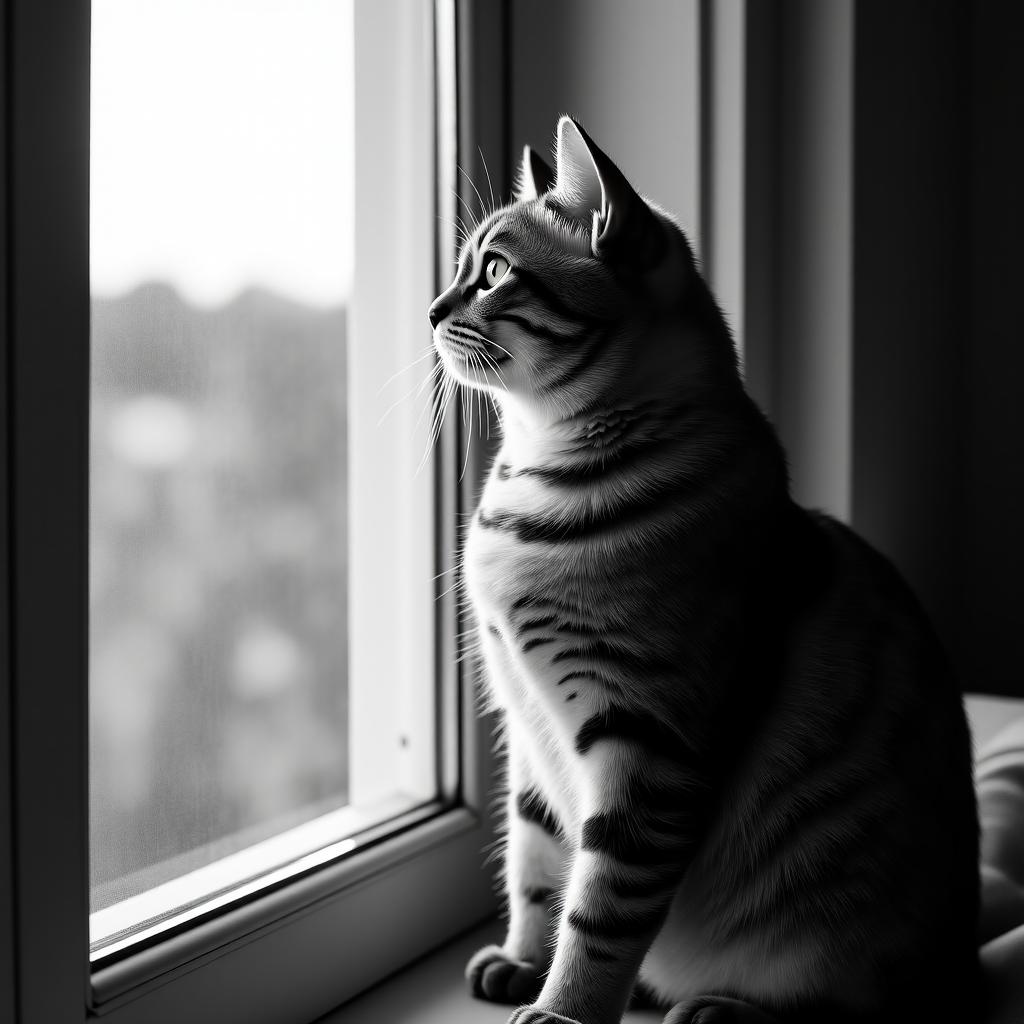}}%
        \fbox{\includegraphics[width=\mjhqfluximgwidth]{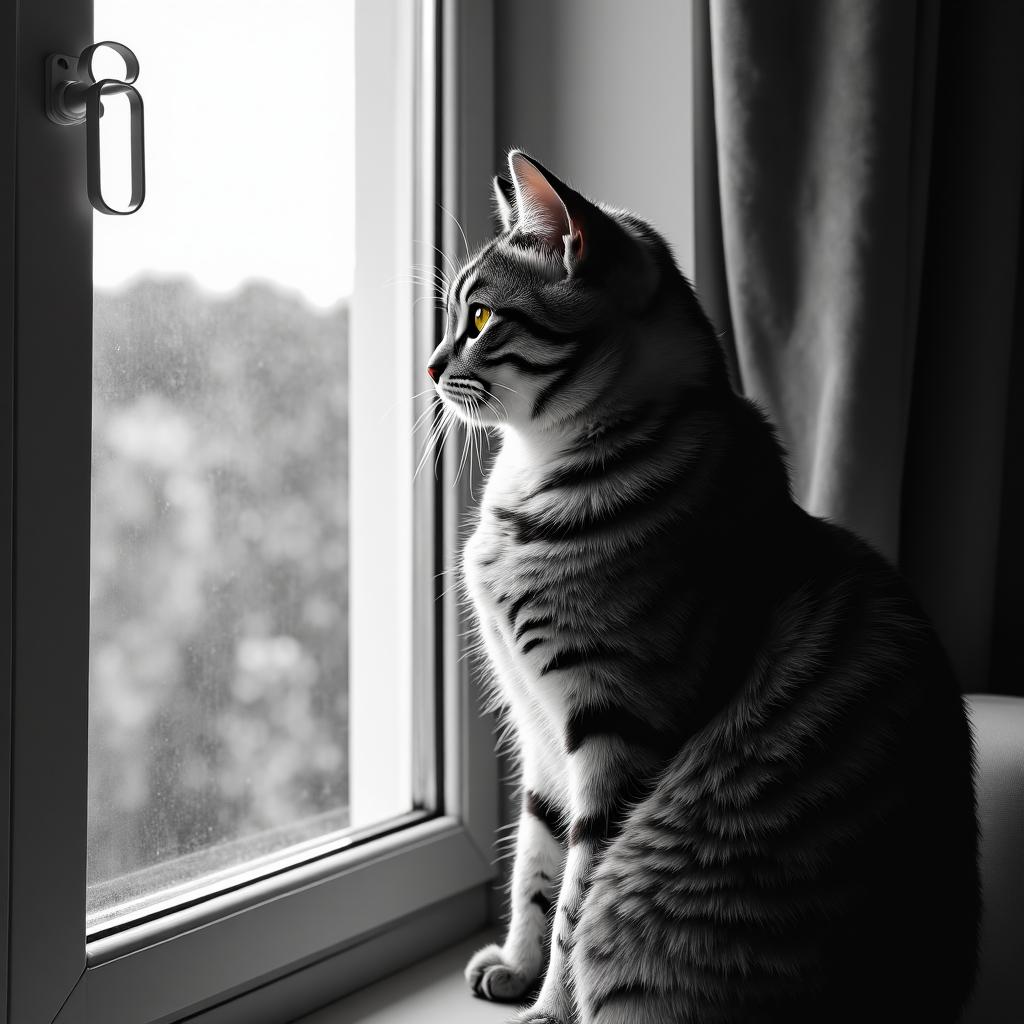}}%
        \fbox{\includegraphics[width=\mjhqfluximgwidth]{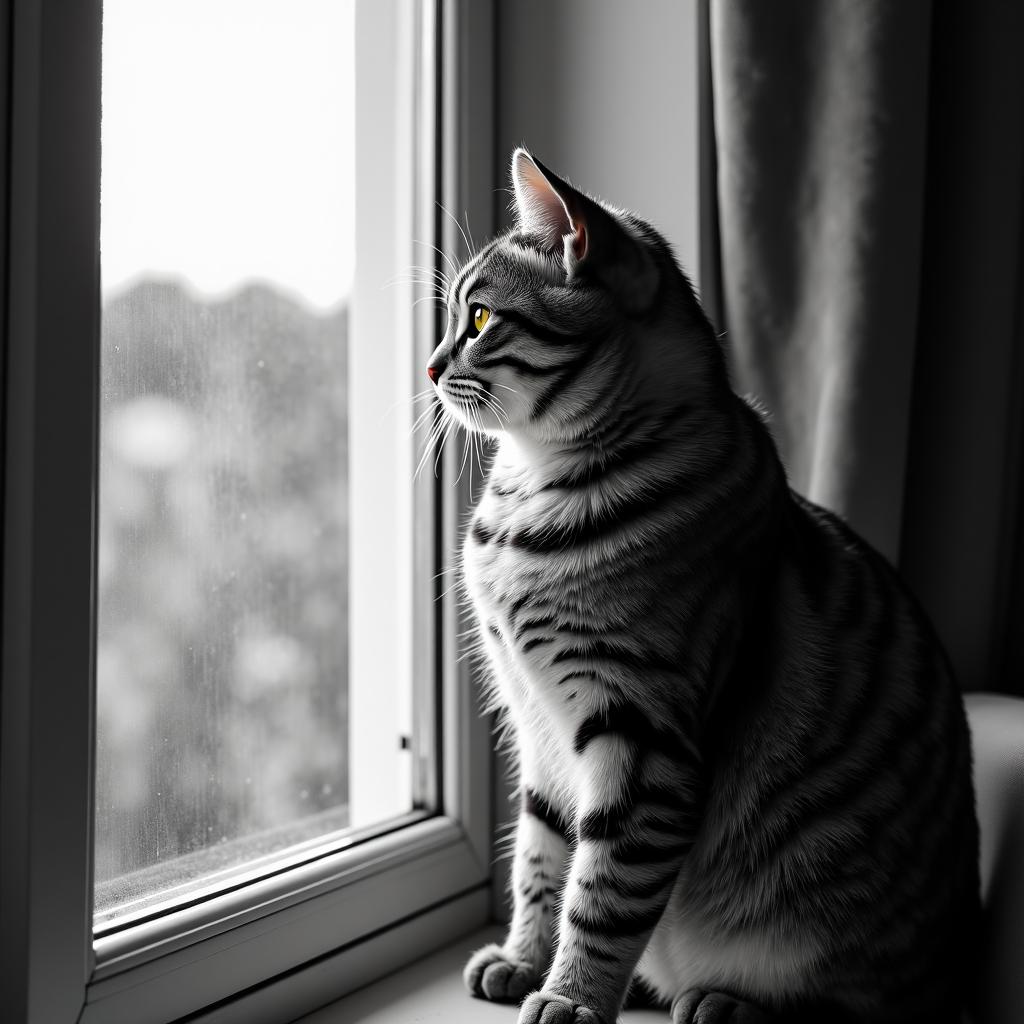}}\\[0.5ex]
        \hfill
    \end{minipage}
    \hfill
    \begin{minipage}[t]{0.395\textwidth}
        \centering
        \hfill
        \fbox{\includegraphics[width=\mjhqfluxrightimgwidth]{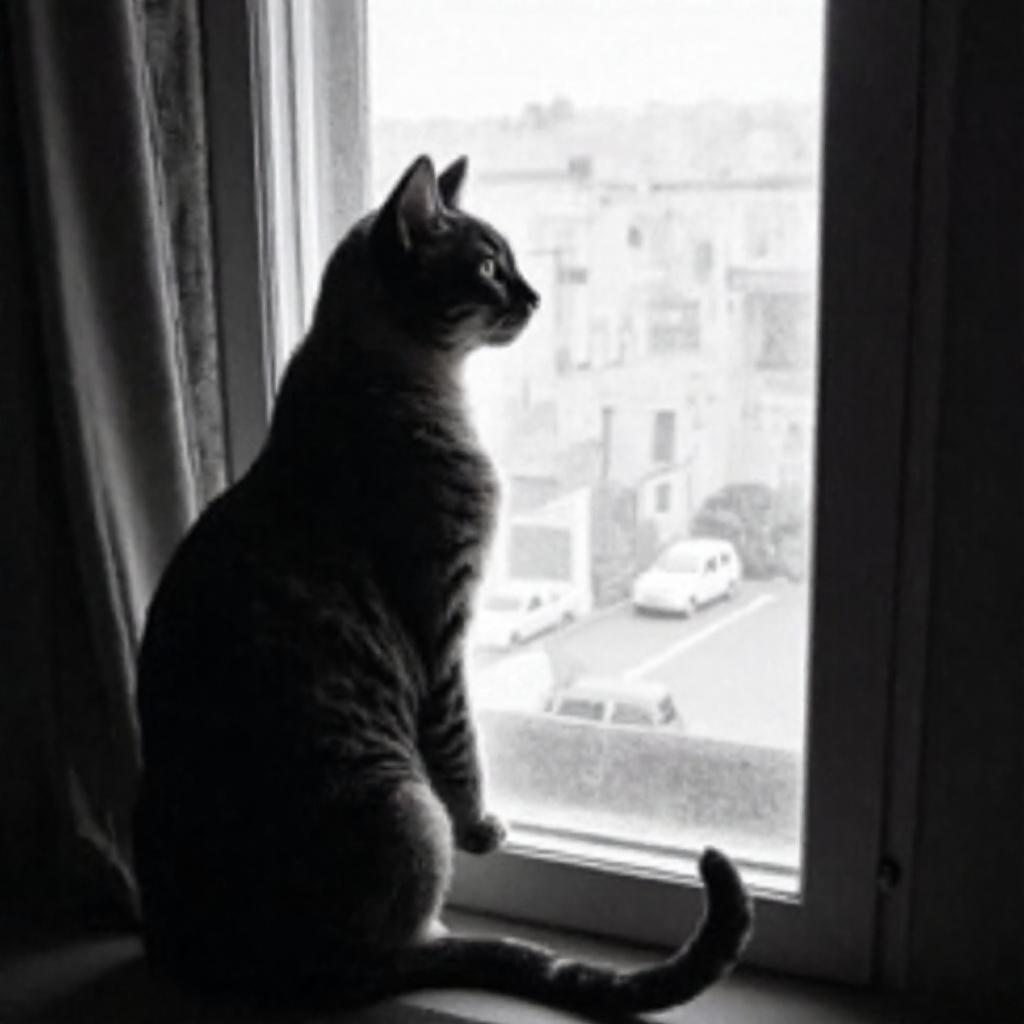}}%
        \fbox{\includegraphics[width=\mjhqfluxrightimgwidth]{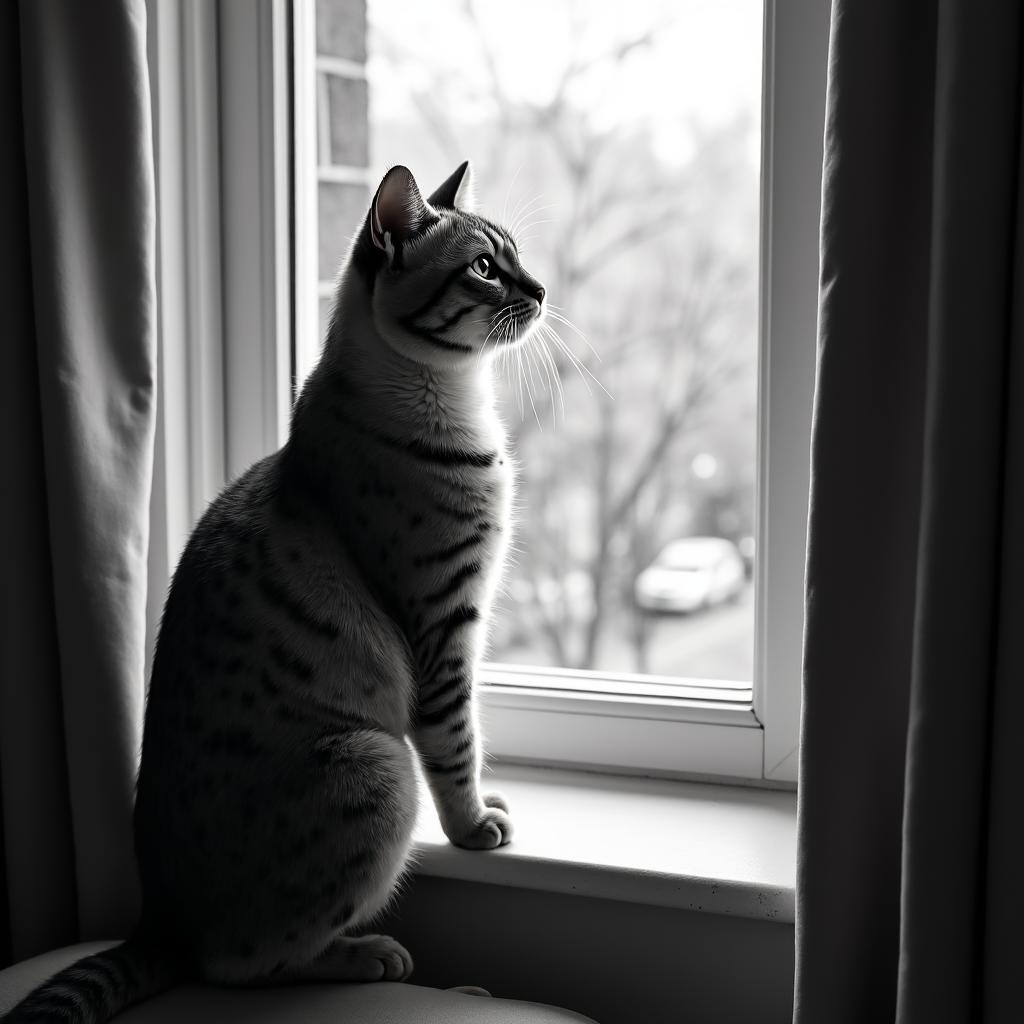}}%
        \fbox{\includegraphics[width=\mjhqfluxrightimgwidth]{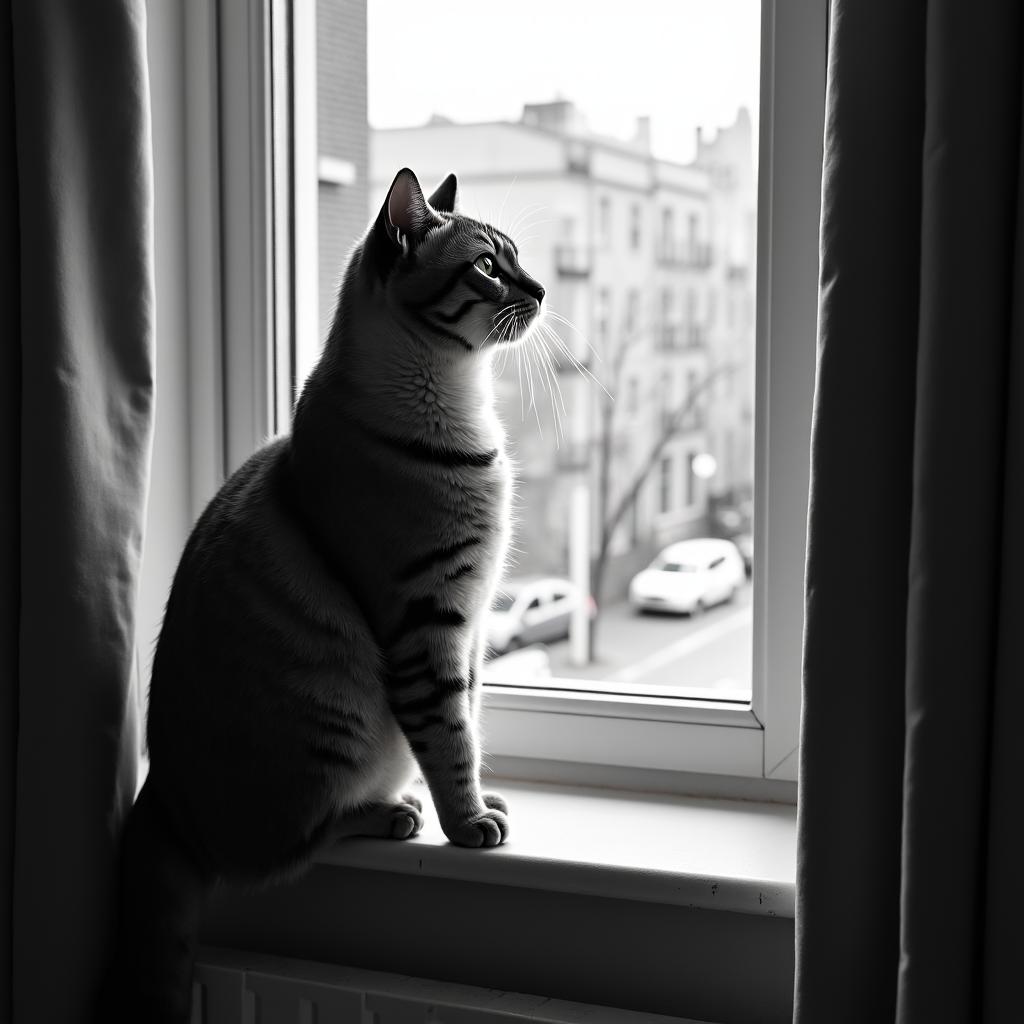}}%
        \hfill
        \fbox{\includegraphics[width=\mjhqfluxrightimgwidth]{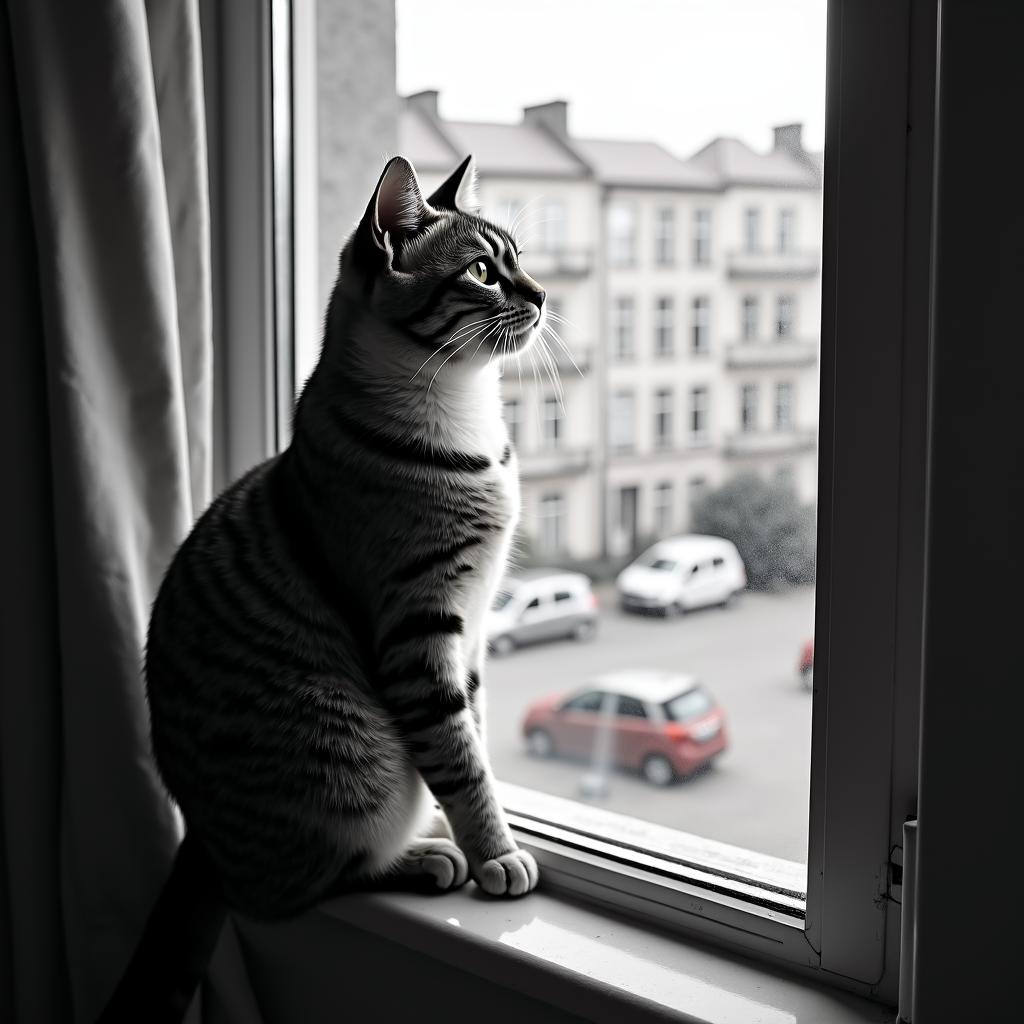}}%
        \\[0.5ex]
    \end{minipage}
    \vspace{-15pt}
    \caption*{
        \begin{minipage}{\mjhqfluxcapwidth}
        \centering
            \footnotesize{Prompt: \textit{a cat, its face looks like Winston Churchill, standing up like a human, in a black and white photo, looking out of a window.}}
        \end{minipage}
    }

    \vspace{0.2cm}
    \begin{minipage}[t]{0.595\textwidth}
        \centering
        \fbox{\includegraphics[width=\mjhqfluximgwidth]{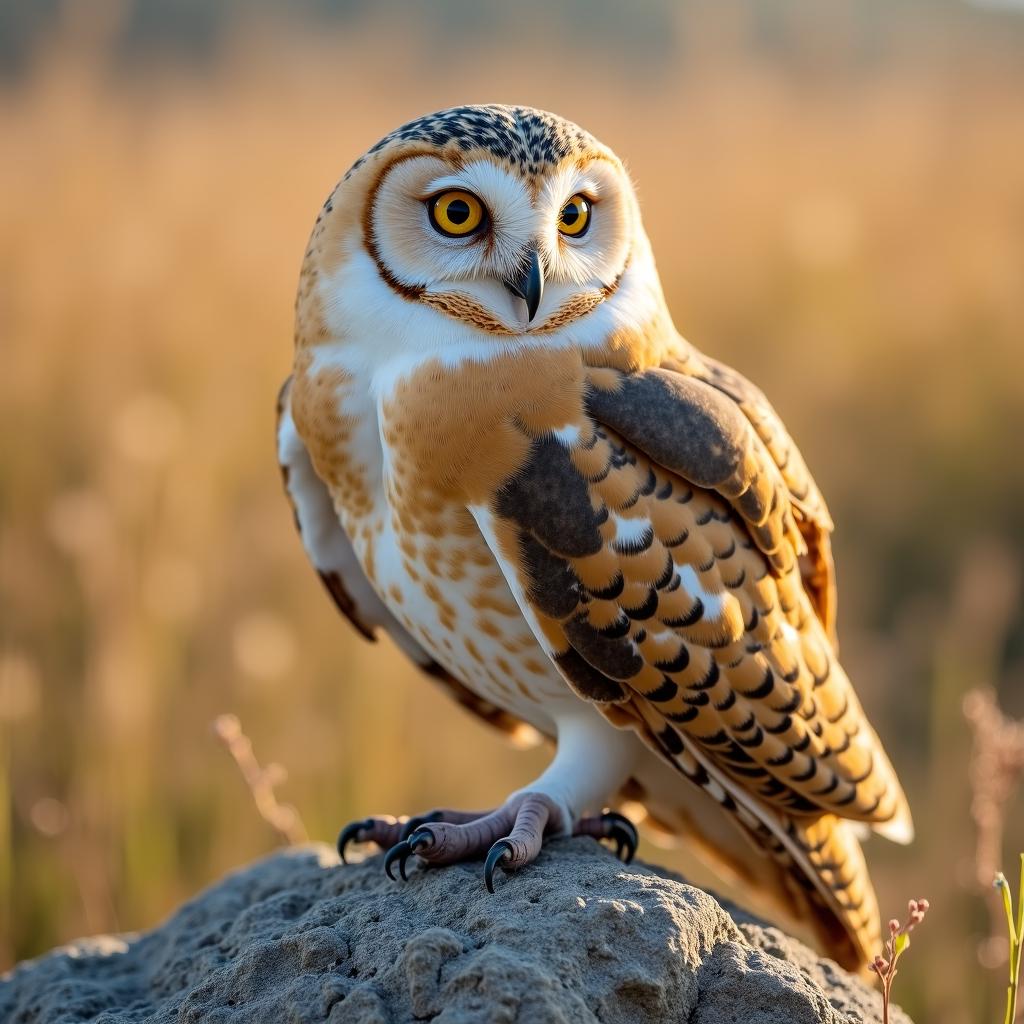}}
        \hspace{0.5mm}
        \fbox{\includegraphics[width=\mjhqfluximgwidth]{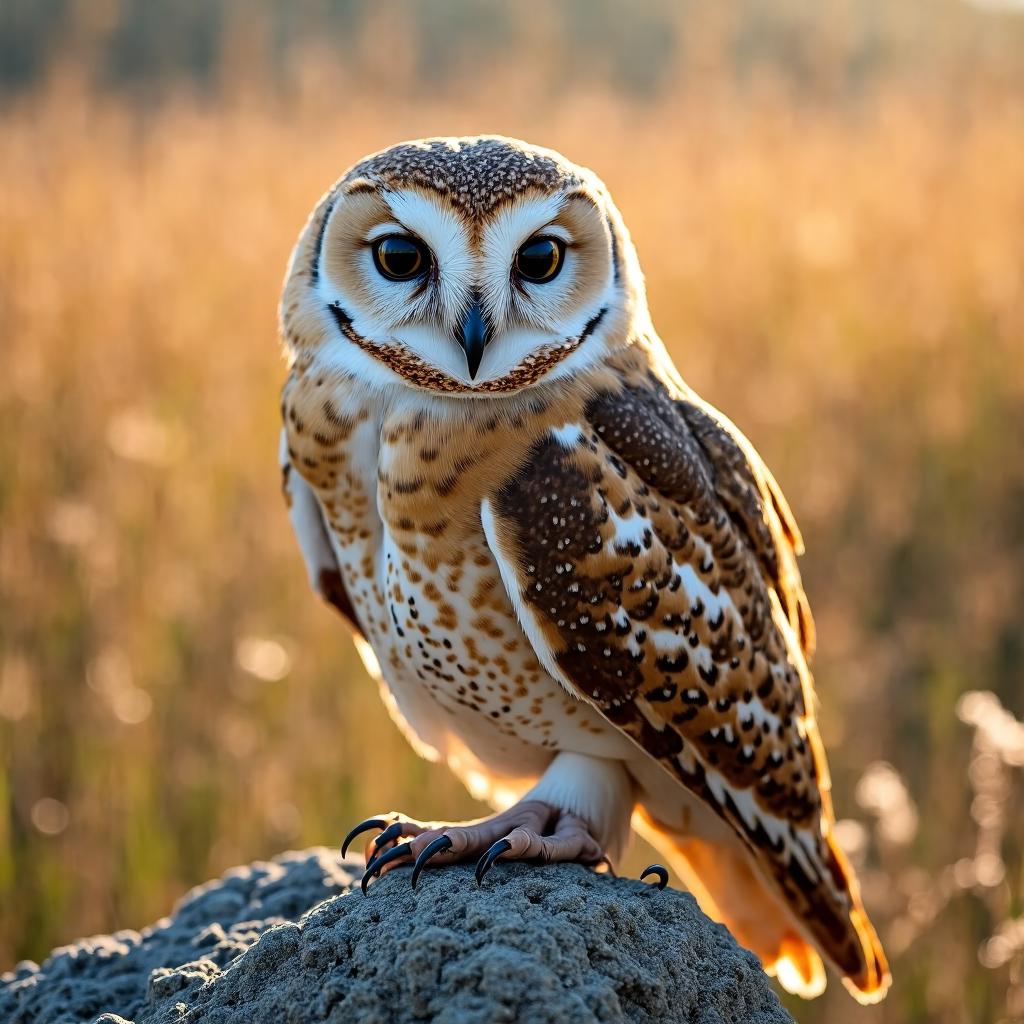}}%
        \fbox{\includegraphics[width=\mjhqfluximgwidth]{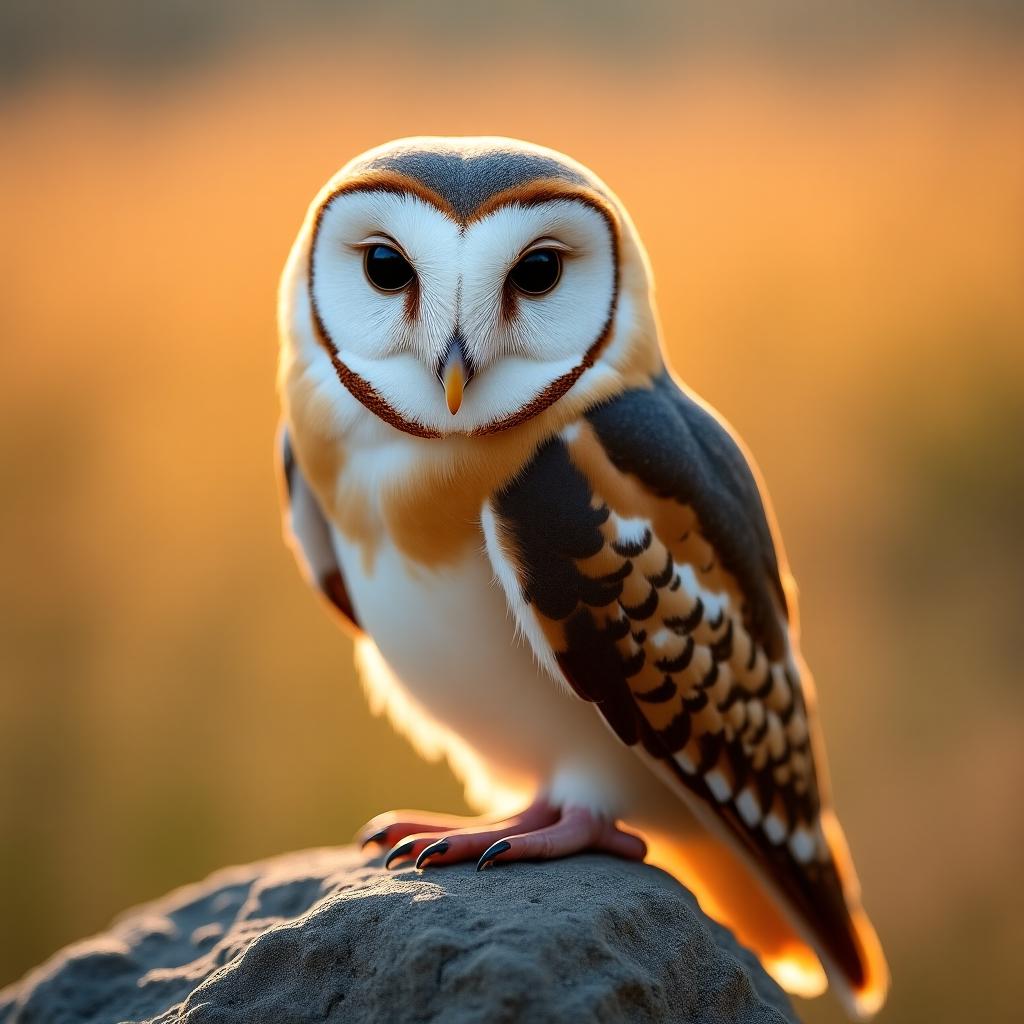}}%
        \fbox{\includegraphics[width=\mjhqfluximgwidth]{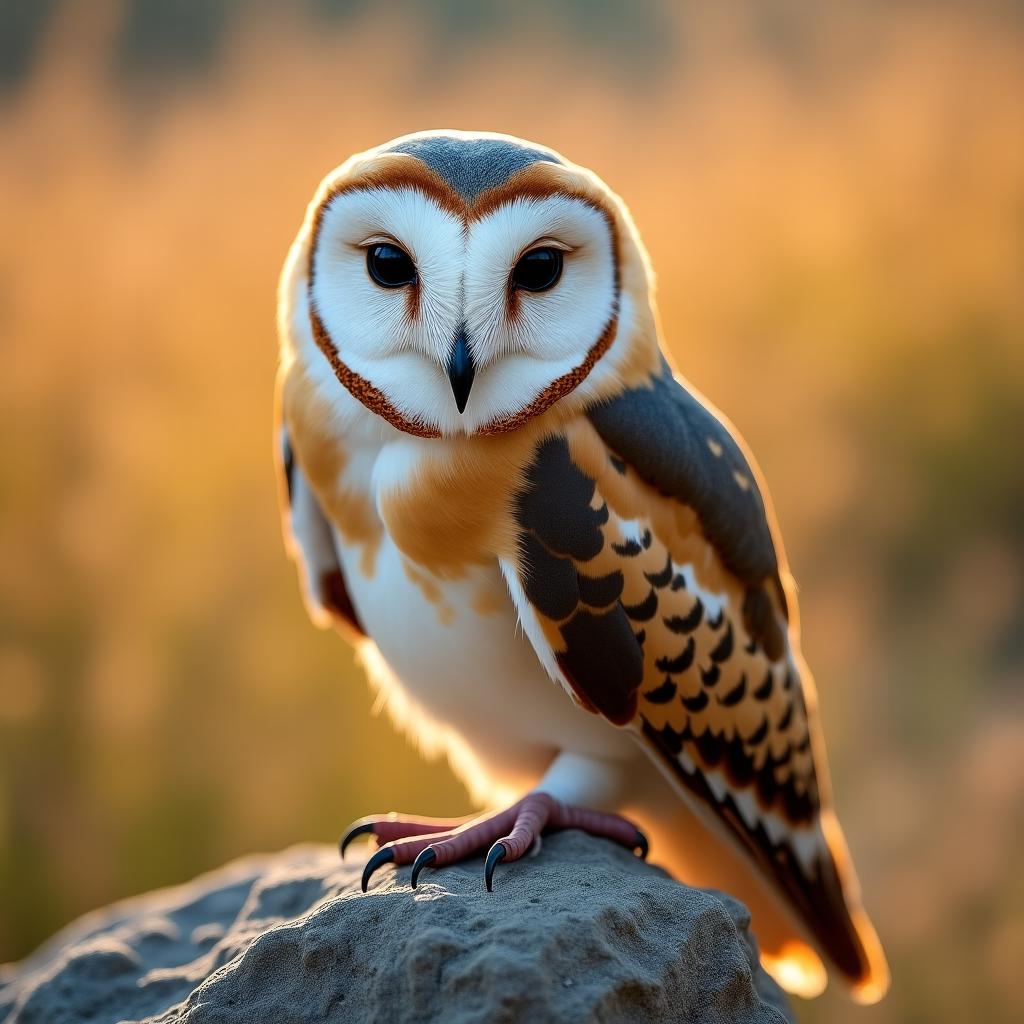}}%
        \fbox{\includegraphics[width=\mjhqfluximgwidth]{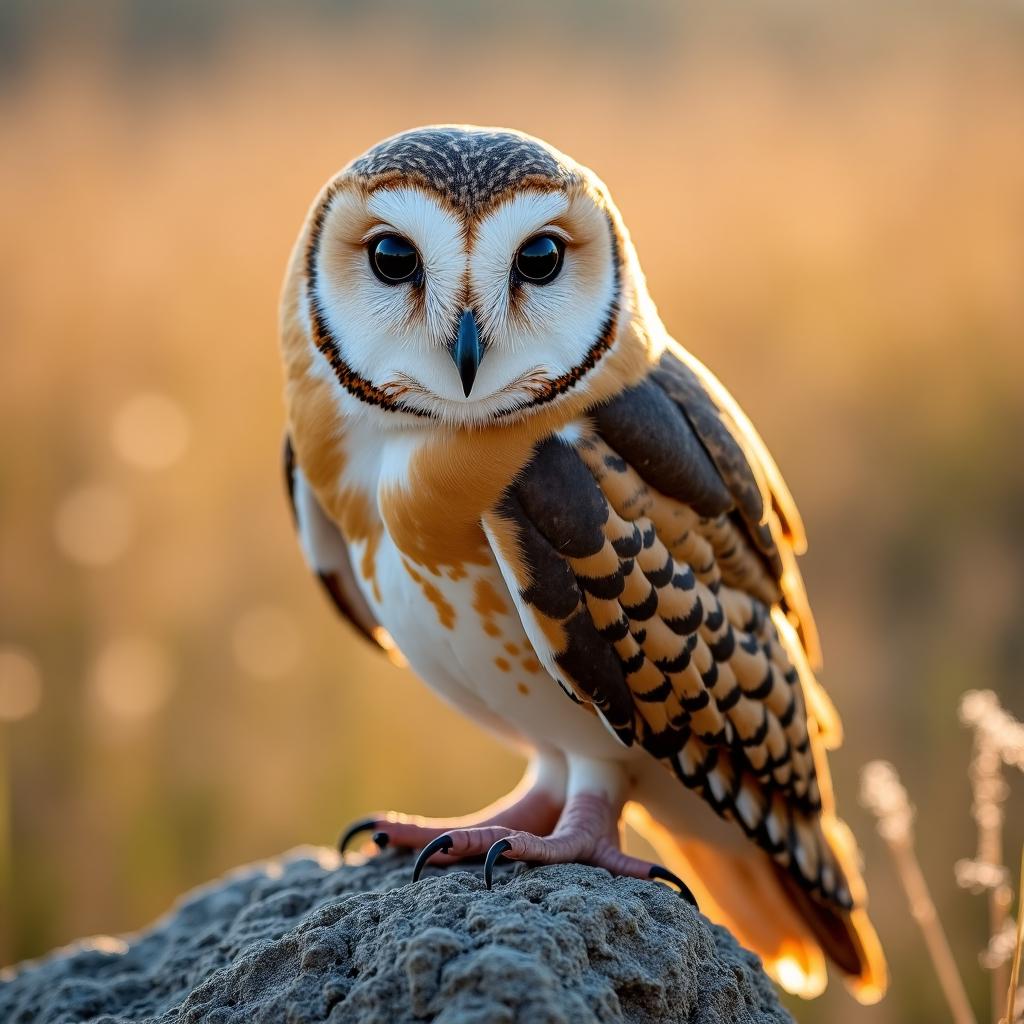}}%
        \fbox{\includegraphics[width=\mjhqfluximgwidth]{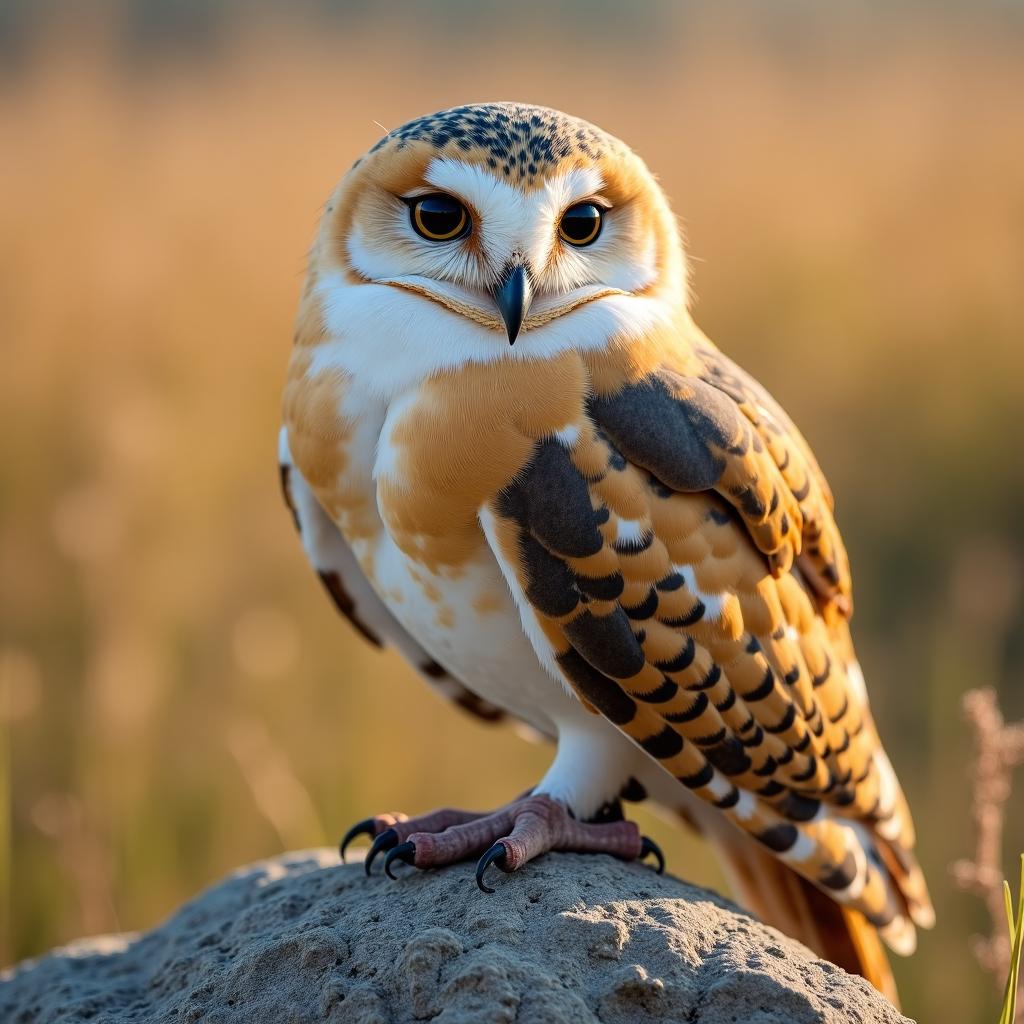}}\\[0.5ex]
        \hfill
    \end{minipage}
    \hfill
    \begin{minipage}[t]{0.395\textwidth}
        \centering
        \hfill
        \fbox{\includegraphics[width=\mjhqfluxrightimgwidth]{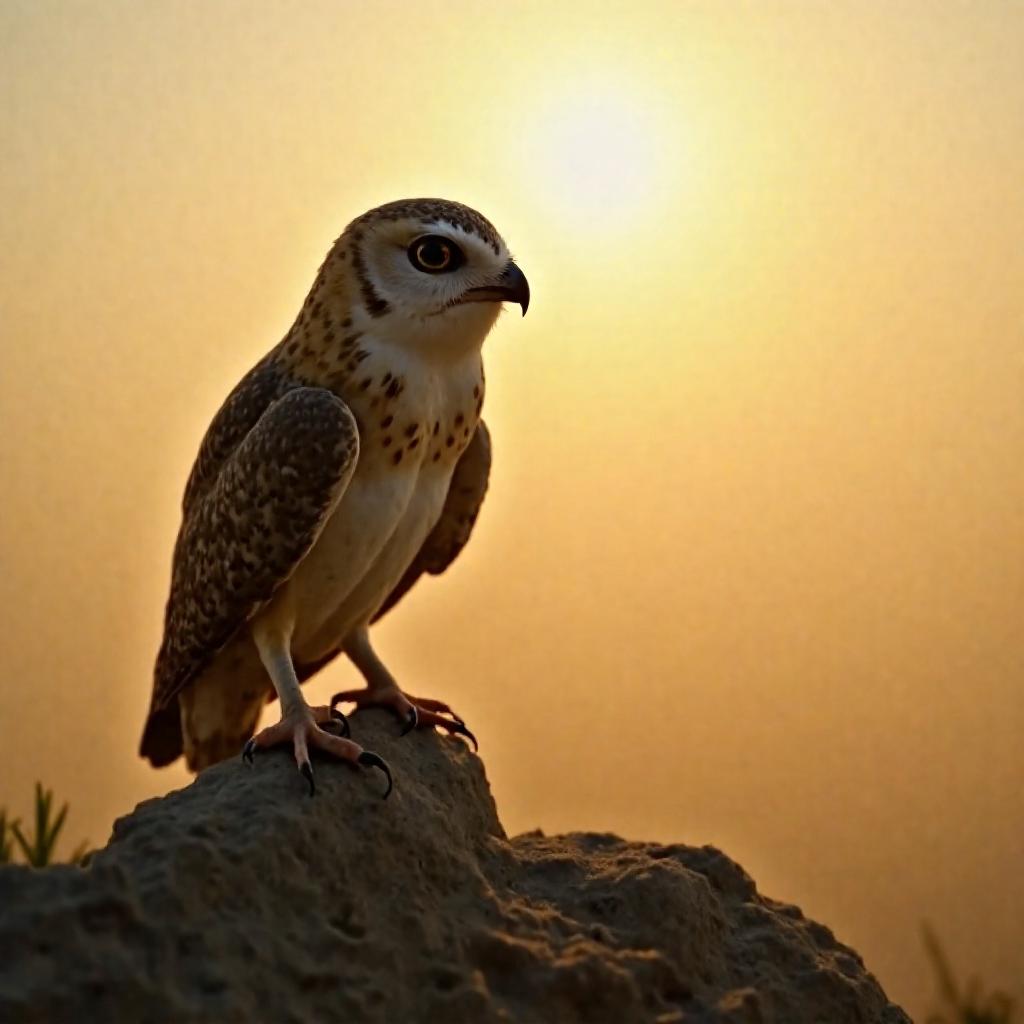}}%
        \fbox{\includegraphics[width=\mjhqfluxrightimgwidth]{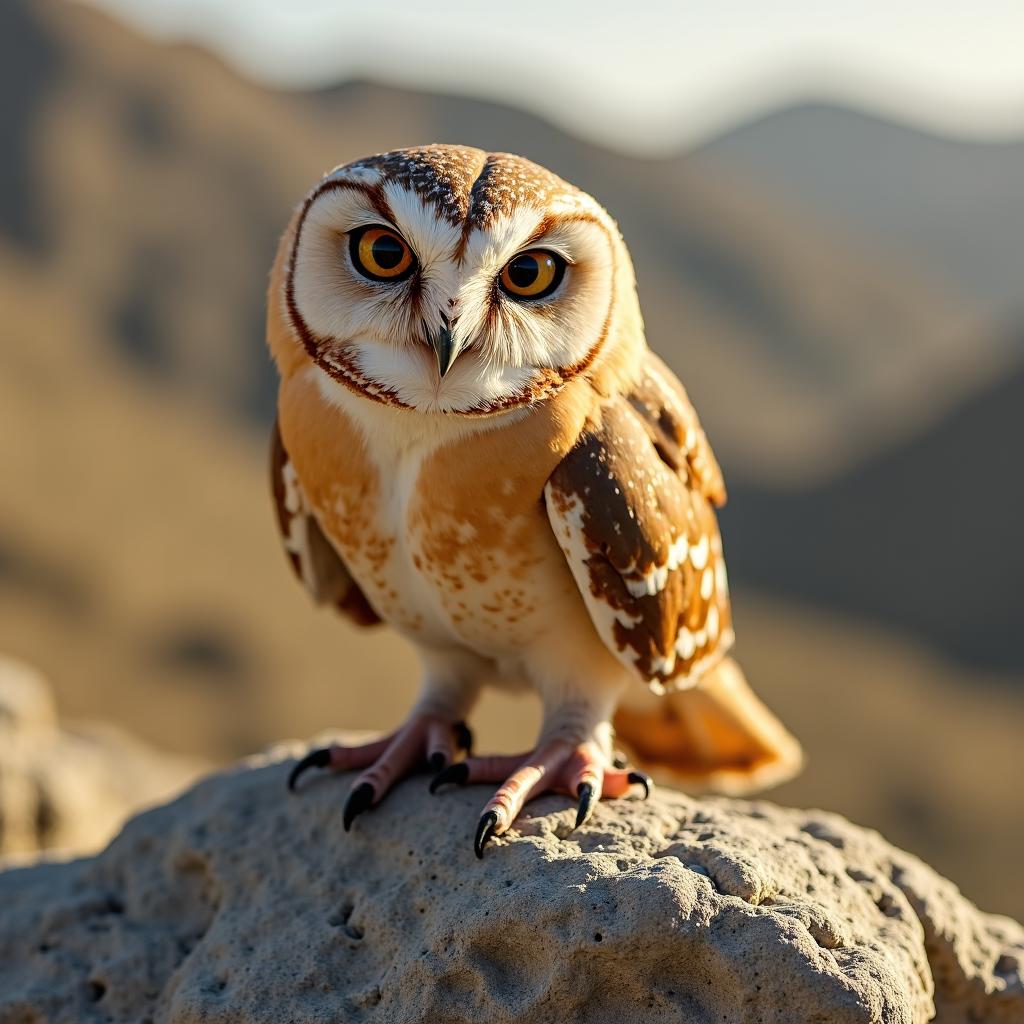}}%
        \fbox{\includegraphics[width=\mjhqfluxrightimgwidth]{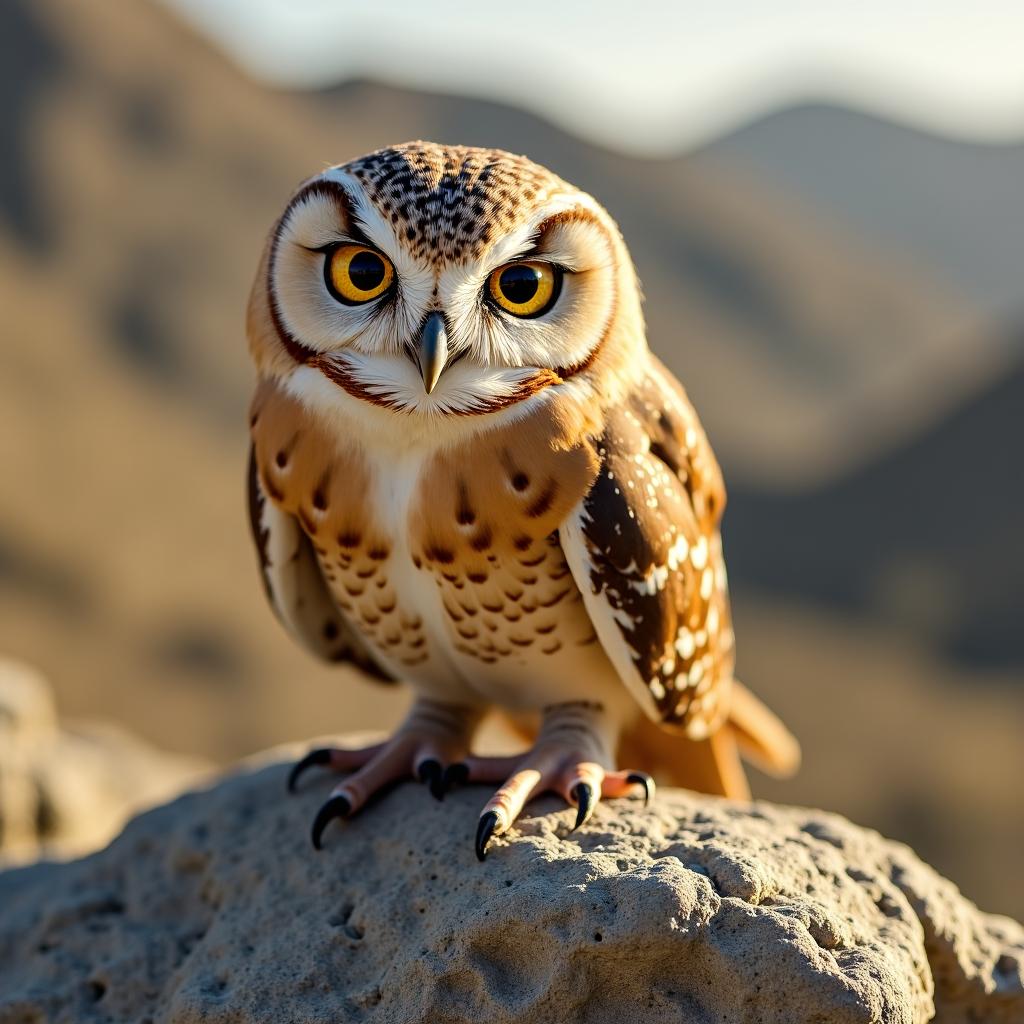}}%
        \hfill
        \fbox{\includegraphics[width=\mjhqfluxrightimgwidth]{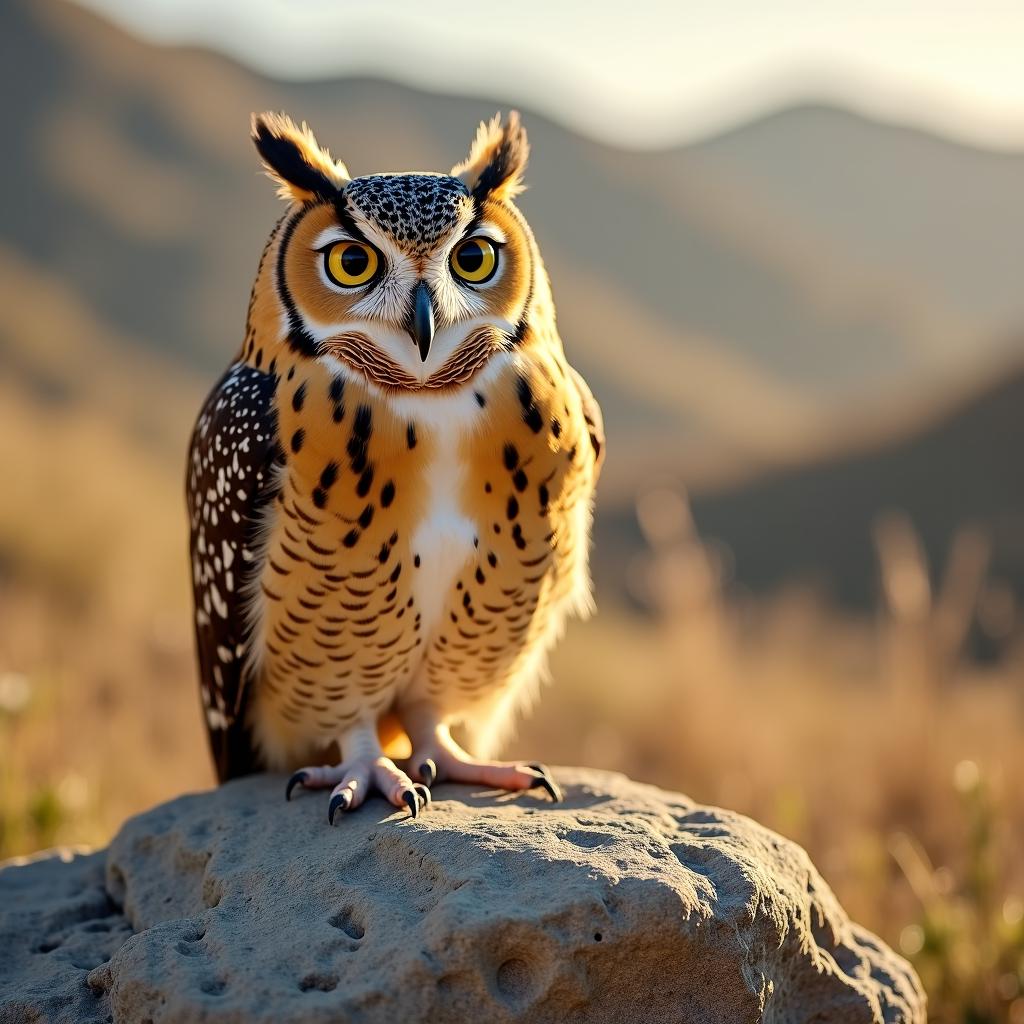}}\\[0.5ex]
    \end{minipage}
    \vspace{-15pt}
    \caption*{
        \begin{minipage}{\mjhqfluxcapwidth}
        \centering
            \footnotesize{Prompt: \textit{an owl sits on a rock and looks outward, in the style of nikon d850, light beige and amber, exaggerated poses, explosive wildlife, dansaekhwa, multiple points of view.}}
        \end{minipage}
    }

    \vspace{0.2cm}
    \begin{minipage}[t]{0.595\textwidth}
        \centering
        \fbox{\includegraphics[width=\mjhqfluximgwidth]{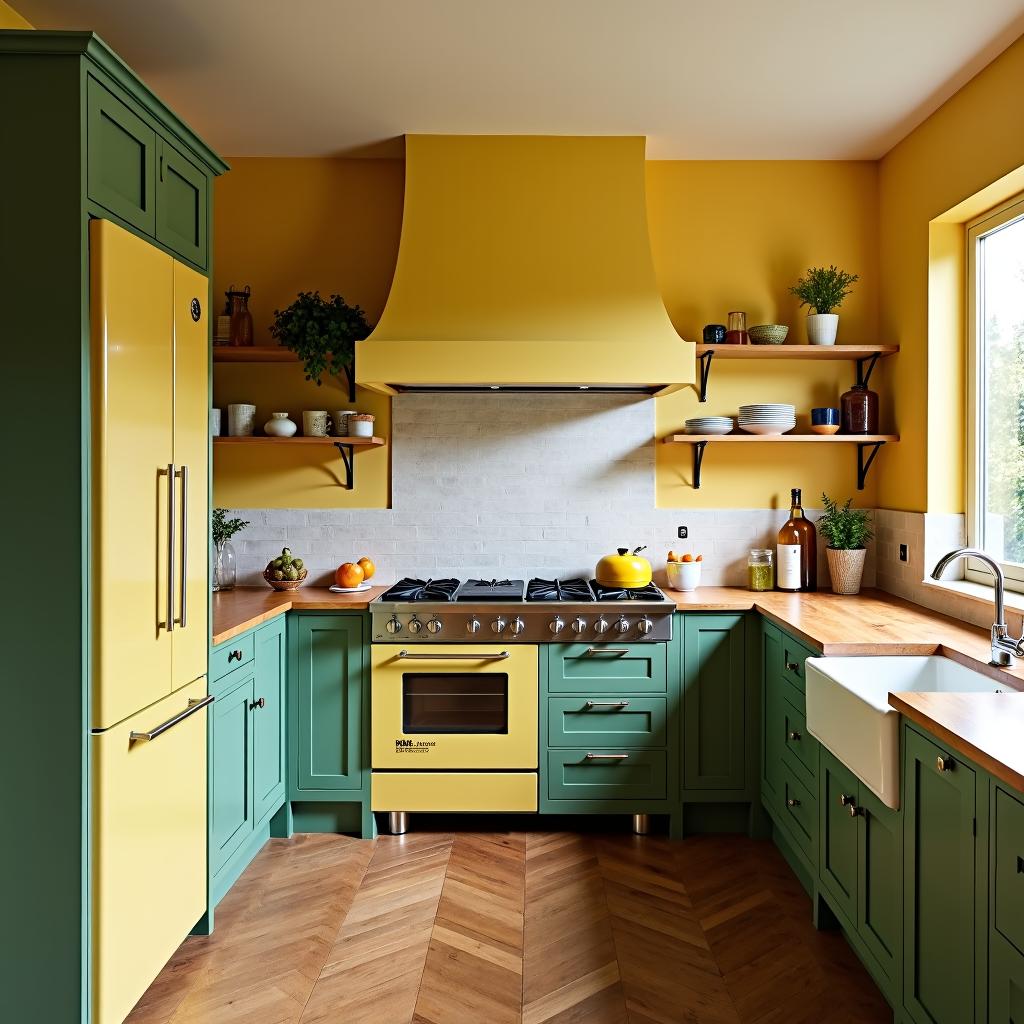}}
        \hspace{0.5mm}
        \fbox{\includegraphics[width=\mjhqfluximgwidth]{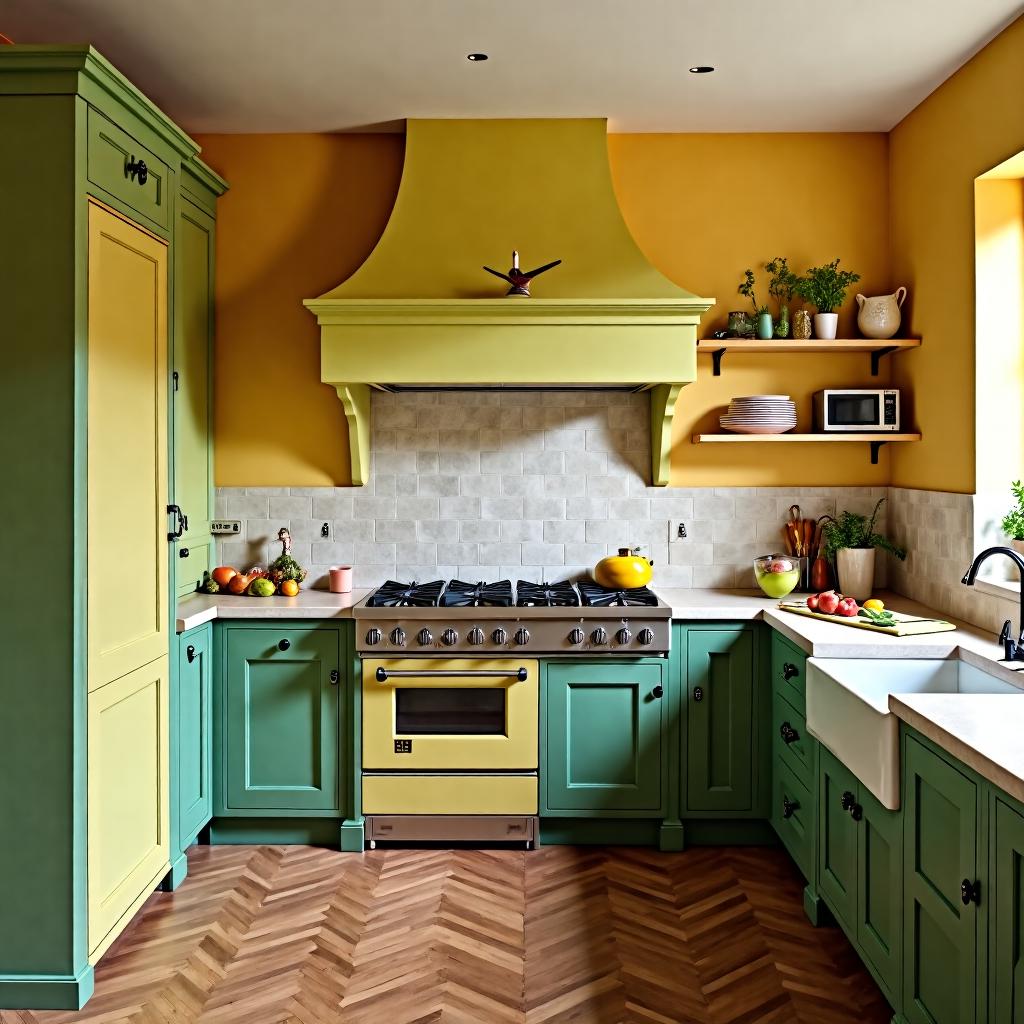}}%
        \fbox{\includegraphics[width=\mjhqfluximgwidth]{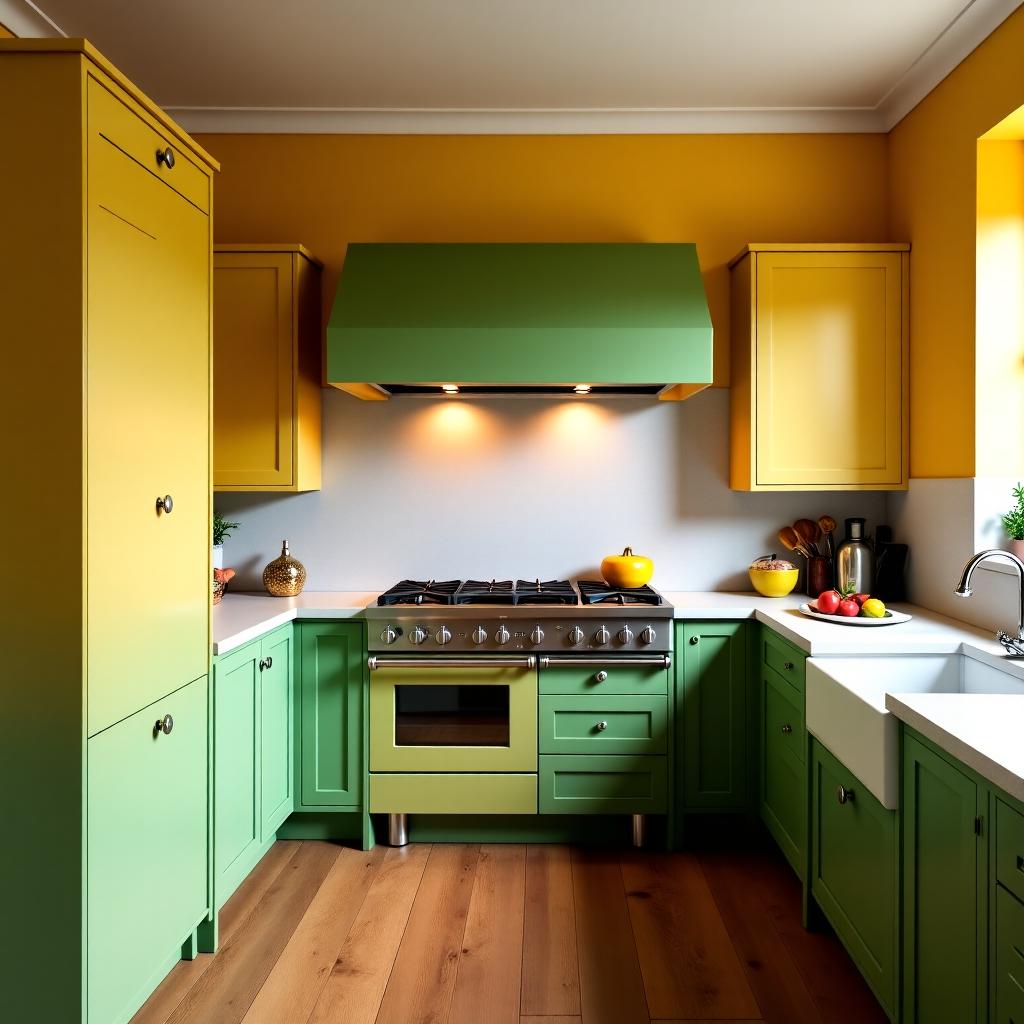}}%
        \fbox{\includegraphics[width=\mjhqfluximgwidth]{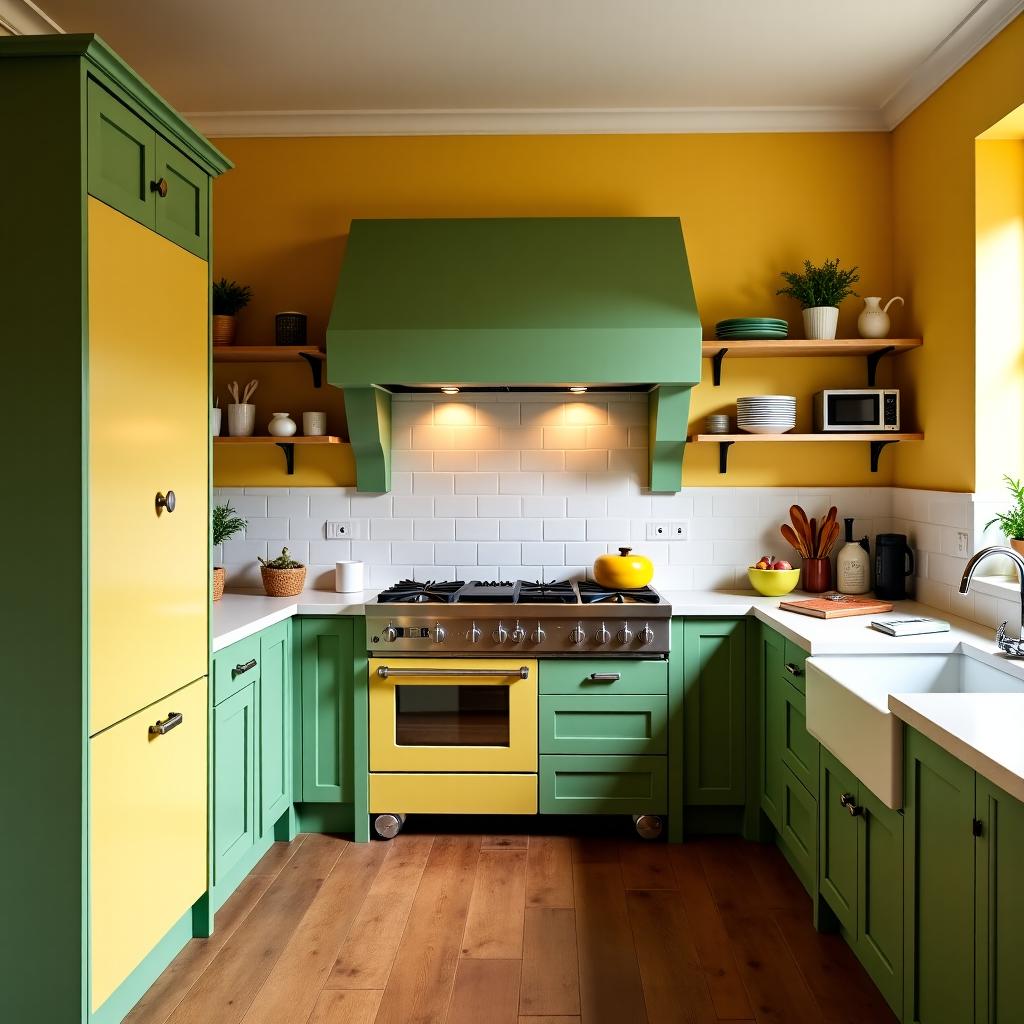}}%
        \fbox{\includegraphics[width=\mjhqfluximgwidth]{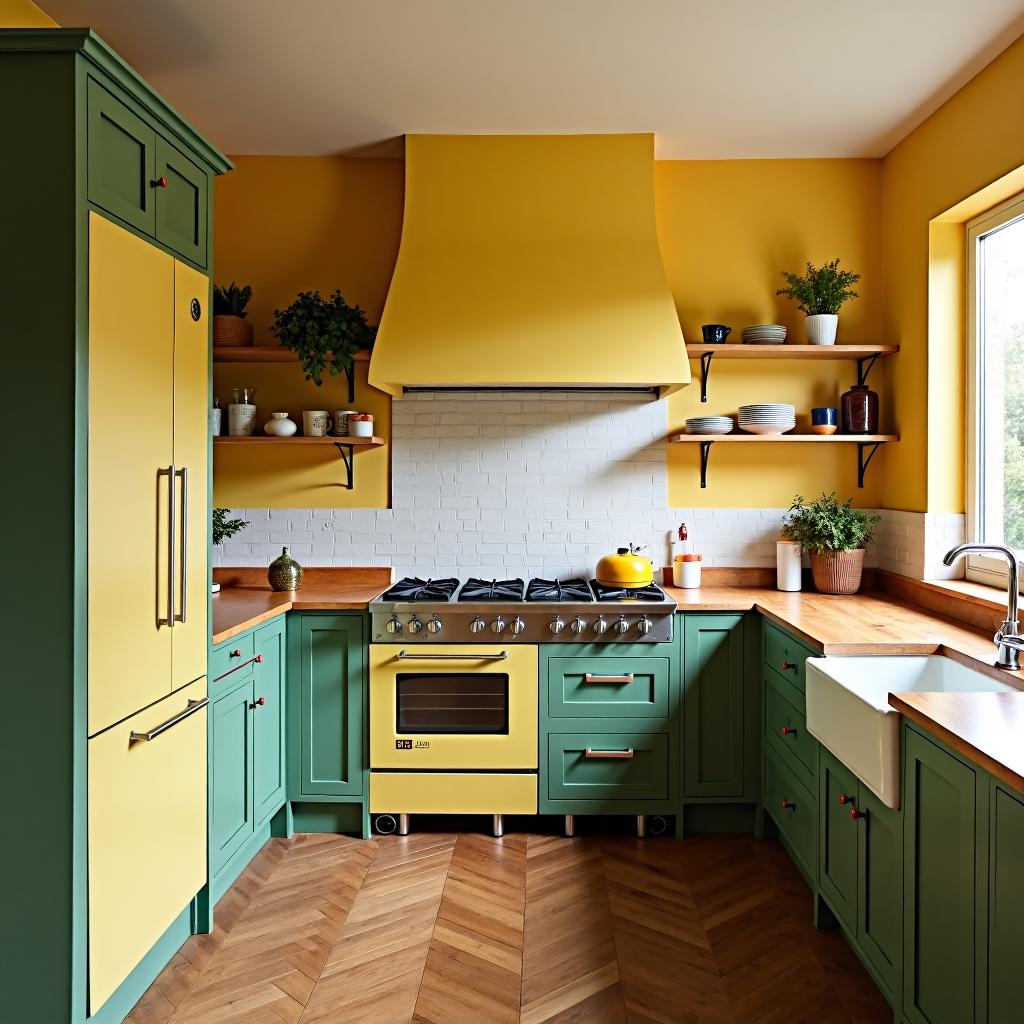}}%
        \fbox{\includegraphics[width=\mjhqfluximgwidth]{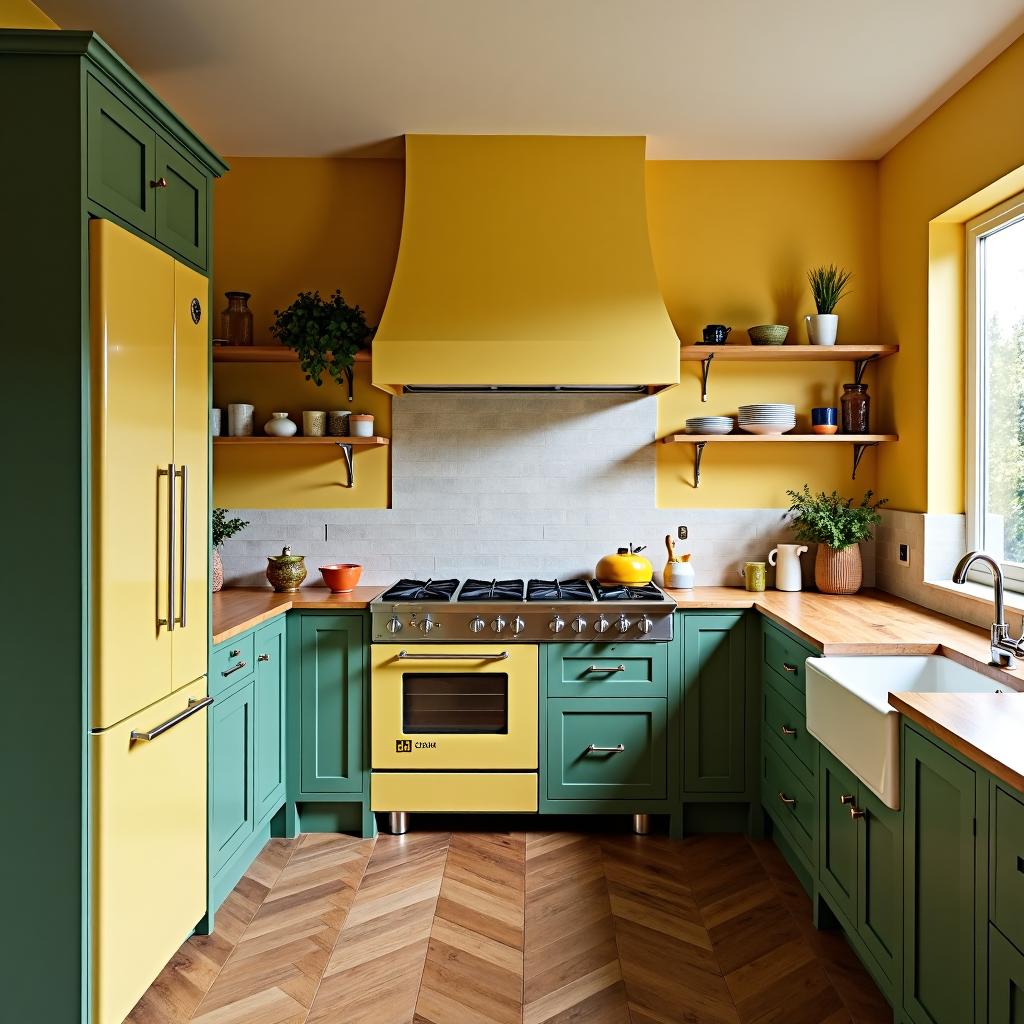}}\\[0.5ex]
        \hfill
    \end{minipage}
    \hfill
    \begin{minipage}[t]{0.395\textwidth}
        \centering
        \hfill
        \fbox{\includegraphics[width=\mjhqfluxrightimgwidth]{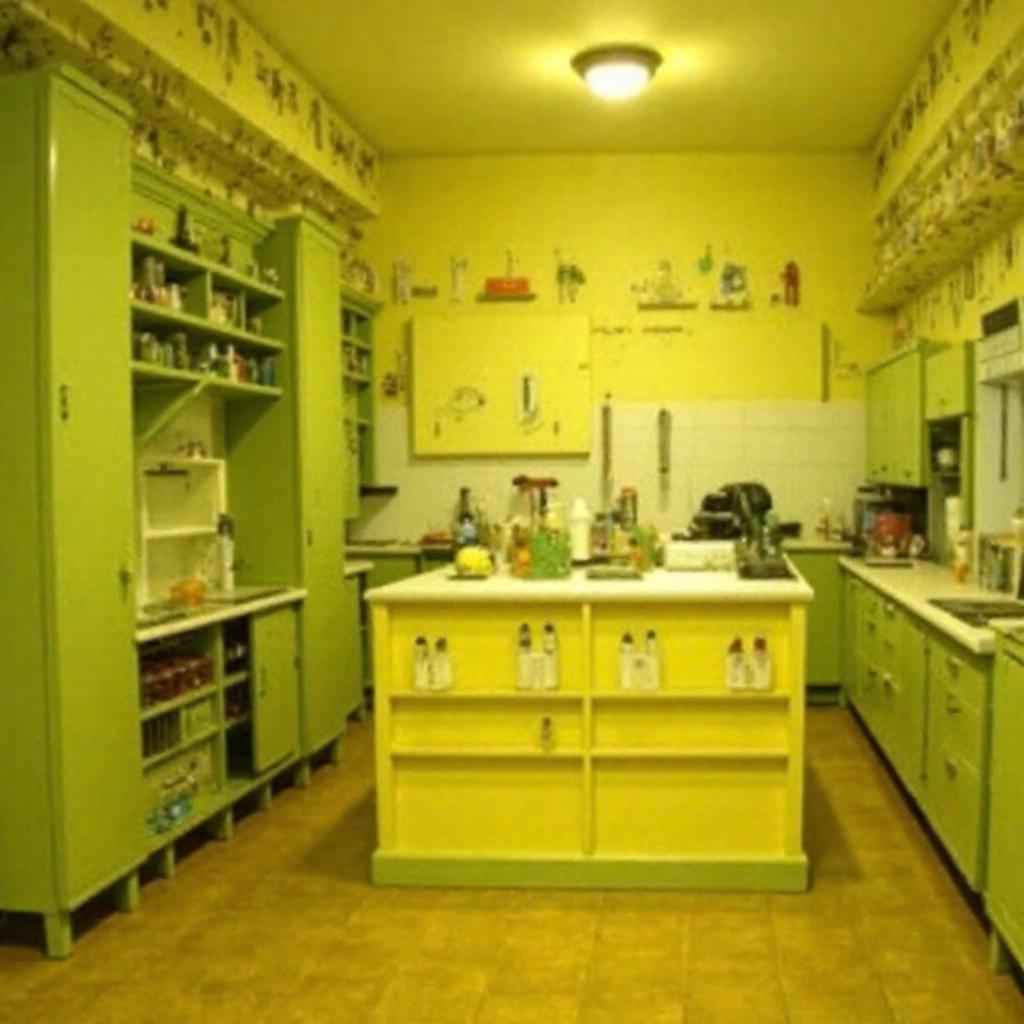}}%
        \fbox{\includegraphics[width=\mjhqfluxrightimgwidth]{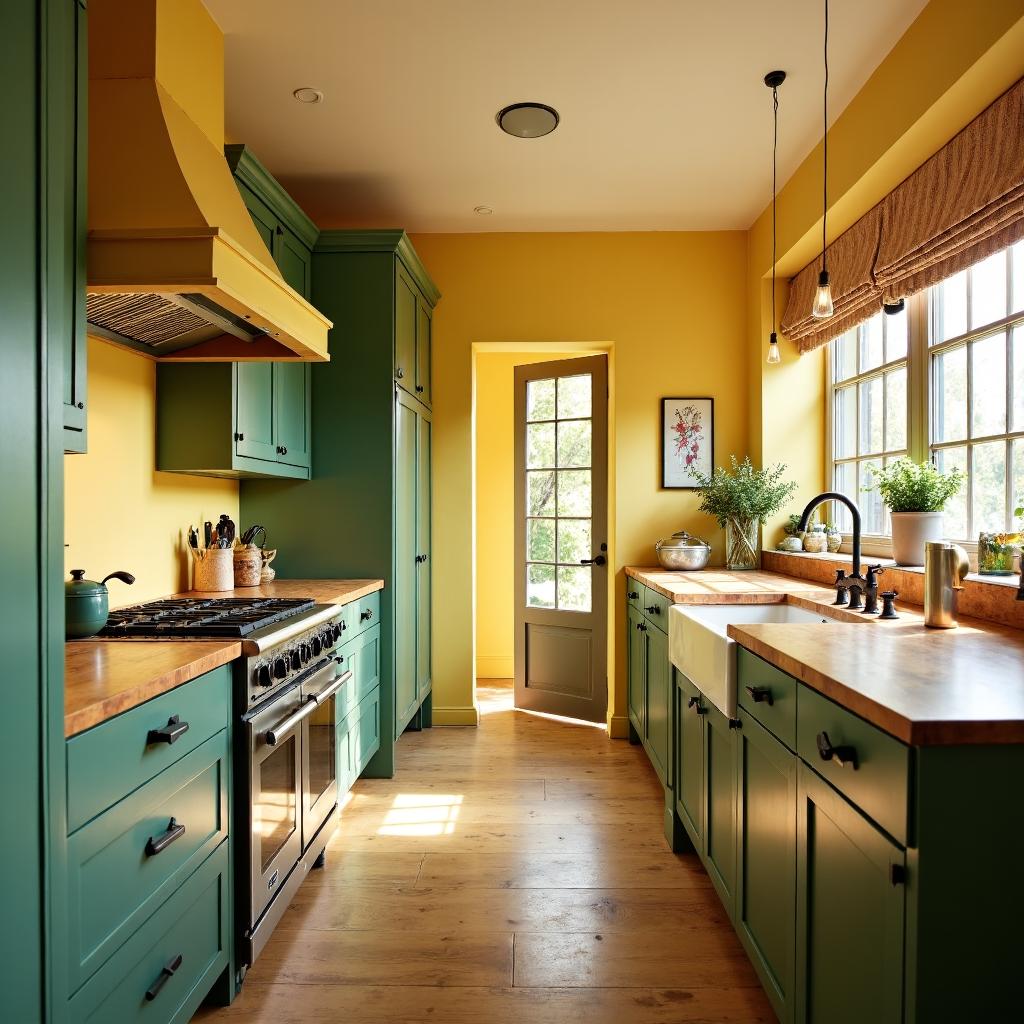}}%
        \fbox{\includegraphics[width=\mjhqfluxrightimgwidth]{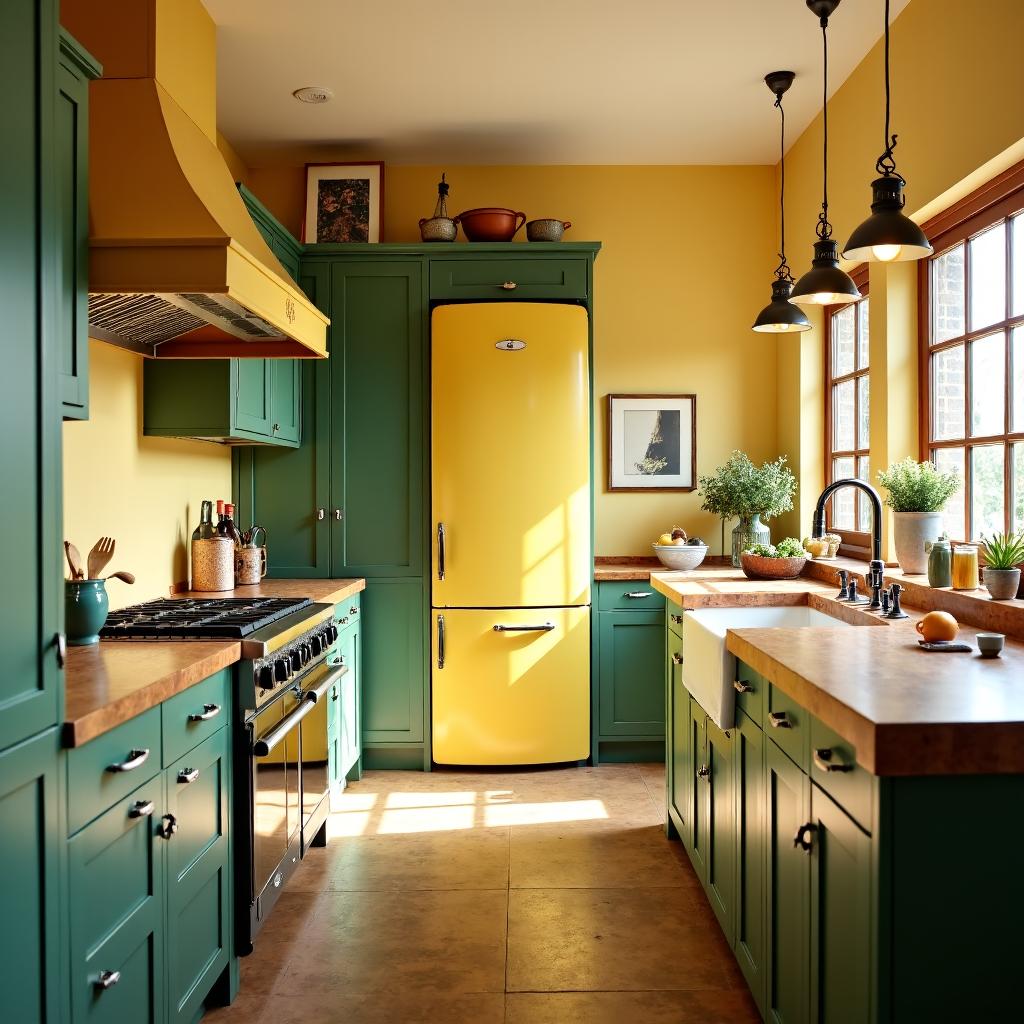}}%
        \hfill
        \fbox{\includegraphics[width=\mjhqfluxrightimgwidth]{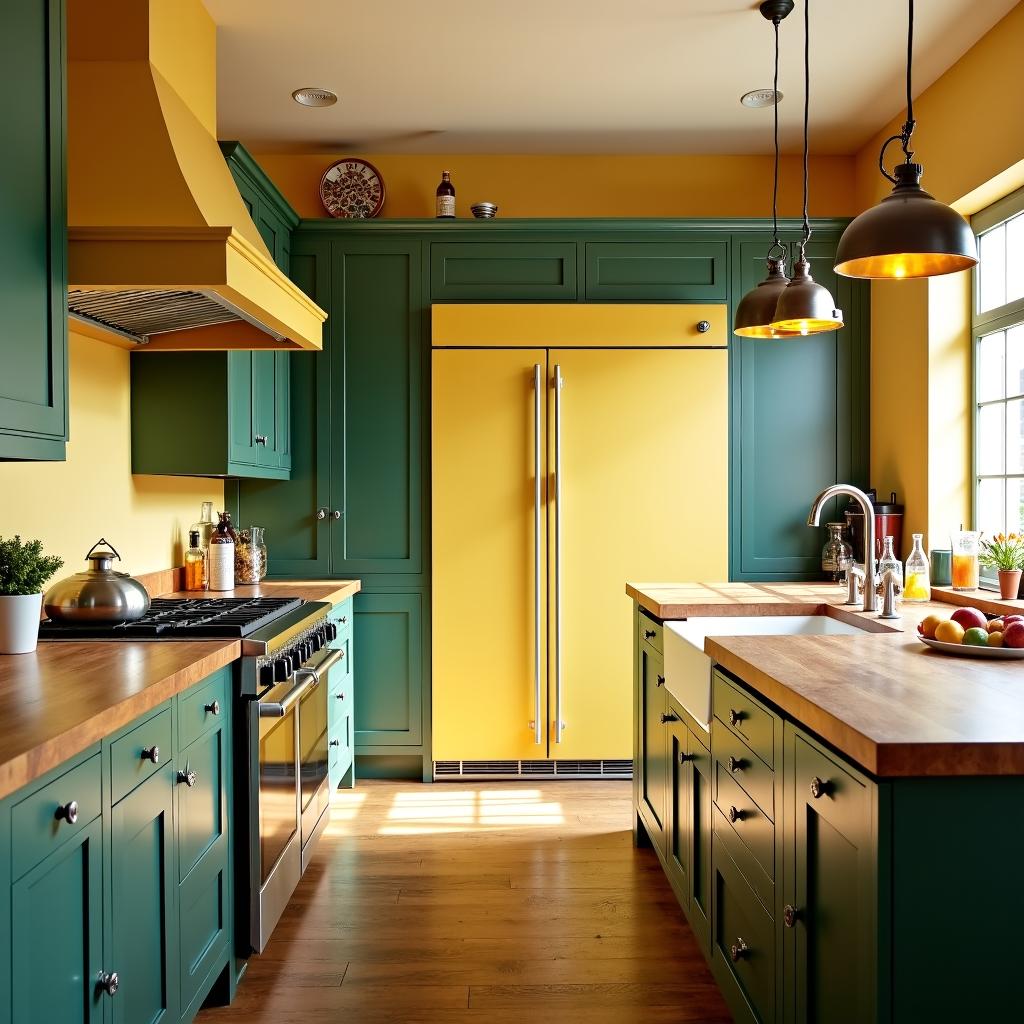}}\\[0.5ex]
    \end{minipage}
    \vspace{-15pt}
    \caption*{
        \begin{minipage}{\mjhqfluxcapwidth}
        \centering
            \footnotesize{Prompt: \textit{medium sized kitchen, bombay furniture, horizontal ombre yellow to green, French provincial style, braai in central hearth, refrigerator, large butcher block, golden hour.}}
        \end{minipage}
    }

    \vspace{0.2cm}
    \begin{minipage}[t]{0.595\textwidth}
        \centering
        \fbox{\includegraphics[width=\mjhqfluximgwidth]{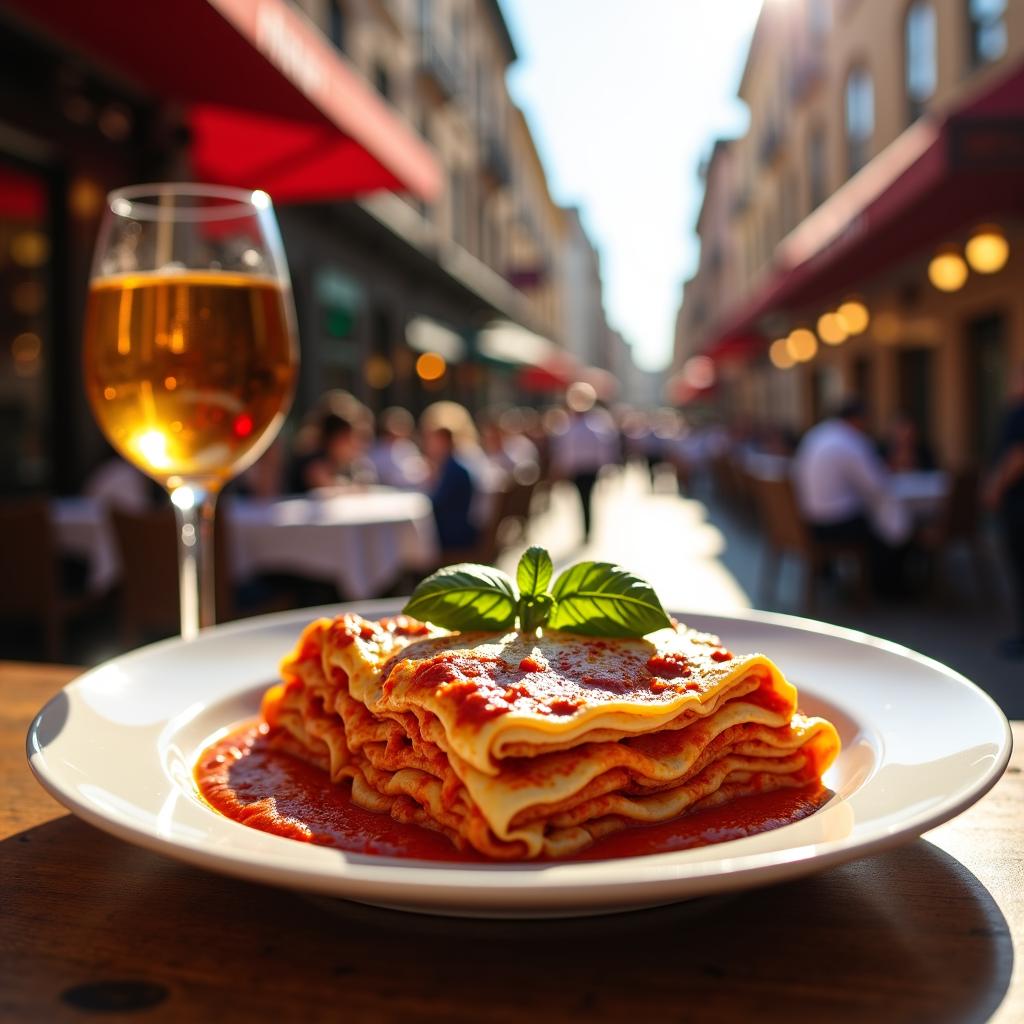}}
        \hspace{0.5mm}
        \fbox{\includegraphics[width=\mjhqfluximgwidth]{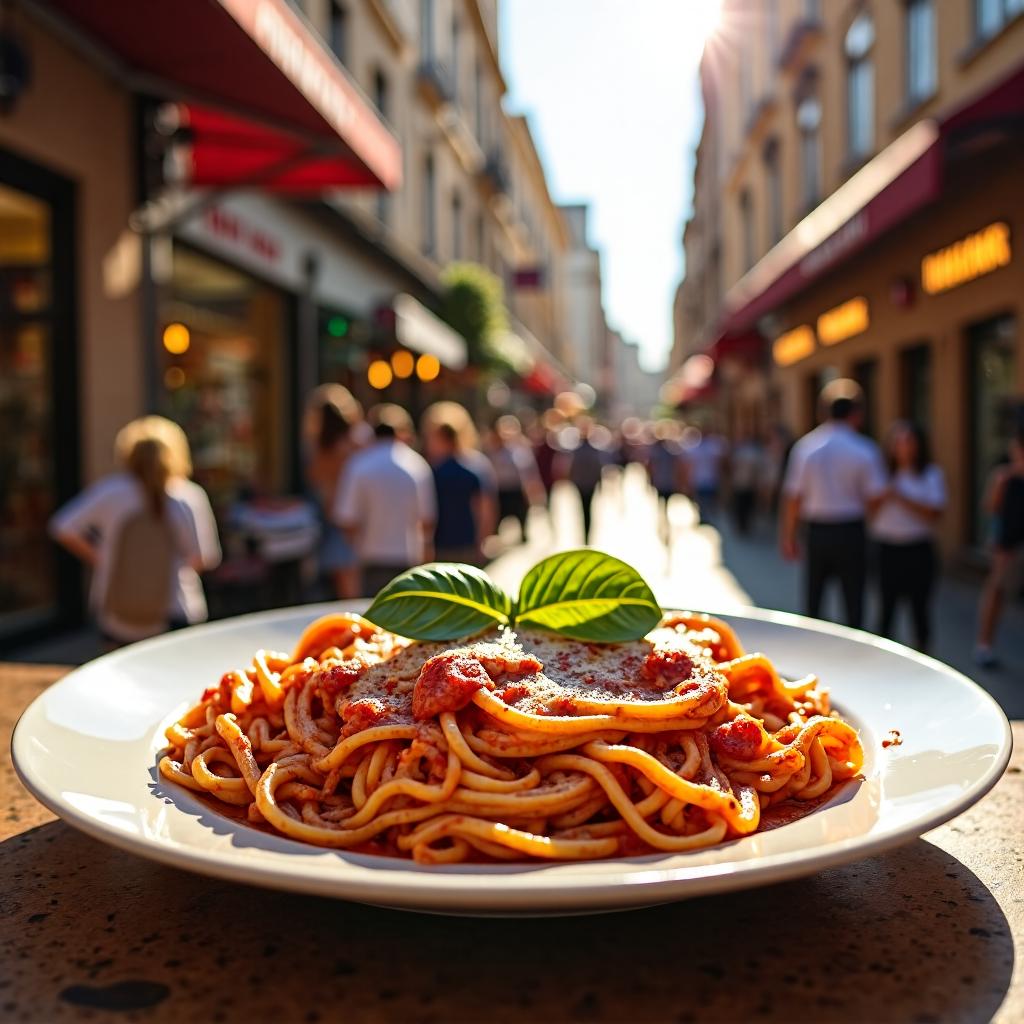}}%
        \fbox{\includegraphics[width=\mjhqfluximgwidth]{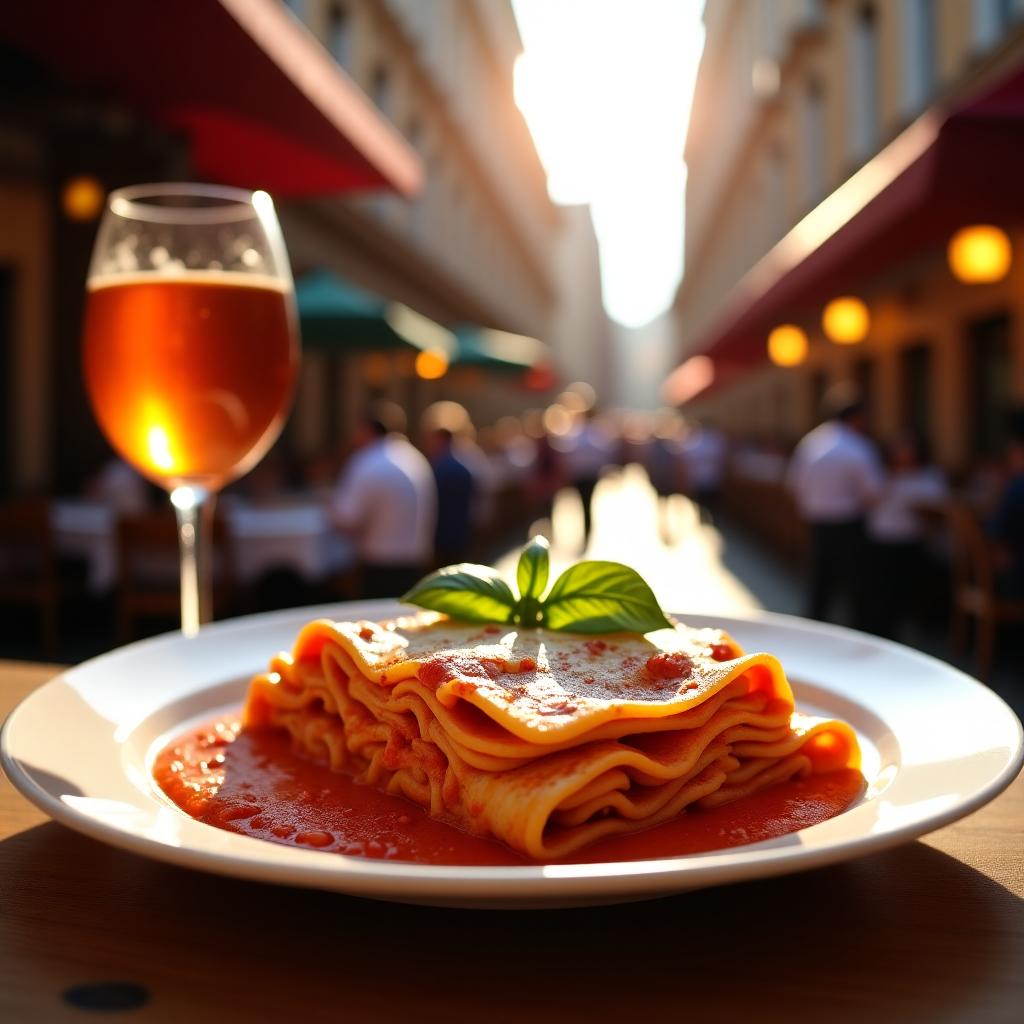}}%
        \fbox{\includegraphics[width=\mjhqfluximgwidth]{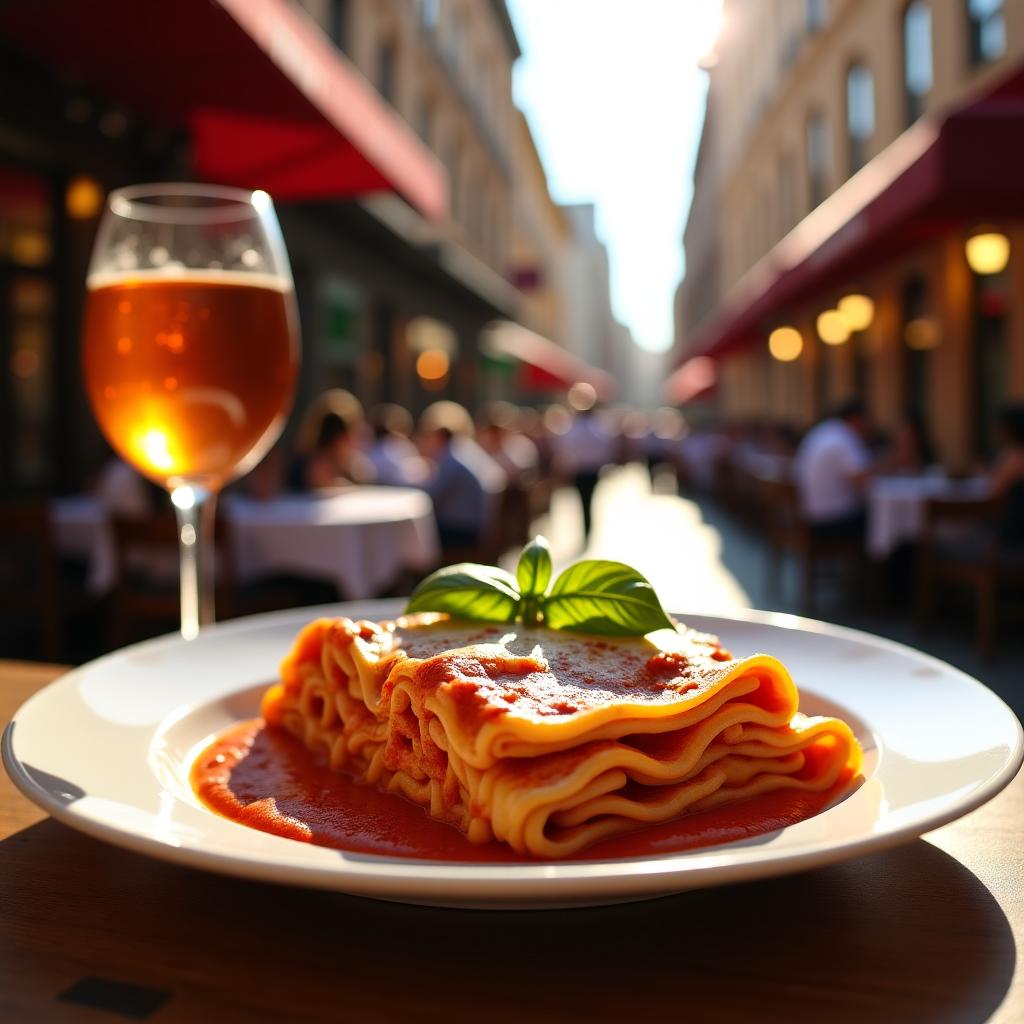}}%
        \fbox{\includegraphics[width=\mjhqfluximgwidth]{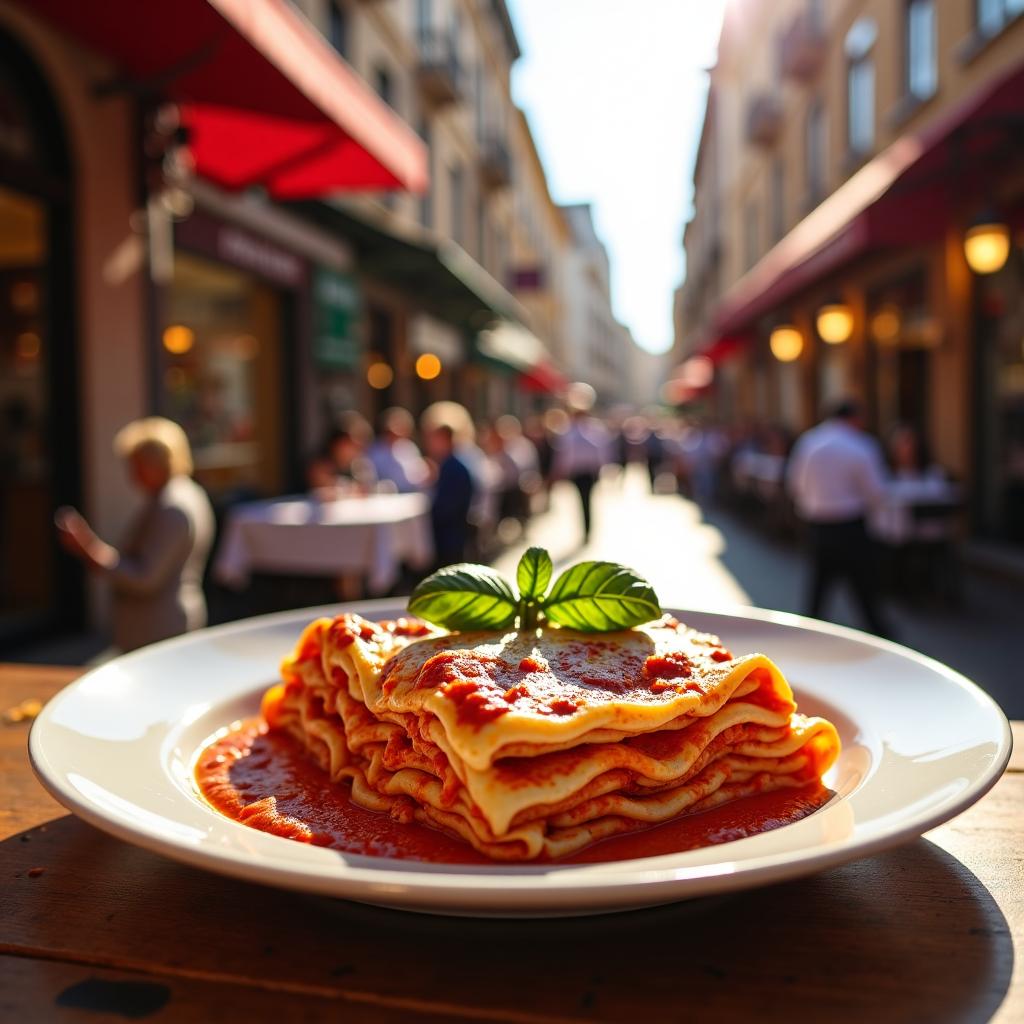}}%
        \fbox{\includegraphics[width=\mjhqfluximgwidth]{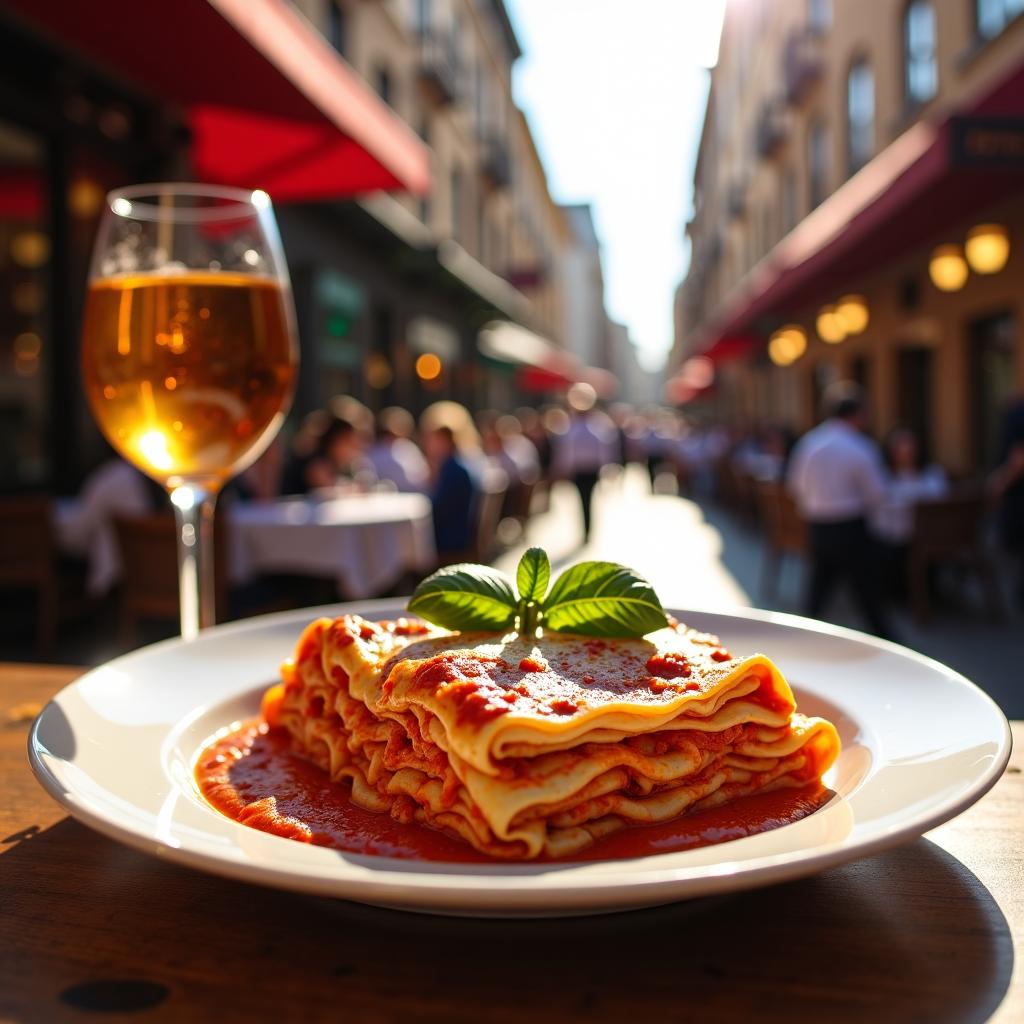}}\\[0.5ex]
        \hfill
    \end{minipage}
    \hfill
    \begin{minipage}[t]{0.395\textwidth}
        \centering
        \hfill
        \fbox{\includegraphics[width=\mjhqfluxrightimgwidth]{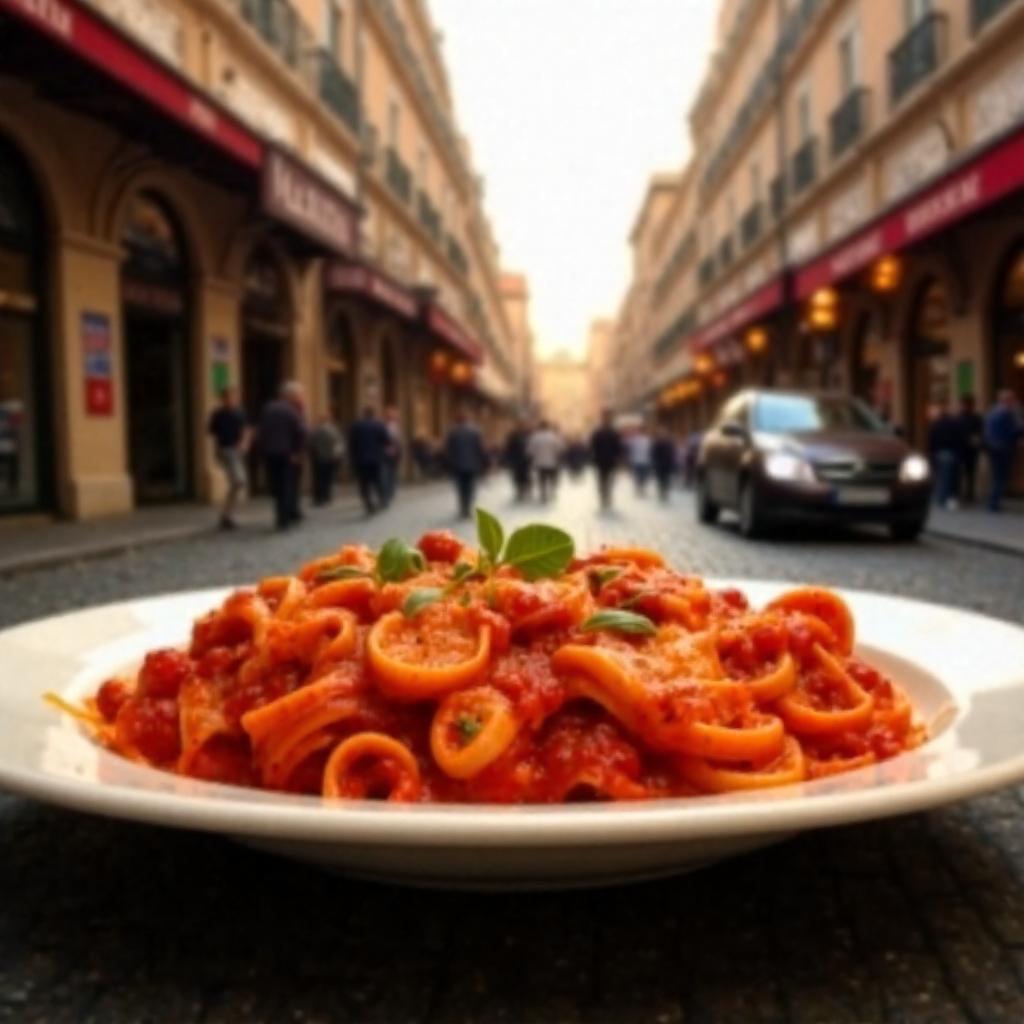}}%
        \fbox{\includegraphics[width=\mjhqfluxrightimgwidth]{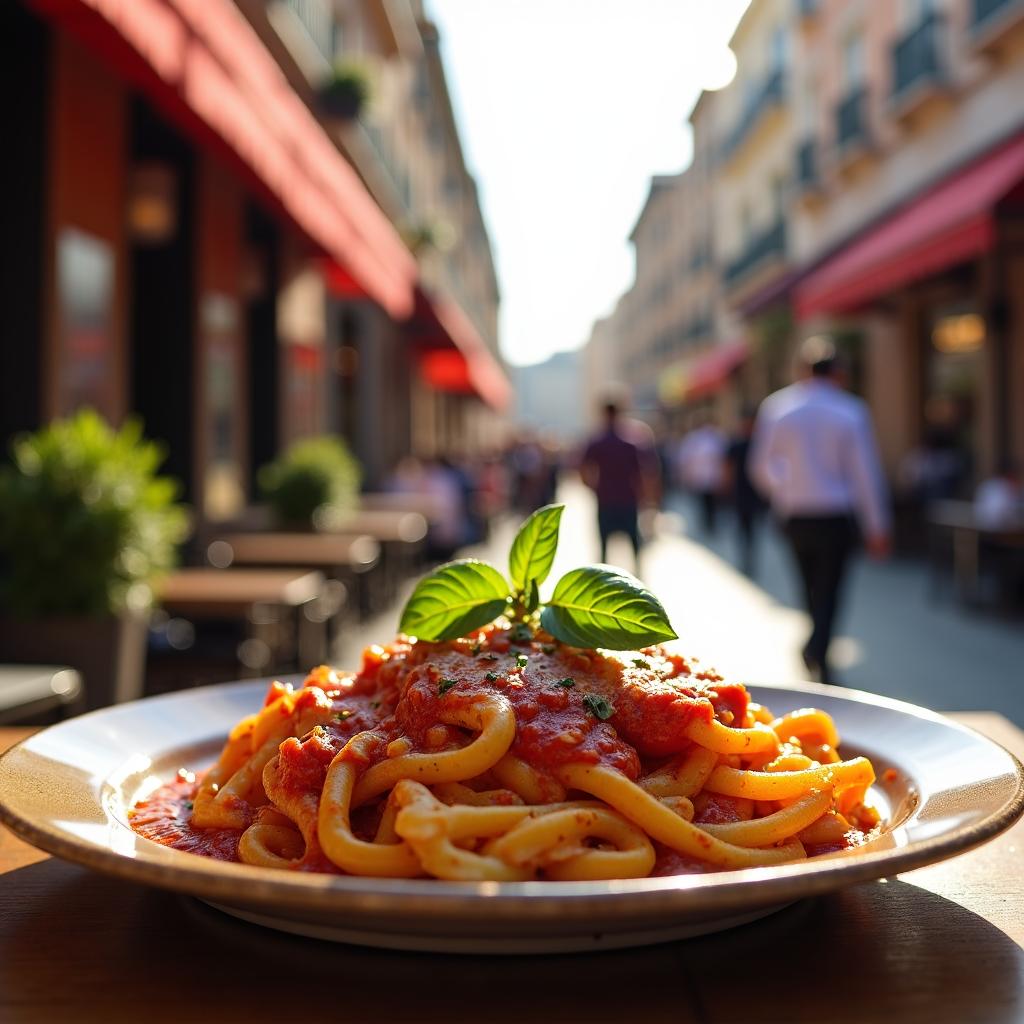}}%
        \fbox{\includegraphics[width=\mjhqfluxrightimgwidth]{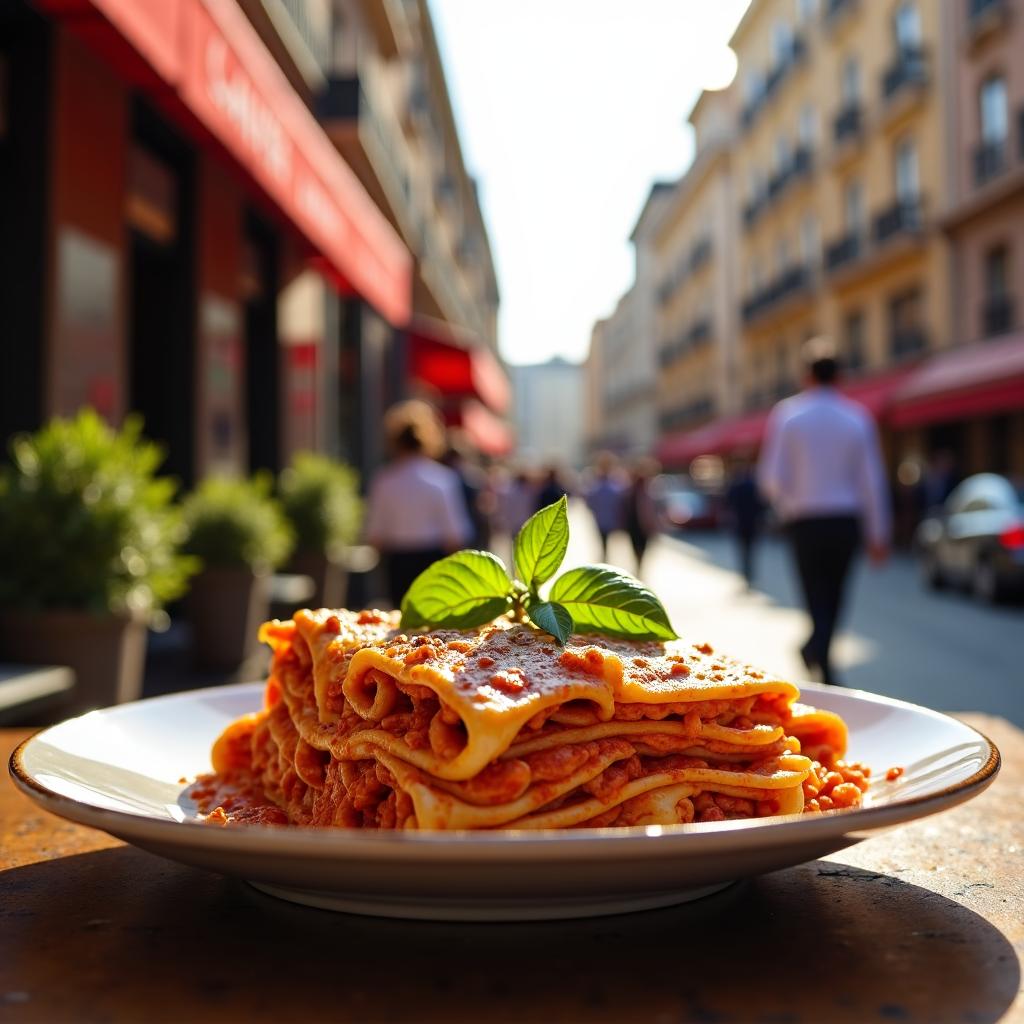}}%
        \hfill
        \fbox{\includegraphics[width=\mjhqfluxrightimgwidth]{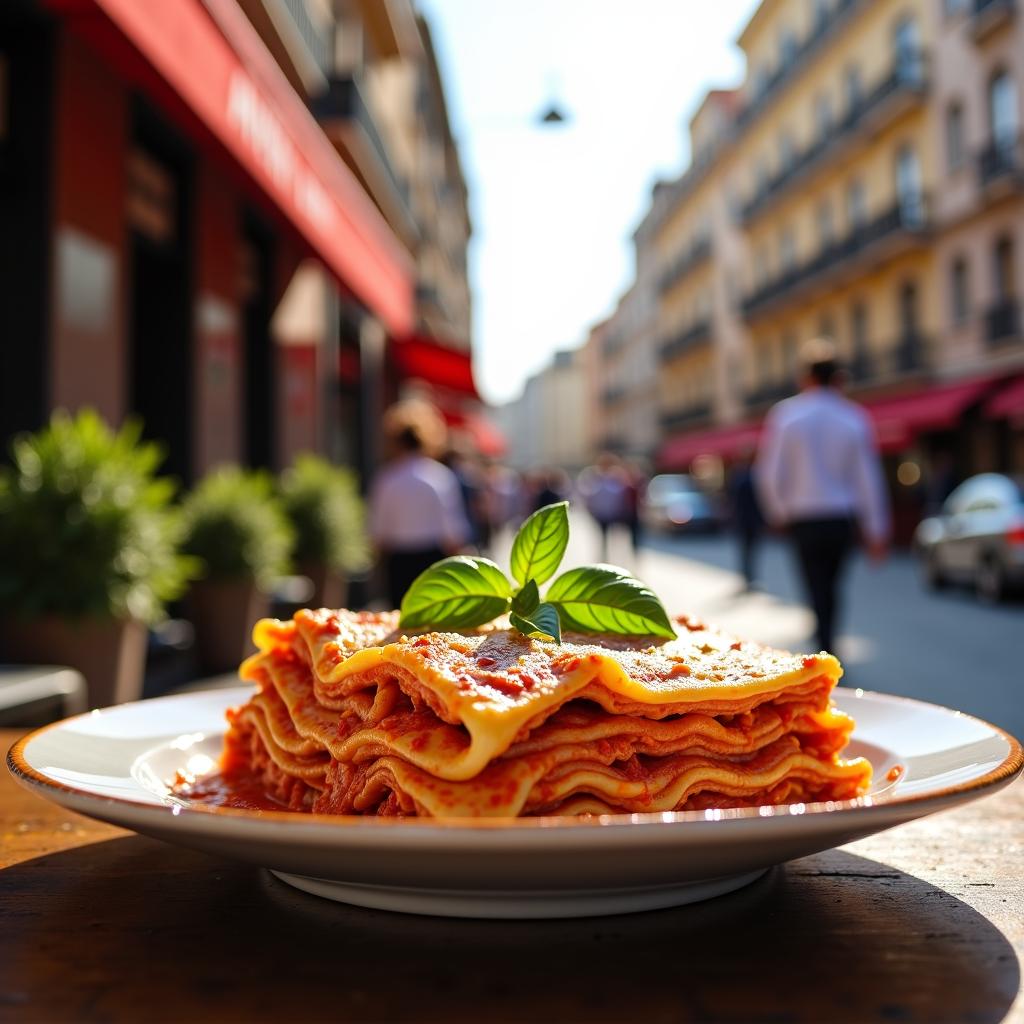}}\\[0.5ex]
    \end{minipage}
    \vspace{-15pt}
    \caption*{
        \begin{minipage}{\mjhqfluxcapwidth}
        \centering
            \footnotesize{Prompt: \textit{a lasagne outside an italian restaurant in the city bologna at midday with bright light. Ultra High Definition. High detailed. HD. Photorealistic.}}
        \end{minipage}
    }

    \vspace{0.2cm}
    \begin{minipage}[t]{0.595\textwidth}
        \centering
        \fbox{\includegraphics[width=\mjhqfluximgwidth]{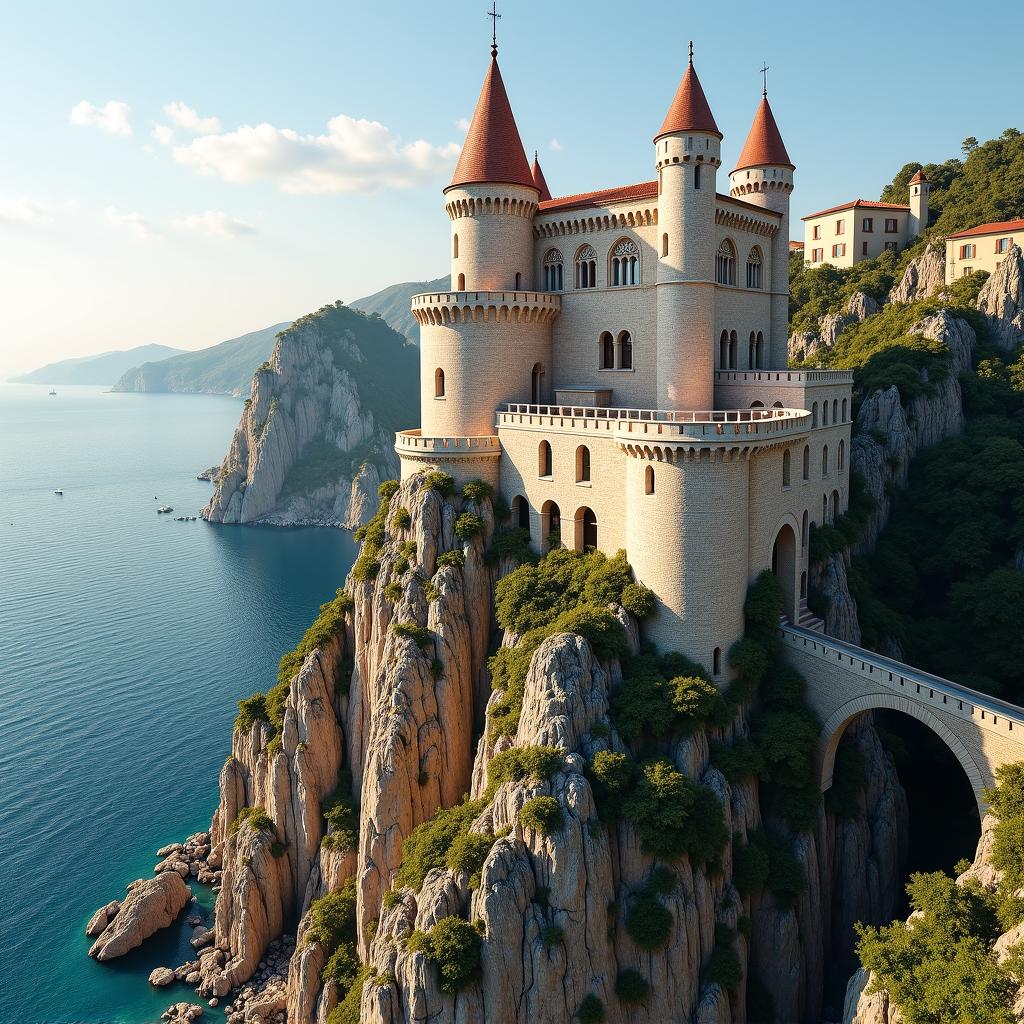}}
        \hspace{0.5mm}
        \fbox{\includegraphics[width=\mjhqfluximgwidth]{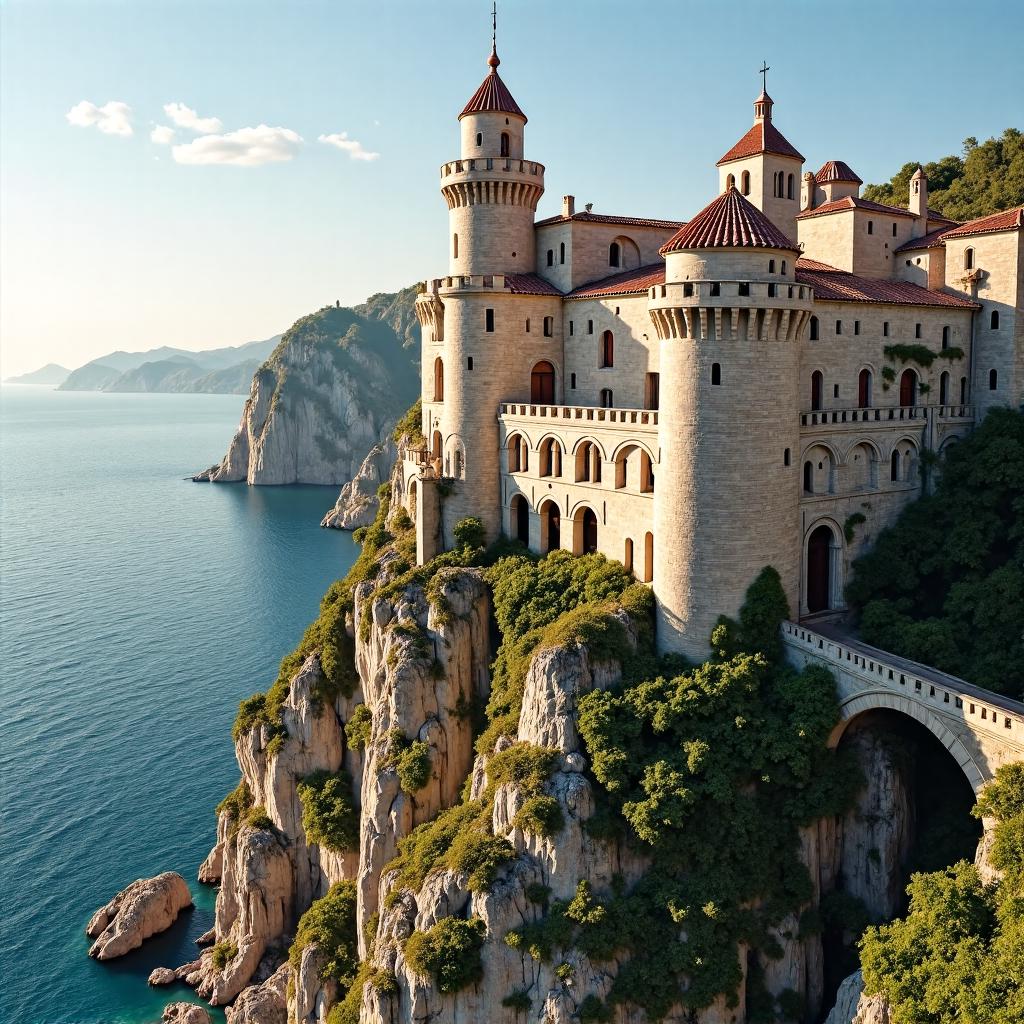}}%
        \fbox{\includegraphics[width=\mjhqfluximgwidth]{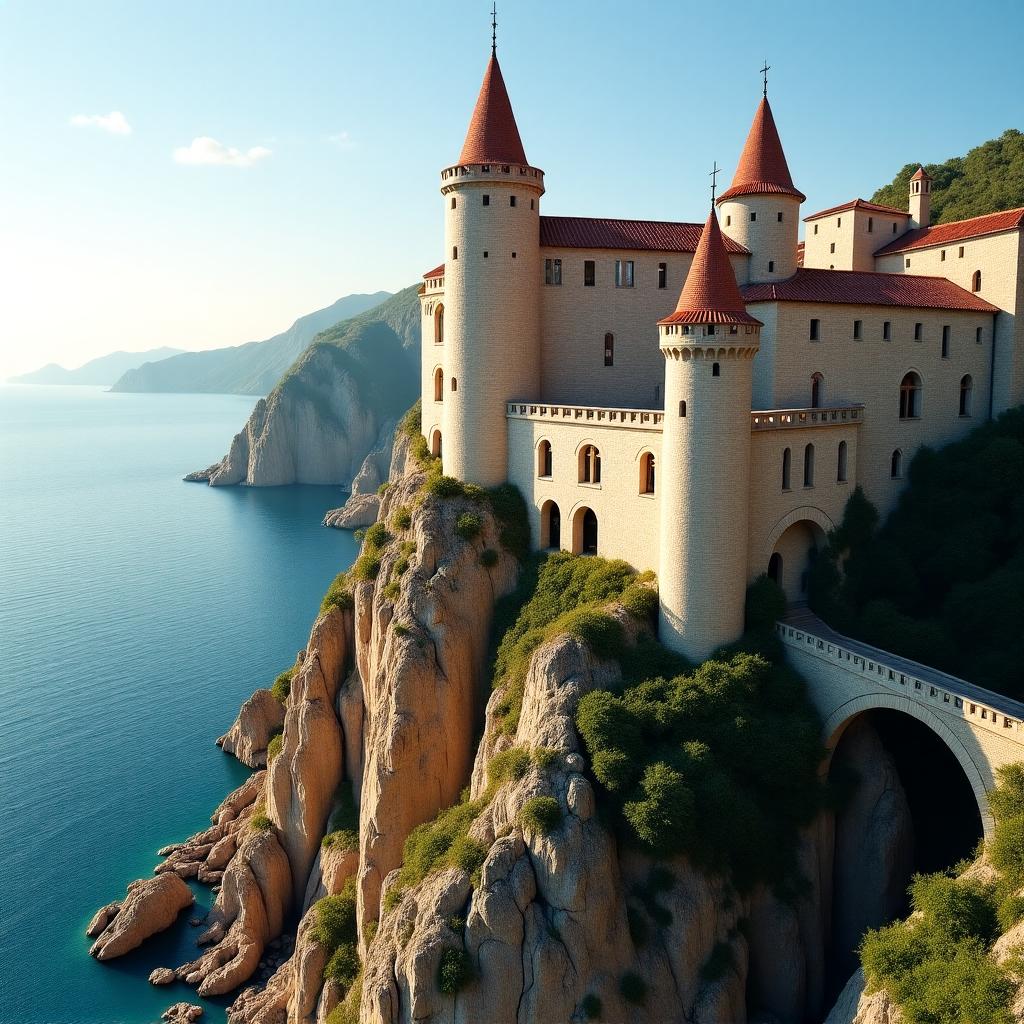}}%
        \fbox{\includegraphics[width=\mjhqfluximgwidth]{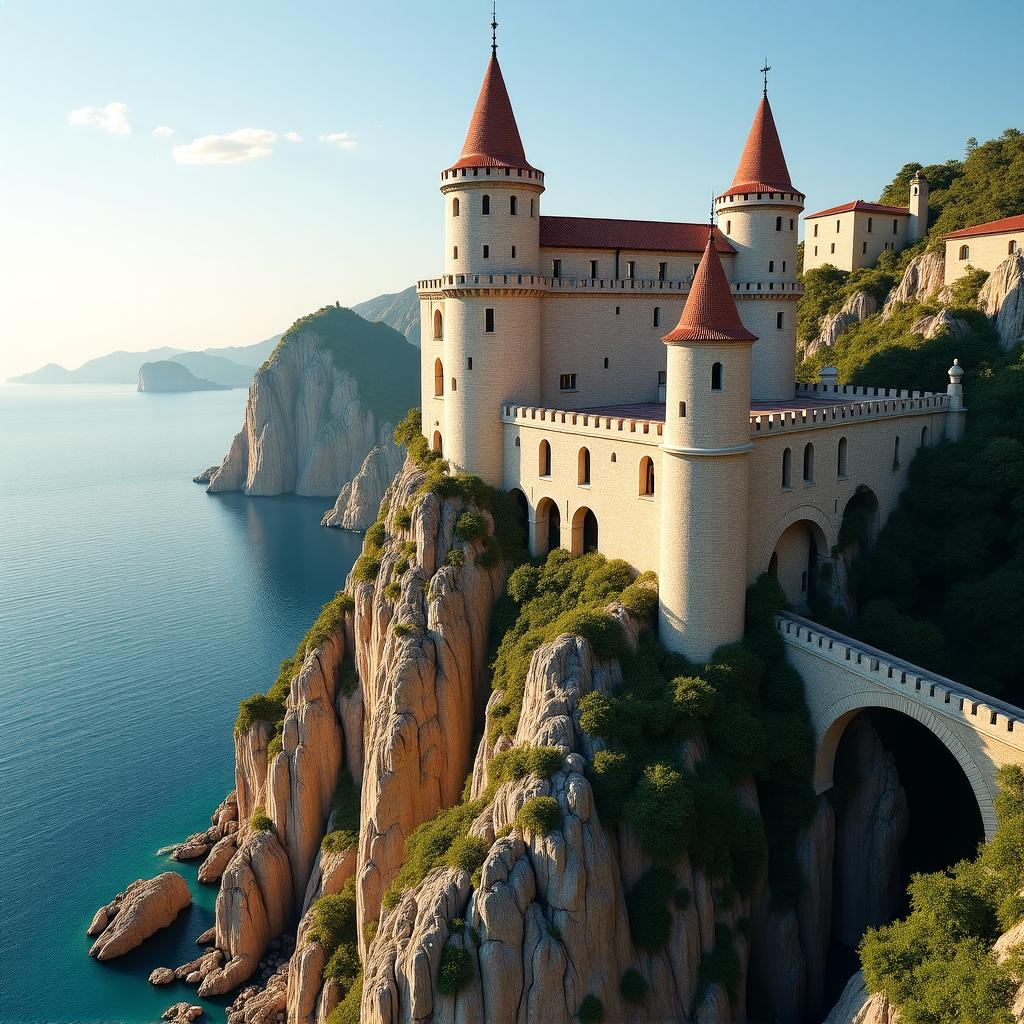}}%
        \fbox{\includegraphics[width=\mjhqfluximgwidth]{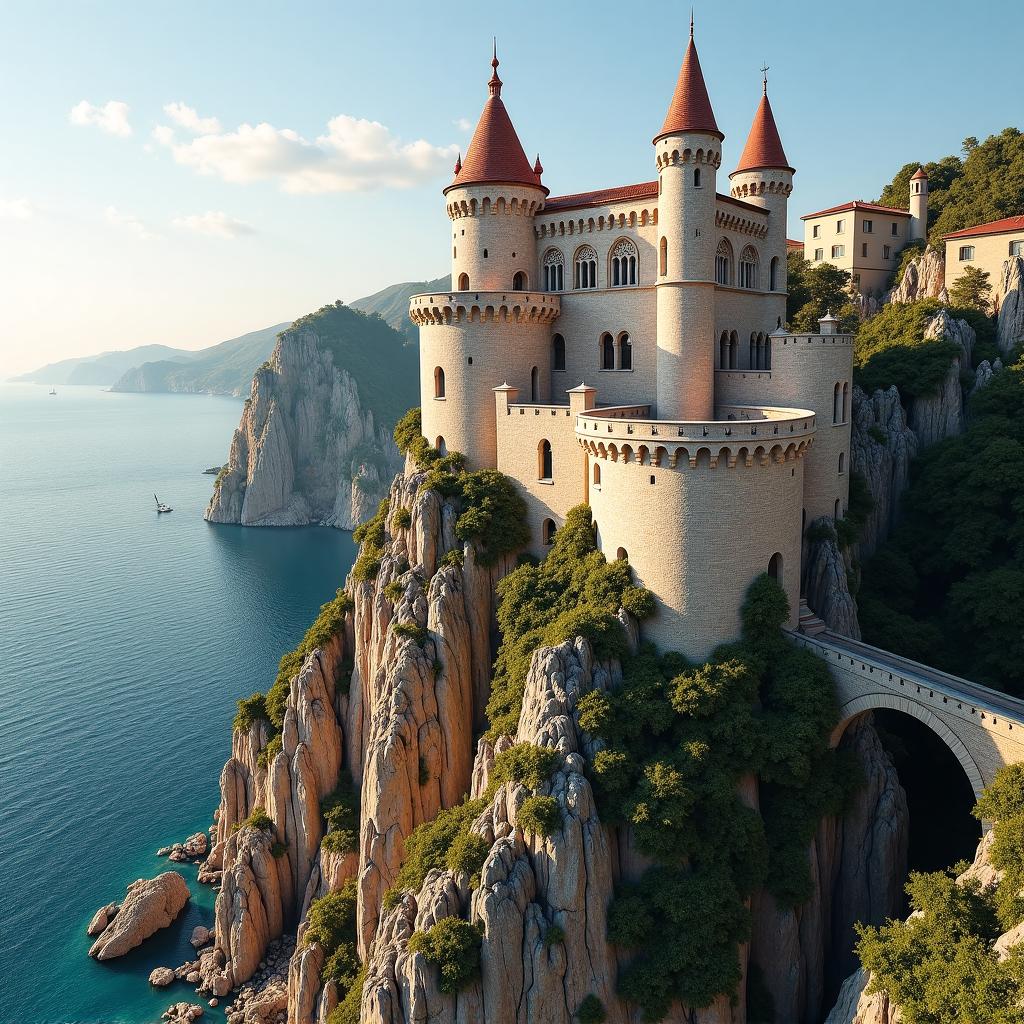}}%
        \fbox{\includegraphics[width=\mjhqfluximgwidth]{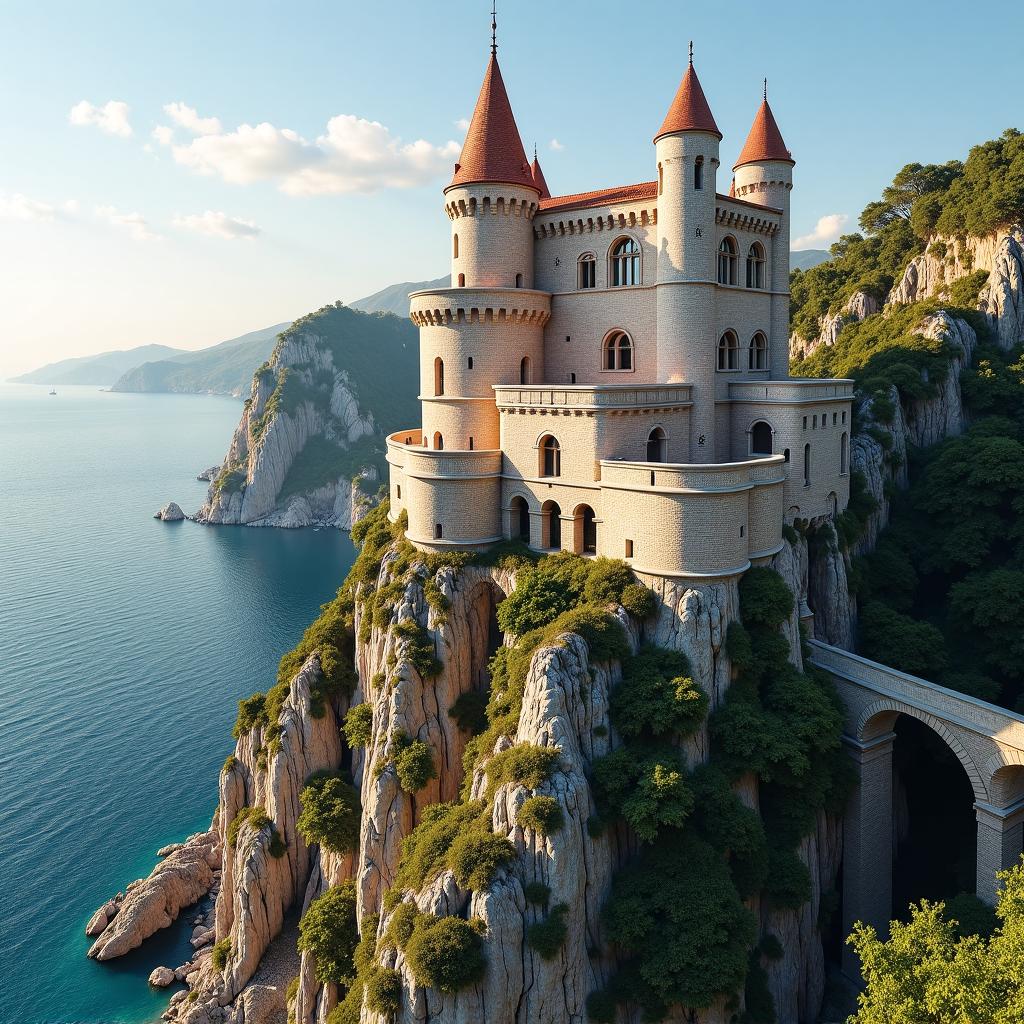}}\\[0.5ex]
        \hfill
    \end{minipage}
    \hfill
    \begin{minipage}[t]{0.395\textwidth}
        \centering
        \hfill
        \fbox{\includegraphics[width=\mjhqfluxrightimgwidth]{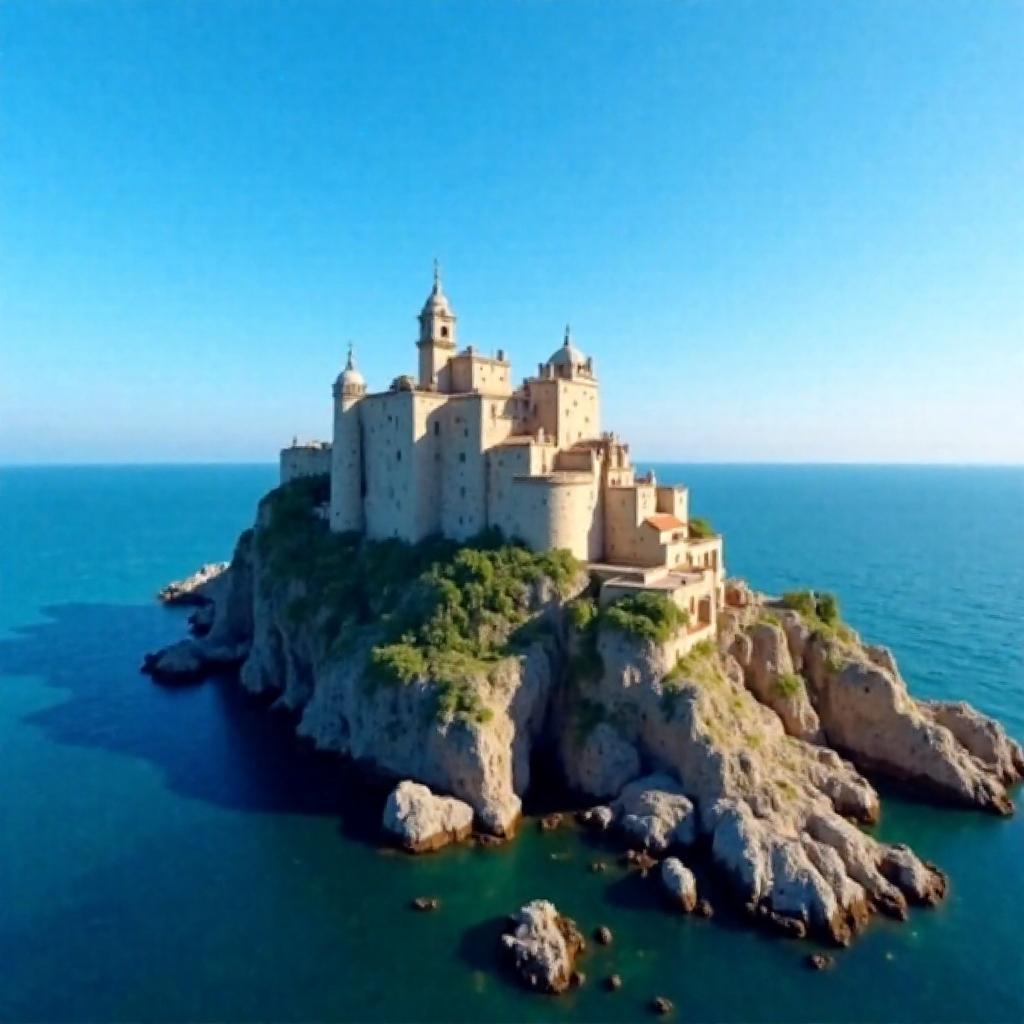}}%
        \fbox{\includegraphics[width=\mjhqfluxrightimgwidth]{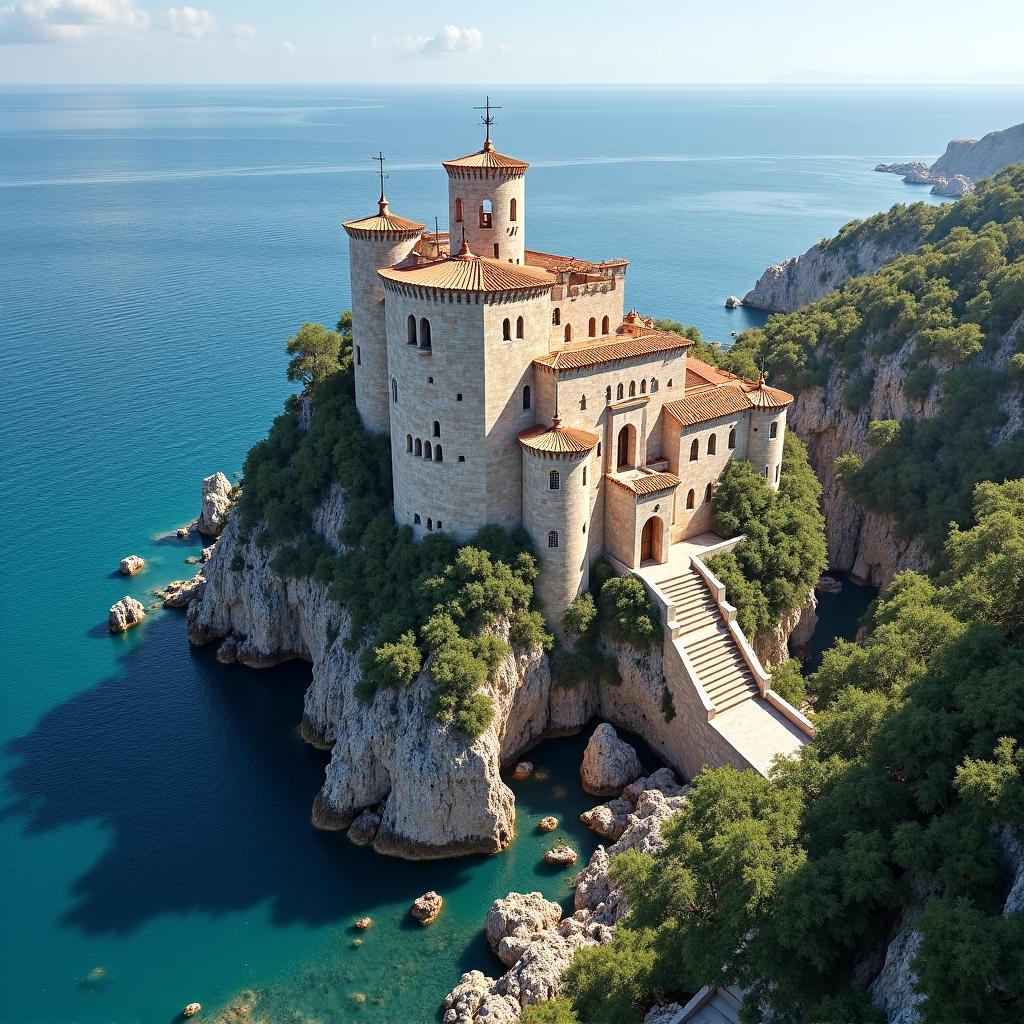}}%
        \fbox{\includegraphics[width=\mjhqfluxrightimgwidth]{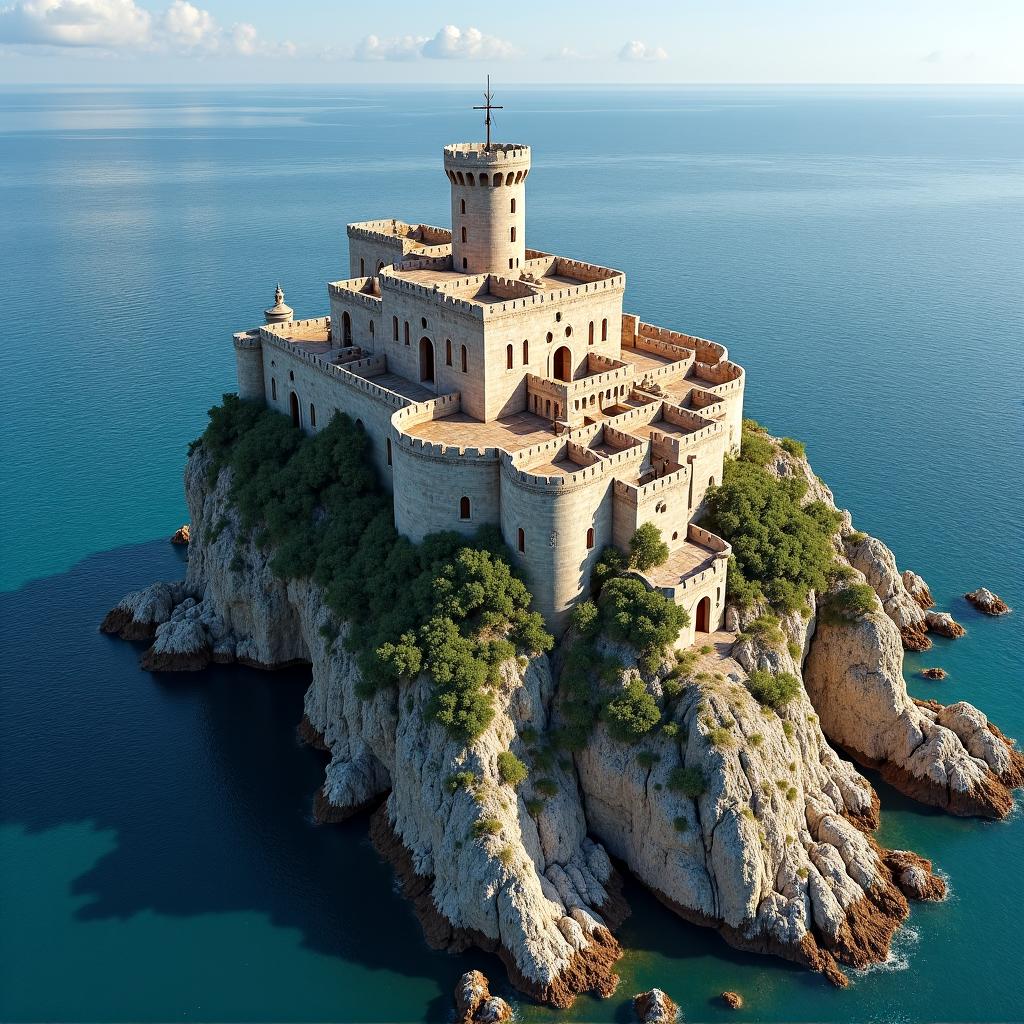}}%
        \hfill
        \fbox{\includegraphics[width=\mjhqfluxrightimgwidth]{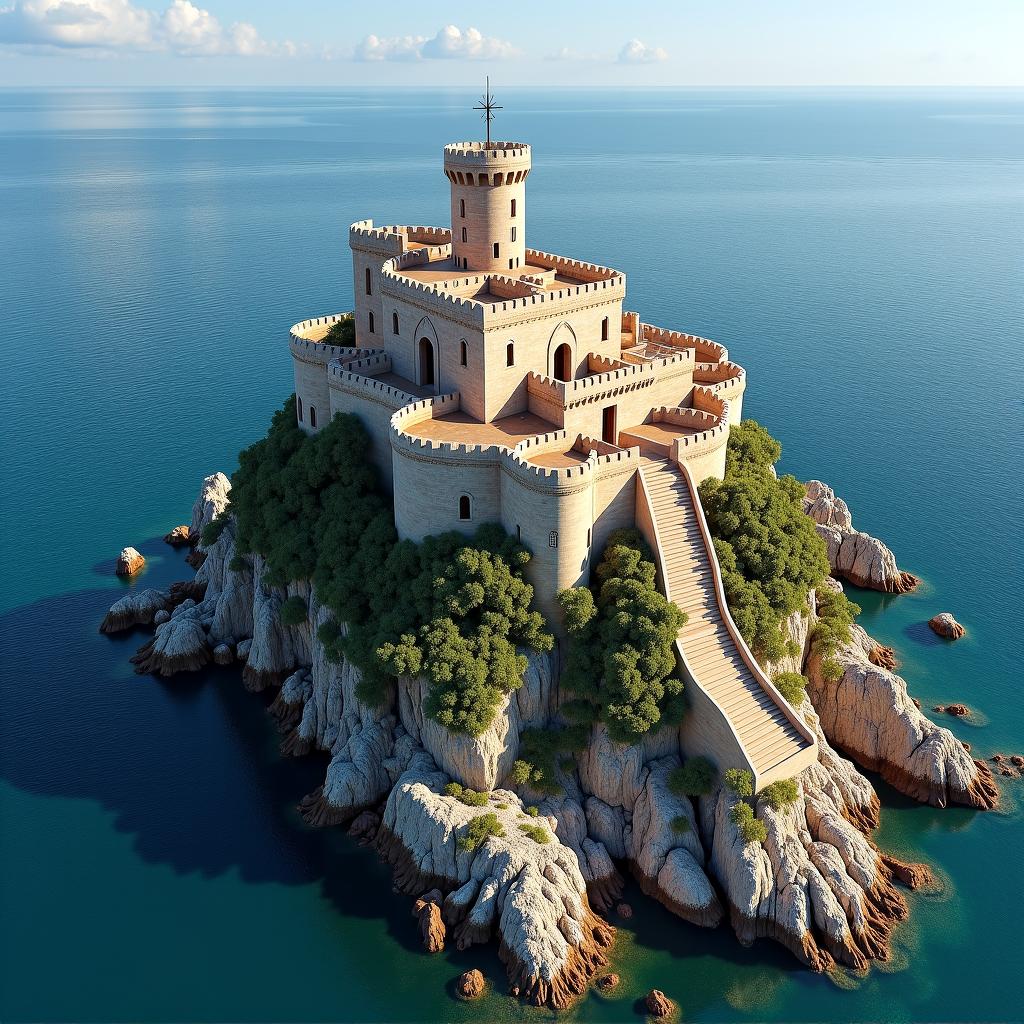}}\\[0.5ex]
    \end{minipage}
    \vspace{-15pt}
    \caption*{
        \begin{minipage}{\mjhqfluxcapwidth}
        \centering
            \footnotesize{Prompt: \textit{birdseye view of Mediterranean castle with Romanesque revival influences, photorealistic, 8k highest resolution, horizon line visible in the back, rocky cliff tropical and clear sky.}}
        \end{minipage}
    }

    \vspace{0.2cm}
    \begin{minipage}[t]{0.595\textwidth}
        \centering
        \fbox{\includegraphics[width=\mjhqfluximgwidth]{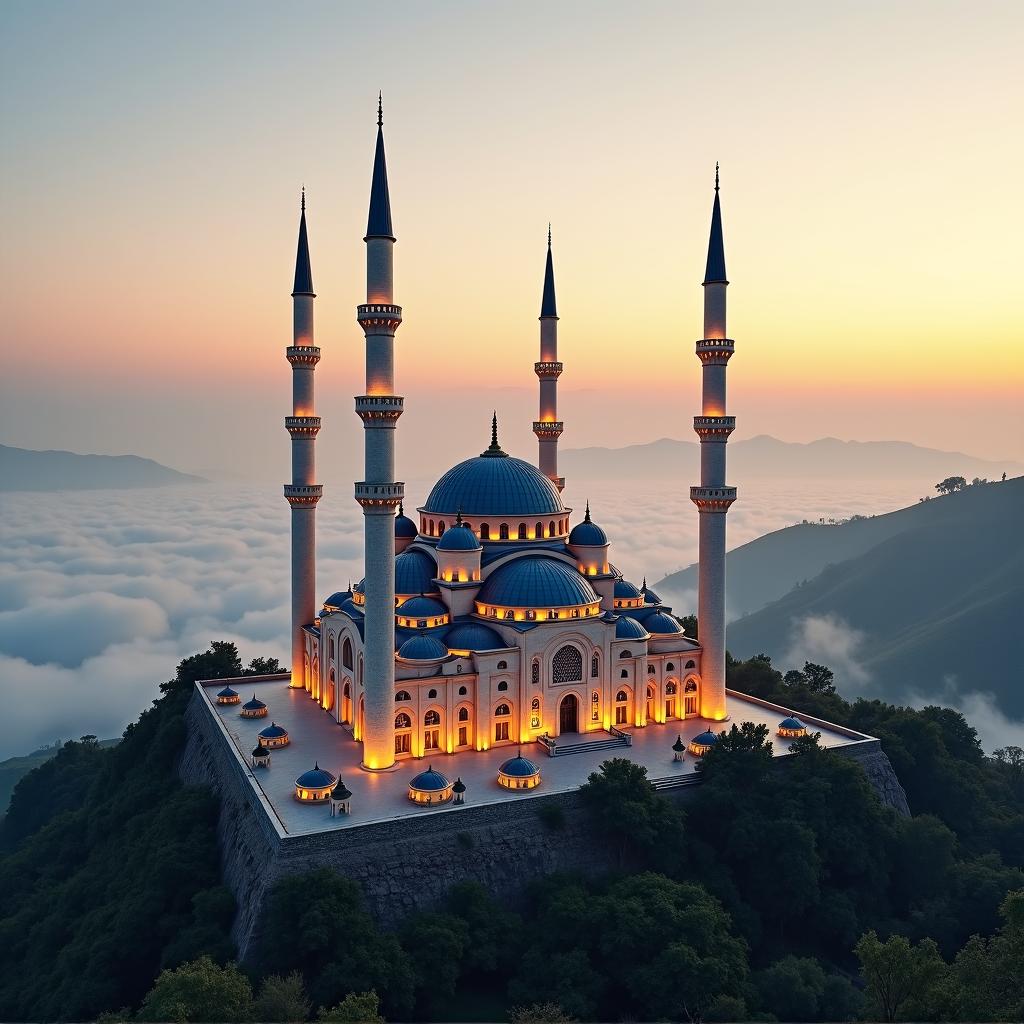}}
        \hspace{0.5mm}
        \fbox{\includegraphics[width=\mjhqfluximgwidth]{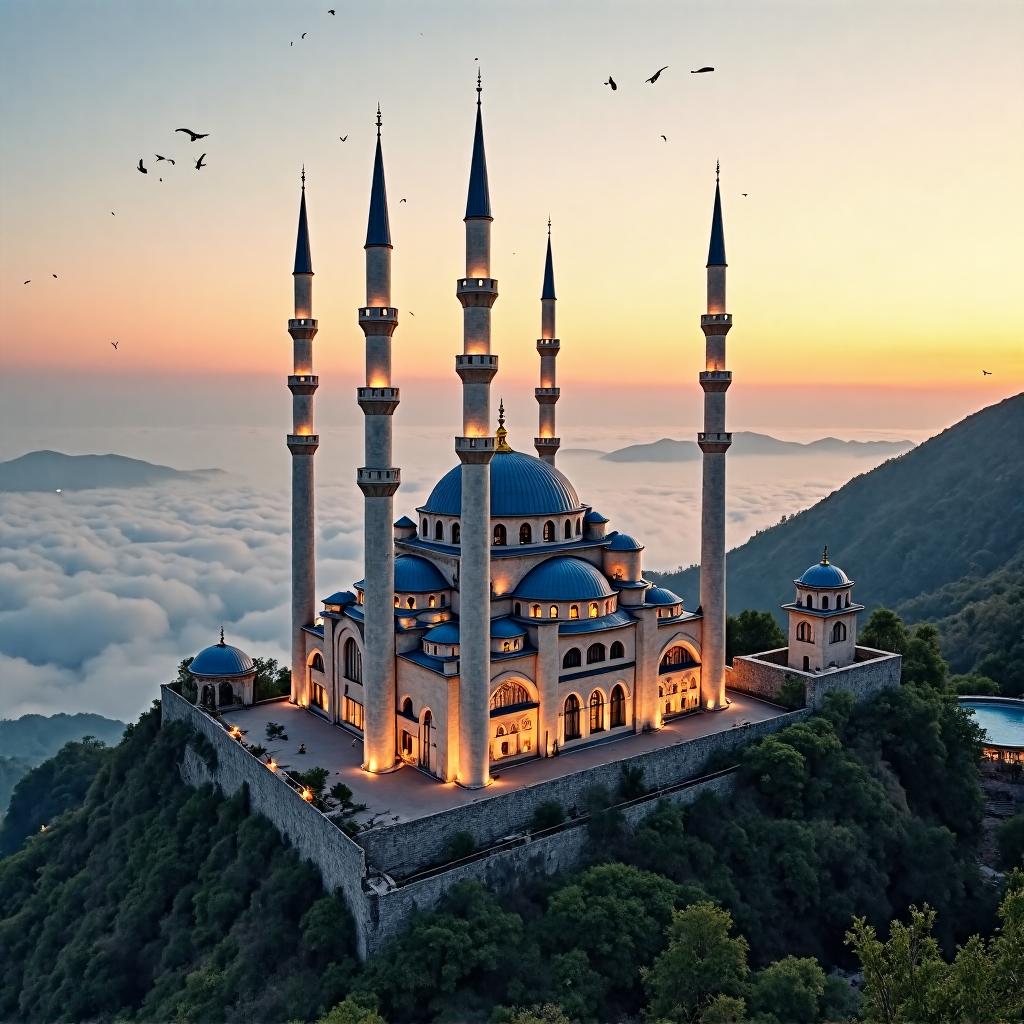}}%
        \fbox{\includegraphics[width=\mjhqfluximgwidth]{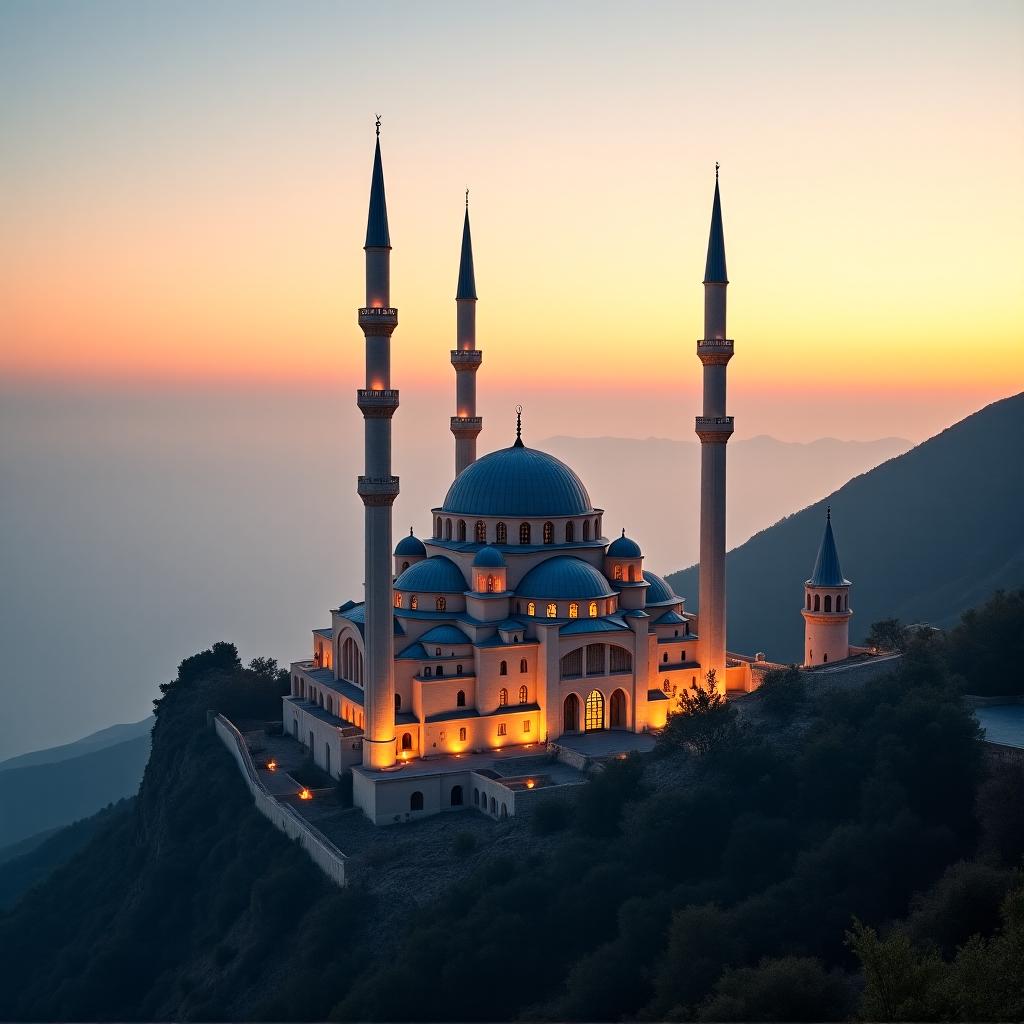}}%
        \fbox{\includegraphics[width=\mjhqfluximgwidth]{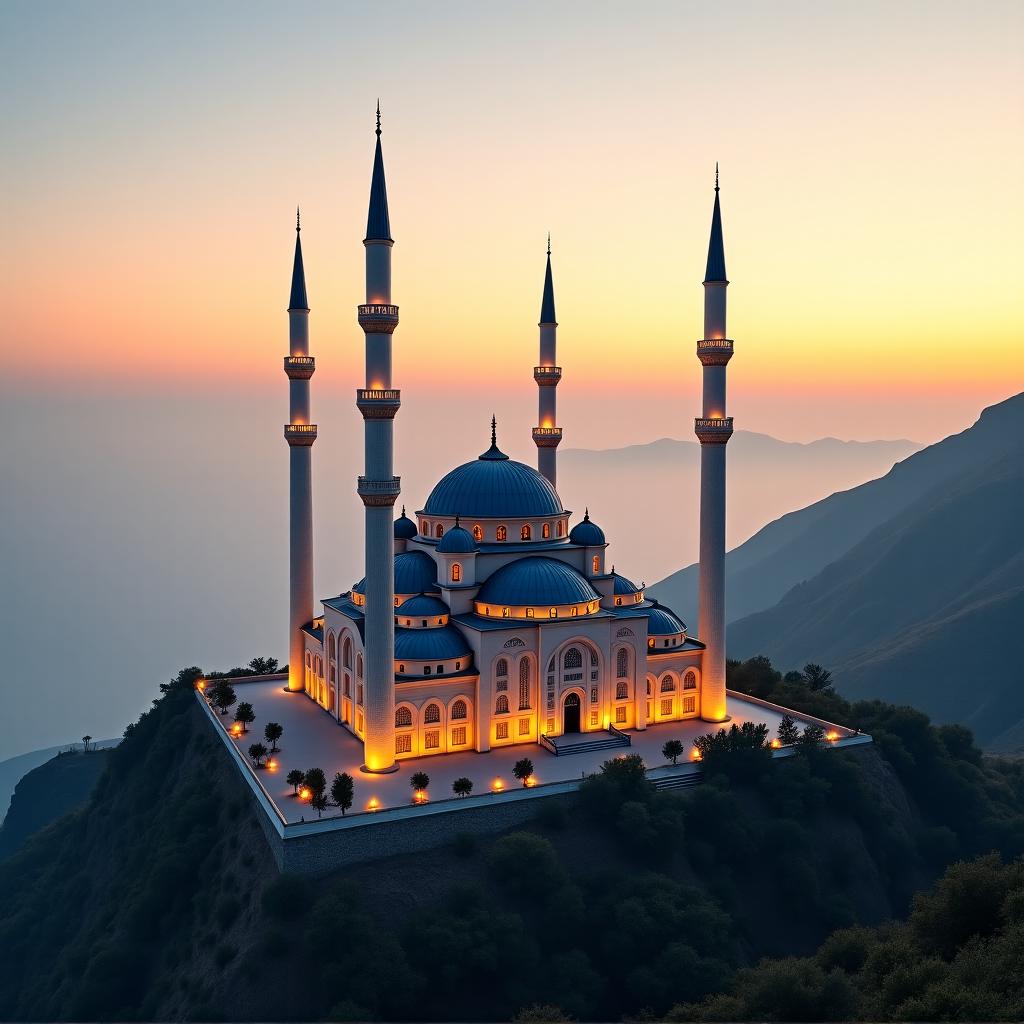}}%
        \fbox{\includegraphics[width=\mjhqfluximgwidth]{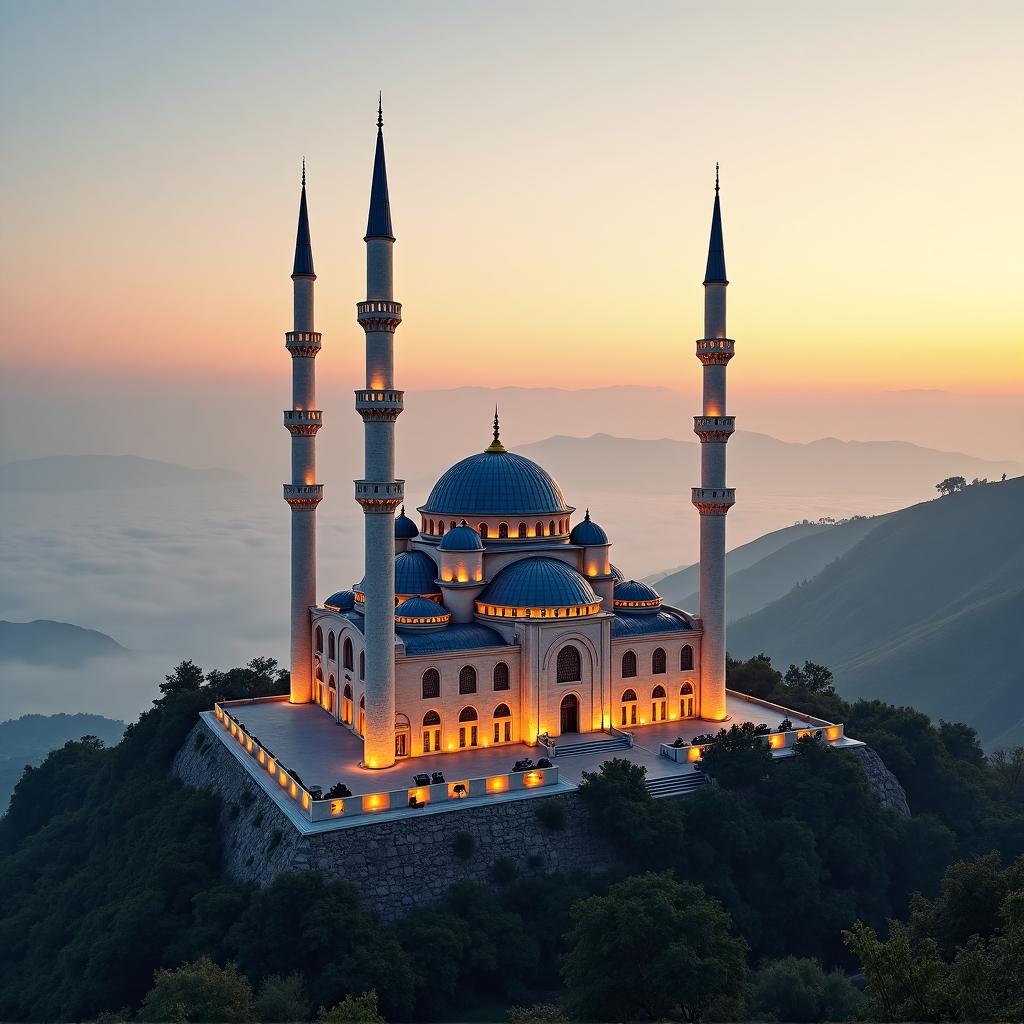}}%
        \fbox{\includegraphics[width=\mjhqfluximgwidth]{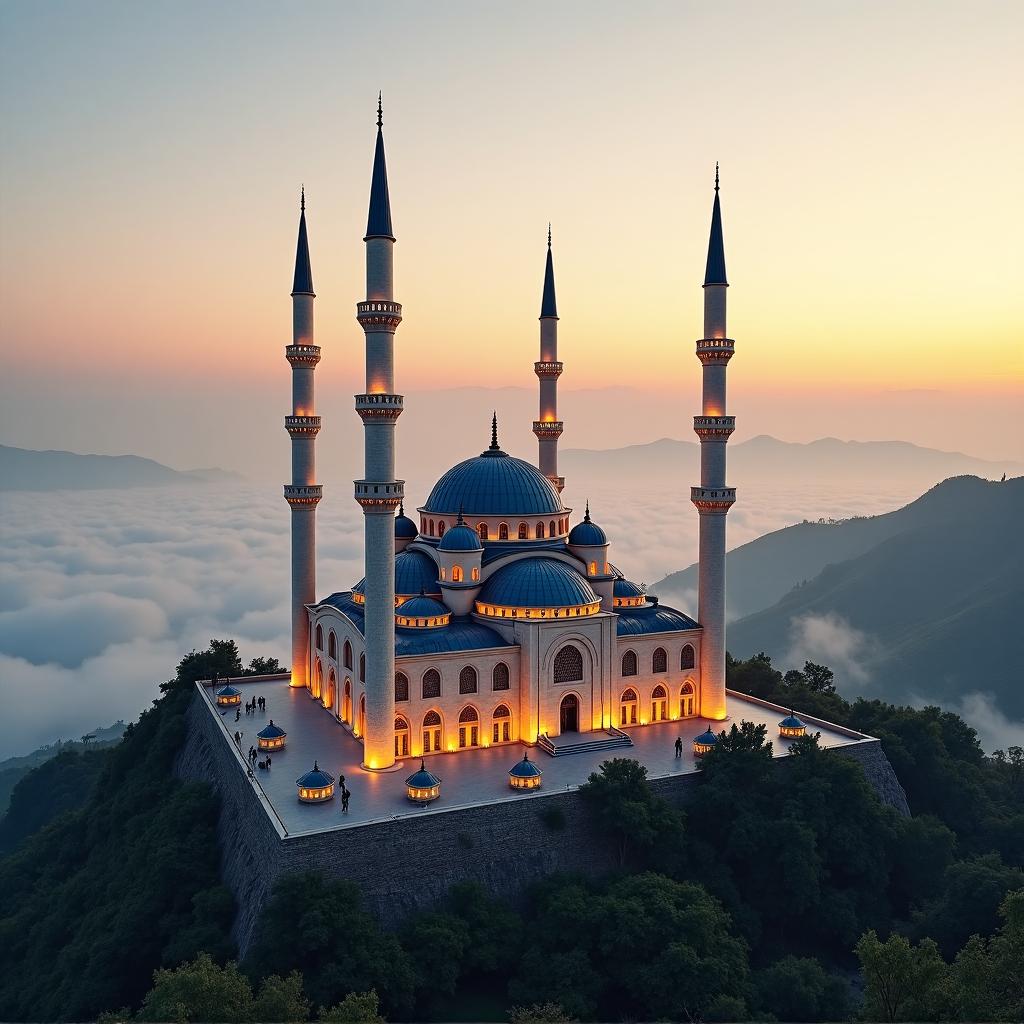}}\\[0.5ex]
        \hfill
    \end{minipage}
    \hfill
    \begin{minipage}[t]{0.395\textwidth}
        \centering
        \hfill
        \fbox{\includegraphics[width=\mjhqfluxrightimgwidth]{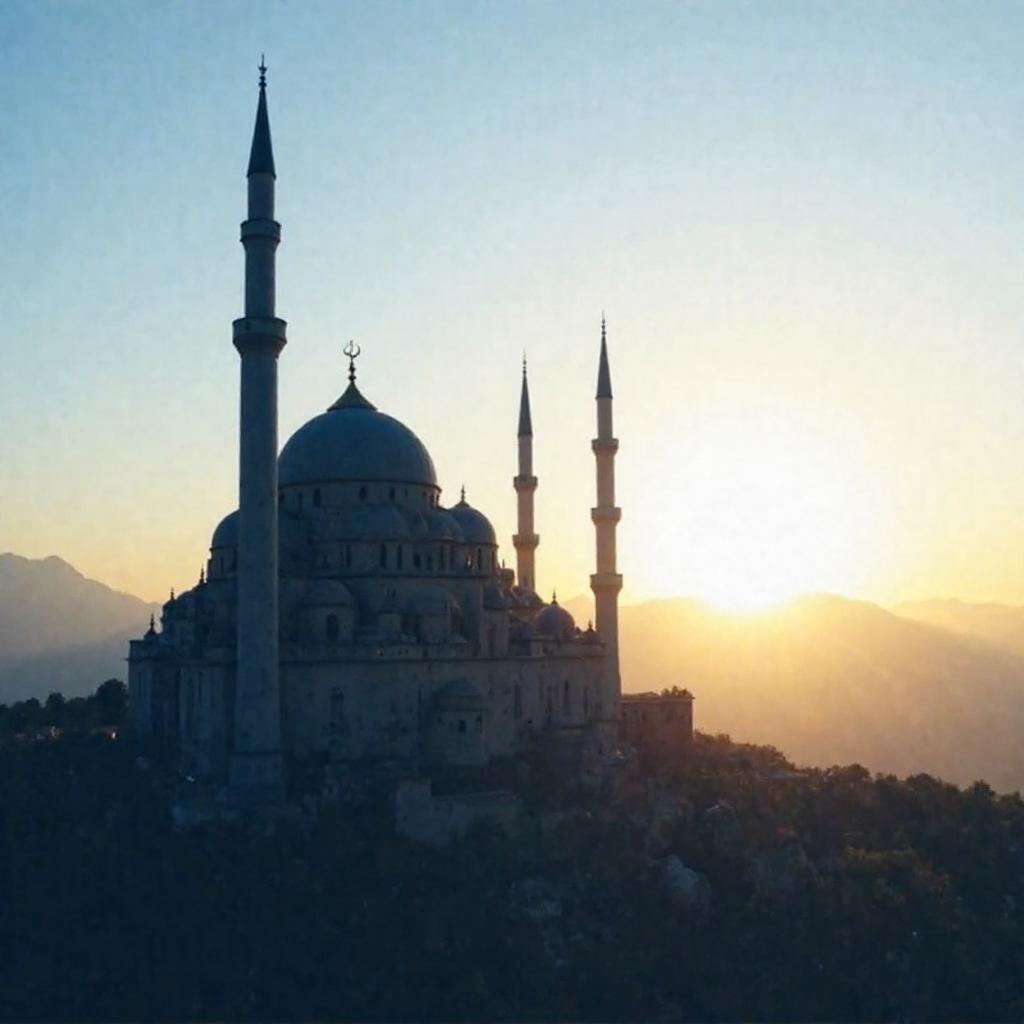}}%
        \fbox{\includegraphics[width=\mjhqfluxrightimgwidth]{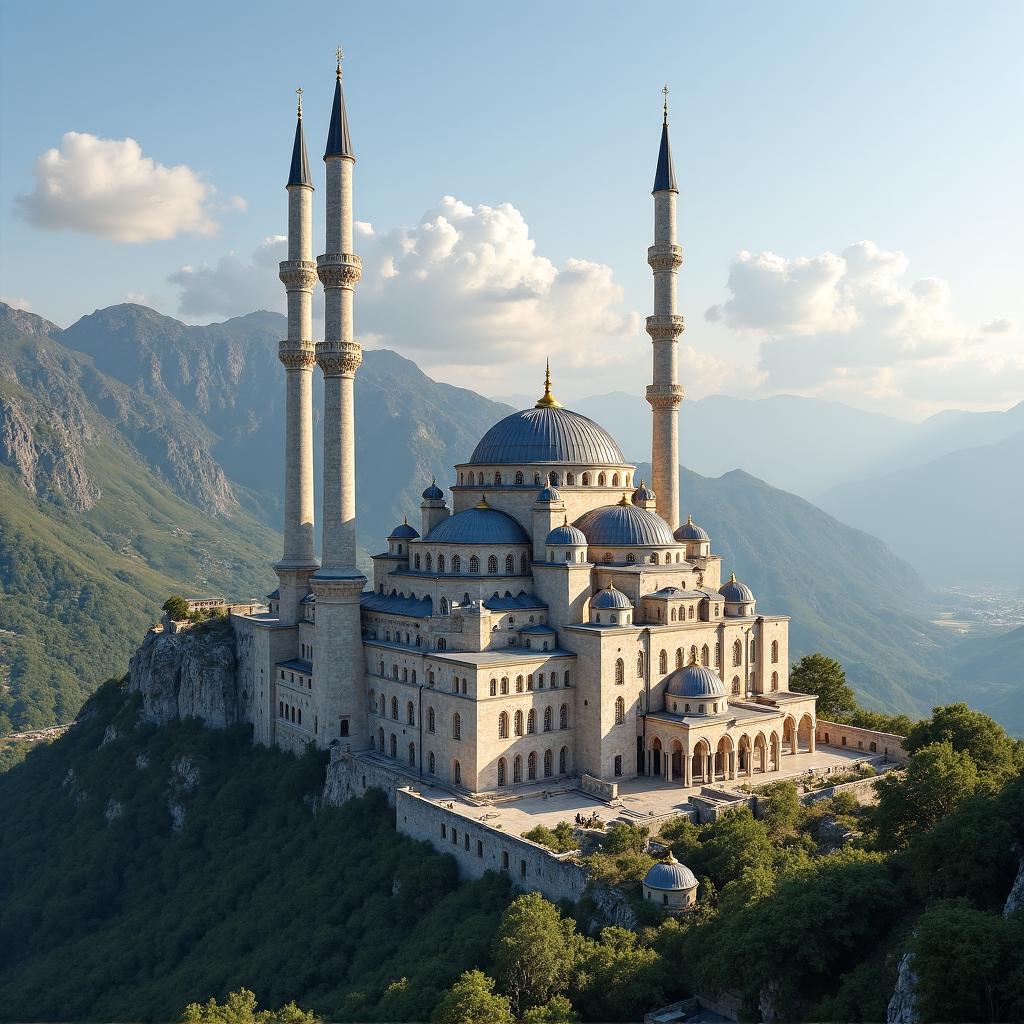}}%
        \fbox{\includegraphics[width=\mjhqfluxrightimgwidth]{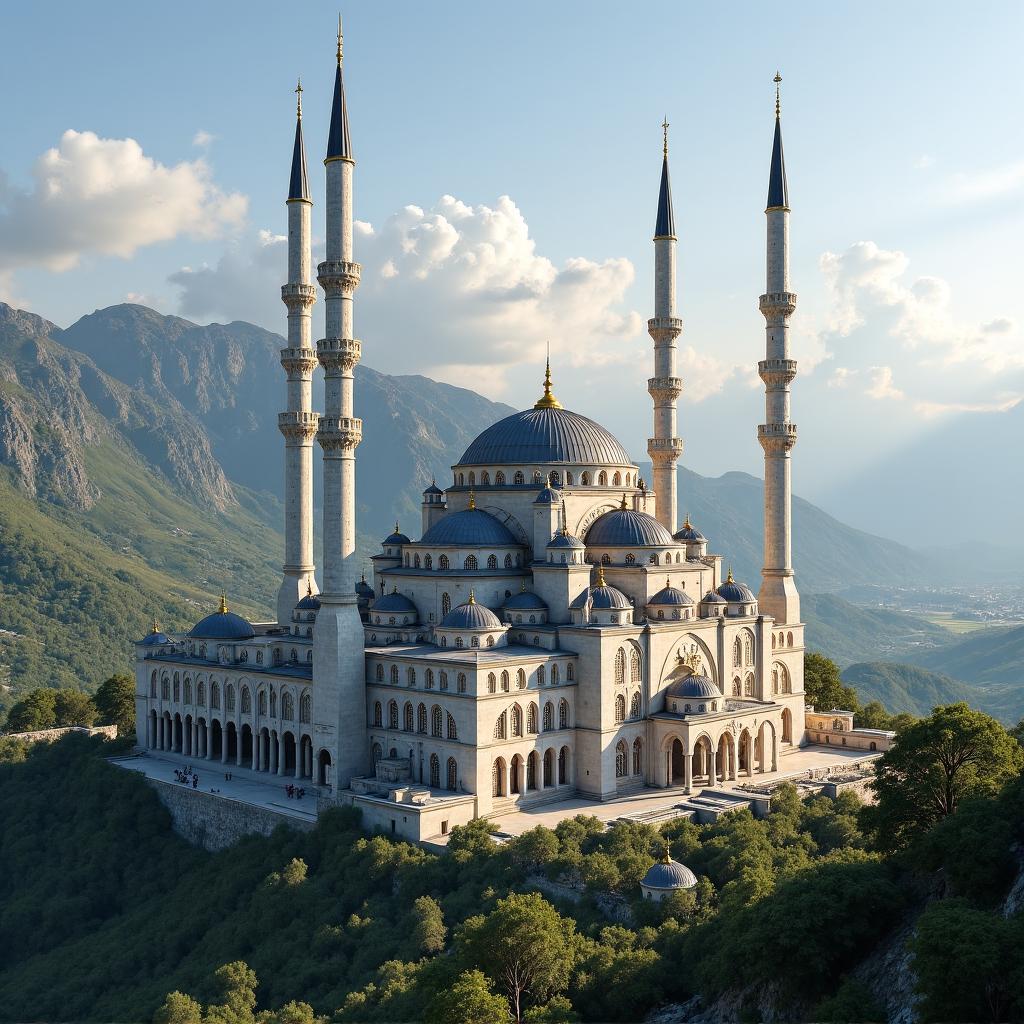}}%
        \hfill
        \fbox{\includegraphics[width=\mjhqfluxrightimgwidth]{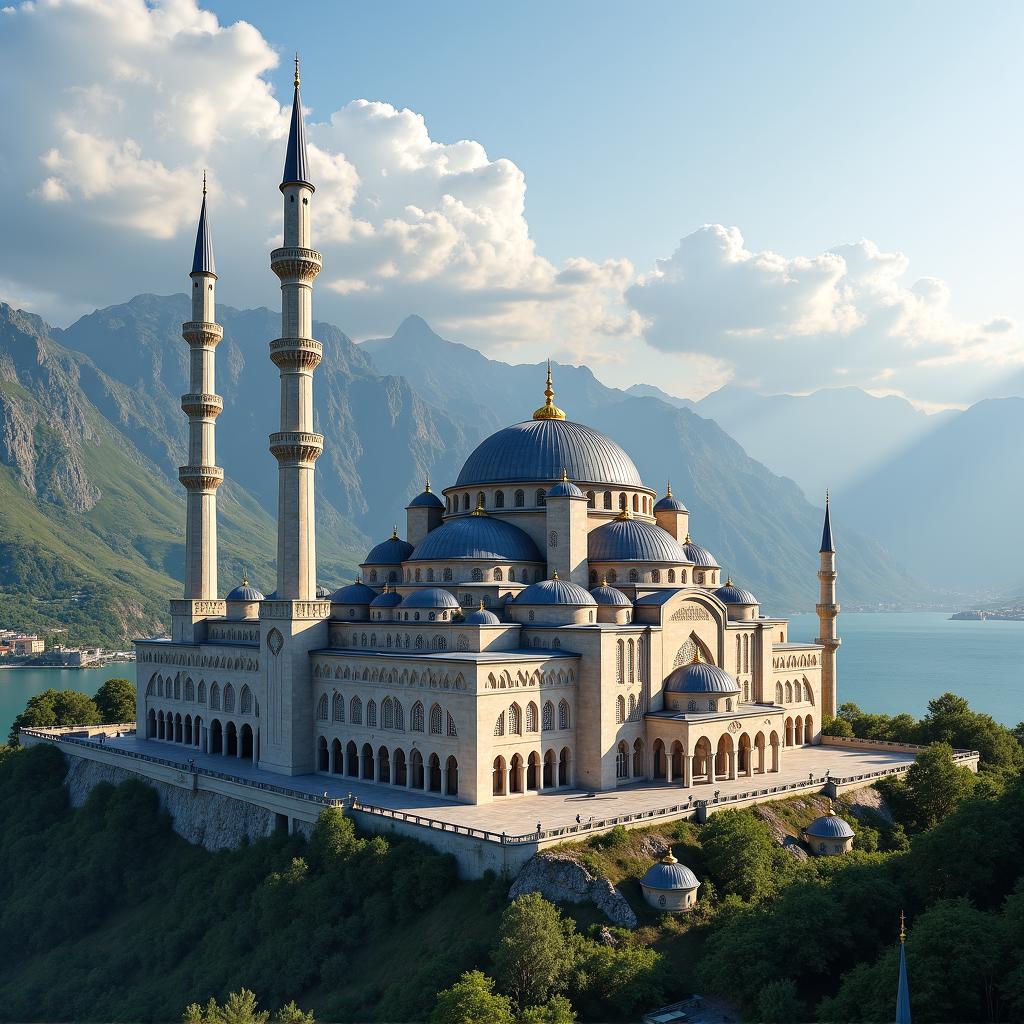}}\\[0.5ex]
    \end{minipage}
    \vspace{-15pt}
    \caption*{
        \begin{minipage}{\mjhqfluxcapwidth}
        \centering
            \footnotesize{Prompt: \textit{majestic islamic mosque in the mountain landscape beautiful.}}
        \end{minipage}
    }
    
    \caption{
        Additional visualizations of the main results on the MJHQ dataset, comparing FLUX and W8A8 DiT (left) with W4A8 (right) quantization.
    }
    \label{fig:app-mjhq-flux}
\end{figure}

\newcommand{\dcifluximgwidth}{0.123\textwidth}
\newcommand{\dcifluxcapwidth}{\textwidth}
\setlength{\fboxsep}{0pt}
\setlength{\fboxrule}{0.2pt}
\begin{figure}[h!]
    \centering
    \begin{minipage}[t]{\textwidth}
        \centering
        \begin{minipage}[t]{\dcifluximgwidth}
            \centering \scriptsize{FP16}
        \end{minipage}
        \hfill
        \begin{minipage}[t]{\dcifluximgwidth}
            \centering \scriptsize{Q-Diffusion(8/8)}
        \end{minipage}%
        \begin{minipage}[t]{\dcifluximgwidth}
            \centering \scriptsize{PTQ4DiT(8/8)}
        \end{minipage}%
        \begin{minipage}[t]{\dcifluximgwidth}
            \centering \scriptsize{\textbf{SegQuant-A(8/8)}}
        \end{minipage}%
        \begin{minipage}[t]{\dcifluximgwidth}
            \centering \scriptsize{\textbf{SegQuant-G(8/8)}}
        \end{minipage}%
        \begin{minipage}[t]{\dcifluximgwidth}
            \centering \scriptsize{Q-Diffusion(4/8)}
        \end{minipage}%
        \begin{minipage}[t]{\dcifluximgwidth}
            \centering \scriptsize{SVDQuant(4/8)}
        \end{minipage}%
        \begin{minipage}[t]{\dcifluximgwidth}
            \centering \scriptsize{\textbf{SegQuant-G(4/8)}}
        \end{minipage}
    \end{minipage}

    \vspace{0.6ex}

    \begin{minipage}[t]{\textwidth}
        \centering
        \fbox{\includegraphics[width=\dcifluximgwidth]{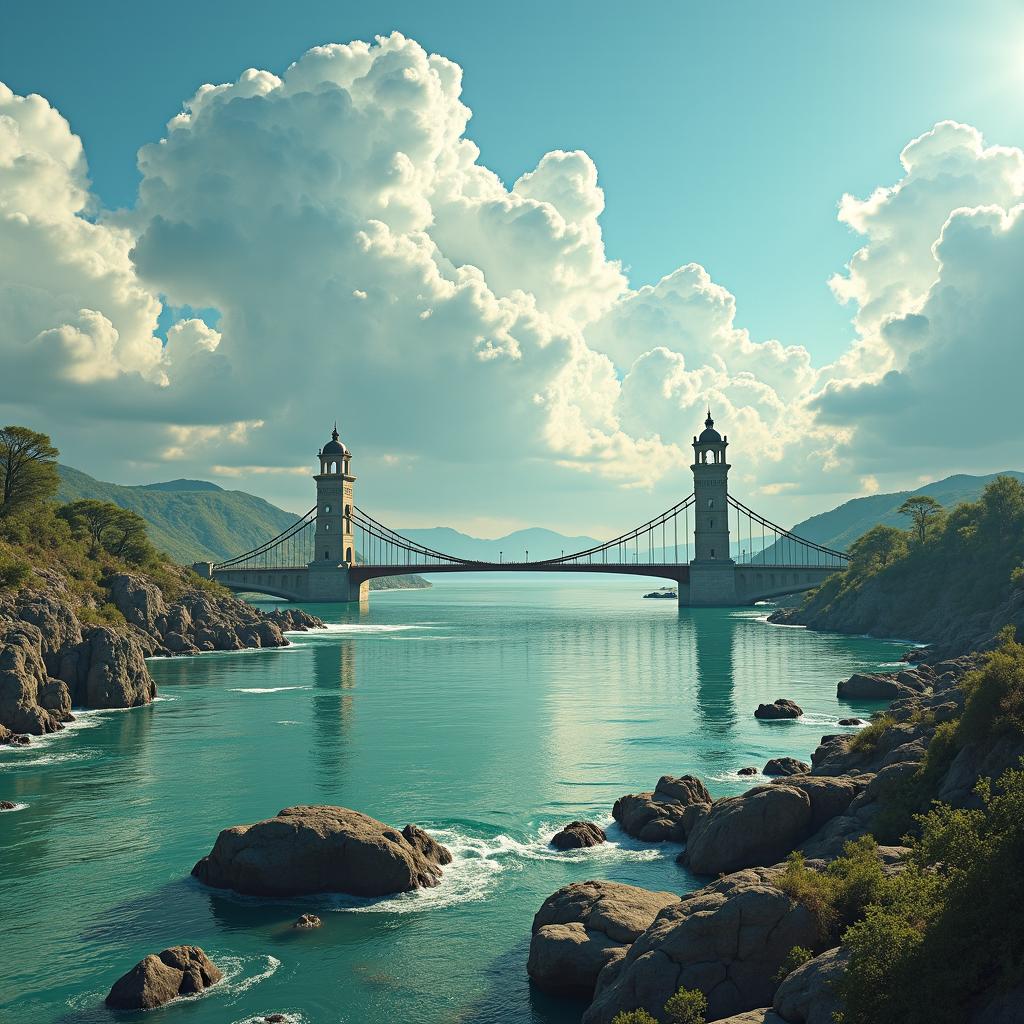}}
        \hfill
        \fbox{\includegraphics[width=\dcifluximgwidth]{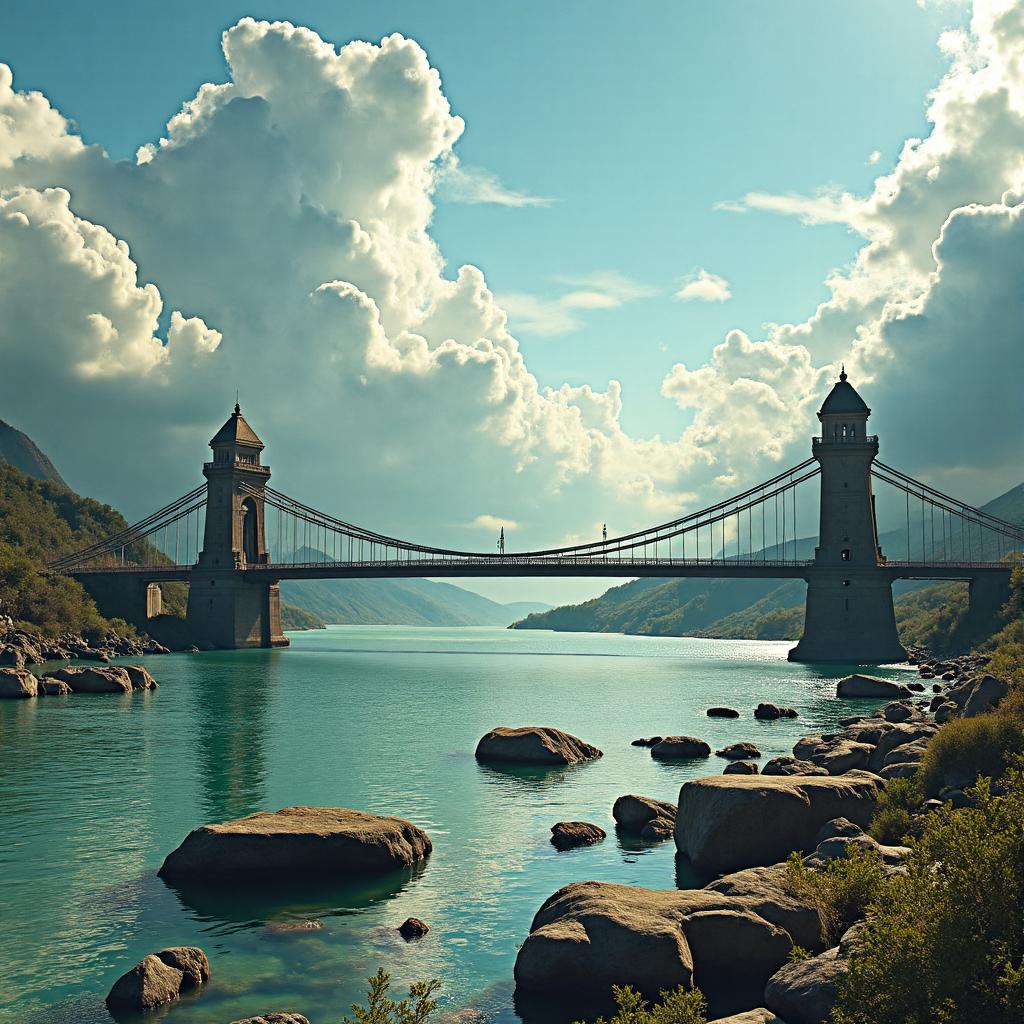}}%
        \fbox{\includegraphics[width=\dcifluximgwidth]{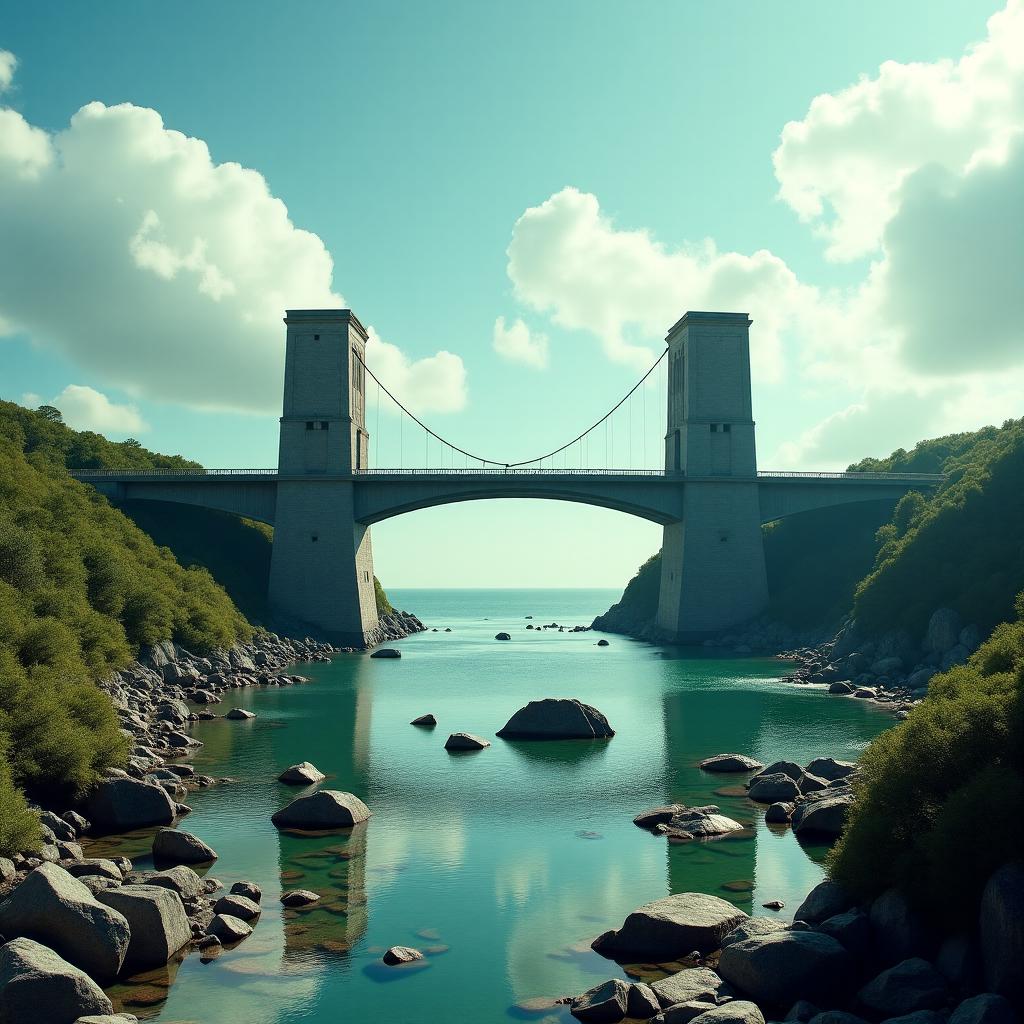}}%
        \fbox{\includegraphics[width=\dcifluximgwidth]{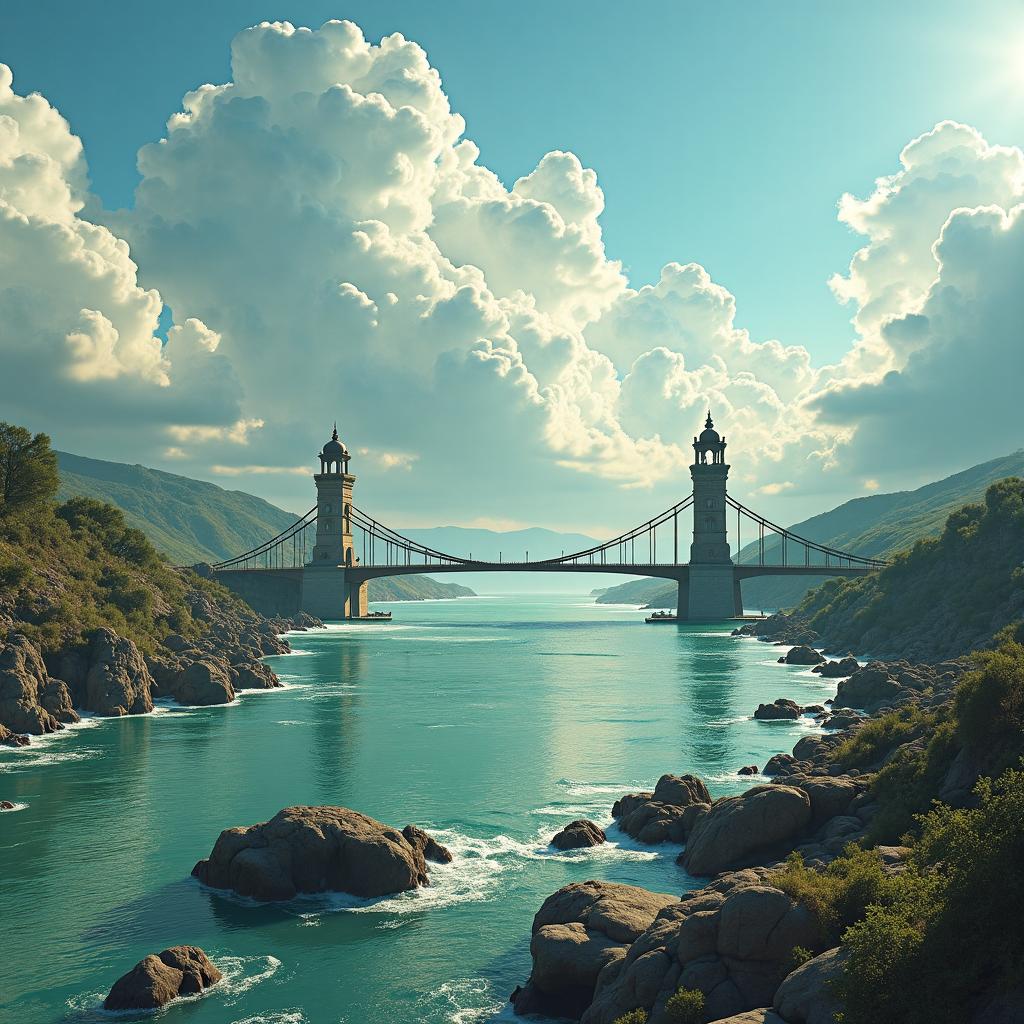}}%
        \fbox{\includegraphics[width=\dcifluximgwidth]{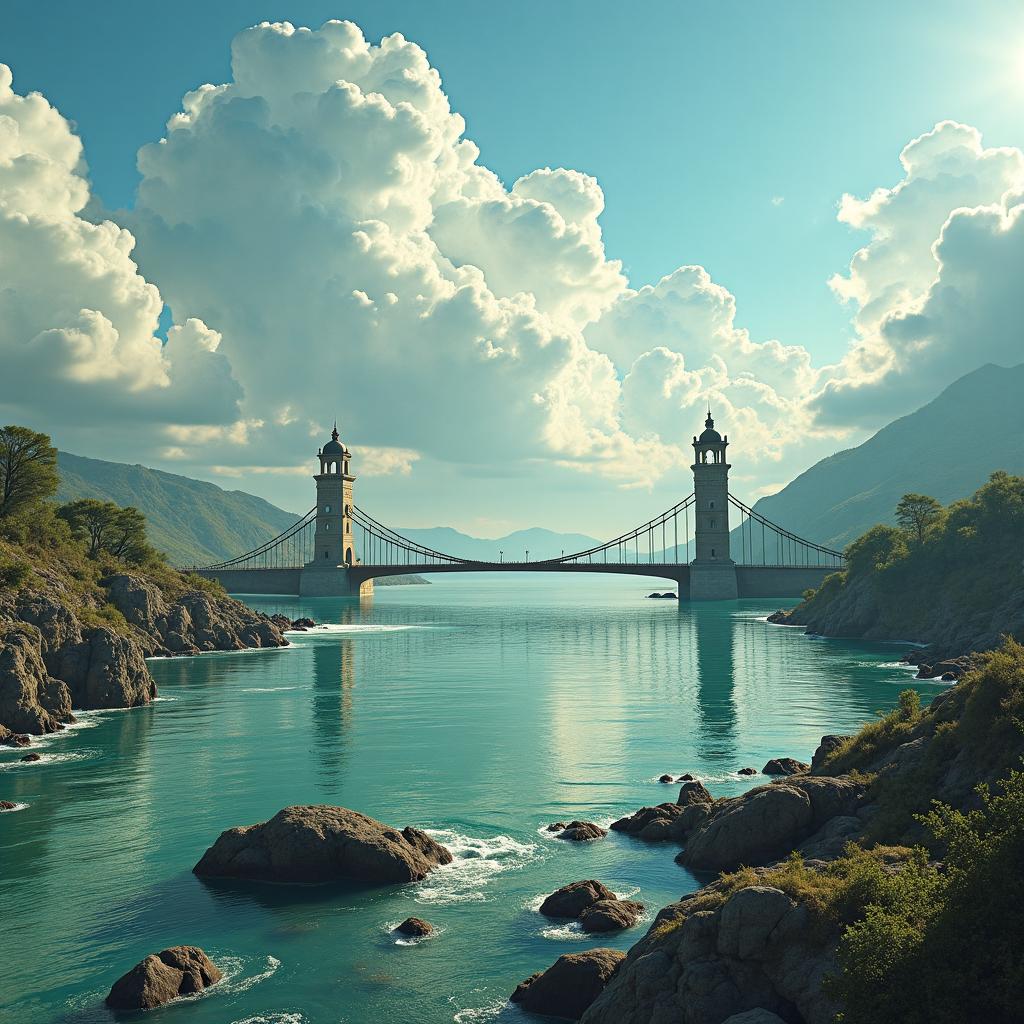}}%
        \fbox{\includegraphics[width=\dcifluximgwidth]{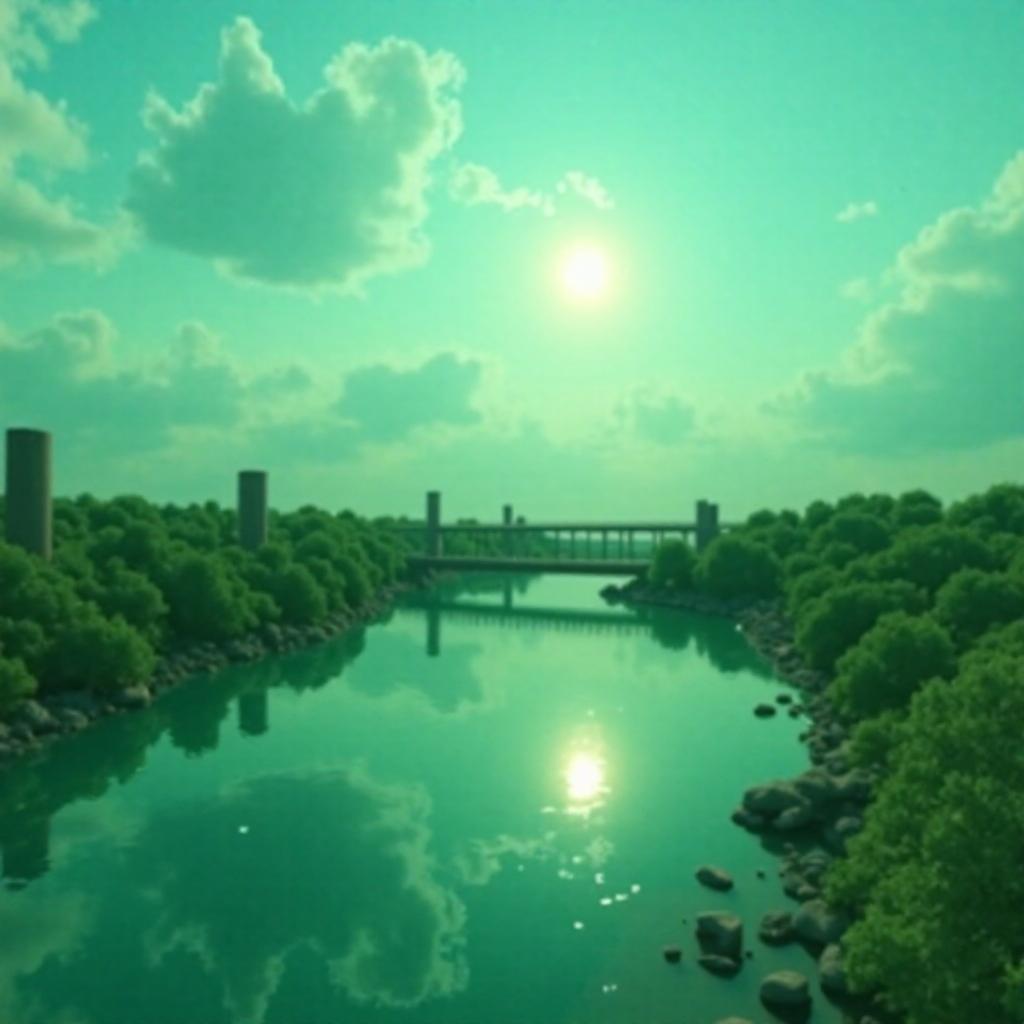}}%
        \fbox{\includegraphics[width=\dcifluximgwidth]{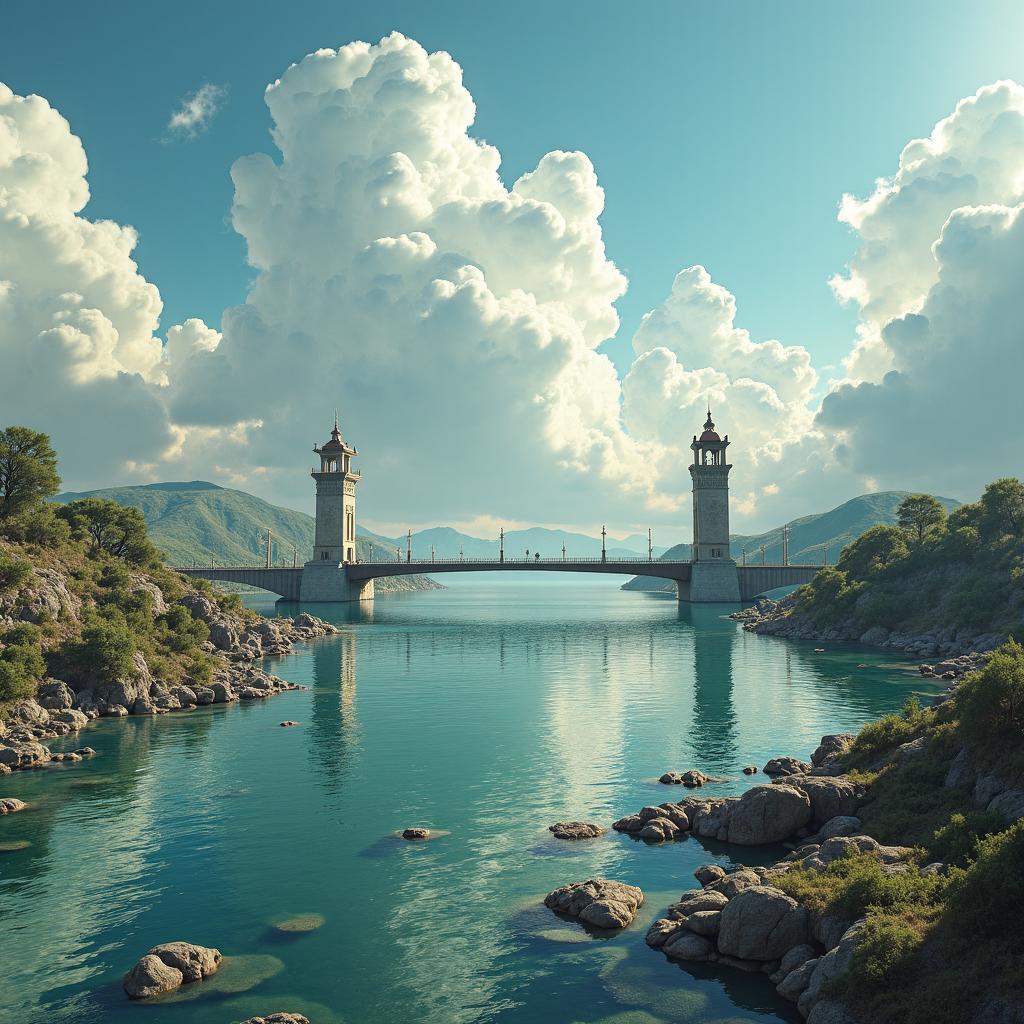}}%
        \fbox{\includegraphics[width=\dcifluximgwidth]{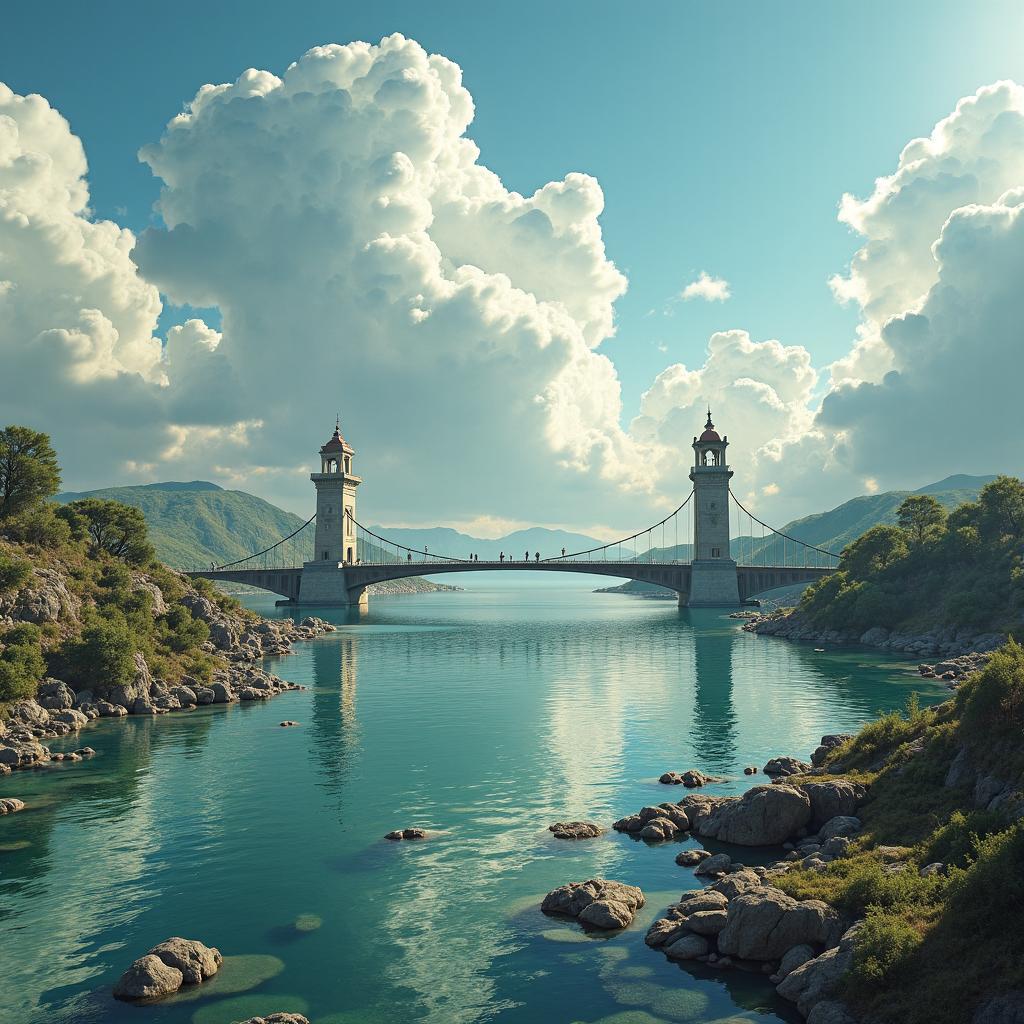}}\\[0.5ex]
        \hfill
    \end{minipage}
    \vspace{-18pt}
    \caption*{
        \begin{minipage}{\dcifluxcapwidth}
        \centering
            \tiny{Prompt: \textit{A body of water it is brown and green the sky is bright and there are many clouds in it A bright day the sky is very blue also there are clouds here that are very thick and very full also in the center of the sky is the biggest clouds. In the center of the image there is a large body of water that is green and blue. Going across the body of water is a large gray bridge with four towers on it two on each side and connecting each set of towers is a large beam that is a walkway. The  walkway has windows that allow looking out over the bay. There is a large bed of rocks that is close to the camera of different colors dark gray brown white and light gray also there is also a yellow water bottle that has ben deposited amongst the rocks.}}
        \end{minipage}
    }

    \vspace{0.4cm}
    \begin{minipage}[t]{\textwidth}
        \centering
        \fbox{\includegraphics[width=\dcifluximgwidth]{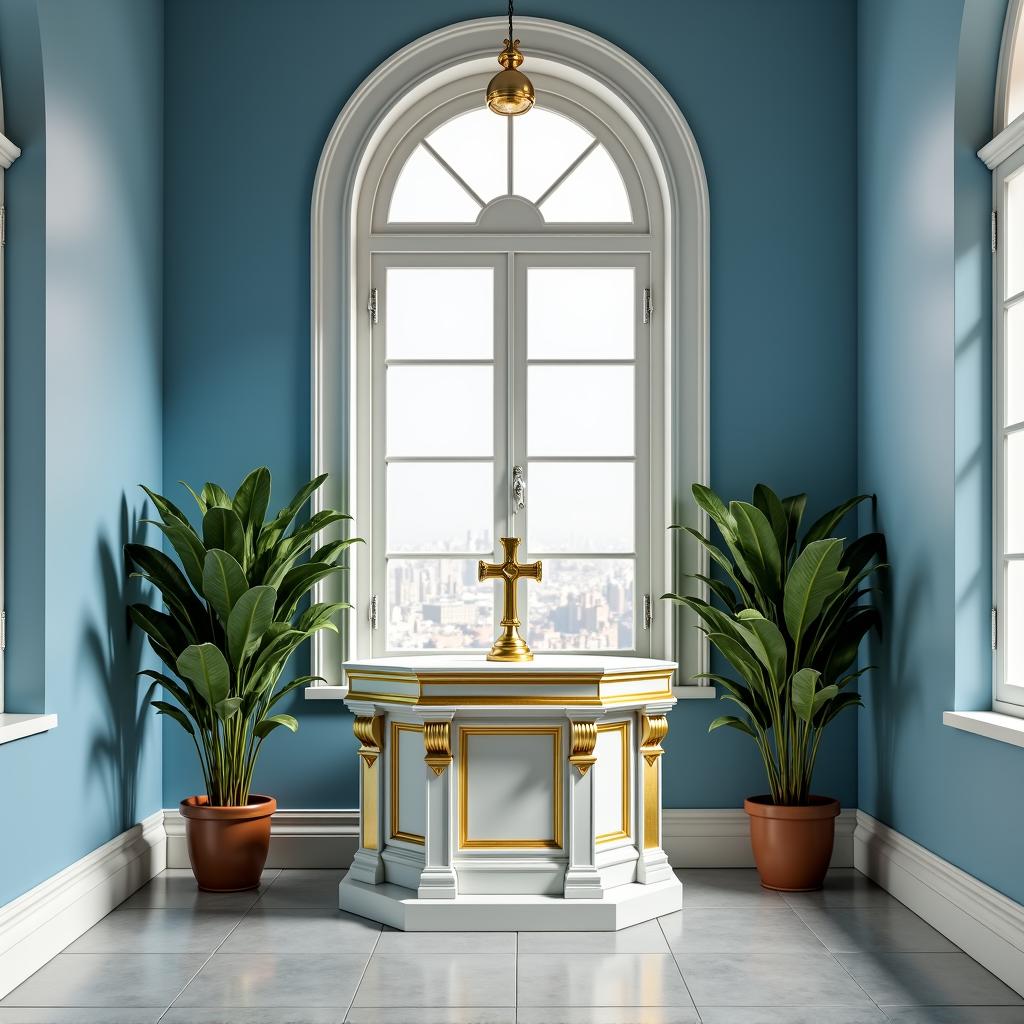}}
        \hfill
        \fbox{\includegraphics[width=\dcifluximgwidth]{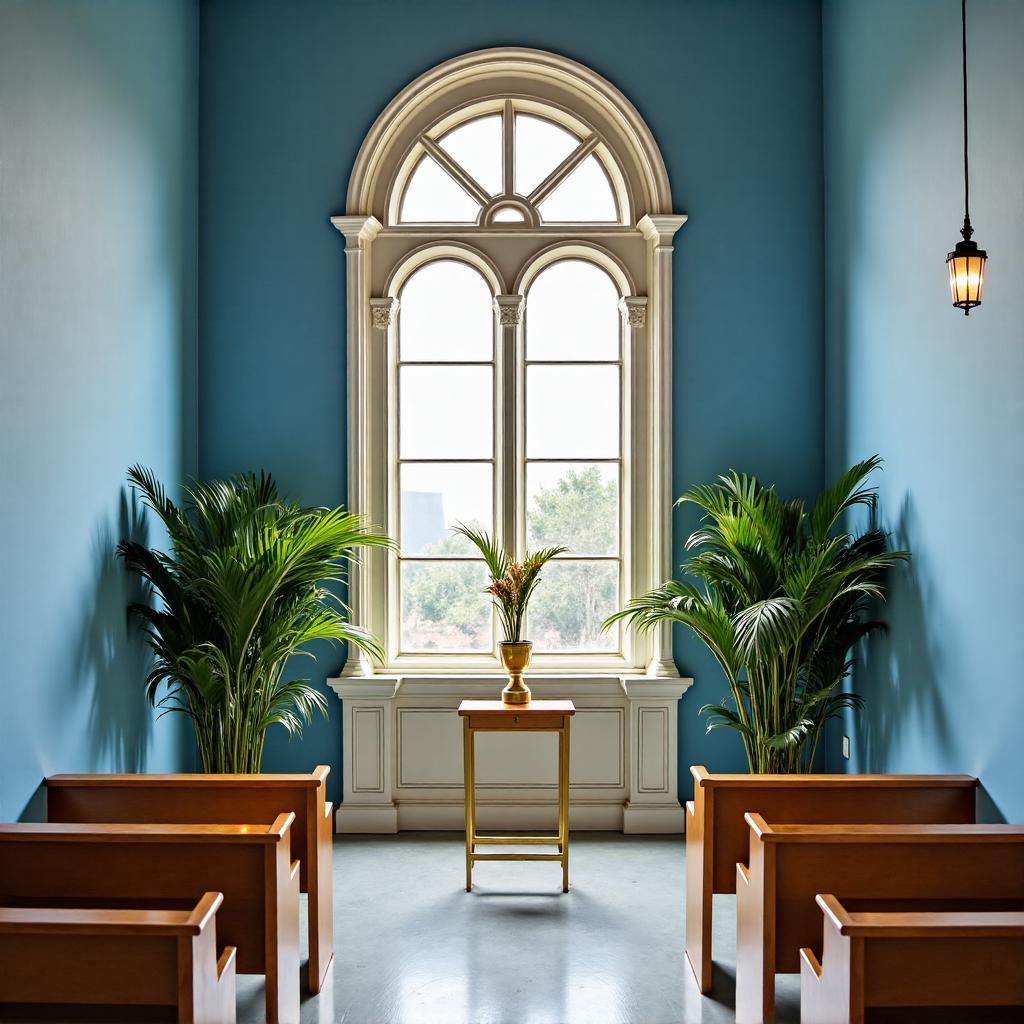}}%
        \fbox{\includegraphics[width=\dcifluximgwidth]{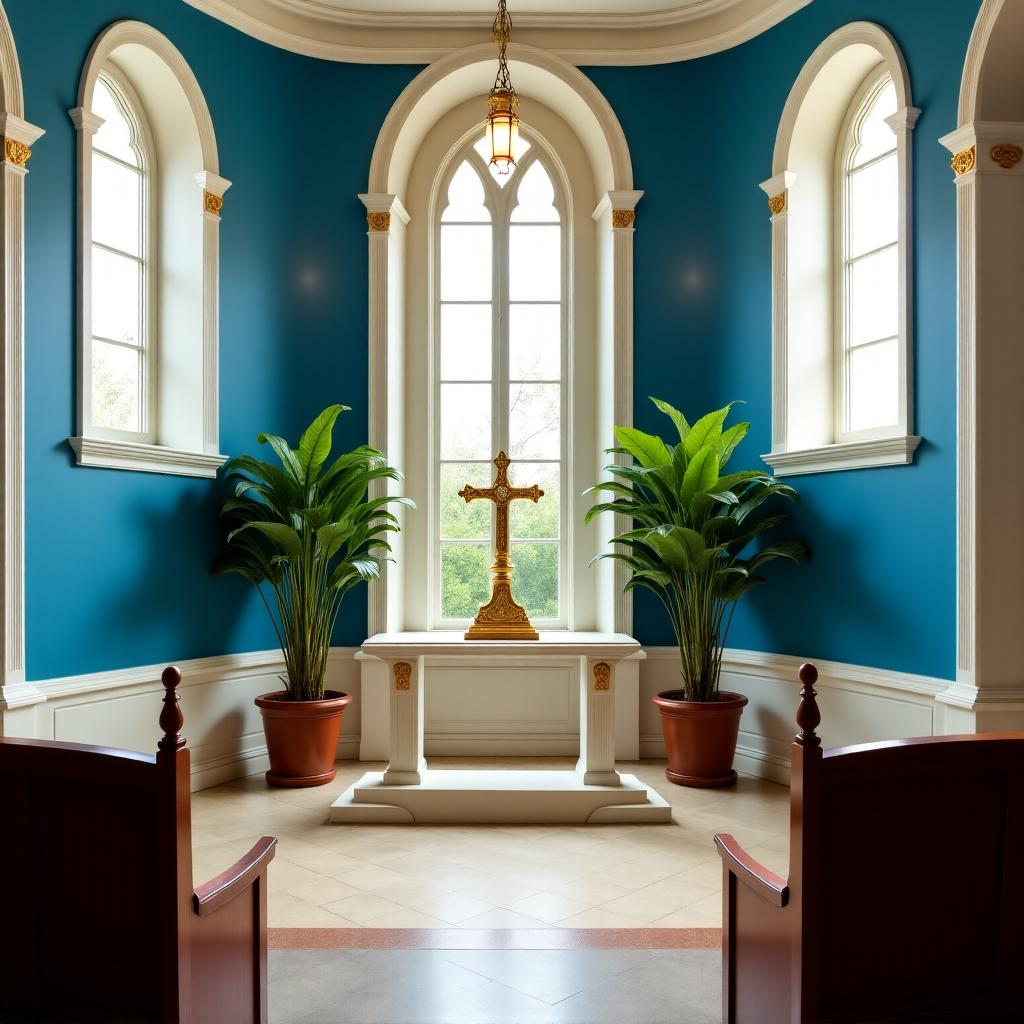}}%
        \fbox{\includegraphics[width=\dcifluximgwidth]{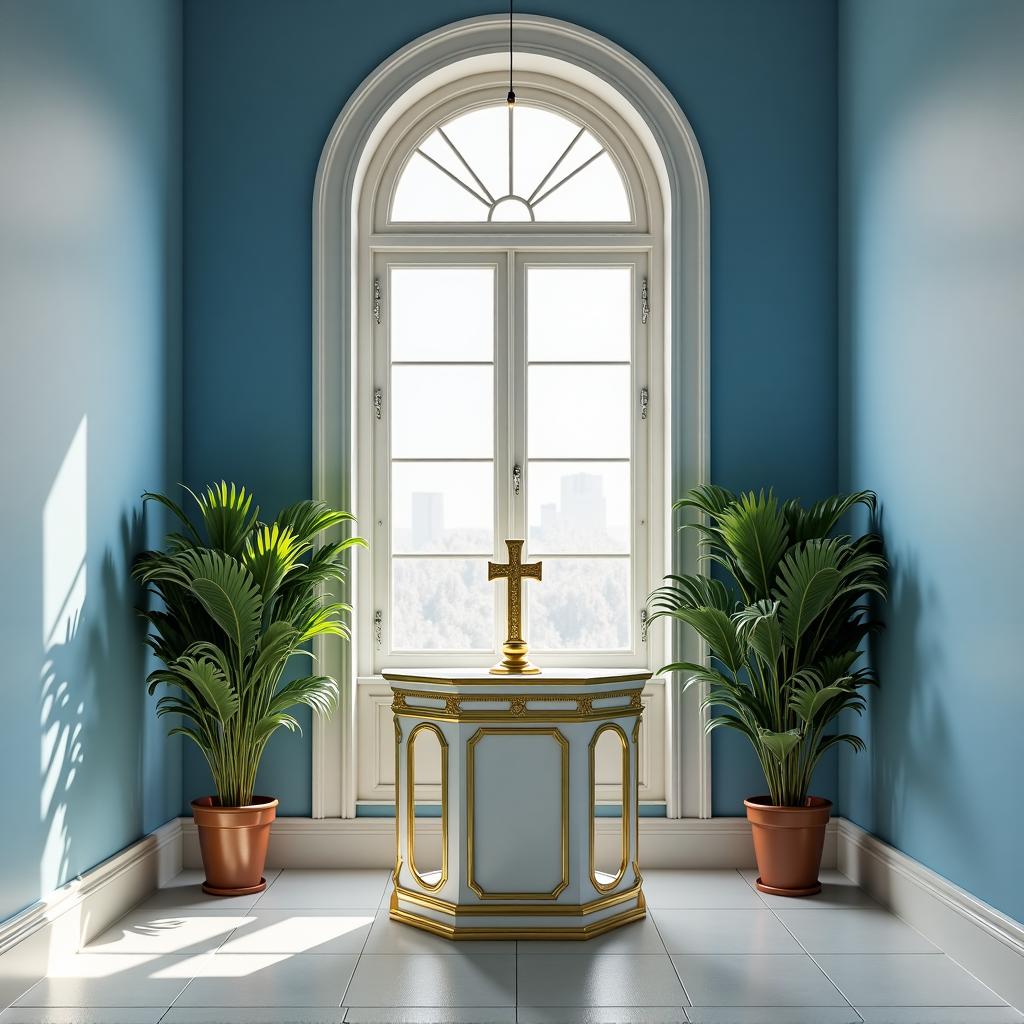}}%
        \fbox{\includegraphics[width=\dcifluximgwidth]{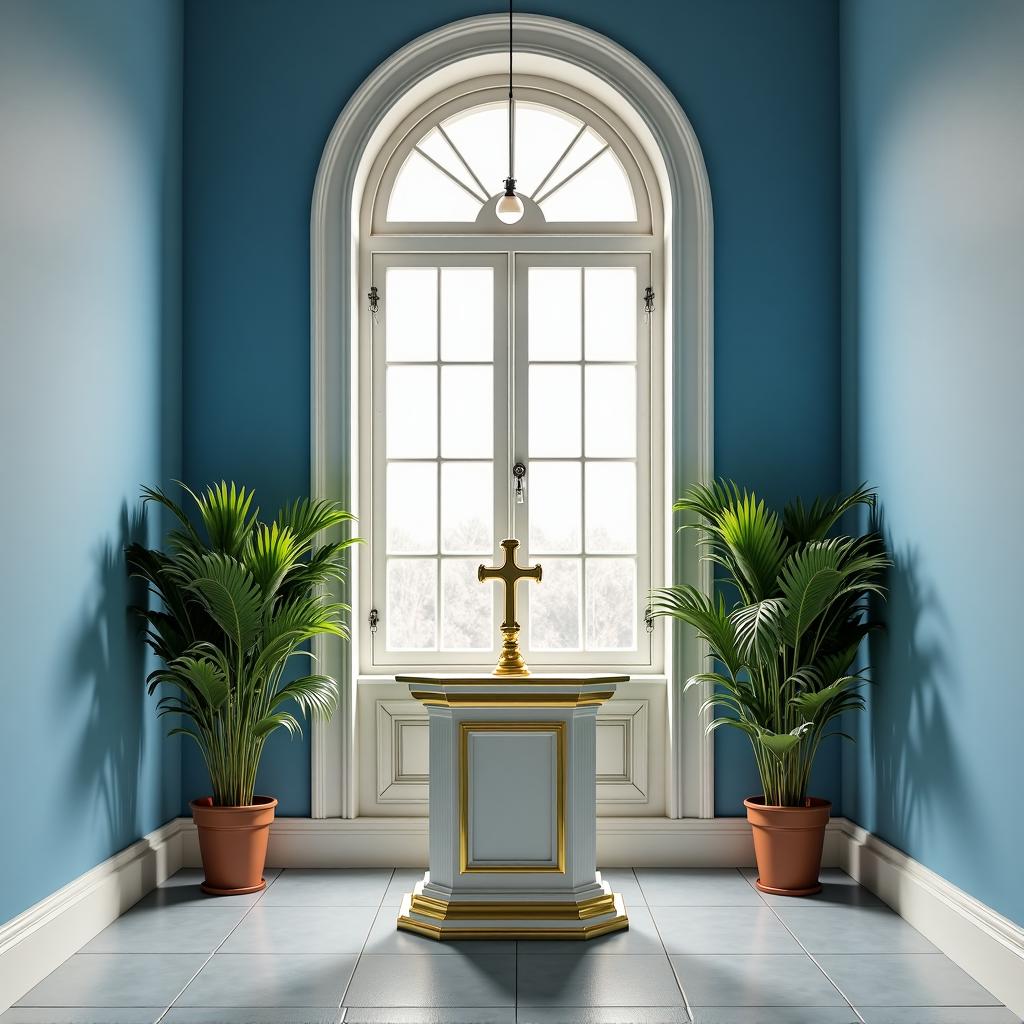}}%
        \fbox{\includegraphics[width=\dcifluximgwidth]{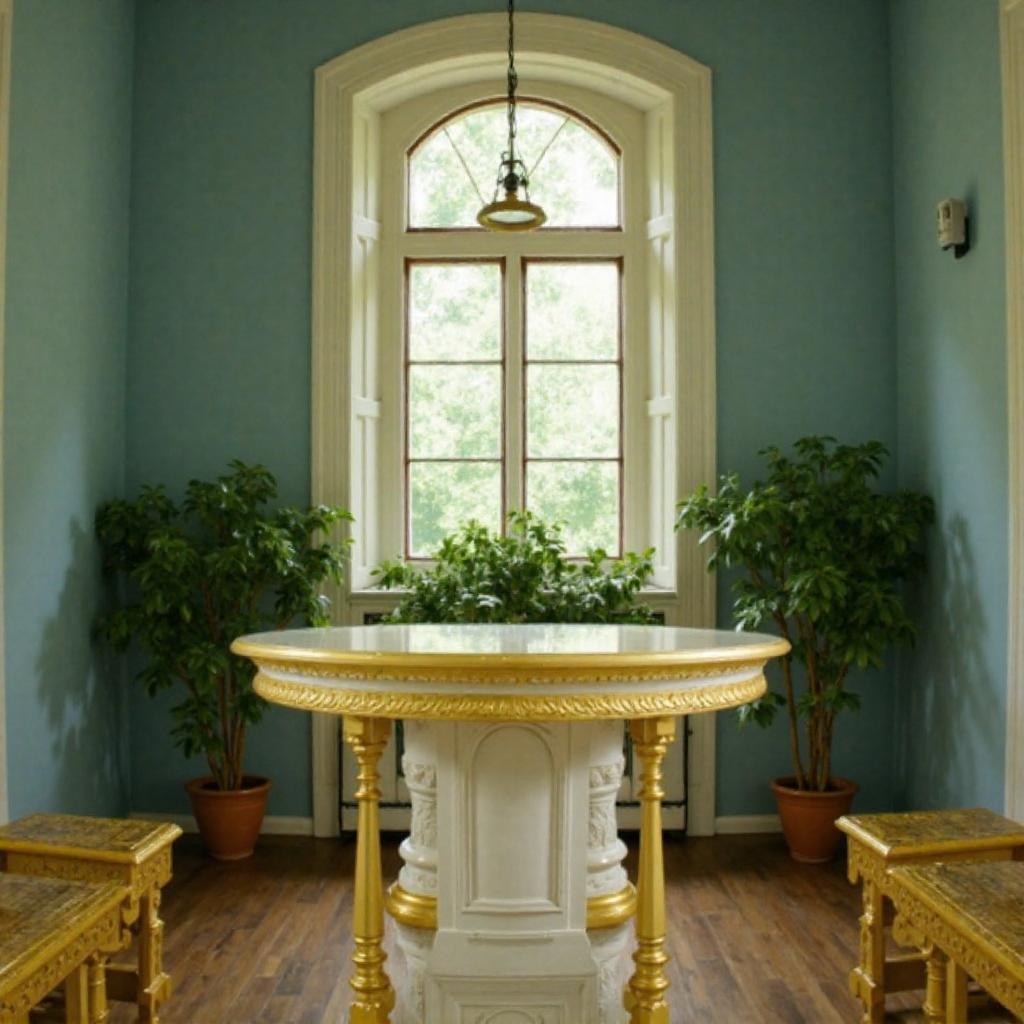}}%
        \fbox{\includegraphics[width=\dcifluximgwidth]{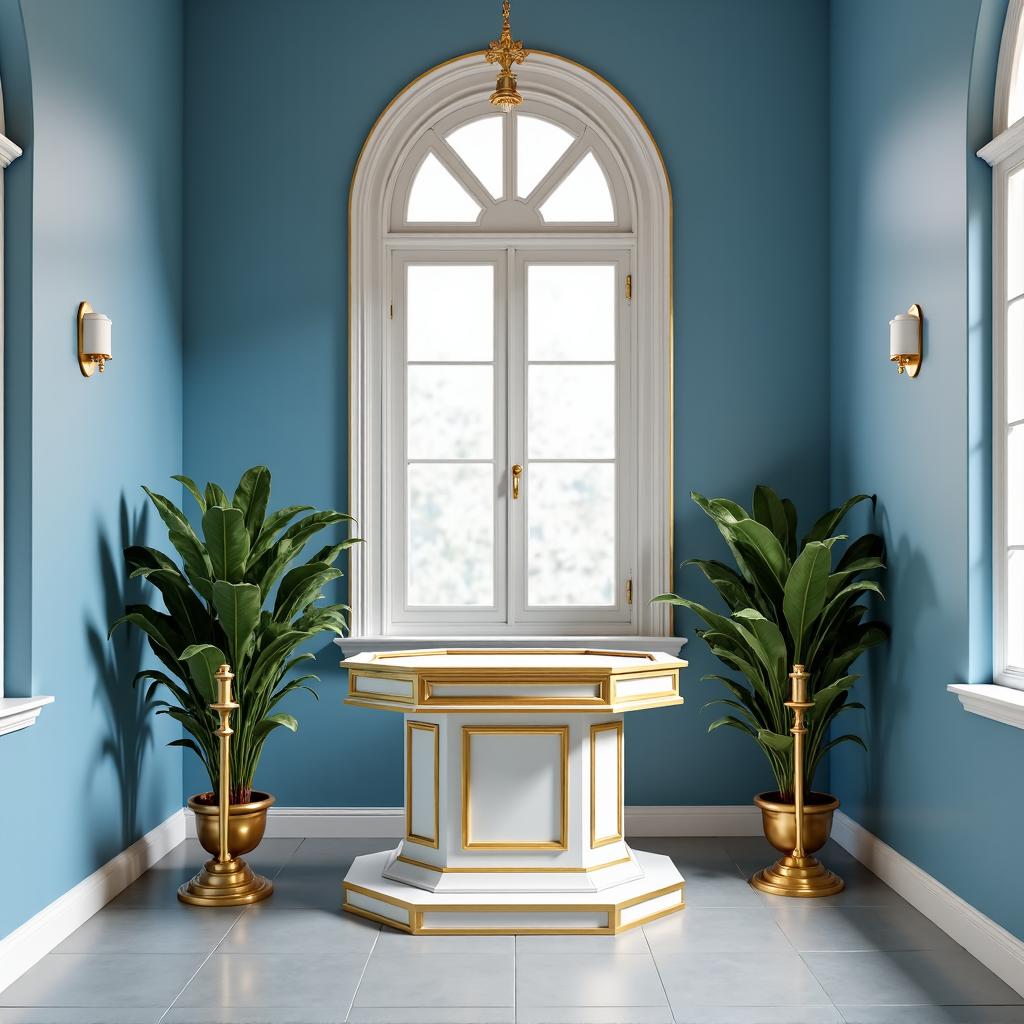}}%
        \fbox{\includegraphics[width=\dcifluximgwidth]{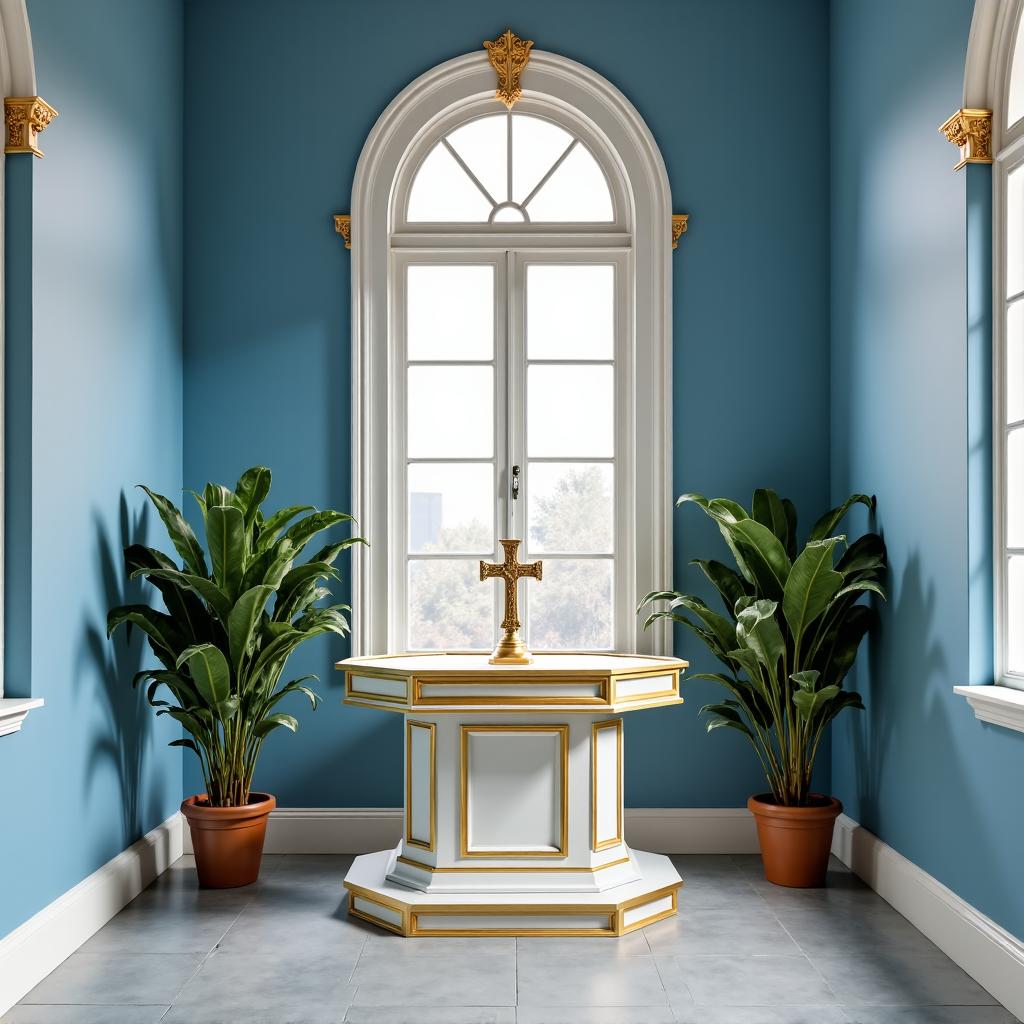}}\\[0.5ex]
        \hfill
    \end{minipage}
    \vspace{-18pt}
    \caption*{
        \begin{minipage}{\dcifluxcapwidth}
        \centering
            \tiny{Prompt: \textit{This is a small room with blue walls inside of a church with a white alter with elaborate gold trim in the center. Two gold stands are in the room near the front of the alter. There is a bay window in the back. There is a plant in front of each window. The top of the alter is round with gold trim and has a gold cross on the top. A light is hanging on the top right side of the room.}}
        \end{minipage}
    }

    \vspace{0.4cm}
    \begin{minipage}[t]{\textwidth}
        \centering
        \fbox{\includegraphics[width=\dcifluximgwidth]{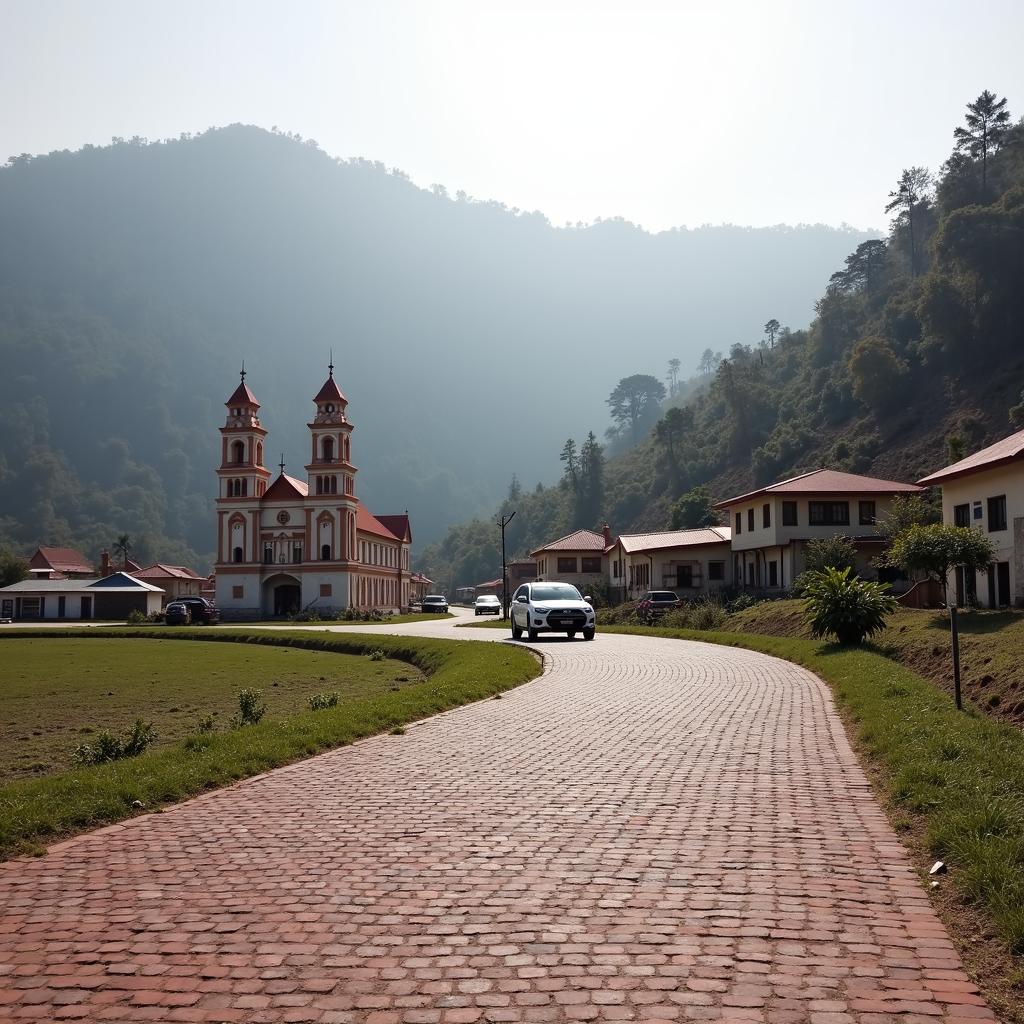}}
        \hfill
        \fbox{\includegraphics[width=\dcifluximgwidth]{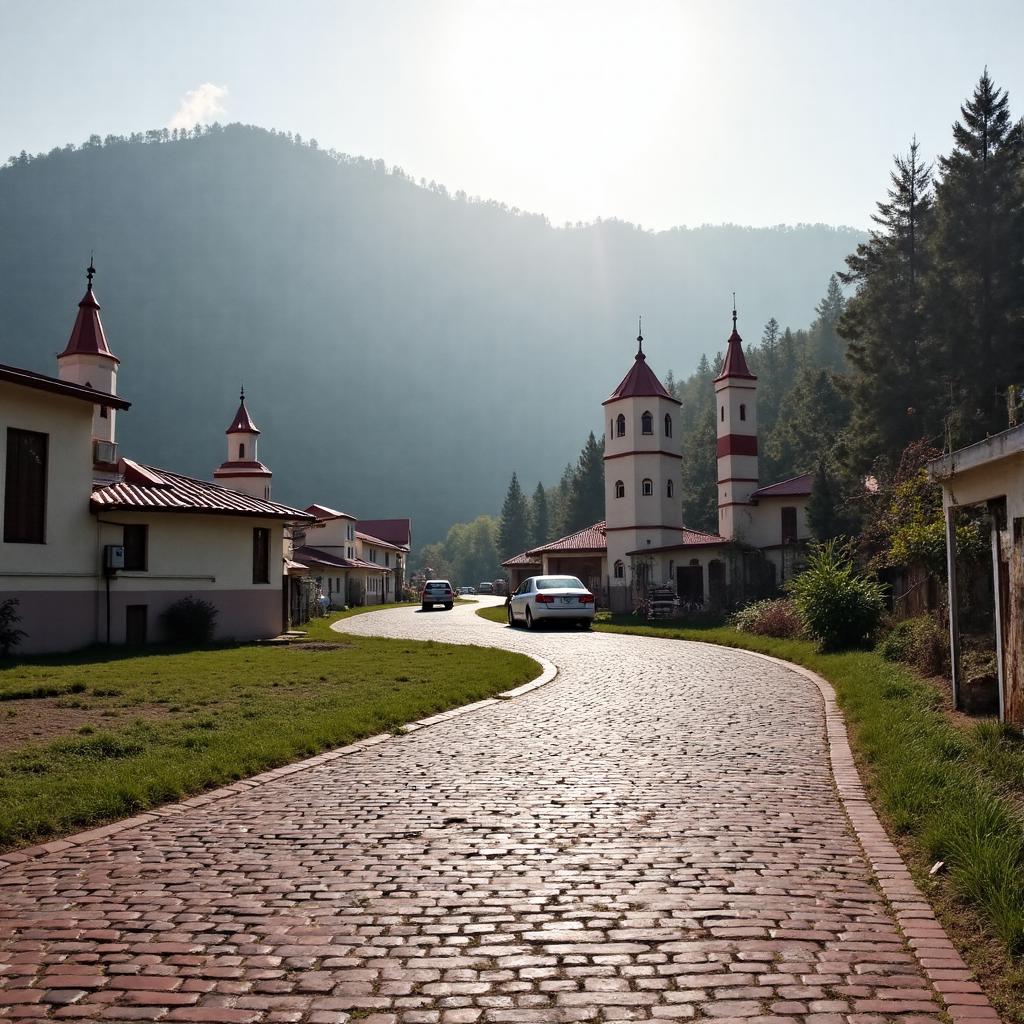}}%
        \fbox{\includegraphics[width=\dcifluximgwidth]{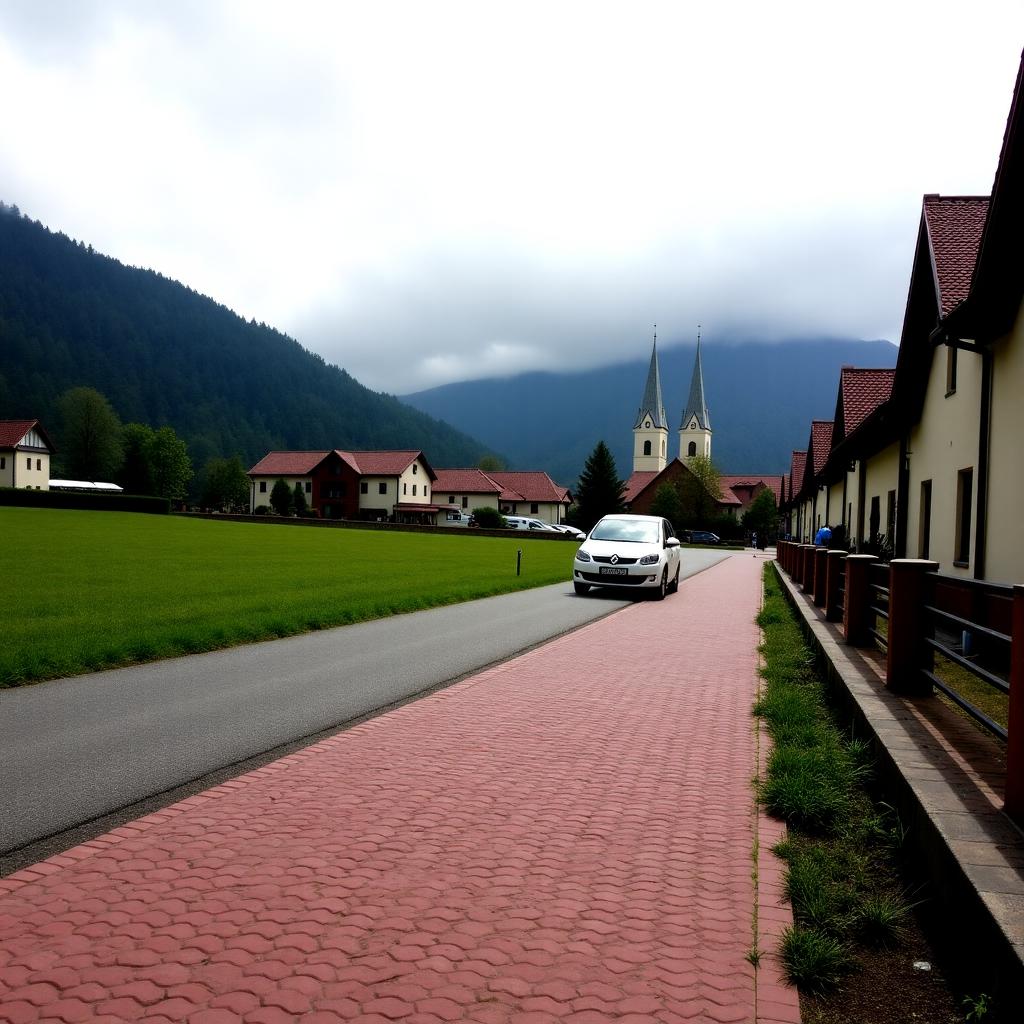}}%
        \fbox{\includegraphics[width=\dcifluximgwidth]{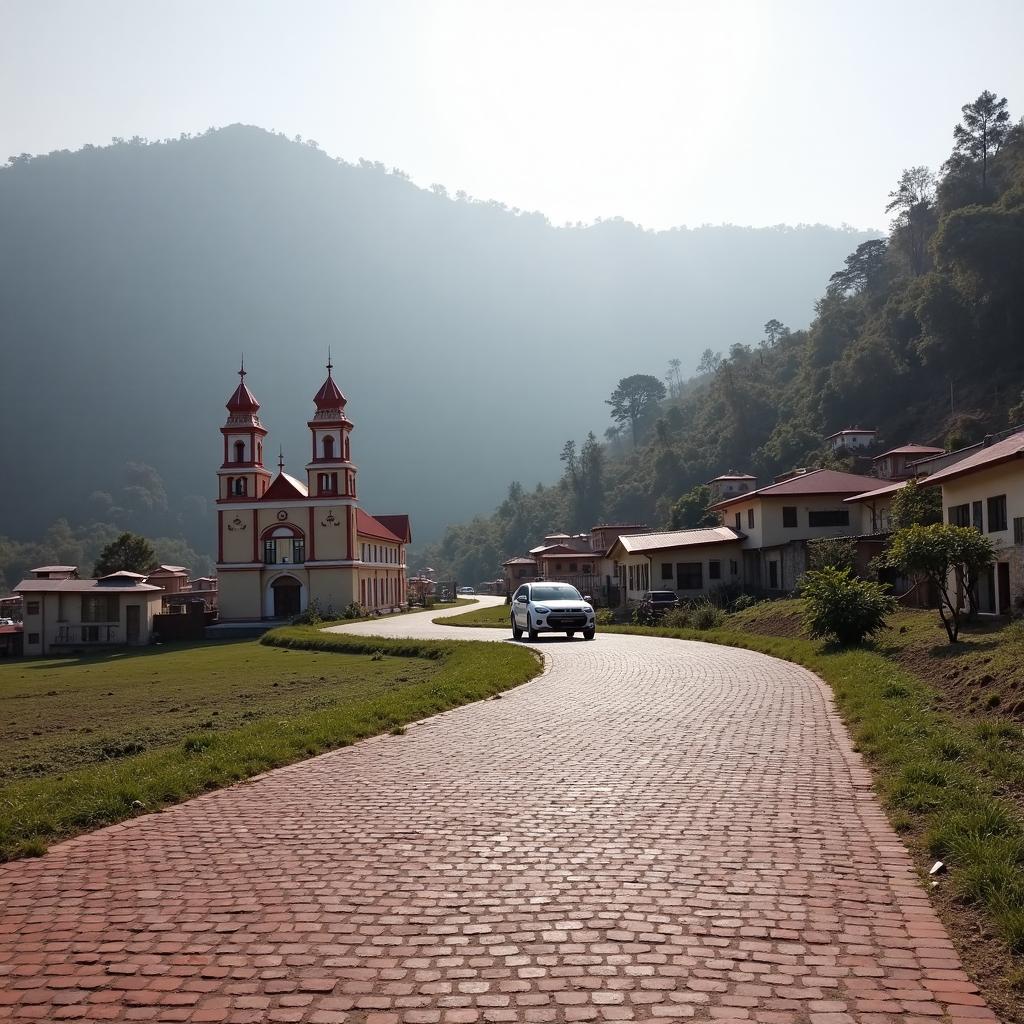}}%
        \fbox{\includegraphics[width=\dcifluximgwidth]{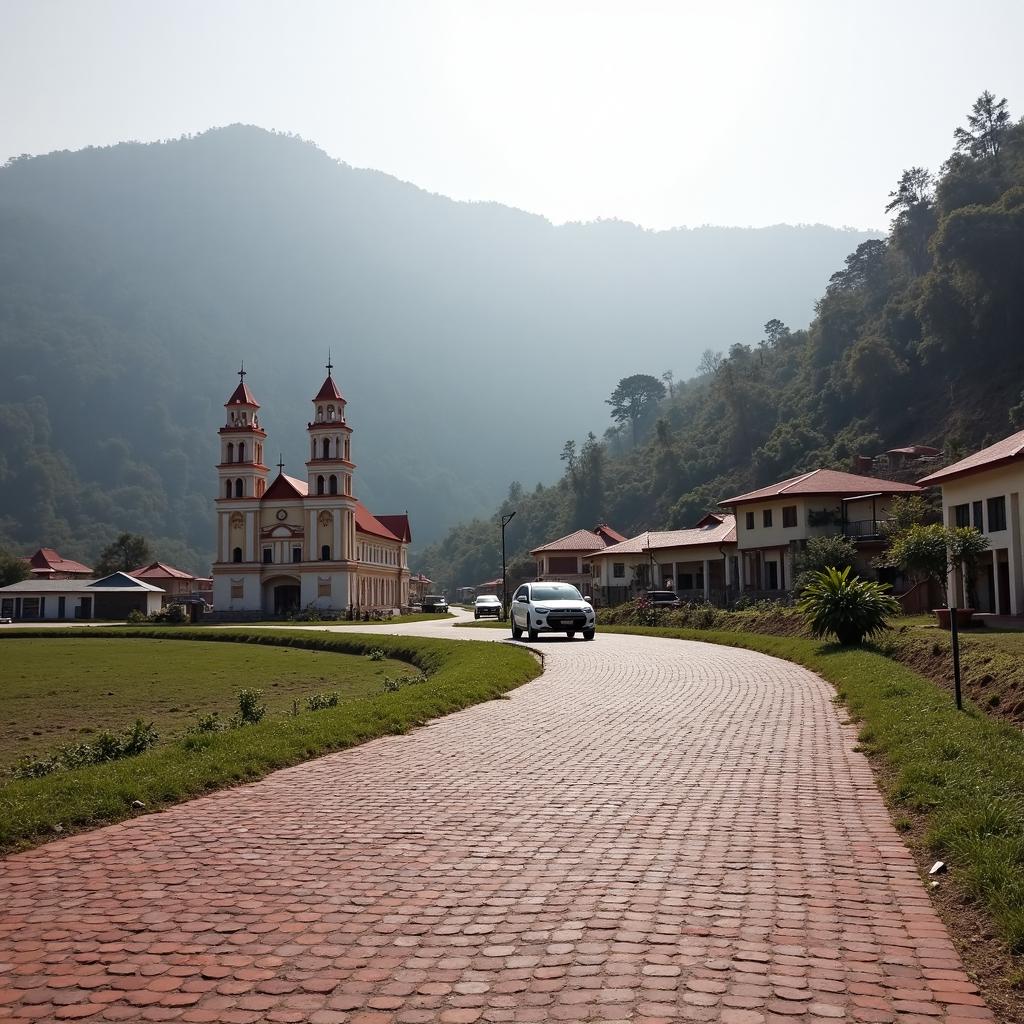}}%
        \fbox{\includegraphics[width=\dcifluximgwidth]{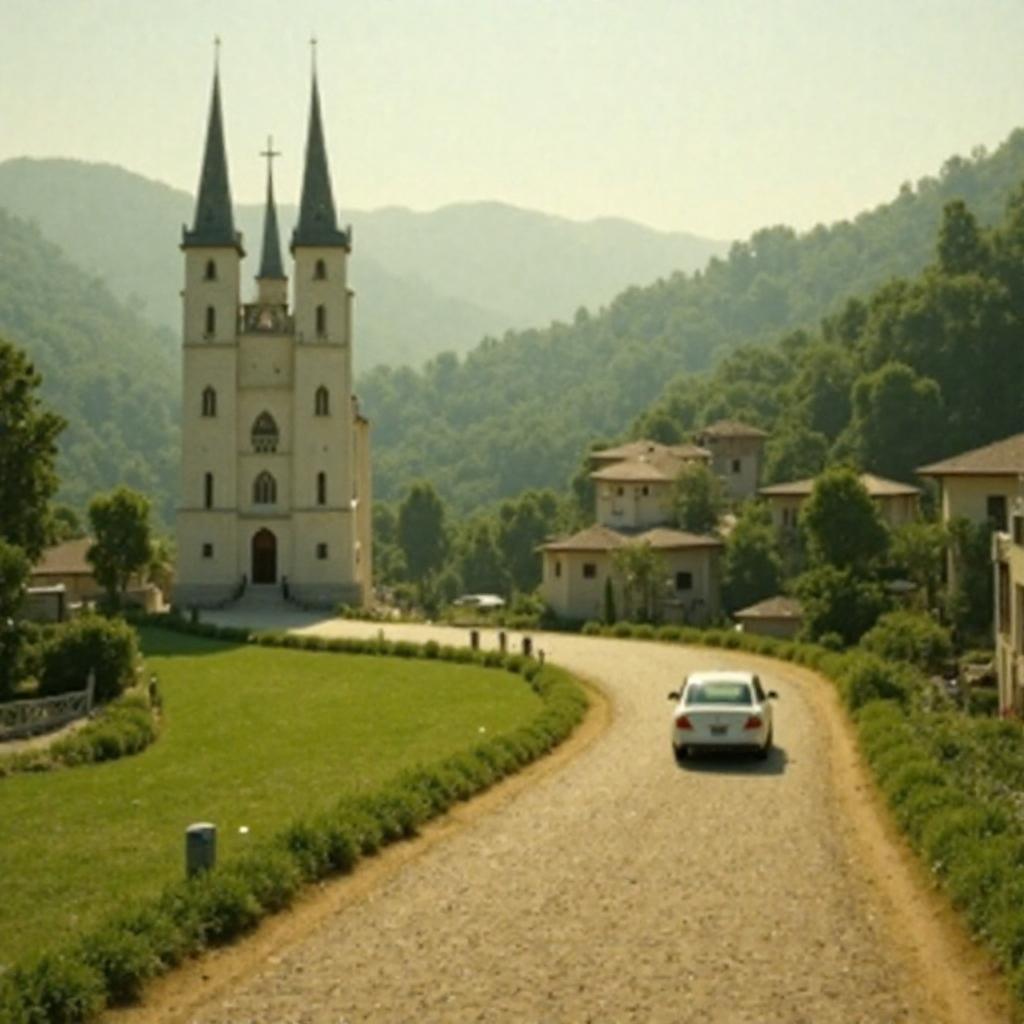}}%
        \fbox{\includegraphics[width=\dcifluximgwidth]{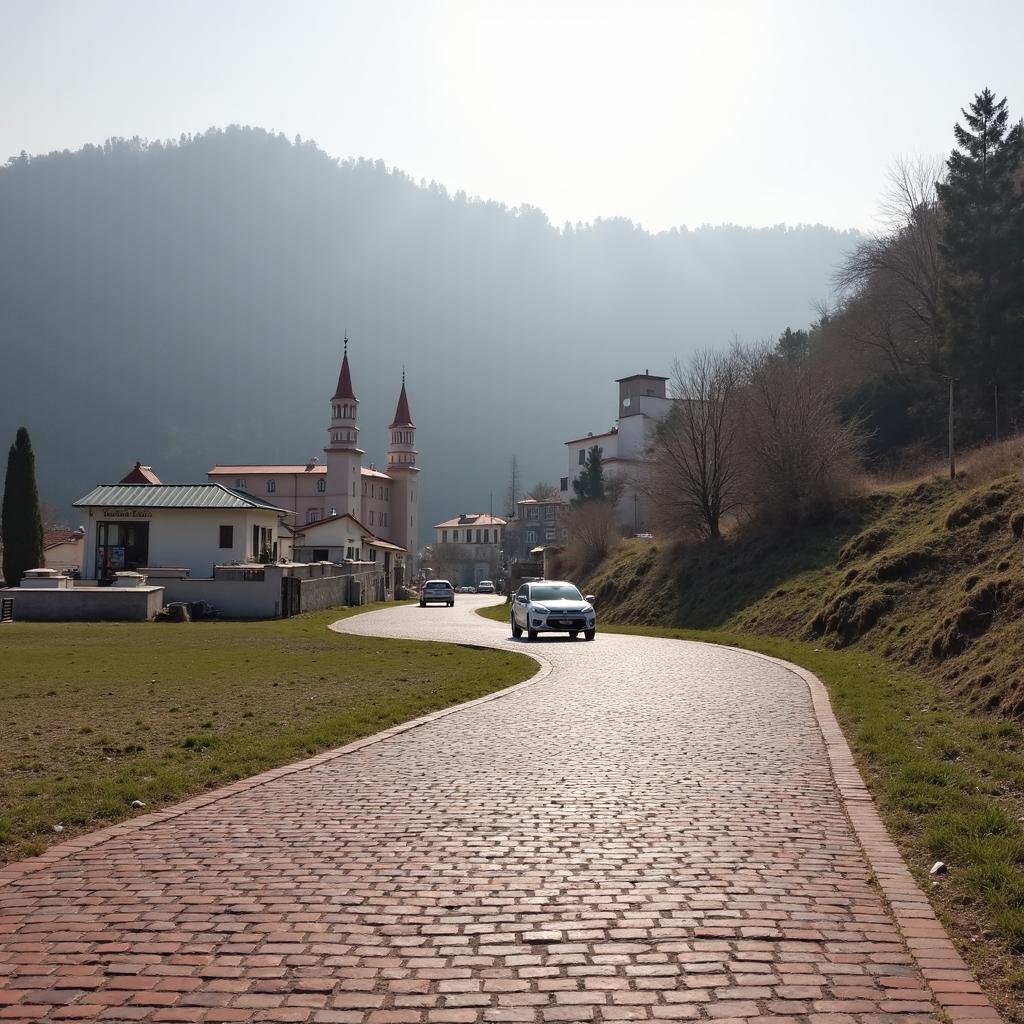}}%
        \fbox{\includegraphics[width=\dcifluximgwidth]{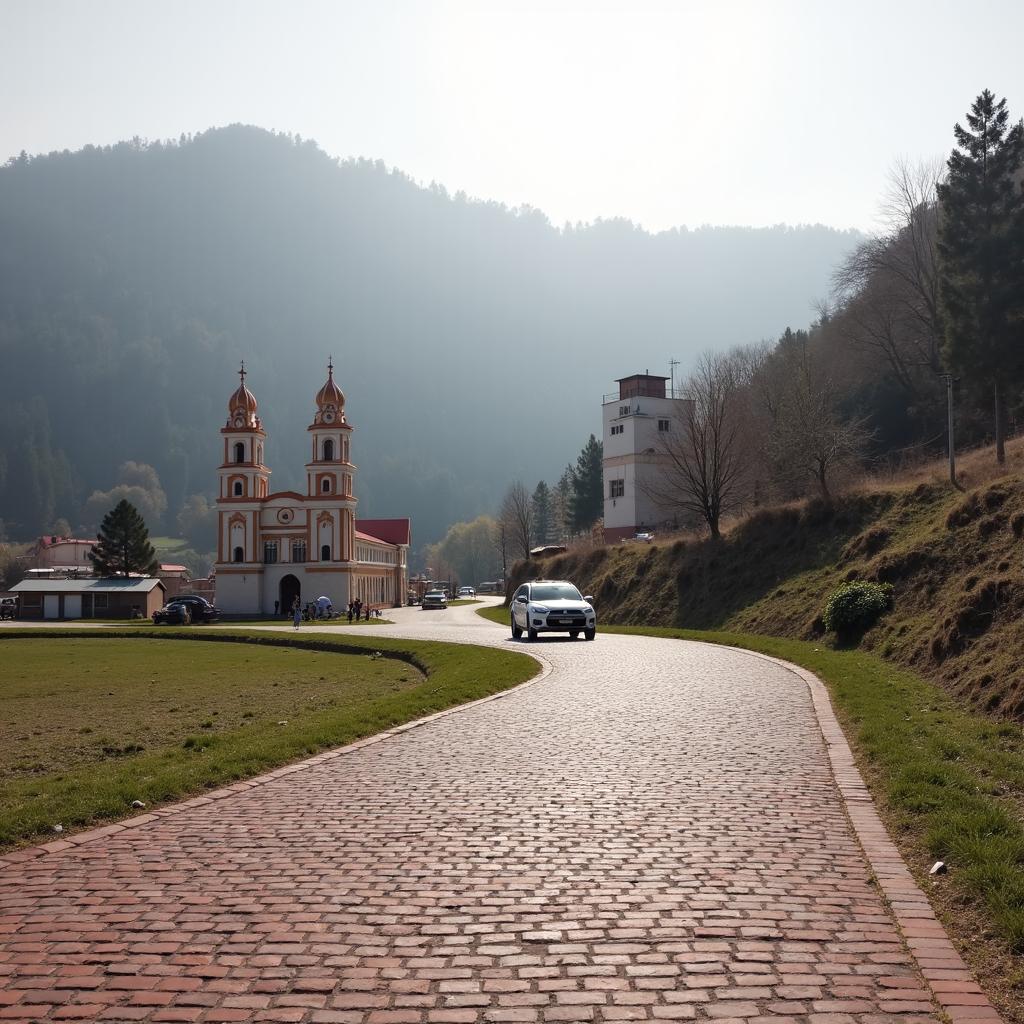}}\\[0.5ex]
        \hfill
    \end{minipage}
    \vspace{-18pt}
    \caption*{
        \begin{minipage}{\dcifluxcapwidth}
        \centering
            \tiny{Prompt: \textit{A brick road that starts in the foreground goes down a slope and into the background. There are buildings on the left and right sides of the road.   A brick road that starts in the foreground extends back and to the left into the background. There is a white vehicle traveling along the road. Buildings can be seen lining the area to the right of the road. There is a field to the left of the road, and buildings to the left of it in the background. There is a religious building in the background that has two towers rising up from it. A wooded area can be seen on the right hand side of the background. A mountainside can be seen on the left hand side of the background. There are trees growing on the mountainside, and mist in the air to the right of the mountainside.}}
        \end{minipage}
    }

    \vspace{0.4cm}
    \begin{minipage}[t]{\textwidth}
        \centering
        \fbox{\includegraphics[width=\dcifluximgwidth]{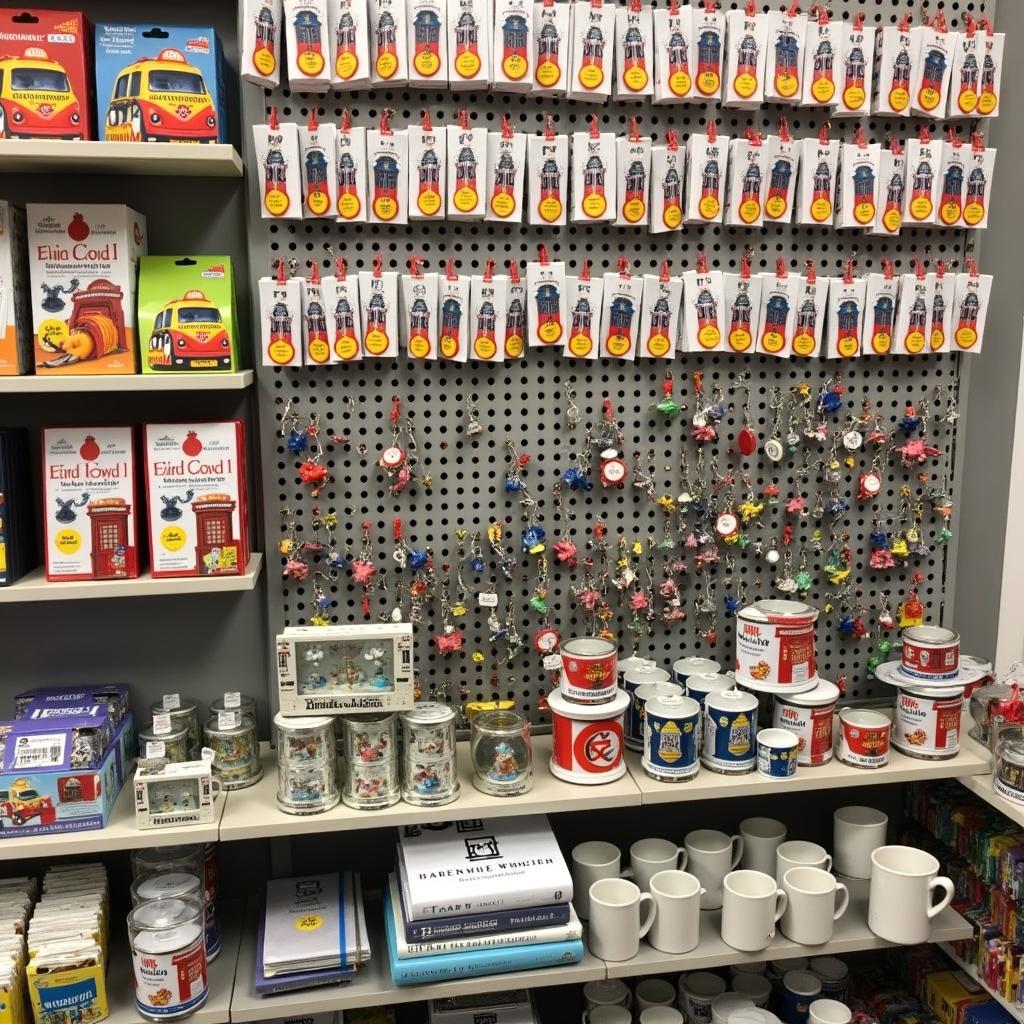}}
        \hfill
        \fbox{\includegraphics[width=\dcifluximgwidth]{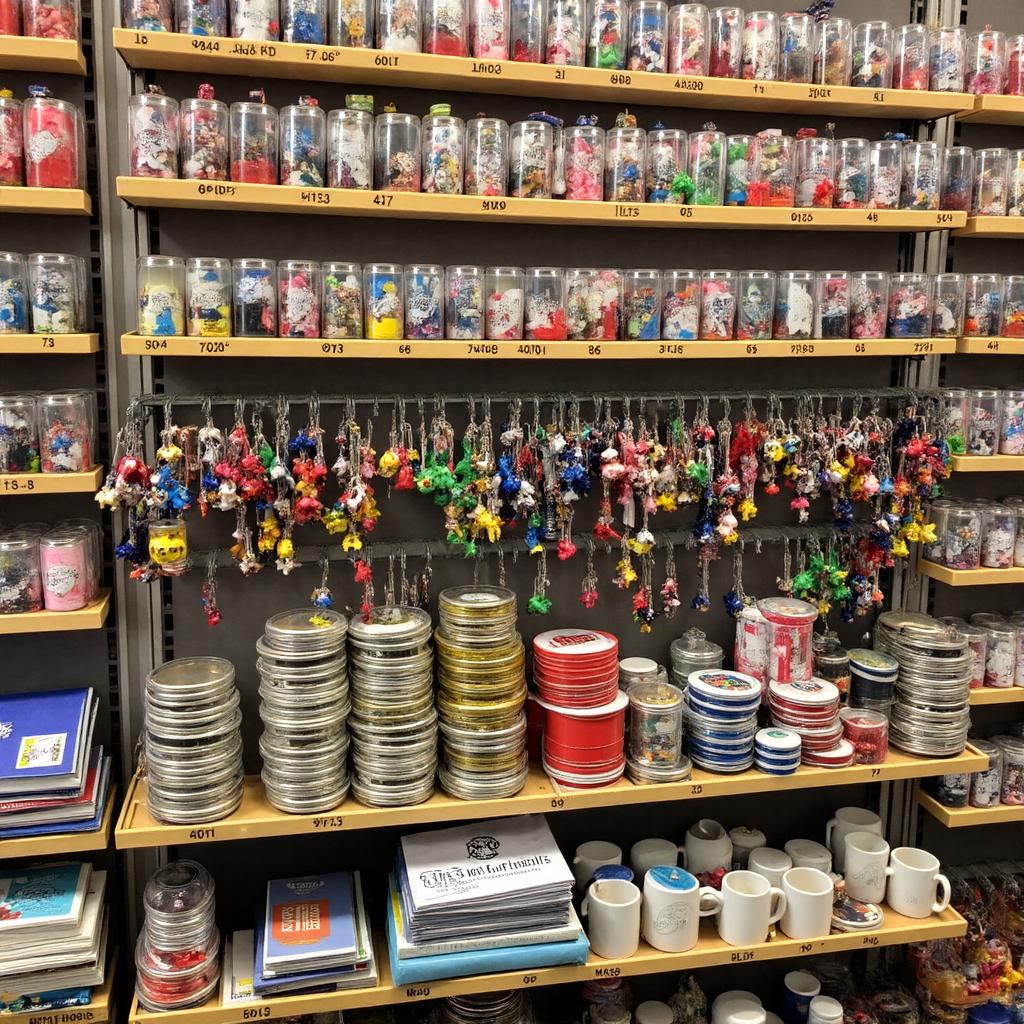}}%
        \fbox{\includegraphics[width=\dcifluximgwidth]{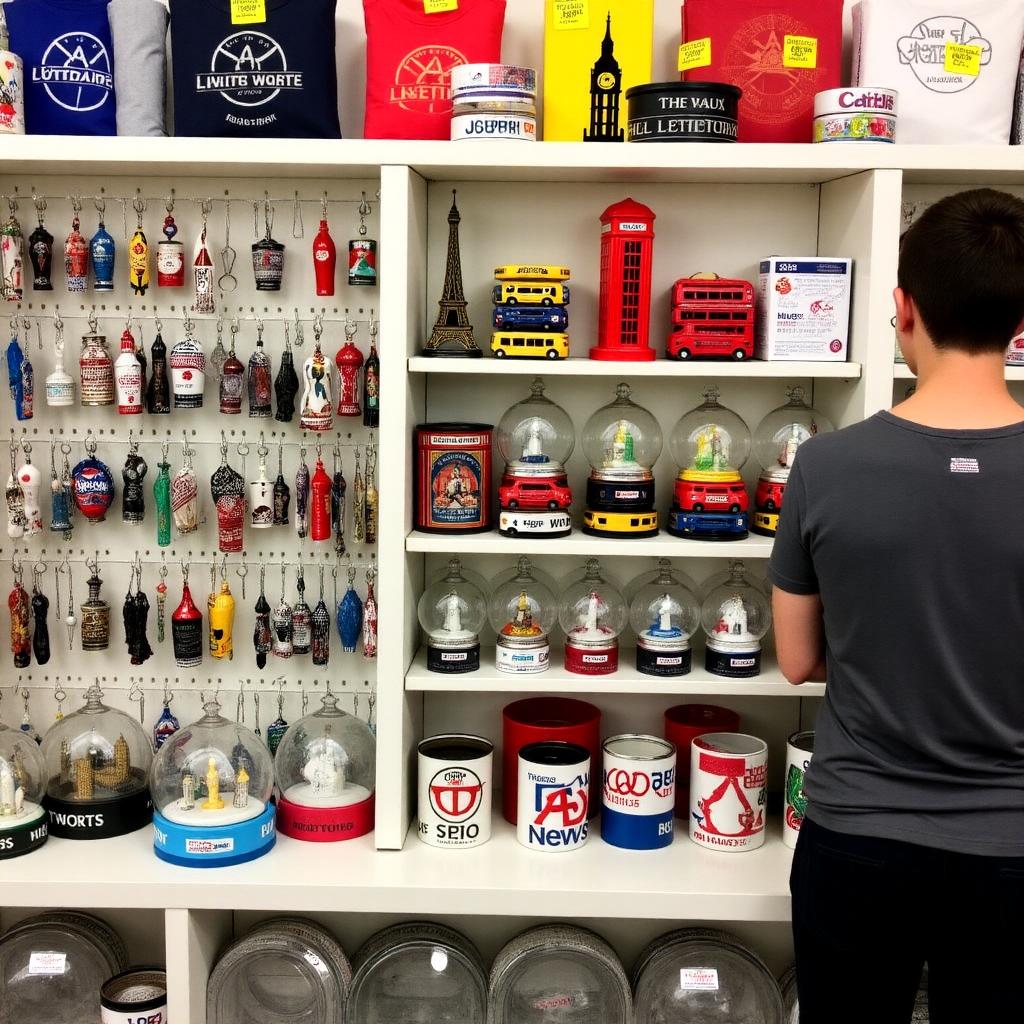}}%
        \fbox{\includegraphics[width=\dcifluximgwidth]{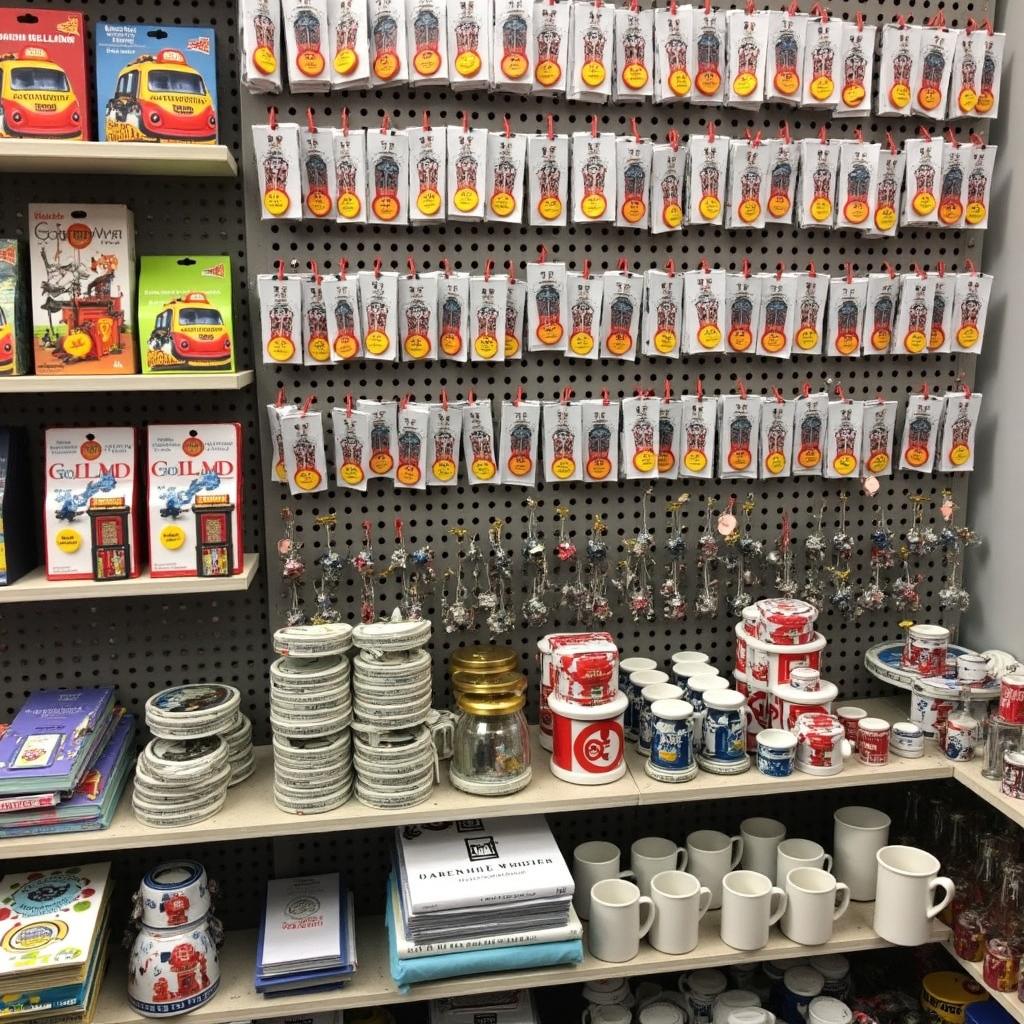}}%
        \fbox{\includegraphics[width=\dcifluximgwidth]{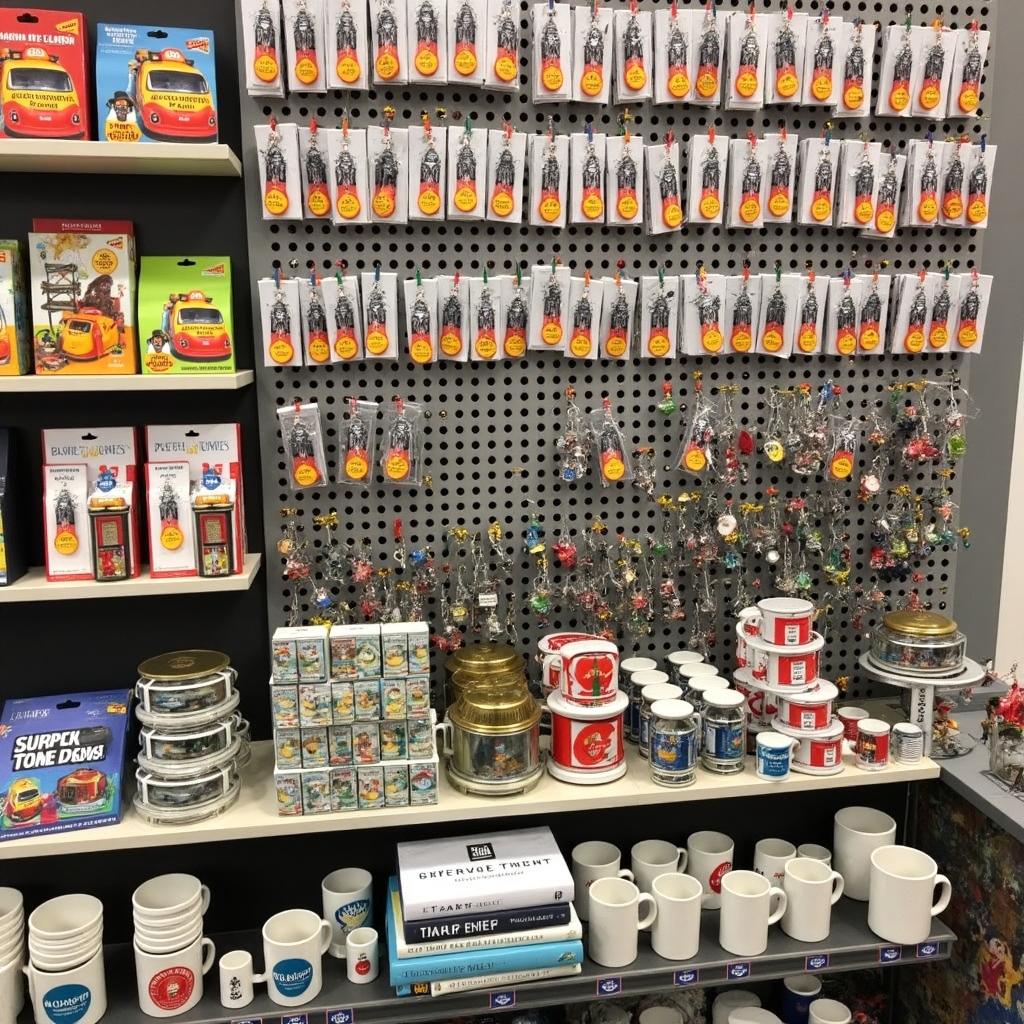}}%
        \fbox{\includegraphics[width=\dcifluximgwidth]{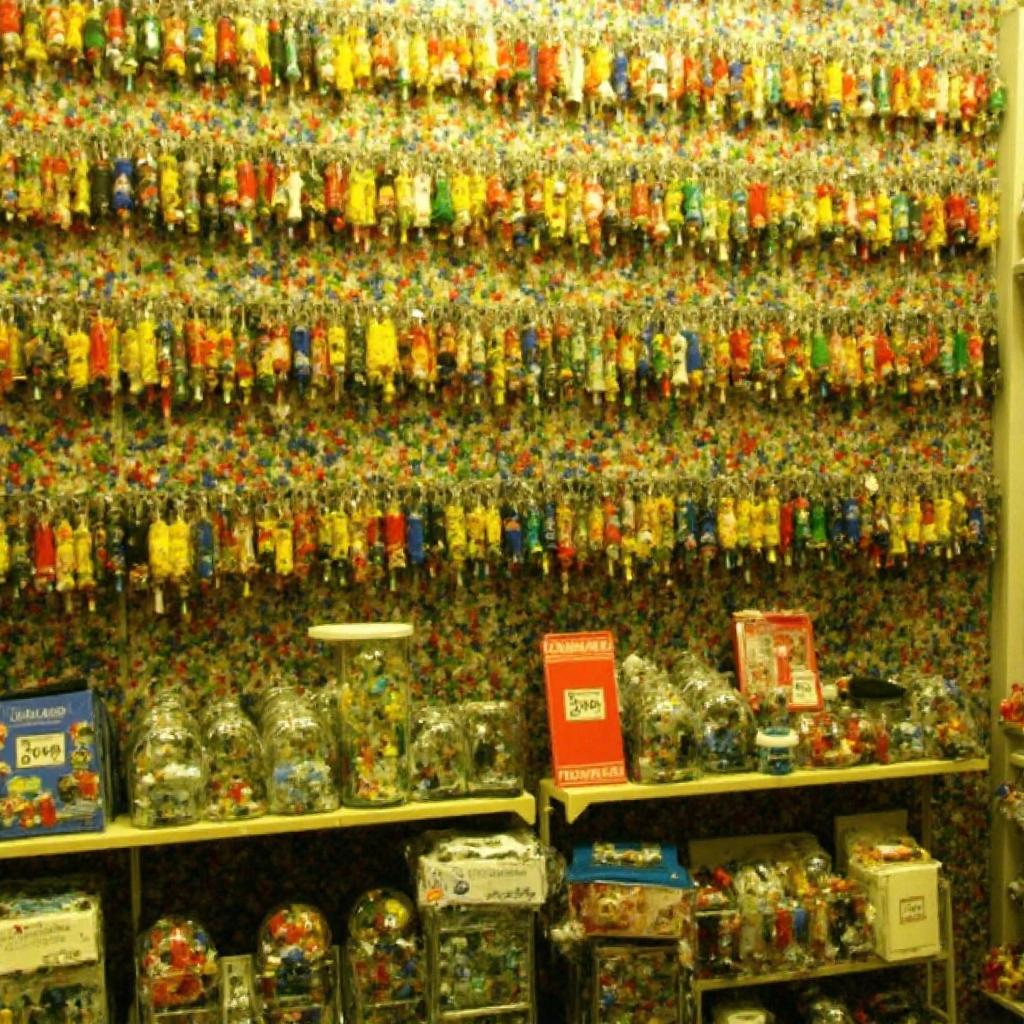}}%
        \fbox{\includegraphics[width=\dcifluximgwidth]{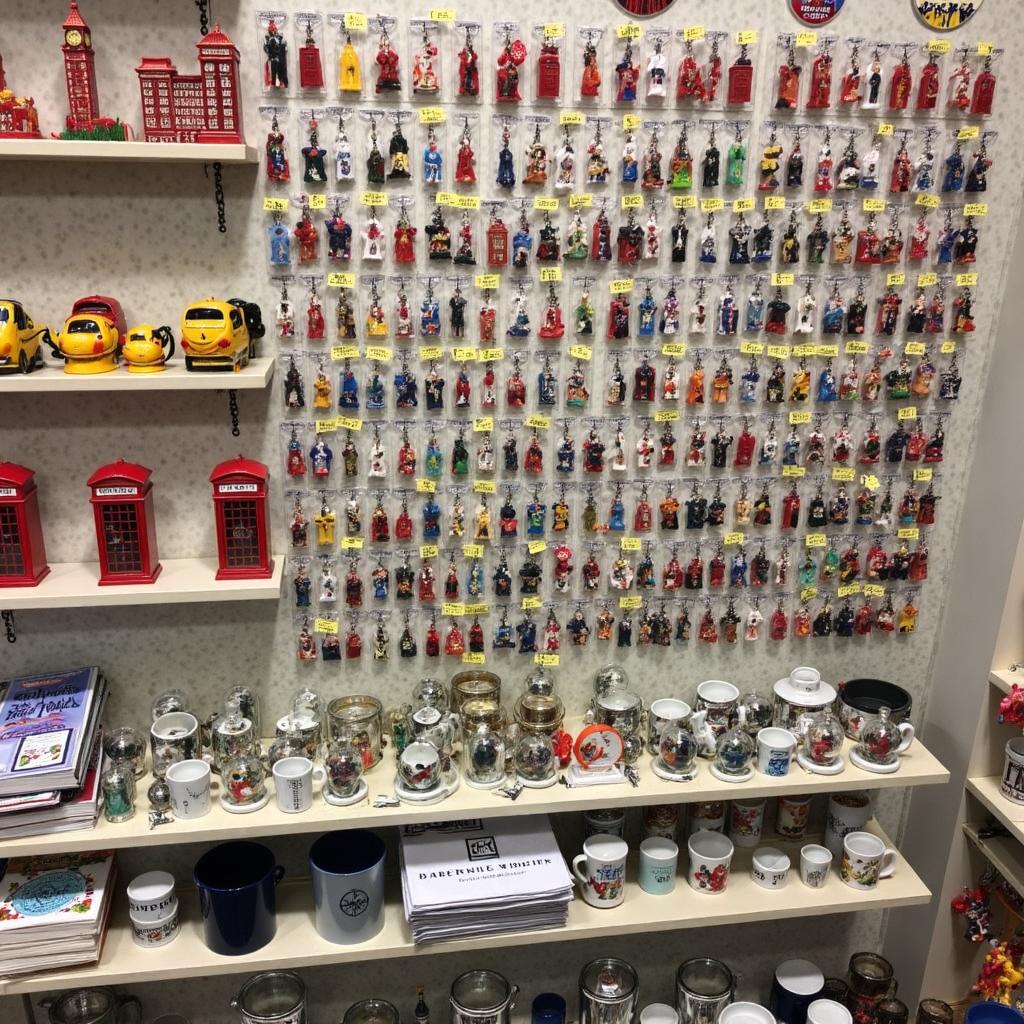}}%
        \fbox{\includegraphics[width=\dcifluximgwidth]{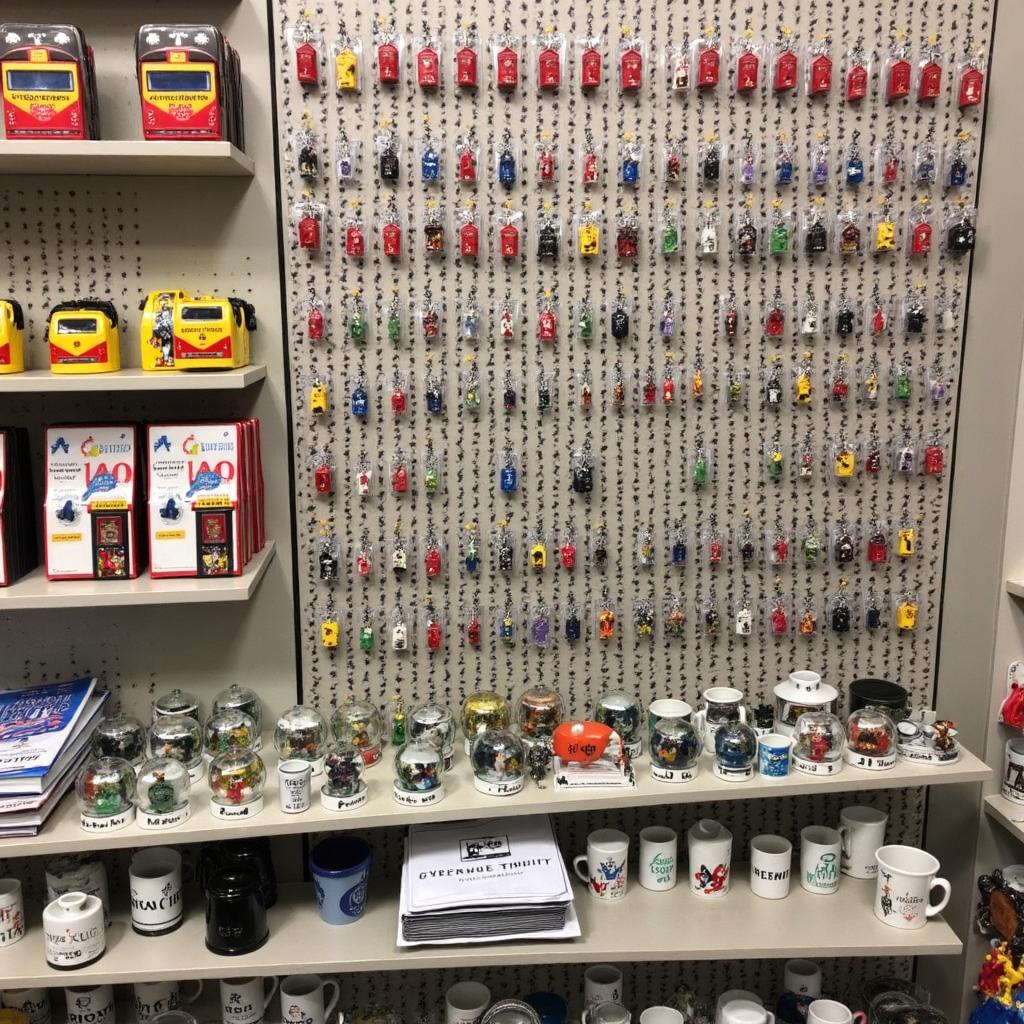}}\\[0.5ex]
        \hfill
    \end{minipage}
    \vspace{-18pt}
    \caption*{
        \begin{minipage}{\dcifluxcapwidth}
        \centering
            \tiny{Prompt: \textit{Tourist merchandise is hanging on a wall on display for people to buy. There are shirts on the left, magnets in the middle, and various objects on the right. Five rows of various shirts that are folded into squares and inside of clear plastic are on the left side of the image. A wall in the middle section is covered in rows of bright colored magnets for sale. There is a round yellow and black price tag in the magnet section. The right side of the image has shelves at the top with miniature figures of the clock tower and red phone booth. Four rows of keychains are hanging below the miniature figures. A shelf with stacks of toy taxis and busses are below the key chains. A book is on its side on a shelf below the toys along with round silver plates. A shelf filled with various sizes of snow globes is below the book. The bottom shelf on the right has decorative plates on stands. A line of mugs is between the magnets and the items on the right.}}
        \end{minipage}
    }

    \vspace{0.4cm}
    \begin{minipage}[t]{\textwidth}
        \centering
        \fbox{\includegraphics[width=\dcifluximgwidth]{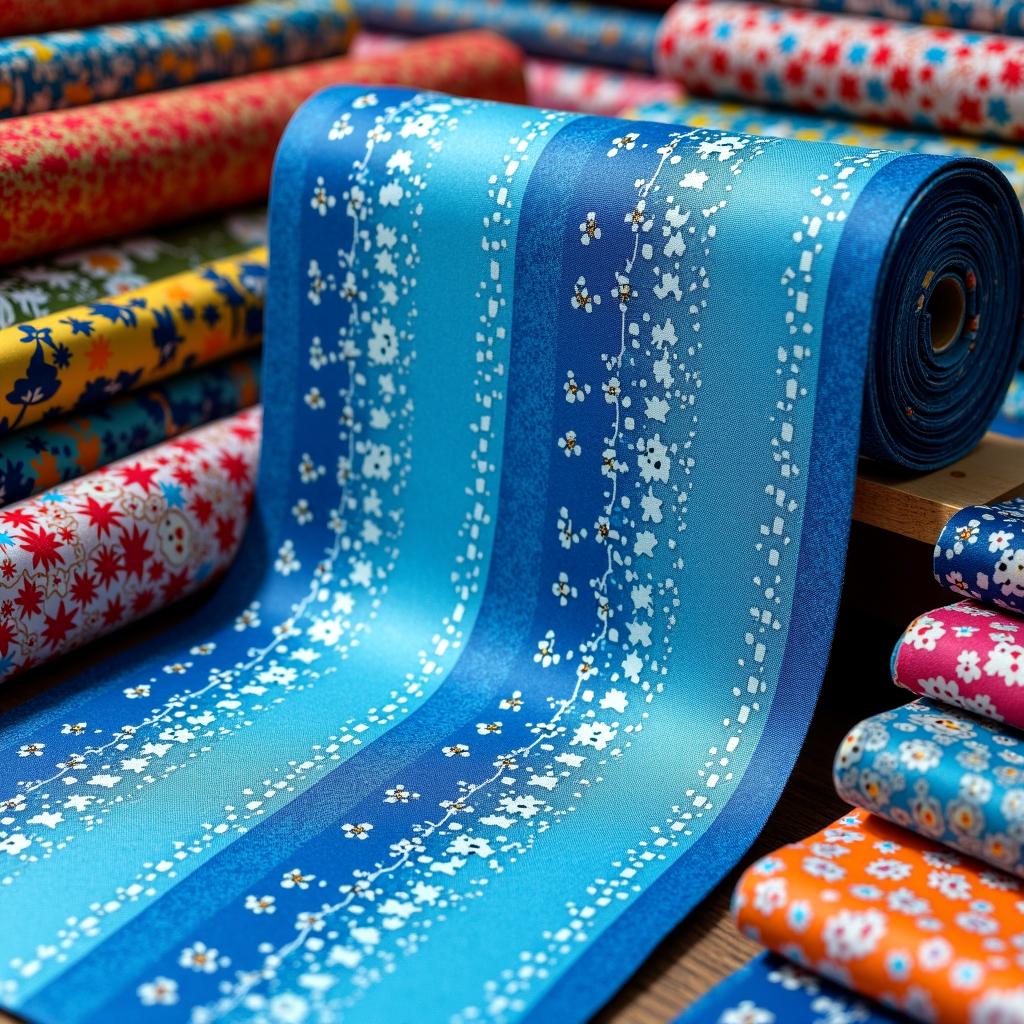}}
        \hfill
        \fbox{\includegraphics[width=\dcifluximgwidth]{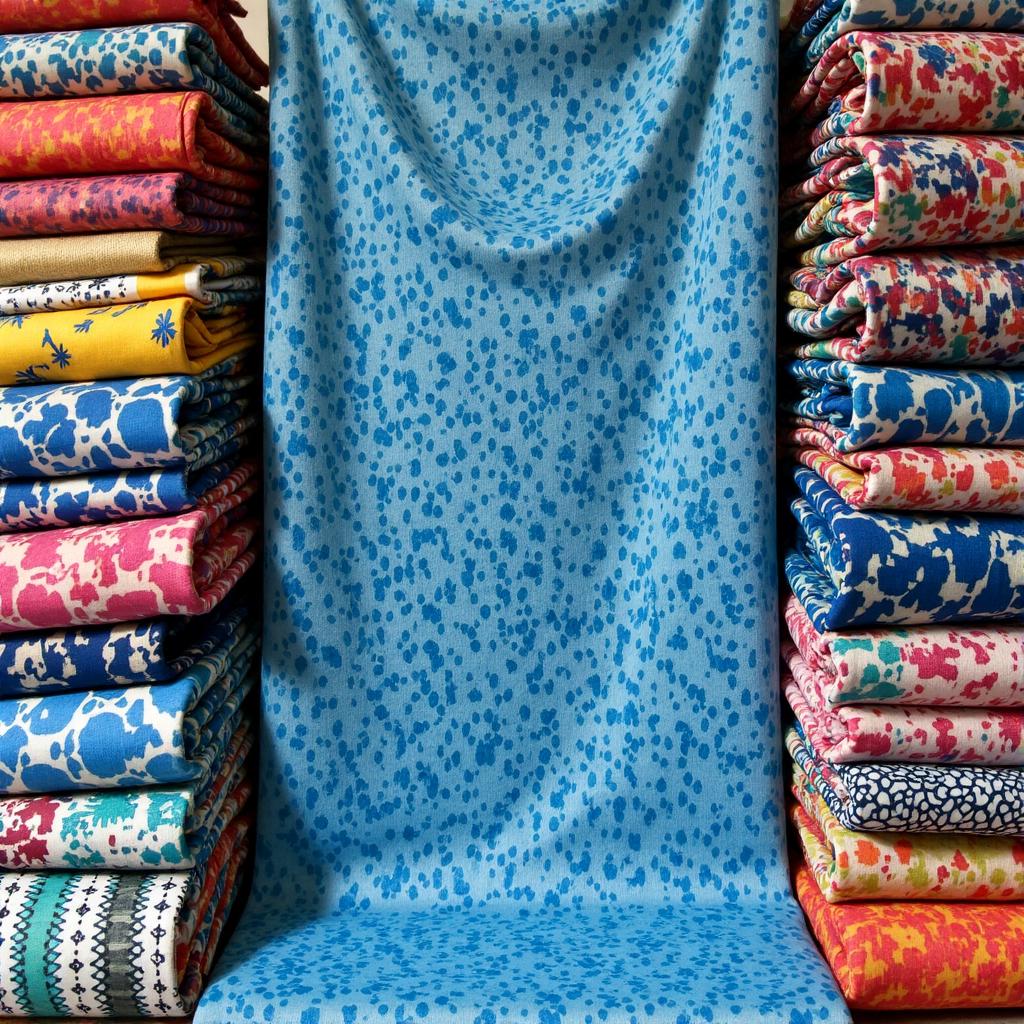}}%
        \fbox{\includegraphics[width=\dcifluximgwidth]{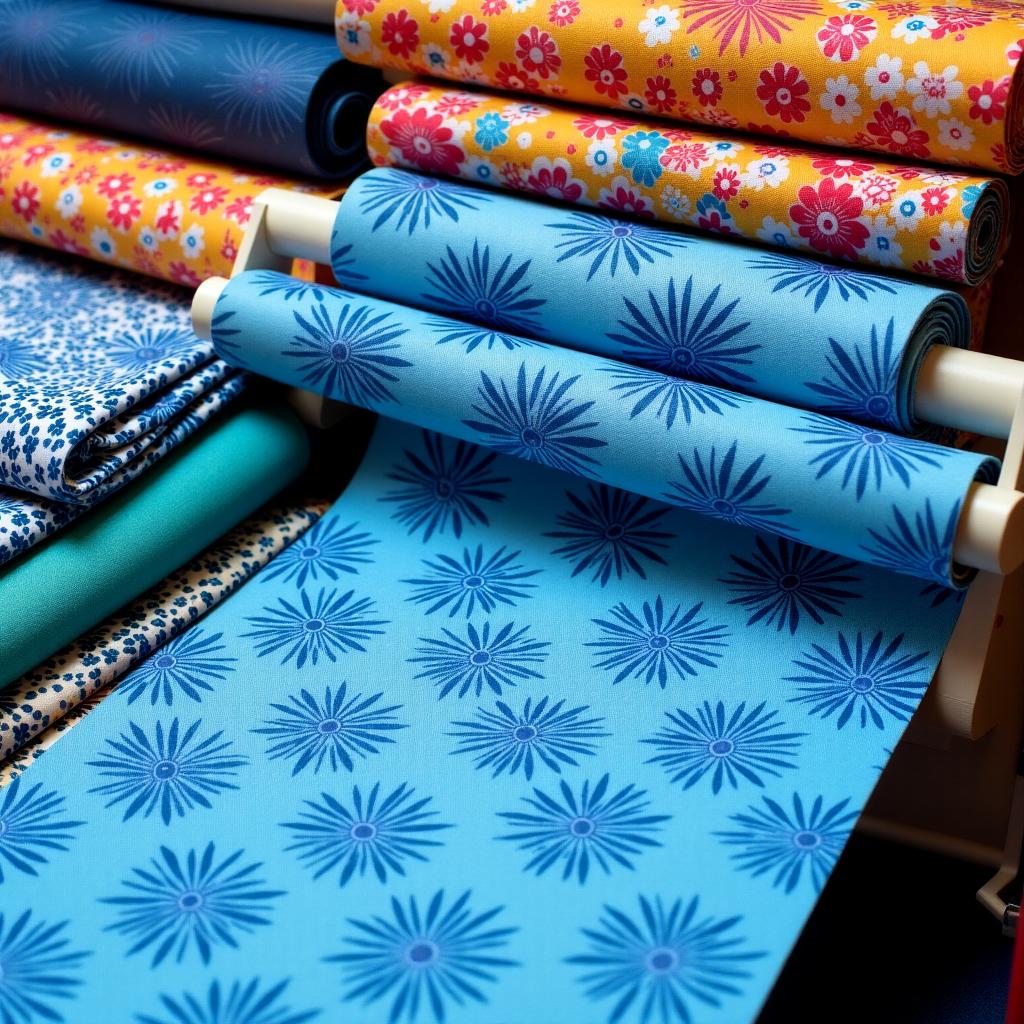}}%
        \fbox{\includegraphics[width=\dcifluximgwidth]{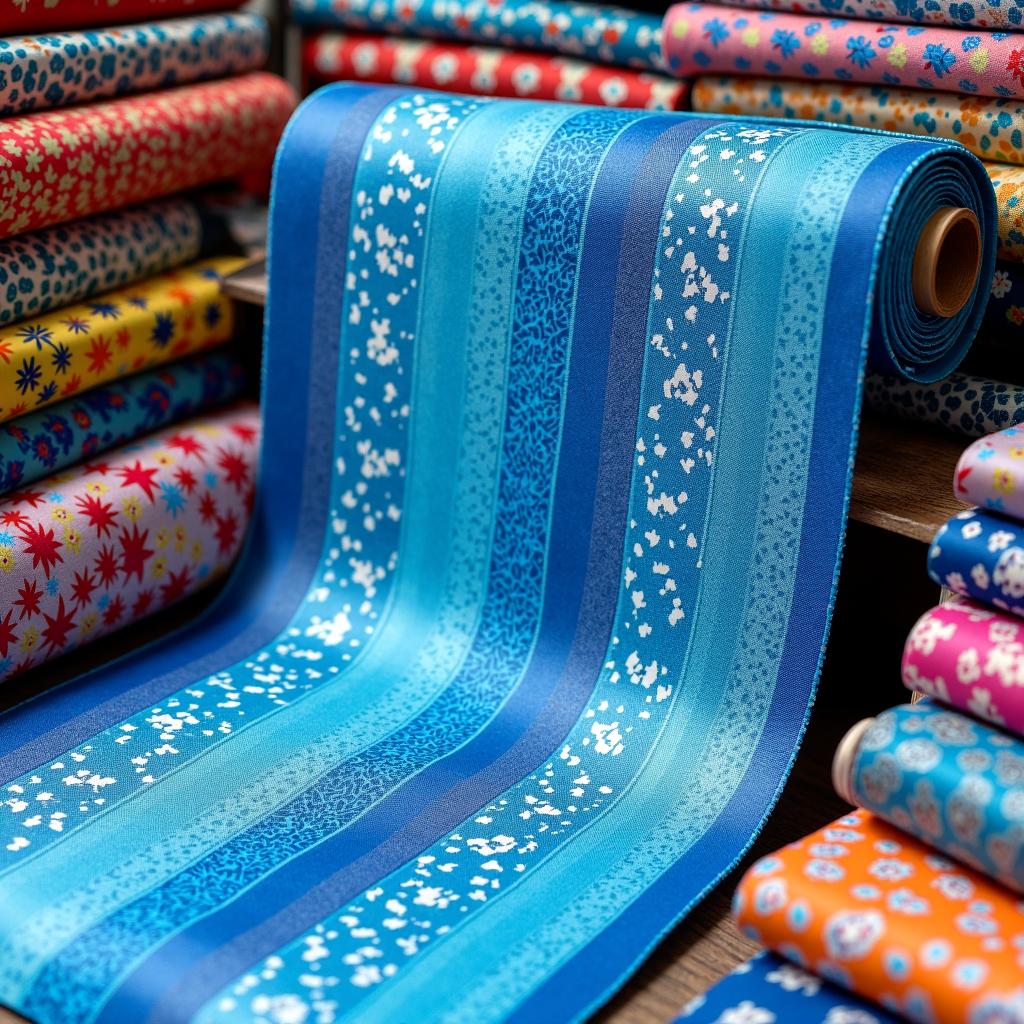}}%
        \fbox{\includegraphics[width=\dcifluximgwidth]{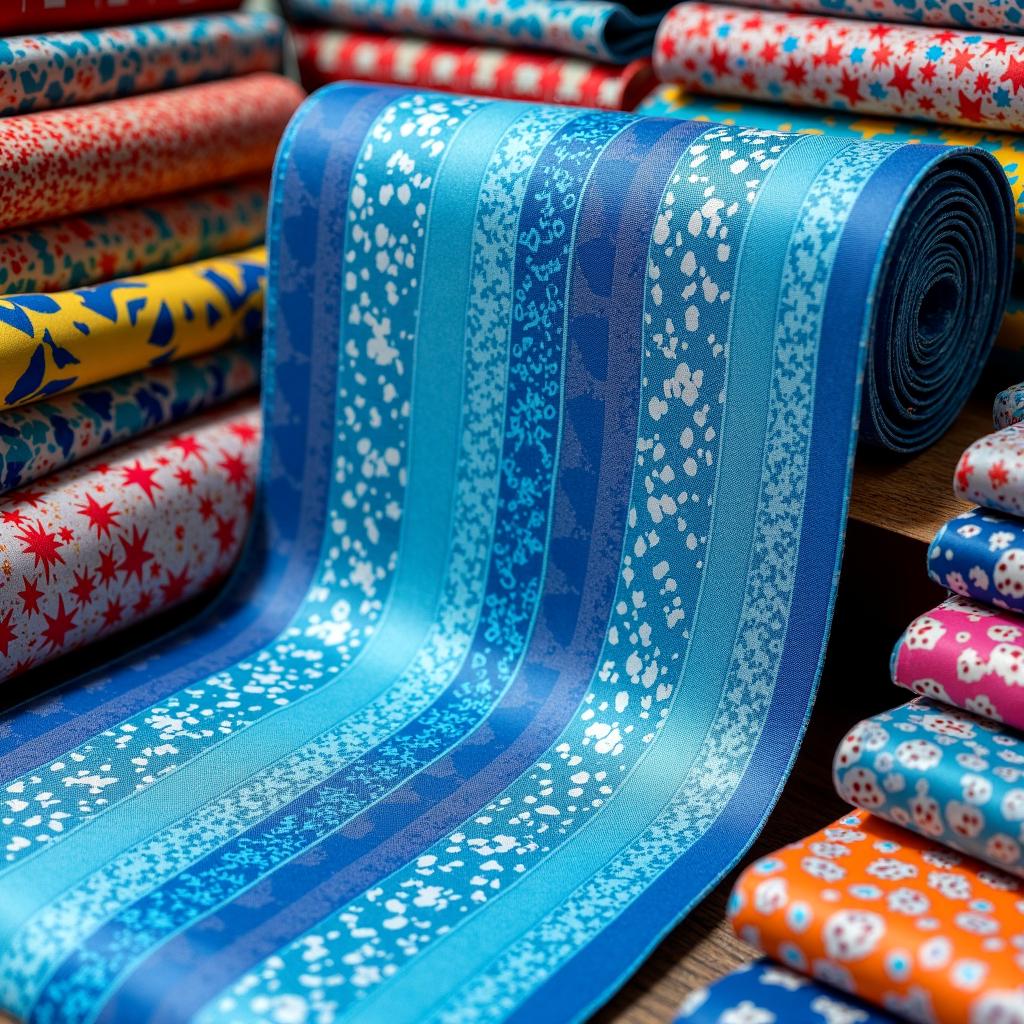}}%
        \fbox{\includegraphics[width=\dcifluximgwidth]{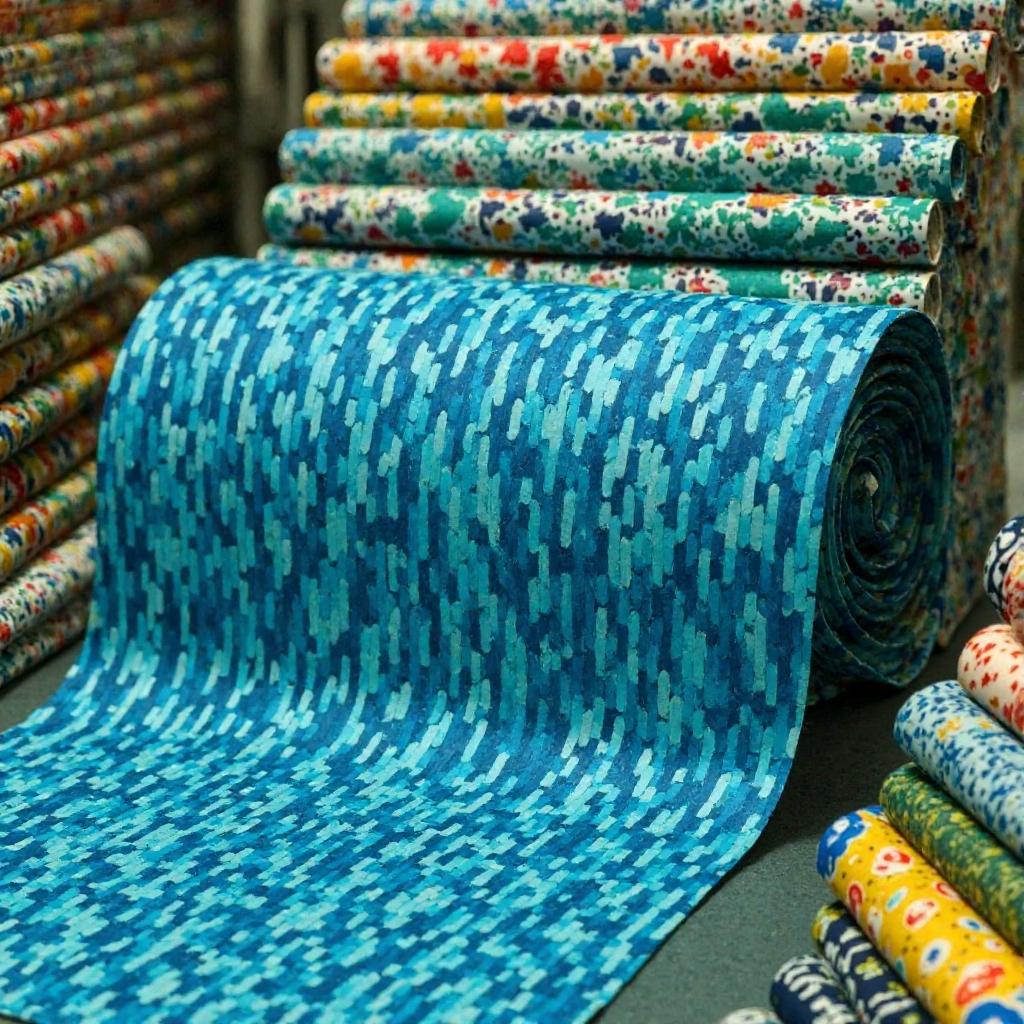}}%
        \fbox{\includegraphics[width=\dcifluximgwidth]{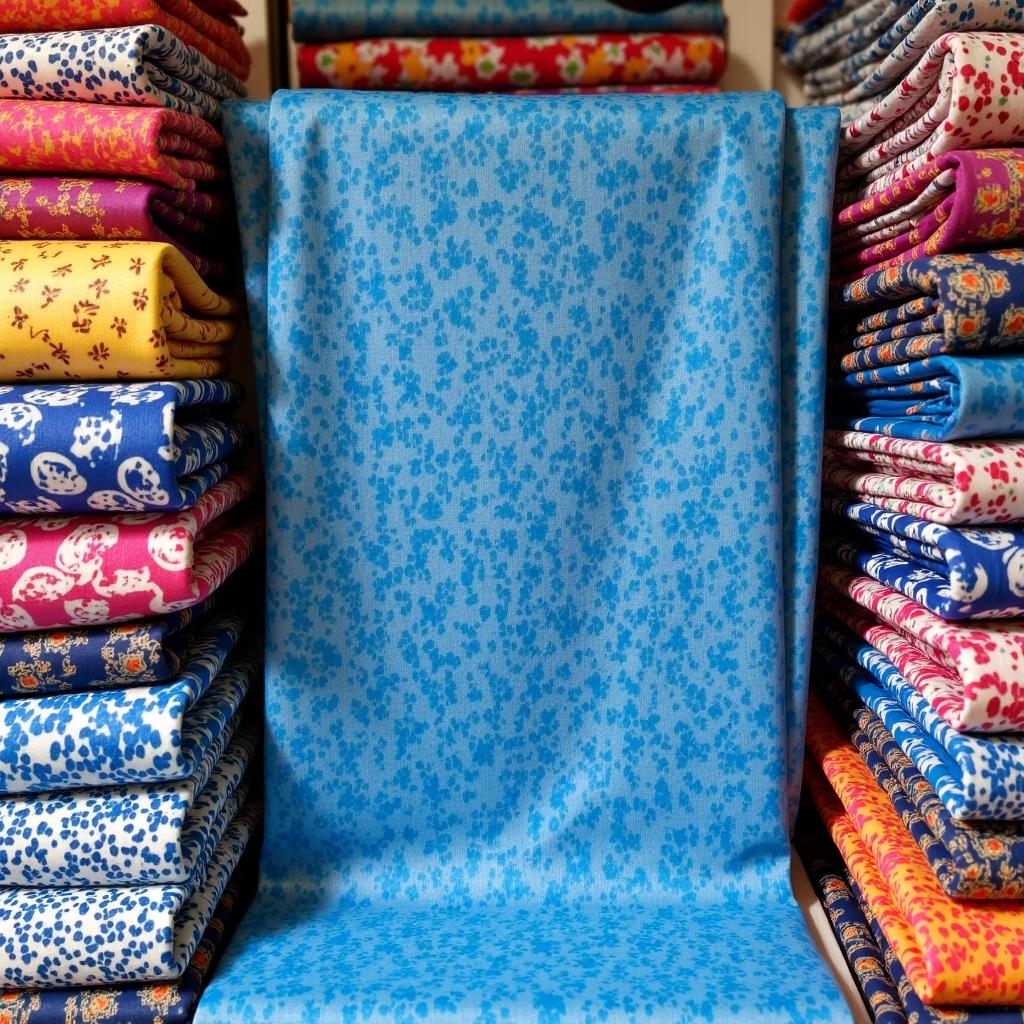}}%
        \fbox{\includegraphics[width=\dcifluximgwidth]{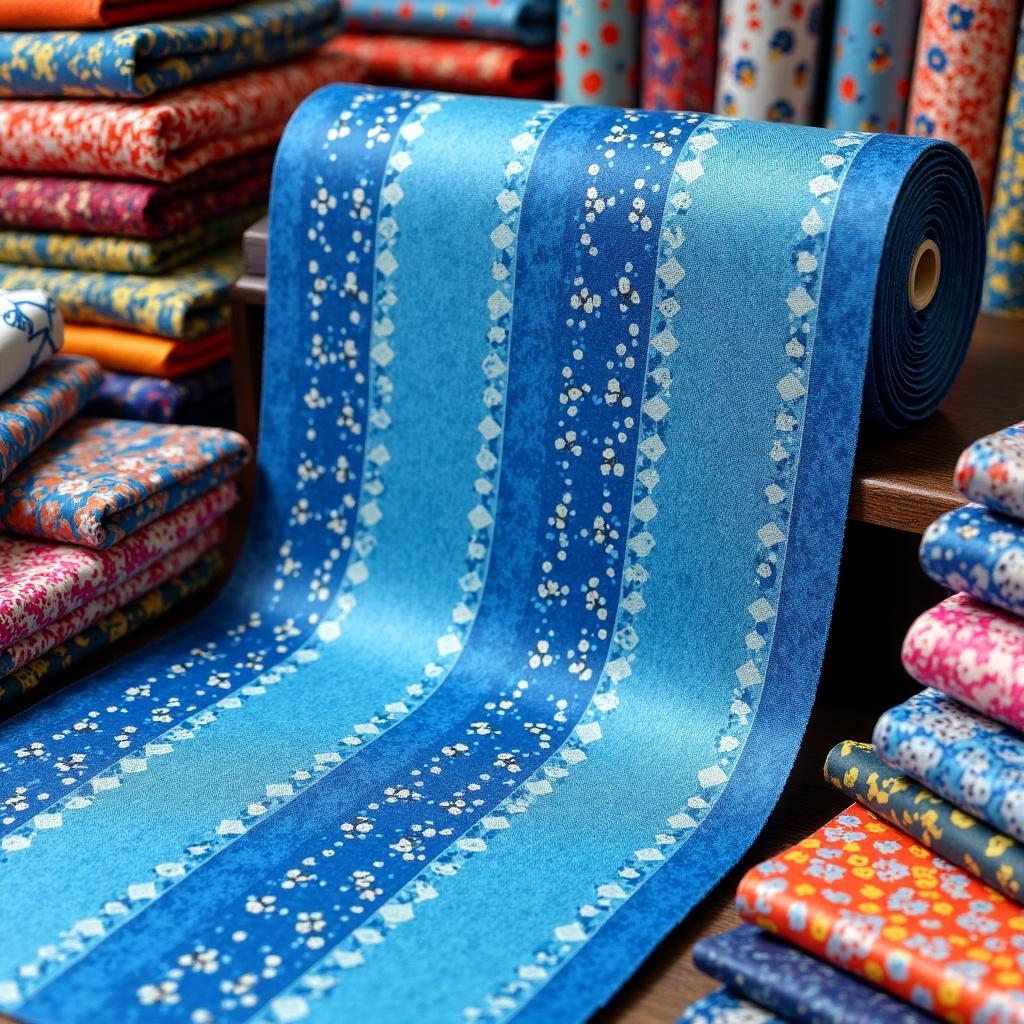}}\\[0.5ex]
        \hfill
    \end{minipage}
    \vspace{-18pt}
    \caption*{
        \begin{minipage}{\dcifluxcapwidth}
        \centering
            \tiny{Prompt: \textit{This is a roll of a fabric that is multiple shades of blue and has a leaf pattern woven into it. It is surrounded by other bright and intricate patterned rolls of fabric. This is a roll of a light blue with dark blue pattern, woven fabric surrounded by rolls and stacks of folded brightly patterned fabrics.}}
        \end{minipage}
    }

    \vspace{0.4cm}
    \begin{minipage}[t]{\textwidth}
        \centering
        \fbox{\includegraphics[width=\dcifluximgwidth]{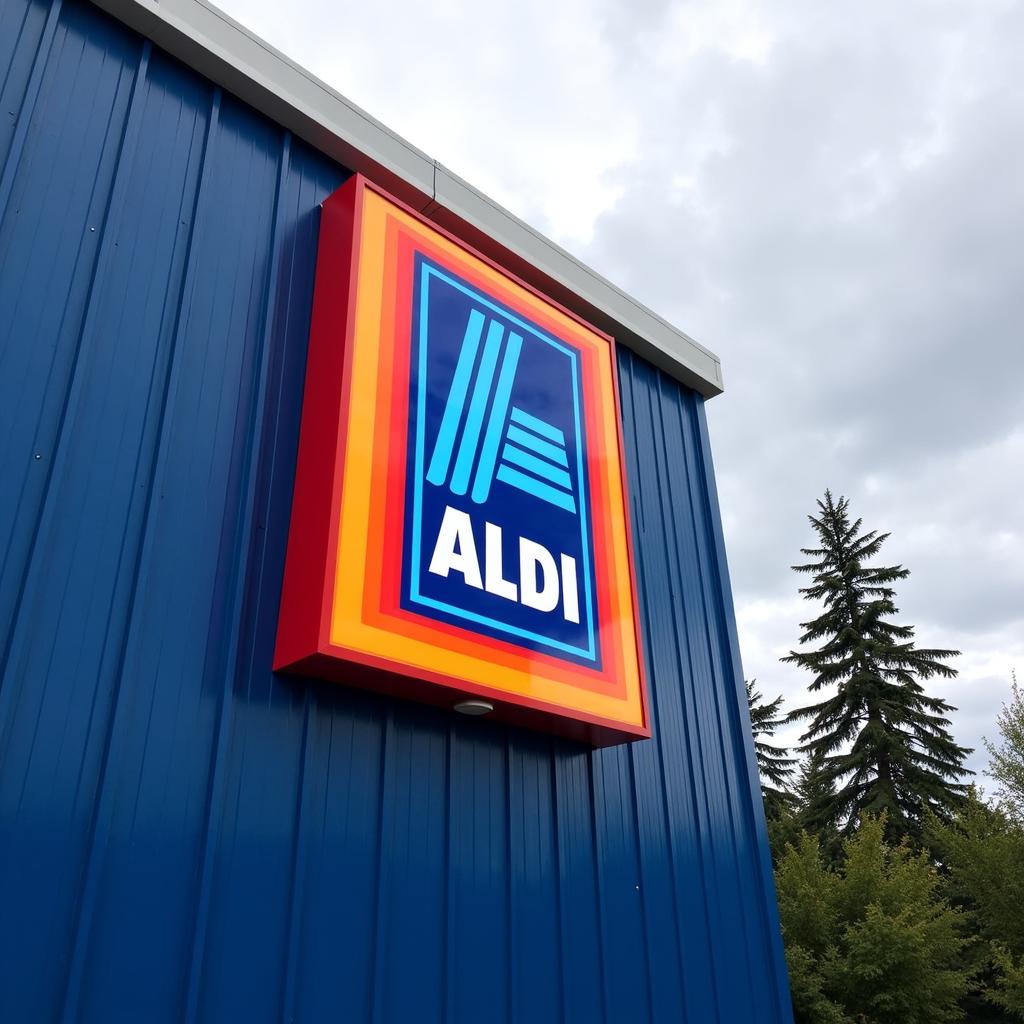}}
        \hfill
        \fbox{\includegraphics[width=\dcifluximgwidth]{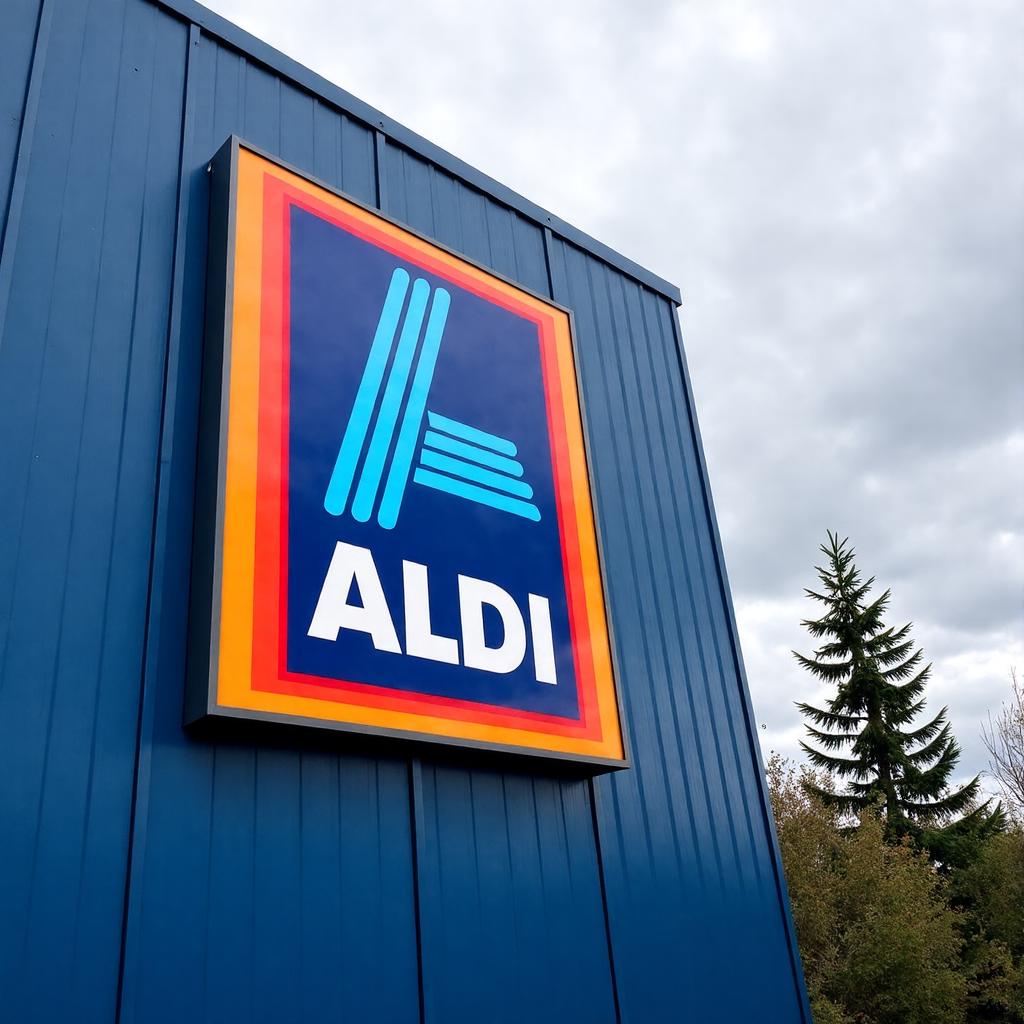}}%
        \fbox{\includegraphics[width=\dcifluximgwidth]{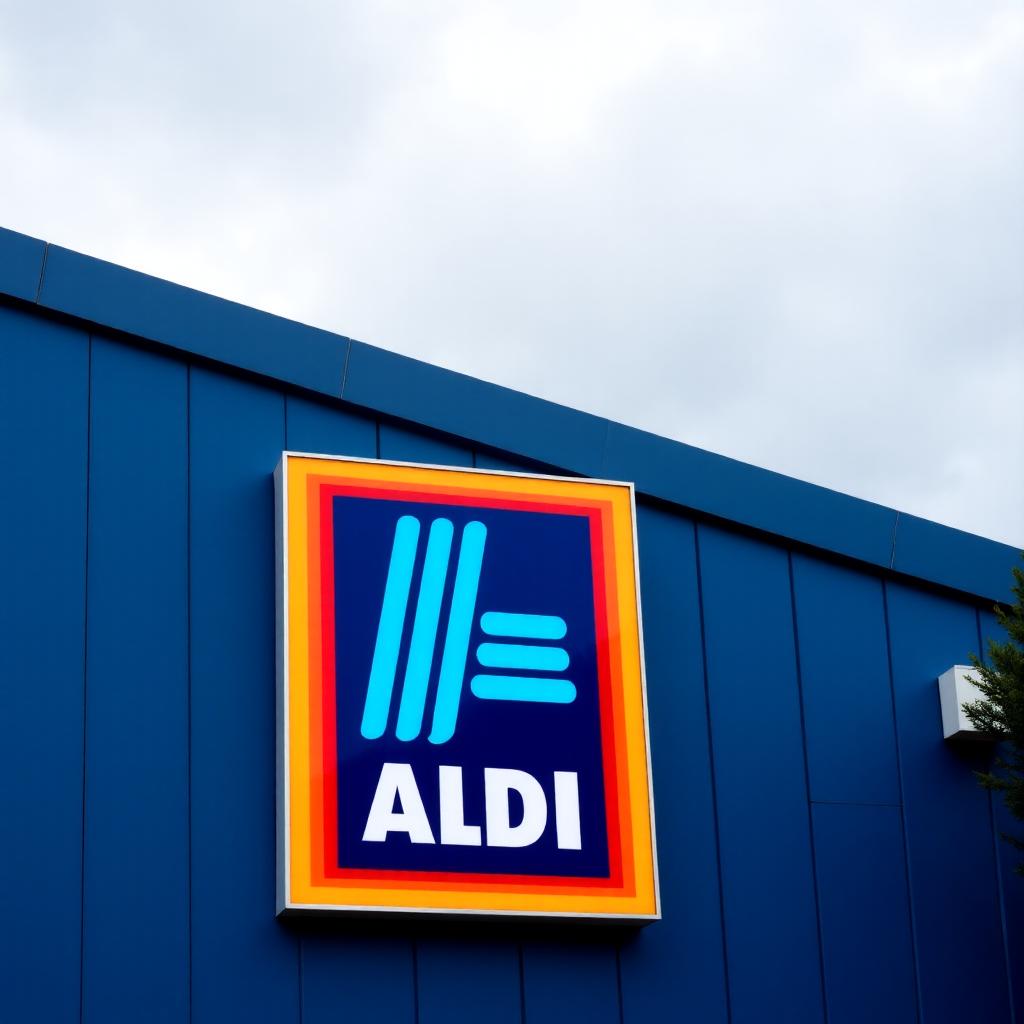}}%
        \fbox{\includegraphics[width=\dcifluximgwidth]{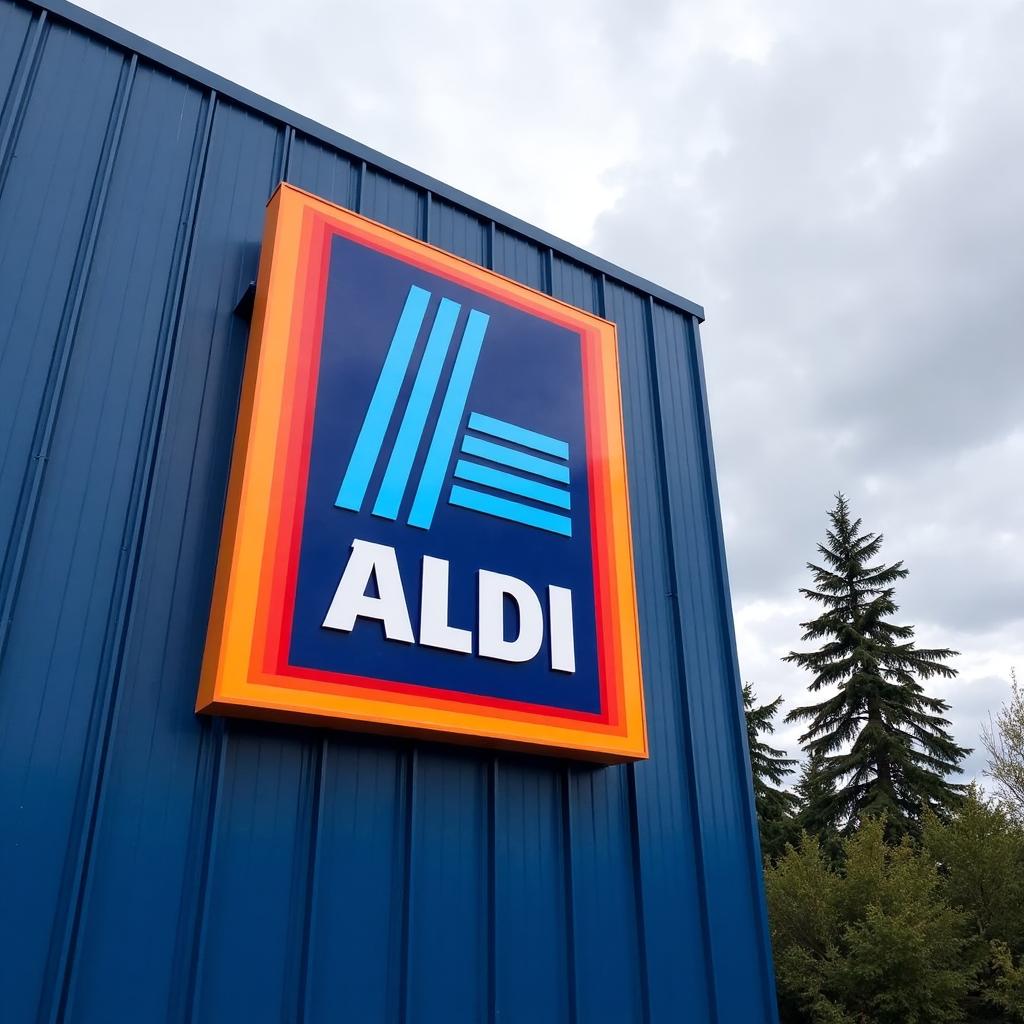}}%
        \fbox{\includegraphics[width=\dcifluximgwidth]{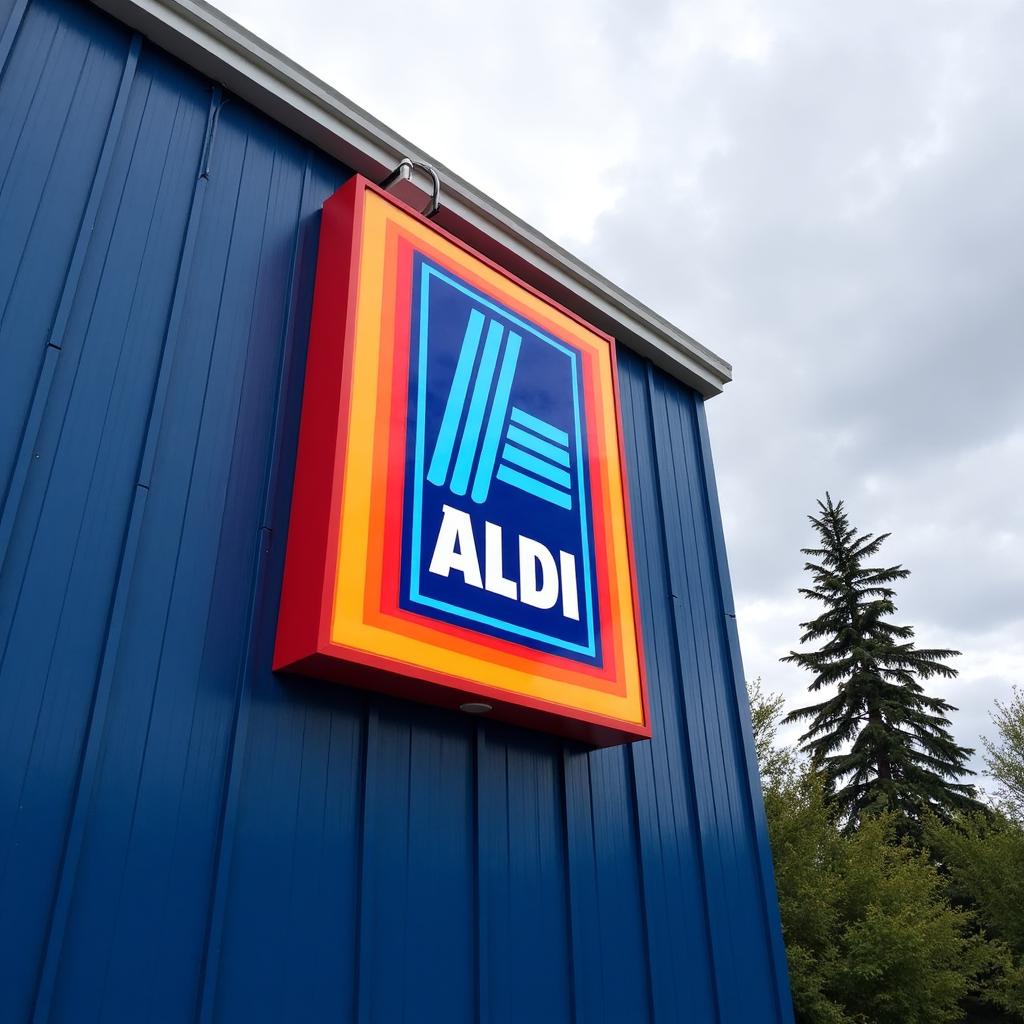}}%
        \fbox{\includegraphics[width=\dcifluximgwidth]{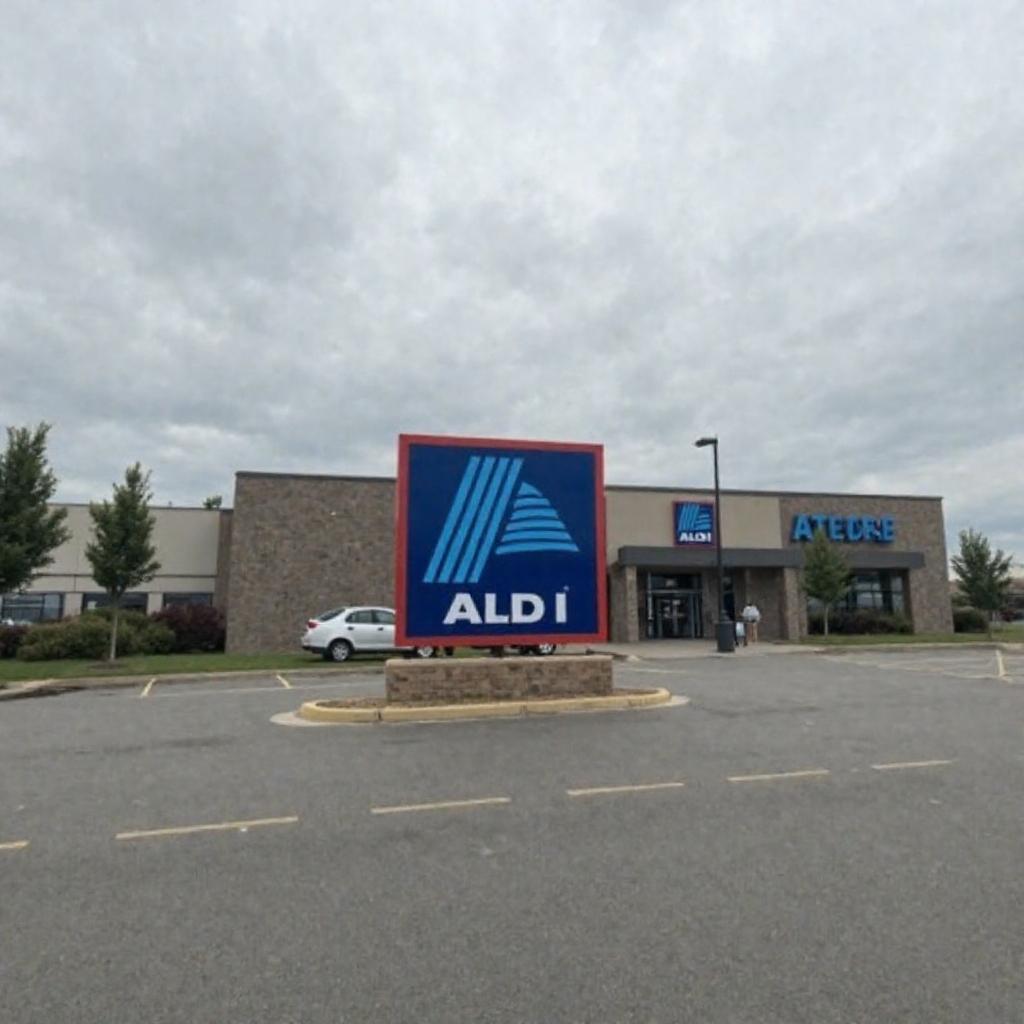}}%
        \fbox{\includegraphics[width=\dcifluximgwidth]{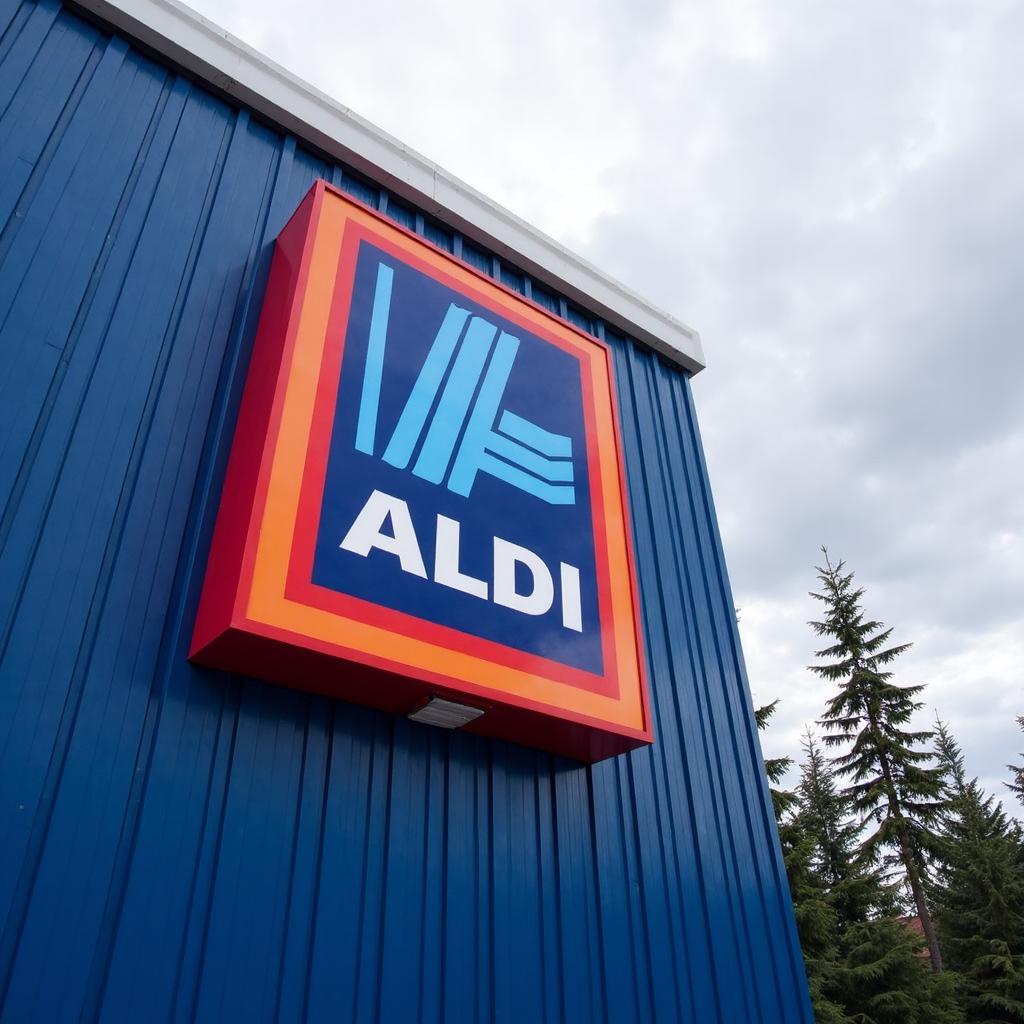}}%
        \fbox{\includegraphics[width=\dcifluximgwidth]{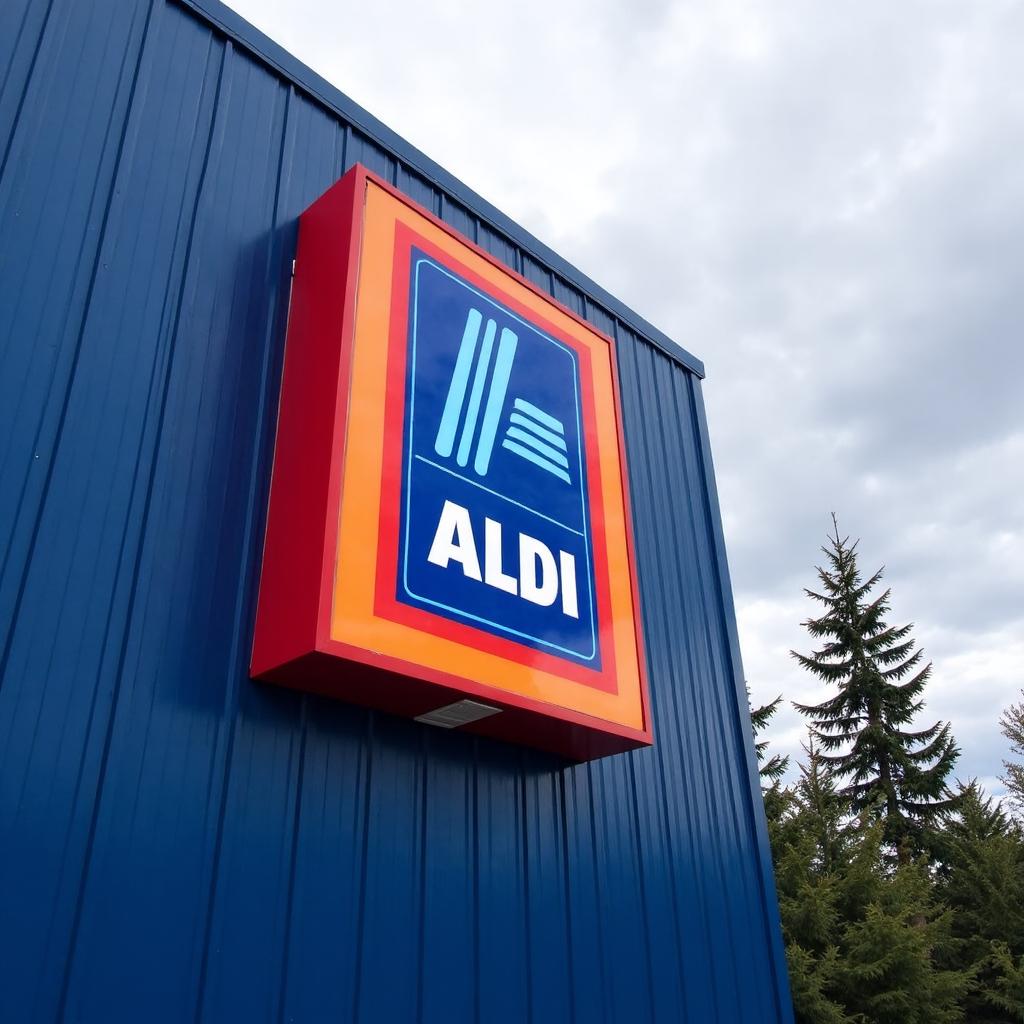}}\\[0.5ex]
        \hfill
    \end{minipage}
    \vspace{-18pt}
    \caption*{
        \begin{minipage}{\dcifluxcapwidth}
        \centering
            \tiny{Prompt: \textit{A rectangular sign that contains the logo for Aldi grocery stores can be seen on the side of a building. There are several trees rising up behind the building. A blue-colored wall of a building that has several lines going down its surface can be seen going across the bottom half of the image. There is a rectangular, white-colored sign hanging off of the wall. The sign has a red border going around its edges. There is a large letter “A” in the middle of the sign that is made out of curving blue and light blue stripes. There is a blue rectangle going across the bottom of the sign that has “ALDI” going across it in white lettering. There is a white-colored object on the right side of the wall that has a rounded red top. Several trees can be seen rising up from behind the building on the right hand side of the image. It is daytime, and the sky above has gray clouds floating across its surface.}}
        \end{minipage}
    }
    
    \caption{
        Additional visualizations of the main results on the DCI dataset, comparing FLUX and W8A8 DiT with W4A8 quantization.
    }
    \label{fig:app-dci-flux}
\end{figure}

\newcommand{\appimgwidth}{0.235\textwidth}
\newcommand{\appcapwidth}{0.96\textwidth}
\setlength{\fboxsep}{0pt}
\setlength{\fboxrule}{0.2pt}
\begin{figure}[h!]
    \centering

    \begin{minipage}[t]{0.5\textwidth}
        \centering
        \fbox{\includegraphics[width=\appimgwidth]{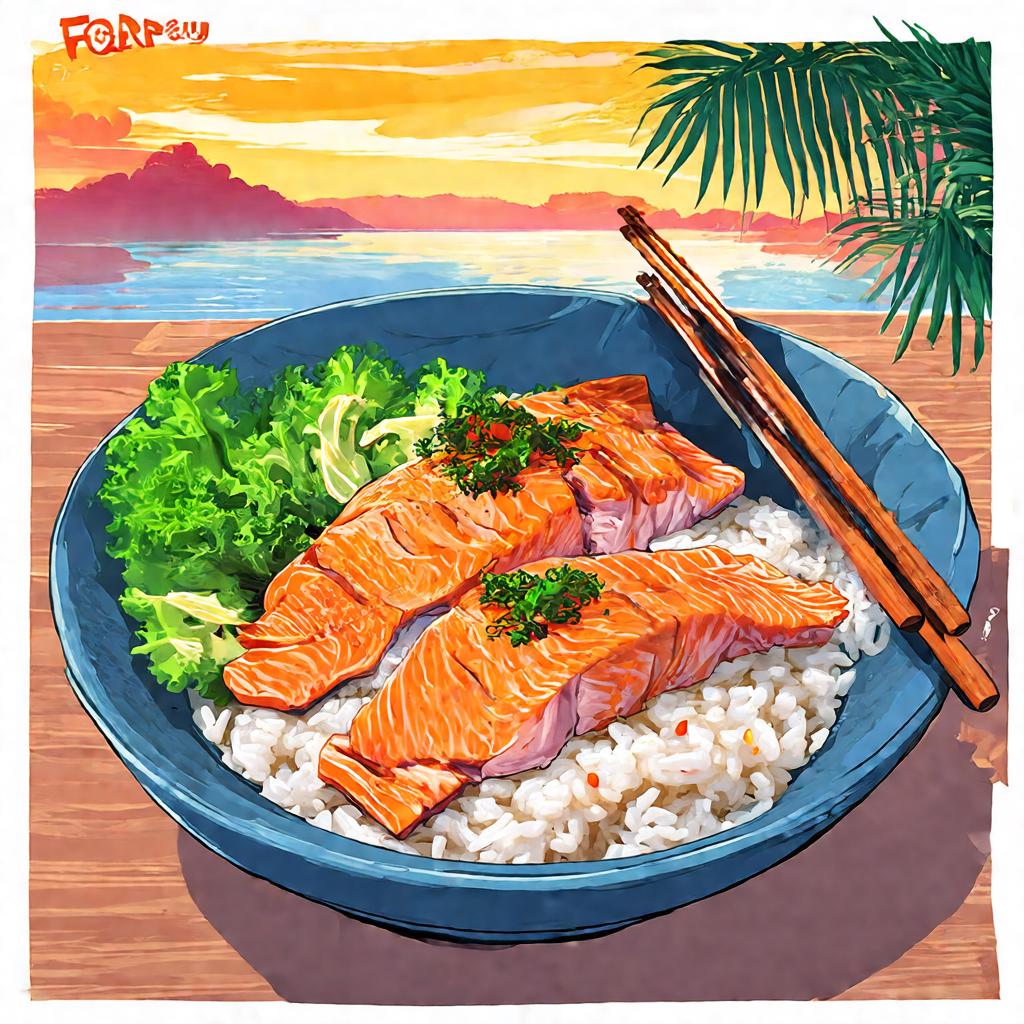}}
        \fbox{\includegraphics[width=\appimgwidth]{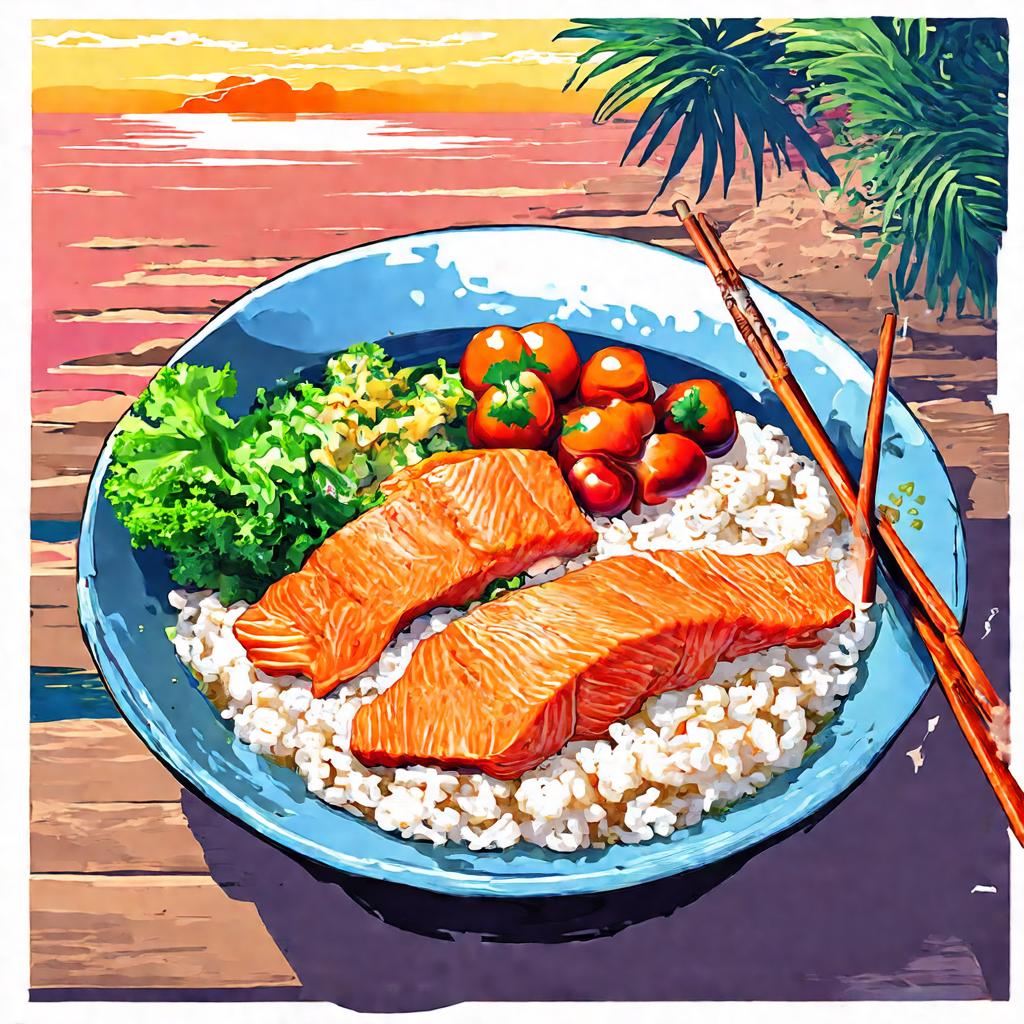}}
        \fbox{\includegraphics[width=\appimgwidth]{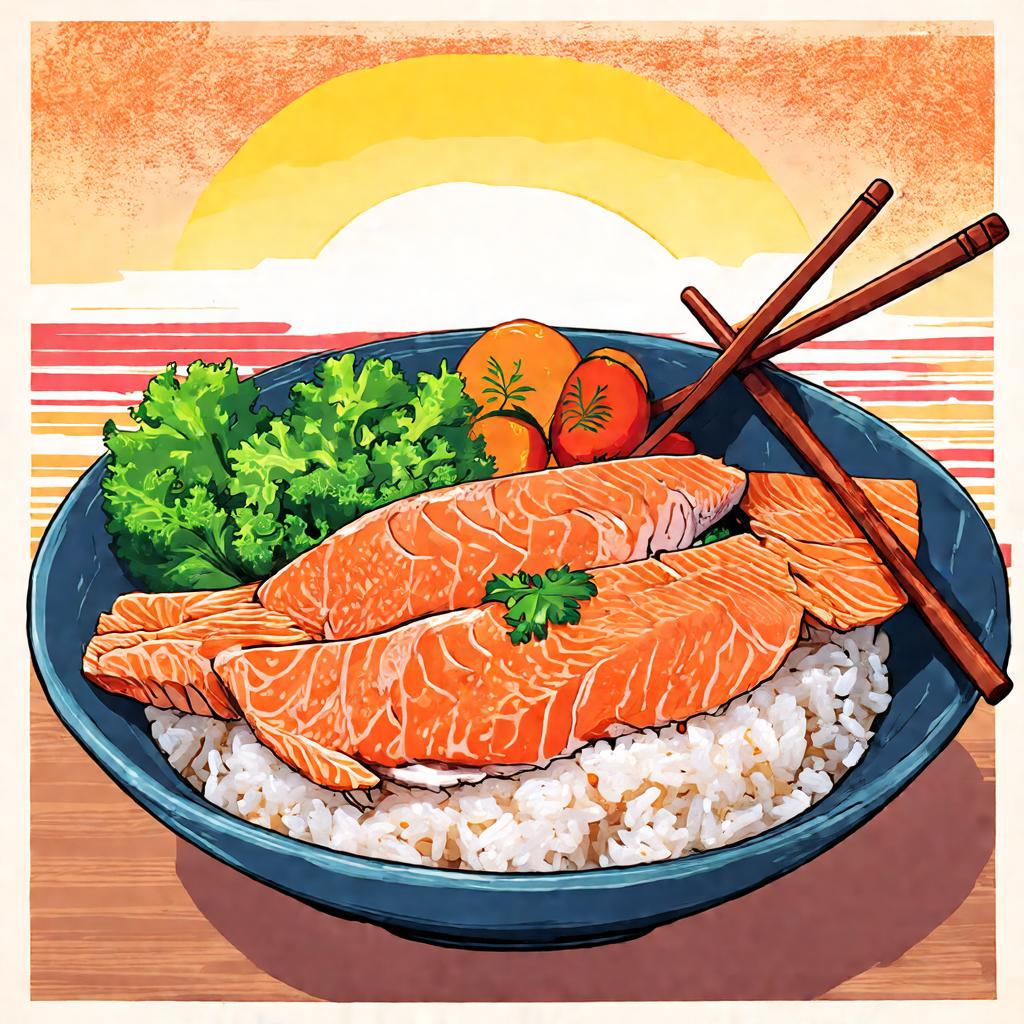}}
        \fbox{\includegraphics[width=\appimgwidth]{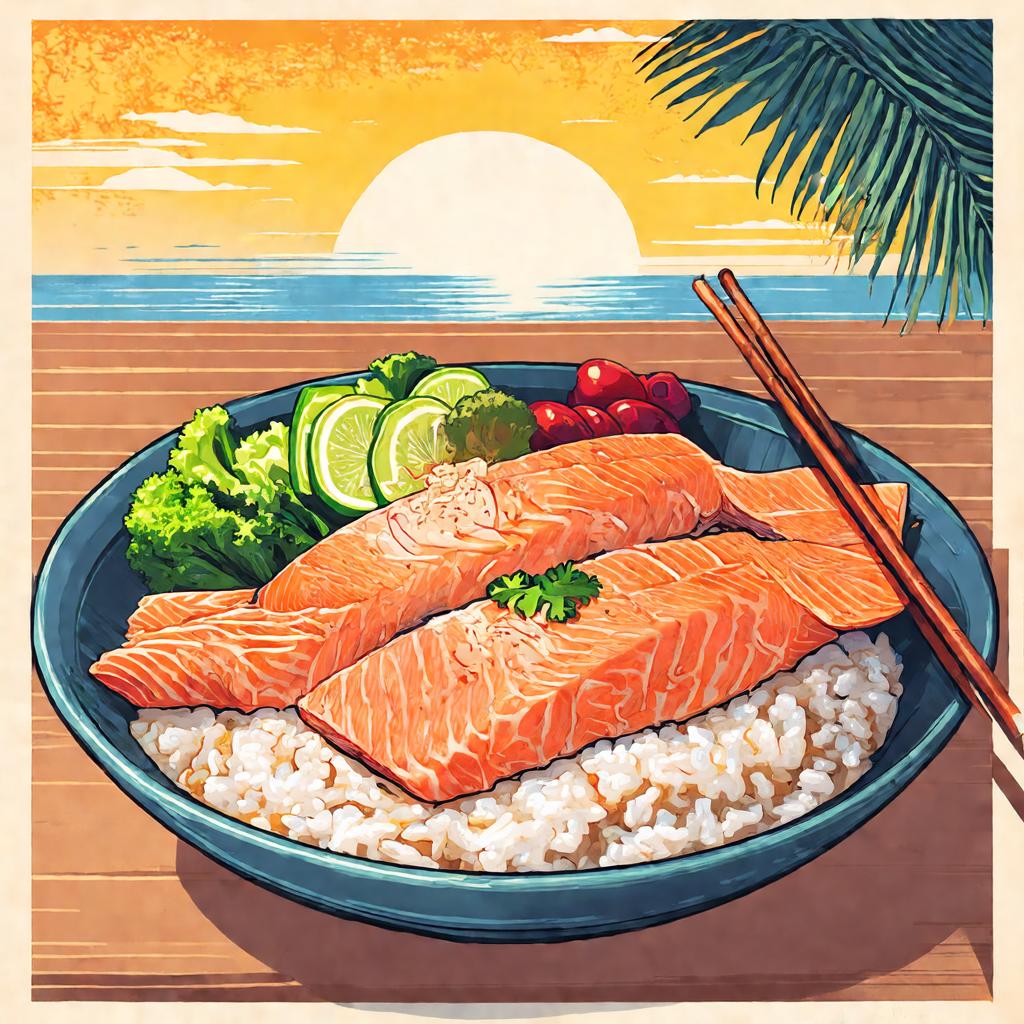}}\\[0.5ex]
        \vspace{-8pt}
        \caption*{
            \begin{minipage}{\appcapwidth}
            \centering
                \tiny{Prompt: \textit{cartoon, vintage poster look, poke bowl, salmon, with fresh vegetables, with a bowl of rice, hand drawn, tropical sunset, chopsticks, 4k}}
            \end{minipage}
        }
    \end{minipage}%
    \begin{minipage}[t]{0.5\textwidth}
        \centering
        \fbox{\includegraphics[width=\appimgwidth]{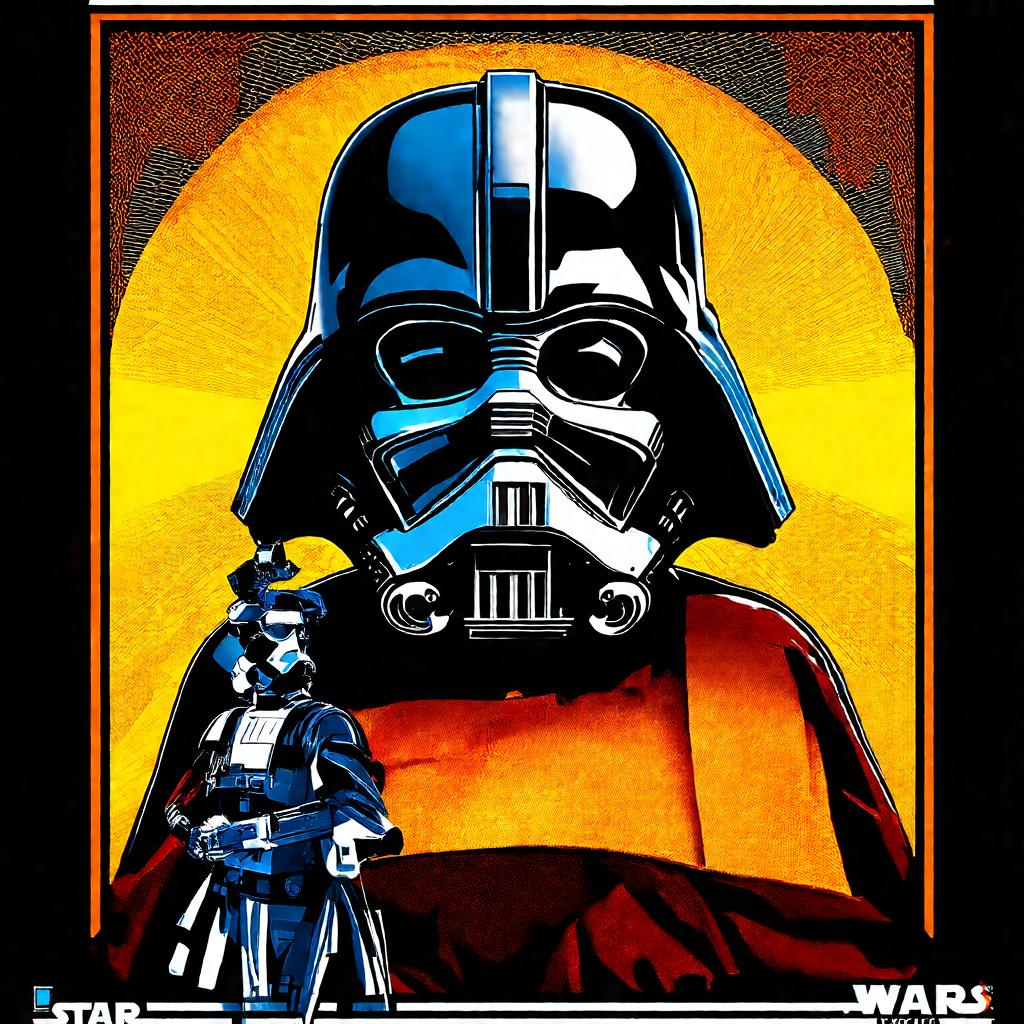}}
        \fbox{\includegraphics[width=\appimgwidth]{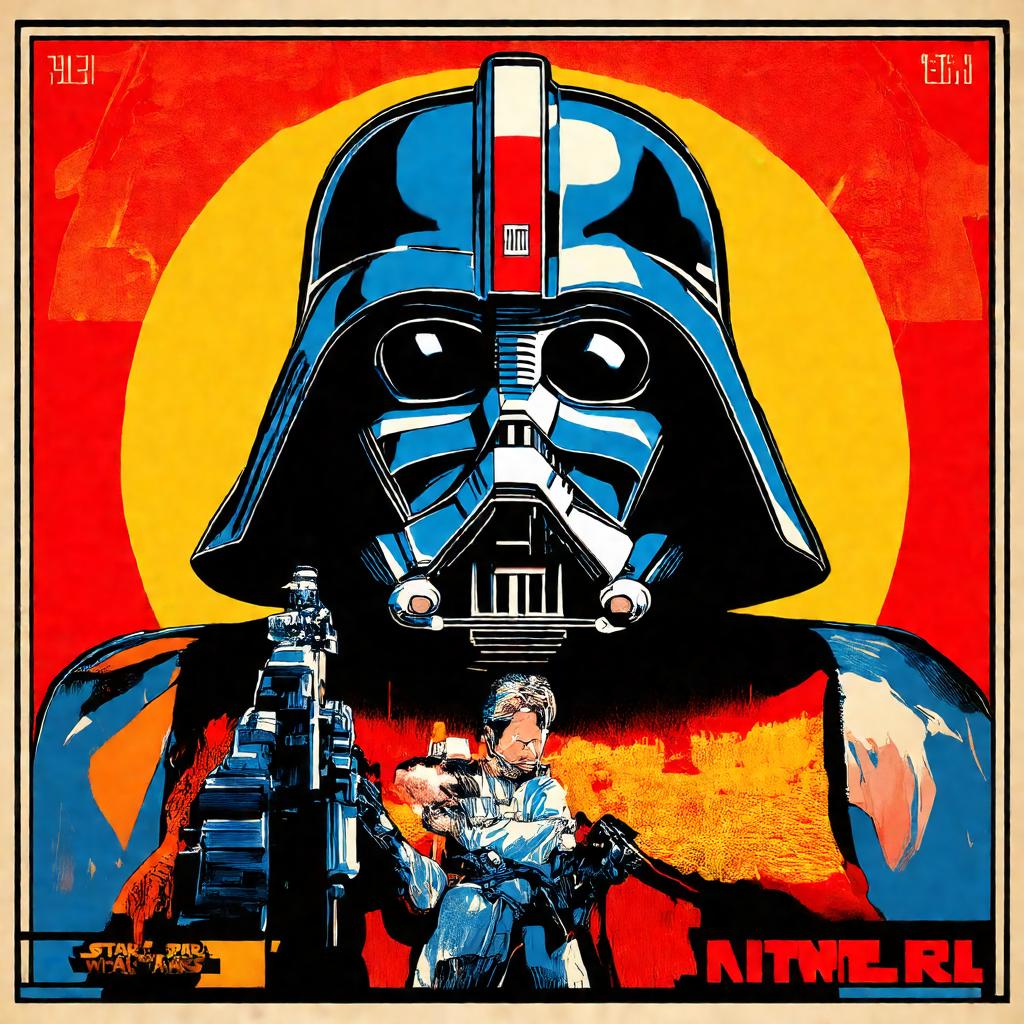}}
        \fbox{\includegraphics[width=\appimgwidth]{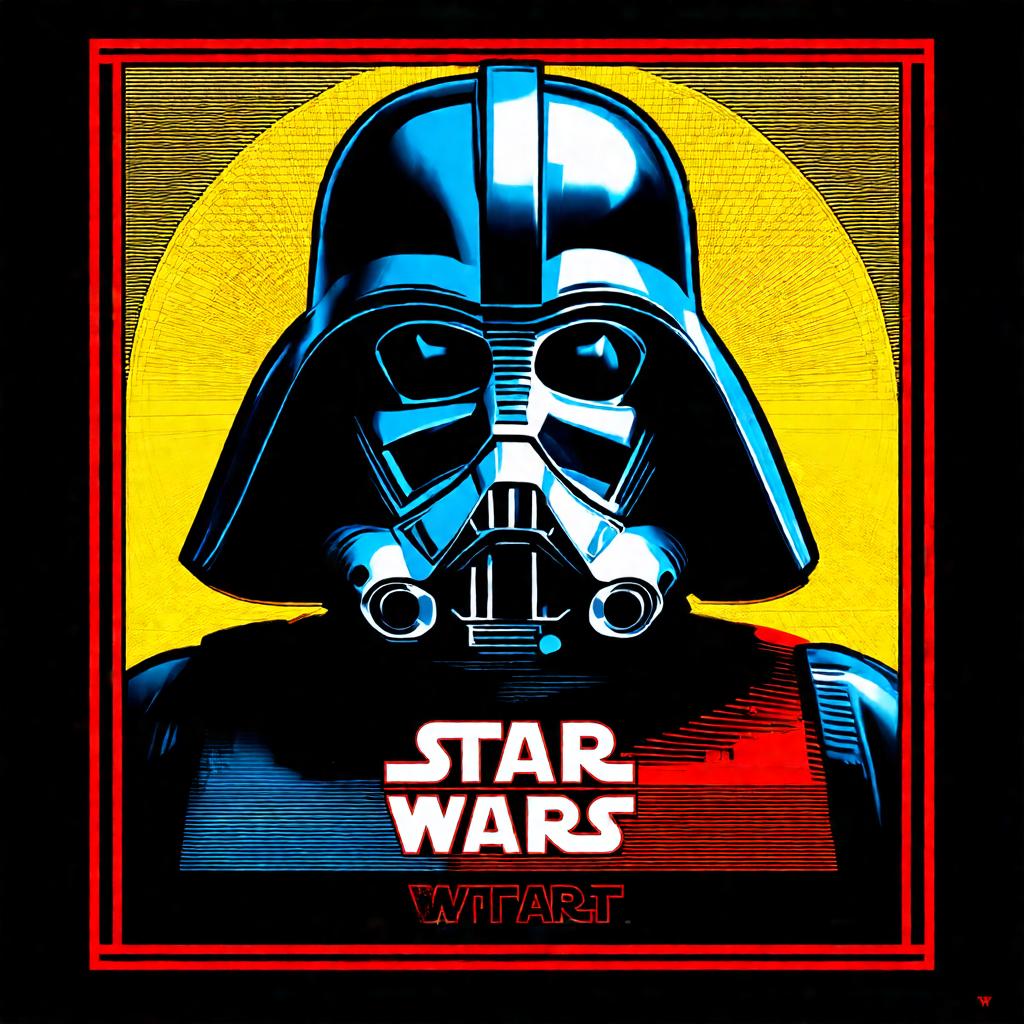}}
        \fbox{\includegraphics[width=\appimgwidth]{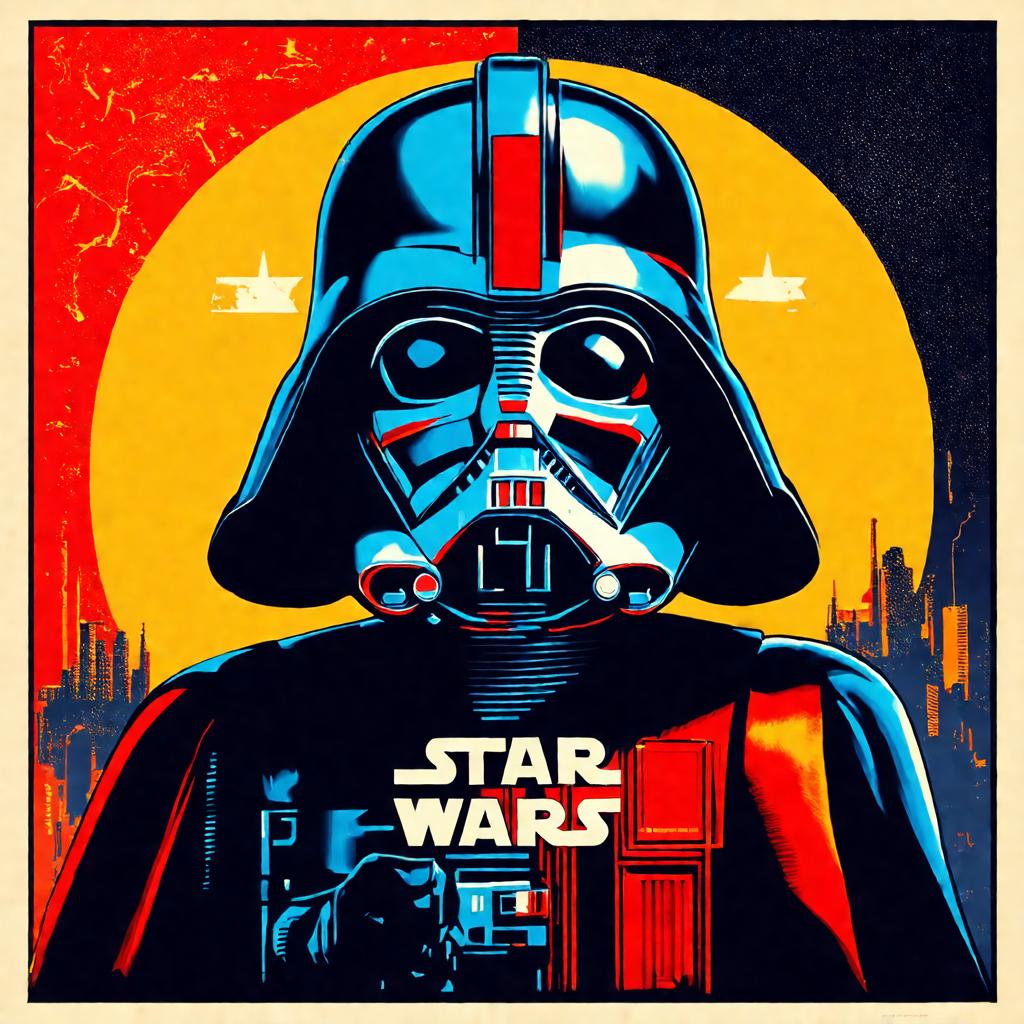}}\\[0.5ex]
        \vspace{-8pt}
        \caption*{
            \begin{minipage}{\appcapwidth}
            \centering
                \tiny{Prompt: \textit{Star Wars poster design by Obey, retro 1980 street color propaganda style, arrangement with golden ratio, high quality render texture}}
            \end{minipage}
        }
    \end{minipage}

    \begin{minipage}[t]{0.5\textwidth}
        \centering
        \fbox{\includegraphics[width=\appimgwidth]{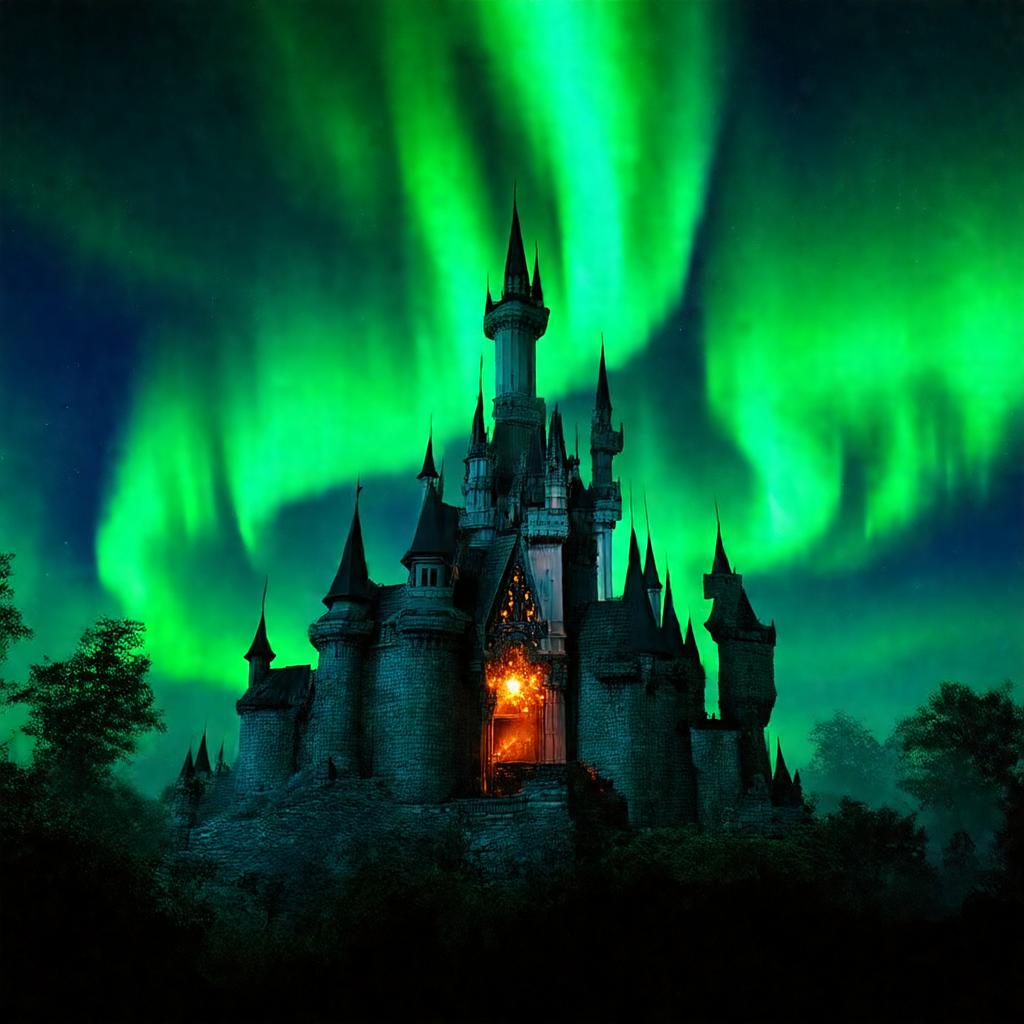}}
        \fbox{\includegraphics[width=\appimgwidth]{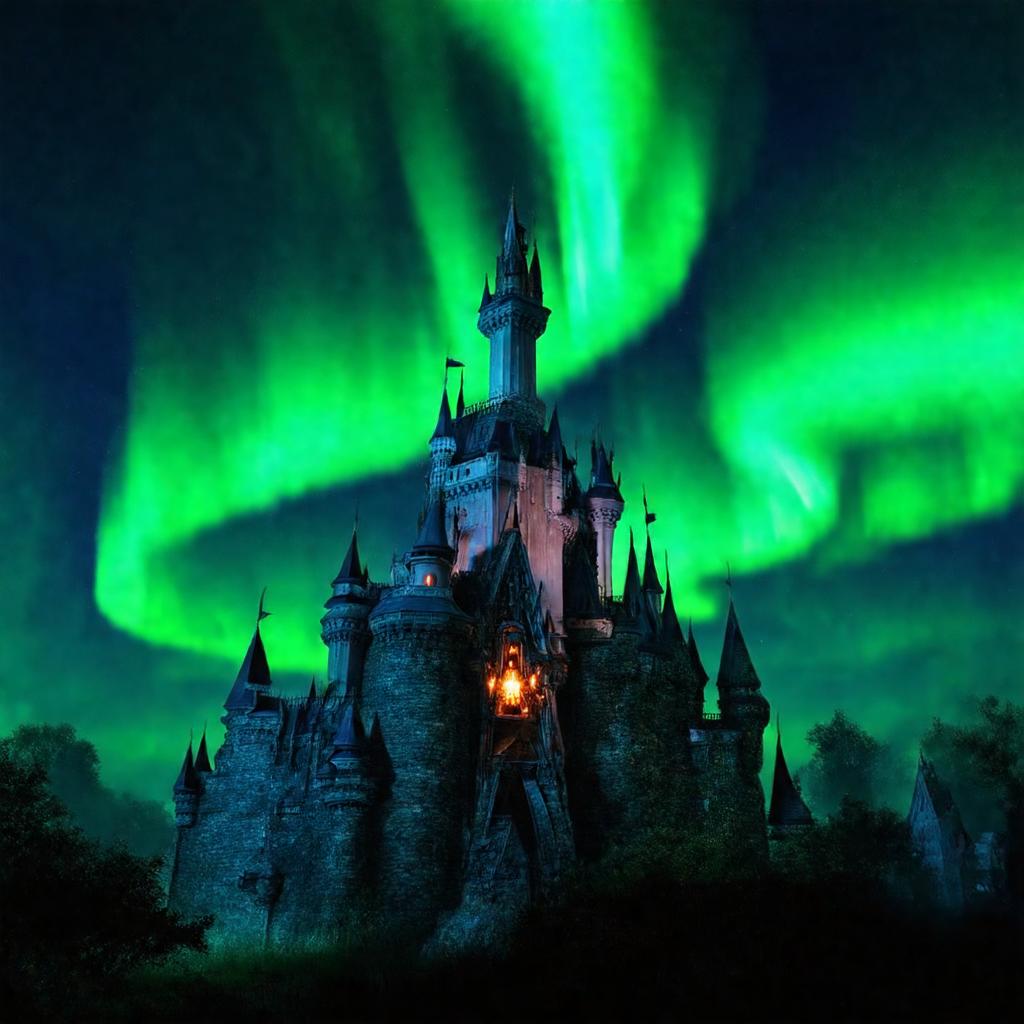}}
        \fbox{\includegraphics[width=\appimgwidth]{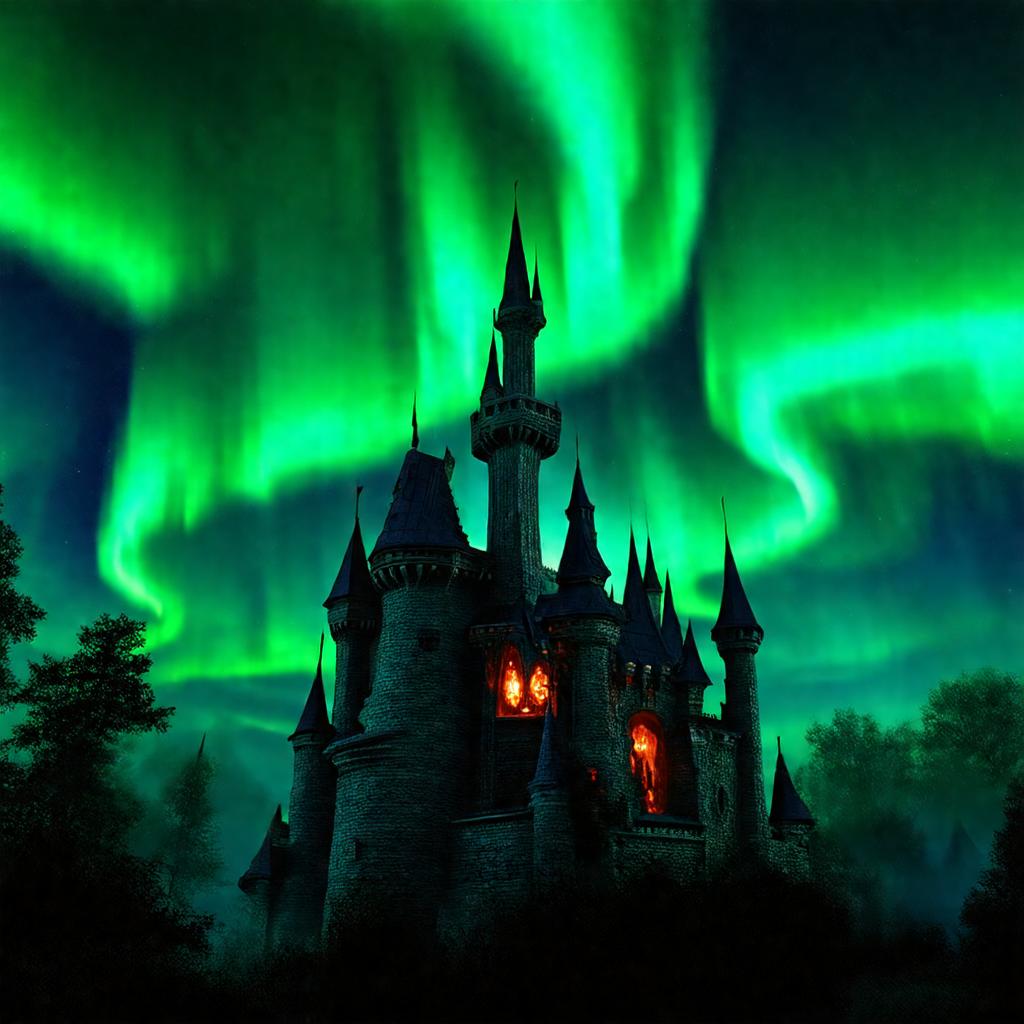}}
        \fbox{\includegraphics[width=\appimgwidth]{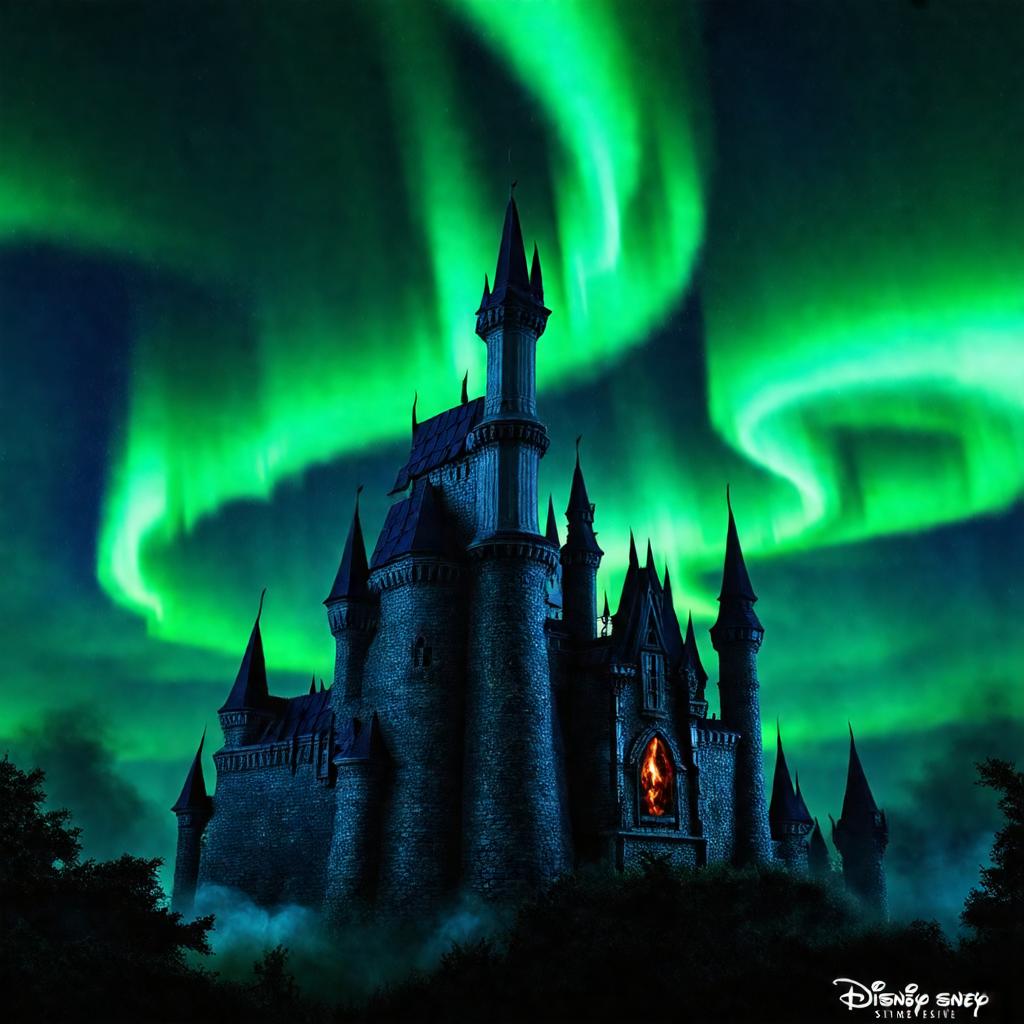}}\\[0.5ex]
        \vspace{-8pt}
        \caption*{
            \begin{minipage}{\appcapwidth}
            \centering
                \tiny{Prompt: \textit{Castle made of smoke, castle, Evil, scary, dark, ghostly, green aurora borealis in the sky, spooky, intricate detail, high resolution, atmospheric, 4K HDR, Film Still, Hyperrealistic, Disney movie}}
            \end{minipage}
        }
    \end{minipage}%
    \begin{minipage}[t]{0.5\textwidth}
        \centering
        \fbox{\includegraphics[width=\appimgwidth]{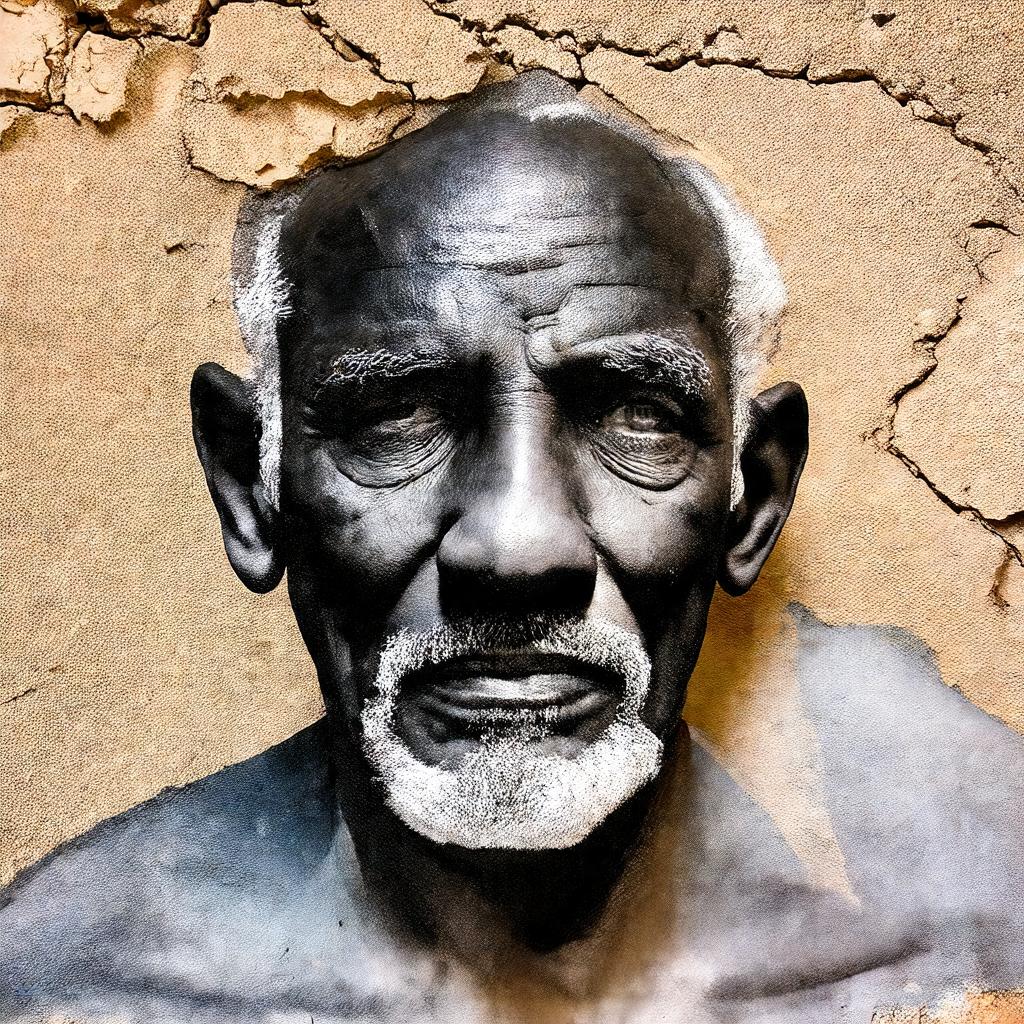}}
        \fbox{\includegraphics[width=\appimgwidth]{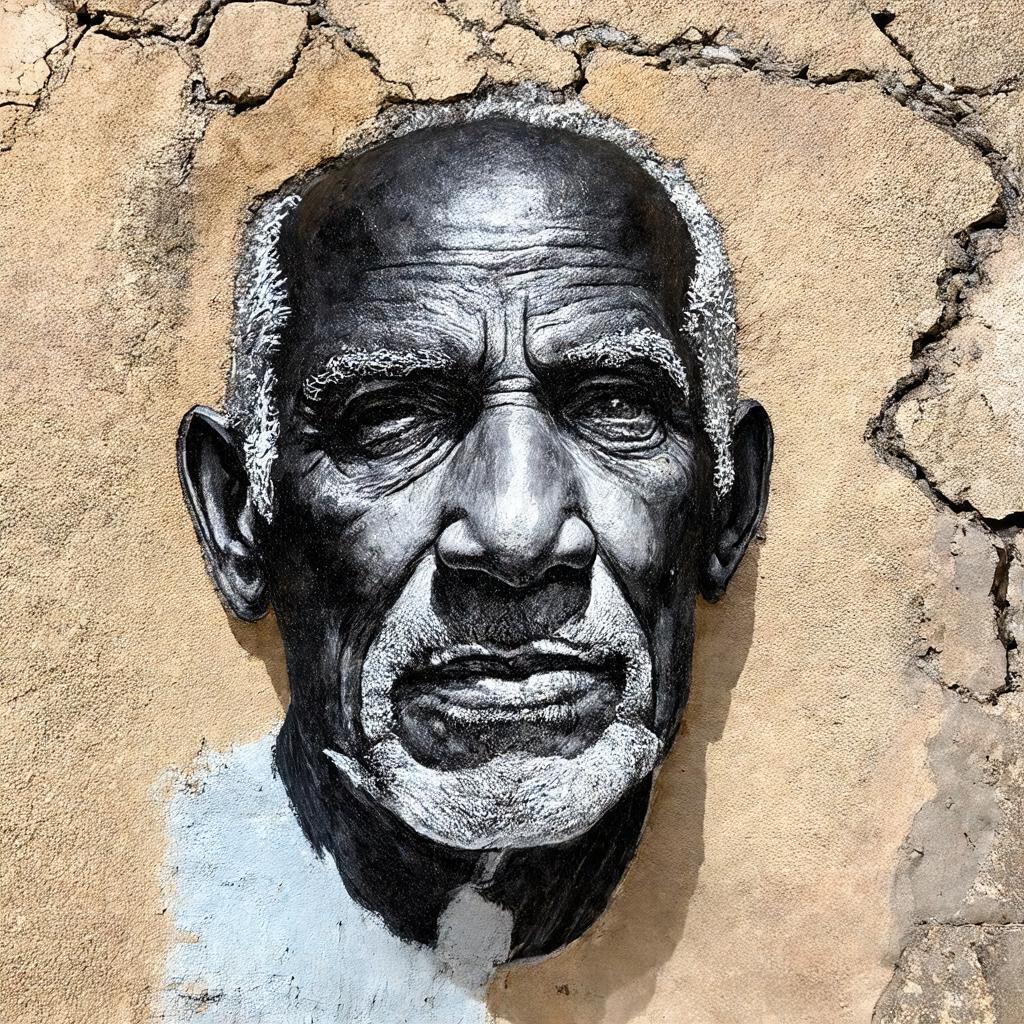}}
        \fbox{\includegraphics[width=\appimgwidth]{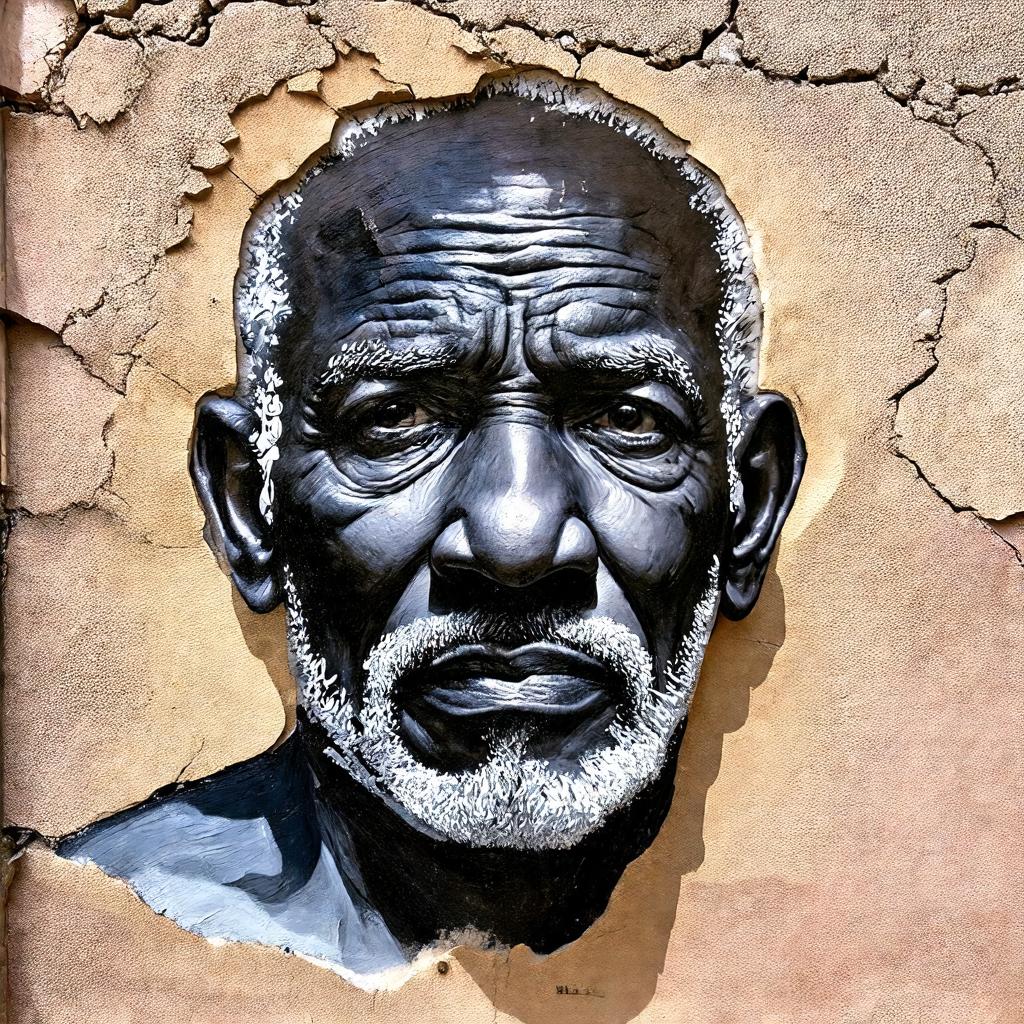}}
        \fbox{\includegraphics[width=\appimgwidth]{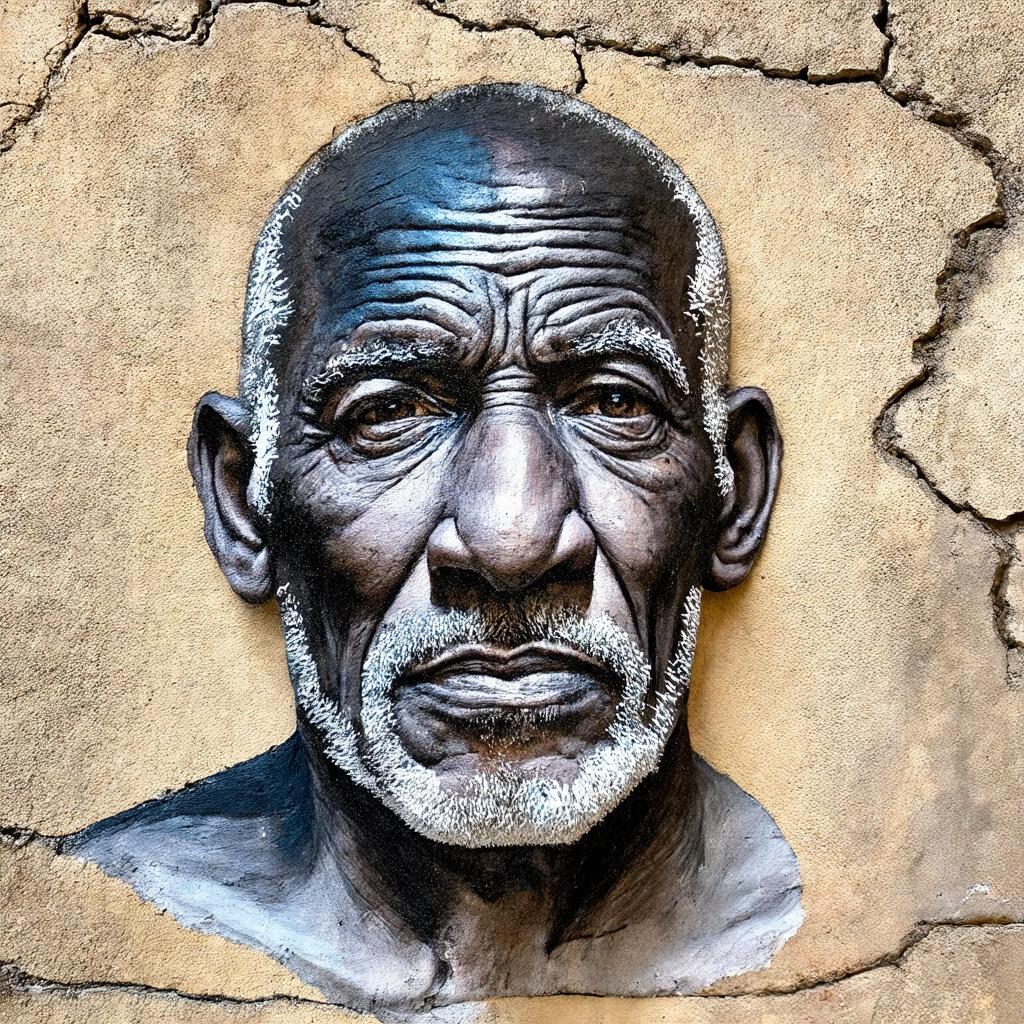}}\\[0.5ex]
        \vspace{-8pt}
        \caption*{
            \begin{minipage}{\appcapwidth}
            \centering
                \tiny{Prompt: \textit{an old black mans face painted on a wall with some cracks, in the style of topographical realism, grit and grain, dorothea tanning, matte photo, stone, symbolic images, grungy patchwork, wide angle view}}
            \end{minipage}
        }
    \end{minipage}
    
    \begin{minipage}[t]{0.5\textwidth}
        \centering
        \fbox{\includegraphics[width=\appimgwidth]{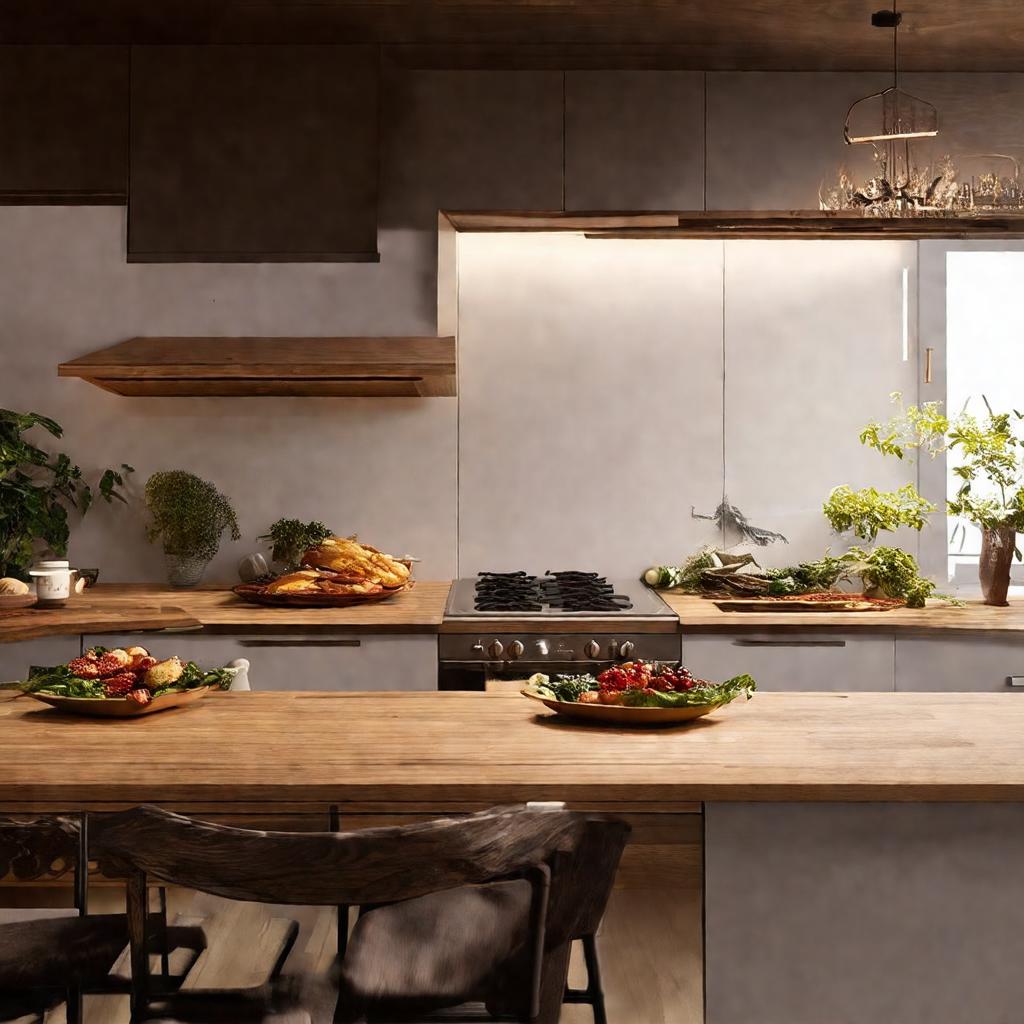}}
        \fbox{\includegraphics[width=\appimgwidth]{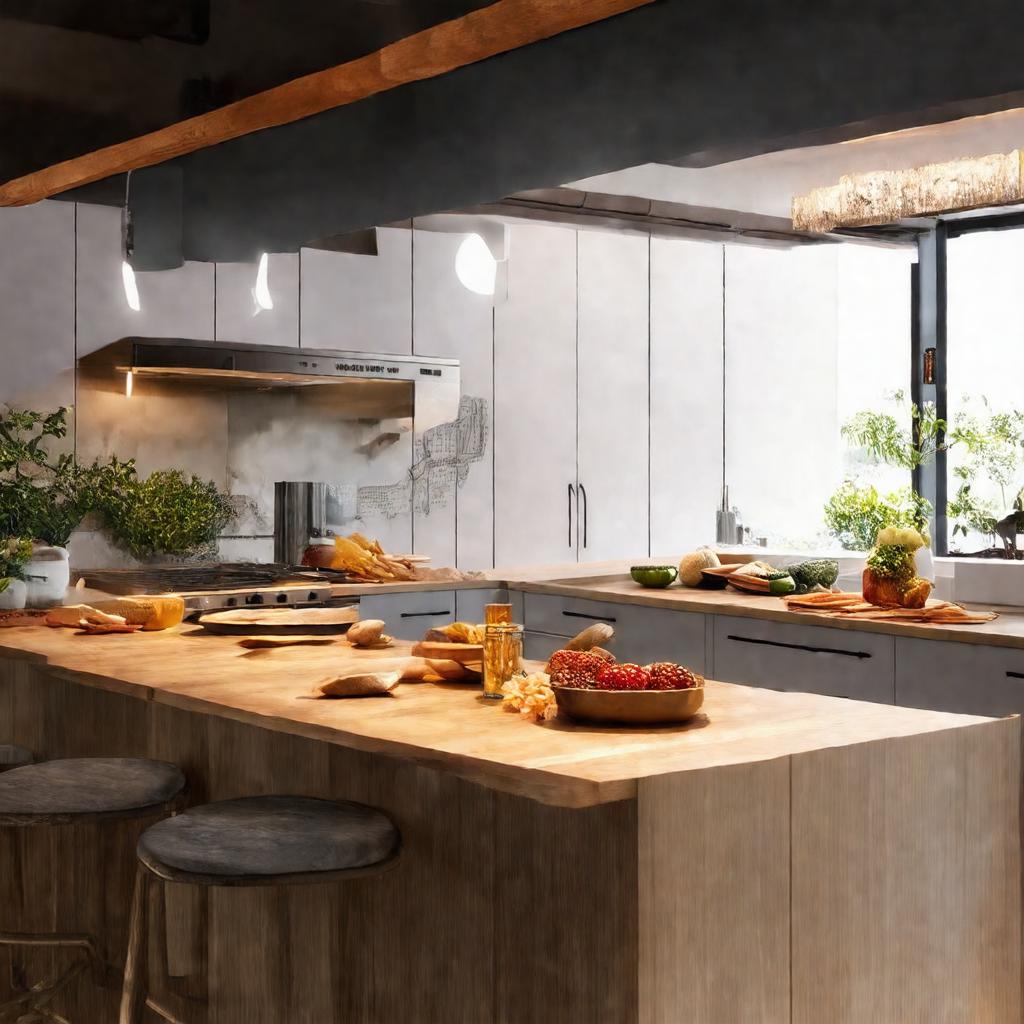}}
        \fbox{\includegraphics[width=\appimgwidth]{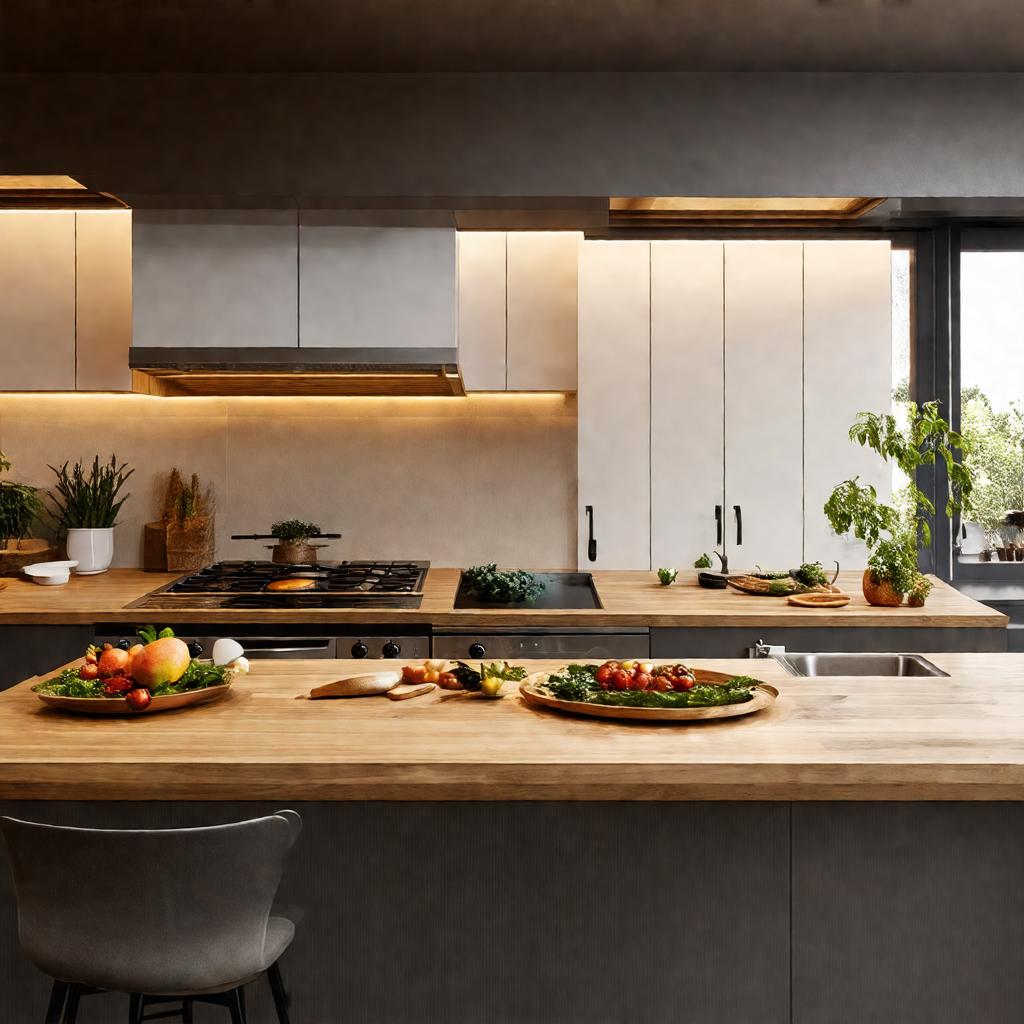}}
        \fbox{\includegraphics[width=\appimgwidth]{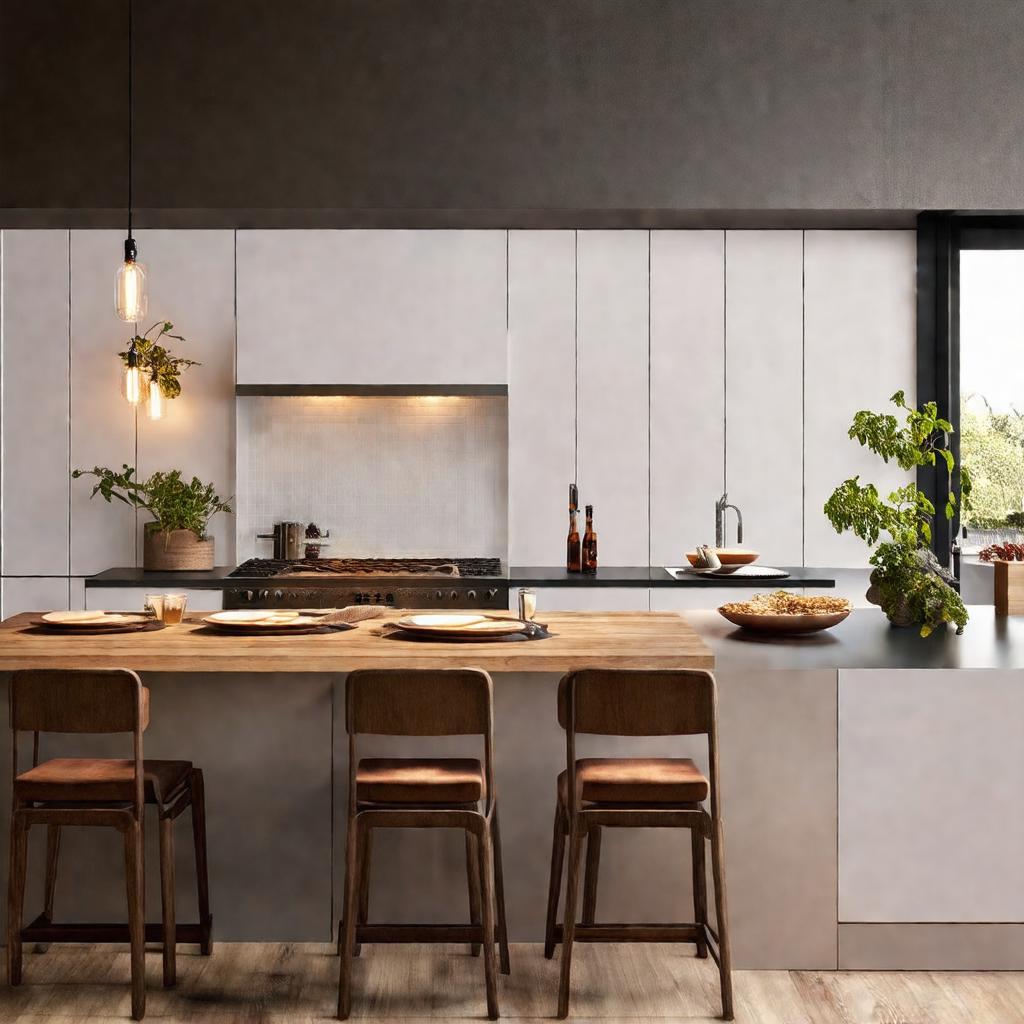}}\\[0.5ex]
        \vspace{-8pt}
        \caption*{
            \begin{minipage}{\appcapwidth}
            \centering
                \tiny{Prompt: \textit{kitchen design thinking studio, cinematic lighting, phorealistic, 8k}}
            \end{minipage}
        }
    \end{minipage}%
    \begin{minipage}[t]{0.5\textwidth}
        \centering
        \fbox{\includegraphics[width=\appimgwidth]{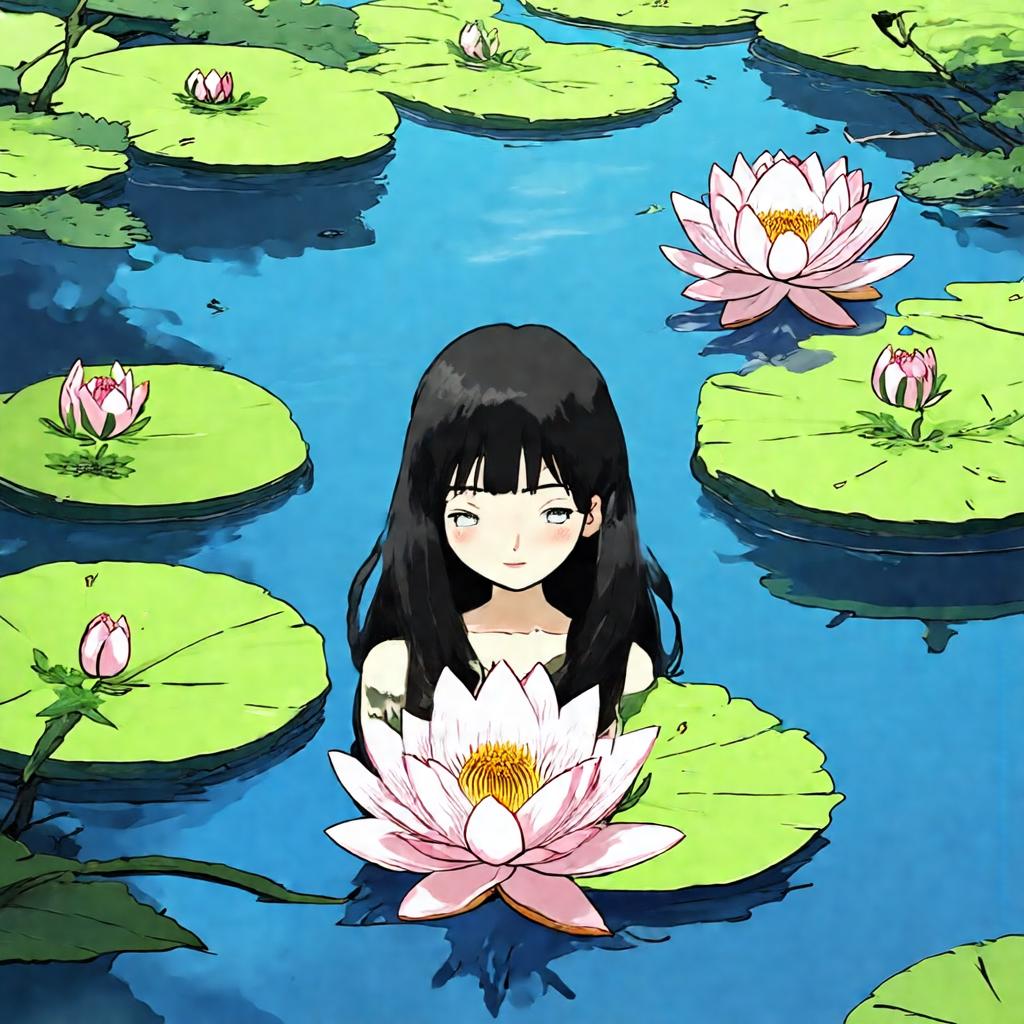}}
        \fbox{\includegraphics[width=\appimgwidth]{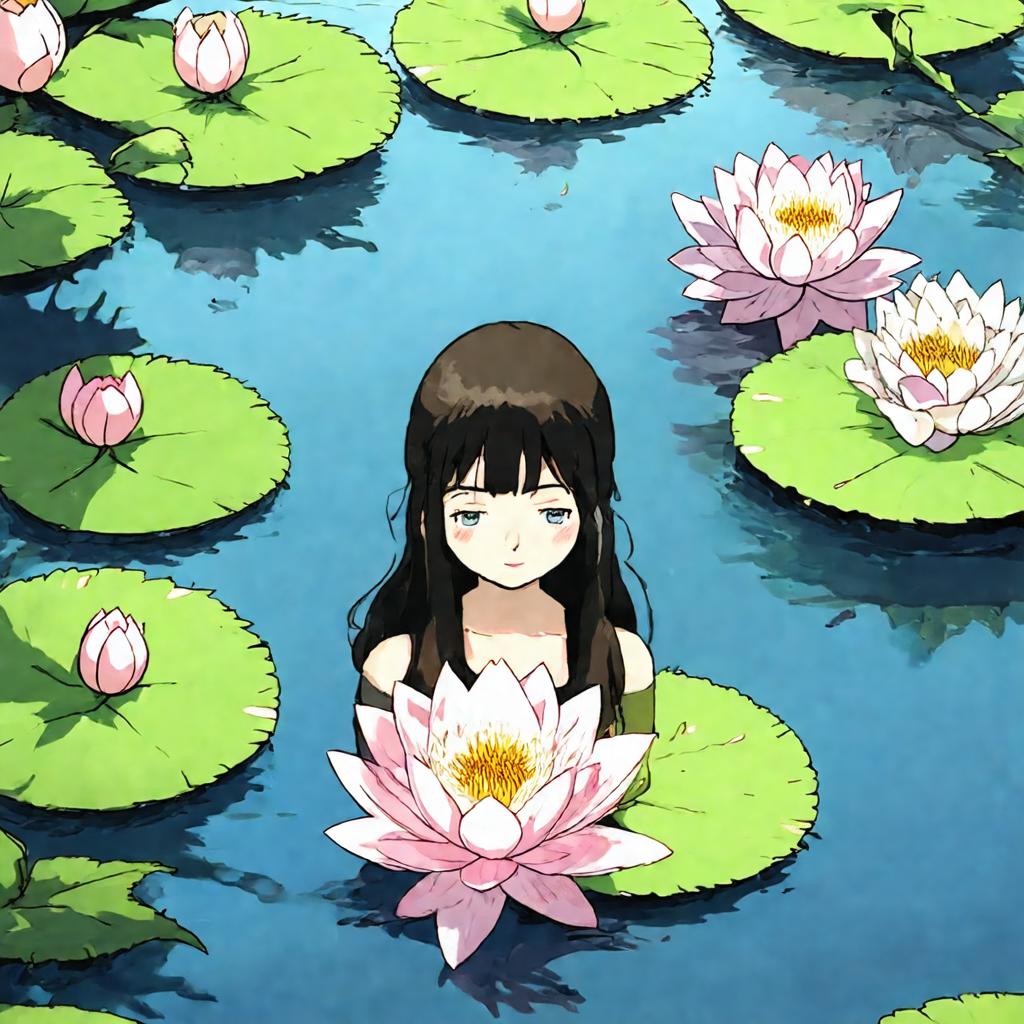}}
        \fbox{\includegraphics[width=\appimgwidth]{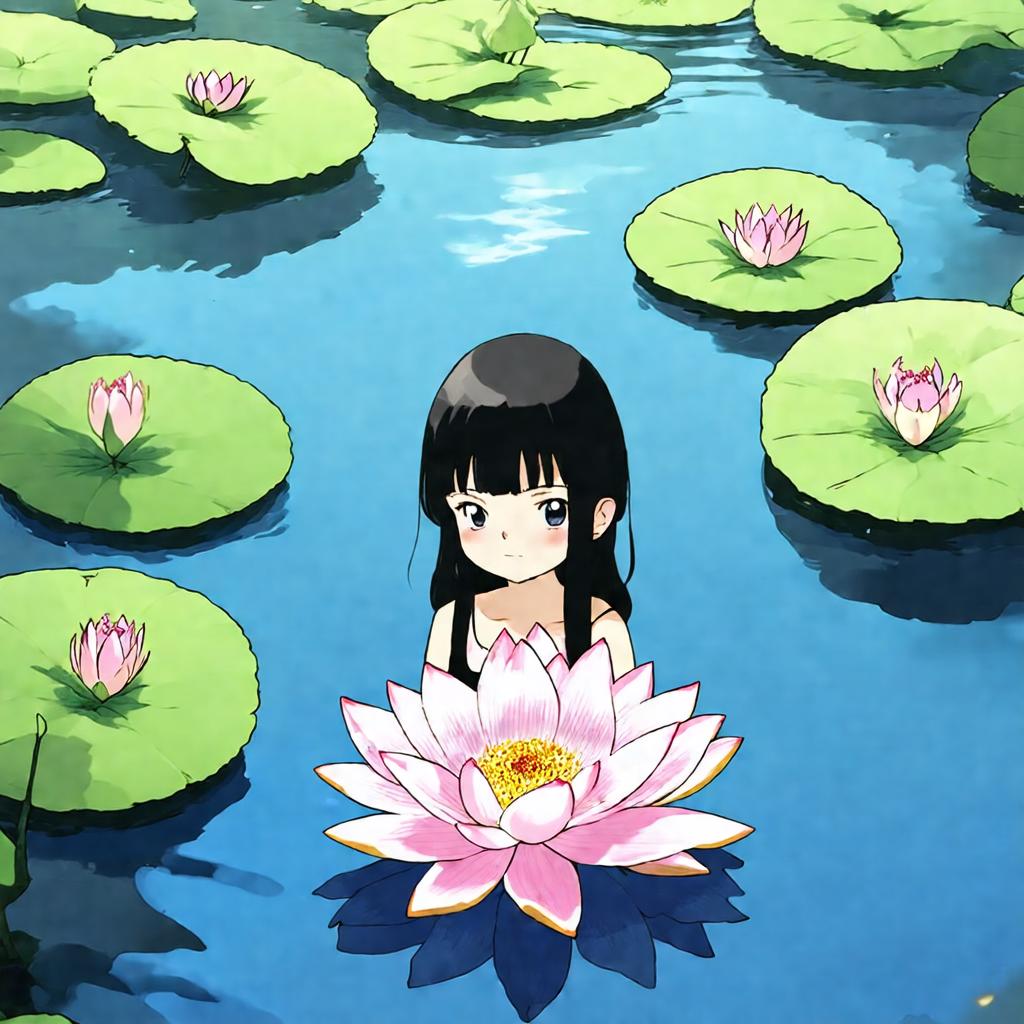}}
        \fbox{\includegraphics[width=\appimgwidth]{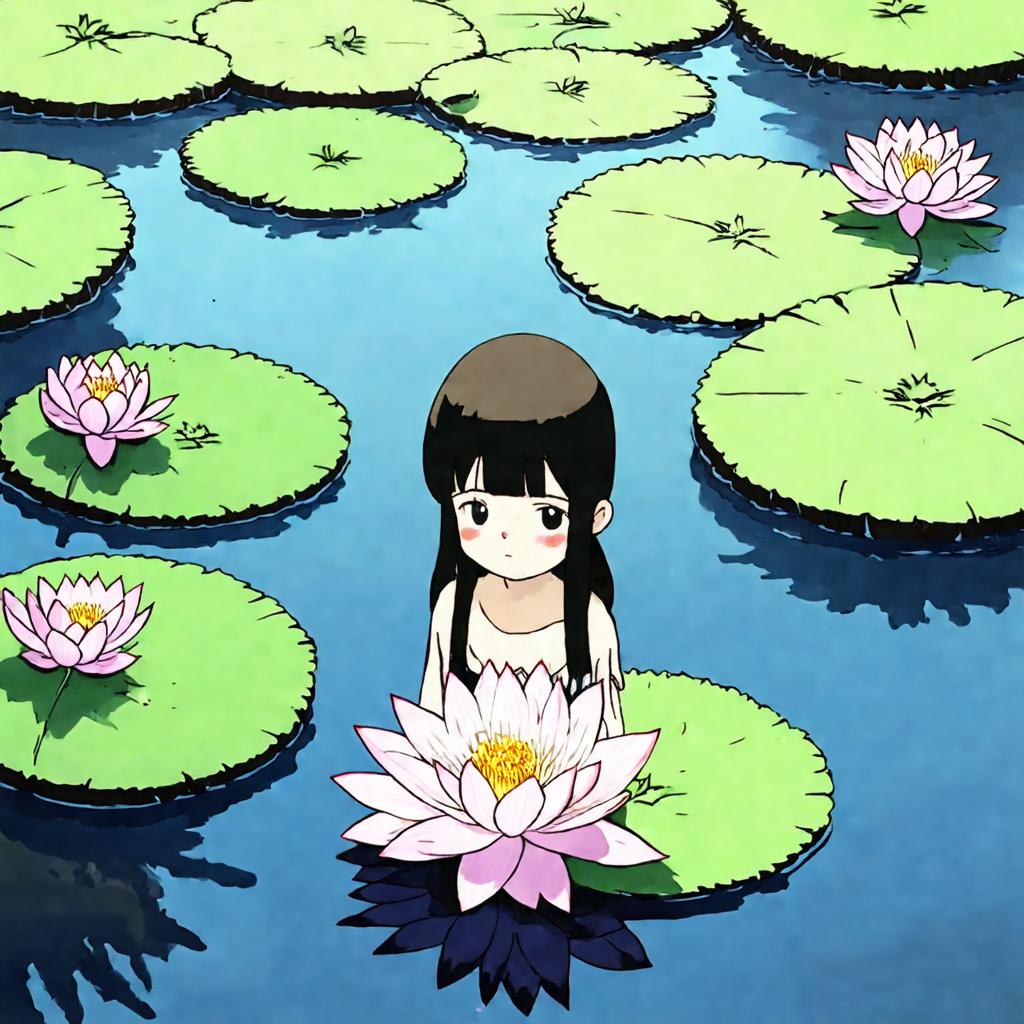}}\\[0.5ex]
        \vspace{-8pt}
        \caption*{
            \begin{minipage}{\appcapwidth}
            \centering
                \tiny{Prompt: \textit{lotus blossoms ghibli style animation}}
            \end{minipage}
        }
    \end{minipage}

    \begin{minipage}[t]{0.5\textwidth}
        \centering
        \fbox{\includegraphics[width=\appimgwidth]{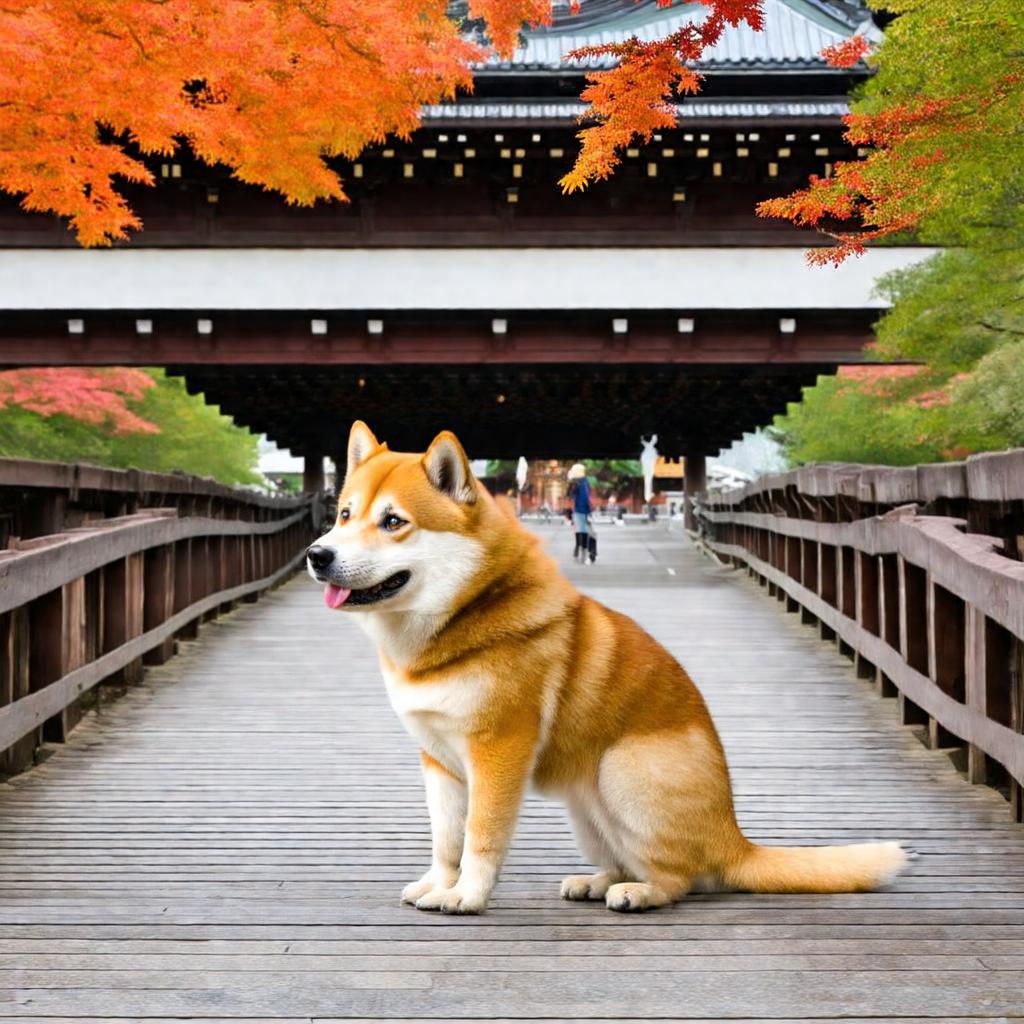}}
        \fbox{\includegraphics[width=\appimgwidth]{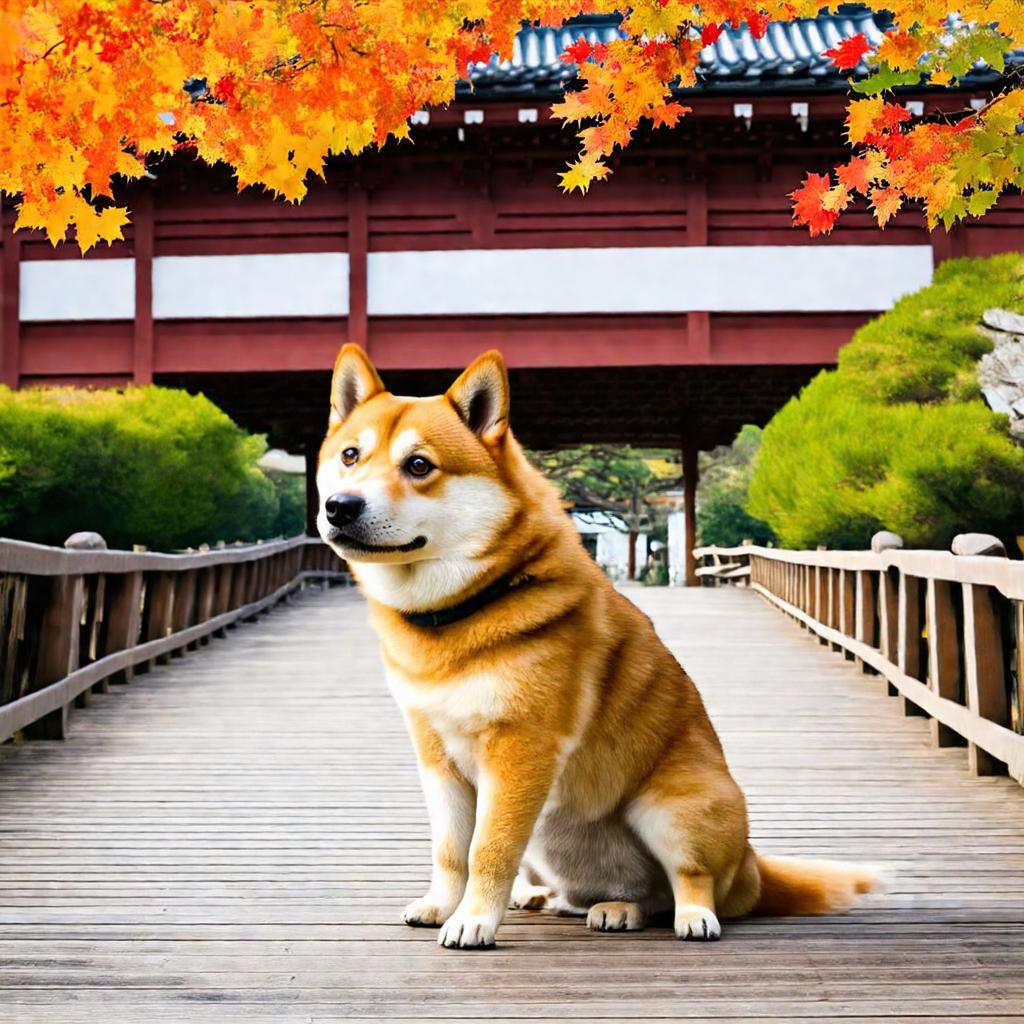}}
        \fbox{\includegraphics[width=\appimgwidth]{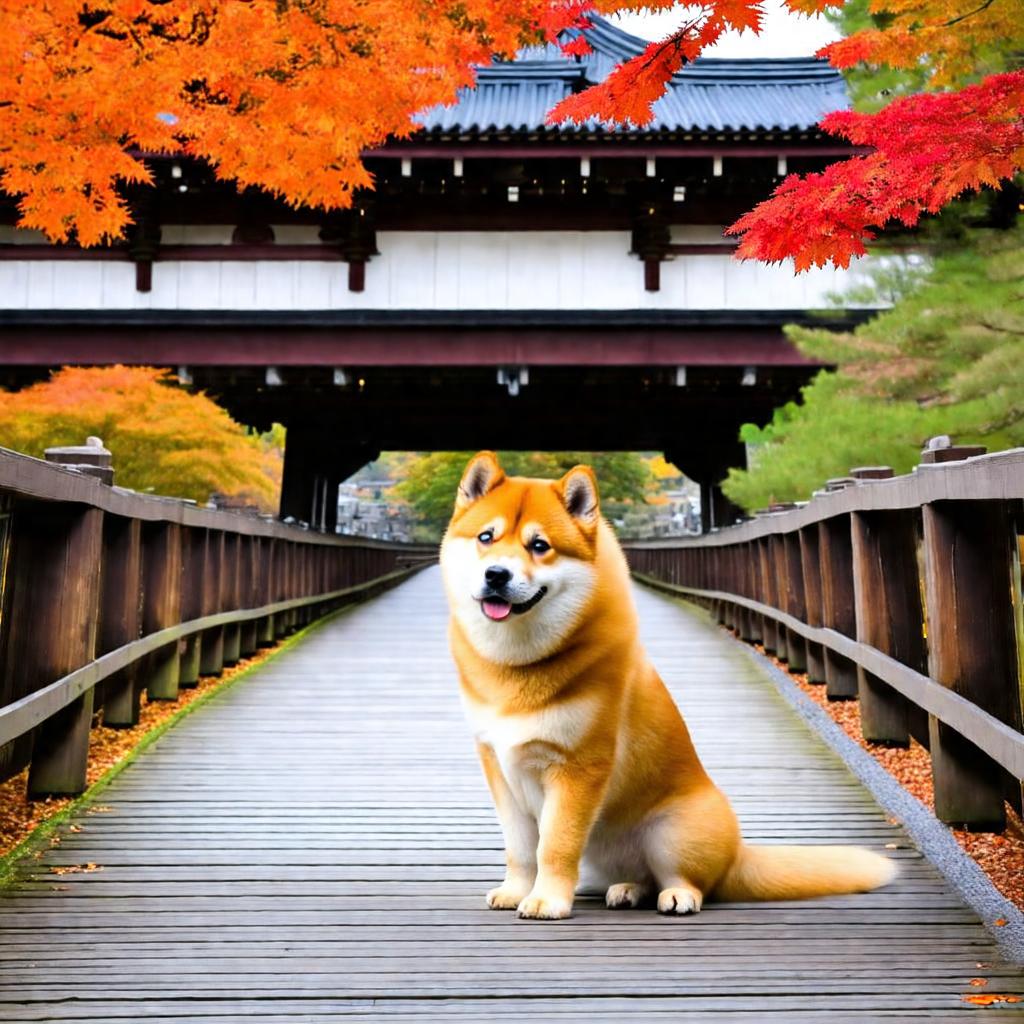}}
        \fbox{\includegraphics[width=\appimgwidth]{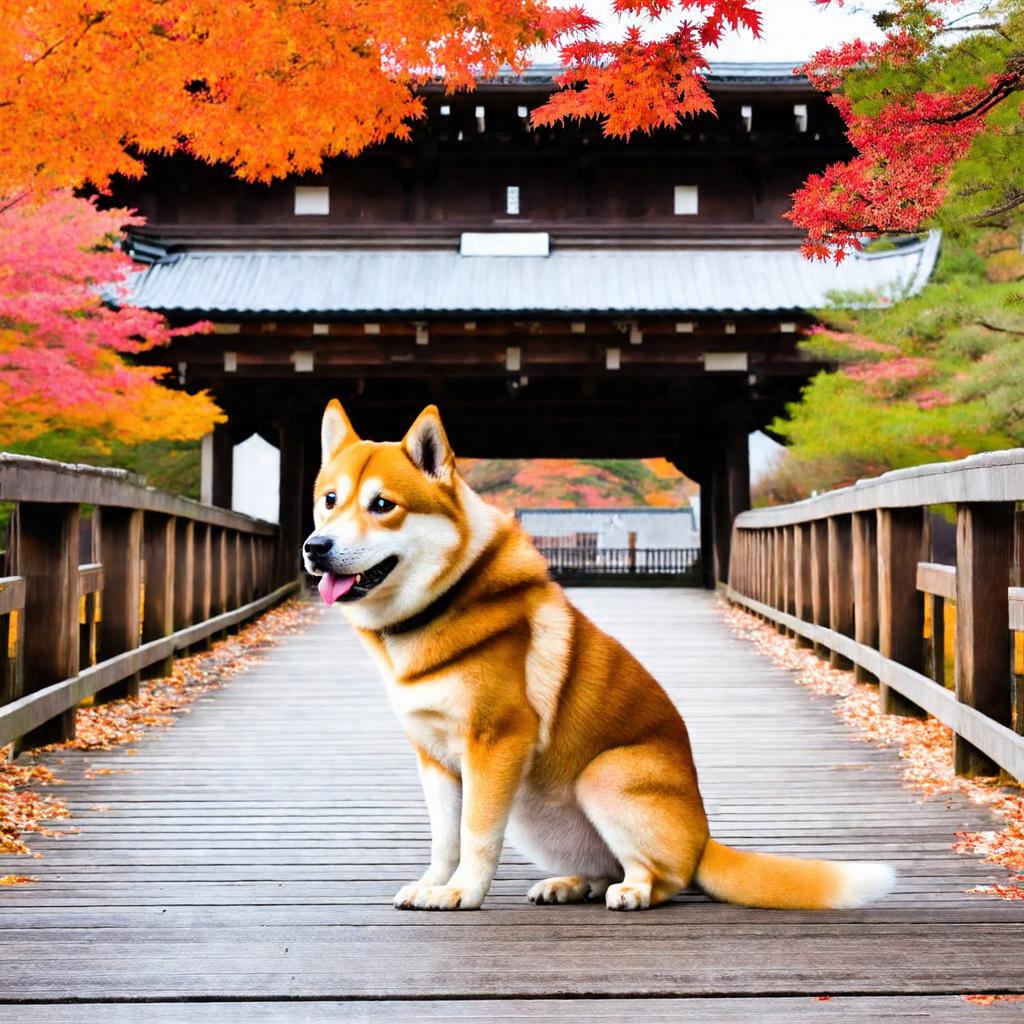}}\\[0.5ex]
        \vspace{-8pt}
        \caption*{
            \begin{minipage}{\appcapwidth}
            \centering
                \tiny{Prompt: \textit{shibainu, dog, japan, dog on the bridge, behind of bridge exists japanese temple, leaves changing color, ziburi style}}
            \end{minipage}
        }
    \end{minipage}%
    \begin{minipage}[t]{0.5\textwidth}
        \centering
        \fbox{\includegraphics[width=\appimgwidth]{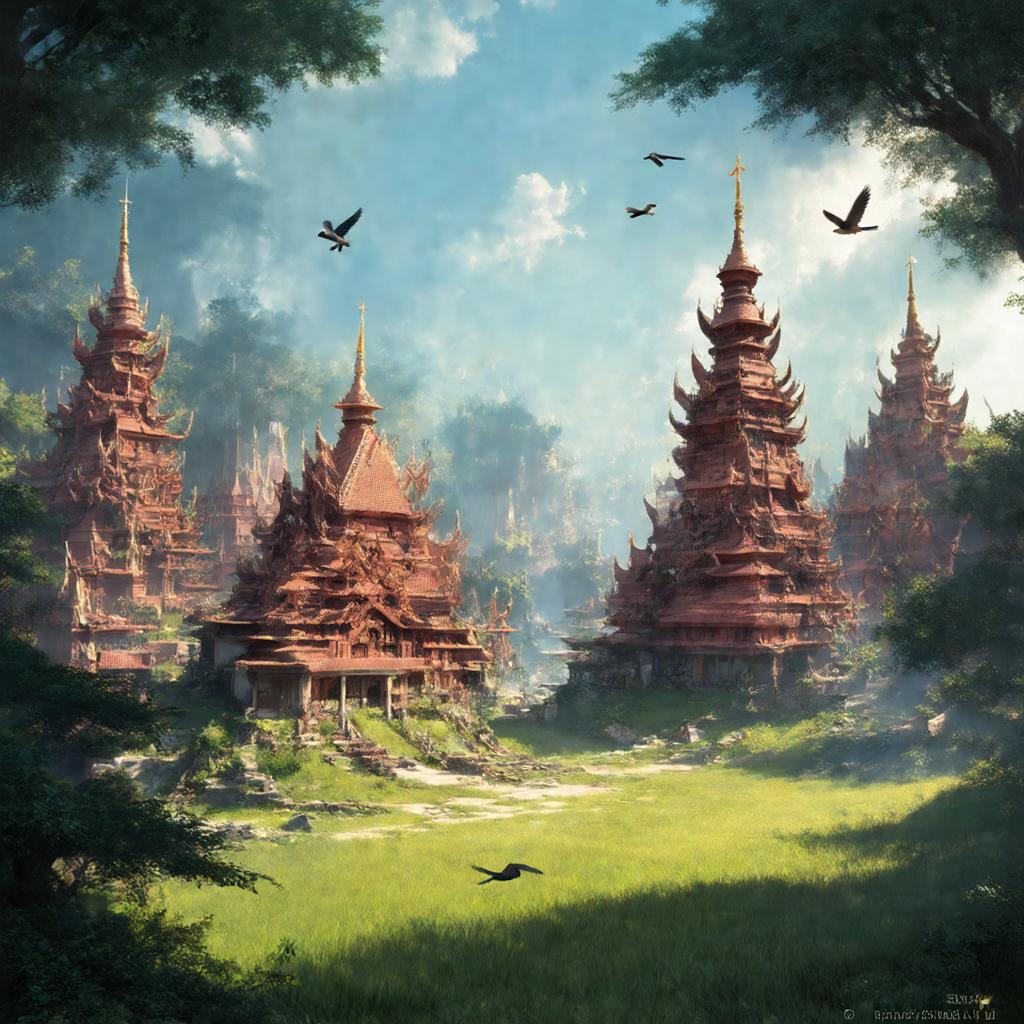}}
        \fbox{\includegraphics[width=\appimgwidth]{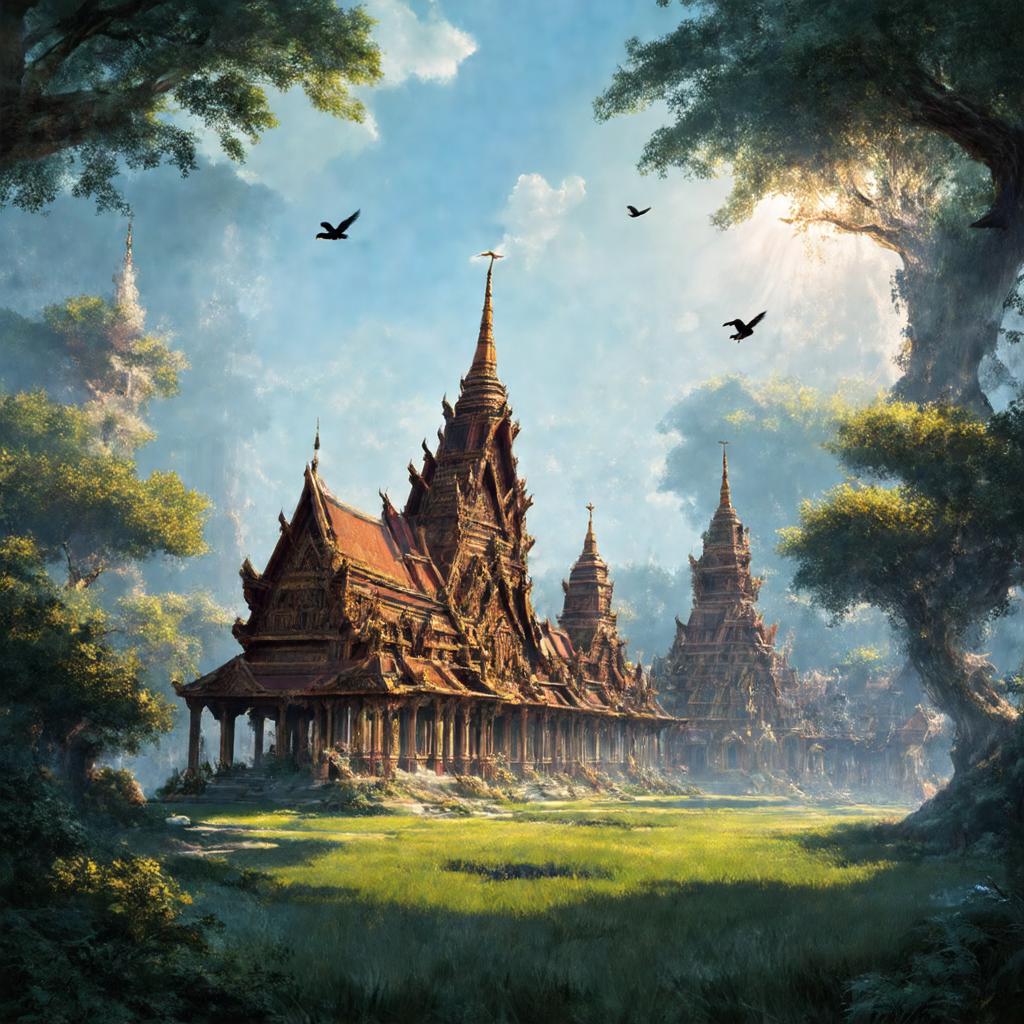}}
        \fbox{\includegraphics[width=\appimgwidth]{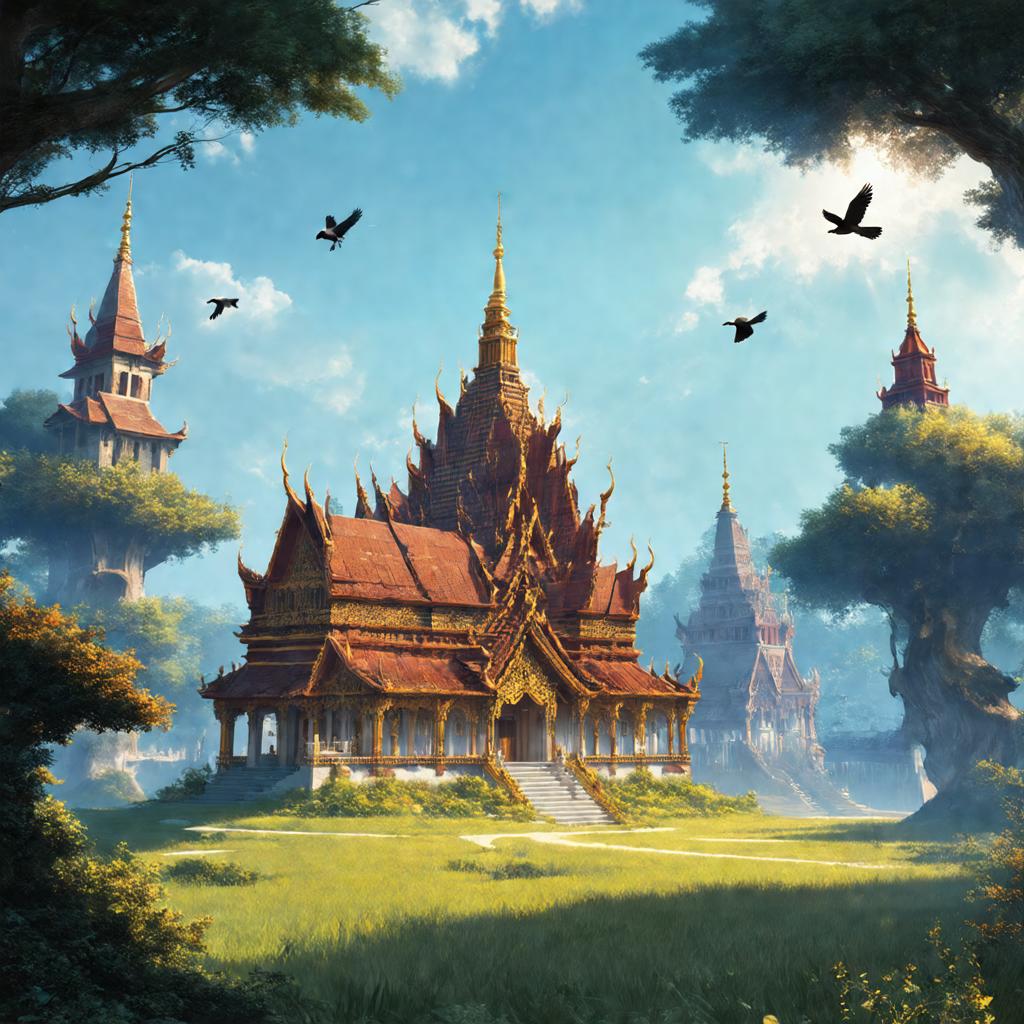}}
        \fbox{\includegraphics[width=\appimgwidth]{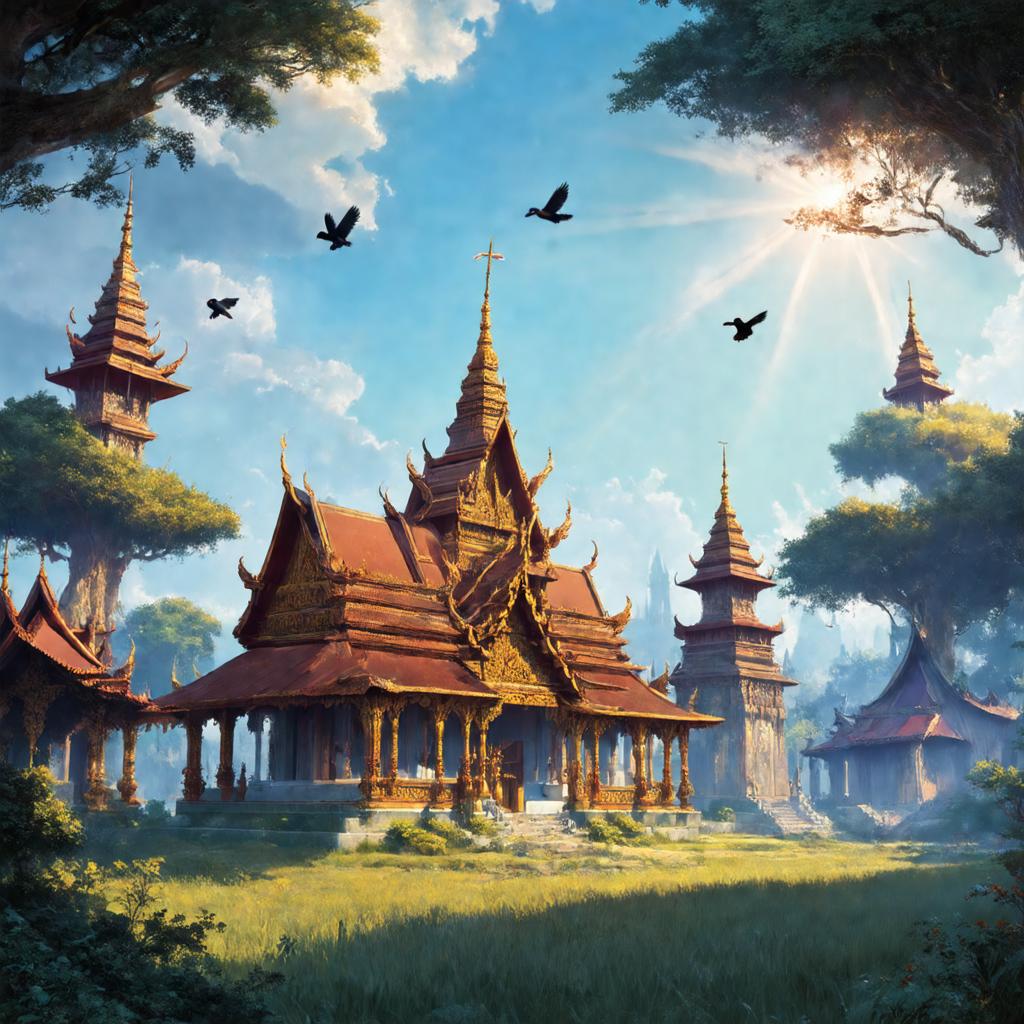}}\\[0.5ex]
        \vspace{-8pt}
        \caption*{
            \begin{minipage}{\appcapwidth}
            \centering
                \tiny{Prompt: \textit{temples of various religions stand as separate buildings in a beautiful forest on which sunlight falls and birds fly}}
            \end{minipage}
        }
    \end{minipage}
    \begin{minipage}[t]{0.5\textwidth}
        \centering
        \fbox{\includegraphics[width=\appimgwidth]{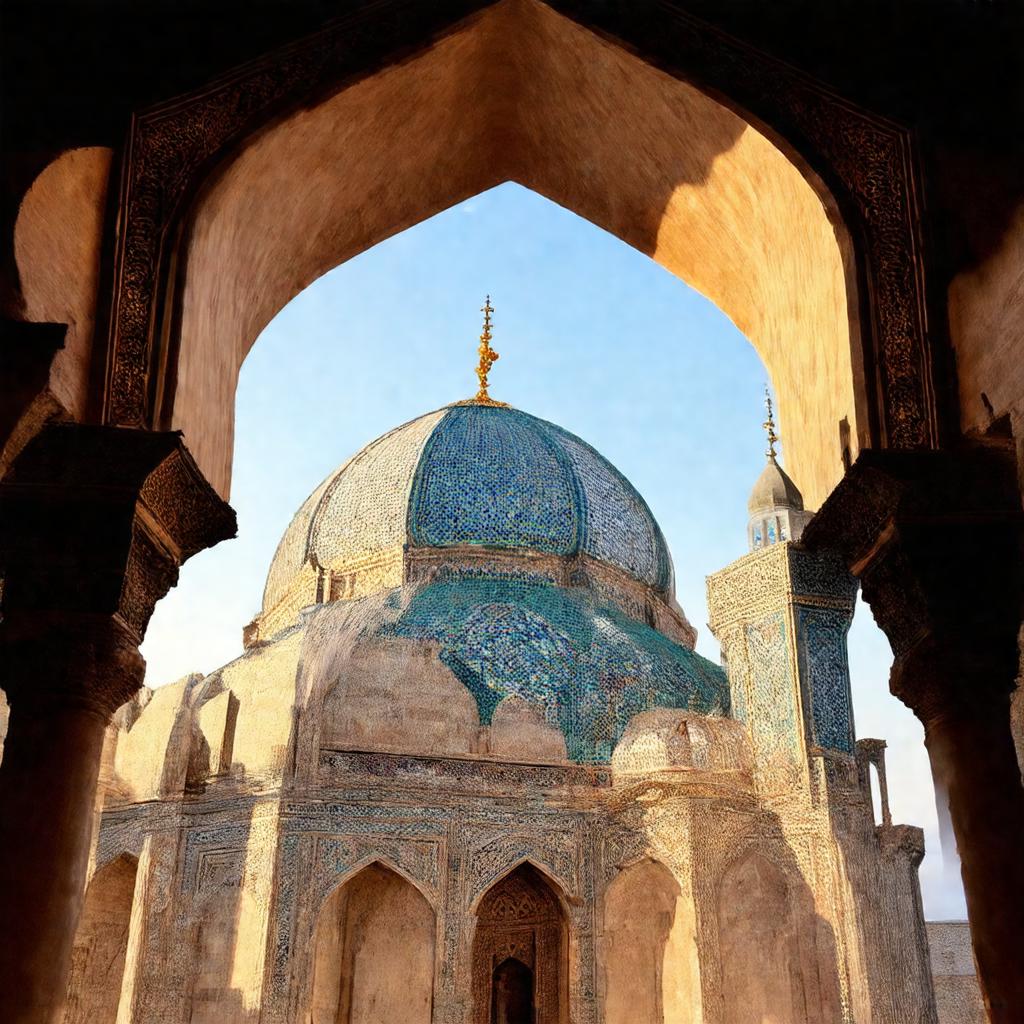}}
        \fbox{\includegraphics[width=\appimgwidth]{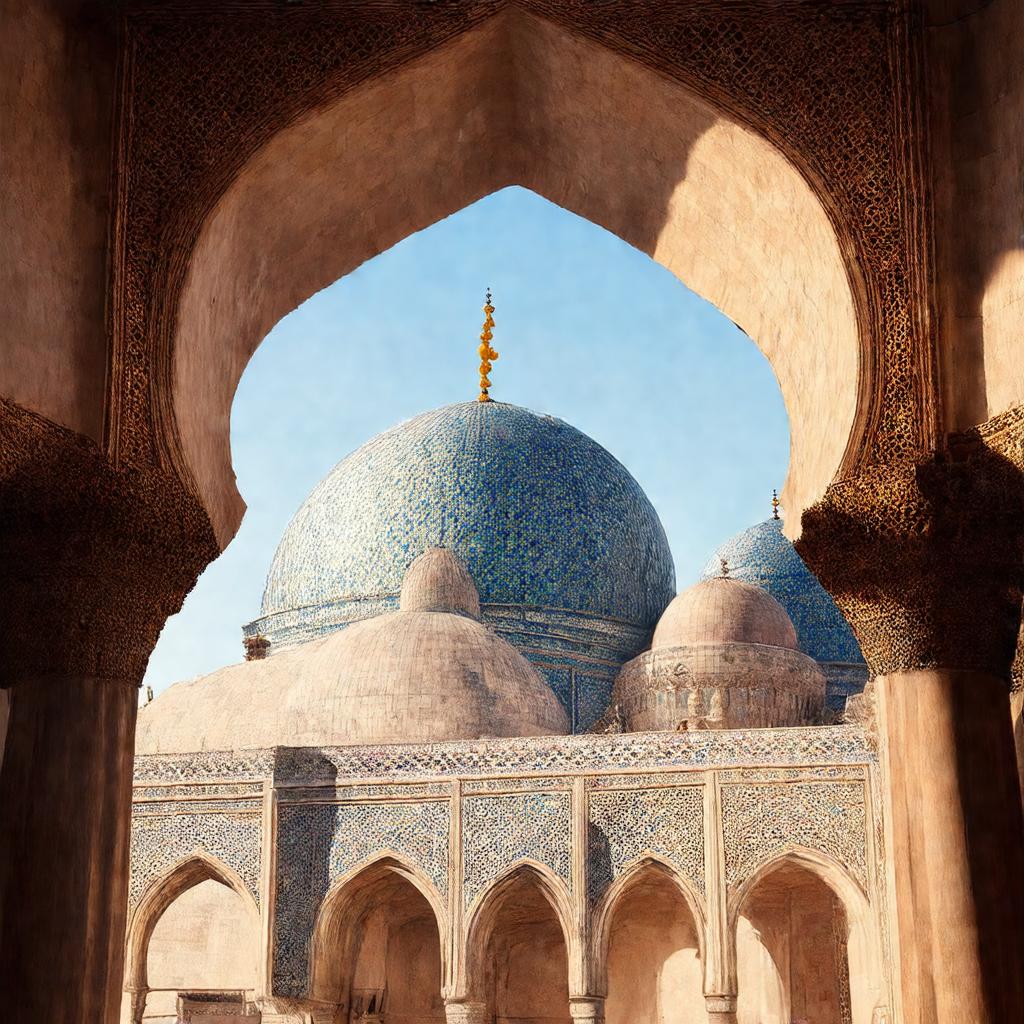}}
        \fbox{\includegraphics[width=\appimgwidth]{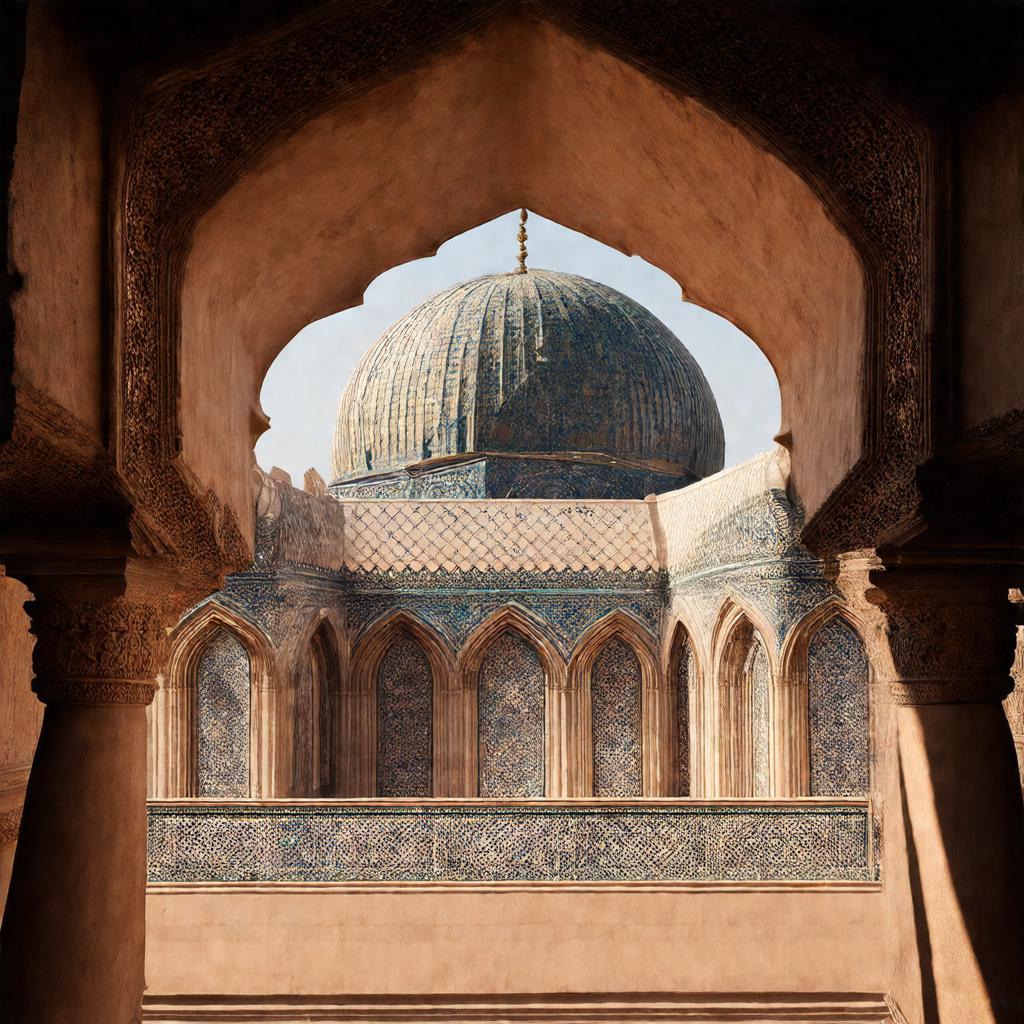}}
        \fbox{\includegraphics[width=\appimgwidth]{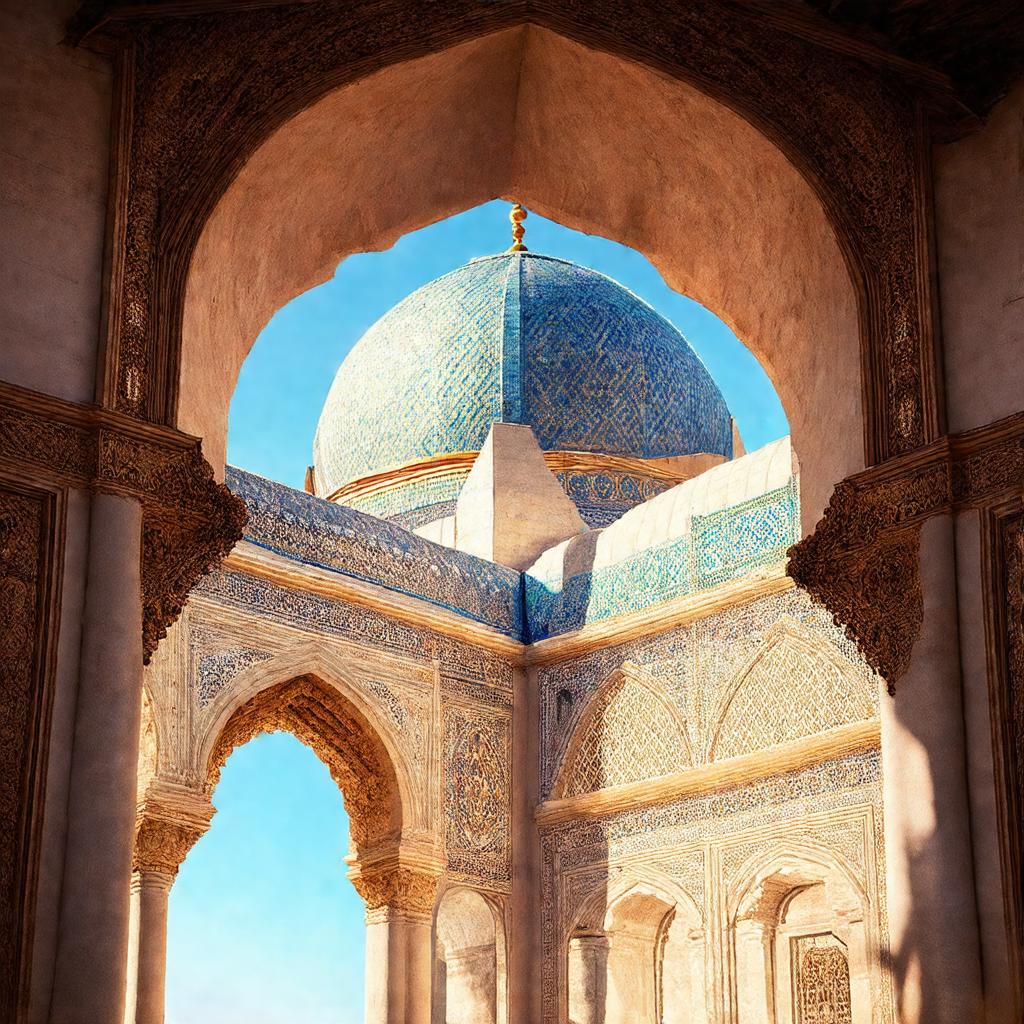}}\\[0.5ex]
        \vspace{-8pt}
        \caption*{
            \begin{minipage}{\appcapwidth}
            \centering
                \tiny{Prompt: \textit{Create a hyperrealistic portrayal of arabesque architecture, showcasing the elaborate archways and domes adorned with intricate carvings and mosaics. Ensure that the camera lens captures the depth and shadows of the design, highlighting the architectural details.}}
            \end{minipage}
        }
    \end{minipage}%
    \begin{minipage}[t]{0.5\textwidth}
        \centering
        \fbox{\includegraphics[width=\appimgwidth]{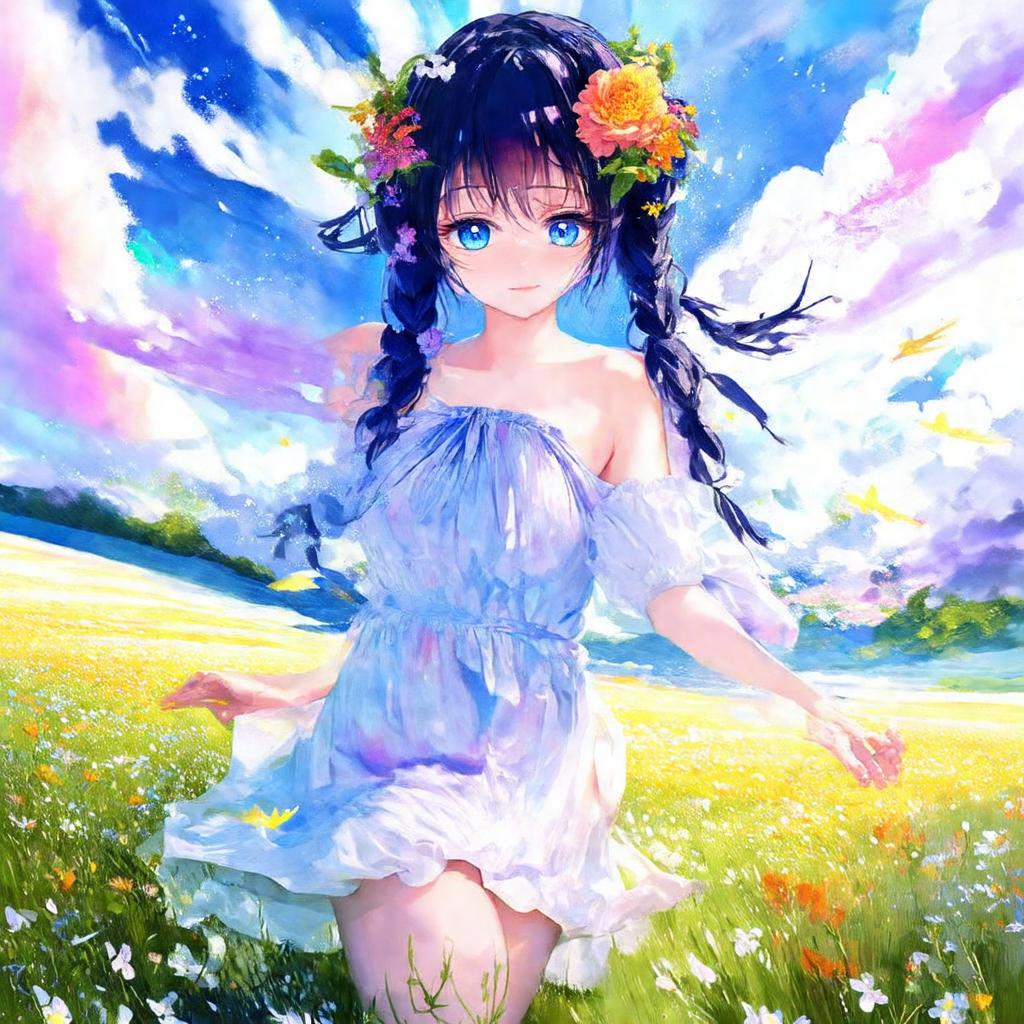}}
        \fbox{\includegraphics[width=\appimgwidth]{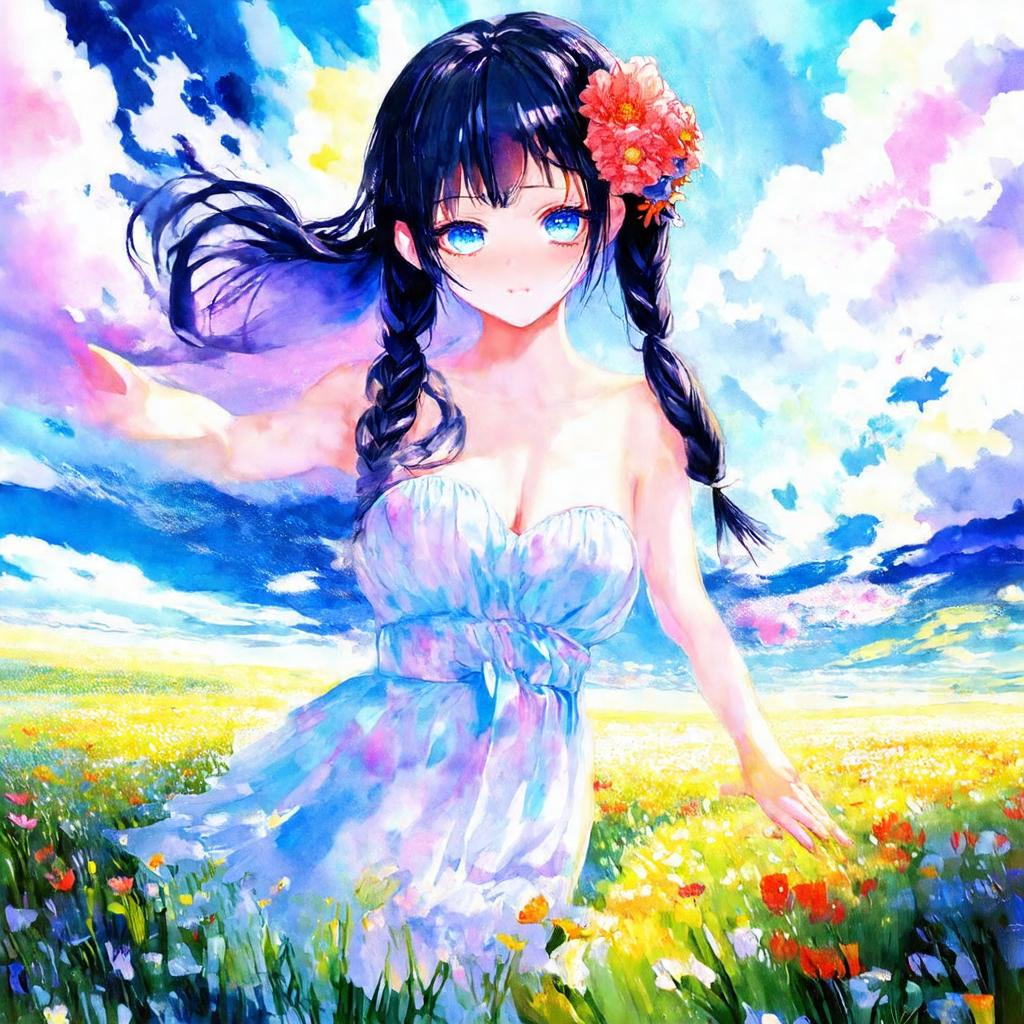}}
        \fbox{\includegraphics[width=\appimgwidth]{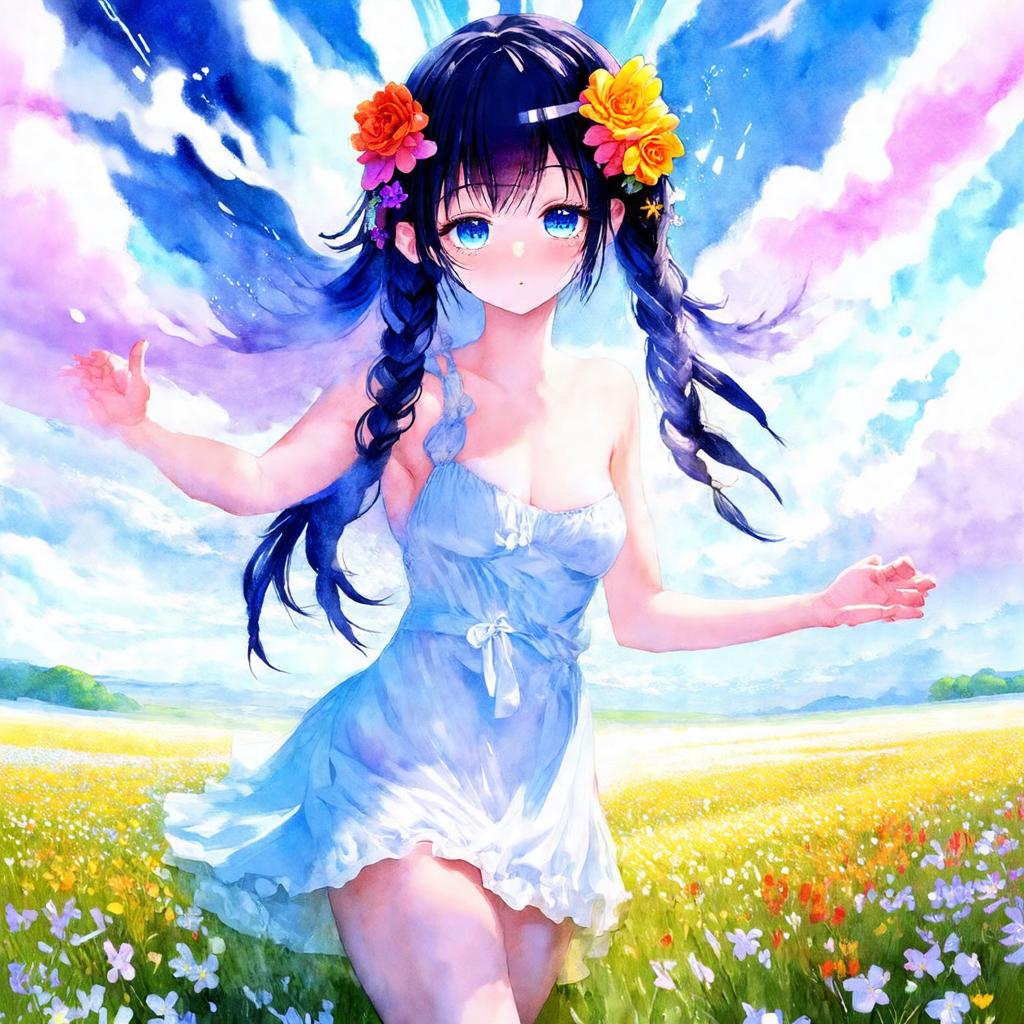}}
        \fbox{\includegraphics[width=\appimgwidth]{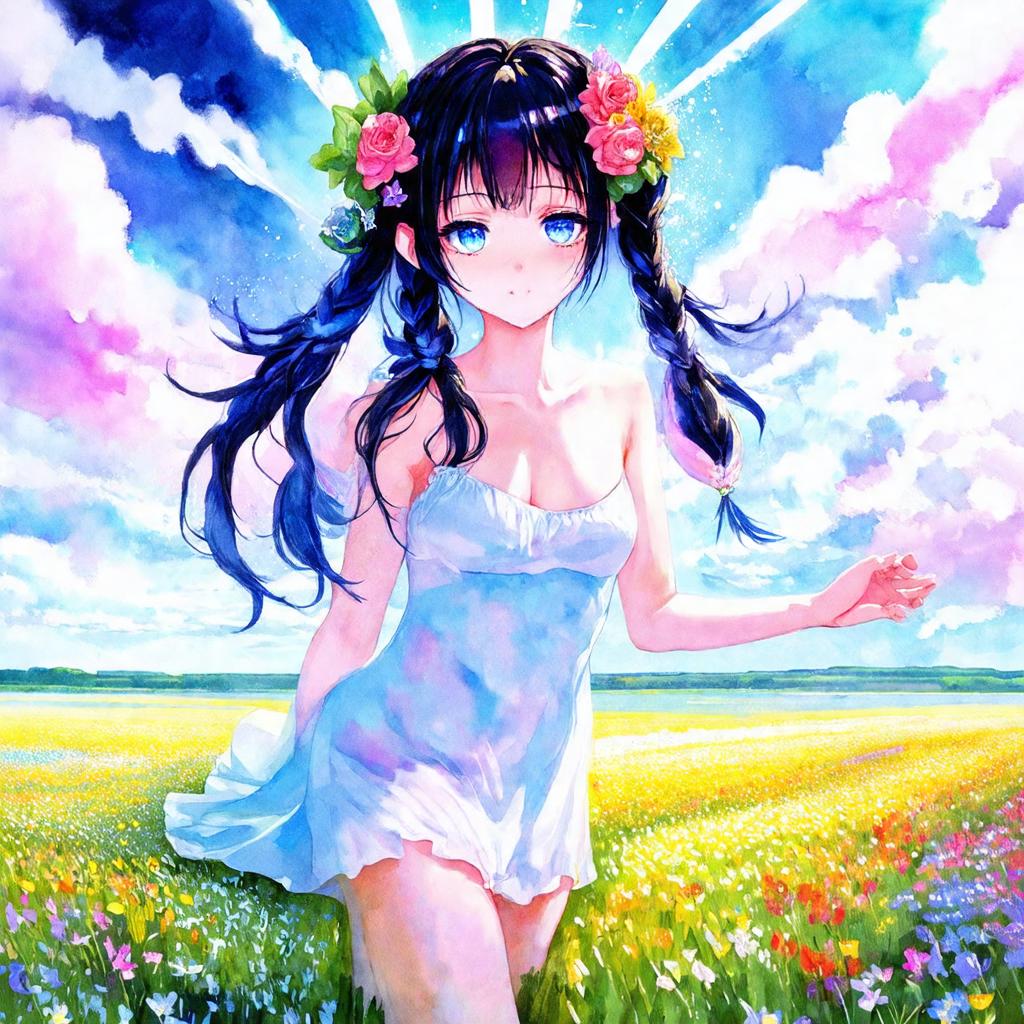}}\\[0.5ex]
        \vspace{-8pt}
        \caption*{
            \begin{minipage}{\appcapwidth}
            \centering
                \tiny{Prompt: \textit{pretty anime girl. Kawaii Face Style. flowing black hair. sparkling blue eyes. dancing in a dreamy field of flowers. rays of shimmering light. polar stratospheric clouds in the background. wearing bohemianchic clothing. braids and flowers in her hair. vibrant hues. distinctive and eyecatching fine art. splash watercolor. masterpiece}}
            \end{minipage}
        }
    \end{minipage}

    \begin{minipage}[t]{\textwidth}
        \centering
        \fbox{\includegraphics[width=\appimgwidth]{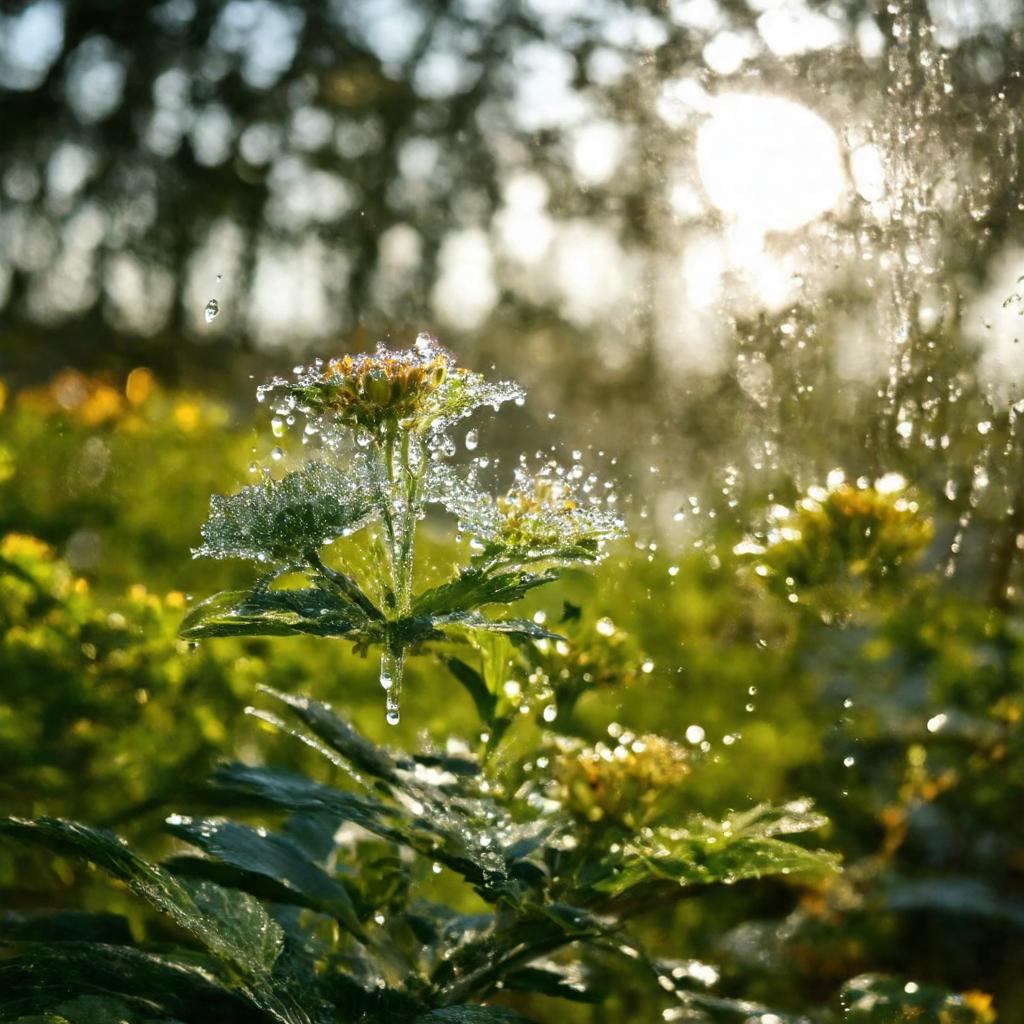}}
        \fbox{\includegraphics[width=\appimgwidth]{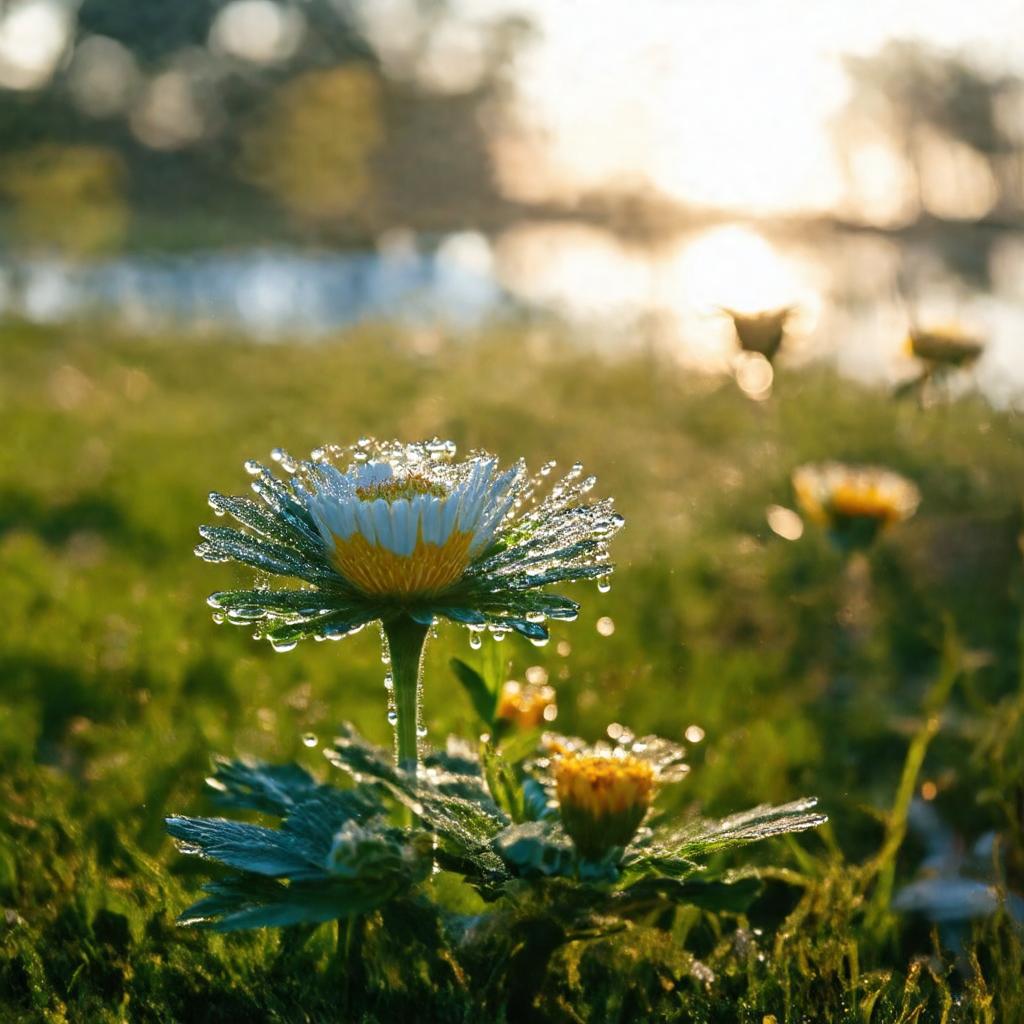}}
        \fbox{\includegraphics[width=\appimgwidth]{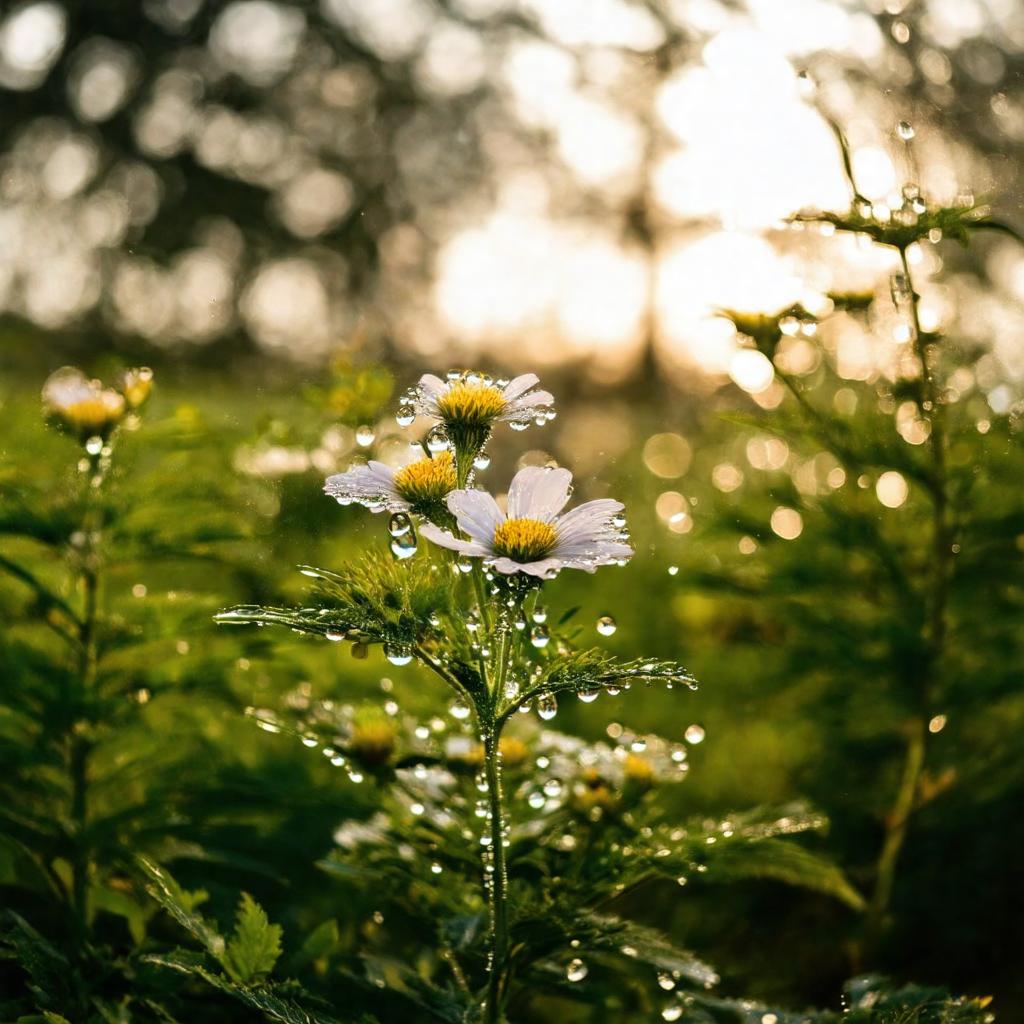}}
        \fbox{\includegraphics[width=\appimgwidth]{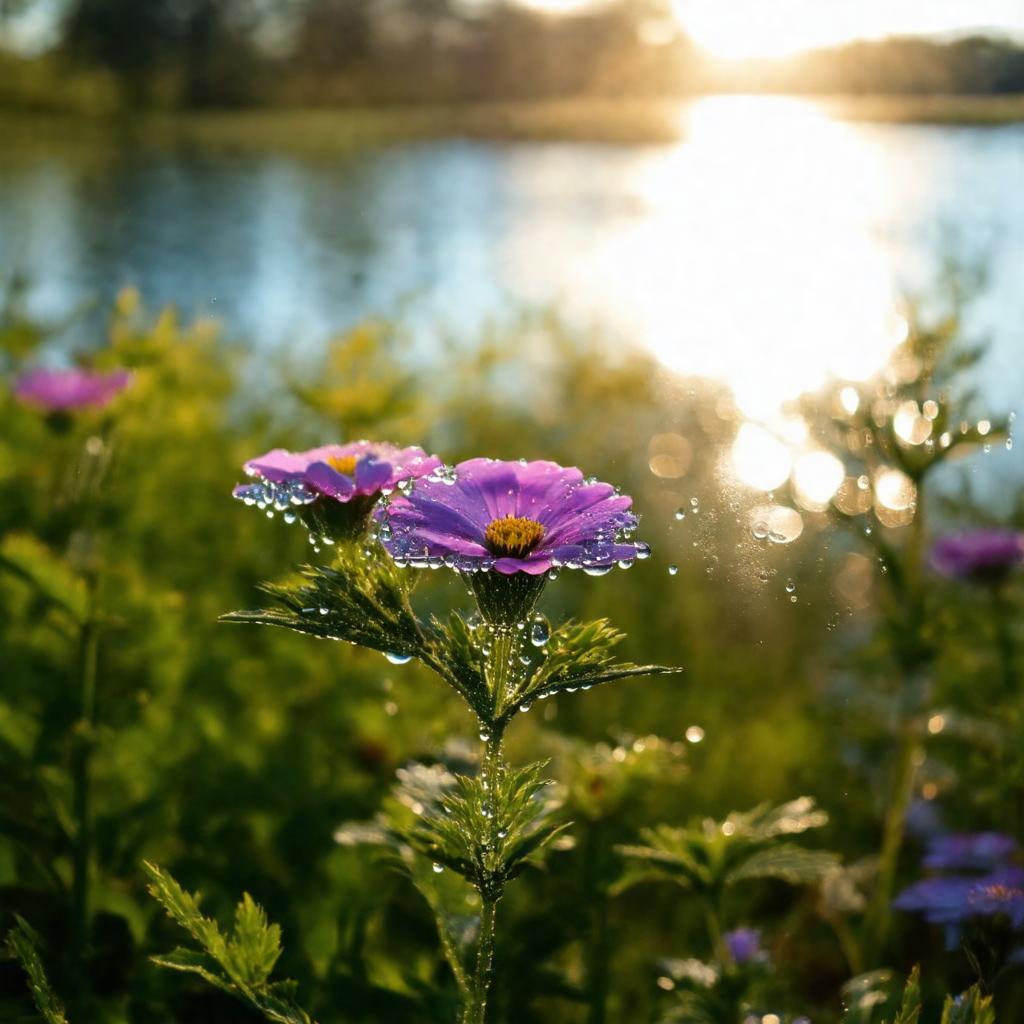}}\\[0.5ex]
        \vspace{-8pt}
        \caption*{
            \begin{minipage}{1\textwidth}
            \centering
                \tiny{Prompt: \textit{beautiful nature scene with dew dripping from flowers The photo was skillfully taken with a Nikon camera. The D850 DSLR paired with the versatile Nikkor 2470mm f2.8 lens, renowned for its sharpness and exceptional color reproduction. The f8 aperture is chosen to provide a deep depth of field and sharp detail capture of the entire scene. The ISO sensitivity is set to 200 and the shutter speed is 1500 second. photography uses bright, natural sunlight reflecting off a lake, illuminating the entire scene with harsh, cool light and highlighting the contrasting shadows that define the contours of the landscape.}}
            \end{minipage}
        }
    \end{minipage}

    \caption{
        More visualization of ablation results on the MJHQ dataset with SD3.5 and W8A8 DiT quantization. From left to right: baseline, \textsc{Seg.}, \textsc{Dual.} and \textsc{Seg.}+\textsc{Dual.}
    }
    \label{fig:app-mjhq-abl}
\end{figure}

\end{document}